\documentclass[twoside,11pt]{article}
    
    \usepackage{tikz}
    \usepackage{tcolorbox}
    \newtcolorbox{mybox}[3][]
{
  colframe = #2!25,
  colback  = #2!10,
  coltitle = #2!20!black,  
  title    = {#3},
  #1,
}

    \usepackage{jmlr2e}
    \usepackage{natbib}
    \usepackage{graphicx}
    \usepackage{tabularx}
    \usepackage{multirow}
    \usepackage{amsmath}
    \usepackage{amsfonts}

    \usepackage{amsthm}
    \usepackage{mathtools} 
    \usepackage{xfrac}
    \usepackage{wrapfig}
    \usepackage{xcolor, color, colortbl} 
    \usepackage{subfig}
    \usepackage{amssymb}
    \usepackage{pifont}
    \newcommand{\cmark}{\ding{51}}%
    \usepackage{booktabs}
    \usepackage{scalerel,stackengine}
    \stackMath

    \usepackage{algorithm}
    \usepackage{algorithmic}

    \NewDocumentCommand{\rot}{O{45} O{1em} m}{\makebox[#2][l]{\rotatebox{#1}{#3}}}%
    
    \DeclareRobustCommand{\&}{%
      \ifdim\fontdimen1\font>0pt
        \textsl{\symbol{`\&}}%
      \else
        \symbol{`\&}%
      \fi
    }

    \def\one{\mbox{1\hspace{-4.25pt}\fontsize{12}{14.4}\selectfont\textrm{1}}} 
    
    \jmlrheading{1}{2020}{1-48}{4/00}{10/00}{Antoine Grosnit, Alexander I. Cowen-Rivers, Rasul Tutunov, Ryan-Rhys Griffiths, Jun Wang, Haitham Bou-Ammar}
    
    \ShortHeadings{CompBO: A Case for Compositional Optimisation in Bayesian Optimisation}{}

    \firstpageno{1}
    
\begin{document}
    
    \title{Are we Forgetting about Compositional Optimisers in Bayesian Optimisation?}
    
    \author{\name Antoine Grosnit \thanks{Equal contribution} \email antoine.grosnit@huawei.com \\
       \name Alexander I. Cowen-Rivers \footnotemark[1] \email alexander.cowen.rivers@huawei.com \\
       \name Rasul Tutunov \footnotemark[1] \email rasul.tutunov@huawei.com \\
       \addr Huawei R\&D UK\\
       \AND
       \name Ryan-Rhys Griffiths \email rrg27@cam.ac.uk \\
       \addr Huawei R\&D UK\\
       University of Cambridge\\
       \AND
       \name Jun Wang \email w.j@huawei.com \\
       \name Haitham Bou-Ammar \thanks{Honorary position at UCL} \email haitham.ammar@huawei.com \\
       \addr Huawei R\&D UK\\
       University College London}
    
    \editor{}
    
    \maketitle
    
    \begin{abstract}
    Bayesian optimisation presents a sample-efficient methodology for global optimisation. Within this framework, a crucial performance-determining subroutine is the maximisation of the acquisition function, a task complicated by the fact that acquisition functions tend to be non-convex and thus nontrivial to optimise. In this paper, we undertake a comprehensive empirical study of approaches to maximise the acquisition function. Additionally, by deriving novel, yet mathematically equivalent, compositional forms for popular acquisition functions, we recast the maximisation task as a compositional optimisation problem, allowing us to benefit from the extensive literature in this field. We highlight the empirical advantages of the compositional approach to acquisition function maximisation across 3958 individual experiments comprising synthetic optimisation tasks as well as tasks from Bayesmark. Given the generality of the acquisition function maximisation subroutine, we posit that the adoption of compositional optimisers has the potential to yield performance improvements across all domains in which Bayesian optimisation is currently being applied.
    \end{abstract}
    
    
    \section{Introduction}
    
    Bayesian optimisation is a method for optimising black-box objective functions \citep{1964_Kushner, 1975_Mockus, 1998_Jones}. The black-box optimisation (BBO) problem describes the search for the global maximiser $\mathbf{x}^*$ of an unknown objective function $f(\mathbf{x})$. The objective function is unknown in the sense that an analytical form is unavailable. However, the objective may still be evaluated pointwise at arbitrary query locations within the bounds of the design space. A further characteristic of the BBO problem is that each query is expensive in terms of time, and as such, it is desirable to query as few points as possible in the search for the global maximiser.
    
    Real world examples of BBO problems are ubiquitous. Illustrative examples include hyperparameter tuning in machine learning \citep{2018_Falkner, 2018_Kandasamy, 2019_White, 2020_Gabillon}, where the black-box objective is the mapping between a set of model hyperparameters $\mathbf{x}$ and the validation set performance $f(\mathbf{x})$, as well as automatic chemical design \citep{2018_Gomez, 2020_Korovina, 2020_FlowMO, 2020_Griffiths}, where the black-box objective is the mapping between a molecule $\mathbf{x}$ and its suitability as a drug candidate $f(\mathbf{x})$. Further examples of BBO problems appear as subroutines of optimisation algorithms such as immune optimisation \citep{2015_Zhang, 2015_Mahapatra}, ant colony optimisation \citep{2014_Yoo, 2015_Speranskii} and genetic algorithms \citep{2015_Peng}, in reinforcement learning when accounting for safety \citep{cowen2020samba, abdullah2019wasserstein}, in multi-agent systems to compute Nash equilibria \citep{yang2019alpha, 2018_Aprem}, in speech recognition \citep{2020_boffin} and more broadly across domains spanning architecture \citep{2015_Costa}, chemical engineering \citep{2018_Ploskas} and biology \citep{2012_Shah, 2020_Moss}.
    
    \begin{wrapfigure}{l}{0.5\textwidth}
      \centering
      \includegraphics[trim = 1em 0em 1.3em 2.06em , clip = true , width=0.5\textwidth]{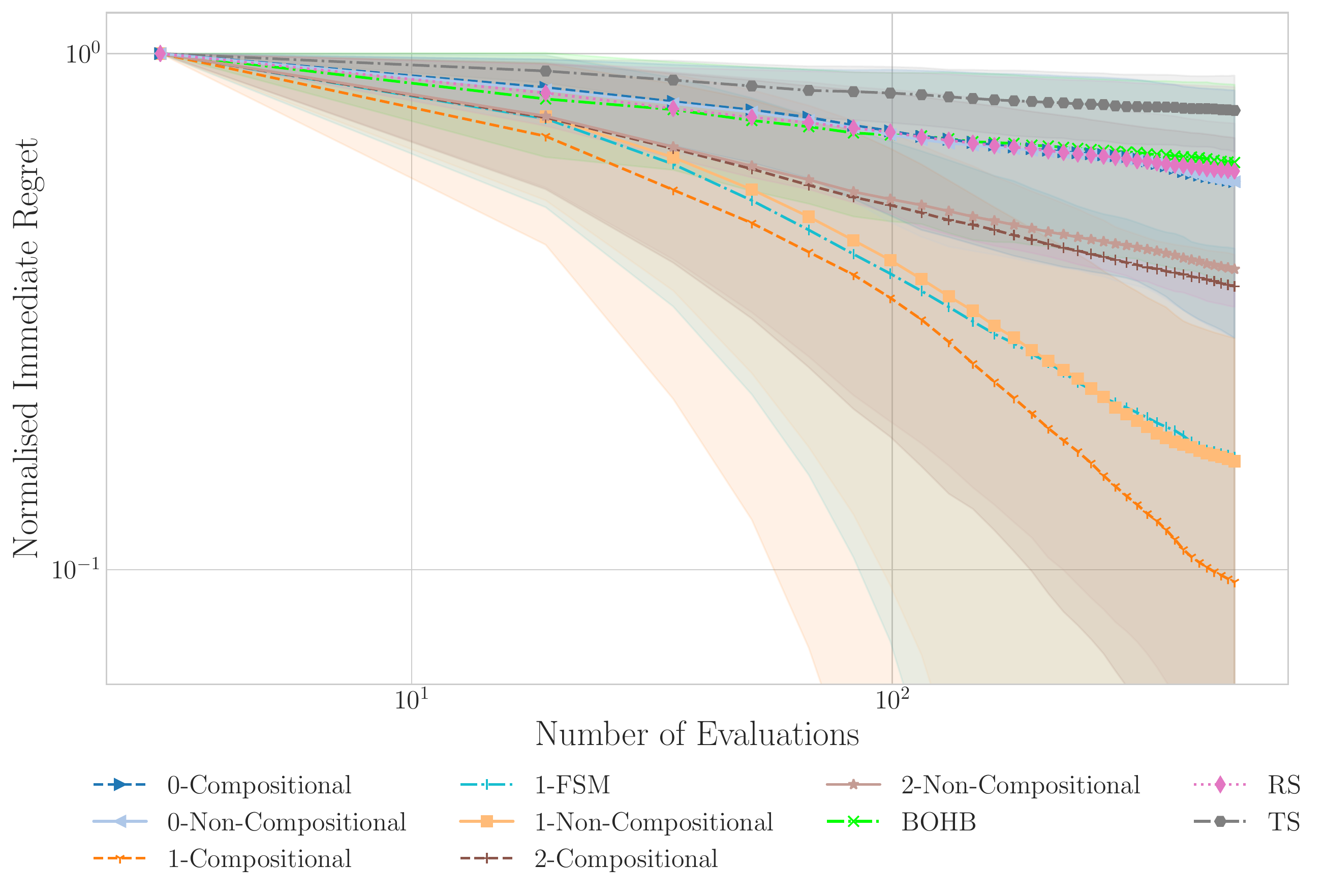}
      \caption{Summary plot for 3100 synthetic BBO experiments showing that first-order compositional optimisers outperform others. Lower regret is indicative of better performance.}
        \label{fig:summary-synthetic}
    \end{wrapfigure}
    
    Various strategies exist for optimising black-box objective functions including zero-order methods \citep{2013_Valko, 2015_Grill, 2020_Gabillon}, resource allocation methods \citep{2017_Li, 2018_Falkner} and surrogate model-based methods \citep{2012_Snoek, 2016_Shahriari, 2018_Frazier}. In this paper, we focus on Bayesian optimisation, a sequential, data-efficient, surrogate model-based approach that is particularly effective when function evaluations are costly. The two core components of the Bayesian optimisation algorithm are a probabilistic surrogate model and an acquisition function. The probabilistic surrogate model facilitates data efficiency by making use of the full optimisation history to represent the black-box function and additionally leverages uncertainty estimates to guide exploration. Given that the true sequential risk describing the optimality of a sequence of queries is computationally intractable, an acquisition function is a myopic heuristic which acts as a proxy to the true sequential risk. The acquisition function measures the utility of a query point $\mathbf{x}$ by its mean value under the surrogate model (exploitation) as well as its uncertainty under the surrogate model (exploration). At each round of the Bayesian optimisation algorithm, the acquisition function is maximised to select the next query point.
    
    \begin{wrapfigure}{l}{0.5\textwidth}
    \begin{center}
        \vspace{-20pt}
        \includegraphics[width=0.5\textwidth]{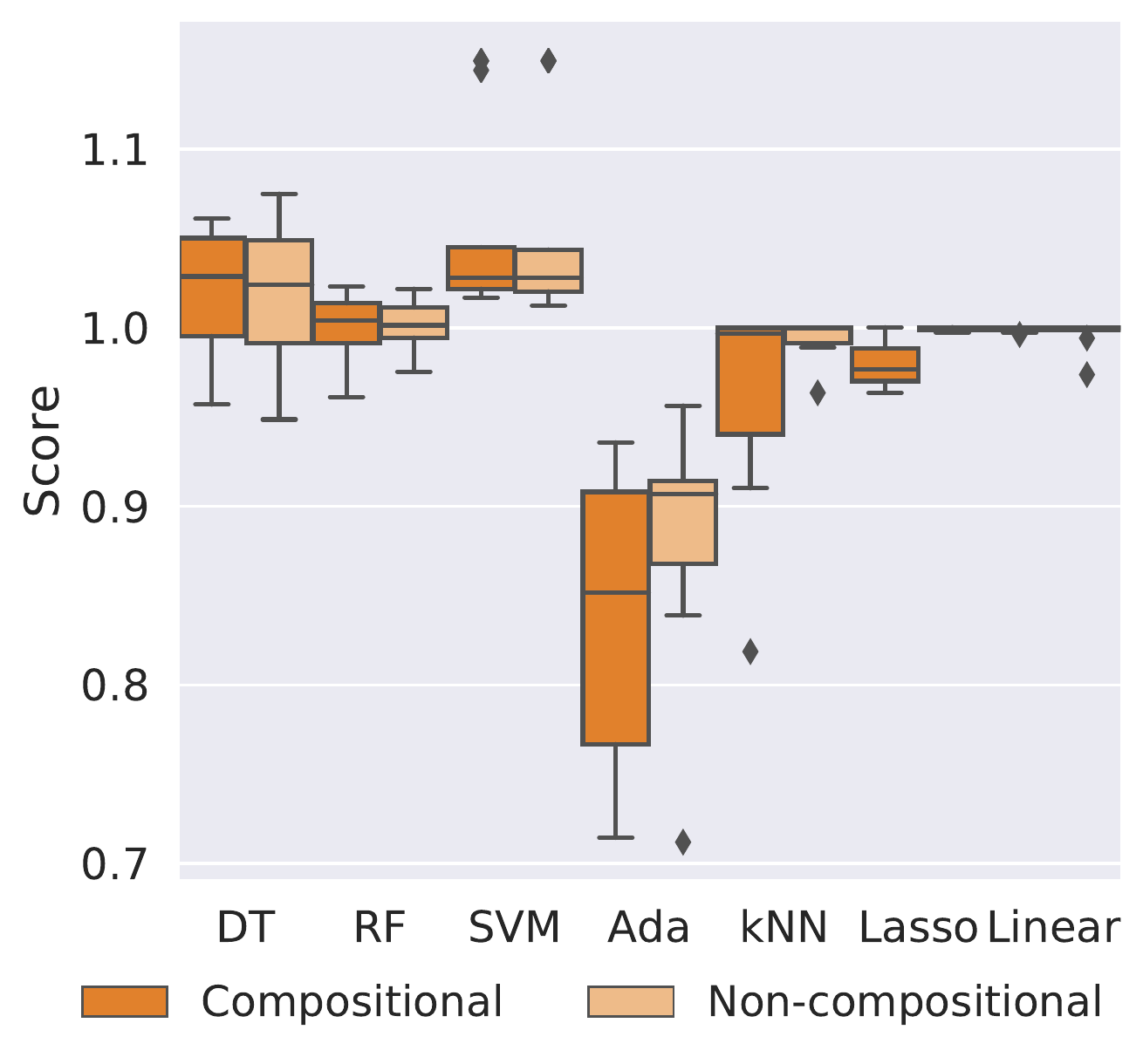}
        \caption{{Bayesmark regression summary. Comparison of compositional and non-compositional optimisers. Higher score is better.}}
        \vspace{-20pt}
        \label{fig:summary_competition_reg}
    \end{center}
    \end{wrapfigure}
    
    It has been argued that maximisation of the acquisition function is an important, yet neglected determinant of the performance of Bayesian optimisation schemes \citep{2018_Wilson}. The vast majority of acquisition functions however, constitute a serious challenge from the standpoint of optimisation; a characteristic exacerbated in the batch setting, where acquisition functions are routinely non-convex, high-dimensional and intractable \citep{2018_Wilson}. Many strategies exist for optimising acquisition functions including gradient-based methods \citep{Duchi_2011_adagrad, 2012_Hinton, ADAM}, evolutionary methods \citep{2006_Igel, 2006_Jastrebski, 2016_Hansen} as well as variations of random search \citep{1968_Schumer, 1976_Schrack, 2012_Bergstra}. In this work, we choose to focus on gradient-based methods which were recently shown to be highly effective for optimising a wide class of Monte Carlo acquisition functions \citep{2018_Wilson}.
    
    The most commonly-used acquisition functions in practical applications \citep{2012_Snoek} are Monte Carlo acquisition functions in the sense that they are formulated as integrals with respect to the current probabilistic belief over the unknown function $f$ \citep{2016_Shahriari, 2018_Wilson}; these integrals are typically intractable and as such are approximated by the corresponding Monte Carlo (MC) estimate. In order to admit gradient-based optimisation, a reparametrisation trick \citep{2014_Kingma, 2014_Rezende}, introduced first as infinitesimal perturbation analysis \citep{1985_Cao, 1988_Glasserman}, is applied to facilitate differentiation through the MC estimates with respect to the parameters of the surrogate model. It was shown in \citep{2018_Wilson} that acquisition functions estimated via MC integration are consistently amenable to gradient-based optimisation via standard first and second-order methods including SGA \citep{2007_Bottou}, Adam \citep{ADAM}, RMSprop \citep{2012_Hinton}, AdaGrad \citep{Duchi_2011_adagrad} and L-BFGS-B \citep{1997_Zhu}.
    
    In this work, we exploit the observation that most common acquisition functions exhibit compositional structure and hence can be equivalently reformulated in a compositional form  \citep{2017_Wang}. Such a reformulation allows a broader class of optimisation techniques to be applied for acquisition function optimisation \citep{tutunov2020cadam, ghadimi2020single, 2017_MWang} and in practice can more often enable better numerical performance to be achieved in comparison with standard first and second-order methods. The compositional form is achieved for the expected improvement (EI), simple regret (SR), upper confidence bound (UCB) and probability of improvement (PI) acquisition functions by first exposing the finite-sum form of the reparameterised acquisition functions derived by \citep{2018_Wilson} and second introducing a deterministic outer function when considering the problem from a matrix-vector perspective. It should be noted that reformulating the acquisition function in a compositional form is distinct from the setting where the black-box function has a compositional form \citep{2019_Astudillo}.
    
    In order to both improve and analyse the optimisation performance on the compositional form of the acquisition function, we introduce several algorithmic adaptations. Firstly, we present (C)L-BFGS; a modification to the L-BFGS algorithm to enable the handling of nested compositional forms. Secondly, we develop AdamOS, a variant of the Adam optimiser \citep{ADAM} which borrows the hyperparameter settings of CAdam \citep{tutunov2020cadam} and facilitates performance comparison between compositional and non-compositional optimisers. Lastly, we formulate a generalised iterative update rule for first-order compositional optimisers and show how the updates of a number of first-order optimisers may be expressed in this manner.
    
    

    
    \begin{wrapfigure}{r}{0.5\textwidth}
    \begin{center}
        \vspace{-30pt}
        \includegraphics[width=0.5\textwidth]{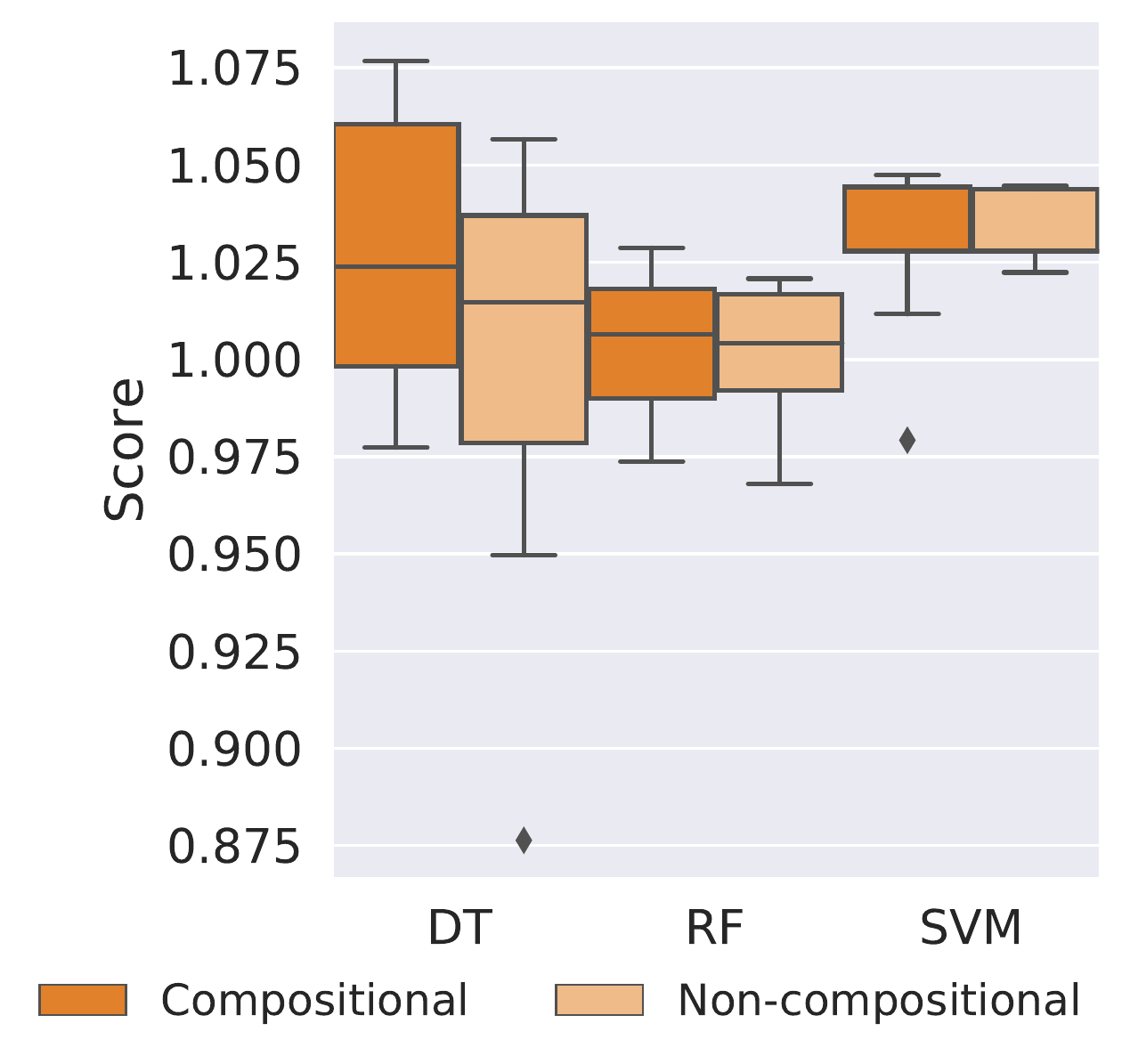}
        \caption{Bayesmark classification summary. Comparison of compositional and non-compositional optimisers. Higher score is better.}
        \vspace{-30pt}
        \label{fig:summary_competition_class}
    \end{center}
    \end{wrapfigure}
    
    In our empirical study, we seek to identify the most effective means of optimising the acquisition function under a range of experimental conditions including input dimensionality, presence or absence of observation noise and choice of acquisition function. We investigate twenty-eight optimisation schemes, spanning zeroth, first and second-order optimisers as well as both compositional and non-compositional methods. Additionally, we seek to answer the following questions: Are there benefits to the finite-sum formulation of the reparameterised acquisition functions compared to the more frequently-encountered empirical risk minimisation formulation? Are compositional or non-compositional approaches to optimisation more effective and if so, under what conditions are they more effective? What are the performance-related trade-offs in memory-efficient implementations of compositional acquisition functions? How does the wall-clock time of compositional optimisation methods compare to non-compositional optimisation methods, and how does this vary with the dimensionality of the input space? How do compositional optimisers fare when faced with noisy observations?
    
    In order to answer these questions, we first perform a set of experiments across five noiseless synthetic function tasks. Using this set of noiseless experiments as a filter for the most effective optimisers, we then perform a second set of experiments on the Bayesmark datasets which are noisy and bear a closer resemblance to real-world problems than the synthetic tasks. Our results for the synthetic experiments are summarised in \autoref{fig:summary-synthetic} whilst our results for the Bayesmark datasets are summarised in \autoref{fig:summary_competition_reg} and \autoref{fig:summary_competition_class} for the regression and classification challenges respectively. In sum total, our empirical study comprises 3958 individual experiments.
    

    The paper is organised as follows: First, we introduce the necessary background on the Bayesian optimisation framework. Second, we hone in on the acquisition function maximisation subroutine of Bayesian optimisation with the intent to understand the efficacy of compositional optimisation schemes. We provide a general overview of compositional optimisation and derive compositional forms for the four most popular myopic acquisition functions. Third, we discuss state-of-the-art compositional solvers, namely CAdam, NASA, SCGA and ASCGA. Fourth, we detail our experimental setup and present the empirical results. Fifth, we analyse the experimental results, draw conclusions and indicate avenues for future work as well as descriptions of open problems in acquisition function maximisation.

    
    \section{Bayesian Optimisation}
    We consider a sequential decision approach to the global optimisation of smooth functions $f: \mathcal{X} \rightarrow \mathbb{R}$ over a bounded input domain $\mathcal{X} \subseteq \mathbb{R}^{d}$. At each decision round, $i$, we select an input $\textbf{x}_{i} \in \mathcal{X}$ and observe the value of the \emph{black-box} function $f(\textbf{x}_{i})$. We allow the returned value to be either deterministic i.e., $y_i = f(\textbf{x}_{i})$ or stochastic with $y_i = f(\textbf{x}_{i}) + \epsilon_{i}$, where $\epsilon_{i}$ denotes a bounded-variance random variable. Our goal is to rapidly (in terms of regret) approach the maximum $\textbf{x}^{\star} = \arg\max_{\textbf{x} \in \mathcal{X}} f(\textbf{x})$. Since both $f(\cdot)$ and $\textbf{x}^{\star}$ are unknown, solvers need to trade off exploitation and exploration during the search process.  
    
    To reason about the unknown function, typical Bayesian optimisation algorithms assume smoothness and adopt Bayesian modelling as a principle to carry out inference about the properties of $f(\cdot)$ in light of the observations. Here, one introduces a prior to encode beliefs over the smoothness properties and an observation model to describe collected data, $\mathcal{D}_{i} = \left\{\textbf{x}_{l}, y_{l}\right\}_{l=1}^{{n}_{i}}$, up to the $i^{th}$ round with $n_{i}$ denoting the total acquired data so far. Using these two components in addition to Bayes rule, we can then compute a posterior  $p(f(\cdot)|\mathcal{D}_{i}
    )$ to encode all knowledge of $f(\cdot)$ allowing us to account for the location of the maximum.
    
    \subsection{Bayesian Optimisation with Gaussian Processes}\label{Sec:BO_GP}
    A Gaussian process (GP) offers a flexible and sample-efficient procedure for placing priors over unknown functions \citep{2006_Williams}. These models are fully specified by a mean function $m(\textbf{x})$ and a covariance function, or kernel, $k(\textbf{x}, \textbf{x}^{\prime})$ that encodes the smoothness assumptions on $f(\cdot)$. Given any finite collection of inputs $\textbf{x}_{1:n_{i}}$, the outputs are jointly Gaussian given by: 
    
    \begin{equation*}
        f(\textbf{x}_{1:n_{i}})|\boldsymbol{\theta} \sim \mathcal{N}\left(m(\textbf{x}_{1:n_i}), \textbf{K}_{\boldsymbol{\theta}}(\textbf{x}_{1:n_i}, \textbf{x}_{1:n_i})\right),
    \end{equation*}
    
    where $[m(\textbf{x}_{1:n_i})]_{k} = m(\textbf{x}_{k})$ denotes the mean vector, and $\textbf{K}_{\boldsymbol{\theta}}(\textbf{x}_{1:n_i}, \textbf{x}_{1:i}) \in \mathbb{R}^{n_{i} \times n_i}$ the covariance matrix with its $(k, l)^{th}$ entry computed as $[\textbf{K}_{\boldsymbol{\theta}}(\textbf{x}_{1:n_i}, \textbf{x}_{1:n_i})]_{k, l} = k_{\boldsymbol{\theta}}(\textbf{x}_{k}, \textbf{x}_{l})$. Here, $k_{\boldsymbol{\theta}}(\cdot, \cdot)$ represents a parameterised kernel with unknown hyperparameters $\boldsymbol{\theta}$ corresponding to lengthscales or signal amplitudes for example. For ease of presentation following \citep{2006_Williams}, we use a zero-mean prior in our notation here. In terms of the choice of Gaussian process kernel, there are a wide array of options which encode prior modelling assumptions about the latent function. Two of the most commonly-encountered kernels in the Bayesian optimisation literature are the squared exponential (SE) and $\text{Mat\'{e}rn}(5/2)$ kernels

    \begin{align*}
       [\textbf{K}^{\text{SE}}_{\boldsymbol{\theta}}(\textbf{x}_{1:n_i}, \textbf{x}_{1:n_i})]_{k, l} &=  k_{\boldsymbol{\theta}}^{\text{SE}}(\textbf{x}_{k}, \textbf{x}_{l}) = \exp\left(-\frac{1}{2} r^{2} \right) \\
       [\textbf{K}^{\text{Mat\'{e}rn}(5/2)}_{\boldsymbol{\theta}}(\textbf{x}_{1:n_i}, \textbf{x}_{1:n_i})]_{k, l} &=  k_{\boldsymbol{\theta}}^{\text{Mat\'{e}rn}(5/2)}(\textbf{x}_{k}, \textbf{x}_{l}) = \exp\left(-\sqrt{5} r \right)\left(1 + \sqrt{5}r + \frac{5}{3} r^{2}\right),
    \end{align*}\label{eq:matern52}
    
    \noindent where $r = \sqrt{\left(\textbf{x}_{k} - \textbf{x}_{l}\right)^{\mathsf{T}}\text{diag}\left(\boldsymbol{\theta}^{2}\right)^{-1}\left(\textbf{x}_{k} - \textbf{x}_{l}\right)}$ and $\boldsymbol{\theta} \in \mathbb{R}^{d}$ denotes the $d$-dimensional hyperparameters with $\boldsymbol{\theta}^{2}$ executed element-wise. As noted in \citep{2006_Williams}, both these kernels are suited for situations where little is known about the latent function in question. The Mat\'{e}rn kernel, however, is arguably suitable for a broader class of real-world Bayesian optimisation problems as it imposes less restrictive smoothness assumptions on $f(\cdot)$ \citep{2012_Stein}. Following initial experimentation with linear, cosine, squared exponential and various Mat\'{e}rn kernels, we chose the $\text{Mat\'{e}rn}(5/2)$ kernel to perform all experiments with. 
    
    Given the data $\mathcal{D}_{i}$, and assuming Gaussian-corrupted observations $y_{i} = f(\textbf{x}_{i}) + \epsilon_{i}$ with $\epsilon_{i} \sim \mathcal{N}(0, \sigma^{2})$, we can write the joint distribution over the data and an arbitrary evaluation input $\textbf{x}$ as: 
    
    \begin{equation*}
        \left[\begin{array}{c}
      \textbf{y}_{1:n_i}  \\
      f(\textbf{x}) 
        \end{array}
        \right] \Bigg| \ \boldsymbol{\theta} \sim \mathcal{N}\left(\textbf{0}, \left[\begin{array}{cc}
      \textbf{K}_{\boldsymbol{\theta}}^{(i)} + \sigma^{2} \textbf{I} &  \textbf{k}^{(i)}_{\boldsymbol{\theta}}(\textbf{x})  \\
      \textbf{k}^{(i), \mathsf{T}}_{\boldsymbol{\theta}}(\textbf{x}) & k_{\boldsymbol{\theta}}(\textbf{x}, \textbf{x})  
        \end{array}
        \right]\right),
    \end{equation*}
    
    \noindent where $\textbf{K}_{\boldsymbol{\theta}}^{(i)} = \textbf{K}_{\boldsymbol{\theta}}(\textbf{x}_{1:n_{i}}, \textbf{x}_{1:n_{i}})$ and $\textbf{k}_{\boldsymbol{\theta}}^{(i)}(\textbf{x}) = \textbf{k}_{\boldsymbol{\theta}}(\textbf{x}_{1:n_{i}}, \textbf{x})$. With the above joint distribution derived, we can now easily compute the predictive posterior through marginalisation \citep{2006_Williams} leading us to $
        f(\textbf{x})|\mathcal{D}_{i}, \boldsymbol{\theta} \sim \mathcal{N}\left(\boldsymbol{\mu}_{i}(\textbf{x}; \boldsymbol{\theta}), \sigma_{i}(\textbf{x}; \boldsymbol{\theta})^{2}\right)$ with: 
    \begin{align*}
    {\mu}_{i}(\textbf{x}; \boldsymbol{\theta}) &= \textbf{k}_{\boldsymbol{\theta}}^{(i)}(\textbf{x})^{\mathsf{T}} (\textbf{K}_{\boldsymbol{\theta}}^{(i)} + \sigma^{2} \textbf{I})^{-1}\textbf{y}_{1:n_{i}}\\  
    \sigma_{i}(\textbf{x};\boldsymbol{\theta})^{2} &= k_{\boldsymbol{\theta}}(\textbf{x}, \textbf{x}) - \textbf{k}_{\boldsymbol{\theta}}^{(i)}(\textbf{x})^{\mathsf{T}} (\textbf{K}_{\boldsymbol{\theta}}^{(i)} + \sigma^{2} \textbf{I})^{-1} \textbf{k}_{\boldsymbol{\theta}}^{(i)}(\textbf{x}).
    \end{align*}
    Of course, the above can be generalised to the case when a predictive posterior over $q$ arbitrary evaluation points, $\textbf{x}^{\star}_{1:q}$, needs to be computed as is the case in batched adaptations of Bayesian optimisation. In such a setting $\textbf{f}(\textbf{x}^{\star}_{1:q})|\mathcal{D}_{i}, \boldsymbol{\theta} \sim \mathcal{N}(\boldsymbol{\mu}_i(\textbf{x}_{1:q}^{\star}; \boldsymbol{\theta}), \boldsymbol{\Sigma}_i(\textbf{x}_{1:q}^{\star}; \boldsymbol{\theta}))$ with: 
    \begin{align*}
        \boldsymbol{\mu}_i(\textbf{x}_{1:q}^{\star}; \boldsymbol{\theta}) & = \textbf{K}_{\boldsymbol{\theta}}^{(i)}(\textbf{x}^{\star}_{1:q}, \textbf{x}_{1:n_i})(\textbf{K}_{\boldsymbol{\theta}}^{(i)} + \sigma^{2} \textbf{I})^{-1}\textbf{y}_{1:n_i}\\
        \boldsymbol{\Sigma}_i(\textbf{x}_{1:q}^{\star}; \boldsymbol{\theta}) & = \textbf{K}_{\boldsymbol{\theta}}^{(i)}(\textbf{x}_{1:q}^{\star}, \textbf{x}_{1:q}^{\star}) - \textbf{K}_{\boldsymbol{\theta}}^{(i)}(\textbf{x}^{\star}_{1:q}, \textbf{x}_{1:n_i})(\textbf{K}_{\boldsymbol{\theta}}^{(i)} + \sigma^{2} \textbf{I})^{-1}\textbf{K}_{\boldsymbol{\theta}}^{\mathsf{T}, (i)}(\textbf{x}^{\star}_{1:q}, \textbf{x}_{1:n_i}).
    \end{align*}
    
    \noindent The remaining ingredient needed in a GP pipeline is a process to determine the unknown hyperparameters $\boldsymbol{\theta}$ given a set of observation $\mathcal{D}_{i}$. In standard GPs \citep{2006_Williams}, $\boldsymbol{\theta}$ are fit by minimising the negative log marginal likelihood (NLML) leading us to the following optimisation problem:
    
    \begin{equation}
    \label{eq:ml-kernel-params}
    \min_{\boldsymbol{\theta}} \mathcal{J}(\boldsymbol{\theta}) = \frac{1}{2}\text{det}\left(\textbf{C}^{(i)}_{\boldsymbol{\theta}}\right) + \frac{1}{2}\textbf{y}^{\mathsf{T}}_{1:n_{i
    }}\textbf{C}^{(i), -1}_{\boldsymbol{\theta}}\textbf{y}_{1:n_{i}} + \frac{n_{i}}{2} \log 2 \pi,  \ \text{with $\textbf{C}_{\boldsymbol{\theta}}^{(i)} = \textbf{K}_{\boldsymbol{\theta}}^{(i)} + \sigma^{2} \textbf{I}$}. 
    \end{equation}
    
    The objective in Equation~\ref{eq:ml-kernel-params} represents a non-convex optimisation problem making GPs susceptible to local minima. Various off-the-shelf optimisation solvers ranging from first-order \citep{ADAM, 2007_Bottou} to second-order \citep{1997_Zhu, 1998_Amari} methods have been rigorously studied in the literature. In our experiments, we made use of a set of implementations provided in GPyTorch \citep{2018_Gardner} that relied on a scipy \citep{virtanen2020scipy} implementation of L-BFGS-B \citep{1997_Zhu} for determining $\boldsymbol{\theta}$. It is also worth noting that gradients of the loss in Equation~\ref{eq:ml-kernel-params} require inverting an $n_{i} \times n_{i}$ covariance matrix leading to an order of $\mathcal{O}(n_{i}^{3})$ complexity in each optimisation step. In large data regimes, variational GPs have proved to be a scalable methodology through the usage of $m << n_i$ inducing points \citep{2009_Titsias, 2013_Hensman}. 
    
    In Bayesian optimisation however, data is typically sparse due to the expense of evaluating even one query of the black-box function, which makes the application of sparse GPs less attractive in these scenarios. While other scalable surrogate models such as Bayesian neural networks (BNNs) and Random Forest have featured in the literature, each come with disadvantages. Many BNN-based approaches rely on approximate inference, and hence uncertainty estimates may deteriorate in quality relative to exact GPs while the Random-Forest-based SMAC algorithm is not amenable to gradient-based optimisation due to a discontinuous response surface \citep{2011_hutter, 2016_Shahriari}. As such, we restrict our focus to exact GPs and direct the reader to external sources for discussion on alternative surrogate models such as sparse GPs \citep{2016_McIntire}, BNNs \citep{2016_Springenberg, 2017_Lobato}, neural processes \citep{2018_Kim} as well as heteroscedastic GPs \citep{2017_Calandra, 2019_Griffiths}.

    \subsection{Acquisition Functions}\label{Sec:AcqFun}
    Having introduced a distribution over latent black-box functions and specified mechanisms for updating hyperparameters, we now discuss the process by which novel query points are suggested for collection in order to improve the surrogate model's best guess for the global optimiser $\textbf{x}^{\star}$. In Bayesian optimisation, proposing novel query points is performed through maximising an acquisition function $\alpha(\cdot|\mathcal{D}_{i})$ that trades off exploration and exploitation by utilising statistics from $p(f(\cdot)|\mathcal{D}_{i})$, i.e., $\textbf{x}_{i+1} = \arg\max_{\textbf{x}} \alpha(\textbf{x}|\mathcal{D}_i) $.  
    Generally, acquisition functions are taxonimised into myopic and non-myopic forms. The former class involves integrals defined in terms of beliefs over unknown outcomes from the black-box function, while the latter class constitutes more complicated nested integrals. Due to the difficulty associated in acquiring unbiased estimates of nested integrals and the lack of widespread usage, in this paper we focus on the myopic acquisition functions that we detail next. 
    
    \paragraph{Expected Improvement:} One of the most popular acquisition functions is expected improvement \citep{1975_Mockus, 1998_Jones}, which determines new query points by maximising expected gain relative to the function values observed so far. Formally, denote by $\textbf{x}_{i}^{+}$ an input point in $\mathcal{D}_{i}$ for which $f(\cdot)$ is maximised, i.e., $\textbf{x}_{i}^{+} = \arg\max_{\textbf{x} \in \textbf{x}_{1:n_{i}}}f(\textbf{x})$. Given $\textbf{x}_{i}^{+}$, we define an expected improvement acquisition to compute the expected positive gain in function value compared to the best incumbent point in $\mathcal{D}_{i}$ as: 
    \begin{equation*}
        \alpha_{\text{EI}}(\textbf{x}|\mathcal{D}_{i}) = \mathbb{E}_{f(\textbf{x})|\mathcal{D}_{i}, \boldsymbol{\theta}} \left[\max\{(f(\textbf{x})-f(\textbf{x}_{i}^{+})), 0\}\right] = \mathbb{E}_{f(\textbf{x})|\mathcal{D}_{i}, \boldsymbol{\theta}} \left[\text{ReLU}(f(\textbf{x})-f(\textbf{x}_{i}^{+}))\right], 
    \end{equation*}
    where ReLU represents a rectified linear unit with $\text{ReLU}(a) = \max\{0, a\}$. The above can be generalised to support a batch form generating $\textbf{x}_{1:q}$ query points as introduced in~\citep{ginsbourger2008multi}. Here, we first compute the multi-dimensional predictive posterior $f(\textbf{x}_{1:q})|\mathcal{D}_{i}, \boldsymbol{\theta}$ as described in Section~\ref{Sec:BO_GP} and then define the maximal gain across all $q$-batches as: 
    \begin{equation}
    \label{Eq:q_EI}
        \alpha_{\text{q-EI}}(\textbf{x}_{1:q}|\mathcal{D}_i) = \mathbb{E}_{\textbf{f}(\textbf{x}_{1:q})|\mathcal{D}_{i}, \boldsymbol{\theta}}\left[\max_{j \in 1:q}\{\text{ReLU}(\textbf{f}(\textbf{x}_{1:q}) - f(\textbf{x}_{i}^{+})\textbf{1}_{q})\}\right],
    \end{equation}
    where $\textbf{1}_{q}$ denotes a $q$-dimensional vector of ones and as such, the $\text{ReLu}(\cdot)$ is to be executed element-wise. In words, Equation~\ref{Eq:q_EI} simply computes the expected maximal improvement across all $q$-dimensional predictions compared to the best incumbent point in $\mathcal{D}_{i}$.  
    
    \paragraph{Probability of Improvement:} Another commonly-used acquisition function in Bayesian optimisation is the probability of improvement criterion which measures the probability of acquiring gains in the function value compared to $f(\textbf{x}_{i}^{+})$~\citep{1964_Kushner}. Such a probability is measured through an expected Heaviside step function as follows: 
    \begin{equation*}
        \alpha_{\text{PI}}(\textbf{x}|\mathcal{D}_{i}) = \mathbb{E}_{f(\textbf{x})|\mathcal{D}_{i}, \boldsymbol{\theta}}\left[\one\{f(\textbf{x}) - f(\textbf{x}_{i}^{+})\}\right],
    \end{equation*}
    with $\one\{f(\textbf{x}) - f(\textbf{x}_{i}^{+})\} =1$ if $f(\textbf{x}) \geq f(\textbf{x}_{i}^{+})$ and zero otherwise. Analogous to expected improvement, we can extend $\alpha_{\text{PI}}(\textbf{x}|\mathcal{D}_{i})$ to a batch form by generalising the step function to support-vectored random variables in addition to adopting maximal gain across all batches as an improvement metric: 
    \begin{equation}
    \label{Eq:p_EI}
     \alpha_{\text{q-PI}}(\textbf{x}_{1:q}|\mathcal{D}_{i}) = \mathbb{E}_{\textbf{f}(\textbf{x}_{1:q})|\mathcal{D}_{i}, \boldsymbol{\theta}}\left[\max_{j\in 1:q}\left\{\one\{\textbf{f}(\textbf{x}_{1:q}) - f(\textbf{x}_{i}^{+})\textbf{1}_{q}\}\right\}\right],
    \end{equation}
    where $\one\{\textbf{f}(\textbf{x}_{1:q}) - f(\textbf{x}_{i}^{+})\textbf{1}_{q}\}$ returns a $q$-dimensional binary vector with $[\one\{\textbf{f}(\textbf{x}_{1:q}) - f(\textbf{x}_{i}^{+})\}]_{j} = 1$ if $[\textbf{f}(\textbf{x}_{1:q})]_{j} \geq [f(\textbf{x}_{i}^{+})\textbf{1}_{q}]_{j}$ and zero otherwise for all $j \in \{1, \dots, q\}$.

    \paragraph{Simple Regret:} In simple regret, new query points are determined by maximising expected outcomes, i.e., $\alpha_{\text{SR}}(\textbf{x}|\mathcal{D}_{i}) = \mathbb{E}_{f(\textbf{x})|\mathcal{D}_{i}, \boldsymbol{\theta}}[f(\textbf{x})]$. This, in turn, can also be generalised to a batch mode by considering the maximal improvement across all $q$ batches leading to: 
    \begin{equation*}
        \alpha_{\text{q-SR}}(\textbf{x}_{1:q}|\mathcal{D}_{i}) = \mathbb{E}_{\textbf{f}(\textbf{x}_{1:q})|\mathcal{D}_{i}, \boldsymbol{\theta}}\left[\max_{j \in 1:q}\left\{\textbf{f}(\textbf{x}_{1:q})\right\}\right]. 
    \end{equation*}
    
    \paragraph{Upper Confidence Bound:} In this type of acquisition, the learner trades off the mean and variance of the predictive distribution to gather new query points for function evaluation~\citep{srinivas2009gaussian}. In the standard form, an upper-confidence bound acquisition can simply be written as: $\alpha_{\text{UCB}}(\textbf{x}|\mathcal{D}_{i}) = \mu_{i}(\textbf{x}; \boldsymbol{\theta}) + \sqrt{\beta} \sigma_{i}(\textbf{x};\boldsymbol{\theta})$ with $\beta \in \mathbb{R}$ being a free tuneable hyperparameter. Although widely used, such a form of the upper-confidence bound is not directly amendable to parallelism. To circumvent this problem, the authors in \citep{2018_Wilson} have shown an equivalent form for the expectation by exploiting reparameterisation leading to: 
    \begin{equation*}
        \alpha_{\text{UCB}}(\textbf{x}|\mathcal{D}_{i}) = \mu_{i}(\textbf{x}; \boldsymbol{\theta}) + \sqrt{\beta} \sigma_{i}(\textbf{x};\boldsymbol{\theta}) = \mathbb{E}_{f(\textbf{x})|\mathcal{D}_{i}, \boldsymbol{\theta}}\left[\mu_i(\textbf{x}; \boldsymbol{\theta}) + \sqrt{\sfrac{\beta \pi}{2}}|\gamma_{i}(\textbf{x}; \boldsymbol{\theta})|\right], 
    \end{equation*}
    with $\gamma_{i}(\textbf{x}; \boldsymbol{\theta}) = f(\textbf{x}) - \mu_{i}(\textbf{x}; \boldsymbol{\theta})$. Given such a formulation, we can now follow similar reasoning to previous generalisations of acquisition functions and consider a batched version by taking the maximum over all $q$ query points: 
    \begin{equation*}
        \alpha_{\text{q-UCB}}(\textbf{x}_{1:q}|\mathcal{D}_{i}) = \mathbb{E}_{\textbf{f}(\textbf{x}_{1:q})|\mathcal{D}_{i}, \boldsymbol{\theta}}\left[\max_{j\in1:q}\left\{\boldsymbol{\mu}_i(\textbf{x}_{1:q}; \boldsymbol{\theta}) + \sqrt{\sfrac{\beta \pi}{2}}|\boldsymbol{\gamma}_{i}(\textbf{x}_{1:q}; \boldsymbol{\theta})|\right\}\right],
    \end{equation*}
    where $\boldsymbol{\gamma}_{i}(\textbf{x}_{1:q}; \boldsymbol{\theta}) = \textbf{f}(\textbf{x}_{1:q}) - \boldsymbol{\mu}_{i}(\textbf{x}_{1:q}; \boldsymbol{\theta})$. \\
    
    Following the introduction of GP surrogate models and acquisition functions, we are now ready to present a canonical template for the Bayesian optimisation algorithm. The main steps are summarised in the pseudocode of Algorithm~\ref{Algo:BO}.

    \begin{algorithm}
    \caption{Batched Bayesian Optimisation with GPs}
    \label{Algo:BO}
    \begin{algorithmic}[1]
    \STATE \textbf{Inputs:} Total number of outer iterations $N$, initial randomly-initialised dataset $\mathcal{D}_{0} = \{\textbf{x}_{l}, y_{l}\equiv f(\textbf{x}_{l})\}_{l=1}^{n_{0}}$, batch size $q$, acquisition function type
    \STATE \textbf{for} $i= 0 : N-1$: 
    \STATE \hspace{1em} Fit the GP model to the current dataset $\mathcal{D}_{i}$ by $\min_{\boldsymbol{\theta}} \mathcal{J}(\boldsymbol{\theta})$ from Equation~\ref{eq:ml-kernel-params}
    \STATE \hspace{1em} Find $q$ points by solving $\textbf{x}^{(\text{new})}_{1:q} =  \arg\max_{\textbf{x}_{1:q}}\alpha_{\text{q-type}}(\textbf{x}_{1:q}|\mathcal{D}_{i})$ \label{eq:acq_func_maximisation} (Round if categorical)
    \STATE \hspace{1em} Evaluate new inputs by querying the black-box to acquire $\textbf{y}^{(\text{new})}_{1:q} = f(\textbf{x}^{(\text{new})}_{1:q})$
    \STATE \hspace{1em} Update the dataset creating $\mathcal{D}_{i+1} = \mathcal{D}_{i} \cup \{\textbf{x}^{(\text{new})}_{l}, y^{(\text{new})}_{l}\}_{l=1}^{q}$
    \STATE \textbf{end for}
    \STATE \textbf{Output:} Return the best-performing query point from the data $\textbf{x}^{\star} = \arg\max_{\textbf{x} \in \mathcal{D}_{N}} f(\textbf{x})$
    \end{algorithmic}
    \end{algorithm}
    First, a GP model is fit to the available data (line 3) enabling the computation of the predictive distribution needed to maximise the acquisition function (see line 4 of Algorithm~\ref{Algo:BO}). Having acquired new query points, the learner then updates the dataset $\mathcal{D}_{i}$ after which the above process repeats until a total number of iterations $N$ is reached. At the end of the main loop, Algorithm~\ref{Algo:BO} outputs $\textbf{x}^{\star}$, the best performing input from all acquired data $\mathcal{D}_{N}$. 
    
    Clearly, maximising acquisition functions plays a crucial role in Bayesian optimisation as this step constitutes the process by which the learner yields concrete exploratory actions to improve the guess for the global optimum $\textbf{x}^{\star}$. The majority of acquisition functions, however, are often intractable, posing formidable challenges during the optimisation step in line 4 of Algorithm~\ref{Algo:BO}. In order to tackle these challenges, researchers have proposed a plethora of methods that can generally be categorised into three main groups. Approximation techniques, the first group, replace the quantity of interest with a more readily-computable one e.g. \citep{2011_Cunningham} apply expectation propagation \citep{2001_Minka, 2001_Minka_phd, 2001_Opper} as an approximate integration method while \citep{2017_Jegelka} apply a mean field approximation to enable a Gumbel sampling approximation to their max-value entropy search acquisition function. As noted in \citep{2018_Wilson}, these methods tend to work well in practice but may not converge to the true value of the optimiser. On the other hand, solutions provided in the second group \citep{2013_Chevalier} derive near-analytic expressions in the sense that they contain terms such as low-dimensional multivariate normal cumulative density functions that cannot be computed exactly but for which high-quality estimators exist \citep{1992_Genz, 2004_Genz}. As noted again by \citep{2018_Wilson}, these methods rarely scale to high dimensions. Finally, the third group comprises Monte Carlo (MC) methods \citep{2009_Osborne, 2012_Hennig, 2012_Snoek} which provide unbiased estimators to $\alpha(\cdot|\mathcal{D}_{i})$. MC methods have been successfully used in the context of acquisition function maximisation to the extent that they form the backbone of modern Bayesian optimisation libraries such as BoTorch \citep{balandat2019botorch}. 
    
    As such, given their prevalence in present-day implementations, we restrict our attention to MC techniques and note three classes of widely-used optimisers. Zeroth-order procedures~\citep{Elad, 2020_Gabillon}, such as evolutionary algorithms~\citep{CMAES, pymoo}, only use function value information for determining the maximum of the acquisition. First-order methods \citep{ADAM, 2007_Bottou}, on the other hand, utilise gradient information during the ascent step, while second-order methods exploit (approximations to) Hessians~\citep{byrd1995limited, 1997_Zhu, boyd_vandenberghe_2004,tutunov2015distributed, tutunov2019distributed} in their update. During the implementation of first and second-order optimisers, one realises the need for differentiating through an MC estimator with respect to the parameters of the generative distribution $\mathcal{P}(\cdot)$. As described in \citep{2018_Wilson}, this can be achieved through reparameterisation in two steps: 1) reparameterising samples from $\mathcal{P}(\cdot)$ as draws from a simpler distribution $\hat{\mathcal{P}}(\cdot)$, and 2) interchanging integration and differentiation by exploiting sample-path derivatives. After reparameterisation, the designer faces two implementation choices which we refer to as ERM-BO and FSM-BO akin to the distinction between empirical risk minimisation~\citep{ERM} and finite sum~\citep{FSM} optimisation forms\footnote{Of course, an empirical risk and a finite sum formulation become equivalent as samples grow large. In reality, infinite samples cannot possibly be acquired hence our two-class categorisation.}. 
    
    In an ERM-BO construction, samples from $\hat{\mathcal{P}}(\cdot)$ are acquired at every iteration of the optimisation algorithm as needed. In contrast, in an FSM-BO setting, all samples from $\hat{\mathcal{P}}(\cdot)$ are obtained upfront and mini-batched during gradient computations. Due to memory consideration, especially in high-dimensional scenarios, the ERM-BO version has been mostly preferred and studied in the literature~\citep{knudde2017gpflowopt, balandat2019botorch}. 
    
    In this paper however, we are interested in both views and desire to shed light on best practices when optimising acquisition functions. To accomplish such a goal, we carefully probe both settings and realise that an FSM-BO implementation enables a novel connection to a compositional (nested expectation) formulation that sanctions new compositional solvers not previously attempted. Next, we derive such a connection, present memory-efficient optimisation algorithms for FSM-BO, and demonstrate empirical gains in large-scale experiments. For ease of exposition, we summarise the main derivations of the coming section in Figure~\ref{Fig:Summary} to demonstrate the three steps of reparameterisation, Monte-Carlo estimation for finite-sum forms, and matrix-vector considerations for compositional objectives. 
    
    \section{Acquisition Function Maximisation} \label{Sec:AcqOp}
    \begingroup
    \setlength{\intextsep}{-17.5pt}%
    \begin{wrapfigure}{r}{0.5\textwidth}
      \begin{center}
        \includegraphics[trim = 52em 26em 52em 17em, clip=true, width=0.48\textwidth]{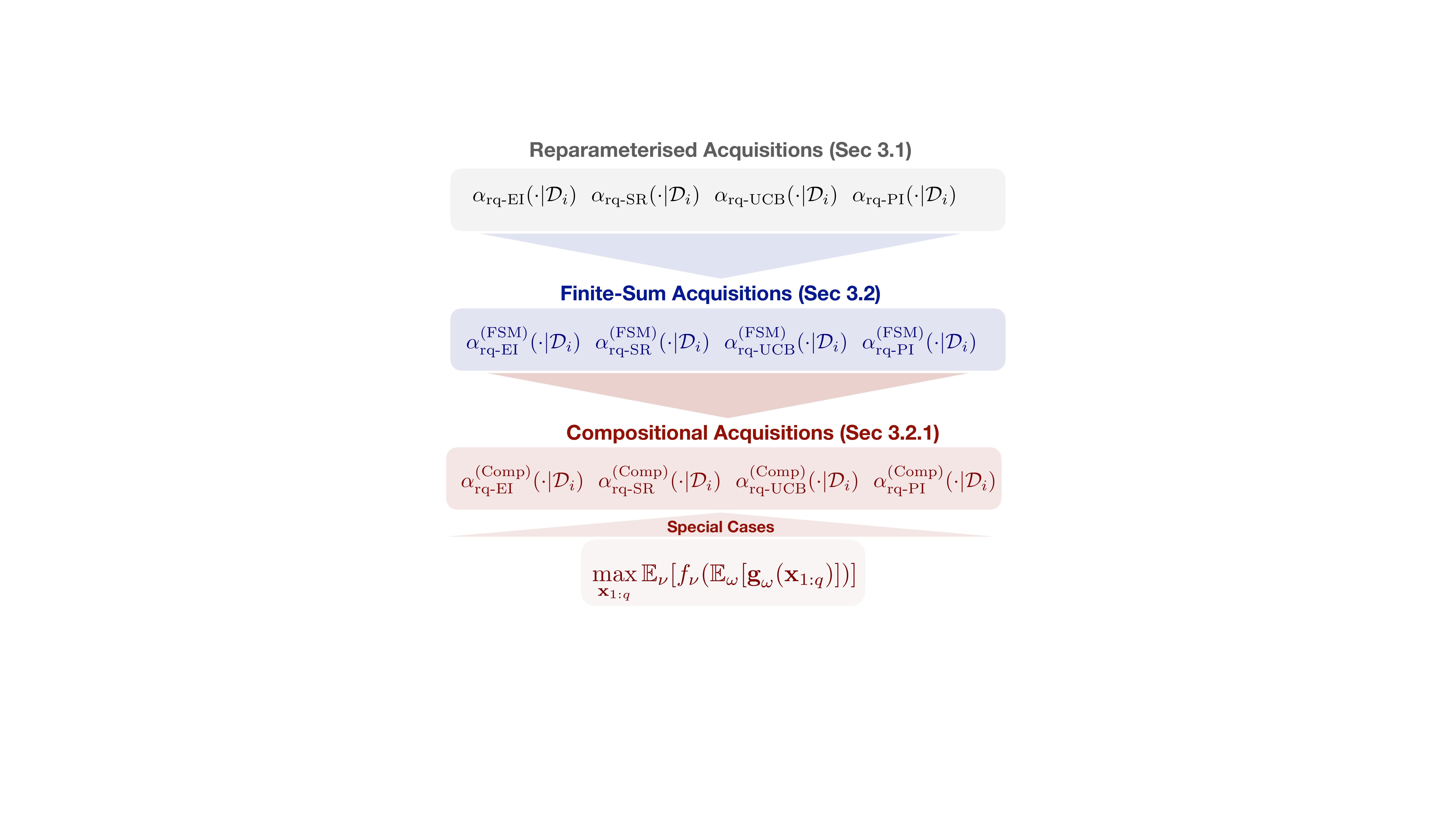}
      \end{center}
      \vspace{-1em}
      \caption{BO acquisitions in this paper.}
      \label{Fig:Summary}
    \end{wrapfigure}
    
    The first step in investigating different implementations of BO is to derive relevant reparameterised forms of the acquisition functions in Section~\ref{Sec:AcqFun}. When reparameterising one reinterprets samples $\textbf{y}_{k} \sim \mathcal{P}(\textbf{y}; \boldsymbol{\theta})$ as a deterministic map $\boldsymbol{\lambda}_{\boldsymbol{\theta}}(\cdot)$ of a simpler random variable $\textbf{z}_{k} \sim \hat{\mathcal{P}}(\textbf{z})$, that is $\textbf{y} = \lambda_{\boldsymbol{\theta}}(\textbf{z})$. Under these conditions, the expectation of some loss $\mathcal{L}(\cdot)$ under $\textbf{y}$ can be rewritten in terms of $\hat{\mathcal{P}}(\textbf{z})$ as $\mathbb{E}_{\textbf{y} \sim \mathcal{P}(y; \boldsymbol{\theta})}[\mathcal{L}(\textbf{y})] = \mathbb{E}_{\textbf{z} \sim \hat{\mathcal{P}}(\textbf{z})}[\mathcal{L}(\lambda_{\boldsymbol{\theta}}(\textbf{z}))]$ allowing us, under further technical conditions~\citep{2018_Wilson}, to push gradients inside expectations when needed. 
    
    Before diving into ascent direction computation, we first present reparameterised acquisition formulations as derived in \citep{2018_Wilson}. First, we realise that all batched acquisition functions in Section~\ref{Sec:AcqFun} involve an expectation over the GP's predictive posterior $\textbf{f}(\textbf{x}_{1:q})|\mathcal{D}_{i}, \boldsymbol{\theta} \sim \mathcal{N}(\boldsymbol{\mu}_{i}(\textbf{x}_{1:q};\boldsymbol{\theta}), \boldsymbol{\Sigma}_{i}(\textbf{x}_{1:q};\boldsymbol{\theta}))$. Second, we recall that if a random variable is Gaussian distributed, one can reparameterise by choosing $\textbf{z} \sim \mathcal{N}(\textbf{0}, \textbf{I})$ and then applying $\lambda_{\boldsymbol{\theta}}(\textbf{z}) = \boldsymbol{\mu}_{i}(\textbf{x}_{1:q}; \boldsymbol{\theta}) + \textbf{L}_{i}(\textbf{x}_{1:q}; \boldsymbol{\theta})\textbf{z}$ with $\textbf{L}_{i}(\textbf{x}_{1:q}; \boldsymbol{\theta})\textbf{L}^{\mathsf{T}}_{i}(\textbf{x}_{1:q}; \boldsymbol{\theta}) = \boldsymbol{\Sigma}_{i}(\textbf{x}_{1:q};\boldsymbol{\theta})$. Using such a deterministic transformation $\boldsymbol{\lambda}_{\boldsymbol{\theta}}(\textbf{z})$, the original random variable's distribution remains unchanged indicating a mean $\boldsymbol{\mu}_{i}(\textbf{x}_{1:q}; \boldsymbol{\theta})$ and covariance $\boldsymbol{\Sigma}_{i}(\textbf{x}_{1:q};\boldsymbol{\theta})$. Now, we can easily replace $\lambda_{\boldsymbol{\theta}}(\textbf{z})$ in each of the expected improvement, simple regret, and upper confidence bound acquisitions leading us to the following batch-reparameterised formulations: 
    \begin{align}
        \alpha_{\text{rq-EI}}(\textbf{x}_{1:q}|\mathcal{D}_{i}) & = \mathbb{E}_{\textbf{z} \sim \mathcal{N}(\textbf{0}, \textbf{I})}\left[\max_{j\in1:q}\left\{\text{ReLU}\left(\boldsymbol{\mu}_{i}(\textbf{x}_{1:q}; \boldsymbol{\theta}) + \textbf{L}_{i}(\textbf{x}_{1:q}; \boldsymbol{\theta})\textbf{z} - f(\textbf{x}_{i}^{+})\textbf{1}_{q}\right)\right\}\right], \label{eq:reparam-EI}\\
        \alpha_{\text{rq-SR}}(\textbf{x}_{1:q}|\mathcal{D}_{i}) & =  \mathbb{E}_{\textbf{z} \sim \mathcal{N}(\textbf{0}, \textbf{I})}\left[\max_{j\in1:q}\left\{\boldsymbol{\mu}_{i}(\textbf{x}_{1:q}; \boldsymbol{\theta}) + \textbf{L}_{i}(\textbf{x}_{1:q}; \boldsymbol{\theta})\textbf{z}\right\}\right],\label{eq:reparam-SR} \\
        \alpha_{\text{rq-UCB}}(\textbf{x}_{1:q}|\mathcal{D}_{i}) & =  \mathbb{E}_{\textbf{z} \sim \mathcal{N}(\textbf{0}, \textbf{I})}\left[\max_{j\in1:q}\left\{\boldsymbol{\mu}_i(\textbf{x}_{1:q}; \boldsymbol{\theta}) + \sqrt{\sfrac{\beta \pi}{2}}|\textbf{L}_{i}(\textbf{x}_{1:q}; \boldsymbol{\theta})\textbf{z}|\right\}\right].\label{eq:reparam-UCB}
    \end{align}
    When it comes to probability of improvement, the direct insertion of $\lambda_{\boldsymbol{\theta}}(\textbf{z})$ into Equation~\ref{Eq:p_EI} is difficult due to the discrete nature of the utility measure that violates differentiablity assumptions in reparameterisation~\citep{jang2016categorical}. To overcome this issue, we follow~\citep{2018_Wilson} and adopt the concrete (continuous to discrete) approximation to replace the discontinuous mapping~\citep{maddison2016concrete} such that transformed and original variables are close in distribution. Sticking to the formulation presented~\citep{2018_Wilson}, we loosen the indicator part of $\alpha_{\text{q-PI}}(\cdot)$ from Equation~\ref{Eq:p_EI} and write: 
    \begin{equation*}
       \max_{j\in 1:q}\left\{\one\{\textbf{f}(\textbf{x}_{1:q}) - f(\textbf{x}_{i}^{+})\textbf{1}_{q}\}\right\} \approx \max_{j\in 1:q}\left\{\text{Sig}\left(\frac{\textbf{f}(\textbf{x}_{1:q}) - f(\textbf{x}_{i}^{+})\textbf{1}_{q}}{\tau}\right)\right\},
    \end{equation*}
    where $\text{Sig}(\cdot)$ is executed component-wise and denotes the sigmoid function with $\tau \in \mathbb{R}_{+}$ representing its temperature parameter that yields an exact approximation as $\tau \rightarrow 0$. Given the approximation above and using a multivariate standard normal (instead of a uniform, see~\citep{maddison2016concrete}) as $\hat{\mathcal{P}}(\textbf{z})$, we derive the following reparameterised form for the probability of improvement acquisition:
    \begin{equation}\label{eq:qPI-smooth}
        \alpha_{\text{rq-PI}}(\textbf{x}_{1:q}|\mathcal{D}_{i}) =  \mathbb{E}_{\textbf{z} \sim \mathcal{N}(\textbf{0}, \textbf{I})}\left[\max_{j\in1:q}\left\{\text{Sig}\left(\frac{\boldsymbol{\mu}_{i}(\textbf{x}_{1:q}; \boldsymbol{\theta}) + \textbf{L}_{i}(\textbf{x}_{1:q}; \boldsymbol{\theta})\textbf{z} - f(\textbf{x}_{i}^{+})\textbf{1}_{q}}{\tau}\right)\right\}\right].
    \end{equation}
    
    Given reparameterised acquisitions, we now turn our attention to ERM- and FSM-BO depicting both implementations and presenting novel compositional procedures that are sample and memory efficient. 
    
    \subsection{ERM-BO using Stochastic Optimisation}\label{Sec:ERM_BO}
    Mainstream implementations of BO cast the inner optimisation problem (line 4 in Algorithm~\ref{Algo:BO}) in an empirical risk form $\max_{\textbf{x}_{1:q}} \mathbb{E}_{\textbf{z}\sim\mathcal{N}(\textbf{0}, \textbf{I})}[\mathcal{L}(\textbf{x}_{1:q}; \textbf{z})]$ with $\mathcal{L}(\textbf{x}_{1:q}; \textbf{z})$ dependent on the acquisition's type, e.g., $\max_{j\in1:q}\left\{\boldsymbol{\mu}_{i}(\textbf{x}_{1:q}; \boldsymbol{\theta}) + \textbf{L}_{i}(\textbf{x}_{1:q}; \boldsymbol{\theta})\textbf{z}\right\}$ in the simple regret case. Such a connection enables tractable optimisation through the usage of numerous zero, first, and second-order optimisers developed in the literature~\citep{CMAES, bottou2018optimization, Survey}. Since such an implementation is fairly common in practice~\citep{knudde2017gpflowopt, balandat2019botorch} and not to burden the reader with unnecessary notation, we defer the exact details of the optimisers used in our experiments to appendices~\ref{App:ZeroOrder},~\ref{App:FirstOrder} and~\ref{App:SecondOrder}. Here, we briefly mention that we surveyed three zero-order optimisers, eight first-order algorithms and one well-known approximate second-order method.  
    \paragraph{Zeroth-Order Optimisers in ERM-BO:} Zeroth-order methods optimise objectives based on function value information and have emerged from many different fields. In the online learning literature, for example, development of zeroth-order methods is mostly theoretical aiming at efficient and optimal regret guarantees~\citep{Elad, Bandits, 2020_Gabillon} -- a challenging topic in itself. Empirical successes of such procedures have been achieved in isolated instances~\citep{shalev2007online, viappiani2009regret, contal2013parallel, chen2013combinatorial,  bresler2016collaborative, ariu2020regret, hallak2020regret}. Mainstream implementation of zeroth-order optimisers for BO, however, are of the evolutionary type updating generations of $\textbf{x}$ through a process of adaptation and mutation~\citep{bentley1999evolutionary}. 
    
    In our experiments, we used three such strategies, varying from simple to advanced. The most simple among the three was random search (RS) which acts as a low-memory, low-compute baseline. The second, corresponds to a covariance matrix evolutionary strategy (CMA-ES) that generates updates of the mean and covariance of a multivariate normal based on average sample ranks gathered from function value information~\citep{hansen1996adapting, CMAES}. The third and final algorithm was differential evolution (DE) which is widely considered a go-to in evolutionary optimisation~\citep{price1996differential, baioletti2020differential}, e.g., NSGA I and II~\citep{deb2002fast} as implemented in~\citep{pymoo}. DE continuously updates a population of candidate solutions via component-wise mutation performing selection according to a mutation probability $p_{\text{mutation}}$. More details are available in Appendix~\ref{App:ZeroOrder}.    
    \paragraph{First-Order Optimisers in ERM-BO:} 
    First-order optimisation techniques rely on gradient information to compute updates of $\textbf{x}$. They are iterative in nature running for a total of $T$ iterations and executing a variant of the following rule at each step\footnote{For simplicity in the notation for acquisition functions $\alpha(\textbf{x}_{1:q, 0}|\mathcal{D}_i)$ we drop the subscript with the type.}: 
    \begin{equation}
    \label{Eq:General1}
        \textbf{x}_{1:q, t+1} = \delta_t \textbf{x}_{1:q, t} + \eta_{t}\frac{\boldsymbol{\phi}_{t}^{(1)}\left(\overline{\nabla \alpha(\textbf{x}_{1:q, 0}|\mathcal{D}_i)}, \dots, \overline{\nabla \alpha(\textbf{x}_{1:q, t}|\mathcal{D}_i)}, \left\{\beta_{k}^{(1)}\right\}_{k=0}^{t}\right)}{\boldsymbol{\phi}_{t}^{(2)}\left(\overline{\nabla \alpha(\textbf{x}_{1:q, 0}|\mathcal{D}_{i})}^{2}, \dots, \overline{\nabla \alpha(\textbf{x}_{1:q, t}|\mathcal{D}_{i})}^{2}, \left\{\beta_{k}^{(2)}\right\}_{k=0}^{t}, \epsilon\right)} \ \ \text{(General update),}
    \end{equation}
    where $\delta_{t}$ is a weighting that depends on the type of algorithm used, $\eta_t$ is a typically decaying learning rate, $\boldsymbol{\phi}_{t}^{(1)}(\cdot)$ and $\boldsymbol{\phi}_{t}^{(2)}(\cdot)$ are history-dependent mappings that vary between algorithms with the ratio executed element-wise, $\left\{\beta_{k}^{(1)}\right\}_{k=0}^{t}$ and $\left\{\beta_{k}^{(2)}\right\}_{k=0}^{t}$ are history-weighting parameters, and $\epsilon$ a small positive constant used to avoid division by zero. Additionally, $\overline{\nabla \alpha(\textbf{x}_{1:q, 0}|\mathcal{D}_{i})}, \dots, \overline{\nabla \alpha(\textbf{x}_{1:q, t}|\mathcal{D}_{i})}$ represent sub-sampled gradient estimators that are acquired using Monte-Carlo samples of $\textbf{z} \sim \mathcal{N}(0, \textbf{I})$. It is also worth noting that differentiating through the $\max$ operator that appears in all acquisitions can be performed either using sub-gradients or by propagating through the max value of the corresponding vector.  
    
    To elaborate our generalised form, we realise that one can easily recover Adam's~\citep{ADAM} update equation by setting $\delta_{1} = \dots =  \delta_{T} =1$, $ \beta_{1}^{(1)} = \dots = \beta_{T}^{(1)} = \beta_{1}$,  $ \beta_{1}^{(2)} = \dots = \beta_{T}^{(2)} = \beta_{2}$, and $\boldsymbol{\phi}_{t}^{(1)}$ and $\boldsymbol{\phi}_{t}^{(2)}$ to: 
    \begin{align*}
        \boldsymbol{\phi}_{t}^{(1)}\left(\overline{\nabla \alpha(\textbf{x}_{1:q, 0}|\mathcal{D}_{i})}, \dots, \overline{\nabla \alpha(\textbf{x}_{1:q, t}|\mathcal{D}_{i})}, \beta_{1} \right) & = \frac{1 - \beta_{1}}{1 - \beta_{1}^{t}} \sum_{k=0}^{t} \beta_{1}^{k}\overline{\nabla \alpha(\textbf{x}_{1:q, t-k}|\mathcal{D}_{i})}, \\ 
        \boldsymbol{\phi}_{t}^{(2)}\left(\overline{\nabla \alpha(\textbf{x}_{1:q, 0}|\mathcal{D}_{i})}^{2}, \dots, \overline{\nabla \alpha(\textbf{x}_{1:q, t}|\mathcal{D}_{i})}^{2}, \beta_{2}, \epsilon \right) &= \sqrt{\frac{1 - \beta_{2}}{1 - \beta_{2}^{t}} \sum_{k=0}^{t} \beta_{2}^{k}\overline{\nabla \alpha(\textbf{x}_{1:q, t-k}|\mathcal{D}_{i})}^{2}} + \epsilon.
    \end{align*}
    
    Of course, Adam is yet another special case of Equation~\ref{Eq:General1}. For notational convenience, we defer the detailed derivations of other optimisers including SGA~\citep{RobbMonr51}, RProp~\citep{riedmiller1993direct}, RMSprop \citep{2012_Hinton}, AdamW~\citep{loshchilov_2019_adamw}, AdamOS (an Adam adaptation with new hyperparameters that we propose in this paper), AdaGrad~\citep{Duchi_2011_adagrad}, and AdaDelta~\citep{zeiler_2012_adadelta} to Appendix~\ref{App:FirstOrder}.

    \paragraph{Second-Order Optimisers in ERM-BO:} 
    Along with gradient information, second-order optimisers utilise Hessian (sometimes the Fisher matrix instead~\citep{amari1997neural, pascanu2013revisiting}) information for maximising objective functions. The general iterative update equation for a second-order method is given by:
    \begin{align*}
        \textbf{x}_{1:q,t+1} = \textbf{x}_{1:q,t} - \eta_t\left[\overline{\nabla^2\alpha(\textbf{x}_{1:q,t}|\mathcal{D}_{i})}\right]^{-1}\overline{\nabla \alpha(\textbf{x}_{1:q,t}|\mathcal{D}_{i})}\ \ \ \ \  \text{(General update),}
    \end{align*}
    where $\overline{\nabla^2\alpha(\textbf{x}_{1:q,t}|\mathcal{D}_{i})}$ is an approximation to the true Hessian $\nabla^2\alpha(\textbf{x}_{1:q,t}|\mathcal{D}_{i})$ as evaluated on the current iterate $\textbf{x}_{1:q,t}$, and $\overline{\nabla \alpha(\textbf{x}_{1:q,t}|\mathcal{D}_{i})}$ denotes a gradient estimate that is acquired through Monte Carlo samples as described above. It is worth emphasising the need for the approximation $\overline{\nabla^2\alpha(\textbf{x}_{1:q,t}|\mathcal{D}_{i})}$ to $\nabla^2\alpha(\textbf{x}_{1:q,t}|\mathcal{D}_{i})$ due to the large size of the true Hessian matrix ($\mathbb{R}^{dq\times dq}$ in our case), as well as the necessity to compute an inverse at every iteration of the update. Numerous approximation techniques with varying degrees of accuracy have been proposed in the literature~\citep{shanno1970conditioning, mokhtari2014res,mokhtari2015global, byrd2016stochastic}. In this paper, however, we make use of L-BFGS~\citep{1997_Zhu} due to its widespread adoption in both GPs and BO~\citep{2006_Williams, balandat2019botorch}. Exact details and pseudocode for L-BFGS are comprehensively presented in Appendix~\ref{App:SecondOrder}.

    \subsection{FSM-BO \& Connections to Compositional Optimisation}\label{sec:FSM-BO} 
    Rather than considering the problem of acquisition function maximisation as an instance of empirical risk minimisation, we can follow an alternative route and focus on finite sum approximations. To do so, imagine we acquire $M$ independent and identically-distributed samples from $\mathcal{N}(\textbf{0}, \textbf{I})$, $\{\textbf{z}_{m}\}_{m=1}^{M}$, upfront before the beginning of any acquisition function optimisation step. Assuming fixed samples for now, we can write finite-sum forms of the reparameterised acquisition functions (those from Section~\ref{Sec:AcqOp}) using a simple Monte Carlo estimator as follows: 
    \begin{align}
        \alpha_{\text{rq-EI}}^{(\text{FSM})}(\textbf{x}_{1:q}|\mathcal{D}_{i}) & = \frac{1}{M} \sum_{m=1}^{M} \max_{j\in1:q}\left\{\text{ReLU}\left(\boldsymbol{\mu}_{i}(\textbf{x}_{1:q}; \boldsymbol{\theta}) + \textbf{L}_{i}(\textbf{x}_{1:q}; \boldsymbol{\theta})\textbf{z}_{m} - f(\textbf{x}_{i}^{+})\textbf{1}_{q}\right)\right\}, \label{eq:reparam-EIFSM}\\
        \alpha_{\text{rq-SR}}^{(\text{FSM})}(\textbf{x}_{1:q}|\mathcal{D}_{i}) & =  \frac{1}{M} \sum_{m=1}^{M}\max_{j\in1:q}\left\{\boldsymbol{\mu}_{i}(\textbf{x}_{1:q}; \boldsymbol{\theta}) + \textbf{L}_{i}(\textbf{x}_{1:q}; \boldsymbol{\theta})\textbf{z}_m\right\},\label{eq:reparam-SRFSM} \\
        \alpha_{\text{rq-UCB}}^{(\text{FSM})}(\textbf{x}_{1:q}|\mathcal{D}_{i}) & =  \frac{1}{M}\sum_{m=1}^{M}\max_{j\in1:q}\left\{\boldsymbol{\mu}_i(\textbf{x}_{1:q}; \boldsymbol{\theta}) + \sqrt{\sfrac{\beta \pi}{2}}|\textbf{L}_{i}(\textbf{x}_{1:q}; \boldsymbol{\theta})\textbf{z}_{m}|\right\},\label{eq:reparam-UCBFSM} \\
        \alpha_{\text{rq-PI}}^{(\text{FSM})}(\textbf{x}_{1:q}|\mathcal{D}_{i}) &=  \frac{1}{M}\sum_{m=1}^{M}\max_{j\in1:q}\left\{\text{Sig}\left(\frac{\boldsymbol{\mu}_{i}(\textbf{x}_{1:q}; \boldsymbol{\theta}) + \textbf{L}_{i}(\textbf{x}_{1:q}; \boldsymbol{\theta})\textbf{z}_{m} - f(\textbf{x}_{i}^{+})\textbf{1}_{q}}{\tau}\right)\right\}.\label{eq:reparam-PIFSM} 
    \end{align}
    At this stage, we can execute any off-the-shelf optimiser to maximise the finite sum version of the acquisitions, i.e., Equations~ 9 to 12. Contrary to ERM-BO which samples new $\textbf{z}$ vectors at each iteration, the FSM formulation fixes $\{\textbf{z}_{m}\}_{m=1}^{M}$ and mini-batches from this fixed pool to compute necessary gradients and Hessian estimates for first and second-order methods respectively.  
    At first sight, one might believe that ERM and FSM are the only plausible approximation forms of acquisition functions in BO. Upon further investigation, however, we realise that finite sum myopic acquisitions adhere to yet another configuration that is still to be (well-) explored in the literature. Not only does this new form allow for novel solvers not yet attempted in acquisition function maximisation, but also seems to significantly outperform both ERM-and FSM-BO in practice, cf. Section~\ref{Sec:Exps}. 
    
    \subsubsection{Comp-BO: A Compositional Form for Myopic Acquisition Functions}\label{Sec: Section_Comp_form}
    Recently, the optimisation community has displayed an increased interest in developing specialised algorithms for compositional (or nested) objectives due to their prevalence in subfields of machine learning, e.g., in model-agnostic-meta-learning~\citep{tutunov2020cadam}, semi-implicit variational inference~\citep{yin2018semi}, dynamic programming and reinforcement learning~\citep{2017_MWang}. In each of these examples, compositional solvers have demonstrated efficiency advantages when compared to other algorithms which begs the question as to whether these improvements can be ported to Bayesian  optimisation. 
    
    From a definition perspective, compositional problems involve maximising an objective that consists of a non-linear nesting of expectations of random variables: 
    \begin{equation}
    \label{Eq:Comp}
        \max_{\textbf{x}_{1:q}} \mathbb{E}_{\nu}[f_{\nu}(\mathbb{E}_{\omega}[\textbf{g}_{\omega}(\textbf{x}_{1:q})])],
    \end{equation}
    where $\nu$ and $\omega$ are (not necessarily iid) random variables sampled from $\mathcal{P}_{\nu}(\cdot)$ and $\mathcal{P}_{\omega}(\cdot)$ respectively~\citep{wang2016stochastic}, $f_{\nu}(\cdot)$ a stochastic function, and $\textbf{g}_{\omega}(\cdot)$ is a stochastic map. Hence to benefit from such techniques, our first step consists of transforming the finite-sum versions of the acquisition functions above into a composed (or nested) form that abides by the structure in Equation~\ref{Eq:Comp}. Interestingly, this can easily be achieved if we look at the problem from a matrix-vector perspective. To illustrate, consider $\alpha_{\text{rq-EI}}^{(\text{FSM})}(\textbf{x}_{1:q}|\mathcal{D}_{i})$ and define $\textbf{g}^{(\text{EI})}_{\omega}(\textbf{x}_{1:q})$ to be a $q \times M$ matrix such that the $\omega^{th}$ column is set to $\textbf{v}^{(\text{EI})}_{\omega} = \text{ReLU}\left(\boldsymbol{\mu}_{i}(\textbf{x}_{1:q}; \boldsymbol{\theta}) + \textbf{L}_{i}(\textbf{x}_{1:q}; \boldsymbol{\theta})\textbf{z}_{\omega} - f(\textbf{x}_{i}^{+})\textbf{1}_{q}\right) \in \mathbb{R}^{q}$ with $\omega$ uniformly distributed in $[1:M]$, and set the other columns to $\textbf{0}_{q}$: 
    \begin{equation*}
        \textbf{g}^{(\text{EI})}_{\omega}(\textbf{x}_{1:q})= [\textbf{0}_{q}, \dots, \textbf{v}^{(\text{EI})}_{\omega}, \dots, \textbf{0}_{q}].
    \end{equation*}
    Clearly, if we consider the expectation with respect to $\omega \sim \text{Uniform}([1:M])$, we arrive at the following matrix that sums all information across $\{\textbf{z}_{m}\}_{m=1}^{M}$:
    \begin{equation*}
        \mathbb{E_{\omega}}[\textbf{g}^{(\text{EI})}_{\omega}(\textbf{x}_{1:q})] = \frac{1}{M} [\textbf{v}^{(\text{EI})}_{1}, \dots, \textbf{v}^{(\text{EI})}_{m}, \dots, \textbf{v}^{(\text{EI})}_{M}],
    \end{equation*}
    with $\textbf{v}^{(\text{EI})}_{m} = \text{ReLU}\left(\boldsymbol{\mu}_{i}(\textbf{x}_{1:q}; \boldsymbol{\theta}) + \textbf{L}_{i}(\textbf{x}_{1:q}; \boldsymbol{\theta})\textbf{z}_{m} - f(\textbf{x}_{i}^{+})\textbf{1}_{q}\right)$ being a $q$-dimensional vector. To attain the original form of $\alpha_{\text{rq-EI}}^{(\text{FSM})}(\cdot)$, we further introduce a deterministic outer function $f^{(\text{EI})}: \mathbb{R}^{q \times M} \rightarrow \mathbb{R}$ as follows: 
    \begin{align*}
        \alpha_{\text{rq-EI}}^{(\text{Comp})} (\textbf{x}_{1:q}|\mathcal{D}_{i}) = f^{(\text{EI})} (\mathbb{E}_{\omega}[\textbf{g}^{(\text{EI})}_{\omega}(\textbf{x}_{1:q})]) & = \frac{1}{M} \sum_{m=1}^{M} \max_{j \in 1:q} \textbf{v}^{(\text{EI})}_{m}  = \alpha_{\text{rq-EI}}^{(\text{FSM})}. 
    \end{align*}
    Importantly, the above shows that a finite-sum expected improvement acquisition can be written in a compositional (nested) form with $\alpha_{\text{rq-EI}}^{(\text{FSM})} = f (\mathbb{E}_{\omega}[\textbf{g}_{\omega}(\textbf{x})])$. In our derivations, we have considered a deterministic outer function $f(\cdot)$ leading us to a special case of Equation~\ref{Eq:Comp} where $P_\nu(\cdot)$ is Dirac. Such a consideration is mostly due to the fact that $q$ is typically in the order of tens or hundreds in BO allowing for exact outer summations. In the case of large batch sizes, our formulation can easily be generalised to a stochastic setting exactly matching a compositional form as shown in Appendix~\ref{App:StochastiCCompFormulation}.   
    
    Following the same strategy above, we can now reformulate all other acquisition functions as instances of compositional optimisation. Next, we list these results and refer the reader to Appendix~\ref{App:StochastiCCompFormulation} for a detailed exposition. First, we choose $\omega \sim \text{Uniform}([1:M])$ and then consider the following inner matrix mappings:
    \begin{align*}
        \textbf{g}^{\text{(PI)}}_{\omega}(\textbf{x}_{1:q}) &= [\textbf{0}_{q}, \dots, \textbf{v}^{(\text{PI})}_{\omega}, \dots, \textbf{0}_{q}]\in\mathbb{R}^{q\times M},\\\nonumber
        \textbf{g}^{\text{(SR)}}_{\omega}(\textbf{x}_{1:q}) &= [\textbf{0}_{q}, \dots, \textbf{v}^{(\text{SR})}_{\omega}, \dots, \textbf{0}_{q}]\in\mathbb{R}^{q\times M}\\\nonumber
        \textbf{g}^{\text{(UCB)}}_{\omega}(\textbf{x}_{1:q}) &= [\textbf{0}_{q}, \dots, \textbf{v}^{(\text{UCB})}_{\omega}, \dots, \textbf{0}_{q}]\in\mathbb{R}^{q\times M}
    \end{align*}
    where the $q-$dimensional vectors $\textbf{v}^{(\text{PI})}_{m}, \textbf{v}^{(\text{SR})}_{m}$, and $ \textbf{v}^{(\text{UCB})}_{m}$ are defined as (for $m\in [1:M]$):
    \begin{align*}
        \textbf{v}^{(\text{PI})}_{m} &= \frac{1}{\tau}\left[\boldsymbol{\mu}_{i}(\textbf{x}_{1:q}; \boldsymbol{\theta}) + \textbf{L}_{i}(\textbf{x}_{1:q}; \boldsymbol{\theta})\textbf{z}_{m} - f(\textbf{x}_{i}^{+})\textbf{1}_{q}\right],\\\nonumber
        \textbf{v}^{(\text{SR})}_{m} &= \boldsymbol{\mu}_{i}(\textbf{x}_{1:q}; \boldsymbol{\theta}) + \textbf{L}_{i}(\textbf{x}_{1:q}; \boldsymbol{\theta})\textbf{z}_{m},\\\nonumber
        \textbf{v}^{(\text{UCB})}_{m} & = \boldsymbol{\mu}_{i}(\textbf{x}_{1:q}; \boldsymbol{\theta}) + \sqrt{\sfrac{\beta\pi}{2}}\left|\textbf{L}_{i}(\textbf{x}_{1:q}; \boldsymbol{\theta})\textbf{z}_{m}\right|.
    \end{align*}
    Now, properly selecting the outer functions $f^{(\text{PI})}(\cdot), f^{(\text{SR})}(\cdot)$, and $f^{(\text{UCB})}(\cdot)$ gives us:
    \begin{align*}
        \alpha_{\text{rq-PI}}^{(\text{Comp})} (\textbf{x}_{1:q}|\mathcal{D}_{i})= f^{(\text{PI})}(\mathbb{E}_{\omega}[\textbf{g}^{\text{(PI)}}_{\omega}(\textbf{x}_{1:q})] ) & = \frac{1}{M}\sum_{m=1}^{M}\max_{j \in 1:q}\left\{\text{Sig}\left(\textbf{v}^{(\text{PI})}_m\right)\right\} = \alpha_{\text{rq-PI}}^{(\text{FSM})} (\textbf{x}_{1:q}|\mathcal{D}_{i}),\\
        \alpha_{\text{rq-SR}}^{(\text{Comp})} (\textbf{x}_{1:q}|\mathcal{D}_{i}) = f^{(\text{SR})}(\mathbb{E}_{\omega}[\textbf{g}^{\text{(SR)}}_{\omega}(\textbf{x}_{1:q})]) & = \frac{1}{M}\sum_{m=1}^{M}\max_{j \in 1:q}\left\{\textbf{v}^{(\text{SR})}_m\right\} = \alpha_{\text{rq-SR}}^{(\text{FSM})} (\textbf{x}_{1:q}|\mathcal{D}_{i}),\\ 
        \alpha_{\text{rq-UCB}}^{(\text{Comp})} (\textbf{x}_{1:q}|\mathcal{D}_{i}) = f^{(\text{UCB})}(\mathbb{E}_{\omega}[\textbf{g}^{\text{(UCB)}}_{\omega}(\textbf{x}_{1:q})])& = \frac{1}{M}\sum_{m=1}^M\max_{j\in1:q}\left\{\textbf{v}^{(\text{UCB})}_m\right\} = \alpha_{\text{rq-UCB}}^{(\text{FSM})} (\textbf{x}_{1:q}|\mathcal{D}_{i}).
    \end{align*}
    Clearly, the results above recover the formulations of the acquisition functions given in Equations \ref{eq:reparam-SRFSM} -  \ref{eq:reparam-PIFSM} while making them amenable to compositional solvers, a new class of optimisers not yet well-studied in the Bayesian optimisation literature. We detail such compositional optimisers next.
    
    \paragraph{Zeroth-Order Compositional Solvers for BO:} Of course,  the compositional forms presented above are still suitable for zeroth-order methods (Section \ref{Sec:ERM_BO}). The distinguishing factor from non-compositional forms is the evaluation process of nested objectives which requires careful consideration. In the case of $\alpha_{\text{rq-EI}}^{(\text{Comp})}(\textbf{x}_{1:q}|\mathcal{D}_{i})$, for example, the inner expectation $\mathbb{E}_{\omega}[\textbf{g}^{(\text{EI})}_{\omega}(\textbf{x})]$ in Equation \ref{Eq:Comp} can be evaluated using a Monte Carlo approximation: 
    \begin{equation*}
        \mathbb{E}_{\omega}[\textbf{g}^{(\text{EI})}_{\omega}(\textbf{x}_{1:q})] \approx \frac{1}{K}\sum_{m=1}^{K}\textbf{g}^{(\text{EI})}_{\omega_m}(\textbf{x}_{1:q}), \ \ \text{with $K < M$ being a mini-batch of $\{\textbf{z}_{m}\}_{m=1}^{M}$}. 
    \end{equation*}
    Furthermore, the outer function is estimated by $f^{(\text{EI})}(\mathbb{E}_{\omega}[\textbf{g}^{(\text{EI})}_{\omega}(\textbf{x}_{1:q})]) \approx f^{(\text{EI})}\left(\frac{1}{K}\sum_{m=1}^{K}\textbf{g}^{(\text{EI})}_{\omega_m}(\textbf{x}_{1:q})\right)$, where such an estimate asymptotically ($K\to\infty$) converges to the true expectation due to the continuity of $f^{(\text{EI})}(\cdot)$:
    \begin{align*}
        \lim_{K\to\infty}f^{(\text{EI})}\left(\frac{1}{K}\sum_{m=1}^{K}\textbf{g}^{(\text{EI})}_{\omega_m}(\textbf{x}_{1:q})\right) = f^{(\text{EI})}(\mathbb{E}_{\omega}[\textbf{g}^{(\text{EI})}_{\omega}(\textbf{x}_{1:q})]).
    \end{align*}
    Clearly, this observation allows us to straightforwardly apply any of the three considered zero-order methods (CMA-ES, DE, and RS) for determining updates of $\textbf{x}_{1:q}$. Certainly, such Monte Carlo approximations are not distinctive for $\alpha_{\text{rq-EI}}^{(\text{Comp})}(\textbf{x}_{1:q}|\mathcal{D}_{i})$, allowing us to follow the same scheme for $\alpha_{\text{rq-PI}}^{(\text{Comp})}(\textbf{x}_{1:q}|\mathcal{D}_{i})$, $\alpha_{\text{rq-SR}}^{(\text{Comp})}(\textbf{x}_{1:q}|\mathcal{D}_{i})$, and $\alpha_{\text{rq-UCB}}^{(\text{Comp})}(\textbf{x}_{1:q}|\mathcal{D}_{i})$.

    \paragraph{First-Order Compositional Solvers for BO:} In contrast to zeroth-order compositional methods, where the only difference between them and their non-compositional counterparts is in the evaluation of the objective function, first-order compositional optimisers require more sophisticated techniques due to the difficulty associated in acquiring unbiased gradients of nested objectives. To elaborate, let us carry on with our running example and consider the gradient of $\alpha_{\text{rq-EI}}^{(\text{Comp})}(\textbf{x}_{1:q}|\mathcal{D}_{i}) = f^{(\text{EI})}(\mathbb{E}_{\omega}[\textbf{g}^{(\text{EI})}_{\omega}(\textbf{x}_{1:q})])$. Using the chain rule, we can easily see that such a gradient involves a product of the Jacobian of $\textbf{g}_{\omega}(\textbf{x}_{1:q})$ with the gradient of $f^{(\text{EI})}(\cdot)$ that is to be evaluated around the inner mapping\footnote{Of course, a simple solution corresponds to a Nested Monte Carlo approach that approximates both inner and outer mappings with samples from $\omega$ and $\nu$ and then executes standard off-the-shelf algorithms. In our experiments, we make use of such a technique which we refer to as Adam-Nested (see Section~\ref{Sec:Exps}) but realise that dedicated first-order compositional solvers tend to outperform such a scheme.}:
    \begin{equation*}
        \nabla_{\text{vec}(\textbf{x}_{1:q})} \alpha_{\text{rq-EI}}^{(\text{Comp})}(\textbf{x}_{1:q}|\mathcal{D}_{i}) = \mathbb{E}_{\omega}[\nabla_{\text{vec}(\textbf{x}_{1:q})}\textbf{g}_{\omega}^{(\text{EI})}(\textbf{x}_{1:q})]^{\mathsf{T}}\nabla_{\boldsymbol{\zeta}}f^{(\text{EI})}(\boldsymbol{\zeta})\left.\right\vert_{\boldsymbol{\zeta} = \mathbb{E}_{\omega}[\textbf{g}_{\omega}^{(\text{EI})}(\textbf{x}_{1:q})]},
    \end{equation*}
    where we use $\text{vec}(\textbf{x}_{1:q}) \in \mathbb{R}^{dq}$ to denote an unrolled vector across all dimensions $d$ and batch sizes $q$. When attempting to acquire an unbiased estimate of $    \nabla_{\text{vec}(\textbf{x}_{1:q})} \alpha_{\text{rq-EI}}^{(\text{Comp})}(\textbf{x}_{1:q}|\mathcal{D}_{i})$, we realise that the first term can be approximated by simple Monte Carlo: 
    \begin{equation*}
        \mathbb{E}_{\omega}[\nabla_{\text{vec}(\textbf{x}_{1:q})}\textbf{g}_{\omega}^{(\text{EI})}(\textbf{x}_{1:q})] \approx \frac{1}{K_{1}} \sum_{m=1}^{K_{1}}\nabla_{\text{vec}(\textbf{x}_{1:q})}\textbf{g}_{\omega_{m}}^{(\text{EI})}(\textbf{x}_{1:q}),
    \end{equation*}
    with $K_{1} < M$ being a batch size. The second part, however, is tougher to estimate as it involves a gradient of a non-linear nesting of an expected value, i.e., $\nabla_{\boldsymbol{\zeta}}f^{(\text{EI})}(\boldsymbol{\zeta})\left.\right\vert_{\boldsymbol{\zeta} = \mathbb{E}_{\omega}[\textbf{g}_{\omega}^{(\text{EI})}(\textbf{x}_{1:q})]}$. To resolve this problem, in the compositional optimisation literature ~\citep{2017_Wang, tutunov2020cadam}, typically an auxiliary variable $\textbf{u}$ is introduced and an exponentially-weighted average of $\boldsymbol{\zeta}$ is used, resulting in asymptotically-vanishing biases. To acquire such behaviour, not only do we need to update $\textbf{x}_{1:q}$ but we also need to modify $\textbf{u}$ and our estimation of $\boldsymbol{\zeta}$. As such, most compositional solvers execute three subroutines (main $\textbf{x}_{1:q}$, auxiliary $\textbf{u}$ and $\boldsymbol{\zeta}$) between iterations $t$ and $t+1$ -- the first to generate $\textbf{x}_{1:q, t+1}$, the second for $\textbf{u}_{t+1}$ and the third for $\boldsymbol{\zeta}_{t+1}$. Rather than presenting every subroutine for all utilised algorithms across all acquisition functions, here we keep the exposition general and provide a set of unifying update rules, deferring exact details to Appendix~\ref{App:FirstOrder_Compos}. To that end, we introduce four history-dependent mappings $\boldsymbol{\phi}_{t}^{(1)}(\cdot)$, $\boldsymbol{\phi}_{t}^{(2)}(\cdot)$, $\boldsymbol{\phi}_{t}^{(3)}(\cdot)$ and $\boldsymbol{\phi}_{t}^{(4)}(\cdot)$. $\boldsymbol{\phi}_{t}^{(1)}(\cdot)$ and $\boldsymbol{\phi}_{t}^{(2)}(\cdot)$ act on sub-sampled gradient histories, and their corresponding squares, for updating $\textbf{x}_{1:q, t}$ as follows: 
    \begin{align}
    \label{Eq:General2a}
        &\underline{\textbf{Main variable update: }}\\\nonumber
        &\textbf{x}_{1:q, t+1} =  \textbf{x}_{1:q, t} + \eta_{t}\frac{\boldsymbol{\phi}_{t}^{(1)}\left(\left\{\overline{\nabla_{\text{vec}(\textbf{x}_{1:q})}\alpha^{(\text{Comp})}(\textbf{x}_{1:q, k}, \boldsymbol{\zeta}_k|\mathcal{D}_{i})}\right\}^{t}_{k=0}, \left\{\gamma_{k}^{(1)}\right\}_{k=0}^{t}\right)}{\boldsymbol{\phi}_{t}^{(2)}\left(\left\{\overline{\nabla_{\text{vec}(\textbf{x}_{1:q})}\alpha^{(\text{Comp})}(\textbf{x}_{1:q, k}, \boldsymbol{\zeta}_k|\mathcal{D}_{i})}^2\right\}^{t}_{k=0}, \left\{\gamma_{k}^{(2)}\right\}_{k=0}^{t}, \epsilon\right)},
    \end{align}
    where $\eta_{t}$ is a learning rate, $\{\gamma_{k}^{(1)}\}_{k=0}^{t}$ and $\{\gamma_{k}^{(2)}\}_{k=0}^{t}$ are history-dependent weightings that vary across algorithms. In Equation~\ref{Eq:General2a}, we also use $\overline{\nabla_{\text{vec}(\textbf{x}_{1:q})}\alpha^{(\text{Comp})}(\textbf{x}_{1:q, k}, \boldsymbol{\zeta}_k|\mathcal{D}_{i})}$ to define a compositional gradient estimate that can be written as: 
    \begin{equation}
    \label{Eq:SubGrad}
        \overline{\nabla_{\text{vec}(\textbf{x}_{1:q})}\alpha^{(\text{Comp})}(\textbf{x}_{1:q, k}, \boldsymbol{\zeta}_k|\mathcal{D}_{i})} = \left[\frac{1}{K_1}\sum_{m=1}^{K_1}\nabla_{\text{vec}(\textbf{x}_{1:q})}\textbf{g}^{(\text{type})}_{\omega_m}(\textbf{x}_{1:q, k})\right]^{\mathsf{T}}\nabla_{\boldsymbol{\zeta}}f^{(\text{type})}(\boldsymbol{\zeta}_k),
    \end{equation}
    with $\textbf{g}^{(\text{type})}_{\omega_m}$ and $f^{(\text{type})}$ denoting the inner and outer mapping of a compositional formulation where $\text{type} \in \{\text{EI}, \text{PI}, \text{SR}, \text{UCB}\}$. With $\textbf{x}_{1:q, t+1}$ computed, the next step is to update $\textbf{u}_{t}$ and $\boldsymbol{\zeta}_{t}$ which can be achieved through $\boldsymbol{\phi}_{t}^{(3)}(\cdot)$ and $\boldsymbol{\phi}_{t}^{(4)}(\cdot)$ in the following manner:
    \begin{align}
        \textbf{u}_{t+1} &=  \boldsymbol{\phi}^{(3)}_{t+1}\left(\textbf{x}_{1:q, 0}, \dots,  \textbf{x}_{1:q, t+1}, \{\beta_k\}^{t}_{k=0}\right), \label{Eq:General2b} \\
        \boldsymbol{\zeta}_{t+1} &= \boldsymbol{\phi}^{(4)}_{t+1}\left(\overline{\textbf{g}^{(\text{type})}(\textbf{u}_1)}, \dots,  \overline{\textbf{g}^{(\text{type})}(\textbf{u}_{t+1})},   \{\beta_k\}^{t}_{k=0},\boldsymbol{\zeta}_0, \textbf{u}_0\right), \label{Eq:General2c}
    \end{align}
    where $\{\beta_{k}\}_{k=0}^{t}$ is a set of  free parameters\footnote{It is worth noting that in Appendix~\ref{App:HyperParam} we provide a complete set of all hyperparameters used across all 28 optimisers.}, $\boldsymbol{u}_{0}$ and $\boldsymbol{\zeta}_{0}$ are initialisations that in turn depend on $\textbf{x}_{1:q, 0}$. Furthermore, in Equation~\ref{Eq:General2c} we used $\overline{\textbf{g}^{(\text{type})}(\cdot )}$ to represent a Monte Carlo estimate of the inner mapping, i.e., 
    \begin{equation*}
        \overline{\textbf{g}^{(\text{type})}(\cdot )} = \frac{1}{K_2} \sum_{m=1}^{K_2} \textbf{g}_{\omega_{m}}^{(\text{type})}(\cdot),
    \end{equation*}
    where $K_2 < M$ is a batch size and $\text{type} \in \{\text{EI}, \text{PI}, \text{SR}, \text{UCB}\}$. As an illustrative example, we note that one can recover CAdam~\citep{tutunov2020cadam} by instantiating the above as follows:  
    \begin{align*}
        &\boldsymbol{\phi}_{t}^{(1)}\left(\left\{\overline{\nabla_{\text{vec}(\textbf{x}_{1:q})}\alpha^{(\text{Comp})}(\textbf{x}_{1:q, k}, \boldsymbol{\zeta}_k|\mathcal{D}_{i})}\right\}^{t}_{k=0}, \left\{\gamma_{k}^{(1)}\right\}_{k=0}^{t}\right) =\\\nonumber
        &\hspace{5cm}\sum_{k=0}^t(1 - \gamma^{[1]}_k)\prod_{j=k+1}^{t}\gamma^{[1]}_j\overline{\nabla_{\text{vec}(\textbf{x}_{1:q})}\alpha^{(\text{Comp})}(\textbf{x}_{1:q, k}, \boldsymbol{\zeta}_k|\mathcal{D}_{i})}, \\\nonumber
        &\boldsymbol{\phi}_{t}^{(2)}\left(\left\{\overline{\nabla_{\text{vec}(\textbf{x}_{1:q})}\alpha^{(\text{Comp})}(\textbf{x}_{1:q, k}, \boldsymbol{\zeta}_k|\mathcal{D}_{i})}^2\right\}^{t}_{k=0}, \left\{\gamma_{k}^{(2)}\right\}_{k=0}^{t}, \epsilon\right) = \\\nonumber
        &\hspace{5cm}\sqrt{\sum_{k=0}^t(1 - \gamma^{[2]}_k)\prod_{j=k+1}^{t}\gamma^{[2]}_j\overline{\nabla_{\text{vec}(\textbf{x}_{1:q})}\alpha^{(\text{Comp})}(\textbf{x}_{1:q, k},\boldsymbol{\zeta}_k|\mathcal{D}_{i})}^2} + \epsilon,\\\nonumber
        &\boldsymbol{\phi}^{(3)}_{t}\left(\textbf{x}_{1:q, 0}, \dots,  \textbf{x}_{1:q, t}, \{\beta_k\}^{t-1}_{k=0}\right) = (1 - \beta^{-1}_{t-1})\textbf{x}_{1:q, t-1} + \beta^{-1}_{t-1}\textbf{x}_{1:q, t}, \\\nonumber
        &\boldsymbol{\phi}^{(4)}_t\left(\overline{\textbf{g}^{(\text{type})}(\textbf{u}_1)}, \dots,  \overline{\textbf{g}^{(\text{type})}(\textbf{u}_{t})},   \{\beta_k\}^{t-1}_{k=0},\boldsymbol{\zeta}_0, \textbf{u}_0\right) = \sum_{k=1}^{t}\beta_{k-1}\prod_{j=k}^{t-1}(1 - \beta_j)\overline{\textbf{g}^{(\text{type})}(\textbf{u}_k)}.
    \end{align*}
    Of course, CAdam is just an instance of the generic update rules presented in  Equations~\ref{Eq:General2a}-~\ref{Eq:General2c}. Other first-order compositional methods, such as NASA ~\citep{ghadimi2020single}, ASCGA ~\citep{2017_Wang}, SCGA ~\citep{2017_Wang} and Adam applied to a nested Monte Carlo objective can all be derived from our general form as demonstrated in Appendix~\ref{App:FirstOrder_Compos}.

    \paragraph{Second-Order Compositional Solvers for BO: } For a holistic comparison against ERM-BO, we prefer to use the three same optimisation categories of zero-, first-, and second-order methods in Comp-BO. Although significant progress towards first-order compositional optimisers has been achieved in the literature, second-order techniques tackling the objective in Equation~\ref{Eq:Comp} are yet to be developed. In this paper, we take a first step towards developing second-order compositional methods and propose an adaption of the standard L-BFGS algorithm to handle nested compositional forms. To start, we note that any second-order technique considers function curvature in its update through the usage of Hessian information:
    
    \begin{align*}
        \textbf{x}_{1:q,t+1} = \textbf{x}_{1:q,t} + \eta_t\left[\overline{\nabla^2_{\text{vec}(\textbf{x}_{1:q})\text{vec}(\textbf{x}_{1:q})}\alpha^{(\text{Comp})}(\textbf{x}_{1:q,t}|\mathcal{D}_i)}\right]^{-1}\overline{\nabla_{\text{vec}(\textbf{x}_{1:q})}\alpha^{(\text{Comp})}(\textbf{x}_{1:q,t}|\mathcal{D}_i)},
    \end{align*}
    where $\overline{\nabla^2_{\text{vec}(\textbf{x}_{1:q})\text{vec}(\textbf{x}_{1:q})}\alpha^{(\text{Comp})}(\textbf{x}_{1:q,t}|\mathcal{D}_i)}$ and $\overline{\nabla_{\text{vec}(\textbf{x}_{1:q})}\alpha^{(\text{Comp})}(\textbf{x}_{1:q,t}|\mathcal{D}_i)}$ are stochastic approximations of the Hessian and the gradient of $\alpha^{(\text{Comp})}(\textbf{x}_{1:q,t}|\mathcal{D}_i)$ and $\eta_{t}$ is a learning rate. A compositional structure however, imposes practical limitations for the applicability of any arbitrary second-order method due to two essential difficulties. The first relates to the computation of the Hessian, while the second relates to calculating its inverse. When evaluating $\nabla^2_{\text{vec}(\textbf{x}_{1:q})\text{vec}(\textbf{x}_{1:q})}\alpha^{(\text{Comp})}(\textbf{x}_{1:q}|\mathcal{D}_i)$, we encounter an expensive 3-tensor-vector product -- $\mathcal{O}(d^2 q^3 M)$ with $d$, $q$ and $M$ denoting the dimensionality, batch size of input queries and $\textbf{z}$ respectively -- of the following form:  
    \begin{align*}
        &\nabla^2_{\text{vec}(\textbf{x}_{1:q})\text{vec}(\textbf{x}_{1:q})}\alpha^{(\text{Comp})}(\textbf{x}_{1:q}|\mathcal{D}_i) = \\\nonumber
        &\hspace{2cm}\textbf{J}(\textbf{x}_{1:q})^{\mathsf{T}}\nabla^2_{\boldsymbol{\zeta}\boldsymbol{\zeta}}f(\mathbb{E}_{\omega}[\textbf{g}_{\omega}(\textbf{x}_{1:q})])\textbf{J}(\textbf{x}_{1:q}) + \nabla_{\text{vec}(\textbf{x}_{1:q})}\textbf{J}(\textbf{x}_{1:q})\times_{1}\nabla_{\boldsymbol{\zeta}}f(\mathbb{E}_{\omega}[\textbf{g}_{\omega}(\textbf{x}_{1:q})]),
    \end{align*}
    where $\textbf{J}(\textbf{x}_{1:q}) = \mathbb{E}[\nabla_{\text{vec}(\textbf{x}_{1:q})}\textbf{g}_{\omega}(\textbf{x}_{1:q})]$ is the Jacobian of the inner mapping $\mathbb{E}_{\omega}[\textbf{g}_{\omega}(\textbf{x}_{1:q})]$, the 3-tensor  $\nabla_{\text{vec}(\textbf{x}_{1:q})}\textbf{J}(\textbf{x}_{1:q})$ is the Hessian of $\mathbb{E}_{\omega}[\textbf{g}_{\omega}(\textbf{x}_{1:q})]$, and $\times_1$ is a mode-1 product between a 3-tensor and a vector. Apart from needing such expensive products -- a total of $\mathcal{O}(dq^3M(d+M))$  --  the update rule introduced above further escalates the computational burden by requiring an inverse that is generally cubic in the number of dimensions, i.e., $\mathcal{O}(d^3 q^3)$ in our case. Hence, a feasible approximation for computing $[\nabla^2_{\text{vec}(\textbf{x}_{1:q})\text{vec}(\textbf{x}_{1:q})}\alpha^{(\text{Comp})}(\textbf{x}_{1:q}|\mathcal{D}_i)]^{-1}$ plays a crucial role in the success of any second-order method for compositional objectives. As introduced earlier, BFGS-type methods ameliorate the expense of the calculations by utilising the recursive Sherman-Morison formulae that we also follow here~\citep{riedel1992sherman}. For such an application, we require two curvature pairs $\textbf{s}_{t}$ and $\textbf{h}_{t}$ for recursively approximating the inverse of the Hessian. Namely if $\textbf{s}_{t} = \textbf{x}_{1:q, t} - \textbf{x}_{1:q, t-1}$ and $\textbf{h}_t = \overline{\nabla_{\text{vec}(\textbf{x}_{1:q})}\alpha^{(\text{Comp})}(\textbf{x}_{1:q,t}|\mathcal{D}_i)} - \overline{\nabla_{\text{vec}(\textbf{x}_{1:q})}\alpha^{(\text{Comp})}(\textbf{x}_{1:q, t-1}|\mathcal{D}_i)}$, one can show that 
    \begin{align*}
        \textbf{A}_{t} = \left[\textbf{I} - \frac{\textbf{s}_{t}\textbf{h}^{\mathsf{T}}_{t}}{\textbf{h}^{\mathsf{T}}_{t}\textbf{s}_t}\right]\textbf{A}_{t-1}\left[\textbf{I} - \frac{\textbf{h}_{t}\boldsymbol{s}^{\mathsf{T}}_{t}}{\textbf{h}^{\mathsf{T}}_{t}\textbf{s}_{t}}\right] + \frac{\textbf{s}_t\textbf{s}^{\mathsf{T}}_t}{\textbf{h}^{\mathsf{T}}_t\textbf{s}_t}, 
    \end{align*}
    provides a valid approximation to the $t^{th}$ iteration Hessian inverse when initialising $\textbf{A}_{0} = \textbf{I}$. That is $\textbf{A}_t \approx \left[\nabla^2_{\text{vec}(\textbf{x}_{1:q})\text{vec}(\textbf{x}_{1:q})}\alpha^{(\text{Comp})}(\textbf{x}_{1:q, t}|\mathcal{D}_i)\right]^{-1}$ and memory cost is reduced to $\mathcal{O}(T dq)$, with $T$ being total number of update iterations. Hence, a BFGS-type update can now be written as: 
    \begin{align*}
        \textbf{x}_{1:q, t+1} =  \textbf{x}_{1:q, t+1} + \eta_t \textbf{A}_{t}  \underbrace{\overline{\nabla_{\text{vec}(\textbf{x}_{1:q})}\alpha^{(\text{Comp})}(\textbf{x}_{1:q,t}|\mathcal{D}_i)}}_{\text{ Gradient Monte-Carlo estimate}}. 
    \end{align*}

    \subsubsection{Memory-Efficient Implementations for Comp-BO}\label{Sec:Memory-Efficient}
    Although the ERM-BO and FSM-BO strategies discussed in Sections \ref{Sec:ERM_BO} and \ref{sec:FSM-BO} share commonalities such as the sampling of the reparametrisation variable $\textbf{z}\in\mathbb{R}^q$ and the use of Monte Carlo estimates, one important difference between the approaches is memory complexity - the total amount of space in storage (be that disk or cloud) needed for the complete execution of an optimisation method. It is worthwhile mentioning that the key difference between memory and time resources is that the former can be erased and reused multiple times while the latter cannot, and this distinction plays an important role in the analysis of applied optimisation algorithms.  
    
    For ERM-BO methods, the total amount of required memory is defined by the size of the largest mini-batch sampled during the execution and the memory needed for the iterative update. Since in all ERM-BO algorithms we use mini-batches of a constant size $ K=128 $, and at each iteration $t$ we store only the current iterative value $\textbf{x}_{1:q,t}\in\mathbb{R}^{dq}$ the overall memory complexity is therefore bounded by $\mathcal{O}(Kq + dq)$.
    
    Similarly to empirically-founded techniques, in FSM-BO methods we also store at each step $t$ the current value of the iterate $\textbf{x}_{1:q,t}\in\mathbb{R}^{dq}$ and utilise a mini-batch of samplings of size $K \ll M$. However in contrast to the ERM-BO case, the upfront sampling of $M$ reparameterisation random variables $\textbf{z}$ used in the FSM-BO scenario leads to an $\mathcal{O}(Mq + dq)$ bound for the overall memory capacity. On one hand, large values of $M$ are preferable as they provide a better approximation to the true acquisition functions given in Equations \ref{eq:reparam-EI} - \ref{eq:qPI-smooth}, yet on the other hand, such values of $M$ make finite-sum methods memory stringent.  
    
    To remedy this problem, we propose memory-efficient adaptations of compositional methods: CAdam-ME, NASA-ME and Nested-MC-ME. In a nutshell, all these methods exploit the observation that at any given iteration, stochastic compositional optimisers only require uniform sub-sampling from the fixed collection of $M$ reparametrisation variables $\textbf{z}$. Hence instead of storing $M$ samples upfront, one can draw $K$ of them from $\mathcal{N}(\textbf{0}, \textbf{I})$ at each iteration  resulting in an overall memory complexity given by $\mathcal{O}(Kq + dq)$. For a detailed description of the memory-efficient methods CAdam-ME, NASA-ME and Nested-MC-ME, we refer the reader to Appendix~\ref{App:MemoryEfficient}.
    
    \section{Experiments \& Results}\label{Sec:Exps}

    Having presented a comprehensive set of optimisation techniques suitable for maximising acquisition functions, we now wish to systematically evaluate their empirical performance. Specifically, we design our experimental setup with the intention of answering the following questions:
    

    \begin{enumerate}
        \item Do Finite-Sum Minimisation acquisition functions provide any benefits compared to the more frequently-used Empirical Risk Minimisation versions?
        \item Do compositional optimisers provide any advantages over non-compositional optimisers?
        \item What are the practical savings for using memory-efficient implementations of compositional acquisition functions?
        \item Are compositional methods more computationally expensive than non-compositional optimisation methods and how does runtime scale as a function of the input dimensionality?
        \item How do compositional optimisers perform when optimising real-world black-box functions with noisy evaluations?
    \end{enumerate}

    \noindent In order to answer Questions 1-4, we run twenty-eight optimiser variants on five synthetic, noiseless BBO problems for which the true maxima are known. Knowing the true maxima allows for exact computation of the normalised immediate regret 
    
    \begin{align}
    \label{imm_regret}
        r_t = \frac{|f(\mathbf{\tilde{x_t}}) - f(\mathbf{x^*})|}{|f(\mathbf{\tilde{x_0}}) - f(\mathbf{x^*})|},
    \end{align}
    
    \noindent where $f(\mathbf{x^*})$ is the function value at the global optimiser $\mathbf{x^*}$, $\mathbf{\tilde{x_t}}$ is the algorithm's recommendation at round $t$ and $f(\mathbf{\tilde{x_0}})$ is the regret upon initialisation at round $0$. The use of analytic functions also facilitates the treatment of input dimensionality as an experiment variable. In order to answer question 5, we focus on the tasks from Bayesmark. These tasks possess noise in the evaluations and are more representative of real-world BBO problems. For these latter experiments we take forward the best-performing optimisers observed in the synthetic function experiments. A pictorial summary of the experimental setup is provided in Figure~\ref{fig:exp_variati0ons}.
    
    \begingroup
    \setlength{\intextsep}{0pt}%
    \begin{figure}[h!]
        \centering
        \includegraphics[trim=0em 3em 0em 3em, clip=true, width=\textwidth]{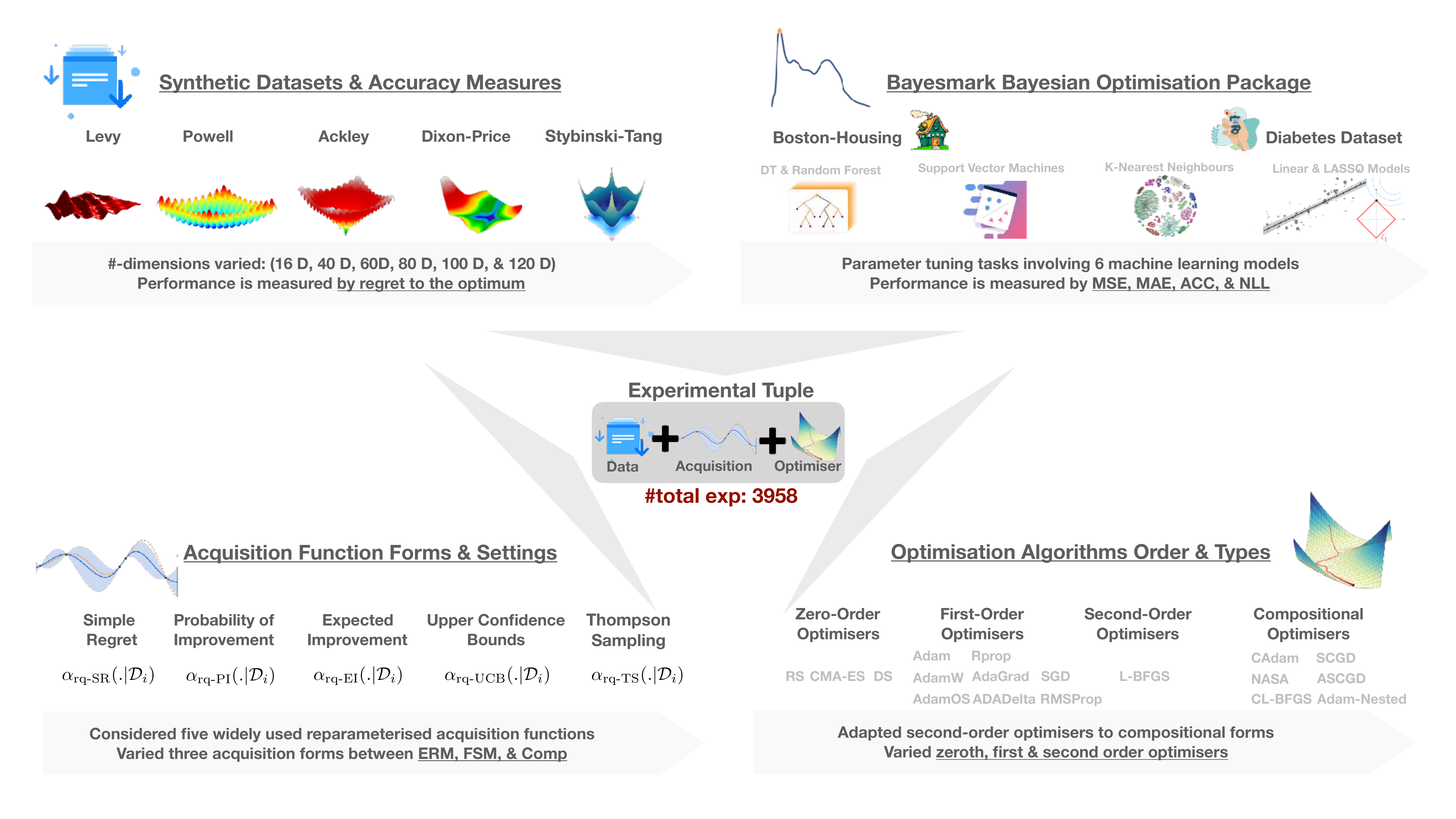}
        \caption{Experiment Overview: \textbf{Top Left}: Synthetic functions (noiseless). \textbf{Top Right}: Bayesmark data (noisy). \textbf{Bottom Left}: Five classes of acquisition function in ERM, Finite-Sum, and Compositional forms. \textbf{Bottom Right}: Four classes of optimiser. Each experiment tuple comprises a dataset, an acquisition function and an optimiser. The study comprises 3958 experiments in total.}
        \label{fig:exp_variati0ons}
    \end{figure}
    
    \paragraph{Surrogate Model: } For all tasks, we use a GP with constant mean function set to the empirical mean of the data, and a $\text{Mat\'{e}rn}(5/2)$ kernel with lengthscale parameter $\boldsymbol{\theta}$. At each acquisition step $k$, the hyperparameters of the GP kernel are estimated  based on the current observed input-output pairs $\mathcal{D}_k$ by optimising the negative log marginal likelihood with a \textit{Gamma} prior over $\boldsymbol{\theta}$. To facilitate the fitting procedure of the surrogate model, we standardise the outputs and apply an affine transformation to the inputs so that the search domain lies in $[0,1]^d$. At the beginning of each experiment, three points are drawn uniformly at random within the search domain to initialise the surrogate model.
    
    Additionally, in order to provide some indication as to how the GP-based surrogate model schemes, endowed with compositional optimisation of the acquisition function, perform against other surrogates, we also compare against the BOHB algorithm \citep{2018_Falkner}, a hybrid approach based on Bayesian optimisation and the Hyperband algorithm \citep{2017_Li}. BOHB has recently been demonstrated to outperform Bayesian optimisation across a range of problems in the multi-fidelity setting, that is where multiple objective functions exist possessing varying degrees of accuracy and cost associated with querying them \citep{2019_Song}. In order to enable comparison in the single-fidelity contexts considered in our experiments, we simply ignore the budget handling from Hyperband.
    
    \paragraph{Acquisition Functions: }
    
    We consider the batched versions of each acquisition function presented in Section \ref{Sec:AcqFun}, namely EI, PI, SR and UCB under ERM, FSM and compositional forms. Additionally, we employ Thompson sampling \citep{1933_Thompson} as a baseline in order to provide an indication as to how the compositionally-optimised acquisition functions perform against another popular batch acquisition function.
    
    \paragraph{Optimisers:} Acquisition function maximisation is carried out using the zero-order optimisers RS, CMA-ES and DE from the \textsc{pymoo} library \citep{pymoo}, the non-compositional first-order optimisers Adadelta, Adagrad, Adam, AdamW, RMSprop, Rprop and  SGA taken from \textsc{PyTorch} \citep{Paszke2019pytorch}, the second-order optimiser L-BFGS-B from the \textsc{SciPy} library \citep{virtanen2020scipy}, as well as the compositional optimisers ASCGA, CAdam, MC-Nested, NASA and SCGA that we implemented on top of the \textsc{BoTorch} library \citep{balandat2019botorch}. Except when using non-memory-efficient compositional methods, we used quasi-MC normal Sobol sequences \citep{owen2003quasi-monte-carlo} instead of \textit{i.i.d.} normal samples in order to obtain lower variance estimates of the value and gradient of the acquisition function as recommended by \cite{balandat2019botorch}. For the L-BFGS-B optimiser, the minibatch of samples was fixed in all cases. To ensure fairness in performance comparison, the same number of optimisation steps $T$ (set to $64$) and minibatch size $m$ (set to $128$), is used for each method at each acquisition step. As acquisition function maximisation is a non-convex problem, it is sensitive to the initialisation set. As such, we use multiple restart points \citep{wang2019parallel} that we first obtain by drawing $1024$ batches uniformly at random in the modified search space $[0,1]^{q \times d}$, and second using the default heuristic from \cite{balandat2019botorch} to select only $32$ promising initialisation batches. Consequently, at each inner optimisation step of BO, the Random Search optimisation strategy is granted $32 \times T \times m$ evaluations of the acquisition function at random batches. Similarly, CMA-ES and DE are run for $64$ evolution and mutation steps, and the aforementioned initialisation strategy is used to generate the $32$ members of the initial population.   
    
    It is known that first-order stochastic optimisers can be very sensitive to the choice of hyperparameter settings \citep{balandat2019botorch, schmidt2020descendingcrowdedvalley}. Therefore, to limit the effect of choice of hyperparameter settings for the different optimisers, we conducted each experiment in two phases. An experiment in this instance is characterised by the 3-tuple consisting of a black-box function, an acquisition function and an optimiser. 
    
    In the first phase, we ran BO hyperparameter tuning to identify the best optimiser hyperparameters, in the sense that these hyperparameters provide the lowest final regret for the given task. This first phase allows us to compare optimisers in their most favourable settings, and therefore we hope that under-performance cannot be the result of a poor choice of hyperparameters but would reflect a real weakness of the considered method in tackling BO's inner optimisation problem.
    
    In the second phase, we ran the black-box maximisation task using the acquisition function and optimiser with hyperparameters fixed to be the best ones identified during the first phase. The set and range of the considered hyperparameters are summarised in Table~\ref{tab:non-compositional-tuning} for non-compositional optimisers, and in Table~\ref{tab:compositional-tuning} for compositional optimisers.

    \subsection{FSM vs. ERM}\label{sec:syntheticbo}
    
    \begin{figure}[t!]
    \centering
      \centering
      \includegraphics[width=.75\linewidth]{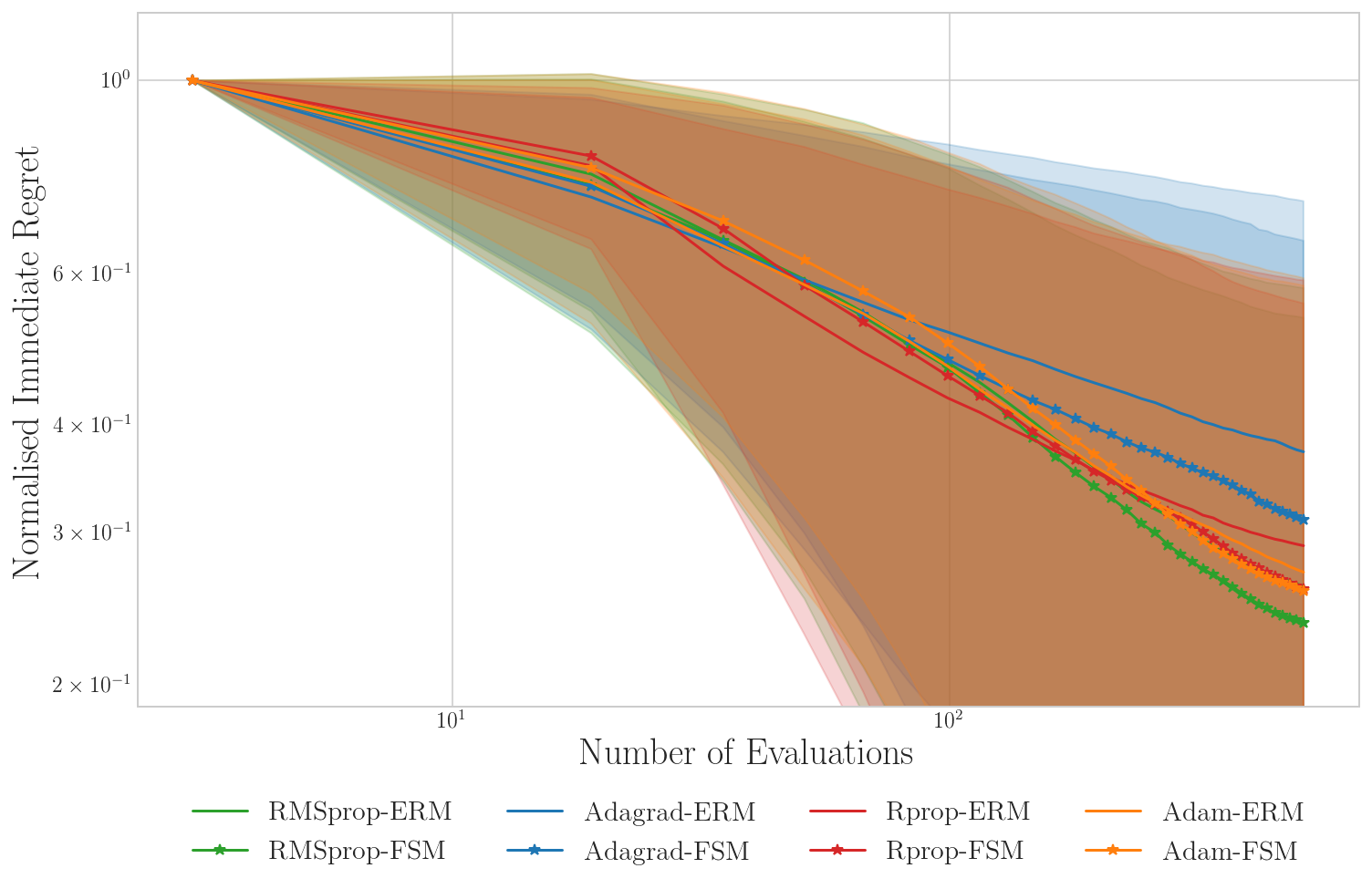}
      \caption{Summary plot comparing the evolution of the normalised immediate regret averaged over all tasks when using first-order methods with either the ERM or FSM formulation of the acquisition function. The results of $960$ experiments are summarised. We observe a small advantage of the FSM formulation over the ERM formulation across every optimiser.}
      \label{fig:erm-v-fsm}
    \end{figure}
    
    In the following experiment, we consider five non-separable, non-convex, synthetic black-box functions chosen to have a variety of optimisation landscapes and that are commonly-used benchmarks for optimisation algorithms \citep{Jamil2013survey-benchmarks, Laguna2005opt-benchmark-levy}. We include the unimodal functions \textit{Dixon-Price} and \textit{Powell} as well as the multimodal \textit{Levy}, \textit{Ackley} and \textit{Styblinski-Tang} functions. We run experiments for (negative) versions of these functions with search domain specified as in \citep{Jamil2013survey-benchmarks, Laguna2005opt-benchmark-levy}. We consider optimisation problems across dimensionalities in the set ($16$D, $40$D, $60$D, $80$D, $100$D and $120$D) in order to observe the impact of the input space dimension on the optimisers' performance. At each acquisition step, a batch of $q=16$ points is acquired as a result of batch acquisition function maximisation. We run each BO algorithm with $32$ acquisition steps and observe the normalised immediate regret from Equation \ref{imm_regret} as the performance metric.
    
    \paragraph{Results Summary} 
    
    Figure \ref{fig:erm-v-fsm} aggregates by optimiser category, (zero-order non-compositional, first-order compositional,$\ldots$), the results of $960$ experiments involving each combination of optimisation task, acquisition function and optimiser. The best performances obtained inside each category are accounted for. Specifically, given a category and an acquisition step, the lowest normalised immediate regrets obtained at this step by an optimiser belonging to this category are included and the average and standard deviation obtained over all optimisation tasks and all acquisition functions, are reported. 
    
    \noindent In light of these results we will now answer Question 1: 
    
    \begin{mybox}{brown}{Question 1}

    ``Do Finite-Sum Minimisation acquisition functions provide any benefits compared to the more frequently-used Empirical Risk Minimisation versions?''.
    
    \end{mybox}

    
   When looking at the top four first-order non-compositional optimisers, Figure~\ref{fig:erm-v-fsm} shows in all cases that the FSM version outperforms the ERM version when averaging the normalised immediate regret scores over all optimisation tasks and acquisition functions. This can be seen in an un-aggregated breakdown in both Figure~\ref{fig:immediate_log_regret_120D} and Figure~\ref{fig:summary-synthetic-acqfunc-inputdim}. This is an interesting discovery, and to the best of our knowledge, we are the first to observe this. We now proceed to our second question.
    
    \begin{table}
        \centering
       \resizebox{\textwidth}{!}{
    
    \begin{tabularx}{1.2\textwidth}{llllXXXXXXXXXXXXXX}
    \toprule
      &   &  & Dim. & \multicolumn{2}{l}{16} & \multicolumn{2}{l}{40} & \multicolumn{2}{l}{60} & \multicolumn{2}{l}{80} & \multicolumn{2}{l}{100} & \multicolumn{2}{l}{120} & \multicolumn{2}{l}{\textbf{Tot.}} \\
      &   &  & {} & \rot[-90]{\#Best (\%)} & \rot[-90]{NFR} & \rot[-90]{\#Best (\%)} & \rot[-90]{NFR} & \rot[-90]{\#Best (\%)} & \rot[-90]{NFR} & \rot[-90]{\#Best (\%)} & \rot[-90]{NFR} & \rot[-90]{\#Best (\%)} & \rot[-90]{NFR} & \rot[-90]{\#Best (\%)} & \rot[-90]{NFR} & \rot[-90]{\#Best (\%)} & \rot[-90]{NFR} \\
    {} & Order & Optimiser & Ref. &    &  &    &  &    &  &    &  &    &  &    &  &    &  \\
    \midrule
    NonComp & 0 &  RS & App.~\ref{App:RS} &  0 & .33 &  0 & .51 &  0 & .60 &  0 & .64 &  0 & .68 &  0 & .75 &  0 & .59 \\
      &   & CMA-ES & App.~\ref{App:CMAES} &  0 & .30 &  0 & .49 &  0 & .76 &  0 & .80 &  0 & .81 &  0 & .85 &  0 & .67 \\
      &   & DE & App.~\ref{App:DE} &  0 & .29 &  0 & .45 &  0 & .61 &  0 & .66 &  0 & .66 &  0 & .70 &  0 & .56 \\
    \noalign{\smallskip}\noalign{\smallskip}\cline{3-18}\noalign{\smallskip}
      &   & \textbf{Subtot.} & &  0 & .31 &  0 & .48 &  0 & .66 &  0 & .70 &  0 & .72 &  0 & .77 &  0 & .61 \\
    \noalign{\smallskip}\cline{2-18}\noalign{\smallskip}
    &   1   & SGA & App.~\ref{App:SGA} &  0 & .18 &  0 & .28 &  0 & .33 &  0 & .42 &  0 & .35 &  0 & .48 &  0 & .34 \\
     &   & Adagrad & App.~\ref{App:AdaGrad} &  5 & .36 &  5 & .55 &  5 & .66 &  5 & .75 &  5 & .87 & 10 & .89 &  6 & .68 \\
        &   & RMSprop & App.~\ref{App:RMSProp} & 10 & .29 &  5 & .45 & 15 & .47 &  0 & .58 &  0 & .53 & \textcolor{red}{\textbf{15}} & .64 &  8 & .49 \\
      
      &   & Adam & App.~\ref{App:ADAM} &  5 & .35 & 15 & .46 &  5 & .51 &  5 & .53 & \textcolor{red}{\textbf{20}} & .61 & 10 & .70 & 10 & .52 \\
      &  & Adadelta & App.~\ref{App:AdaDelta} &  0 & .20 &  0 & .44 &  5 & .32 &  0 & .46 &  0 & .45 &  0 & .48 &  1 & .39 \\
       &   & Rprop & App.~\ref{App:Rprop} &  0 & .36 &  0 & .49 & 10 & .57 &  5 & .61 &  0 & .59 & 10 & .66 &  4 & .55 \\
      
      &   & AdamW & App.~\ref{App:AdamW} &  0 & .18 &  0 & .24 &  5 & .22 &  5 & .22 &  5 & .25 &  5 & .23 &  3 & .22 \\
      &   & Adamos & App.~\ref{App:AdamOS} &  0 & .17 &  0 & .26 &  0 & .26 &  0 & .28 &  5 & .30 &  5 & .34 &  2 & .27 \\
    \noalign{\smallskip}\noalign{\smallskip}\cline{3-18}\noalign{\smallskip}
      &   & \textbf{Subtot.} & & 20 & .26 & 25 & .40 & 45 & .42 & 20 & .48 & 35 & .49 & \textcolor{red}{\textbf{55}} & .55 & 33 & .43 \\
    \noalign{\smallskip}\cline{2-18}\noalign{\smallskip}
      & 2 & L-BFGS-B & App.~\ref{App:SecondOrder} &  0 & .19 &  0 & .29 &  0 & .39 &  0 & .45 &  0 & .45 &  0 & .51 &  0 & .38 \\
    \noalign{\smallskip}\noalign{\smallskip}\cline{3-18}\noalign{\smallskip}
      &   & \textbf{Subtot.} & &  0 & .19 &  0 & .29 &  0 & .39 &  0 & .45 &  0 & .45 &  0 & .51 &  0 & .38 \\
    \noalign{\smallskip}\cline{2-18}\noalign{\smallskip}
      & \textbf{Tot.} &  & & 20 & .27 & 25 & .41 & 45 & .48 & 20 & .53 & 35 & .55 & \textcolor{red}{55} & .60 & 33 & .47 \\
    \noalign{\smallskip}\cline{1-18}\noalign{\smallskip}
    Comp & 0 & CMA-ES & App.~\ref{App:CMAES} &  0 & .30 &  0 & .49 &  0 & .76 &  0 & .82 &  0 & .83 &  0 & .87 &  0 & .68 \\
      &   & DE & App.~\ref{App:DE} &  0 & .30 &  0 & .46 &  0 & .61 &  0 & .64 &  0 & .67 &  0 & .71 &  0 & .57 \\
    \noalign{\smallskip}\noalign{\smallskip}\cline{3-18}\noalign{\smallskip}
      &   & \textbf{Subtot.} & &  0 & .30 &  0 & .47 &  0 & .69 &  0 & .73 &  0 & .75 &  0 & .79 &  0 & .62 \\
    \noalign{\smallskip}\cline{2-18}\noalign{\smallskip}
     &    1  & SCGA & App.~\ref{App:SCGA} & 10 & .12 &  0 & .18 &  0 & .33 &  0 & .44 &  0 & .52 &  0 & .62 &  2 & .37 \\
      &  & ASCGA & App.~\ref{App:ASCGA} &  5 & .11 &  5 & .17 &  0 & .34 &  0 & .48 &  0 & .53 &  0 & .60 &  2 & .37 \\
      &   & CAdam & App.~\ref{App:CADAM} & 20 & .09 & 25 & .12 & \textcolor{red}{\textbf{35}} & .19 & \textcolor{red}{\textbf{25}} & .14 & \textcolor{red}{\textbf{20}} & .14 & 10 & .22 & \textcolor{red}{\textbf{22}} & .15 \\
       &   & NASA & App.~\ref{App:NASA} & \textcolor{red}{\textbf{45}} & .08 & \textcolor{red}{\textbf{35}} & .21 & 15 & .31 & 20 & .39 & 10 & .40 &  5 & .55 & \textcolor{red}{\textbf{22}} & .32 \\
      &   & Nested-MC & App.~\ref{App:NestedMC}  &  0 & .17 & 10 & .22 &  5 & .23 &  5 & .26 &  5 & .29 &  0 & .38 &  4 & .26 \\
      &   & CAdam-ME & App.~\ref{App:CADAMME} &  - &   - &  - &   - &  - &   - & 20 & .14 & 15 & .16 & 20 & .24 & 18 & .18 \\
        
      &   & NASA-ME & App.~\ref{App:NASAME} &  - &   - &  - &   - &  - &   - & 10 & .35 & 10 & .40 &  5 & .52 &  8 & .43 \\
      
      &   & Nested-MC-ME & App.~\ref{App:NestedMCME}  &  - &   - &  - &   - &  - &   - &  0 & .28 &  5 & .29 &  5 & .32 &  3 & .29 \\
     
    \noalign{\smallskip}\noalign{\smallskip}\cline{3-18}\noalign{\smallskip}
      &   & \textbf{Subtot.} & & \textcolor{red}{\textbf{80}} & .12 & \textcolor{red}{\textbf{75}} & .18 & \textcolor{red}{\textbf{55}} & .28 & \textcolor{red}{\textbf{80}} & .31 & \textcolor{red}{\textbf{65}} & .34 & 45 & .43 & \textcolor{red}{\textbf{67}} & .28 \\
    \noalign{\smallskip}\cline{2-18}\noalign{\smallskip}
      & 2 & CL-BFGS-B & Sec.~\ref{sec:FSM-BO} &  0 & .20 &  0 & .28 &  0 & .34 &  0 & .36 &  0 & .44 &  0 & .50 &  0 & .35 \\
    \noalign{\smallskip}\noalign{\smallskip}\cline{3-18}\noalign{\smallskip}
      &   & \textbf{Subtot.} & &  0 & .20 &  0 & .28 &  0 & .34 &  0 & .36 &  0 & .44 &  0 & .50 &  0 & .35 \\
    \noalign{\smallskip}\cline{2-18}\noalign{\smallskip}
      & \textbf{Tot.} &  & & \textcolor{red}{\textbf{80}} & .17 & \textcolor{red}{\textbf{75}} & .27 & \textcolor{red}{\textbf{55}} & .39 & \textcolor{red}{\textbf{80}} & .39 & \textcolor{red}{\textbf{65}} & .43 & 45 & .50 & \textcolor{red}{\textbf{67}} & .36 \\
    \bottomrule
    \end{tabularx}
    }
    \caption{Marginal results over acquisition functions and synthetic black-box optimisation tasks (i.e., $20$ tasks per dimension). For each dimension, the first column, \#Best (\%), indicates the percentage of tasks on which an optimiser yielded the lowest final regret, while the second column reports normalised final regret (NFR). ERM and FSM versions of first-order, non-compositional optimisers are grouped together. We mark the best percentage across each dimension (dim.) in red. Clearly, in 16, 40, 60, 80 and 100 dims, compositional solvers achieve the best performance in at least 55 \% these tasks, with 80 \% in 16 and 80 dims, while first-order non-compositional optimisers outperform others 55 \% of the time in 120 dimensions.}
        \label{tab:summary_perf}
    \end{table}
    
    \subsection{Compositional vs. Non-Compositional Optimisation}
    
    \begin{figure}[t!]
      \centering
      \includegraphics[width=1\linewidth]{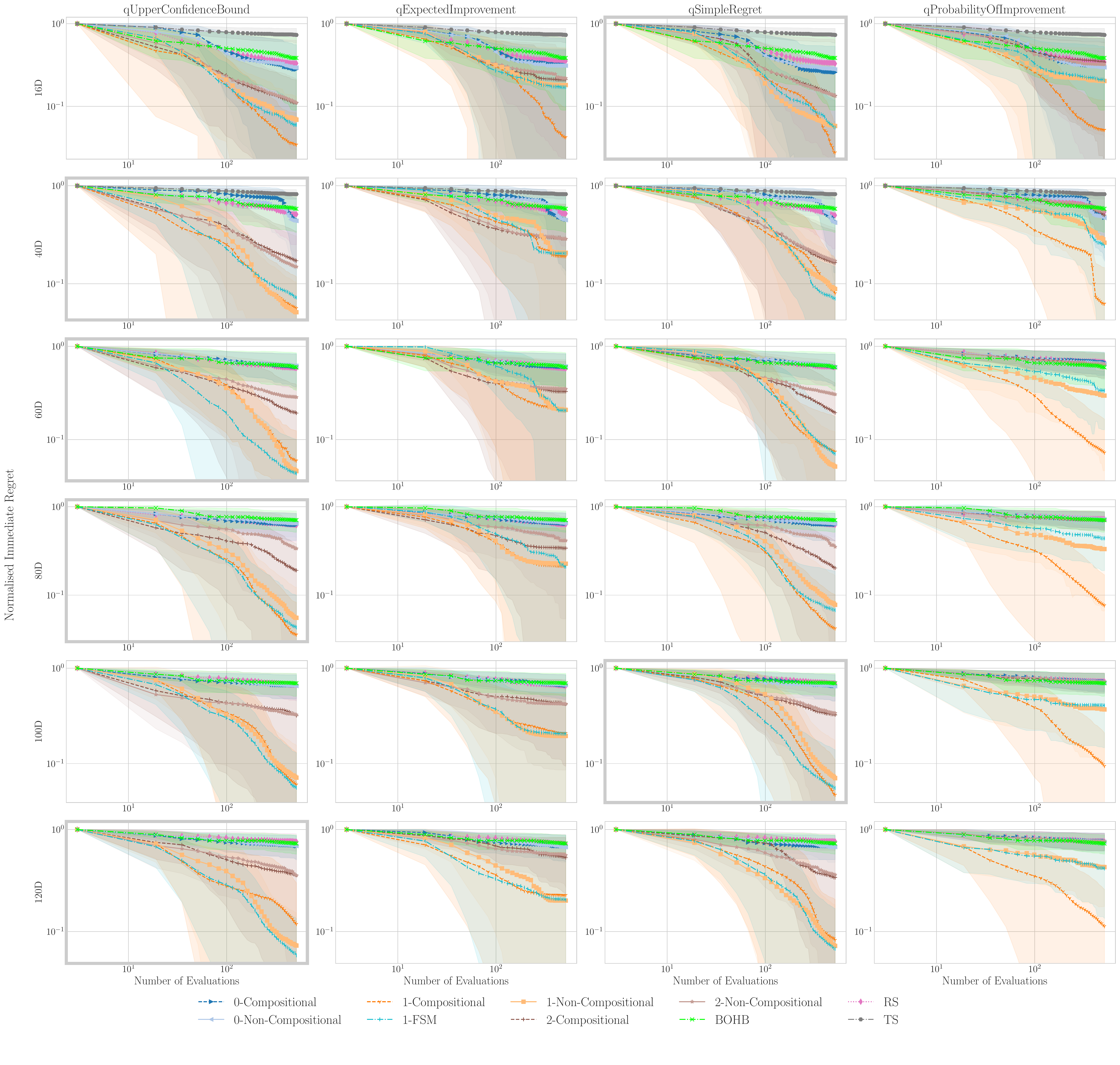}
      \caption{Detailed summary plot for synthetic black-box optimisation showing the best relative improvement for each category of optimiser. Each row corresponds to a domain dimension ($16$D, $40$D, $60$D, $80$D, $100$D and $120$D) and each column is associated with an acquisition function (EI, PI, SR and UCB). Relative improvements yielded by BOHB and TS are also reported (there is no variation across columns as they do not depend on the acquisition function), leading the number of experiments aggregated on this figure to be $3100$. On each row, the graph corresponding to the acquisition function that achieved the lowest regret for the given input dimension has a thick grey border. In $40$, $60$, $80$ and $120$ dimensions, the best performance is achieved using UCB with a first-order optimiser, while in $16$ and $100$ dims, it is SR with a first-order compositional optimiser that led to the largest relative improvement.}
    \label{fig:summary-synthetic-acqfunc-inputdim}
    \end{figure}

    To synthesise the results obtained over all combinations of synthetic function ({Levy}, {Ackley}, {Powell}, {Dixon-Price}, {Styblinski-Tang}), input dimensionality ($16$D, $40$D, $60$D, $80$D, $100$D and $120$D), and acquisition function (EI, PI, SR, UCB), we show in Figure~\ref{fig:summary-synthetic} the evolution of the normalised immediate regret for each category of optimiser. We confirm the observation of \cite{2018_Wilson} that gradient-based approaches outperform zero-order methods. Evolutionary strategies perform comparably to Random Search (which we exclude from its category as a global baseline). The poor performance of zero-order methods can be explained by the dimensionality of the acquisition function domain, ranging from $16 \times 16$ to $16 \times 120$ and the strict limitation on the number of optimisation steps. Results obtained with BOHB are also similar to Random Search, although it is worth mentioning that the experimental setting is single-fidelity and not multi-fidelity where BOHB has been observed to perform well. The performance of Thompson sampling (TS) coincides with the observation in the literature that TS has difficulty scaling beyond 8-10 dimensions \citep{2020_Wilson}. We run GPflow \citep{2017_gpflow} implementations of function-space, weight-space and decoupled TS with the default hyperparameters from \cite{2020_Wilson}. We report these results in our summary plots and note that scaling such information-based acquisition functions constitutes an important direction for future work, see Section~\ref{Sec:Conc}. 
    
    On examining gradient-based methods, we observe that quasi-Newton (C)L-BFGS-B is consistently outperformed by first-order methods, which was not observed in \citep{balandat2019botorch} where only a small-dimensional experiment with no batch acquisition (i.e. $q=1$) was presented. From this global summary, our results favour first-order optimisers, with a relative advantage being given to compositional methods associated with the FSM approximation. On the other hand, non-compositional optimisers do not seem to be amenable to ERM or FSM formulation.
    
    \begin{figure}[t!]
      \centering
      \includegraphics[width=1\linewidth]{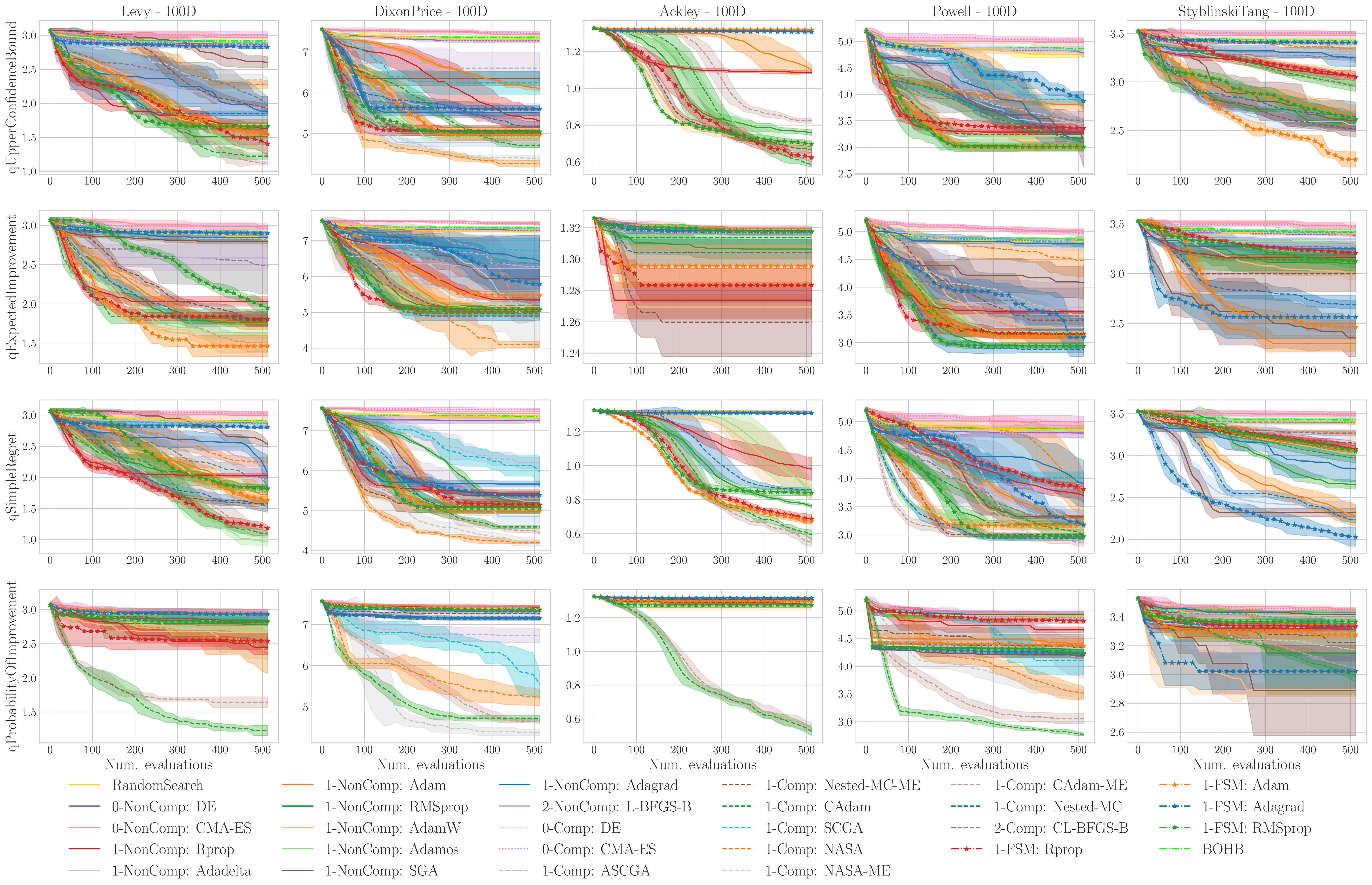}
      \caption{Evolution of immediate log-regret using various acquisition functions and optimisers on $100$D synthetic black-box minimisation tasks. Each row is associated with an acquisition function, and each column corresponds to a black-box function. A total of $545$ experiments have been carried out on $100$ tasks to run BO with all combinations of optimisers, acquisition functions and black-box optimisation functions. We observe that the lowest regret is always achieved with a first-order optimiser, and notably $25$\% of the best performances are obtained when using CAdam.}
      \label{fig:immediate_log_regret_100D}
    \end{figure}

    To show a breakdown of all experiments, we present in Figure~\ref{fig:summary-synthetic-acqfunc-inputdim} the best performances yielded by each category of optimiser for each input dimensionality and acquisition function considered. From this figure, we can first observe that the dimensionality of the BO problem does not seem to have a significant impact on the relative performances between the different types of methods, that is, for any dimension, the best first-order gradient method outperforms the second-order methods, which achieve lower regret than zero-order ones. Aside from this trend at the level of the optimiser order, we do not notice any lower-level trend that may be driven by the input dimensionality.

    An example of the most fine-grained level of analysis (all optimiser performances presented individually) is given in Figure~\ref{fig:immediate_log_regret_100D}. For each task-acquisition pair, we show the log regret over acquisition steps for each optimisation method introduced. We can see that in $65$\% of the experiments that a compositional optimiser outperforms all non-compositional optimisers. As shown in Table~\ref{tab:summary_perf}, the superior performance of compositional optimisers is observed across all task input dimensionalities except for $120$D for which the best optimiser is compositional in only $45$\% of cases.

    Moreover, Figure~\ref{fig:summary-synthetic-acqfunc-inputdim} provides some insight into the comparatively better performance of first-order compositional optimisers observed in the global summary Figure~\ref{fig:summary-synthetic}. Lower regrets are obtained when the PI acquisition function is used. Nevertheless, the shading of the graphs corresponding to the best acquisition function for each dimensionality indicates that PI yields consistently higher regrets than UCB or SR, which encourages the use of these alternative acquisition functions in place of PI with a first-order compositional optimiser.

    \noindent Returning to our second question:
    
    \begin{mybox}{brown}{Question 2}
    Do compositional optimisers provide any advantages over non-compositional optimisers?
    \end{mybox}
    
    The global summary Figure~\ref{fig:summary-synthetic} in addition to Figure~\ref{fig:summary-synthetic-acqfunc-inputdim} indicate that there are a significant number of optimisation task and acquisition function pairs where a compositional optimiser is preferable and as such, compositional schemes warrant much more attention than they are currently receiving in the Bayesian optimisation community. We will now proceed to answer our third question.

\subsection{Memory Efficiency}
    
\begin{figure}[t]
    \centering
    \subfloat{\includegraphics[width=0.45\textwidth]{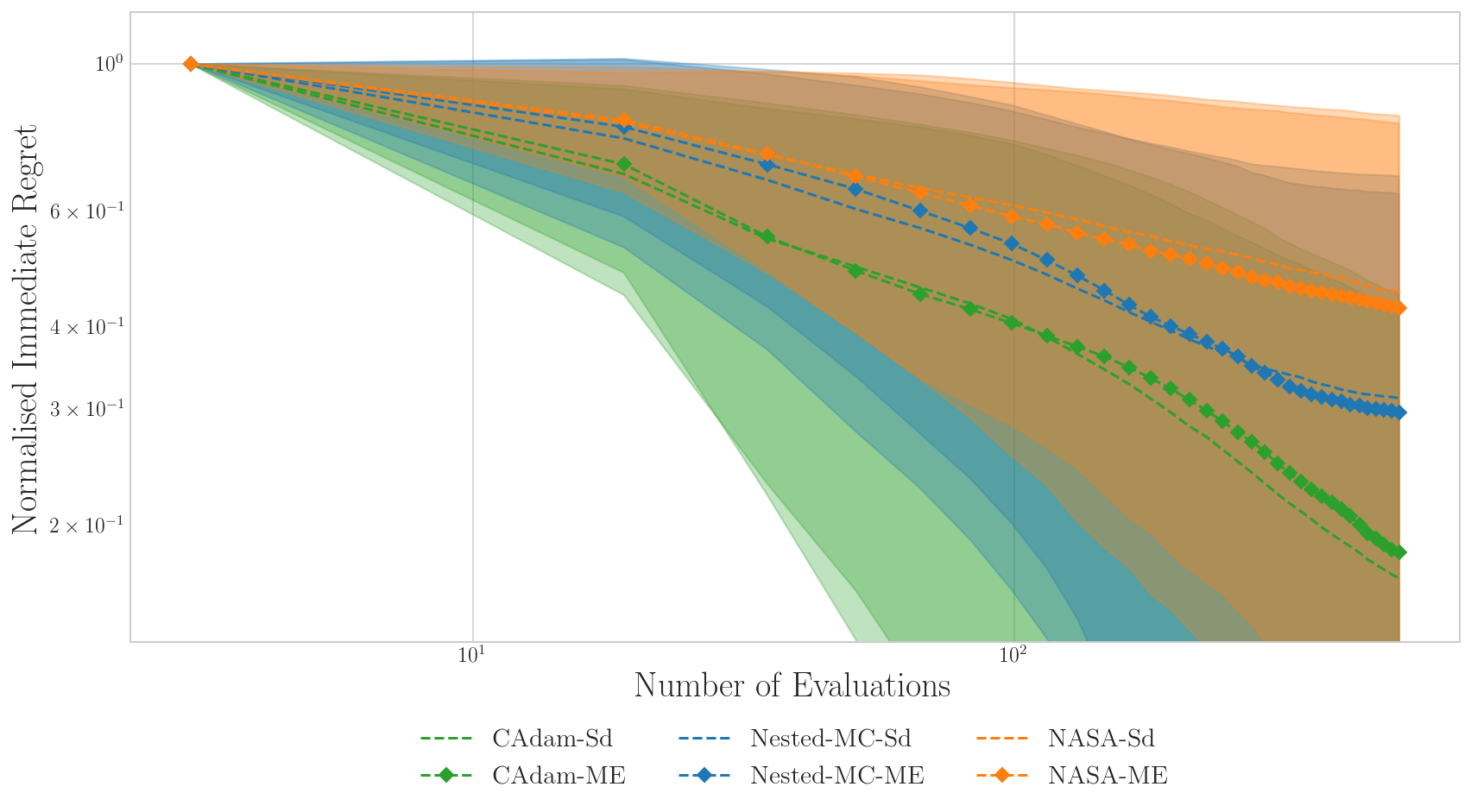}\label{fig:sd-v-me}}
    \hfill
    \subfloat{\includegraphics[width=0.45\textwidth]{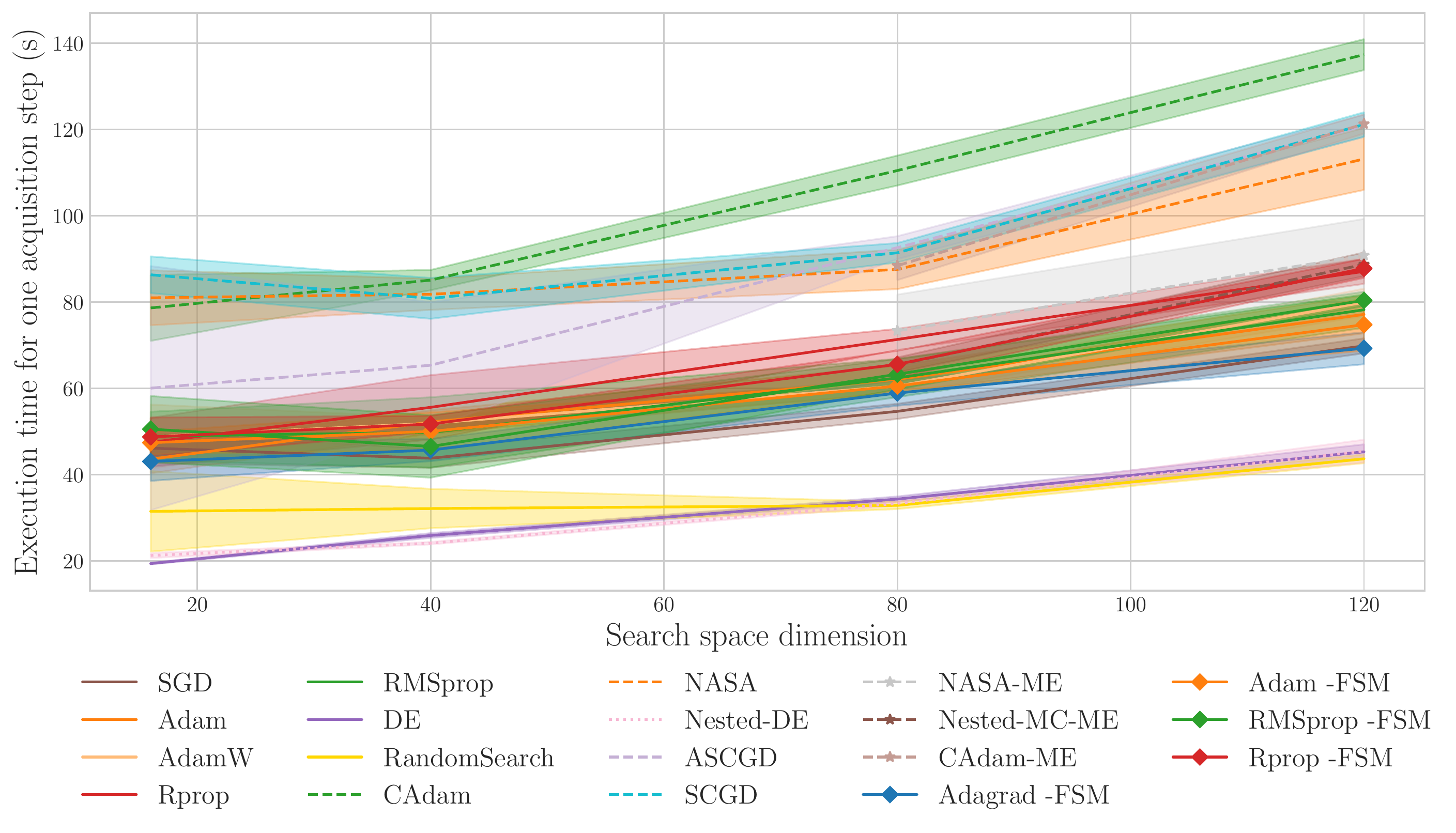}\label{fig:ex-time}}
    \caption{(\ref{fig:sd-v-me}) Summary plot comparing the evolution of the normalised immediate regret averaged over all considered acquisition functions, and optimisation tasks in $80$, $100$ and $120$ dimensions, when using standard (Sd) and memory-efficient (ME) compositional first-order optimisers. From this figure, aggregating the results of $360$ experiments, we can see that memory-efficient optimiser versions perform comparably to standard optimisers, thus making it worthwhile to use memory-efficient implementations due to the large memory savings. (\ref{fig:ex-time}) Execution time of UCB maximisation run on $4$ CPUs. We report the time it takes an optimiser to carry out a single UCB maximisation, and we show the mean and standard deviation observed over $5$ seeds, $32$ acquisition steps and $2$ synthetic black-box functions in $16$, $40$, $80$ and $120$ dims. From this figure, aggregating results of $152$ experiments, we observe that compositional methods take about 1.5-2x the CPU time taken by non-compositional methods. We do not report the execution times measured for (C)L-BFGS-B and CMA-ES as they are an order of magnitude greater than those observed for non-compositional, first-order methods. We provide complementary results in Figure \ref{fig:ex-time-full} in Appendix G.}
    \label{fig:mem_time}
\end{figure}

    
    Compositional acquisition function maximisation requires considerably larger memory relative to ERM. However, by introducing a simple trick whereby we do not store all the auxiliary variables and adopt an alternative sampling scheme, we can dramatically reduce the memory requirements to be equivalent to those of ERM. In answer to question 3:
    
    \begin{mybox}{brown}{Question 3}
    What are the practical savings for using memory-efficient implementations of compositional acquisition functions? 
    \end{mybox}

    Figure~\ref{fig:sd-v-me}, which aggregates results obtained on tasks in $80$, $100$ and $120$ dimensions using both memory-efficient and standard versions of CAdam, NASA and Nested-MC to maximise the acquisition function, shows that CAdam is negatively impacted by the ME implementation, whereas NASA and Nested-MC are positively impacted by memory efficiency. In all cases, the impact on going from standard to memory-efficient implementations is minor enough that we believe it warrants the use of the ME implementation as the de facto standard. We now proceed to answer Question 4:

    \subsection{Runtime Efficiency}
    
    
    Runtime efficiency is of great importance for many applications. As such, we wish to see how the execution time required for a single acquisition function optimisation varies across compositional optimisers and input dimensionality. We fix the acquisition function to UCB as this choice has negligible effect on overall timings and we run the BO algorithm for $32$ acquisition steps on two black-box maximisation tasks using all available optimisers, repeating each experiment five times. In answer to Question 4:
    
        \begin{mybox}{brown}{Question 4}
Are compositional methods more computationally expensive than non-compositional optimisation methods and how does runtime scale as a function of the input dimensionality?
    \end{mybox}

  There is a marked difference between the execution times reported in Figure~\ref{fig:ex-time} for compositional and non-compositional methods with compositional methods being slower relative to non-compositional. Additionally, ME methods are faster than standard compositional methods. We can also see that as the input dimensionality increases, a steeper incline in the execution time for compositional methods relative to non-compositional methods may be observed; a feature to be expected given the extra backward passes required by compositional optimisers. Due to these additional backward passes, compositional methods are $1.5$-$2$ times slower per iteration in terms of wall-clock time. This being said, it should be noted that compositional optimisers may require fewer iterations in total to converge to a specified accuracy and in this case overall wall-clock time could be comparatively better for them. Finally, if the black-box system evaluation wall-clock time is factors larger than the optimisation wall-clock time, which is the case in many real-world problems such as molecule synthesis where a single query can take 2-3 weeks \citep{2020_Thawani}, then the differences in runtime between compositional and non-compositional schemes becomes negligible. We now proceed to answer our final question.

    \subsection{Real-World Problems: Noisy Evaluations}
    
    We now examine the performance of optimisers on Bayesmark tasks. All tasks involve hyperparameter tuning for machine learning models. In contrast to the synthetic functions, the Bayesmark datasets possess noise in the evaluations of the black-box function, a feature inherent in the vast majority of real-world BBO problems. As such, these experiments

    \paragraph{Hyperparameter Tuning Tasks:} The Bayesmark tasks consist of both regression and classification tasks on the Boston and Diabetes UCI datasets \citep{2019_Dua} respectively. In terms of hyperparameter tuning the following six models are considered: Decision Tree (DT), Random Forest (RF), K-Nearest Neighbours (kNN), Support Vector Machine (SVM), Linear and Lasso models. the dimensionality of each task varies from $2$ to $9$. In contrast to the synthetic functions, we only have access to noisy evaluation of the black-box functions in this instance. We apply Bayesian optimisation using $16$ iterations of $8$-batch acquisition steps, to optimise the validation loss, mean-squared error (MSE), mean absolute error (MAE), negative log likelihood (NLL) or accuracy depending on the task, plotting the normalised validation loss score (Eq~\ref{eq:score}) for performance comparison. We ran all six models on regression tasks (both MAE and MSE objectives) and we run three models (DT, RF and SVM) on classification tasks (both NLL and accuracy objectives) due to a limited computation budget. The score achieved after $t$ acquisition steps is given by:
    \begin{equation}
    \label{eq:score}
        \textbf{score}_t = \frac{\mathcal{L}_t- \mathcal{L}^*}{\mathcal{L}^{\text{rand}}_t - \mathcal{L}^*}
    \end{equation}
    where $\mathcal{L}_t$ is the best-achieved loss at batch $t$. $\mathcal{L}^*$ is the estimated optimal loss for the task and $\mathcal{L}^{\text{rand}}$ is the mean loss (across multiple runs) acquired from random search at batch $t$.

    \paragraph{Optimisers:} The top three non-compositional optimisers (Adam, RMSprop, Rprop) were selected for performance comparison against compositional optimisers (NASA, CAdam, Adam-Nested). 
    
    \paragraph{Acquisition Functions:} We show results for the four top-performing acquisition functions (SR, EI, PI and UCB) from the synthetic function experiments.
    
    \paragraph{Surrogate Model:} We use the same GP surrogate model as in Sec~\ref{sec:syntheticbo}, with rounding of integer values when either integer or categorical variables are present. Although more sophisticated methods exist to deal with categorical/integer variables \citep{ru2019bayesian, 2020_Daxberger, garrido2020dealing} we do not consider them here as we are interested in solely in performance on acquisition function maximisation. We sample $2 \times D$ points uniformly at random to initialise the model. We run the same form of hyperparameter tuning for the initialisation as in the synthetic experiments, repeating each experiment 5 times in order to compute the variance for individual tasks. 
    
    \begin{figure}[t!]
        \centering
        \includegraphics[width=\textwidth]{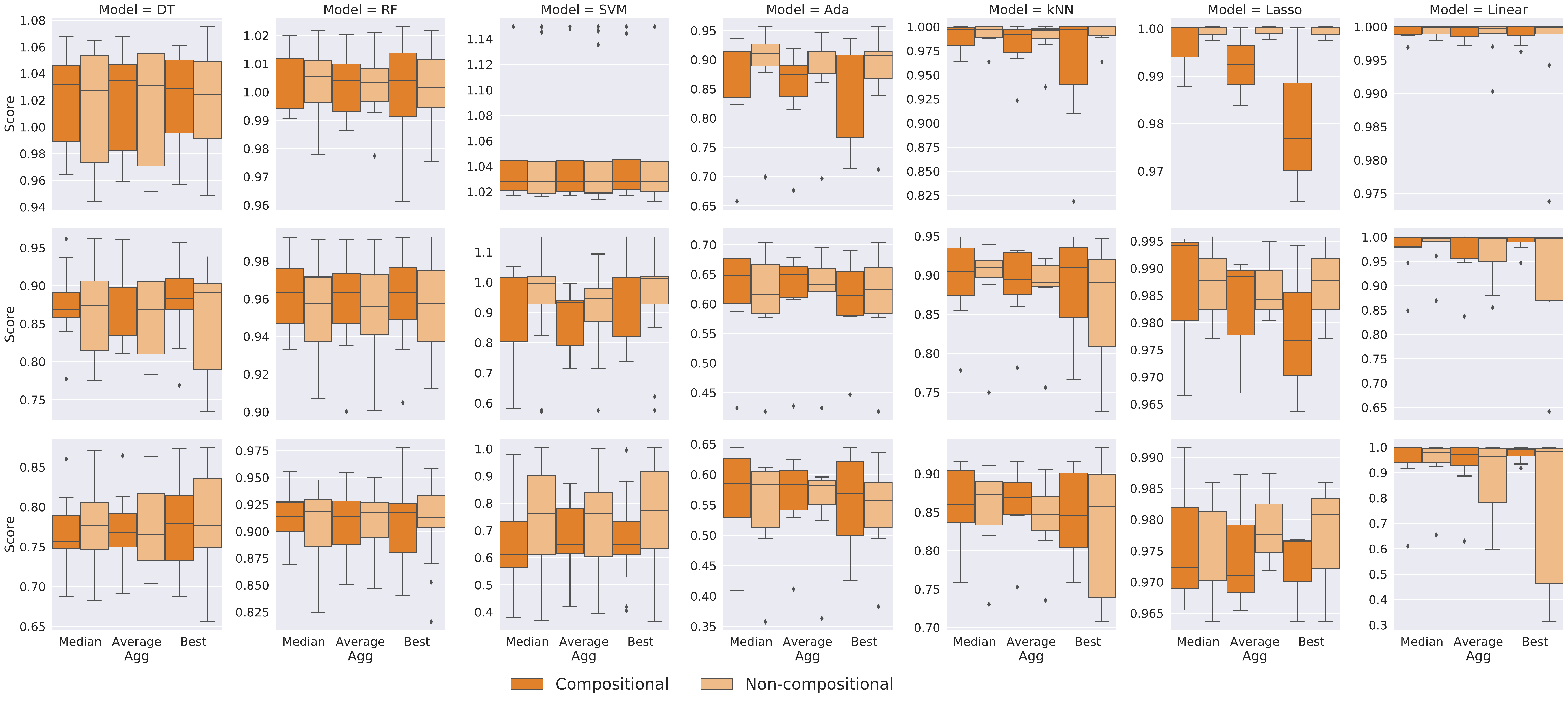}
        \caption{The boxplot shows the quartiles of compositional and non-compositional optimiser performance on the regression hyperparameter tuning task, where the performance metrics are MAE and MSE. For each model, we show a further split of the optimiser class for different aggregation methods. This plot summarises all 672 experiments conducted on regression tasks on the Bayesmark dataset. We observe performance benefits for DT, RF and AdaBoost when using a compositional optimiser, with SVM and kNN showing performance benefits when using a non-compositional optimiser.}
        \label{fig:reg_details}
    \end{figure}

    \paragraph{Results Summary:} 
    
    \noindent In answer to our final question:
    
    \begin{mybox}{brown}{Question 5}
      How do compositional optimisers perform when optimising real-world black-box functions with noisy evaluations?
    \end{mybox}
    
    Figure~\ref{fig:summary_competition_reg} shows a high-level breakdown of compositional and non-compositional optimiser performance on the Bayesmark regression tasks. The best final scores for the model undergoing tuning are pooled across optimisers, tasks, loss functions and acquisition functions. We observe that compositional and non-compositional optimisers perform comparably, with compositional methods performing slightly better for DT, RF and SVM. We see that the mean scores are roughly equivalent for optimiser classes across the kNN, Lasso, linear and AdaBoost models. In an analogous fashion, Figure~\ref{fig:summary_competition_class} pools the scores for all classification experiments. For the DT, and RF models, compositional methods achieve higher mean scores wheraeas comparable performance is observed when tuning the SVM model. In conclusion, compositional vs. non-compositional optimiser performance appears to vary depending on both the model class undergoing tuning as well as the performance metric.
    
    
    \paragraph{Detailed Results}: Figure~\ref{fig:reg_details} depicts a finer-grained breakdown of the pooled results for the Bayesmark regression tasks. Pooling in this case is carried out using the best, median and average optimiser performances across all intra-class optimisers and acquisition functions, where for example the best compositional optimiser for a given model would be the top-scoring optimiser-acquisition pair. For DT and RF, the best results are produced from compositional optimisers, whereas for SVM, AdaBoost, kNN and the linear model, non-compositional methods exhibit better performance. For compositional optimisation of the Lasso model we observe better median performance for a higher number of black-box function evaluations, but deteriorating performance under the best grouping. Figure~\ref{fig:class_details} similarly shows a finer-grained breakdown of the Bayesmark classification tasks. We observe that for certain models, such as RF, compositional methods perform better in each of best, median and average groupings at all steps in the optimisation, namely 8, 16 and 128 evaluations of the black-box system. In the DT experiments we again observe that compositional optimisers perform better in the latter optimisation steps (16 \& 128 evaluations), but worse in the initial stages of the optimisation (8 evaluations). In summary, compositional methods yield better performance in two-thirds of the cases considered in Figure~\ref{fig:class_details}.
    
    \section{Conclusions \& Future Directions}\label{Sec:Conc}
    In this paper, we presented an in-depth study of acquisition function maximisation in Bayesian optimisation. Apart from conventional forms typically used in literature, we demonstrated that acquisition functions adhere to a compositional structure enabling numerous new algorithms that led to favourable empirical results. We verified our claims in a rigorous experimental study involving 3958 tasks and twenty-eight optimisers. We used both synthetic and real-world data gathered from Bayesmark. We demonstrated that compositional optimisers outperform traditional solvers in 67 \% of the time. In the future, we plan to extend our analysis to cover non-myopic acquisition functions, constrained and safe BO, as well as to investigate compositional structures of causal BO.

    
    
    \begin{center}
    \begin{figure}[t!]
    \centering
    \includegraphics[width=0.5\textwidth]{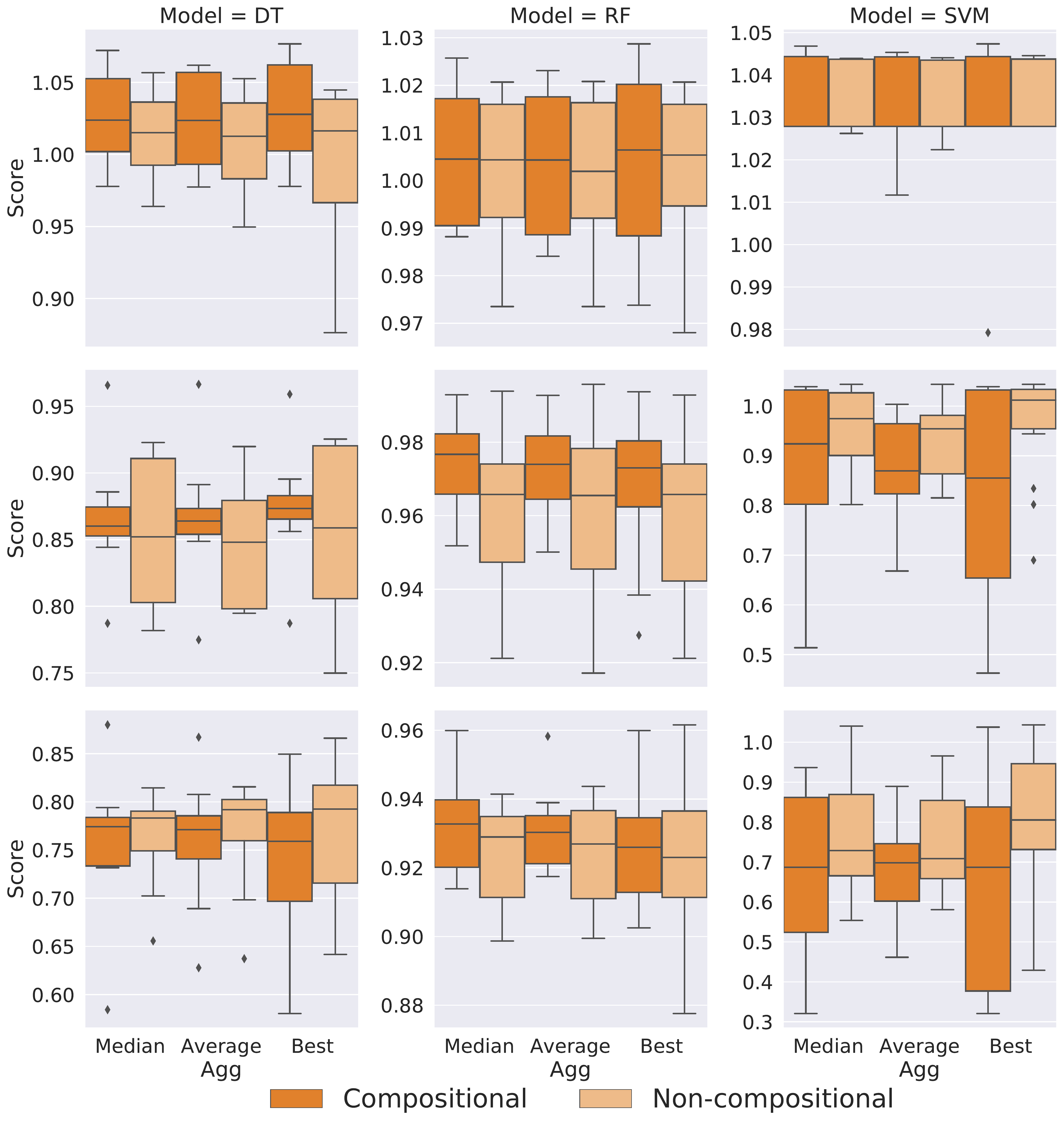}
    \caption{The boxplot shows the quartiles of compositional and non-compositional optimiser performance on the classification hyperparameter tuning task, where the performance metrics are NLL and accuracy. For each model, we show a further split of the optimiser class against different aggregation methods. This plot summarises all 288 experiments conducted on classification tasks for the Bayesmark datasets. We observe that for the DT and RF models, compositional optimisers offer modest performance gains relative to non-compositional optimisers, yet non-compositional optimisers perform better on SVM hyperparameter tuning.}
    \label{fig:class_details}
    \end{figure}
    \end{center}
    
    \newpage
    \bibliography{sample}

\begin{thebibliography}{143}
\providecommand{\natexlab}[1]{#1}
\providecommand{\url}[1]{\texttt{#1}}
\expandafter\ifx\csname urlstyle\endcsname\relax
  \providecommand{\doi}[1]{doi: #1}\else
  \providecommand{\doi}{doi: \begingroup \urlstyle{rm}\Url}\fi

\bibitem[Abdullah et~al.(2019)Abdullah, Ren, Ammar, Milenkovic, Luo, Zhang, and
  Wang]{abdullah2019wasserstein}
Mohammed~Amin Abdullah, Hang Ren, Haitham~Bou Ammar, Vladimir Milenkovic, Rui
  Luo, Mingtian Zhang, and Jun Wang.
\newblock Wasserstein robust reinforcement learning.
\newblock \emph{arXiv preprint arXiv:1907.13196}, 2019.

\bibitem[Amari(1997)]{amari1997neural}
Shun-ichi Amari.
\newblock Neural learning in structured parameter spaces-natural {R}iemannian
  gradient.
\newblock In \emph{Advances In Neural Information Processing Systems}, pages
  127--133, 1997.

\bibitem[Amari(1998)]{1998_Amari}
Shun-ichi Amari.
\newblock Natural gradient works efficiently in learning.
\newblock \emph{Neural {C}omputation}, 10\penalty0 (2):\penalty0 251--276,
  1998.

\bibitem[Amari(2012)]{amari2012differential}
Shun-ichi Amari.
\newblock \emph{Differential-Geometrical Methods in Statistics}.
\newblock Lecture Notes in Statistics. Springer New York, 2012.
\newblock ISBN 9781461250562.
\newblock URL \url{https://books.google.co.uk/books?id=XiDnBwAAQBAJ}.

\bibitem[Amari and Nagaoka(2007)]{amari2007methods}
Shun-ichi Amari and Hiroshi Nagaoka.
\newblock \emph{Methods of Information Geometry}.
\newblock Translations of mathematical monographs. American Mathematical
  Society, 2007.
\newblock ISBN 9780821843024.
\newblock URL \url{https://books.google.co.uk/books?id=vc2FWSo7wLUC}.

\bibitem[Aprem and Roberts(2018)]{2018_Aprem}
Anup Aprem and Stephen Roberts.
\newblock A {B}ayesian optimization approach to compute {N}ash equilibrium of
  potential games using bandit feedback.
\newblock \emph{The Computer Journal}, 2018.

\bibitem[Ariu et~al.(2020)Ariu, Ryu, Yun, and Prouti{\`e}re]{ariu2020regret}
Kaito Ariu, Narae Ryu, Se-Young Yun, and Alexandre Prouti{\`e}re.
\newblock Regret in online recommendation systems.
\newblock \emph{Advances in Neural Information Processing Systems}, 33, 2020.

\bibitem[Astudillo and Frazier(2019)]{2019_Astudillo}
Raul Astudillo and Peter Frazier.
\newblock Bayesian optimization of composite functions.
\newblock In \emph{International Conference on Machine Learning}, pages
  354--363, 2019.

\bibitem[Baioletti et~al.(2020)Baioletti, Di~Bari, Milani, and
  Poggioni]{baioletti2020differential}
Marco Baioletti, Gabriele Di~Bari, Alfredo Milani, and Valentina Poggioni.
\newblock Differential evolution for neural networks optimization.
\newblock \emph{Mathematics}, 8\penalty0 (1):\penalty0 69, 2020.

\bibitem[Balandat et~al.(2020)Balandat, Karrer, Jiang, Daulton, Letham, Wilson,
  and Bakshy]{balandat2019botorch}
Maximilian Balandat, Brian Karrer, Daniel Jiang, Samuel Daulton, Ben Letham,
  Andrew~Gordon Wilson, and Eytan Bakshy.
\newblock Botorch: A framework for efficient {M}onte-{C}arlo {B}ayesian
  optimization.
\newblock \emph{Advances in Neural Information Processing Systems}, 33, 2020.

\bibitem[Bentley(1999)]{bentley1999evolutionary}
Peter~J Bentley.
\newblock \emph{Evolutionary design by computers}.
\newblock Morgan Kaufmann, 1999.

\bibitem[Bergstra and Bengio(2012)]{2012_Bergstra}
James Bergstra and Yoshua Bengio.
\newblock Random search for hyper-parameter optimization.
\newblock \emph{The Journal of Machine Learning Research}, 13\penalty0
  (1):\penalty0 281--305, 2012.

\bibitem[{Blank} and {Deb}(2020)]{pymoo}
Julian. {Blank} and Kalyanmoy. {Deb}.
\newblock Pymoo: Multi-objective optimization in python.
\newblock \emph{IEEE Access}, 8:\penalty0 89497--89509, 2020.

\bibitem[Bottou and Bousquet(2007)]{2007_Bottou}
L{\'e}on Bottou and Olivier Bousquet.
\newblock The tradeoffs of large scale learning.
\newblock \emph{Advances in Neural Information Processing Systems},
  20:\penalty0 161--168, 2007.

\bibitem[Bottou et~al.(2018)Bottou, Curtis, and
  Nocedal]{bottou2018optimization}
L{\'e}on Bottou, Frank~E Curtis, and Jorge Nocedal.
\newblock Optimization methods for large-scale machine learning.
\newblock \emph{Siam Review}, 60\penalty0 (2):\penalty0 223--311, 2018.

\bibitem[Boyd and Vandenberghe(2004)]{boyd_vandenberghe_2004}
Stephen Boyd and Lieven Vandenberghe.
\newblock \emph{Convex Optimization}.
\newblock Cambridge University Press, 2004.

\bibitem[Bresler et~al.(2016)Bresler, Shah, and
  Voloch]{bresler2016collaborative}
Guy Bresler, Devavrat Shah, and Luis~Filipe Voloch.
\newblock Collaborative filtering with low regret.
\newblock In \emph{Proceedings of the 2016 ACM SIGMETRICS International
  Conference on Measurement and Modeling of Computer Science}, pages 207--220,
  2016.

\bibitem[Byrd et~al.(1995)Byrd, Lu, Nocedal, and Zhu]{byrd1995limited}
Richard~H Byrd, Peihuang Lu, Jorge Nocedal, and Ciyou Zhu.
\newblock A limited memory algorithm for bound constrained optimization.
\newblock \emph{SIAM Journal on scientific computing}, 16\penalty0
  (5):\penalty0 1190--1208, 1995.

\bibitem[Byrd et~al.(2016)Byrd, Hansen, Nocedal, and
  Singer]{byrd2016stochastic}
Richard~H Byrd, Samantha~L Hansen, Jorge Nocedal, and Yoram Singer.
\newblock A stochastic quasi-{N}ewton method for large-scale optimization.
\newblock \emph{SIAM Journal on Optimization}, 26\penalty0 (2):\penalty0
  1008--1031, 2016.

\bibitem[Calandra(2017)]{2017_Calandra}
Roberto Calandra.
\newblock \emph{Bayesian modeling for optimization and control in robotics}.
\newblock PhD thesis, Darmstadt, Technische Universit{\"a}t, 2017.

\bibitem[Cao(1985)]{1985_Cao}
Xi-Ren Cao.
\newblock Convergence of parameter sensitivity estimates in a stochastic
  experiment.
\newblock \emph{IEEE Transactions on Automatic Control}, 30\penalty0
  (9):\penalty0 845--853, 1985.

\bibitem[Chen et~al.(2013)Chen, Wang, and Yuan]{chen2013combinatorial}
Wei Chen, Yajun Wang, and Yang Yuan.
\newblock Combinatorial multi-armed bandit: General framework and applications.
\newblock In \emph{International Conference on Machine Learning}, pages
  151--159, 2013.

\bibitem[Chevalier and Ginsbourger(2013)]{2013_Chevalier}
Cl{\'e}ment Chevalier and David Ginsbourger.
\newblock Fast computation of the multi-points expected improvement with
  applications in batch selection.
\newblock In \emph{International Conference on Learning and Intelligent
  Optimization}, pages 59--69. Springer, 2013.

\bibitem[Contal et~al.(2013)Contal, Buffoni, Robicquet, and
  Vayatis]{contal2013parallel}
Emile Contal, David Buffoni, Alexandre Robicquet, and Nicolas Vayatis.
\newblock Parallel {G}aussian process optimization with upper confidence bound
  and pure exploration.
\newblock In \emph{Joint European Conference on Machine Learning and Knowledge
  Discovery in Databases}, pages 225--240. Springer, 2013.

\bibitem[Costa et~al.(2015)Costa, Nannicini, Schroepfer, and
  Wortmann]{2015_Costa}
Alberto Costa, Giacomo Nannicini, Thomas Schroepfer, and Thomas Wortmann.
\newblock Black-box optimization of lighting simulation in architectural
  design.
\newblock In \emph{Complex Systems Design \& Management Asia}, pages 27--39.
  Springer, 2015.

\bibitem[Cowen-Rivers et~al.(2020)Cowen-Rivers, Palenicek, Moens, Abdullah,
  Sootla, Wang, and Ammar]{cowen2020samba}
Alexander~I Cowen-Rivers, Daniel Palenicek, Vincent Moens, Mohammed Abdullah,
  Aivar Sootla, Jun Wang, and Haitham Ammar.
\newblock Samba: Safe model-based \& active reinforcement learning.
\newblock \emph{arXiv preprint arXiv:2006.09436}, 2020.

\bibitem[Cunningham et~al.(2011)Cunningham, Hennig, and
  Lacoste-Julien]{2011_Cunningham}
John~P Cunningham, Philipp Hennig, and Simon Lacoste-Julien.
\newblock Gaussian probabilities and expectation propagation.
\newblock \emph{arXiv preprint arXiv:1111.6832}, 2011.

\bibitem[Daxberger et~al.(2020)Daxberger, Makarova, Turchetta, and
  Krause]{2020_Daxberger}
Erik Daxberger, Anastasia Makarova, Matteo Turchetta, and Andreas Krause.
\newblock Mixed-variable {B}ayesian optimization.
\newblock In \emph{Proceedings of the Twenty-Ninth International Joint
  Conference on Artificial Intelligence, {IJCAI-20}}, pages 2633--2639, 7 2020.

\bibitem[De~G.~Matthews et~al.(2017)De~G.~Matthews, Van Der~Wilk, Nickson,
  Fujii, Boukouvalas, Le{\'o}n-Villagr{\'a}, Ghahramani, and
  Hensman]{2017_gpflow}
Alexander~G De~G.~Matthews, Mark Van Der~Wilk, Tom Nickson, Keisuke Fujii,
  Alexis Boukouvalas, Pablo Le{\'o}n-Villagr{\'a}, Zoubin Ghahramani, and James
  Hensman.
\newblock Gpflow: A {G}aussian process library using {T}ensor{F}low.
\newblock \emph{The Journal of Machine Learning Research}, 18\penalty0
  (1):\penalty0 1299--1304, 2017.

\bibitem[Deb et~al.(2002)Deb, Pratap, Agarwal, and Meyarivan]{deb2002fast}
Kalyanmoy Deb, Amrit Pratap, Sameer Agarwal, and TAMT Meyarivan.
\newblock A fast and elitist multiobjective genetic algorithm: {NSGA-II}.
\newblock \emph{IEEE Transactions on Evolutionary Computation}, 6\penalty0
  (2):\penalty0 182--197, 2002.

\bibitem[Dua and Graff(2017)]{2019_Dua}
Dheeru Dua and Casey Graff.
\newblock {UCI} machine learning repository, 2017.
\newblock URL \url{http://archive.ics.uci.edu/ml}.

\bibitem[Duchi et~al.(2011)Duchi, Hazan, and Singer]{Duchi_2011_adagrad}
John Duchi, Elad Hazan, and Yoram Singer.
\newblock Adaptive subgradient methods for online learning and stochastic
  optimization.
\newblock \emph{Journal of Machine Learning Research}, 12\penalty0
  (61):\penalty0 2121--2159, 2011.

\bibitem[Falkner et~al.(2018)Falkner, Klein, and Hutter]{2018_Falkner}
Stefan Falkner, Aaron Klein, and Frank Hutter.
\newblock {BOHB}: {R}obust and efficient hyperparameter optimization at scale.
\newblock In \emph{International Conference on Machine Learning}, pages
  1437--1446, 2018.

\bibitem[Frazier(2018)]{2018_Frazier}
Peter~I Frazier.
\newblock A tutorial on {B}ayesian optimization.
\newblock \emph{arXiv preprint arXiv:1807.02811}, 2018.

\bibitem[Gabillon et~al.(2020)Gabillon, Tutunov, Valko, and
  Bou-Ammar]{2020_Gabillon}
Victor Gabillon, Rasul Tutunov, Michal Valko, and Haitham Bou-Ammar.
\newblock Derivative-free \& order-robust optimisation.
\newblock In \emph{Proceedings of the Twenty Third International Conference on
  Artificial Intelligence and Statistics}, volume 108, pages 2293--2303. PMLR,
  2020.

\bibitem[Gardner et~al.(2018)Gardner, Pleiss, Bindel, Weinberger, and
  Wilson]{2018_Gardner}
Jacob~R Gardner, Geoff Pleiss, David Bindel, Kilian~Q Weinberger, and
  Andrew~Gordon Wilson.
\newblock Gpytorch: Blackbox matrix-matrix {G}aussian process inference with
  {GPU} acceleration.
\newblock In \emph{Advances in {N}eural {I}nformation {P}rocessing {S}ystems},
  2018.

\bibitem[Garrido-Merch{\'a}n and
  Hern{\'a}ndez-Lobato(2020)]{garrido2020dealing}
Eduardo~C Garrido-Merch{\'a}n and Daniel Hern{\'a}ndez-Lobato.
\newblock Dealing with categorical and integer-valued variables in {B}ayesian
  optimization with {G}aussian processes.
\newblock \emph{Neurocomputing}, 380:\penalty0 20--35, 2020.

\bibitem[Genz(1992)]{1992_Genz}
Alan Genz.
\newblock Numerical computation of multivariate normal probabilities.
\newblock \emph{Journal of computational and graphical statistics}, 1\penalty0
  (2):\penalty0 141--149, 1992.

\bibitem[Genz(2004)]{2004_Genz}
Alan Genz.
\newblock Numerical computation of rectangular bivariate and trivariate normal
  and t probabilities.
\newblock \emph{Statistics and Computing}, 14\penalty0 (3):\penalty0 251--260,
  2004.

\bibitem[Ghadimi et~al.(2020)Ghadimi, Ruszczynski, and Wang]{ghadimi2020single}
Saeed Ghadimi, Andrzej Ruszczynski, and Mengdi Wang.
\newblock A single timescale stochastic approximation method for nested
  stochastic optimization.
\newblock \emph{SIAM Journal on Optimization}, 30\penalty0 (1):\penalty0
  960--979, 2020.

\bibitem[Ginsbourger et~al.(2008)Ginsbourger, Le~Riche, and
  Carraro]{ginsbourger2008multi}
David Ginsbourger, Rodolphe Le~Riche, and Laurent Carraro.
\newblock {A Multi-points Criterion for Deterministic Parallel Global
  Optimization based on Gaussian Processes}.
\newblock Technical report, March 2008.
\newblock URL \url{https://hal.archives-ouvertes.fr/hal-00260579}.

\bibitem[Glasserman(1988)]{1988_Glasserman}
Paul Glasserman.
\newblock Performance continuity and differentiability in {M}onte {C}arlo
  optimization.
\newblock In \emph{1988 Winter Simulation Conference Proceedings}, pages
  518--524. IEEE, 1988.

\bibitem[G{\'o}mez-Bombarelli et~al.(2018)G{\'o}mez-Bombarelli, Wei, Duvenaud,
  Hern{\'a}ndez-Lobato, S{\'a}nchez-Lengeling, Sheberla, Aguilera-Iparraguirre,
  Hirzel, Adams, and Aspuru-Guzik]{2018_Gomez}
Rafael G{\'o}mez-Bombarelli, Jennifer~N Wei, David Duvenaud, Jos{\'e}~Miguel
  Hern{\'a}ndez-Lobato, Benjam{\'\i}n S{\'a}nchez-Lengeling, Dennis Sheberla,
  Jorge Aguilera-Iparraguirre, Timothy~D Hirzel, Ryan~P Adams, and Al{\'a}n
  Aspuru-Guzik.
\newblock Automatic chemical design using a data-driven continuous
  representation of molecules.
\newblock \emph{ACS central science}, 4\penalty0 (2):\penalty0 268--276, 2018.

\bibitem[Gonen and Shalev-Shwartz(2017)]{ERM}
Alon Gonen and Shai Shalev-Shwartz.
\newblock Fast rates for empirical risk minimization of strict saddle problems.
\newblock In \emph{Conference on Learning Theory}, pages 1043--1063, 2017.

\bibitem[Griffiths and Hern{\'a}ndez-Lobato(2020)]{2020_Griffiths}
Ryan-Rhys Griffiths and Jos{\'e}~Miguel Hern{\'a}ndez-Lobato.
\newblock Constrained {B}ayesian optimization for automatic chemical design
  using variational autoencoders.
\newblock \emph{Chemical Science}, 11\penalty0 (2):\penalty0 577--586, 2020.

\bibitem[Griffiths et~al.(2019)Griffiths, Garcia-Ortegon, Aldrick, and
  Lee]{2019_Griffiths}
Ryan-Rhys Griffiths, Miguel Garcia-Ortegon, Alexander~A Aldrick, and Alpha~A
  Lee.
\newblock Achieving robustness to aleatoric uncertainty with heteroscedastic
  {B}ayesian optimisation.
\newblock \emph{arXiv preprint arXiv:1910.07779}, 2019.

\bibitem[Grill et~al.(2015)Grill, Valko, and Munos]{2015_Grill}
Jean-Bastien Grill, Michal Valko, and R{\'e}mi Munos.
\newblock Black-box optimization of noisy functions with unknown smoothness.
\newblock In \emph{Proceedings of the 28th International Conference on Neural
  Information Processing Systems-Volume 1}, pages 667--675, 2015.

\bibitem[Hallak et~al.(2020)Hallak, Mertikopoulos, and
  Cevher]{hallak2020regret}
Nadav Hallak, Panayotis Mertikopoulos, and Volkan Cevher.
\newblock Regret minimization in stochastic non-convex learning via a
  proximal-gradient approach.
\newblock \emph{arXiv preprint arXiv:2010.06250}, 2020.

\bibitem[Hansen(2016)]{2016_Hansen}
Nikolaus Hansen.
\newblock The {CMA} evolution strategy: A tutorial.
\newblock \emph{arXiv preprint arXiv:1604.00772}, 2016.

\bibitem[Hansen and Ostermeier(1996)]{hansen1996adapting}
Nikolaus Hansen and Andreas Ostermeier.
\newblock Adapting arbitrary normal mutation distributions in evolution
  strategies: The covariance matrix adaptation.
\newblock In \emph{Proceedings of IEEE international conference on evolutionary
  computation}, pages 312--317. IEEE, 1996.

\bibitem[Hazan(2016)]{Elad}
Elad Hazan.
\newblock Introduction to online convex optimization.
\newblock \emph{Found. Trends Optim.}, 2\penalty0 (3-4):\penalty0 157--325,
  2016.

\bibitem[Hennig and Schuler(2012)]{2012_Hennig}
Philipp Hennig and Christian~J Schuler.
\newblock Entropy search for information-efficient global optimization.
\newblock \emph{The Journal of Machine Learning Research}, 13\penalty0
  (1):\penalty0 1809--1837, 2012.

\bibitem[Hensman et~al.(2013)Hensman, Fusi, and Lawrence]{2013_Hensman}
James Hensman, Nicol{\`o} Fusi, and Neil~D Lawrence.
\newblock Gaussian processes for {B}ig data.
\newblock In \emph{Proceedings of the Twenty-Ninth Conference on Uncertainty in
  Artificial Intelligence}, pages 282--290, 2013.

\bibitem[Hern{\'a}ndez-Lobato et~al.(2017)Hern{\'a}ndez-Lobato, Requeima,
  Pyzer-Knapp, and Aspuru-Guzik]{2017_Lobato}
Jos{\'e}~Miguel Hern{\'a}ndez-Lobato, James Requeima, Edward~O Pyzer-Knapp, and
  Al{\'a}n Aspuru-Guzik.
\newblock Parallel and distributed {T}hompson sampling for large-scale
  accelerated exploration of chemical space.
\newblock In \emph{International Conference on Machine Learning}, pages
  1470--1479, 2017.

\bibitem[Hinton et~al.(2012)Hinton, Srivastava, and Swersky]{2012_Hinton}
Geoffrey Hinton, Nitish Srivastava, and Kevin Swersky.
\newblock Neural networks for machine learning lecture 6a overview of
  mini-batch gradient descent.
\newblock Coursera: Neural Networks for Machine Learning, 2012.

\bibitem[Hutter et~al.(2011)Hutter, Hoos, and Leyton-Brown]{2011_hutter}
Frank Hutter, Holger~H Hoos, and Kevin Leyton-Brown.
\newblock Sequential model-based optimization for general algorithm
  configuration.
\newblock In \emph{International Conference on Learning and Intelligent
  Optimization}, pages 507--523. Springer, 2011.

\bibitem[Igel et~al.(2006)Igel, Suttorp, and Hansen]{2006_Igel}
Christian Igel, Thorsten Suttorp, and Nikolaus Hansen.
\newblock A computational efficient covariance matrix update and a (1+ 1)-{CMA}
  for evolution strategies.
\newblock In \emph{Proceedings of the 8th Annual Conference on Genetic and
  Evolutionary Computation}, pages 453--460, 2006.

\bibitem[Jamil and Yang(2013)]{Jamil2013survey-benchmarks}
Momin Jamil and Xin-She Yang.
\newblock A literature survey of benchmark functions for global optimisation
  problems.
\newblock \emph{International Journal of Mathematical Modelling and Numerical
  Optimisation}, 4\penalty0 (2):\penalty0 150--194, 2013.

\bibitem[Jang et~al.(2017)Jang, Gu, and Poole]{jang2016categorical}
Eric Jang, Shixiang Gu, and Ben Poole.
\newblock Categorical reparameterization with {G}umbel-softmax.
\newblock In \emph{International Conference on Learning Representations}, 2017.

\bibitem[Jastrebski and Arnold(2006)]{2006_Jastrebski}
Grahame~A Jastrebski and Dirk~V Arnold.
\newblock Improving evolution strategies through active covariance matrix
  adaptation.
\newblock In \emph{2006 IEEE International Conference on Evolutionary
  Computation}, pages 2814--2821. IEEE, 2006.

\bibitem[Jones et~al.(1998)Jones, Schonlau, and Welch]{1998_Jones}
Donald~R Jones, Matthias Schonlau, and William~J Welch.
\newblock Efficient global optimization of expensive black-box functions.
\newblock \emph{Journal of Global Optimization}, 13\penalty0 (4):\penalty0
  455--492, 1998.

\bibitem[Kandasamy et~al.(2018)Kandasamy, Krishnamurthy, Schneider, and
  P{\'o}czos]{2018_Kandasamy}
Kirthevasan Kandasamy, Akshay Krishnamurthy, Jeff Schneider, and Barnab{\'a}s
  P{\'o}czos.
\newblock Parallelised {B}ayesian optimisation via {T}hompson sampling.
\newblock In \emph{International Conference on Artificial Intelligence and
  Statistics}, pages 133--142, 2018.

\bibitem[Kelley(1999)]{BFGSone}
C.T. Kelley.
\newblock \emph{Iterative Methods for Optimization}.
\newblock Frontiers in Applied Mathematics. Society for Industrial and Applied
  Mathematics, 1999.
\newblock ISBN 9780898714333.
\newblock URL \url{https://books.google.co.uk/books?id=Bq6VcmzOe1IC}.

\bibitem[Khosla et~al.(2020)Khosla, Teterwak, Wang, Sarna, Tian, Isola,
  Maschinot, Liu, and Krishnan]{khosla2020supervised}
Prannay Khosla, Piotr Teterwak, Chen Wang, Aaron Sarna, Yonglong Tian, Phillip
  Isola, Aaron Maschinot, Ce~Liu, and Dilip Krishnan.
\newblock Supervised contrastive learning, 2020.

\bibitem[Kim et~al.(2018)Kim, Mnih, Schwarz, Garnelo, Eslami, Rosenbaum,
  Vinyals, and Teh]{2018_Kim}
Hyunjik Kim, Andriy Mnih, Jonathan Schwarz, Marta Garnelo, Ali Eslami, Dan
  Rosenbaum, Oriol Vinyals, and Yee~Whye Teh.
\newblock Attentive neural processes.
\newblock In \emph{International Conference on Learning Representations}, 2018.

\bibitem[Kingma and Ba(2015)]{ADAM}
Diederik~P. Kingma and Jimmy Ba.
\newblock Adam: {A} method for stochastic optimization.
\newblock In Yoshua Bengio and Yann LeCun, editors, \emph{3rd International
  Conference on Learning Representations, {ICLR} 2015, San Diego, CA, USA, May
  7-9, 2015, Conference Track Proceedings}, 2015.

\bibitem[Kingma and Welling(2014)]{2014_Kingma}
Diederik~P. Kingma and Max Welling.
\newblock Auto-encoding variational {B}ayes.
\newblock In Yoshua Bengio and Yann LeCun, editors, \emph{2nd International
  Conference on Learning Representations, {ICLR} 2014, Banff, AB, Canada, April
  14-16, 2014, Conference Track Proceedings}, 2014.

\bibitem[Knudde et~al.(2017)Knudde, van~der Herten, Dhaene, and
  Couckuyt]{knudde2017gpflowopt}
Nicolas Knudde, Joachim van~der Herten, Tom Dhaene, and Ivo Couckuyt.
\newblock Gpflowopt: A {B}ayesian optimization library using {T}ensor{F}low,
  2017.

\bibitem[Korovina et~al.(2020)Korovina, Xu, Kandasamy, Neiswanger, Poczos,
  Schneider, and Xing]{2020_Korovina}
Ksenia Korovina, Sailun Xu, Kirthevasan Kandasamy, Willie Neiswanger, Barnabas
  Poczos, Jeff Schneider, and Eric Xing.
\newblock Chembo: {B}ayesian optimization of small organic molecules with
  synthesizable recommendations.
\newblock In \emph{International Conference on Artificial Intelligence and
  Statistics}, pages 3393--3403. PMLR, 2020.

\bibitem[Kushner(1964)]{1964_Kushner}
Harold~J. Kushner.
\newblock {A New Method of Locating the Maximum Point of an Arbitrary Multipeak
  Curve in the Presence of Noise}.
\newblock \emph{Journal of Basic Engineering}, 86\penalty0 (1):\penalty0
  97--106, 03 1964.

\bibitem[Laguna and Marti(2005)]{Laguna2005opt-benchmark-levy}
Manuel Laguna and Rafael Marti.
\newblock Experimental testing of advanced scatter search designs for global
  optimization of multimodal functions.
\newblock \emph{Journal of Global Optimization}, 33:\penalty0 235--255, 10
  2005.
\newblock \doi{10.1007/s10898-004-1936-z}.

\bibitem[Lattimore and Szepesvári(2020)]{Bandits}
Tor Lattimore and Csaba Szepesvári.
\newblock \emph{Bandit Algorithms}.
\newblock Cambridge University Press, 2020.

\bibitem[Li et~al.(2017)Li, Jamieson, DeSalvo, Rostamizadeh, and
  Talwalkar]{2017_Li}
Lisha Li, Kevin Jamieson, Giulia DeSalvo, Afshin Rostamizadeh, and Ameet
  Talwalkar.
\newblock Hyperband: A novel bandit-based approach to hyperparameter
  optimization.
\newblock \emph{The Journal of Machine Learning Research}, 18\penalty0
  (1):\penalty0 6765--6816, 2017.

\bibitem[Loshchilov and Hutter(2019)]{loshchilov_2019_adamw}
Ilya Loshchilov and Frank Hutter.
\newblock Decoupled weight decay regularization.
\newblock In \emph{International Conference on Learning Representations}, 2019.

\bibitem[Maddison et~al.(2017)Maddison, Mnih, and Teh]{maddison2016concrete}
Chris~J Maddison, Andriy Mnih, and Yee~Whye Teh.
\newblock The concrete distribution: A continuous relaxation of discrete random
  variables.
\newblock In \emph{International Conference on Learning Representations}, 2017.

\bibitem[Mahapatra et~al.(2015)Mahapatra, Ganguli, and Kumar]{2015_Mahapatra}
Prasant~Kumar Mahapatra, Susmita Ganguli, and Amod Kumar.
\newblock A hybrid particle swarm optimization and artificial immune system
  algorithm for image enhancement.
\newblock \emph{Soft Computing}, 19\penalty0 (8):\penalty0 2101--2109, 2015.

\bibitem[McIntire et~al.(2016)McIntire, Ratner, and Ermon]{2016_McIntire}
Mitchell McIntire, Daniel Ratner, and Stefano Ermon.
\newblock Sparse {G}aussian processes for {B}ayesian optimization.
\newblock In \emph{Proceedings of the Thirty-Second Conference on Uncertainty
  in Artificial Intelligence}, pages 517--526, 2016.

\bibitem[Minka(2001{\natexlab{a}})]{2001_Minka}
Thomas~P Minka.
\newblock Expectation propagation for approximate {B}ayesian inference.
\newblock In \emph{Proceedings of the Seventeenth conference on Uncertainty in
  artificial intelligence}, pages 362--369, 2001{\natexlab{a}}.

\bibitem[Minka(2001{\natexlab{b}})]{2001_Minka_phd}
Thomas~P Minka.
\newblock \emph{A family of algorithms for approximate {B}ayesian inference}.
\newblock PhD thesis, Massachusetts Institute of Technology,
  2001{\natexlab{b}}.

\bibitem[Mo{\v{c}}kus(1975)]{1975_Mockus}
Jonas Mo{\v{c}}kus.
\newblock On {B}ayesian methods for seeking the extremum.
\newblock In \emph{Optimization Techniques IFIP Technical Conference}, pages
  400--404. Springer, 1975.

\bibitem[Mokhtari and Ribeiro(2014)]{mokhtari2014res}
Aryan Mokhtari and Alejandro Ribeiro.
\newblock {RES}: {R}egularized stochastic {BFGS} algorithm.
\newblock \emph{IEEE Transactions on Signal Processing}, 62\penalty0
  (23):\penalty0 6089--6104, 2014.

\bibitem[Mokhtari and Ribeiro(2015)]{mokhtari2015global}
Aryan Mokhtari and Alejandro Ribeiro.
\newblock Global convergence of online limited memory {BFGS}.
\newblock \emph{The Journal of Machine Learning Research}, 16\penalty0
  (1):\penalty0 3151--3181, 2015.

\bibitem[Moss et~al.(2020{\natexlab{a}})Moss, Leslie, Beck, Gonzalez, and
  Rayson]{2020_Moss}
Henry Moss, David Leslie, Daniel Beck, Javier Gonzalez, and Paul Rayson.
\newblock Boss: Bayesian optimization over string spaces.
\newblock \emph{Advances in Neural Information Processing Systems}, 33,
  2020{\natexlab{a}}.

\bibitem[Moss and Griffiths(2020)]{2020_FlowMO}
Henry~B Moss and Ryan-Rhys Griffiths.
\newblock Gaussian process molecule property prediction with flowmo.
\newblock \emph{arXiv preprint arXiv:2010.01118}, 2020.

\bibitem[Moss et~al.(2020{\natexlab{b}})Moss, Aggarwal, Prateek, Gonz{\'a}lez,
  and Barra-Chicote]{2020_boffin}
Henry~B Moss, Vatsal Aggarwal, Nishant Prateek, Javier Gonz{\'a}lez, and
  Roberto Barra-Chicote.
\newblock Boffin tts: Few-shot speaker adaptation by bayesian optimization.
\newblock In \emph{ICASSP 2020-2020 IEEE International Conference on Acoustics,
  Speech and Signal Processing (ICASSP)}, pages 7639--7643. IEEE,
  2020{\natexlab{b}}.

\bibitem[Opper et~al.(2001)Opper, Winther, et~al.]{2001_Opper}
Manfred Opper, Ole Winther, et~al.
\newblock From naive mean field theory to the tap equations.
\newblock \emph{Advanced mean field methods: theory and practice}, pages 7--20,
  2001.

\bibitem[Osborne et~al.(2009)Osborne, Garnett, and Roberts]{2009_Osborne}
Michael~A Osborne, Roman Garnett, and Stephen~J Roberts.
\newblock Gaussian processes for global optimization.
\newblock In \emph{3rd International Conference on Learning and Intelligent
  Optimization (LION3)}, pages 1--15, 2009.

\bibitem[Owen(2003)]{owen2003quasi-monte-carlo}
Art~B Owen.
\newblock Quasi-{M}onte {C}arlo sampling.
\newblock \emph{Monte Carlo Ray Tracing: Siggraph 2003 Course 44}, pages
  69--88, 2003.

\bibitem[Pascanu and Bengio(2014)]{pascanu2013revisiting}
Razvan Pascanu and Yoshua Bengio.
\newblock Revisiting natural gradient for deep networks.
\newblock In \emph{International Conference on Learning Representations}, 2014.

\bibitem[Paszke et~al.(2019)Paszke, Gross, Massa, Lerer, Bradbury, Chanan,
  Killeen, Lin, Gimelshein, Antiga, Desmaison, Kopf, Yang, DeVito, Raison,
  Tejani, Chilamkurthy, Steiner, Fang, Bai, and Chintala]{Paszke2019pytorch}
Adam Paszke, Sam Gross, Francisco Massa, Adam Lerer, James Bradbury, Gregory
  Chanan, Trevor Killeen, Zeming Lin, Natalia Gimelshein, Luca Antiga, Alban
  Desmaison, Andreas Kopf, Edward Yang, Zachary DeVito, Martin Raison, Alykhan
  Tejani, Sasank Chilamkurthy, Benoit Steiner, Lu~Fang, Junjie Bai, and Soumith
  Chintala.
\newblock Pytorch: An imperative style, high-performance deep learning library.
\newblock In H.~Wallach, H.~Larochelle, A.~Beygelzimer, F.~d\textquotesingle
  Alch\'{e}-Buc, E.~Fox, and R.~Garnett, editors, \emph{Advances in Neural
  Information Processing Systems 32}, pages 8024--8035. Curran Associates,
  Inc., 2019.

\bibitem[Peng and Li(2015)]{2015_Peng}
Bo~Peng and Lei Li.
\newblock An improved localization algorithm based on genetic algorithm in
  wireless sensor networks.
\newblock \emph{Cognitive Neurodynamics}, 9\penalty0 (2):\penalty0 249--256,
  2015.

\bibitem[Pennington et~al.(2014)Pennington, Socher, and Manning]{GloveNLP}
Jeffrey Pennington, Richard Socher, and Christopher Manning.
\newblock {G}lo{V}e: Global vectors for word representation.
\newblock In \emph{Proceedings of the 2014 Conference on Empirical Methods in
  Natural Language Processing ({EMNLP})}, pages 1532--1543, Doha, Qatar,
  October 2014. Association for Computational Linguistics.
\newblock \doi{10.3115/v1/D14-1162}.
\newblock URL \url{https://www.aclweb.org/anthology/D14-1162}.

\bibitem[Ploskas et~al.(2018)Ploskas, Laughman, Raghunathan, and
  Sahinidis]{2018_Ploskas}
Nikolaos Ploskas, Christopher Laughman, Arvind~U Raghunathan, and Nikolaos~V
  Sahinidis.
\newblock Optimization of circuitry arrangements for heat exchangers using
  derivative-free optimization.
\newblock \emph{Chemical Engineering Research and Design}, 131:\penalty0
  16--28, 2018.

\bibitem[Price(1996)]{price1996differential}
Kenneth~V Price.
\newblock Differential evolution: a fast and simple numerical optimizer.
\newblock In \emph{Proceedings of North American Fuzzy Information Processing},
  pages 524--527. IEEE, 1996.

\bibitem[Rasmussen and Williams(2006)]{2006_Williams}
Carl~Edward Rasmussen and Christopher~KI Williams.
\newblock \emph{Gaussian Processes for Machine Learning}, volume~2.
\newblock MIT press Cambridge, MA, 2006.

\bibitem[Rezende et~al.(2014)Rezende, Mohamed, and Wierstra]{2014_Rezende}
Danilo~Jimenez Rezende, Shakir Mohamed, and Daan Wierstra.
\newblock Stochastic backpropagation and approximate inference in deep
  generative models.
\newblock In \emph{Proceedings of the 31st International Conference on Machine
  Learning-Volume 32}, pages II--1278, 2014.

\bibitem[Riedel(1992)]{riedel1992sherman}
Kurt~S Riedel.
\newblock A {S}herman--{M}orrison--{W}oodbury identity for rank augmenting
  matrices with application to centering.
\newblock \emph{SIAM Journal on Matrix Analysis and Applications}, 13\penalty0
  (2):\penalty0 659--662, 1992.

\bibitem[Riedmiller and Braun(1993)]{riedmiller1993direct}
Martin Riedmiller and Heinrich Braun.
\newblock A direct adaptive method for faster backpropagation learning: The
  {R}prop algorithm.
\newblock In \emph{IEEE International Conference on Neural Networks}, 1993.

\bibitem[Robbins and Monro(1951)]{RobbMonr51}
Herbert Robbins and Sutton Monro.
\newblock A stochastic approximation method.
\newblock \emph{The Annals of Mathematical Statistics}, 22\penalty0
  (3):\penalty0 400--407, 1951.

\bibitem[Ru et~al.(2019)Ru, Alvi, Nguyen, Osborne, and Roberts]{ru2019bayesian}
Binxin Ru, Ahsan~S Alvi, Vu~Nguyen, Michael~A Osborne, and Stephen~J Roberts.
\newblock Bayesian optimisation over multiple continuous and categorical
  inputs.
\newblock \emph{arXiv preprint arXiv:1906.08878}, 2019.

\bibitem[Schmidt et~al.(2017)Schmidt, Le~Roux, and Bach]{FSM}
Mark Schmidt, Nicolas Le~Roux, and Francis Bach.
\newblock Minimizing finite sums with the stochastic average gradient.
\newblock \emph{Mathematical Programming}, 162\penalty0 (1-2):\penalty0
  83--112, 2017.

\bibitem[Schmidt et~al.(2020)Schmidt, Schneider, and
  Hennig]{schmidt2020descendingcrowdedvalley}
Robin~M Schmidt, Frank Schneider, and Philipp Hennig.
\newblock Descending through a crowded valley--benchmarking deep learning
  optimizers.
\newblock \emph{arXiv preprint arXiv:2007.01547}, 2020.

\bibitem[Schrack and Choit(1976)]{1976_Schrack}
G{\"u}nther Schrack and Mark Choit.
\newblock Optimized relative step size random searches.
\newblock \emph{Mathematical Programming}, 10\penalty0 (1):\penalty0 230--244,
  1976.

\bibitem[Schumer and Steiglitz(1968)]{1968_Schumer}
MA~Schumer and Kenneth Steiglitz.
\newblock Adaptive step size random search.
\newblock \emph{IEEE Transactions on Automatic Control}, 13\penalty0
  (3):\penalty0 270--276, 1968.

\bibitem[Shah and Sahinidis(2012)]{2012_Shah}
Shweta~B Shah and Nikolaos~V Sahinidis.
\newblock {SAS}-{P}ro: {S}imultaneous residue assignment and structure
  superposition for protein structure alignment.
\newblock \emph{PloS one}, 7\penalty0 (5):\penalty0 e37493, 2012.

\bibitem[Shahriari et~al.(2016)Shahriari, Swersky, Wang, Adams, and
  de~Freitas]{2016_Shahriari}
Bobak Shahriari, Kevin Swersky, Ziyu Wang, Ryan~P Adams, and Nando de~Freitas.
\newblock Taking the human out of the loop: A review of {B}ayesian
  optimization.
\newblock \emph{Proceedings of the IEEE}, 1\penalty0 (104):\penalty0 148--175,
  2016.

\bibitem[Shalev-Shwartz and Singer(2007)]{shalev2007online}
Shai Shalev-Shwartz and Yoram Singer.
\newblock Online learning: Theory, algorithms, and applications.
\newblock 2007.

\bibitem[Shanno(1970)]{shanno1970conditioning}
David~F Shanno.
\newblock Conditioning of quasi-{N}ewton methods for function minimization.
\newblock \emph{Mathematics of computation}, 24\penalty0 (111):\penalty0
  647--656, 1970.

\bibitem[Snoek et~al.(2012)Snoek, Larochelle, and Adams]{2012_Snoek}
Jasper Snoek, Hugo Larochelle, and Ryan~P Adams.
\newblock Practical {B}ayesian optimization of machine learning algorithms.
\newblock In \emph{Advances in Neural Information Processing Systems}, pages
  2951--2959, 2012.

\bibitem[{Song} et~al.(2020){Song}, {Hu}, {Ding}, {Di}, and {Song}]{IR}
A.~{Song}, Q.~{Hu}, X.~{Ding}, X.~{Di}, and Z.~{Song}.
\newblock Similar face recognition using the ie-cnn model.
\newblock \emph{IEEE Access}, 8:\penalty0 45244--45253, 2020.
\newblock \doi{10.1109/ACCESS.2020.2978938}.

\bibitem[Song et~al.(2019)Song, Chen, and Yue]{2019_Song}
Jialin Song, Yuxin Chen, and Yisong Yue.
\newblock A general framework for multi-fidelity {B}ayesian optimization with
  {G}aussian processes.
\newblock In \emph{The 22nd International Conference on Artificial Intelligence
  and Statistics}, pages 3158--3167. PMLR, 2019.

\bibitem[Speranskii(2015)]{2015_Speranskii}
Dmitrii~V Speranskii.
\newblock Ant colony optimization algorithms for digital device diagnostics.
\newblock \emph{Automatic Control and Computer Sciences}, 49\penalty0
  (2):\penalty0 82--87, 2015.

\bibitem[Springenberg et~al.(2016)Springenberg, Klein, Falkner, and
  Hutter]{2016_Springenberg}
Jost~Tobias Springenberg, Aaron Klein, Stefan Falkner, and Frank Hutter.
\newblock Bayesian optimization with robust {B}ayesian neural networks.
\newblock In \emph{Advances in Neural Information Processing Systems}, pages
  4134--4142, 2016.

\bibitem[Srinivas et~al.(2010)Srinivas, Krause, Kakade, and
  Seeger]{srinivas2009gaussian}
Niranjan Srinivas, Andreas Krause, Sham Kakade, and Matthias Seeger.
\newblock Gaussian process optimization in the bandit setting: no regret and
  experimental design.
\newblock In \emph{Proceedings of the 27th International Conference on Machine
  Learning}, pages 1015--1022, 2010.

\bibitem[Stein(2012)]{2012_Stein}
Michael~L Stein.
\newblock \emph{Interpolation of spatial data: some theory for kriging}.
\newblock Springer Science \& Business Media, 2012.

\bibitem[Sun et~al.(2019)Sun, Cao, Zhu, and Zhao]{Survey}
Shiliang Sun, Zehui Cao, Han Zhu, and Jing Zhao.
\newblock A survey of optimization methods from a machine learning perspective.
\newblock \emph{IEEE Transactions on Cybernetics}, 50\penalty0 (8):\penalty0
  3668--3681, 2019.

\bibitem[Thawani et~al.(2020)Thawani, Griffiths, Jamasb, Bourached, Jones,
  McCorkindale, Aldrick, and Lee]{2020_Thawani}
Aditya~R Thawani, Ryan-Rhys Griffiths, Arian Jamasb, Anthony Bourached,
  Penelope Jones, William McCorkindale, Alexander~A Aldrick, and Alpha~A Lee.
\newblock The photoswitch dataset: A molecular machine learning benchmark for
  the advancement of synthetic chemistry.
\newblock \emph{arXiv preprint arXiv:2008.03226}, 2020.

\bibitem[Thompson(1933)]{1933_Thompson}
William~R Thompson.
\newblock On the likelihood that one unknown probability exceeds another in
  view of the evidence of two samples.
\newblock \emph{Biometrika}, 25\penalty0 (3/4):\penalty0 285--294, 1933.

\bibitem[Tieleman and Hinton(2012)]{Tieleman2012}
T.~Tieleman and G.~Hinton.
\newblock {Lecture 6.5---{RMS}prop: Divide the gradient by a running average of
  its recent magnitude}.
\newblock Coursera: Neural Networks for Machine Learning, 2012.

\bibitem[Titsias(2009)]{2009_Titsias}
Michalis Titsias.
\newblock Variational learning of inducing variables in sparse {G}aussian
  processes.
\newblock In \emph{Artificial Intelligence and Statistics}, pages 567--574,
  2009.

\bibitem[Tutunov et~al.(2015)Tutunov, Bou-Ammar, and
  Jadbabaie]{tutunov2015distributed}
Rasul Tutunov, Haitham Bou-Ammar, and Ali Jadbabaie.
\newblock Distributed {SDDM} solvers: {T}heory \& applications.
\newblock \emph{arXiv preprint arXiv:1508.04096}, 2015.

\bibitem[Tutunov et~al.(2019)Tutunov, Bou-Ammar, and
  Jadbabaie]{tutunov2019distributed}
Rasul Tutunov, Haitham Bou-Ammar, and Ali Jadbabaie.
\newblock Distributed {N}ewton method for large-scale consensus optimization.
\newblock \emph{IEEE Transactions on Automatic Control}, 64\penalty0
  (10):\penalty0 3983--3994, 2019.

\bibitem[Tutunov et~al.(2020)Tutunov, Li, Wang, and
  Bou-Ammar]{tutunov2020cadam}
Rasul Tutunov, Minne Li, Jun Wang, and Haitham Bou-Ammar.
\newblock Compositional {ADAM}: {A}n adaptive compositional solver.
\newblock \emph{arXiv preprint arXiv:2002.03755}, 2020.

\bibitem[Valko et~al.(2013)Valko, Carpentier, and Munos]{2013_Valko}
Michal Valko, Alexandra Carpentier, and R{\'e}mi Munos.
\newblock Stochastic simultaneous optimistic optimization.
\newblock In \emph{International Conference on Machine Learning}, pages 19--27,
  2013.

\bibitem[van Rijn et~al.(2016)van Rijn, Wang, van Leeuwen, and B{\"a}ck]{CMAES}
Sander van Rijn, Hao Wang, Matthijs van Leeuwen, and Thomas B{\"a}ck.
\newblock Evolving the structure of evolution strategies.
\newblock In \emph{2016 IEEE Symposium Series on Computational Intelligence
  (SSCI)}, pages 1--8. IEEE, 2016.

\bibitem[Viappiani and Boutilier(2009)]{viappiani2009regret}
Paolo Viappiani and Craig Boutilier.
\newblock Regret-based optimal recommendation sets in conversational
  recommender systems.
\newblock In \emph{Proceedings of the third ACM conference on Recommender
  systems}, pages 101--108, 2009.

\bibitem[Virtanen et~al.(2020)Virtanen, Gommers, Oliphant, Haberland, Reddy,
  Cournapeau, Burovski, Peterson, Weckesser, Bright, et~al.]{virtanen2020scipy}
Pauli Virtanen, Ralf Gommers, Travis~E Oliphant, Matt Haberland, Tyler Reddy,
  David Cournapeau, Evgeni Burovski, Pearu Peterson, Warren Weckesser, Jonathan
  Bright, et~al.
\newblock Scipy 1.0: fundamental algorithms for scientific computing in python.
\newblock \emph{Nature methods}, 17\penalty0 (3):\penalty0 261--272, 2020.

\bibitem[Wang et~al.(2020)Wang, Clark, Liu, and Frazier]{wang2019parallel}
Jialei Wang, Scott~C Clark, Eric Liu, and Peter~I Frazier.
\newblock Parallel {B}ayesian global optimization of expensive functions.
\newblock \emph{Operations Research}, 2020.

\bibitem[Wang and Liu(2016)]{wang2016stochastic}
Mengdi Wang and Ji~Liu.
\newblock A stochastic compositional gradient method using {M}arkov samples.
\newblock In \emph{2016 Winter Simulation Conference (WSC)}, pages 702--713.
  IEEE, 2016.

\bibitem[Wang et~al.(2017{\natexlab{a}})Wang, Fang, and Liu]{2017_Wang}
Mengdi Wang, Ethan~X Fang, and Han Liu.
\newblock Stochastic compositional gradient descent: algorithms for minimizing
  compositions of expected-value functions.
\newblock \emph{Mathematical Programming}, 161\penalty0 (1-2):\penalty0
  419--449, 2017{\natexlab{a}}.

\bibitem[Wang et~al.(2017{\natexlab{b}})Wang, Liu, and Fang]{2017_MWang}
Mengdi Wang, Ji~Liu, and Ethan~X Fang.
\newblock Accelerating stochastic composition optimization.
\newblock \emph{The Journal of Machine Learning Research}, 18\penalty0
  (1):\penalty0 3721--3743, 2017{\natexlab{b}}.

\bibitem[Wang and Jegelka(2017)]{2017_Jegelka}
Zi~Wang and Stefanie Jegelka.
\newblock Max-value entropy search for efficient {B}ayesian optimization.
\newblock In \emph{International Conference on Machine Learning}, pages
  3627--3635, 2017.

\bibitem[White et~al.(2019)White, Neiswanger, and Savani]{2019_White}
Colin White, Willie Neiswanger, and Yash Savani.
\newblock {BANANAS}: {B}ayesian optimization with neural architectures for
  neural architecture search.
\newblock \emph{arXiv preprint arXiv:1910.11858}, 2019.

\bibitem[Wilson et~al.(2018{\natexlab{a}})Wilson, Roelofs, Stern, Srebro, and
  Recht]{wilson2018marginal}
Ashia~C. Wilson, Rebecca Roelofs, Mitchell Stern, Nathan Srebro, and Benjamin
  Recht.
\newblock The marginal value of adaptive gradient methods in machine learning,
  2018{\natexlab{a}}.

\bibitem[Wilson et~al.(2018{\natexlab{b}})Wilson, Hutter, and
  Deisenroth]{2018_Wilson}
James Wilson, Frank Hutter, and Marc Deisenroth.
\newblock Maximizing acquisition functions for {B}ayesian optimization.
\newblock In \emph{Advances in Neural Information Processing Systems}, pages
  9884--9895, 2018{\natexlab{b}}.

\bibitem[Wilson et~al.(2020)Wilson, Borovitskiy, Terenin, Mostowsky, and
  Deisenroth]{2020_Wilson}
James~T Wilson, Viacheslav Borovitskiy, Alexander Terenin, Peter Mostowsky, and
  Marc~Peter Deisenroth.
\newblock Efficiently sampling functions from {G}aussian process posteriors.
\newblock In \emph{International Conference on Machine Learning}, 2020.

\bibitem[Wu et~al.(2020)Wu, Guo, and Moustafa]{wu2020pelican}
Peilun Wu, Hui Guo, and Nour Moustafa.
\newblock Pelican: A deep residual network for network intrusion detection,
  2020.

\bibitem[Yang et~al.(2020)Yang, Tutunov, Sakulwongtana, and
  Ammar]{yang2019alpha}
Yaodong Yang, Rasul Tutunov, Phu Sakulwongtana, and Haitham~Bou Ammar.
\newblock $\alpha^{\alpha}$-{R}ank: Practically scaling $\alpha$-rank through
  stochastic optimisation.
\newblock In \emph{Proceedings of the 19th International Conference on
  Autonomous Agents and MultiAgent Systems}, pages 1575--1583, 2020.

\bibitem[Yin and Zhou(2018)]{yin2018semi}
Mingzhang Yin and Mingyuan Zhou.
\newblock Semi-implicit variational inference.
\newblock In \emph{International Conference on Machine Learning}, pages
  5660--5669, 2018.

\bibitem[Yoo and Han(2014)]{2014_Yoo}
Kwang-Seon Yoo and Seog-Young Han.
\newblock Modified ant colony optimization for topology optimization of
  geometrically nonlinear structures.
\newblock \emph{International Journal of Precision Engineering and
  Manufacturing}, 15\penalty0 (4):\penalty0 679--687, 2014.

\bibitem[Zeiler(2012)]{zeiler_2012_adadelta}
Matthew~D Zeiler.
\newblock Adadelta: {A}n adaptive learning rate method.
\newblock \emph{arXiv preprint arXiv:1212.5701}, 2012.

\bibitem[Zhang et~al.(2015)Zhang, Wang, and Long]{2015_Zhang}
Zhuhong Zhang, Lei Wang, and Fei Long.
\newblock Immune optimization approach solving multi-objective
  chance-constrained programming.
\newblock \emph{Evolving Systems}, 6\penalty0 (1):\penalty0 41--53, 2015.

\bibitem[Zhu et~al.(1997)Zhu, Byrd, Lu, and Nocedal]{1997_Zhu}
Ciyou Zhu, Richard~H Byrd, Peihuang Lu, and Jorge Nocedal.
\newblock Algorithm 778: {L-BFGS-B}: {F}ortran subroutines for large-scale
  bound-constrained optimization.
\newblock \emph{ACM Transactions on Mathematical Software (TOMS)}, 23\penalty0
  (4):\penalty0 550--560, 1997.

\end{thebibliography}
    \appendix
    \section{Compositional Construction of Acquisition Functions}\label{App:StochastiCCompFormulation} 
    Given a collection of $M$ i.i.d samples $\{\textbf{z}_m\}^M_{m=1}$ and following finite sum approximations for the acquisition functions given in Equations \ref{eq:reparam-EIFSM} - \ref{eq:reparam-PIFSM}, we now provide detailed compositional reformulations for them (see  Section \ref{Sec: Section_Comp_form}). Let $\omega$ be a random variable distributed uniformly on a collection $\{1, \ldots, M\}$, i.e. $\omega\sim\text{Uniform}([1:M])$:
    
    \subsection{Expected Improvement:}
    Consider an inner stochastic mapping $\textbf{g}^{(\text{EI})}_{\omega}:\mathbb{R}^{dq}\to\mathbb{R}^{q\times M}$, such that:
    \begin{align*}
        \textbf{g}^{(\text{EI})}_{\omega}(\textbf{x}_{1:q}) = [\textbf{0}_q,\ldots, \textbf{v}^{(\text{EI})}_{\omega}, \ldots, \textbf{0}_q]. 
    \end{align*}
    where $\textbf{v}^{(\text{EI})}_{m} = \text{ReLU}\left(\boldsymbol{\mu}_{i}(\textbf{x}_{1:q}; \boldsymbol{\theta}) + \textbf{L}_{i}(\textbf{x}_{1:q}; \boldsymbol{\theta})\textbf{z}_{m} - f(\textbf{x}_{i}^{+})\textbf{1}_{q}\right)\in\mathbb{R}^{q}$ for $m=1:M$. Hence, taking the expectation with respect to $\omega$ gives 
    \begin{align*}
        \mathbb{E}_{\omega}[\textbf{g}^{(\text{EI})}_{\omega}(\textbf{x}_{1:q})] = \frac{1}{M}[\textbf{v}^{(\text{EI})}_{1},\ldots, \textbf{v}^{(\text{EI})}_{M}]
    \end{align*}
    Now let us consider an outer deterministic mapping $f^{(\text{EI})}:\mathbb{R}^{q\times M}\to\mathbb{R}$, such that for a given $q\times M$ input matrix:
    \begin{align*}
        &f^{(\text{EI})}\left(\begin{bmatrix}
     a_{11} & a_{12} & \ldots &a_{1M} \\
     a_{21} & a_{22} & \ldots &a_{2M} \\
     \vdots&\vdots &\vdots &\vdots\\
     a_{q1} & a_{q2} & \ldots &a_{qM} \\
     \end{bmatrix}\right) = \sum_{m=1}^{M} \max\{a_{1m},\ldots, a_{qm}\}.
    \end{align*}
    Therefore,
    \begin{align*}
        &f^{(\text{EI})}(\mathbb{E}_{\omega}[\textbf{g}^{(\text{EI})}_{\omega}(\textbf{x}_{1:q})]) =\frac{1}{M}\sum_{m=1}^M\max_{j\in1:q}\left\{\text{ReLU}\left(\boldsymbol{\mu}_{i}(\textbf{x}_{1:q}; \boldsymbol{\theta}) + \textbf{L}_{i}(\textbf{x}_{1:q}; \boldsymbol{\theta})\textbf{z}_{m} - f(\textbf{x}_{i}^{+})\textbf{1}_{q}\right)\right\} = \alpha^{(\text{FSM})}_{\text{rq-EI}}.
    \end{align*}

    \subsection{Probability of  Improvement:} Consider an inner stochastic mapping $\textbf{g}^{(\text{PI})}_{\omega}:\mathbb{R}^{dq}\to\mathbb{R}^{q\times M}$, such that:
    \begin{align*}
        \textbf{g}^{(\text{PI})}_{\omega}(\textbf{x}_{1:q}) = [\textbf{0}_q,\ldots, \textbf{v}^{(\text{PI})}_{\omega}, \ldots, \textbf{0}_q]. 
    \end{align*}
    where $\textbf{v}^{(\text{PI})}_{m} = \frac{1}{\tau}\left[\boldsymbol{\mu}_{i}(\textbf{x}_{1:q}; \boldsymbol{\theta}) + \textbf{L}_{i}(\textbf{x}_{1:q}; \boldsymbol{\theta})\textbf{z}_{m} - f(\textbf{x}_{i}^{+})\textbf{1}_{q}\right]\in\mathbb{R}^{q}$ for $m=1:M$. Hence, taking the expectation with respect to $\omega$ gives 
    \begin{align*}
        \mathbb{E}_{\omega}[\textbf{g}^{(\text{PI})}_{\omega}(\textbf{x}_{1:q})] = \frac{1}{M}[\textbf{v}^{(\text{PI})}_{1},\ldots, \textbf{v}^{(\text{PI})}_{M}]
    \end{align*}
    Now let us consider an outer deterministic mapping $f^{(\text{PI})}:\mathbb{R}^{q\times M}\to\mathbb{R}$, such that for a given $q\times M$ input matrix:
    \begin{align*}
        &f^{(\text{PI})}\left(\begin{bmatrix}
     a_{11} & a_{12} & \ldots &a_{1M} \\
     a_{21} & a_{22} & \ldots &a_{2M} \\
     \vdots&\vdots &\vdots &\vdots\\
     a_{q1} & a_{q2} & \ldots &a_{qM} \\
     \end{bmatrix}\right) =  \frac{1}{M}\sum_{m=1}^{M} \max_{j\in 1:q}\left\{\text{Sig}(M[a_{1m},\ldots,a_{qm}])\right\}
    \end{align*}
    Therefore,
    \begin{align*}
        &f^{(\text{PI})}(\mathbb{E}_{\omega}[\textbf{g}^{(\text{PI})}_{\omega}(\textbf{x}_{1:q})]) =\frac{1}{M}\sum_{m=1}^{M}\max_{j\in1:q}\left\{\text{Sig}\left(\frac{\boldsymbol{\mu}_{i}(\textbf{x}_{1:q}; \boldsymbol{\theta}) + \textbf{L}_{i}(\textbf{x}_{1:q}; \boldsymbol{\theta})\textbf{z}_{m} - f(\textbf{x}_{i}^{+})\textbf{1}_{q}}{\tau}\right)\right\} = \alpha^{(\text{FSM})}_{\text{rq-PI}}.
    \end{align*}
    
    \subsection{Simple Regret:} Consider an inner stochastic mapping $\textbf{g}^{(\text{PI})}_{\omega}:\mathbb{R}^{dq}\to\mathbb{R}^{q\times M}$, such that:
    \begin{align*}
        \textbf{g}^{(\text{SR})}_{\omega}(\textbf{x}_{1:q}) = [\textbf{0}_q,\ldots, \textbf{v}^{(\text{SR})}_{\omega}, \ldots, \textbf{0}_q]. 
    \end{align*}
    where $\textbf{v}^{(\text{SR})}_{m} = \boldsymbol{\mu}_{i}(\textbf{x}_{1:q}; \boldsymbol{\theta}) + \textbf{L}_{i}(\textbf{x}_{1:q}; \boldsymbol{\theta})\textbf{z}_{m}\in\mathbb{R}^{q}$ for $m=1:M$. Hence, taking the expectation with respect to $\omega$ gives 
    \begin{align*}
        \mathbb{E}_{\omega}[\textbf{g}^{(\text{SR})}_{\omega}(\textbf{x}_{1:q})] = \frac{1}{M}[\textbf{v}^{(\text{SR})}_{1},\ldots, \textbf{v}^{(\text{SR})}_{M}]
    \end{align*}
    Now let us consider an outer deterministic mapping $f^{(\text{SR})}:\mathbb{R}^{q\times M}\to\mathbb{R}$, such that for a given $q\times M$ input matrix:
    \begin{align*}
        &f^{(\text{SR})}\left(\begin{bmatrix}
     a_{11} & a_{12} & \ldots &a_{1M} \\
     a_{21} & a_{22} & \ldots &a_{2M} \\
     \vdots&\vdots &\vdots &\vdots\\
     a_{q1} & a_{q2} & \ldots &a_{qM} \\
     \end{bmatrix}\right) =  \sum_{m=1}^{M} \max\left\{a_{1m}, \ldots, a_{qm}\right\}.
    \end{align*}
    Therefore,
    \begin{align*}
        &f^{(\text{SR})}(\mathbb{E}_{\omega}[\textbf{g}^{(\text{SR})}_{\omega}(\textbf{x}_{1:q})]) =\frac{1}{M} \sum_{m=1}^{M}\max_{j\in1:q}\left\{\boldsymbol{\mu}_{i}(\textbf{x}_{1:q}; \boldsymbol{\theta}) + \textbf{L}_{i}(\textbf{x}_{1:q}; \boldsymbol{\theta})\textbf{z}_m\right\} = \alpha^{(\text{FSM})}_{\text{rq-SR}}.
    \end{align*}
    
    \subsection{Upper Confidence Bound:} Consider an inner stochastic mapping $\textbf{g}^{(\text{UCB})}_{\omega}:\mathbb{R}^{dq}\to\mathbb{R}^{q\times M}$, such that:
    \begin{align*}
        \textbf{g}^{(\text{UCB})}_{\omega}(\textbf{x}_{1:q}) = [\textbf{0}_q,\ldots, \textbf{v}^{(\text{UCB})}_{\omega}, \ldots, \textbf{0}_q].
    \end{align*}
    where $\textbf{v}^{(\text{UCB})}_{m} = \boldsymbol{\mu}_{i}(\textbf{x}_{1:q}; \boldsymbol{\theta}) + \sqrt{\sfrac{\beta\pi}{2}}\left|\textbf{L}_{i}(\textbf{x}_{1:q}; \boldsymbol{\theta})\textbf{z}_{m}\right|\in\mathbb{R}^{q}$ for $m=1:M$. Hence, taking the expectation with respect to $\omega$ gives 
    \begin{align*}
        \mathbb{E}_{\omega}[\textbf{g}^{(\text{UCB})}_{\omega}(\textbf{x}_{1:q})] = \frac{1}{M}[\textbf{v}^{(\text{UCB})}_{1},\ldots, \textbf{v}^{(\text{UCB})}_{M}]
    \end{align*}
    Now let us consider an outer deterministic mapping $f^{(\text{UCB})}:\mathbb{R}^{q\times M}\to\mathbb{R}$, such that for a given $q\times M$ input matrix:
    \begin{align*}
        &f^{(\text{UCB})}\left(\begin{bmatrix}
     a_{11} & a_{12} & \ldots &a_{1M} \\
     a_{21} & a_{22} & \ldots &a_{2M} \\
     \vdots&\vdots &\vdots &\vdots\\
     a_{q1} & a_{q2} & \ldots &a_{qM} \\
     \end{bmatrix}\right) =  \sum_{m=1}^{M} \max\left\{a_{1m}, \ldots, a_{qm}\right\}.
    \end{align*}
    Therefore,
    \begin{align*}
        &f^{(\text{UCB})}(\mathbb{E}_{\omega}[\textbf{g}^{(\text{UCB})}_{\omega}(\textbf{x}_{1:q})]) =\frac{1}{M}\sum_{m=1}^{M}\max_{j\in1:q}\left\{\boldsymbol{\mu}_i(\textbf{x}_{1:q}; \boldsymbol{\theta}) + \sqrt{\sfrac{\beta \pi}{2}}|\textbf{L}_{i}(\textbf{x}_{1:q}; \boldsymbol{\theta})\textbf{z}_{m}|\right\} = \alpha^{(\text{FSM})}_{\text{rq-UCB}}.
    \end{align*}
    
    \section{Zero-Order Optimisation Algorithms for ERM-BO}\label{App:ZeroOrder}
    \subsection{Random Search:} \label{App:RS}
    The most simple zeroth-order strategy we attempted in our experiments was random search (RS), where a new batch of query points is constructed by sampling $q$ candidates $\textbf{x}_{1:q}$ uniformly at random from a bounded search domain. Though simple, RS has been shown to be an effective optimisation scheme in certain settings~\citep{2012_Bergstra, 2017_Li} and can serve as an essential low-memory, low-compute baseline for any acquisition optimiser.   
    
    \subsection{CMA-ES:} \label{App:CMAES}
    In the covariance matrix adaptation evolution strategy (CMA-ES)~\citep{CMAES, pymoo}, a population of new search points is
    generated by sampling a multivariate normal distribution, which for generations $g=0, 1, \dots, $ can be written as: 
    \begin{equation}
    \label{Eq:CMAES}
        \text{vec}\left(\textbf{x}\right)_{l}^{(g+1)} \sim \boldsymbol{\mu}_{\text{CMA-ES}}^{(g)} + \sigma_{\text{CMA-ES}}^{(g)} \mathcal{N}\left(\textbf{0}, \boldsymbol{\Sigma}_{\text{CMA-ES}}^{(g)}\right) \ \ \text{for $l \in [1:\#\text{off-springs}]$,}
    \end{equation}
    where $\boldsymbol{\mu}_{\text{CMA-ES}}^{(g)},  \sigma_{\text{CMA-ES}}^{(g)}$ and $\boldsymbol{\Sigma}_{\text{CMA-ES}}^{(g)}$ are the distribution's hyperparameters that will be updated based on function value information. Also, $\#\text{off-springs} > 2$ represents the number of individuals sampled from a population, e.g., the number of optimiser restarts in our case. Moreover, the usage of the $\text{vec}(\textbf{x}) \in \mathbb{R}^{dq}$ notation denotes a vector of inputs across all batches and dimensions.  
    
    Starting from an initialisation $\boldsymbol{\mu}_{\text{CMA-ES}}^{(0)},  \sigma_{\text{CMA-ES}}^{(0)}$ and $\boldsymbol{\Sigma}_{\text{CMA-ES}}^{(0)}$, CMA-ES updates each of the hyperparameters based on fitness or function values to improve the guess of $\textbf{x}^{\star}$. At some generation $g+1$, the algorithm first samples $\text{vec}(\textbf{x})_{1}^{(g+1)}, \dots, \text{vec}({x})_{\#\text{off-springs}}^{(g+1)}$ according to Equation~\ref{Eq:CMAES} and then ranks individual samples in a descending order based on their acquisition evaluation such that\footnote{Please note that we use $\alpha_{\text{rq-type}}(\cdot)$ to denote one of the reparameterised acquisitions (i.e., EI, PI, UCB, and SR).} $\alpha_{\text{rq-type}}(\text{vec}(\textbf{x})_{1^{\star}}^{(g+1)}|\mathcal{D}_{i})\leq \dots \leq \alpha_{\text{rq-type}}(\text{vec}(\textbf{x})_{ \text{$\#$off-springs}^{\star}}^{(g+1)}|\mathcal{D}_{i})$, where $\text{vec}(\textbf{x})^{(g+1)}_{j^{\star}}$ is the $j^{th}$ best sample vector (according to its acquisition value) from $\text{vec}(\textbf{x})_{1}^{(g+1)}, \dots, \text{vec}(\textbf{x})_{\#\text{off-springs}}^{(g+1)}$. With samples ordered, the algorithm updates $\boldsymbol{\mu}_{\text{CMA-ES}}^{(g+1)}$ as an average of $\kappa \leq \#\text{off-springs}$ selected points:
    \begin{equation}
    \label{Eq:Mean}
        \boldsymbol{\mu}_{\text{CMA-ES}}^{(g+1)} = \boldsymbol{\mu}_{\text{CMA-ES}}^{(g)} + \eta_{\boldsymbol{\mu}_{\text{CMA-ES}}} \sum_{i=1}^{\kappa} w_{i} \left(\text{vec}_{i^{\star}}^{(g+1)} - \boldsymbol{\mu}_{\text{CMA-ES}}^{(g)}\right),
    \end{equation}
    with $\eta_{\boldsymbol{\mu}_{\text{CMA-ES}}} < 1$ being a learning rate, and $w_{i}\propto \kappa - i + 1$. In words, Equation~\ref{Eq:Mean} attempts to shift the distribution's mean closer to a weighted average of the best samples seen so far, which, in turn, can be reinterpreted as maximising a log-data-likelihood conditioned on $\boldsymbol{\mu}_{\text{CMA-ES}}^{(g)}$ as noted in~\citep{2016_Hansen}. 
    
    When it comes to $\sigma_{\text{CMA-ES}}^{(g)}$, a process of cumulative step-size adaptation (CSA) -- also referred to as path length control -- is applied to derive $\sigma_{\text{CMA-ES}}^{(g+1)}$. First, CSA computes an (isotropic) ``evolutionary path'' $\text{path}_{\sigma}$ using: 
    \begin{equation}
    \label{Eq:EvolPath}
        \text{path}_{\sigma} = (1- c_{\sigma} )\text{path}_{\sigma} + \sqrt{1 - (1-c_{\sigma})^{2}}\sqrt{\kappa_{w}}\boldsymbol{\Sigma}_{\text{CMA-ES}}^{(g), - \frac{1}{2}} \frac{\boldsymbol{\mu}_{\text{CMA-ES}}^{(g+1)} - \boldsymbol{\mu}_{\text{CMA-ES}}^{(g)}}{\sigma_{\text{CMA-ES}}^{(g)}}, 
    \end{equation}
    where $c_{\sigma}$ is a constant typically set to $d/3$, and $\kappa_{w}$ is a variance-related constant abiding by $1 \leq \kappa_{w}\leq \kappa$. Given Equation~\ref{Eq:EvolPath}, CSA now updates $\sigma_{\text{CMA-ES}}^{(g+1)}$ by executing\footnote{It is worth noting that the update of $\sigma_{\text{CMA-ES}}^{(g+1)}$ requires the computation of $\mathbb{E}[||\mathcal{N}(0,1)||]$. Such an expectation can be approximated using a Gamma distribution as shown in~\citep{CMAES}}: 
    \begin{equation*}
        \sigma_{\text{CMA-ES}}^{(g+1)} = \sigma_{\text{CMA-ES}}^{(g)}\exp\left(\frac{c_{\sigma}}{d_{\sigma}}\left(\frac{||\text{path}_{\sigma}||}{\mathbb{E}[||\mathcal{N}(0,1)||]} - 1\right)\right), \ \ \text{with $d_{\sigma}$ being a damping value.}
    \end{equation*}
    Similarly, $\boldsymbol{\Sigma}_{\text{CMA-ES}}^{(g)}$ is adapted by following a two-step process, where an (anisotropic) evolutionary path, $\text{path}_{\boldsymbol{\Sigma}_{\text{CMA-ES}}}$, is used to in $\boldsymbol{\Sigma}_{\text{CMA-ES}}^{(g+1)}$ as follows: 
    \begin{align*}
        \text{path}_{\boldsymbol{\Sigma}_{\text{CMA-ES}}} &=  (1 - c_{\boldsymbol{\Sigma}_{\text{CMA-ES}}}) \text{path}_{\boldsymbol{\Sigma}_{\text{CMA-ES}}} + \one_{[0, \eta \sqrt{d}]}(||\text{path}_{\sigma}||)\sqrt{1 - (1 - c_{\boldsymbol{\Sigma}_{\text{CMA-ES}}})^{2}}\sqrt{\kappa_w} \\
        & \hspace{26em} \frac{\boldsymbol{\mu}_{\text{CMA-ES}}^{(g+1)} - \boldsymbol{\mu}_{\text{CMA-ES}}^{(g)}}{\sigma_{\text{CMA-ES}}^{(g)}} \\ 
        \boldsymbol{\Sigma}_{\text{CMA-ES}}^{(g+1 )} & = \gamma  \boldsymbol{\Sigma}_{\text{CMA-ES}}^{(g)} + c_1 \text{path}_{\boldsymbol{\Sigma}_{\text{CMA-ES}}} \text{path}^{\mathsf{T}}_{\boldsymbol{\Sigma}_{\text{CMA-ES}}}, \\
        & \hspace{5em} + \eta_{\boldsymbol{\Sigma}_{\text{CMA-ES}}} \sum_{i=1}^{\kappa} w_{i} \left(\frac{\text{vec}_{i^{\star}}^{(g+1)} - \boldsymbol{\mu}_{\text{CMA-ES}}^{(g)}}{\sigma_{\text{CMA-ES}}^{(g)}}\right) \left(\frac{\text{vec}_{i^{\star}}^{(g+1)} - \boldsymbol{\mu}_{\text{CMA-ES}}^{(g)}}{\sigma_{\text{CMA-ES}}^{(g)}}\right)^{\mathsf{T}}, 
    \end{align*}
    where $\gamma$ is a discount factor, $c_{\boldsymbol{\Sigma}_{\text{CMA-ES}}}$, $c_1$, and $\eta_{\boldsymbol{\Sigma}_{\text{CMA-ES}}}$ are tuneable hyperparameters. Finally, we used $\one_{[0, \eta\sqrt{d}]}(\cdot)$ to denote the indicator function with $\eta$ typically set to $\approx 1.5$. 
    
    \subsection{DE:} \label{App:DE}
    In differential evolution (DE)~\citep{pymoo}, a new set of input probes is generated from a previous population via component-wise mutation. The initial population $\mathcal{D}^{(0)} =  \{\text{vec}(\textbf{x})^{(0)}_1, \ldots, \text{vec}(\textbf{x})^{(0)}_{\#\text{population}} \}$ is given as a collection of $K$ vectors, where each $\text{vec}(\textbf{x})^{(0)}_j\in\mathbb{R}^{dq}$. Each vector $\text{vec}(\textbf{x})^{(g+1)}_{j}$ in the next population $\mathcal{D}^{(g+1)}$ undergoes a component-wise random mutation process consisting of three sequential steps. First, for each   $\text{vec}(\textbf{x})^{(g)}_j\in\mathcal{D}^{(g)}$, DE randomly picks a collection of three different candidates $\textbf{a},\textbf{b},\textbf{c}\in\mathbb{R}^{dq}$ that belong to the current population $\mathcal{D}^{(g)}$. These candidates will play the role of building blocks for a component-wise mutation process generating a candidate  $\mathcal{C} \in\mathbb{R}^{dq}$ for the next population. In the second step, DE randomly picks a component $l\in[1,\ldots, dq]$ of $\text{vec}(\textbf{x})^{(g)}_j$ which will be deterministically mutated with others undergoing a mutation with some fixed probability $p_{\text{mutation}}$:
    \begin{align*}
        \left[\mathcal{C} \right]_l = [\textbf{a}]_l + \text{F}([\textbf{b}]_l - [\textbf{c}]_j), \ \ \ \text{ and } \ \ \ \left[\mathcal{C} \right]_i = \left\{
     \begin{array}{ll}
      [\textbf{a}]_i + \text{F}([\textbf{b}]_i - [\textbf{c}]_i), & \text{w. p. } p_{\text{mutation}}\\
      \left[\text{vec}(\textbf{x})^{(g)}_j\right]_i & \text{w. p. } 1 - p_{\text{mutation}}
     \end{array}
        \right.  
    \end{align*}
    where $\text{F}\in[0,2]$ is a  scaling mutation parameter, and $[\textbf{v}]_i$ is used to denote the $i^{\text{th}}$ component of vector $\textbf{v}$. In the last step the algorithm makes a choice on whether to add $\mathcal{C}$ to the new population based on the acquisition function value information. In case the mutated vector  achieves a better solution than $\text{vec}(\textbf{x})^{(g)}_j$, then  $\mathcal{C}$ is added to a the new population $\mathcal{D}^{(g+1)}$, otherwise $\text{vec}(\textbf{x})^{(g)}_j$ is preserved. After the algorithm terminates, DE reports the best solution out of all constructed populations $\cup_{g\ge 0} \mathcal{D}^{(g)}$. 
    
    \section{First-Order Optimisers for ERM-BO}\label{App:FirstOrder}
     First-order optimisation techniques rely on gradient information to compute updates of $\textbf{x}$. They are iterative in nature, running for a total of $T$ iterations and executing a variant of the following rule at each step: 
    \begin{equation}
    \label{Eq:General}
        \textbf{x}_{1:q, t+1} = \delta_t \textbf{x}_{1:q, t} + \eta_{t}\frac{\boldsymbol{\phi}_{t}^{(1)}\left(\overline{\nabla \alpha(\textbf{x}_{1:q, 0}| \mathcal{D}_i)}, \dots, \overline{\nabla \alpha(\textbf{x}_{1:q, t}|\mathcal{D}_i)}, \left\{\beta_{k}^{(1)}\right\}_{k=0}^{t}\right)}{\boldsymbol{\phi}_{t}^{(2)}\left(\overline{\nabla \alpha(\textbf{x}_{1:q, 0}|\mathcal{D}_i)}^{2}, \dots, \overline{\nabla \alpha(\textbf{x}_{1:q, t}|\mathcal{D}_i)}^{2}, \left\{\beta_{k}^{(2)}\right\}_{k=0}^{t}, \epsilon\right)} \ \ \text{(Generalised update),}
    \end{equation}
    where $\delta_{t}$ is a weighting that depends on the class of algorithm used, $\eta_t$ is a (typically) decaying learning rate, $\boldsymbol{\phi}_{t}^{(1)}(\cdot)$ and $\boldsymbol{\phi}_{t}^{(2)}(\cdot)$ are history-dependent mappings that differ between algorithms with the ratio between them computed element-wise. $\left\{\beta_{k}^{(1)}\right\}_{k=0}^{t}$ and $\left\{\beta_{k}^{(2)}\right\}_{k=0}^{t}$ are history weighting parameters, and $\epsilon$ is a small positive constant used to avoid division by zero. Additionally, $\overline{\nabla \alpha(\textbf{x}_{1:q, 0}| \mathcal{D}_i)}, \dots, \overline{\nabla \alpha(\textbf{x}_{1:q, t}|\mathcal{D}_i)}$ represent sub-sampled gradient estimators that are acquired using Monte Carlo samples of $\textbf{z} \sim \mathcal{N}(0, \textbf{I})$. It is also worth noting that differentiating through the $\max$ operator that appears in all acquisitions can be performed either using sub-gradients or by propagating through the max value of the corresponding vector.  
    
    \subsection{SGA:} \label{App:SGA}
    Stochastic gradient ascent (SGA) is a cornerstone of the optimisation algorithm literature~\citep{RobbMonr51}, simply using gradients to ascend the objective function. Though it requires a large number of iterations to converge, recent studies demonstrate that stochastic gradients~\citep{wilson2018marginal} exhibit better generalisation capabilities when compared to other methods in machine learning applications. We can attain SGA's update from Equation~\ref{Eq:General} by setting constant weightings $\delta_{1} = \dots =  \delta_{T} =1$, $ \left\{\beta_{k}^{(1)}\right\}_{k=0}^{t} = \emptyset$,   $ \left\{\beta_{k}^{(2)}\right\}_{k=0}^{t} = \emptyset$, and defining $\boldsymbol{\phi}_{t}^{(1)}(\cdot)$ and $\boldsymbol{\phi}_{t}^{(2)}(\cdot)$ as: 
    \begin{align*}
        &\boldsymbol{\phi}_{t}^{(1)}\left(\overline{\nabla \alpha(\textbf{x}_{1:q, 0}|\mathcal{D}_i)}, \dots, \overline{\nabla \alpha(\textbf{x}_{1:q, t}|\mathcal{D}_i)} \right) = \overline{\nabla \alpha(\textbf{x}_{1:q, t}|\mathcal{D}_i)}, \\\nonumber
        &\boldsymbol{\phi}_{t}^{(2)}\left(\overline{\nabla \alpha(\textbf{x}_{1:q, 0}|\mathcal{D}_i)}^{2}, \dots, \overline{\nabla \alpha(\textbf{x}_{1:q, t}|\mathcal{D}_i)}^{2} \right) = \textbf{1}_{dq}.
    \end{align*}
    
    \subsection{AdaGrad:} \label{App:AdaGrad}
    In adaptive gradients (AdaGrad), SGA is modified so as to exhibit per-parameter learning rates~\citep{Duchi_2011_adagrad}. Intuitively, AdaGrad increases learning rates for sparse parameters and decreases them for denser ones. Such a strategy has been shown to be successful in settings where the data is sparse, and where sparse parameters convey more information (e.g., natural language processing~\citep{GloveNLP} and image recognition tasks~\citep{IR}). AdaGrad's update can also be extracted from Equation~\ref{Eq:General} by choosing $\delta_{1} = \dots  = \delta_{T} =1$, $ \left\{\beta_{k}^{(1)}\right\}_{k=0}^{t} = \emptyset$,   $ \left\{\beta_{k}^{(2)}\right\}_{k=0}^{t} = \emptyset$, and: \begin{align*}
        \boldsymbol{\phi}_{t}^{(1)}\left(\overline{\nabla \alpha(\textbf{x}_{1:q, 0}|\mathcal{D}_i)}, \dots, \overline{\nabla \alpha(\textbf{x}_{1:q, t}|\mathcal{D}_i)} \right) &= \overline{\nabla \alpha(\textbf{x}_{1:q, t}|\mathcal{D}_i)}, \\  \boldsymbol{\phi}_{t}^{(2)}\left(\overline{\nabla \alpha(\textbf{x}_{1:q, 0}|\mathcal{D}_i)}^{2}, \dots, \overline{\nabla \alpha(\textbf{x}_{1:q, t}|\mathcal{D}_i)}^{2}, \epsilon \right) &= \sqrt{\sum_{k=0}^{t} \overline{\nabla \alpha(\textbf{x}_{1:q, k}|\mathcal{D}_i)}^{2} +\epsilon} \ .
    \end{align*}
    
    \subsection{RMSprop:} \label{App:RMSProp}
    In root mean-square propagation (RMSprop), learning rates are also adapted to each of the parameters. Here, the idea is to divide the learning rate for a parameter by a running average of the magnitudes of recent gradients for that specific parameter~\citep{Tieleman2012}. RMSprop has enjoyed considerable success in machine learning~\citep{khosla2020supervised, wu2020pelican}. To arrive at its update rule, we set $\delta_{1} = \dots = \delta_{T} =1$, $\left\{\beta_{k}^{(1)}\right\}_{k=0}^{t} = \emptyset$, $\beta_{1}^{(2)}= \dots = \beta_{T}^{(2)} = \gamma$ with $\gamma$ denoting a forgetting factor. Furthermore, a constant learning rate $\eta$ is typically adopted in RMSprop, i.e., $\eta_{1} = \dots = \eta_{T} = \eta$, and $\boldsymbol{\phi}_{t}^{(1)}(\cdot)$ and $\boldsymbol{\phi}_{t}^{(2)}(\cdot)$ defined as: 
    \begin{align*}
        \boldsymbol{\phi}_{t}^{(1)}\left(\overline{\nabla \alpha(\textbf{x}_{1:q, 0}|\mathcal{D}_i)}, \dots, \overline{\nabla \alpha(\textbf{x}_{1:q, t}|\mathcal{D}_i)} \right) &= \overline{\nabla \alpha(\textbf{x}_{1:q, t}|\mathcal{D}_i)}, \\  \boldsymbol{\phi}_{t}^{(2)}\left(\overline{\nabla \alpha(\textbf{x}_{1:q, 0}|\mathcal{D}_i)}^{2}, \dots, \overline{\nabla \alpha(\textbf{x}_{1:q, t}|\mathcal{D}_i)}^{2}, \gamma, \epsilon \right) &= \sqrt{(1-\gamma) \sum_{k=0}^{t} \gamma^{k} \overline{\nabla \alpha(\textbf{x}_{1:q, k}|\mathcal{D}_i)}^{2} +\epsilon} \ .
    \end{align*}
     
    \subsection{Adam:} \label{App:ADAM} 
    Adam ~\citep{ADAM} is one of the most successful and widely-used algorithms in machine learning applications. The method computes individual adaptive learning rates for
    different parameters from estimates of the first and second moments of the gradients. In terms of Equation~\ref{Eq:General}, we can derive Adam's update as a special case using the following settings: 1) constant weightings $\delta_{1} = \dots =  \delta_{T} =1$, $ \beta_{1}^{(1)} = \dots = \beta_{T}^{(1)} = \beta_{1}$,  $ \beta_{1}^{(2)} = \dots = \beta_{T}^{(2)} = \beta_{2}$, and 2) $\boldsymbol{\phi}_{t}^{(1)}$ and $\boldsymbol{\phi}_{t}^{(2)}$ defined as: 
    \begin{align*}
        \boldsymbol{\phi}_{t}^{(1)}\left(\overline{\nabla \alpha(\textbf{x}_{1:q, 0}|\mathcal{D}_i)}, \dots, \overline{\nabla \alpha(\textbf{x}_{1:q, t}|\mathcal{D}_i)}, \beta_1 \right) & = \frac{1 - \beta_{1}}{1 - \beta_{1}^{t}} \sum_{k=0}^{t} \beta_{1}^{k}\overline{\nabla \alpha(\textbf{x}_{1:q, t-k}|\mathcal{D}_i)}, \\ 
        \boldsymbol{\phi}_{t}^{(2)}\left(\overline{\nabla \alpha(\textbf{x}_{1:q, 0}|\mathcal{D}_i)}^{2}, \dots, \overline{\nabla \alpha(\textbf{x}_{1:q, t}|\mathcal{D}_i)}^{2}, \beta_2, \epsilon \right) &= \sqrt{\frac{1 - \beta_{2}}{1 - \beta_{2}^{t}} \sum_{k=0}^{t} \beta_{2}^{k}\overline{\nabla \alpha(\textbf{x}_{1:q, t-k}|\mathcal{D}_i)}^{2}} + \epsilon.
    \end{align*}
    
    \subsection{AdaDelta:} \label{App:AdaDelta}
    The AdaDelta algorithm can be viewed as a robust extension of the AdaGrad method \citep{zeiler_2012_adadelta}. AdaDelta adapts learning rates based on a moving window of gradient updates. This window-based modification is implemented in an efficient manner by recursively defining the sum of the gradients as a decaying average of all past squared gradients. Following the general update rule introduced in Equation \ref{Eq:General}, AdaDelta can be formulated by setting $\delta_1 = \dots =  \delta_T = 1$,   $\beta^{[1]}_1 = \epsilon$, $\beta^{[1]}_2 = \eta$,   $\beta^{[1]}_3 = \dots =  \beta^{[1]}_T = \beta^{[2]}_1 = \dots = \beta^{[2]}_T = \gamma$, $\eta_1 = \dots =  \eta_T = \eta$, and 
    \begin{align*}
        &\boldsymbol{\phi}_{t}^{(1)}\left(\overline{\nabla \alpha(\textbf{x}_{1:q, 0}|\mathcal{D}_i)}, \dots, \overline{\nabla \alpha(\textbf{x}_{1:q, t}|\mathcal{D}_i)}, \gamma, \eta, \epsilon \right) = \\\nonumber &\overline{\nabla \alpha(\textbf{x}_{1:q, t}|\mathcal{D}_i)}\sqrt{\sum_{k=0}^{t-1}\frac{\gamma^k\overline{\nabla \alpha(\textbf{x}_{1:q, t-k-1}|\mathcal{D}_i)}^2}{\sum_{j=0}^{t-k-1}\gamma^j\overline{\nabla \alpha(\textbf{x}_{1:q, t-k-1-j}|\mathcal{D}_i)}^2+\frac{\epsilon}{(1-\gamma)}} + \frac{\epsilon}{\eta^2}},\\\nonumber
        &\boldsymbol{\phi}_{t}^{(2)}\left(\overline{\nabla \alpha(\textbf{x}_{1:q, 0}|\mathcal{D}_i)}^{2}, \dots, \overline{\nabla \alpha(\textbf{x}_{1:q, t}|\mathcal{D}_i)}^{2}, \gamma, \epsilon \right) = \sqrt{(1-\gamma)\sum_{k=0}^t\gamma^k\overline{\nabla \alpha(\textbf{x}_{1:q, t-k}|\mathcal{D}_i)}^2 + \epsilon}.
    \end{align*}
    
    \subsection{RProp:} \label{App:Rprop} 
    To overcome the inherent disadvantages of pure gradient descent/ascent techniques in terms of tuning the learning rate, \cite{riedmiller1993direct} propose RProp, an algorithm that takes into account only the sign of the corresponding partial derivative value. In terms of Equation \ref{Eq:General}, RProp can be defined by choosing  $\delta_{1} = \dots =  \delta_{T} =1$, $ \left\{\beta_{k}^{(1)}\right\}_{k=0}^{t} = \emptyset$,   $ \left\{\beta_{k}^{(2)}\right\}_{k=0}^{t} = \emptyset$, and  $\boldsymbol{\phi}_{t}^{(1)}(\cdot)$ and $\boldsymbol{\phi}_{t}^{(2)}(\cdot)$ as:  
    \begin{align*}
        &\boldsymbol{\phi}_{t}^{(1)}\left(\overline{\nabla \alpha(\textbf{x}_{1:q, 0}|\mathcal{D}_i)}, \dots, \overline{\nabla \alpha(\textbf{x}_{1:q, t}|\mathcal{D}_i)} \right) = \text{sign}\left(\overline{\nabla \alpha(\textbf{x}_{1:q, t}|\mathcal{D}_i)}\right), \\\nonumber
        &\boldsymbol{\phi}_{t}^{(2)}\left(\overline{\nabla \alpha(\textbf{x}_{1:q, 0}|\mathcal{D}_i)}^{2}, \dots, \overline{\nabla \alpha(\textbf{x}_{1:q, t}|\mathcal{D}_i)}^{2} \right) = \textbf{1}_{dq}.
    \end{align*}
    
    \subsection{AdamW:} \label{App:AdamW} 
    ~\cite{loshchilov_2019_adamw} propose a variation of Adam optimisation algorithm with decoupled weight decay regularisation to improve its  generalisation properties. AdamW can be written in the form of  Equation \ref{Eq:General} by specifying $\delta_t = (1 - \lambda\eta_t)$, $\beta^{[1]}_1 = \dots =  \beta^{[1]}_T = \beta_1$, $\beta^{[2]}_1 = \dots =  \beta^{[2]}_T = \beta_2$, and
    \begin{align*}
        &\boldsymbol{\phi}_{t}^{(1)}\left(\overline{\nabla \alpha(\textbf{x}_{1:q, 0}|\mathcal{D}_i)}, \dots, \overline{\nabla \alpha(\textbf{x}_{1:q, t}|\mathcal{D}_i)}, \beta_1 \right) = \frac{(1 - \beta_1)\sum_{k=0}^t\beta^k_1\overline{\nabla \alpha(\textbf{x}_{1:q, t-k}|\mathcal{D}_i)}}{1- \beta^t_1},\\\nonumber
        &\boldsymbol{\phi}_{t}^{(2)}\left(\overline{\nabla \alpha(\textbf{x}_{1:q, 0}|\mathcal{D}_i)}^{2}, \dots, \overline{\nabla \alpha(\textbf{x}_{1:q, t}|\mathcal{D}_i)}^{2},\beta_2, \epsilon \right) = \sqrt{\frac{(1 - \beta_2)\sum_{k=0}^t\beta^k_2\overline{\nabla \alpha(\textbf{x}_{1:q, t-k}|\mathcal{D}_i)}^2}{1- \beta^t_2}} + \epsilon.
    \end{align*}
    
    \subsection{AdamOs:} \label{App:AdamOS}
    To isolate the effect of our compositional reformulation, we consider a variation of the standard Adam optimiser with the parameter setup adopted from its compositional counterpart CAdam. In terms of Equation \ref{Eq:General}, AdamOs can be formulated by setting  $\delta_1 = \dots =  \delta_T = 1$,  $\beta^{[1]}_t = \mathcal{O}(\mu^t)$, $\beta^{[2]}_t = 1 - \frac{(1 - \mathcal{O}(\mu^t))^2}{t^{\eta_{\gamma}}}$, $\eta_t = \mathcal{O}\left(\frac{\sqrt{1 - \beta^{[2]}_t}}{(1 - \mathcal{O}(\mu^{t}))t^{\eta_{\eta}}}\right)$, and 
    \begin{align*}
        &\boldsymbol{\phi}_{t}^{(1)}\left(\overline{\nabla \alpha(\textbf{x}_{1:q, 0}|\mathcal{D}_i)}, \dots, \overline{\nabla \alpha(\textbf{x}_{1:q, t}|\mathcal{D}_i)}, \beta_1 \right) = \frac{1 - \beta_{1}}{1 - \beta_{1}^{t}} \sum_{k=0}^{t} \beta_{1}^{k}\overline{\nabla \alpha(\textbf{x}_{1:q, t-k}|\mathcal{D}_i)},\\\nonumber
        &\boldsymbol{\phi}_{t}^{(2)}\left(\overline{\nabla \alpha(\textbf{x}_{1:q, 0}|\mathcal{D}_i)}^{2}, \dots, \overline{\nabla \alpha(\textbf{x}_{1:q, t}|\mathcal{D}_i)}^{2},\beta_2, \epsilon \right) = \sqrt{\frac{1 - \beta_{2}}{1 - \beta_{2}^{t}} \sum_{k=0}^{t} \beta_{2}^{k}\overline{\nabla \alpha(\textbf{x}_{1:q, t-k}|\mathcal{D}_i)}^{2}} + \epsilon.
    \end{align*}

    \section{Second-Order Optimisers in ERM-BO:}\label{App:SecondOrder}
    Second-order optimisation methods along with gradients utilise second-order information of the objective function, typically\footnote{An alternative is the Fischer Information Matrix ~\citep{amari2007methods} used in the natural gradient decent update equation ~\citep{amari2012differential}.} encoded in the Hessian matrix $\nabla^2\alpha(\cdot|\mathcal{D}_{i})$. The general iterative update for second-order methods is given by:
    \begin{align*}
        \textbf{x}_{1:q,t+1} = \textbf{x}_{1:q,t} - \eta_t\left[\overline{\nabla^2\alpha(\textbf{x}_{1:q,t}|\mathcal{D}_{i})}\right]^{-1}\overline{\nabla \alpha(\textbf{x}_{1:q,t}|\mathcal{D}_{i})}\ \ \ \ \  \text{(Generalised update),}
    \end{align*}
    where $\overline{\nabla^2\alpha(\textbf{x}_{1:q,t}|\mathcal{D}_{i})}$ is an approximation for the Hessian matrix evaluated at a current iterate $\textbf{x}_{1:q,t}$. This approximation is needed due to the size of the real Hessian matrix (in our case $\overline{\nabla^2\alpha(\textbf{x}_{1:q,t}|\mathcal{D}_{i})}\in\mathbb{R}^{dq\times dq}$) as well as the necessity to compute its inverse at each iteration of the above generalised update.
    
    The BFGS algorithm \citep{BFGSone} and its memory-efficient version \citep{byrd1995limited} are the most commonly-used second-order techniques for high-dimensional, non-convex optimisation and are based on the Sherman-Morison formulae for recursive computation of the approximated Hessian inverse:
    \begin{align*}
        &\left[\overline{\nabla^2\alpha(\textbf{x}_{1:q,t}|\mathcal{D}_{i})}\right]^{-1} = \left[\textbf{I} - \frac{\textbf{s}_{t}\textbf{h}^{\mathsf{T}}_{t}}{\textbf{h}^{\mathsf{T}}_{t}\textbf{s}_t}\right]\left[\overline{\nabla^2\alpha(\textbf{x}_{1:q,t-1}|\mathcal{D}_{i})}\right]^{-1}\left[\textbf{I} - \frac{\textbf{h}_{t}\boldsymbol{s}^{\mathsf{T}}_{t}}{\textbf{h}^{\mathsf{T}}_{t}\textbf{s}_{t}}\right] + \frac{\textbf{s}_t\textbf{s}^{\mathsf{T}}_t}{\textbf{h}^{\mathsf{T}}_t\textbf{s}_t}, 
    \end{align*}
    where $\left[\overline{\nabla^2\alpha(\textbf{x}_{1:q,0}|\mathcal{D}_{i})}\right]^{-1} = \textbf{I}$ and curvature pairs  $\textbf{h}_{t} = \overline{\nabla\alpha(\textbf{x}_{1:q,t}|\mathcal{D}_{i})} - \overline{\nabla\alpha(\textbf{x}_{1:q,t-1}|\mathcal{D}_{i})}$, $\textbf{s}_t = \textbf{x}_{1:q,t} - \textbf{x}_{1:q,t-1}$. The recursive expression is beneficial for two reasons: 1) it admits computation of the Hessian inverse approximation while avoiding the inversion of large matrices and 2) it is formulated in terms of curvature pairs $\textbf{s}_t, \textbf{y}_t$ and hence permits computation of the descent direction efficiently with respect to both time and memory.

    \section{First-Order Compositional Optimisers}\label{App:FirstOrder_Compos}
    As discussed in Section \ref{Sec: Section_Comp_form}, first-order compositional methods depend on a stochastic approximation of the gradient of a compositional function $\alpha^{(\text{Comp})}(\textbf{x}_{1:q}|\mathcal{D}_i) = f(\mathbb{E}_{\omega}[\textbf{g}_{\omega}(\textbf{x}_{1:q})])$ given by:
    \begin{align*}
        &\overline{\nabla_{\text{vec}(\textbf{x}_{1:q})}\alpha^{(\text{Comp})}(\textbf{x}_{1:q}|\mathcal{D}_{i})} = \left[\frac{1}{K_2}\sum_{m=1}^{K_2}\nabla_{\text{vec}(\textbf{x}_{1:q})}\textbf{g}_{\omega_m}(\textbf{x}_{1:q})\right]^{\mathsf{T}}\nabla_{\boldsymbol{\zeta}}f(\boldsymbol{\zeta})
    \end{align*}
    where $\textbf{y}$ is an iterative auxiliary variable introduced to approximate the expectation of the inner mapping $\mathbb{E}_{\omega}[\textbf{g}_{\omega}(\textbf{x}_{1:q})]$ in a momentum-based fashion. Generalised update rules for first-order compositional optimisers are iterative in nature and have the following form:    
    \begin{align*}
        &\underline{\textbf{Main variable update: }}\\\nonumber
        &\textbf{x}_{1:q, t+1} =  \textbf{x}_{1:q, t} + \eta_{t}\frac{\boldsymbol{\phi}_{t}^{(1)}\left(\left\{\overline{\nabla_{\text{vec}(\textbf{x}_{1:q})}\alpha^{(\text{Comp})}(\textbf{x}_{1:q, k}, \boldsymbol{\zeta}_k|\mathcal{D}_{i})}\right\}^{t}_{k=0}, \left\{\gamma_{k}^{(1)}\right\}_{k=0}^{t}\right)}{\boldsymbol{\phi}_{t}^{(2)}\left(\left\{\overline{\nabla_{\text{vec}(\textbf{x}_{1:q})}\alpha^{(\text{Comp})}(\textbf{x}_{1:q, k}, \boldsymbol{\zeta}_k|\mathcal{D}_{i})}^2\right\}^{t}_{k=0}, \left\{\gamma_{k}^{(2)}\right\}_{k=0}^{t}, \epsilon\right)},
    \end{align*}
    \hspace{0.9cm}$\underline{\textbf{The second  auxiliary variable update:  }}$
    \begin{align*}
        \textbf{u}_{t+1} =  \boldsymbol{\phi}^{(3)}_{t+1}\left(\textbf{x}_{1:q, 0}, \dots,  \textbf{x}_{1:q, t+1}, \{\beta_k\}^{t}_{k=0}\right),
    \end{align*}
    \hspace{0.9cm}$\underline{\textbf{The first  auxiliary variable update:  }}$
    \begin{align*}
        \boldsymbol{\zeta}_{t+1} = \boldsymbol{\phi}^{(4)}_{t+1}\left(\overline{\textbf{g}(\textbf{u}_1)}, \dots,  \overline{\textbf{g}(\textbf{u}_{t+1})},   \{\beta_k\}^{t}_{k=0},\boldsymbol{\zeta}_0, \textbf{u}_0\right).
    \end{align*}
    where $\overline{\textbf{g}(\textbf{u})} = \frac{1}{K_1}\sum_{m=1}^{K_1}\textbf{g}_{\omega_m}(\textbf{u})$ is a Monte Carlo approximation of $\mathbb{E}_{\omega}[\textbf{g}_{\omega}(\textbf{x}_{1:q})]$. Next, we show how different first-order compositional optimisers can be formulated in terms of the above generalised iterative updates.
    
    \subsection{SCGA: } \label{App:SCGA}
    Stochastic Compositional Gradient Ascent ~\citep{2017_Wang} is the first algorithm which focuses on a quasi-gradient computation and a momentum-based approximation of the inner mapping $\mathbb{E}_{\omega}[\textbf{g}_{\omega}(\textbf{x})]$. Following the generalised update scheme, SCGA can be accessed by setting:
    \begin{align*}
        &\boldsymbol{\phi}_{t}^{(1)}\left(\left\{\overline{\nabla_{\text{vec}(\textbf{x}_{1:q})}\alpha^{(\text{Comp})}(\textbf{x}_{1:q, k},\boldsymbol{\zeta}_k|\mathcal{D}_{i})}\right\}^{t}_{k=0}, \left\{\gamma^{(1)}_k\right\}_{k=0}^{t}\right) = \overline{\nabla_{\text{vec}(\textbf{x}_{1:q})}\alpha^{(\text{Comp})}(\textbf{x}_{1:q, t},\boldsymbol{\zeta}_t|\mathcal{D}_{i})}, \\\nonumber
        &\boldsymbol{\phi}_{t}^{(2)}\left(\left\{\overline{\nabla_{\text{vec}(\textbf{x}_{1:q})}\alpha^{(\text{Comp})}(\textbf{x}_{1:q, k},\boldsymbol{\zeta}_k|\mathcal{D}_{i})}^2\right\}^{t}_{k=0}, \left\{\gamma_{k}^{(2)}\right\}_{k=0}^{t}, \epsilon\right) = \textbf{1}_{dq}, \\\nonumber
        &\boldsymbol{\phi}^{(3)}_{t}\left(\textbf{x}_{1:q, 0}, \dots,  \textbf{x}_{1:q, t}, \{\beta_k\}^{t-1}_{k=0}\right) = \textbf{x}_{1:q, t}, \\\nonumber
        &\boldsymbol{\phi}^{(4)}_t\left(\overline{\textbf{g}(\textbf{u}_1)}, \dots,  \overline{\textbf{g}(\textbf{u}_{t})},   \{\beta_k\}^{t-1}_{k=0},\boldsymbol{\zeta}_0, \textbf{u}_0\right) = \sum_{k=1}^{t}\beta_{k-1}\prod_{j=k}^{t-1}(1 - \beta_j)\overline{\textbf{g}(\textbf{u}_k)}.
    \end{align*}
    
    \subsection{ASCGA: } \label{App:ASCGA}
    \cite{2017_Wang} propose an accelerated stochastic compositional gradient algorithm by evaluating compositional gradients via two-timescale iteration updates. We can attain ASCGA from the generalised update equations by defining $\boldsymbol{\phi}^{(1)}_t$,  $\boldsymbol{\phi}^{(2)}_t$,  $\boldsymbol{\phi}^{(3)}_t$,  $\boldsymbol{\phi}^{(4)}_t$ as:  
    \begin{align*}
        &\boldsymbol{\phi}_{t}^{(1)}\left(\left\{\overline{\nabla_{\text{vec}(\textbf{x}_{1:q})}\alpha^{(\text{Comp})}(\textbf{x}_{1:q, k},\boldsymbol{\zeta}_k|\mathcal{D}_{i})}\right\}^{t}_{k=0}, \left\{\gamma^{(1)}_k\right\}_{k=0}^{t}\right) = \overline{\nabla_{\text{vec}(\textbf{x}_{1:q})}\alpha^{(\text{Comp})}(\textbf{x}_{1:q, t},\boldsymbol{\zeta}_t|\mathcal{D}_{i})}, \\\nonumber
        &\boldsymbol{\phi}_{t}^{(2)}\left(\left\{\overline{\nabla_{\text{vec}(\textbf{x}_{1:q})}\alpha^{(\text{Comp})}(\textbf{x}_{1:q, k},\boldsymbol{\zeta}_k|\mathcal{D}_{i})}^2\right\}^{t}_{k=0}, \left\{\gamma_{k}^{(2)}\right\}_{k=0}^{t}, \epsilon\right) = \textbf{1}_{dq},\\\nonumber
        &\boldsymbol{\phi}^{(3)}_{t}\left(\textbf{x}_{1:q, 0}, \dots,  \textbf{x}_{1:q, t}, \{\beta_k\}^{t-1}_{k=0}\right) = (1 - \beta^{-1}_{t-1})\textbf{x}_{1:q, t-1} + \beta^{-1}_{t-1}\textbf{x}_{1:q, t}, \\\nonumber
        &\boldsymbol{\phi}^{(4)}_t\left(\overline{\textbf{g}(\textbf{u}_1)}, \dots,  \overline{\textbf{g}(\textbf{u}_{t})},   \{\beta_k\}^{t-1}_{k=0},\boldsymbol{\zeta}_0, \textbf{u}_0\right) = \sum_{k=1}^{t}\beta_{k-1}\prod_{j=k}^{t-1}(1 - \beta_j)\overline{\textbf{g}(\textbf{u}_k)}.
    \end{align*}
    \subsection{CAdam:} \label{App:CADAM}
    As mentioned in the main body of the paper, one can recover CAdam ~\citep{tutunov2020cadam} by instantiating the above as follows:
    \begin{align*}
        &\boldsymbol{\phi}_{t}^{(1)}\left(\left\{\overline{\nabla_{\text{vec}(\textbf{x}_{1:q})}\alpha^{(\text{Comp})}(\textbf{x}_{1:q, k}, \boldsymbol{\zeta}_k|\mathcal{D}_{i})}\right\}^{t}_{k=0}, \left\{\gamma_{k}^{(1)}\right\}_{k=0}^{t}\right) =\\\nonumber
        &\hspace{5cm}\sum_{k=0}^t(1 - \gamma^{[1]}_k)\prod_{j=k+1}^{t}\gamma^{[1]}_j\overline{\nabla_{\text{vec}(\textbf{x}_{1:q})}\alpha^{(\text{Comp})}(\textbf{x}_{1:q, k}, \boldsymbol{\zeta}_k|\mathcal{D}_{i})}, \\\nonumber
        &\boldsymbol{\phi}_{t}^{(2)}\left(\left\{\overline{\nabla_{\text{vec}(\textbf{x}_{1:q})}\alpha^{(\text{Comp})}(\textbf{x}_{1:q, k}, \boldsymbol{\zeta}_k|\mathcal{D}_{i})}^2\right\}^{t}_{k=0}, \left\{\gamma_{k}^{(2)}\right\}_{k=0}^{t}, \epsilon\right) = \\\nonumber
        &\hspace{5cm}\sqrt{\sum_{k=0}^t(1 - \gamma^{[2]}_k)\prod_{j=k+1}^{t}\gamma^{[2]}_j\overline{\nabla_{\text{vec}(\textbf{x}_{1:q})}\alpha^{(\text{Comp})}(\textbf{x}_{1:q, k},\boldsymbol{\zeta}_k|\mathcal{D}_{i})}^2} + \epsilon,
    \end{align*}
    \begin{align*}
        &\boldsymbol{\phi}^{(3)}_{t}\left(\textbf{x}_{1:q, 0}, \dots,  \textbf{x}_{1:q, t}, \{\beta_k\}^{t-1}_{k=0}\right) = (1 - \beta^{-1}_{t-1})\textbf{x}_{1:q, t-1} + \beta^{-1}_{t-1}\textbf{x}_{1:q, t}, \\\nonumber
        &\boldsymbol{\phi}^{(4)}_t\left(\overline{\textbf{g}^{(\text{type})}(\textbf{u}_1)}, \dots,  \overline{\textbf{g}^{(\text{type})}(\textbf{u}_{t})},   \{\beta_k\}^{t-1}_{k=0},\boldsymbol{\zeta}_0, \textbf{u}_0\right) = \sum_{k=1}^{t}\beta_{k-1}\prod_{j=k}^{t-1}(1 - \beta_j)\overline{\textbf{g}^{(\text{type})}(\textbf{u}_k)}.
    \end{align*}
    \subsection{NASA: } \label{App:NASA}
    Nested Averaged Stochastic Approximation ~\citep{ghadimi2020single} is a single time-scale stochastic approximation algorithm whereby the problem is transformed to a high-dimensional space and together with the main variable $\textbf{x}$, the behaviour of the gradient of the compositional function $\nabla_{\text{vec}(\textbf{x}_{1:q})}\alpha^{(\text{Comp})}(\textbf{x}_{1:q}|\mathcal{D}_i)$ as well as the value of the inner mapping $\mathbb{E}_{\omega}[\textbf{g}_{\omega}(\textbf{x})]$ are studied. In terms of generalised update rules, the NASA algorithm can be formulated by the following setup:  
    \begin{align*}
        &\boldsymbol{\phi}_{t}^{(1)}\left(\left\{\overline{\nabla_{\text{vec}(\textbf{x}_{1:q})}\alpha^{(\text{Comp})}(\textbf{x}_{1:q, k},\boldsymbol{\zeta}_k|\mathcal{D}_{i})}\right\}^{t}_{k=0}, \left\{\rho\tau_k\right\}_{k=0}^{t}\right) = \\\nonumber
        &\hspace{5cm}\rho\sum_{k=0}^{t-1}\tau_{k-1}\prod_{j=k}^{t}(1 - \rho\tau_j)\overline{\nabla_{\text{vec}(\textbf{x}_{1:q})}\alpha^{(\text{Comp})}(\textbf{x}_{1:q, k},\boldsymbol{\zeta}_k|\mathcal{D}_{i})}, \\\nonumber
        &\boldsymbol{\phi}_{t}^{(2)}\left(\left\{\overline{\nabla_{\text{vec}(\textbf{x}_{1:q})}\alpha^{(\text{Comp})}(\textbf{x}_{1:q, k},\boldsymbol{\zeta}_k|\mathcal{D}_{i})}^2\right\}^{t}_{k=0}, \left\{\gamma_{k}^{(2)}\right\}_{k=0}^{t}, \epsilon\right) = \textbf{1}_{dq}, \\\nonumber
        &\boldsymbol{\phi}^{(3)}_{t}\left(\textbf{x}_{1:q, 0}, \dots,  \textbf{x}_{1:q, t}, \{\beta_k\}^{t-1}_{k=0}\right) = \textbf{x}_{1:q, t}, \\\nonumber
        &\boldsymbol{\phi}^{(4)}_t\left(\overline{\textbf{g}(\textbf{u}_1)}, \dots,  \overline{\textbf{g}(\textbf{u}_{t})},   \{\beta_k\}^{t-1}_{k=0},\boldsymbol{\zeta}_0, \textbf{u}_0\right) = \beta\sum_{k=1}^{t}\tau_{k-1}\prod_{j=k}^{t-1}(1 - \beta\tau_j)\overline{\textbf{g}(\textbf{u}_k)}.
    \end{align*}

    \subsection{Nested-MC: }\label{App:NestedMC}
    To emphasise the effect of a momentum-based update for the auxiliary variable $\textbf{y}$, we also consider a compositional variation of the Adam optimiser, where all involved expectation operators are  approximated by corresponding Monte Carlo estimates. In terms of the generalised update scheme, Nested-MC can be formulated as follows:
    \begin{align*}
        &\boldsymbol{\phi}_{t}^{(1)}\left(\left\{\overline{\nabla_{\text{vec}(\textbf{x}_{1:q})}\alpha^{(\text{Comp})}(\textbf{x}_{1:q, k},\boldsymbol{\zeta}_k|\mathcal{D}_{i})}\right\}^{t}_{k=0}, \beta_1\right) = \\\nonumber
        &\hspace{5cm}\frac{(1 - \beta_1)}{1 - \beta^t_1}\sum_{k=0}^t\beta^k_1\overline{\nabla_{\text{vec}(\textbf{x}_{1:q})}\alpha^{(\text{Comp})}(\textbf{x}_{1:q, t-k},\boldsymbol{\zeta}_{t-k}|\mathcal{D}_{i})}, \\\nonumber
        &\boldsymbol{\phi}_{t}^{(2)}\left(\left\{\overline{\nabla_{\text{vec}(\textbf{x}_{1:q})}\alpha^{(\text{Comp})}(\textbf{x}_{1:q, k},\boldsymbol{\zeta}_k|\mathcal{D}_{i})}^2\right\}^{t}_{k=0}, \beta_2, \epsilon\right) = \\\nonumber
        &\hspace{5cm}\sqrt{\frac{(1 - \beta_2)}{1 - \beta^t_2}\sum_{k=0}^t\beta^k_2\overline{\nabla_{\text{vec}(\textbf{x}_{1:q})}\alpha^{(\text{Comp})}(\textbf{x}_{1:q, t-k},\boldsymbol{\zeta}_{t-k}|\mathcal{D}_{i})}^2} + \epsilon,
    \end{align*}
    \begin{align*}
        &\boldsymbol{\phi}^{(3)}_{t}\left(\textbf{x}_{1:q, 0}, \dots,  \textbf{x}_{1:q, t}, \{\beta_k\}^{t-1}_{k=0}\right) = \textbf{x}_{1:q, t}, \\\nonumber
        &\boldsymbol{\phi}^{(4)}_t\left(\overline{\textbf{g}(\textbf{u}_1)}, \dots,  \overline{\textbf{g}(\textbf{u}_{t})},   \{\beta_k\}^{t-1}_{k=0},\boldsymbol{\zeta}_0, \textbf{u}_0\right) = \overline{\textbf{g}(\textbf{u}_t)}.
    \end{align*}
    
    \section{Memory-Efficient Adaptations for Compositional Optimisers}\label{App:MemoryEfficient}
    As described in Section \ref{Sec:Memory-Efficient}, the necessity of storing all $M$ samples of the reparameterisation random variables $\textbf{z}\sim\mathcal{N}(\textbf{0}, \textbf{I})$ makes compositional optimisers cumbersome with respect to memory capacity. For example, an inner mapping $\textbf{g}^{(\text{type})}(\textbf{x}_{1:q}) = \mathbb{E}_{\omega}\left[\textbf{g}^{(\text{type})}(\textbf{x}_{1:q})\right]\in\mathbb{R}^{q\times M}$, where $\text{type}\in\{\text{EI},\text{PI},\text{SR},\text{UCB}\}$ and each stochastic instance $\textbf{g}^{(\text{type})}_{\omega}(\textbf{x}_{1:q})$ is defined as: 
    \begin{align*}
        \textbf{g}^{(\text{type})}_{\omega}(\textbf{x}_{1:q}) &= [\textbf{0}_{q}, \dots, \textbf{v}^{(\text{type})}_{\omega}, \dots, \textbf{0}_{q}]\in\mathbb{R}^{q\times M}
    \end{align*}
    where each $\textbf{v}^{(\text{type})}_{\omega}\in\mathbb{R}^q$ is formulated in terms of an associated vector $\textbf{z}$ sampled uniformly from a fixed collection, as described in Section \ref{Sec: Section_Comp_form}. As a result, the construction of a Monte Carlo estimate for  $\textbf{g}^{(\text{type})}(\textbf{x}_{1:q})$ involves storing all $\{\textbf{z}_1, \dots, \textbf{z}_M\}$ and therefore gives rise to high memory consumption. In the memory-efficient adaptation however, we remedy this problem by sampling a set of reparameterisation random variables $\textbf{z}_1, \dots, \textbf{z}_K$ directly from a distribution $\mathcal{N}(\textbf{0},\textbf{I})$ rather then from a large fixed collection. As a result, a stochastic instance of the inner mapping can be written as a $q$ by $K$ matrix:
    \begin{align*}
        \overline{\textbf{g}^{(\text{type}), (\text{ME})}(\cdot)} = \left[\textbf{v}^{(\text{type})}_{\textbf{z}_1}(\cdot),\dots,\textbf{v}^{(\text{type})}_{\textbf{z}_K}(\cdot) \right]\in\mathbb{R}^{q\times K}
    \end{align*}
    where the $j^{th}$ column is defined via the associated $\textbf{v}^{(\text{type})}_{\textbf{z}_{j}}(\cdot)$ in an analogous fashion to Section \ref{Sec: Section_Comp_form}. This adjustment immediately allows us to compute stochastic estimates for the Jacobian $\overline{\nabla_{\text{vec}(\textbf{x}_{1:q})}\textbf{g}^{(\text{type}), (\text{ME})}(\cdot)} =  \nabla_{\text{vec}(\textbf{x}_{1:q})}\overline{\textbf{g}^{(\text{type}), (\text{ME})}(\cdot)}$ of the inner mapping in a memory-efficient manner. Finally, the gradient of the compositional objective $\alpha^{(\text{Comp})}(\star|\mathcal{D}_i)$ can be estimated as follows:
    \begin{align*}
        \overline{\nabla_{\text{vec}(\textbf{x}_{1:q})}\alpha^{(\text{Comp}), (\text{ME})}(\star,\ast|\mathcal{D}_i)} 
        &=\left[\overline{\nabla_{\text{vec}(\textbf{x}_{1:q})}\textbf{g}^{(\text{type}), (\text{ME})}(\star)}\right]^{\mathsf{T}}\nabla_{\boldsymbol{\zeta}}f^{(\text{type})}(\ast).
    \end{align*}
    where $\ast$ represents the value of the first auxiliary variable $\boldsymbol{\zeta}$ obtained via the
    exponentially-weighted average of estimates $\overline{\textbf{g}^{(\text{type}), (\text{ME})}(\cdot)}$ (see Section \ref{Sec: Section_Comp_form}). The generalised iterative update equations for memory-efficient compositional optimisers already have a familiar form:
    \begin{align*}
        &\underline{\textbf{Main variable update: }}\\\nonumber
        &\textbf{x}_{1:q, t+1} =  \textbf{x}_{1:q, t} + \eta_{t}\frac{\boldsymbol{\phi}_{t}^{(1)}\left(\left\{\overline{\nabla_{\text{vec}(\textbf{x}_{1:q})}\alpha^{(\text{Comp}),\text{(ME)}}(\textbf{x}_{1:q, k}, \boldsymbol{\zeta}_k|\mathcal{D}_{i})}\right\}^{t}_{k=0}, \left\{\gamma_{k}^{(1)}\right\}_{k=0}^{t}\right)}{\boldsymbol{\phi}_{t}^{(2)}\left(\left\{\overline{\nabla_{\text{vec}(\textbf{x}_{1:q})}\alpha^{(\text{Comp}),\text{(ME)}}(\textbf{x}_{1:q, k}, \boldsymbol{\zeta}_k|\mathcal{D}_{i})}^2\right\}^{t}_{k=0}, \left\{\gamma_{k}^{(2)}\right\}_{k=0}^{t}, \epsilon\right)},
    \end{align*}
    \hspace{0.5cm}$\underline{\textbf{The second  auxiliary variable update:  }}$
    \begin{align*}
        \textbf{u}_{t+1} =  \boldsymbol{\phi}^{(3)}_{t+1}\left(\textbf{x}_{1:q, 0}, \dots,  \textbf{x}_{1:q, t+1}, \{\beta_k\}^{t}_{k=0}\right),
    \end{align*}
    \hspace{0.5cm}$\underline{\textbf{The first  auxiliary variable update:  }}$
    \begin{align*}
        \boldsymbol{\zeta}_{t+1} = \boldsymbol{\phi}^{(4)}_{t+1}\left(\overline{\textbf{g}^{(\text{type}),\text{(ME)}}(\textbf{u}_1)}, \dots,  \overline{\textbf{g}^{(\text{type}),\text{(ME)}}(\textbf{u}_{t+1})},   \{\beta_k\}^{t}_{k=0},\boldsymbol{\zeta}_0, \textbf{u}_0\right).
    \end{align*}
    Next, we show how memory-efficient compositional optimisers can be formulated in terms of the above generalised iterative updates.
    
    \subsection{CAdam-ME} \label{App:CADAMME}
    A memory-efficient version of the CAdam optimiser in terms of the generalised update:
    \begin{align*}
        &\boldsymbol{\phi}_{t}^{(1)}\left(\left\{\overline{\nabla_{\text{vec}(\textbf{x}_{1:q})}\alpha^{(\text{Comp}),\text{(ME)}}(\textbf{x}_{1:q, k}, \boldsymbol{\zeta}_k|\mathcal{D}_{i})}\right\}^{t}_{k=0}, \left\{\gamma_{k}^{(1)}\right\}_{k=0}^{t}\right) =\\\nonumber
        &\hspace{5cm}\sum_{k=0}^t(1 - \gamma^{[1]}_k)\prod_{j=k+1}^{t}\gamma^{[1]}_j\overline{\nabla_{\text{vec}(\textbf{x}_{1:q})}\alpha^{(\text{Comp}),\text{(ME)}}(\textbf{x}_{1:q, k}, \boldsymbol{\zeta}_k|\mathcal{D}_{i})}, \\\nonumber
        &\boldsymbol{\phi}_{t}^{(2)}\left(\left\{\overline{\nabla_{\text{vec}(\textbf{x}_{1:q})}\alpha^{(\text{Comp}),\text{(ME)}}(\textbf{x}_{1:q, k}, \boldsymbol{\zeta}_k|\mathcal{D}_{i})}^2\right\}^{t}_{k=0}, \left\{\gamma_{k}^{(2)}\right\}_{k=0}^{t}, \epsilon\right) = \\\nonumber
        &\hspace{4cm}\sqrt{\sum_{k=0}^t(1 - \gamma^{[2]}_k)\prod_{j=k+1}^{t}\gamma^{[2]}_j\overline{\nabla_{\text{vec}(\textbf{x}_{1:q})}\alpha^{(\text{Comp}),\text{(ME)}}(\textbf{x}_{1:q, k},\boldsymbol{\zeta}_k|\mathcal{D}_{i})}^2} + \epsilon,\\\nonumber
        &\boldsymbol{\phi}^{(3)}_{t}\left(\textbf{x}_{1:q, 0}, \dots,  \textbf{x}_{1:q, t}, \{\beta_k\}^{t-1}_{k=0}\right) = (1 - \beta^{-1}_{t-1})\textbf{x}_{1:q, t-1} + \beta^{-1}_{t-1}\textbf{x}_{1:q, t}, \\\nonumber
        &\boldsymbol{\phi}^{(4)}_t\left(\left\{\overline{\textbf{g}^{(\text{type}),\text{(ME)}}(\textbf{u}_k)}\right\}^t_{k=1},   \{\beta_k\}^{t-1}_{k=0},\boldsymbol{\zeta}_0, \textbf{u}_0\right) =\sum_{k=1}^{t}\beta_{k-1}\prod_{j=k}^{t-1}(1 - \beta_j)\overline{\textbf{g}^{(\text{type}),\text{(ME)}}(\textbf{u}_k)}.
    \end{align*}
    \subsection{NASA-ME} \label{App:NASAME}
    The NASA algorithm also has a  memory-efficient adaptation:
    \begin{align*}
        &\boldsymbol{\phi}_{t}^{(1)}\left(\left\{\overline{\nabla_{\text{vec}(\textbf{x}_{1:q})}\alpha^{(\text{Comp}),\text{(ME)}}(\textbf{x}_{1:q, k},\boldsymbol{\zeta}_k|\mathcal{D}_{i})}\right\}^{t}_{k=0}, \left\{\rho\tau_k\right\}_{k=0}^{t}\right) = \\\nonumber
        &\hspace{5cm}\rho\sum_{k=0}^{t-1}\tau_{k-1}\prod_{j=k}^{t}(1 - \rho\tau_j)\overline{\nabla_{\text{vec}(\textbf{x}_{1:q})}\alpha^{(\text{Comp}),\text{(ME)}}(\textbf{x}_{1:q, k},\boldsymbol{\zeta}_k|\mathcal{D}_{i})}, \\\nonumber
        &\boldsymbol{\phi}_{t}^{(2)}\left(\left\{\overline{\nabla_{\text{vec}(\textbf{x}_{1:q})}\alpha^{(\text{Comp}),\text{(ME)}}(\textbf{x}_{1:q, k},\boldsymbol{\zeta}_k|\mathcal{D}_{i})}^2\right\}^{t}_{k=0}, \left\{\gamma_{k}^{(2)}\right\}_{k=0}^{t}, \epsilon\right) = \textbf{1}_{dq}, \\\nonumber
        &\boldsymbol{\phi}^{(3)}_{t}\left(\textbf{x}_{1:q, 0}, \dots,  \textbf{x}_{1:q, t}, \{\beta_k\}^{t-1}_{k=0}\right) = \textbf{x}_{1:q, t}, \\\nonumber
        &\boldsymbol{\phi}^{(4)}_t\left(\left\{\overline{\textbf{g}^{(\text{type}),\text{(ME)}}(\textbf{u}_k)}\right\}^{t}_{k=1},   \{\beta_k\}^{t-1}_{k=0},\boldsymbol{\zeta}_0, \textbf{u}_0\right) =\beta\sum_{k=1}^{t}\tau_{k-1}\prod_{j=k}^{t-1}(1 - \beta\tau_j)\overline{\textbf{g}^{(\text{type}),\text{(ME)}}(\textbf{u}_k)}.
    \end{align*}

    \subsection{Nested MC-ME} \label{App:NestedMCME}
    Finally, the Nested MC optimiser can also be converted to its memory-efficient form:
    \begin{align*}
        &\boldsymbol{\phi}_{t}^{(1)}\left(\left\{\overline{\nabla_{\text{vec}(\textbf{x}_{1:q})}\alpha^{(\text{Comp}),\text{(ME)}}(\textbf{x}_{1:q, k},\boldsymbol{\zeta}_k|\mathcal{D}_{i})}\right\}^{t}_{k=0}, \beta_1\right) = \\\nonumber
        &\hspace{5cm}\frac{(1 - \beta_1)}{1 - \beta^t_1}\sum_{k=0}^t\beta^k_1\overline{\nabla_{\text{vec}(\textbf{x}_{1:q})}\alpha^{(\text{Comp}),\text{(ME)}}(\textbf{x}_{1:q, t-k},\boldsymbol{\zeta}_{t-k}|\mathcal{D}_{i})},
    \end{align*}
    \begin{align*}
        &\boldsymbol{\phi}_{t}^{(2)}\left(\left\{\overline{\nabla_{\text{vec}(\textbf{x}_{1:q})}\alpha^{(\text{Comp}),\text{(ME)}}(\textbf{x}_{1:q, k},\boldsymbol{\zeta}_k|\mathcal{D}_{i})}^2\right\}^{t}_{k=0}, \beta_2, \epsilon\right) = \\\nonumber
        &\hspace{4cm}\sqrt{\frac{(1 - \beta_2)}{1 - \beta^t_2}\sum_{k=0}^t\beta^k_2\overline{\nabla_{\text{vec}(\textbf{x}_{1:q})}\alpha^{(\text{Comp}),\text{(ME)}}(\textbf{x}_{1:q, t-k},\boldsymbol{\zeta}_{t-k}|\mathcal{D}_{i})}^2} + \epsilon, \\\nonumber
        &\boldsymbol{\phi}^{(3)}_{t}\left(\textbf{x}_{1:q, 0}, \dots,  \textbf{x}_{1:q, t}, \{\beta_k\}^{t-1}_{k=0}\right) = \textbf{x}_{1:q, t}, \\\nonumber
        &\boldsymbol{\phi}^{(4)}_t\left(\left\{\overline{\textbf{g}^{(\text{type}),\text{(ME)}}(\textbf{u}_k)}\right\}^t_{k=1},   \{\beta_k\}^{t-1}_{k=0},\boldsymbol{\zeta}_0, \textbf{u}_0\right) = \overline{\textbf{g}^{(\text{type}),\text{(ME)}}(\textbf{u}_t)}.
    \end{align*}

    \section{Extended Results}
    \subsection{Synthetic tasks}
    
    We provide in Figures \ref{fig:immediate_log_regret_16D}, \ref{fig:immediate_log_regret_40D}, \ref{fig:immediate_log_regret_60D},
    \ref{fig:immediate_log_regret_80D} and \ref{fig:immediate_log_regret_120D} the evolution of immediate regrets obtained using each optimiser and acquisition function on synthetic black-box maximisation tasks in $16$, $40$, $60$, $80$ and $120$ dimensions. These results are summarised in Table \ref{tab:summary_perf}.
    
    \begin{figure}[!h]
      \centering
      \includegraphics[width=1\linewidth]{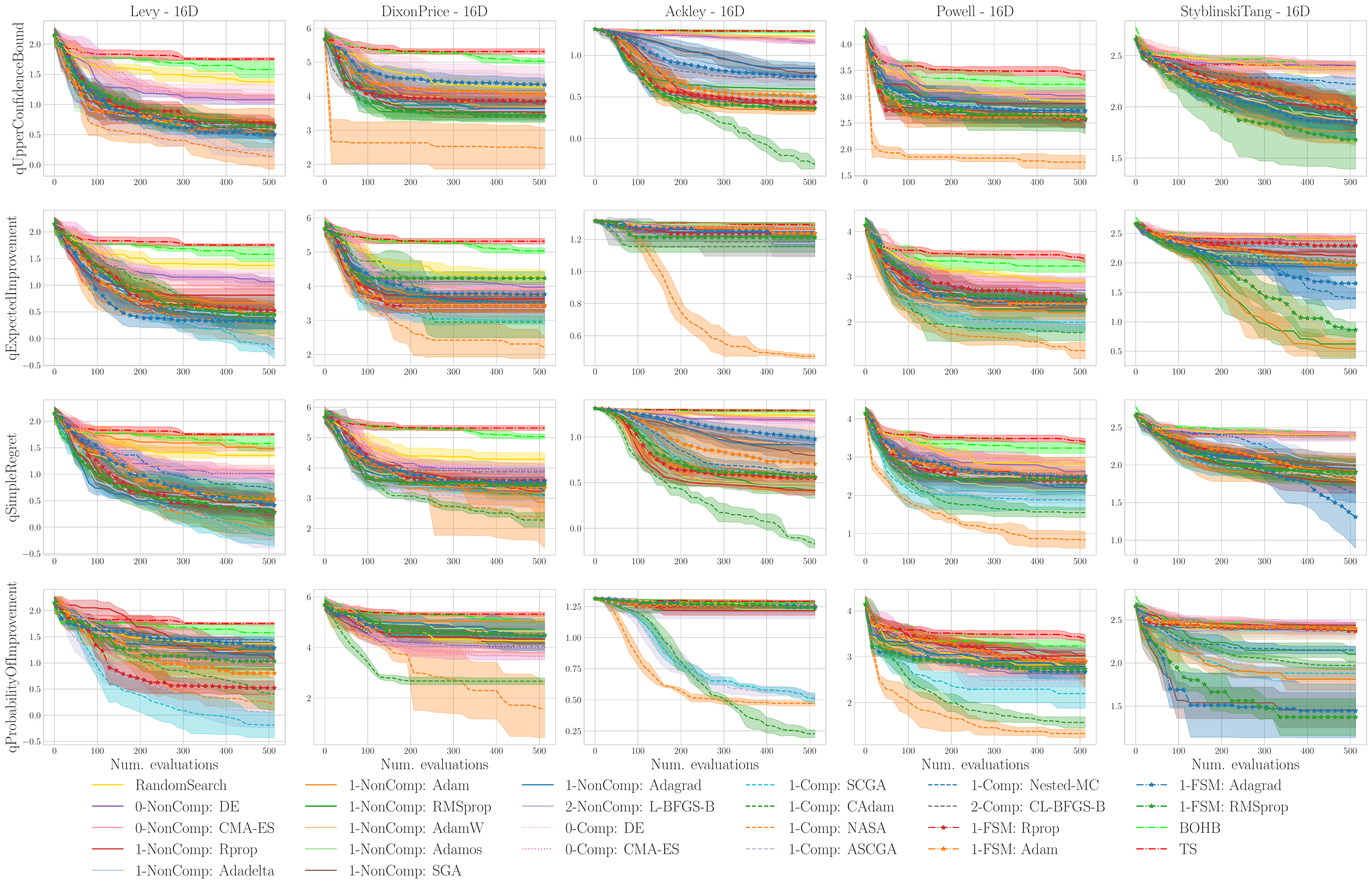}
      \caption{Evolution of immediate log-regret for various acquisition functions and optimisers on $16$D synthetic function maximisation tasks. Results of $490$ experiments are shown on this graph.}
          \label{fig:immediate_log_regret_16D}
\end{figure}

    \begin{figure}
      \centering
      \includegraphics[width=1\linewidth]{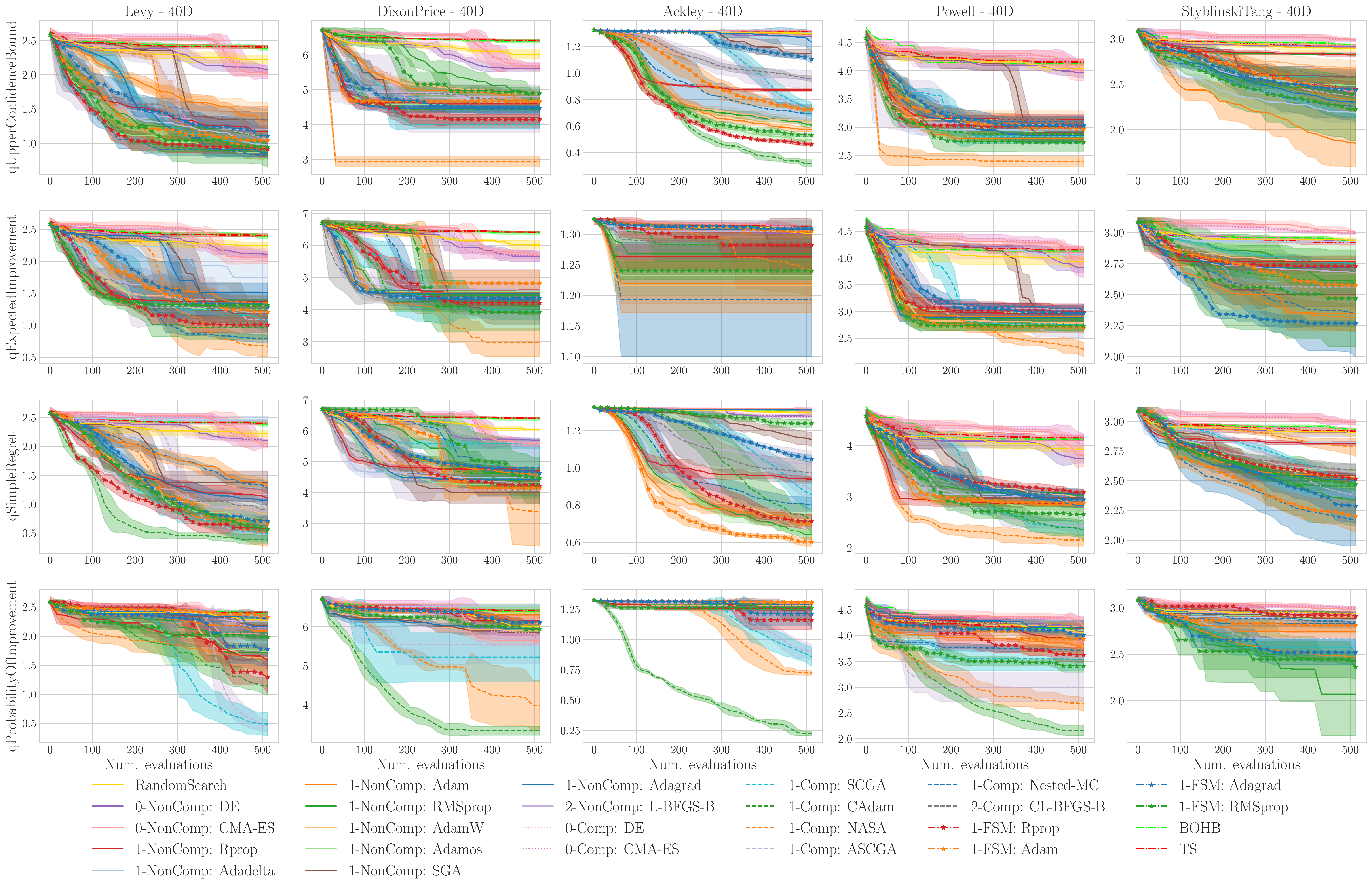}
      \caption{Evolution of immediate log-regret for various acquisition functions and optimisers on $40$D synthetic function maximisation tasks. Results of $490$ experiments are shown on this graph.}
         \label{fig:immediate_log_regret_40D}
 \end{figure}
    
    \begin{figure}
      \centering
      \includegraphics[width=1\linewidth]{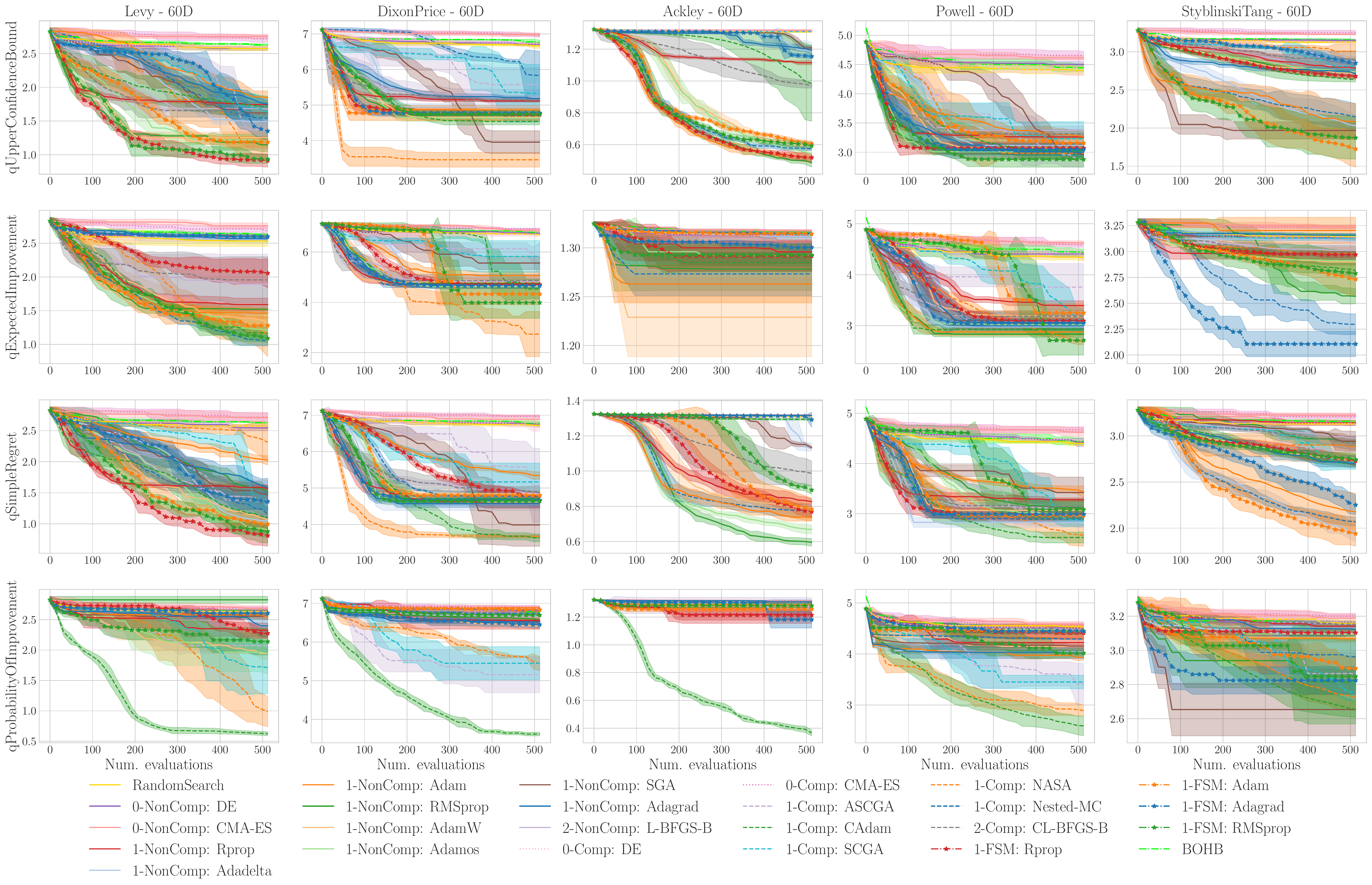}
      \caption{Evolution of immediate log-regret for various acquisition functions and optimisers on $60$D synthetic function maximisation tasks. Results of $485$ experiments are shown on this graph.}
        \label{fig:immediate_log_regret_60D}
  \end{figure}
    
    \begin{figure}
      \centering
      \includegraphics[width=1\linewidth]{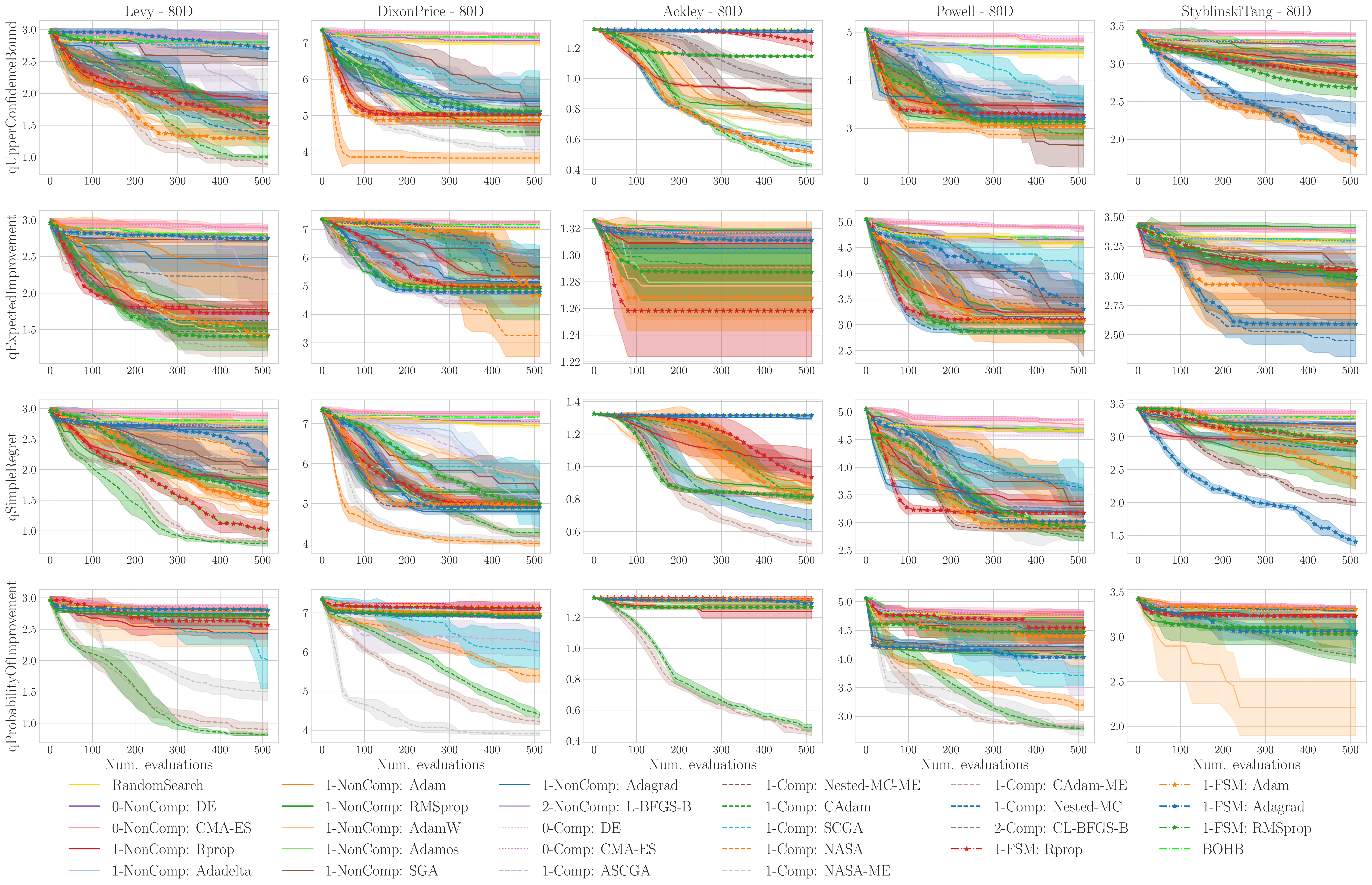}
      \caption{Evolution of immediate log-regret for various acquisition functions and optimisers on $80$D synthetic function maximisation tasks. Results of $545$ experiments are shown on this graph.}
        \label{fig:immediate_log_regret_80D}
  \end{figure}
    
    \begin{figure}
      \centering
      \includegraphics[width=1\linewidth]{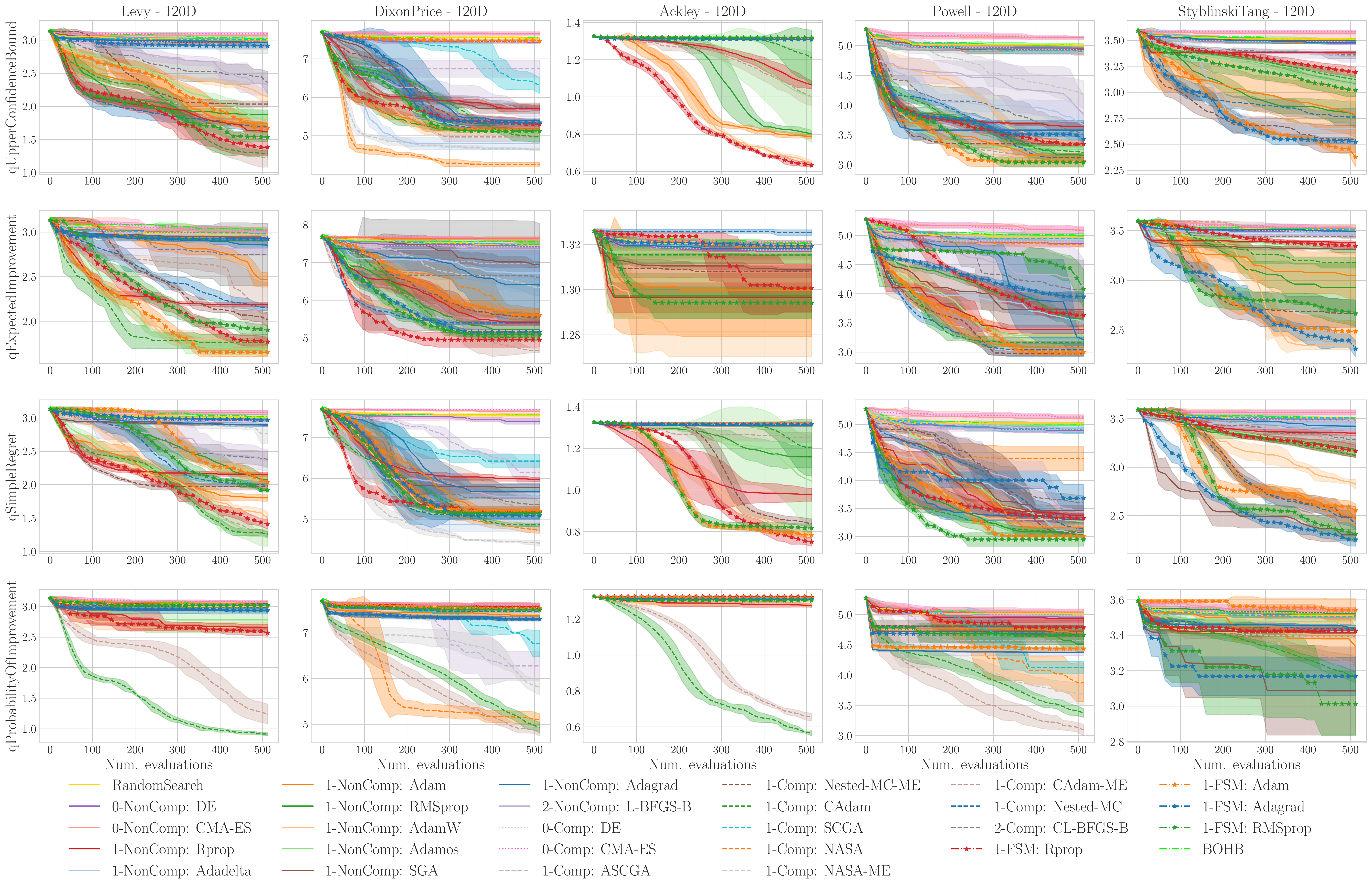}
      \caption{Evolution of immediate log-regret for various acquisition functions and optimisers on $120$D synthetic function maximisation tasks. Results of $545$ experiments are shown on this graph.}
            \label{fig:immediate_log_regret_120D}
    \end{figure}
    
    \begin{figure}
    \centering
      \includegraphics[width=.75\linewidth]{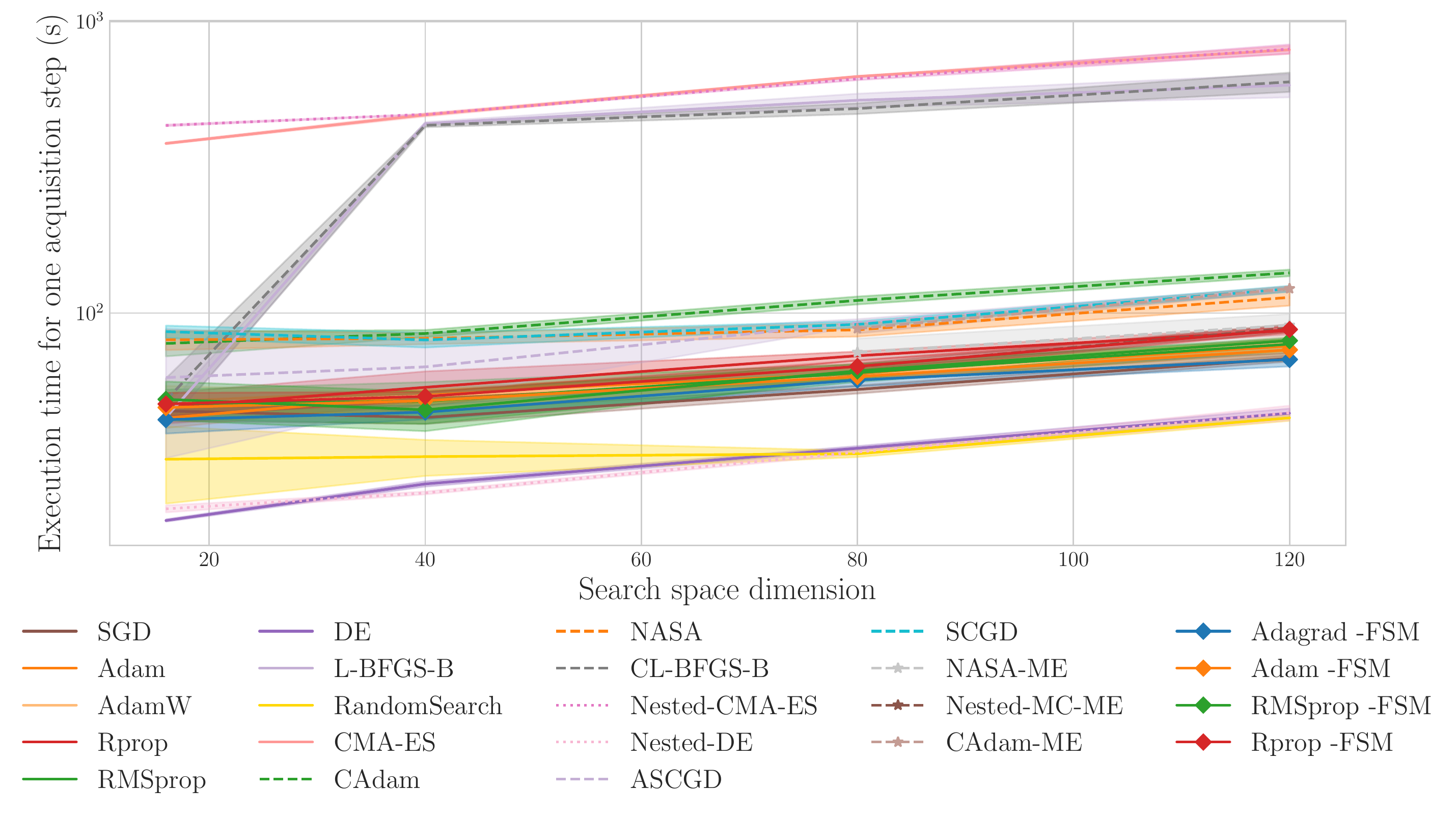}
      \caption{Execution time of UCB maximisation run on $4$ CPUs measured for each optimiser on $2$ synthetic black-box maximisation tasks in $16$, $40$, $80$ and $120$ dims, amounting to $184$ BO experiments.}
      \label{fig:ex-time-full}
    \end{figure}
    
    \subsection{Hyperparameter tuning tasks}

    {\renewcommand{\arraystretch}{0}
    \begin{figure}
    \begin{tabular}{ccccc}
    \subfloat{\includegraphics[width=0.19\columnwidth, trim={0 0.5cm 0 0.4cm}, clip]{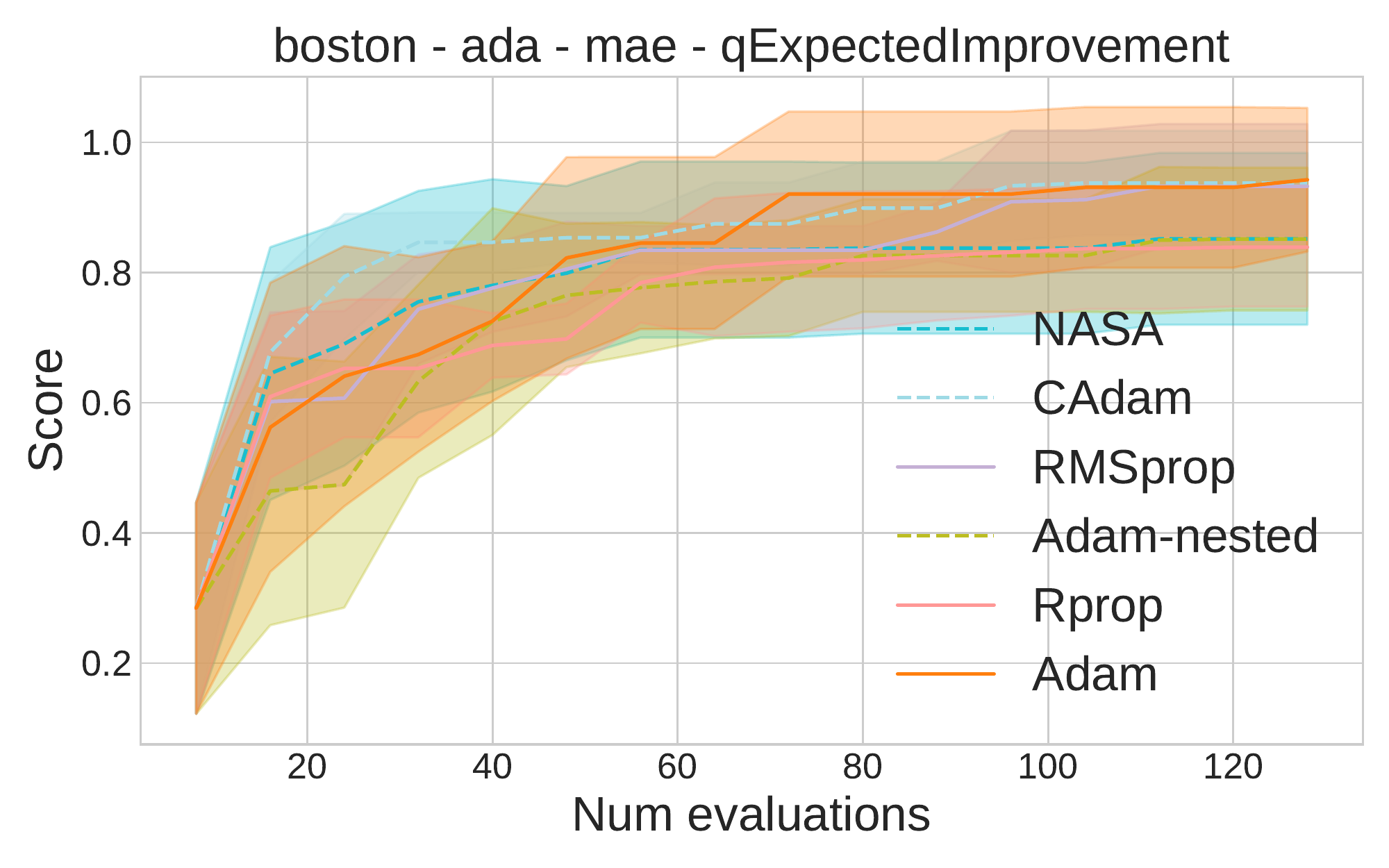}} &  \hspace{-0.5cm}
    \subfloat{\includegraphics[width=0.19\columnwidth, trim={0 0.5cm 0 0.4cm}, clip]{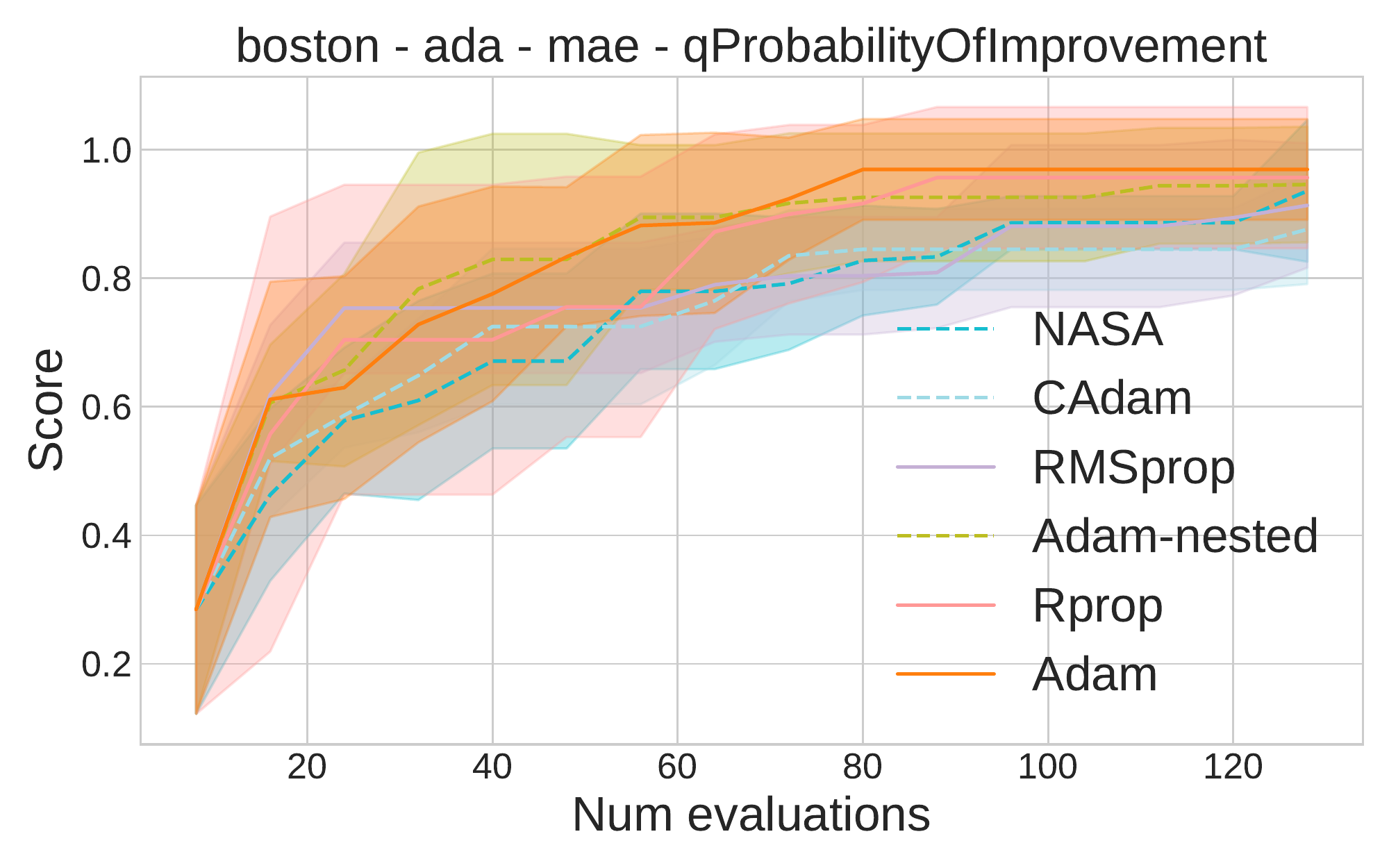}} & \hspace{-0.5cm}
    \subfloat{\includegraphics[width=0.19\columnwidth, trim={0 0.5cm 0 0.4cm}, clip]{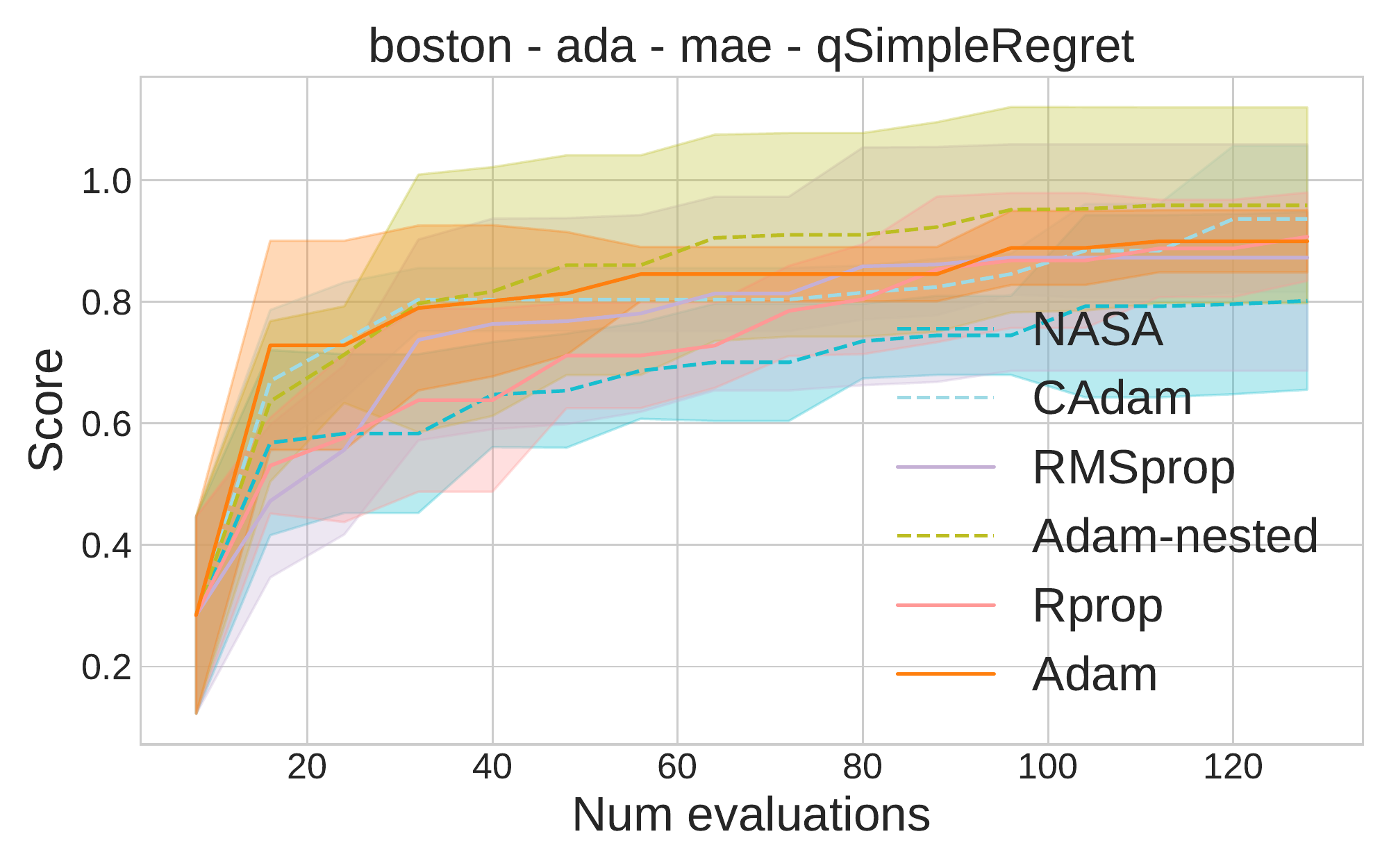}} &  \hspace{-0.5cm}
    \subfloat{\includegraphics[width=0.19\columnwidth, trim={0 0.5cm 0 0.4cm}, clip]{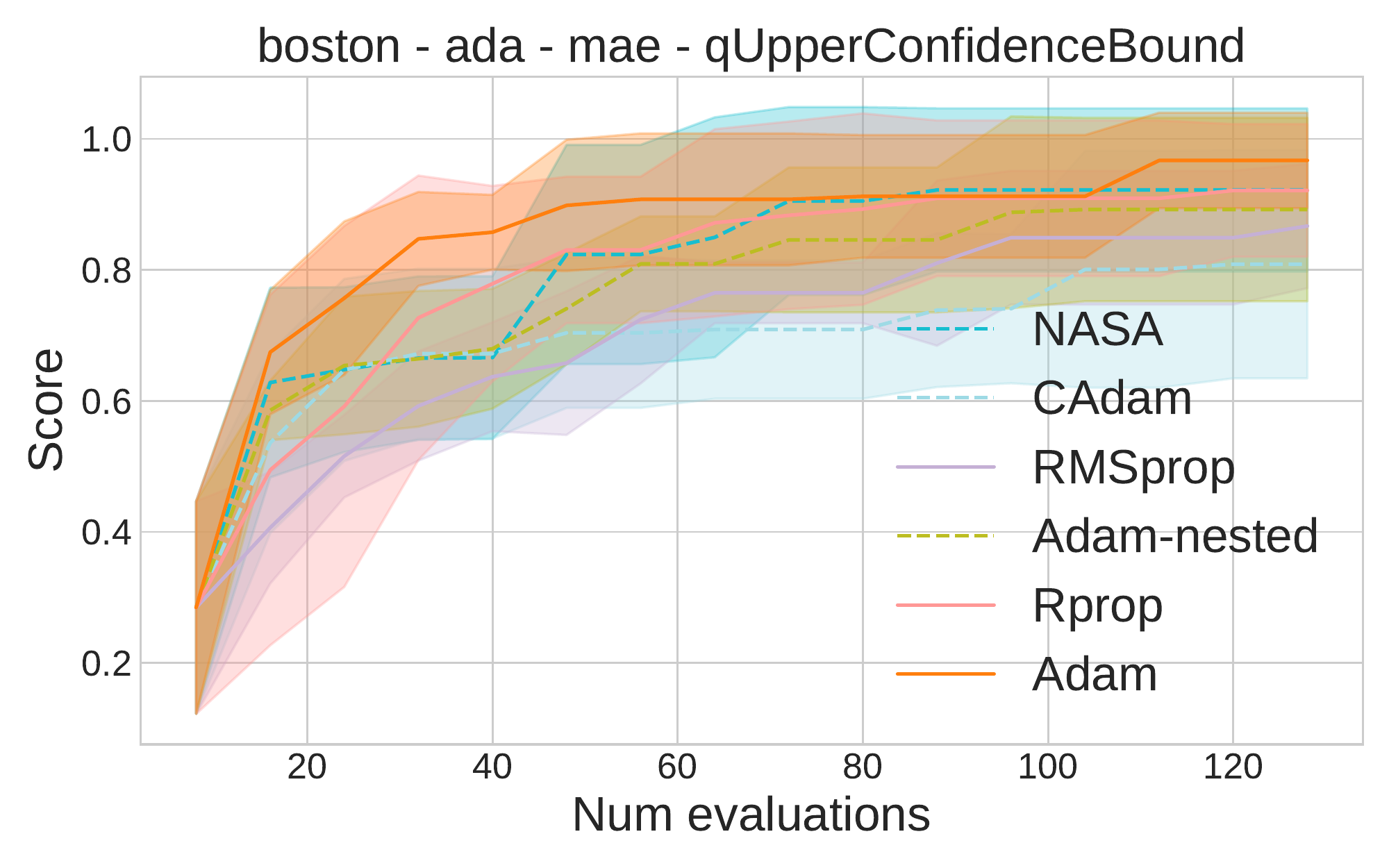}} & \hspace{-0.5cm}
    \subfloat{\includegraphics[width=0.19\columnwidth, trim={0 0.5cm 0 0.4cm}, clip]{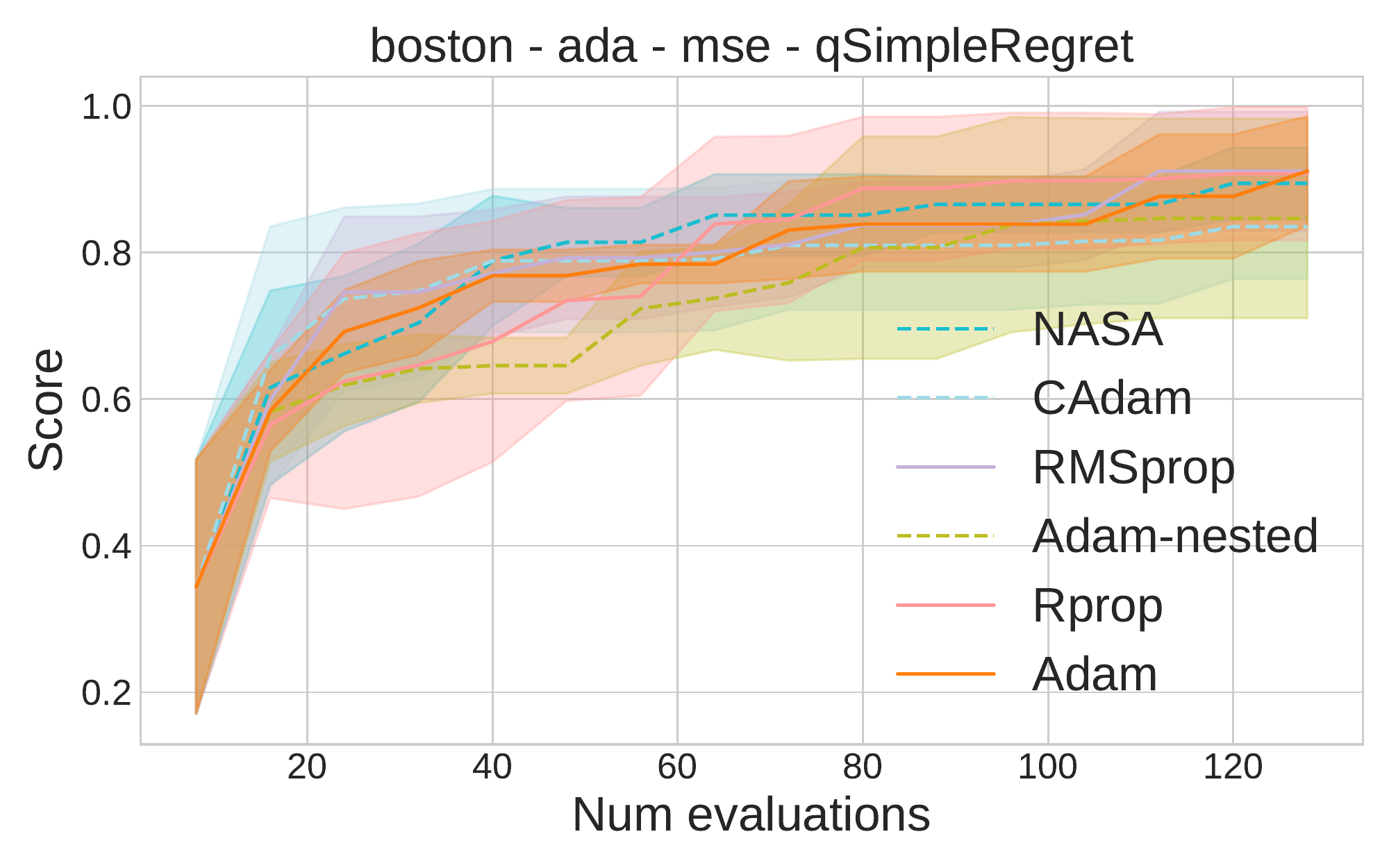}} \\  
    \subfloat{\includegraphics[width=0.19\columnwidth, trim={0 0.5cm 0 0.4cm}, clip]{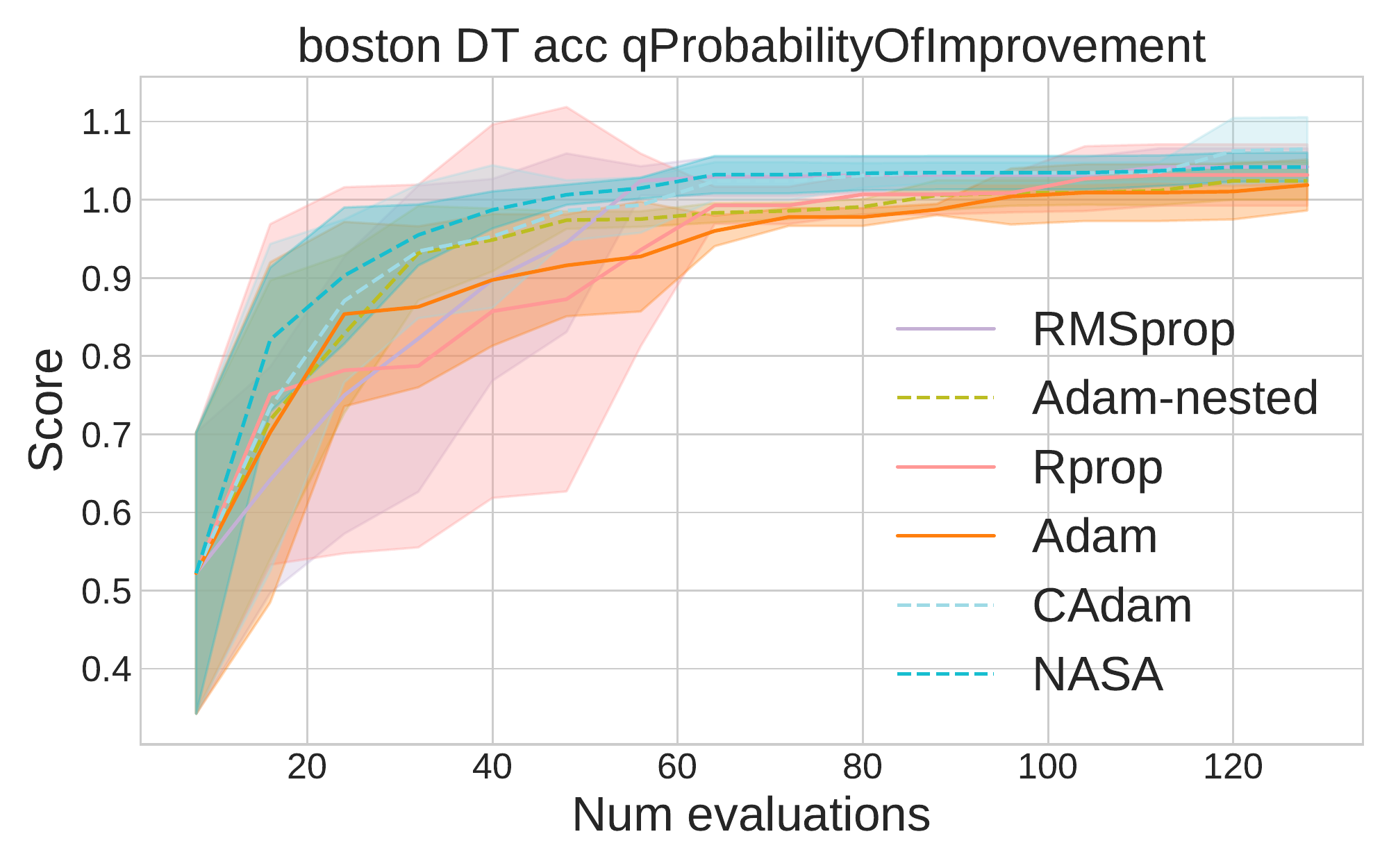}} &   \hspace{-0.5cm} 
    \subfloat{\includegraphics[width=0.19\columnwidth, trim={0 0.5cm 0 0.4cm}, clip]{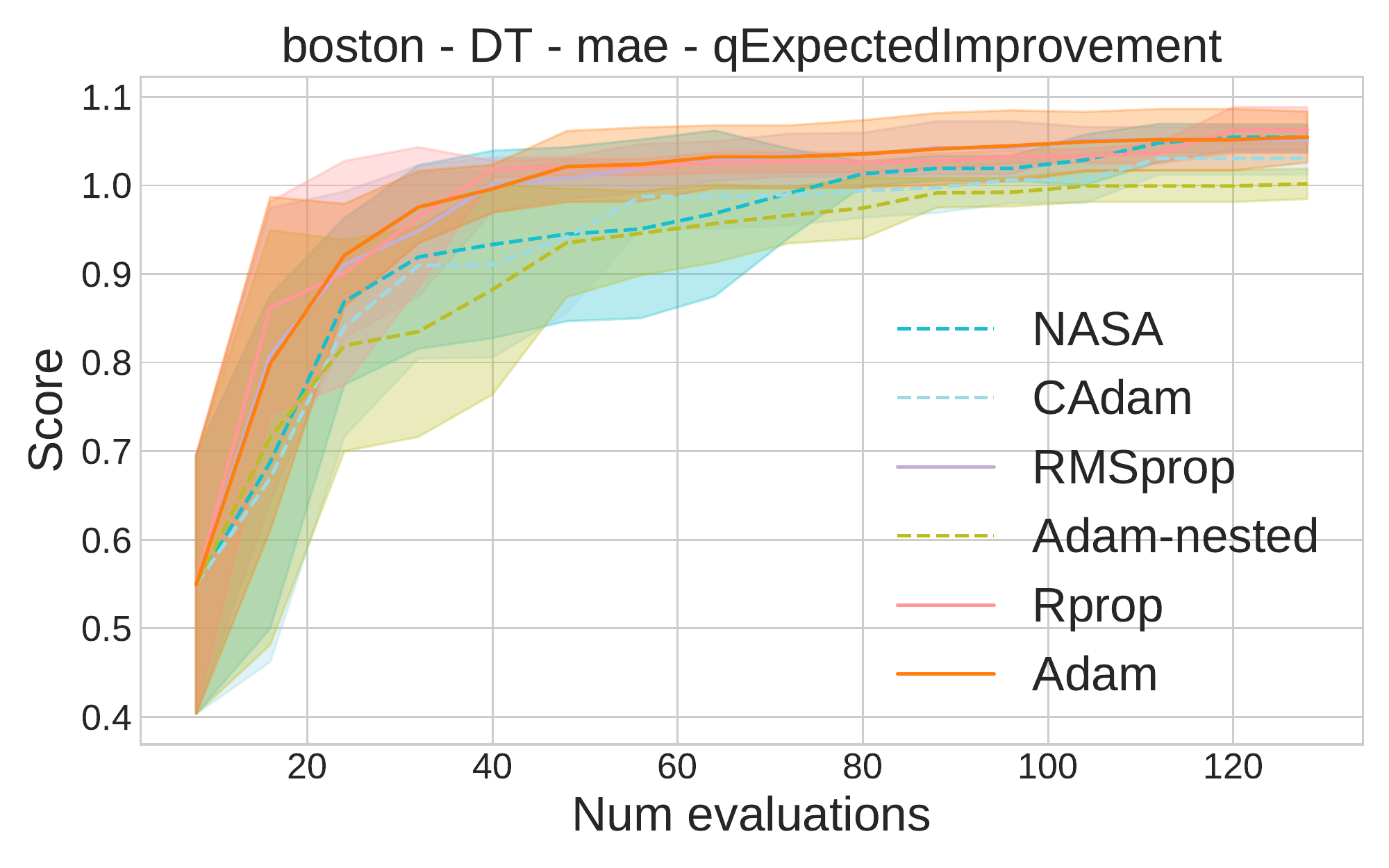}} & \hspace{-0.5cm}
    \subfloat{\includegraphics[width=0.19\columnwidth, trim={0 0.5cm 0 0.4cm}, clip]{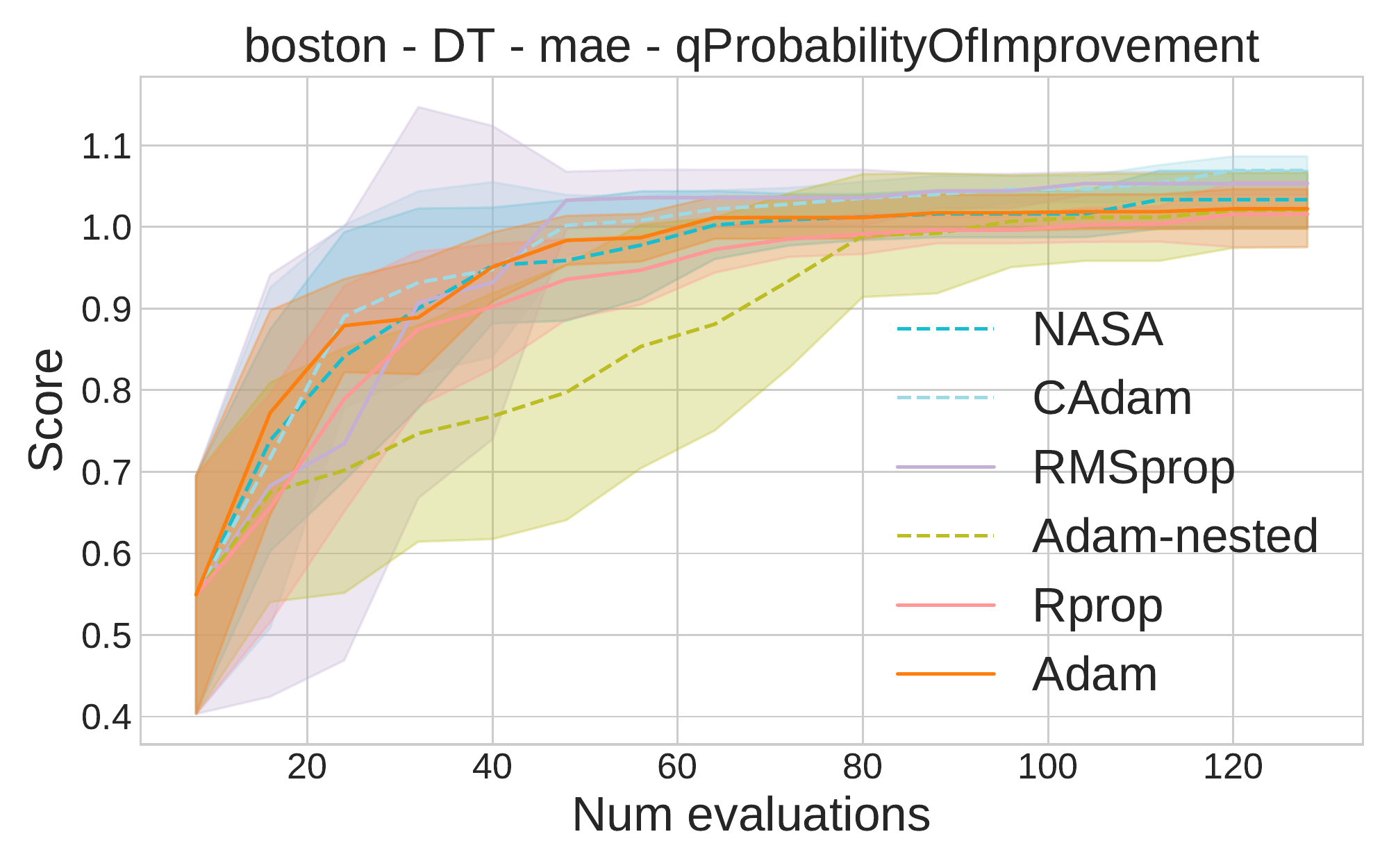}} &  \hspace{-0.5cm}
    \subfloat{\includegraphics[width=0.19\columnwidth, trim={0 0.5cm 0 0.4cm}, clip]{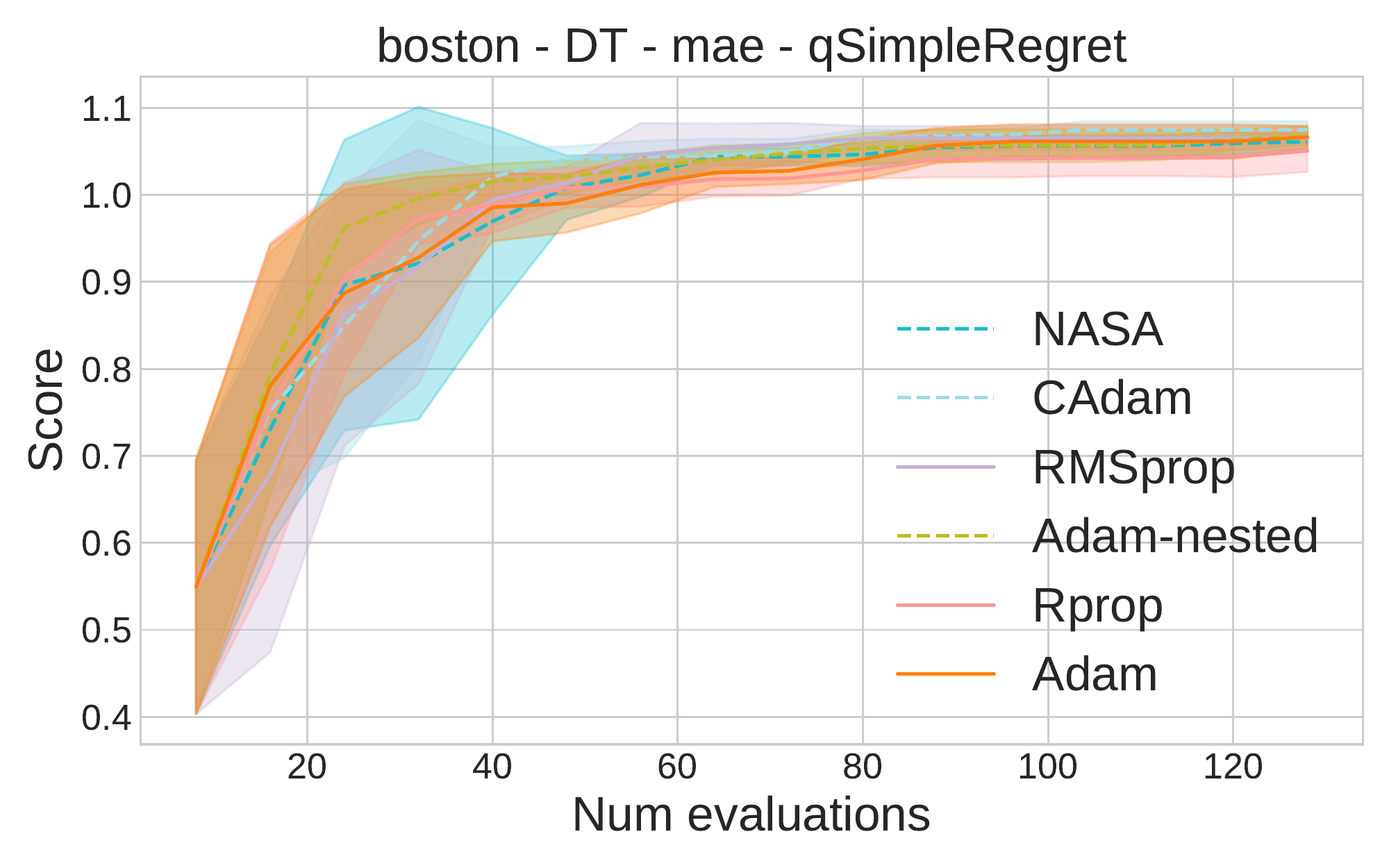}} & \hspace{-0.5cm}
    \subfloat{\includegraphics[width=0.19\columnwidth, trim={0 0.5cm 0 0.4cm}, clip]{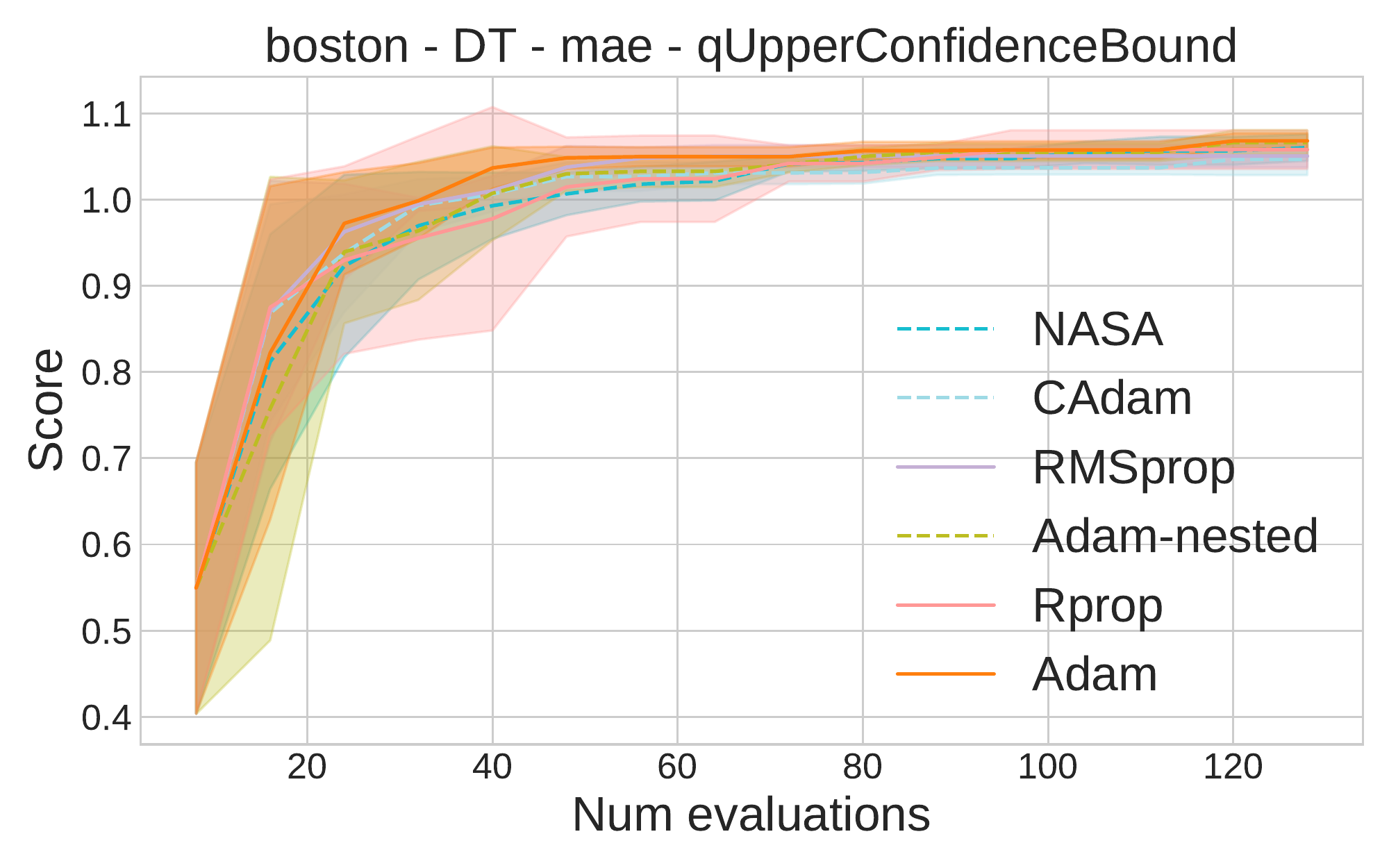}} \\  
    \subfloat{\includegraphics[width=0.19\columnwidth, trim={0 0.5cm 0 0.4cm}, clip]{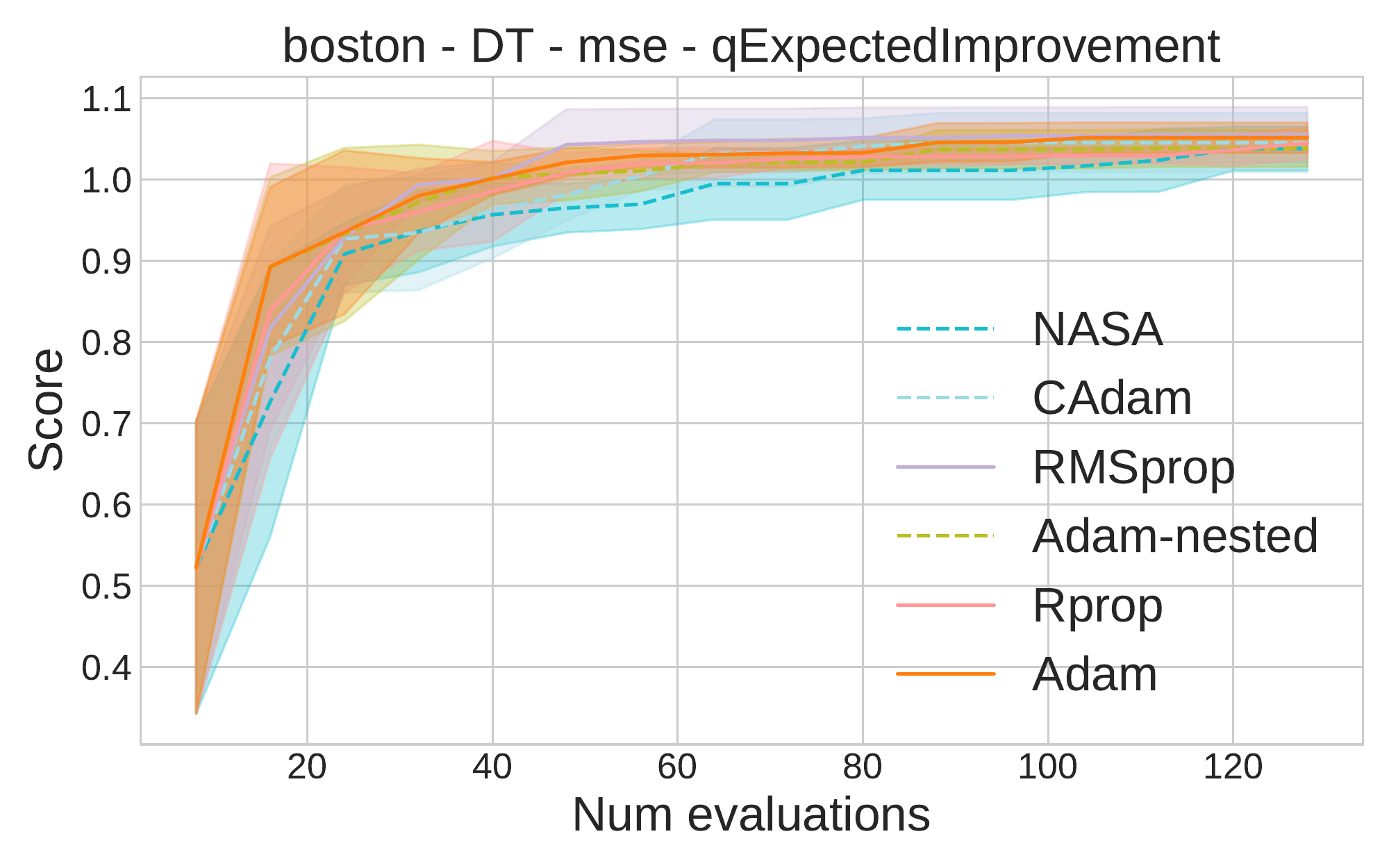}} &   \hspace{-0.5cm} 
    \subfloat{\includegraphics[width=0.19\columnwidth, trim={0 0.5cm 0 0.4cm}, clip]{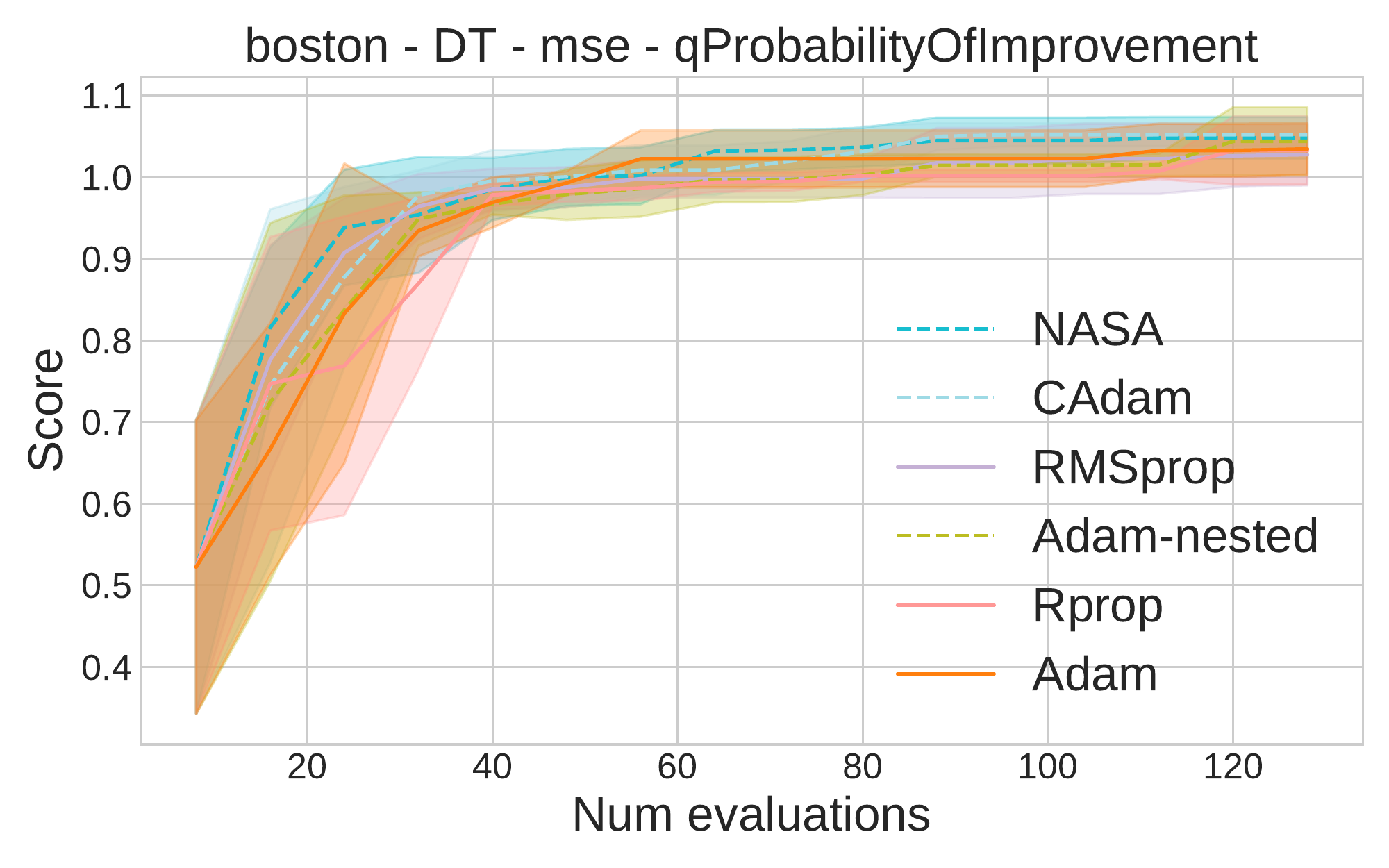}} & \hspace{-0.5cm}
    \subfloat{\includegraphics[width=0.19\columnwidth, trim={0 0.5cm 0 0.4cm}, clip]{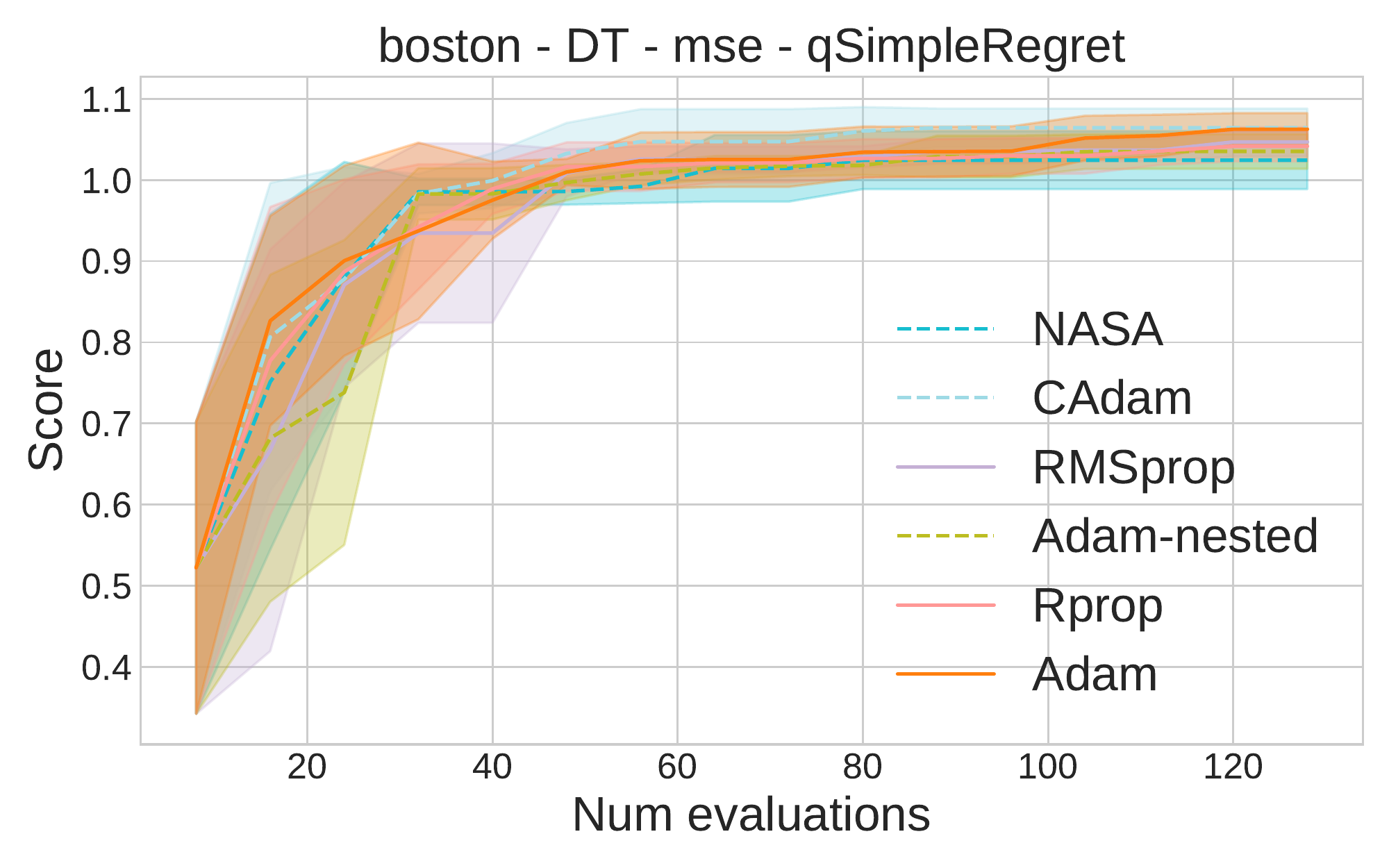}} &  \hspace{-0.5cm}
    \subfloat{\includegraphics[width=0.19\columnwidth, trim={0 0.5cm 0 0.4cm}, clip]{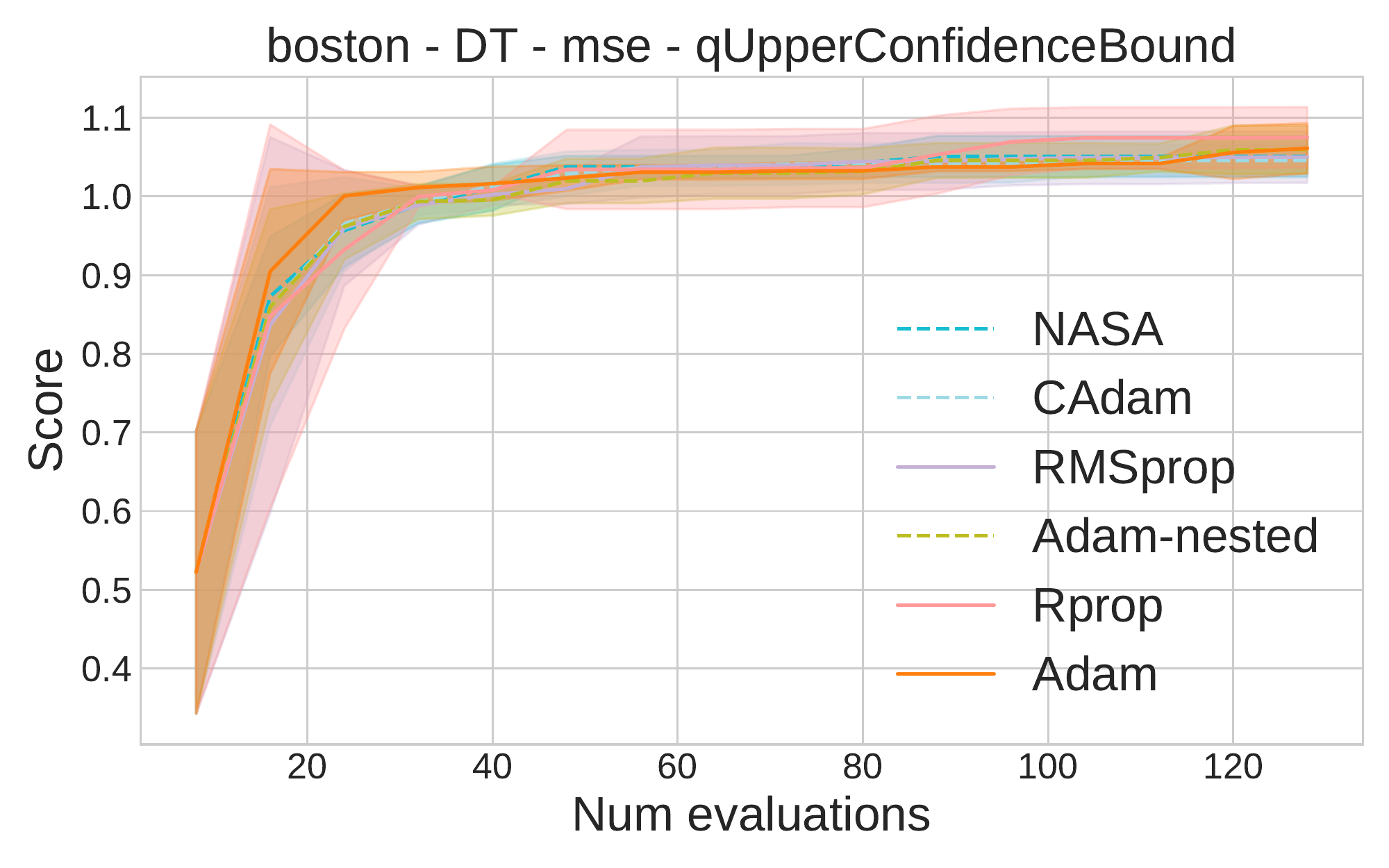}} & \hspace{-0.5cm}
    \subfloat{\includegraphics[width=0.19\columnwidth, trim={0 0.5cm 0 0.4cm}, clip]{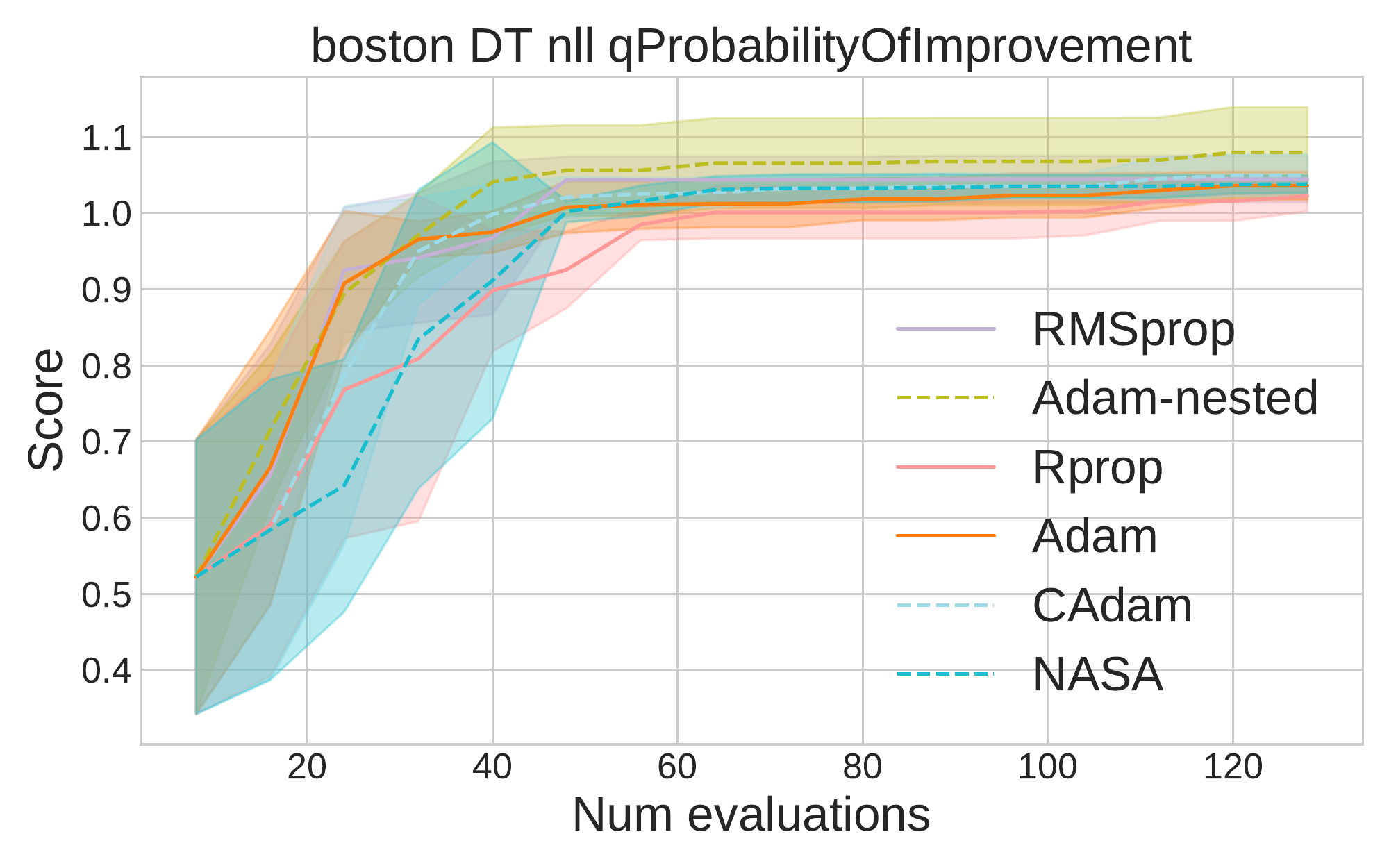}} \\  
    \subfloat{\includegraphics[width=0.19\columnwidth, trim={0 0.5cm 0 0.4cm}, clip]{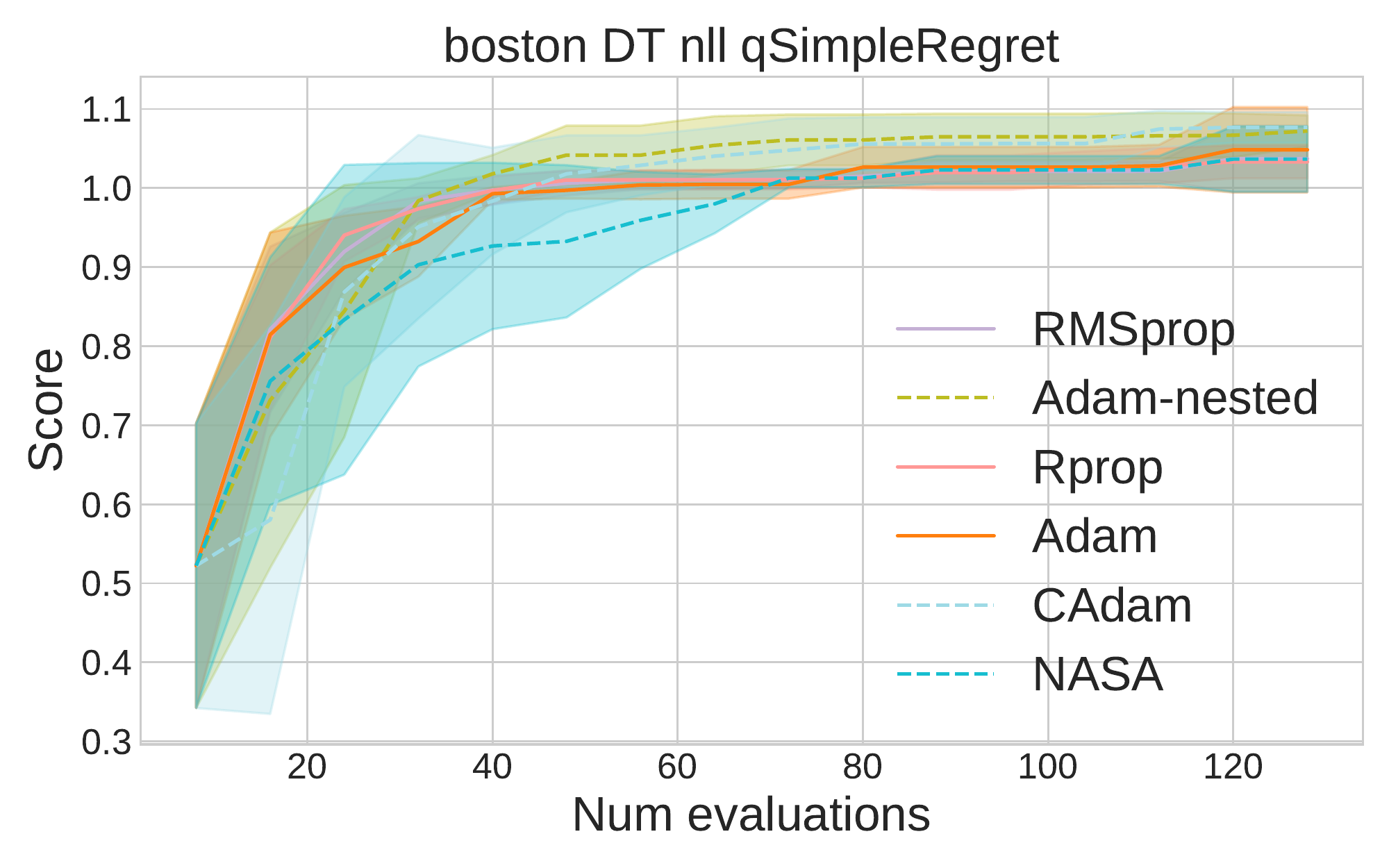}} &   \hspace{-0.5cm} 
    \subfloat{\includegraphics[width=0.19\columnwidth, trim={0 0.5cm 0 0.4cm}, clip]{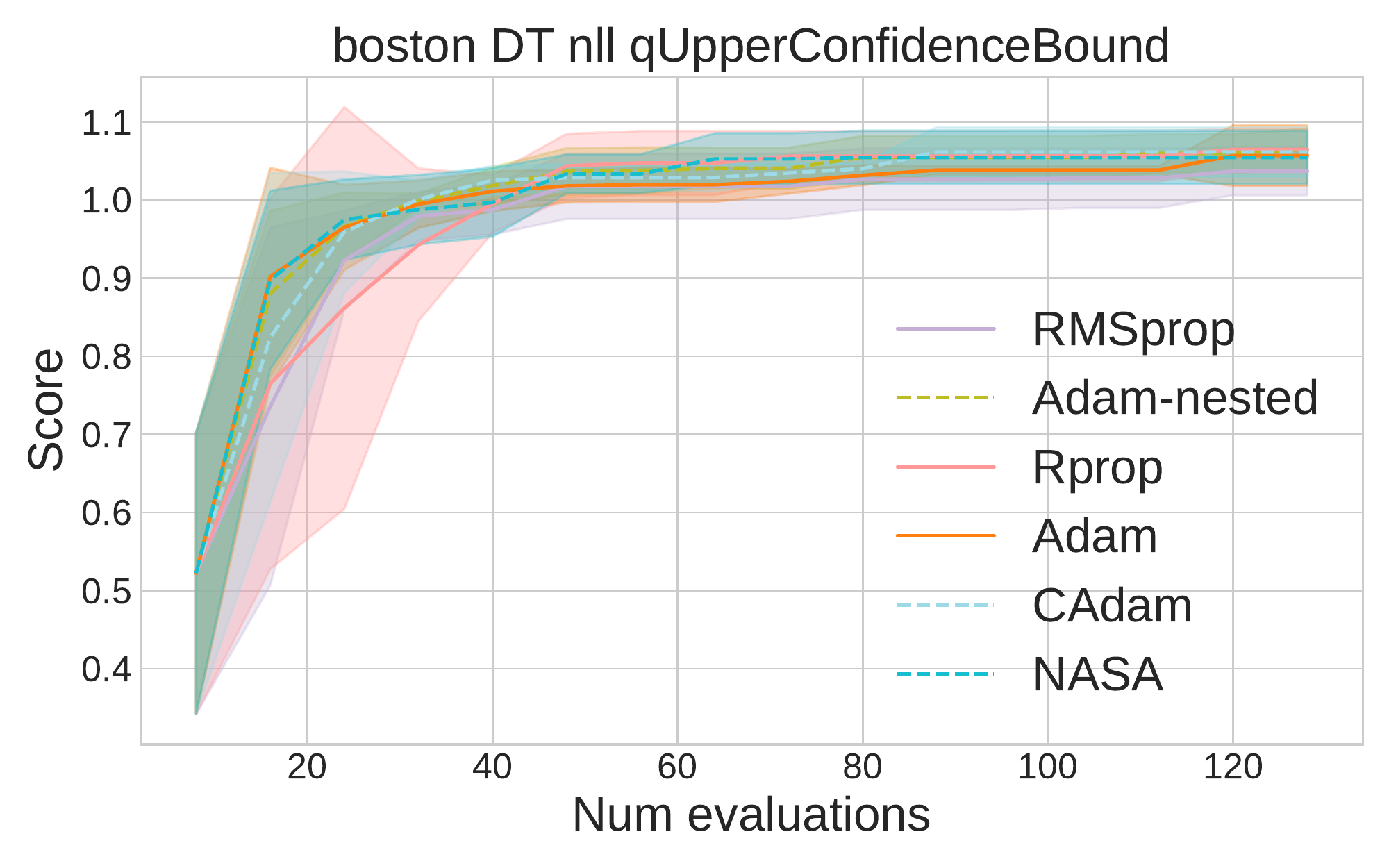}} & \hspace{-0.5cm}
    \subfloat{\includegraphics[width=0.19\columnwidth, trim={0 0.5cm 0 0.4cm}, clip]{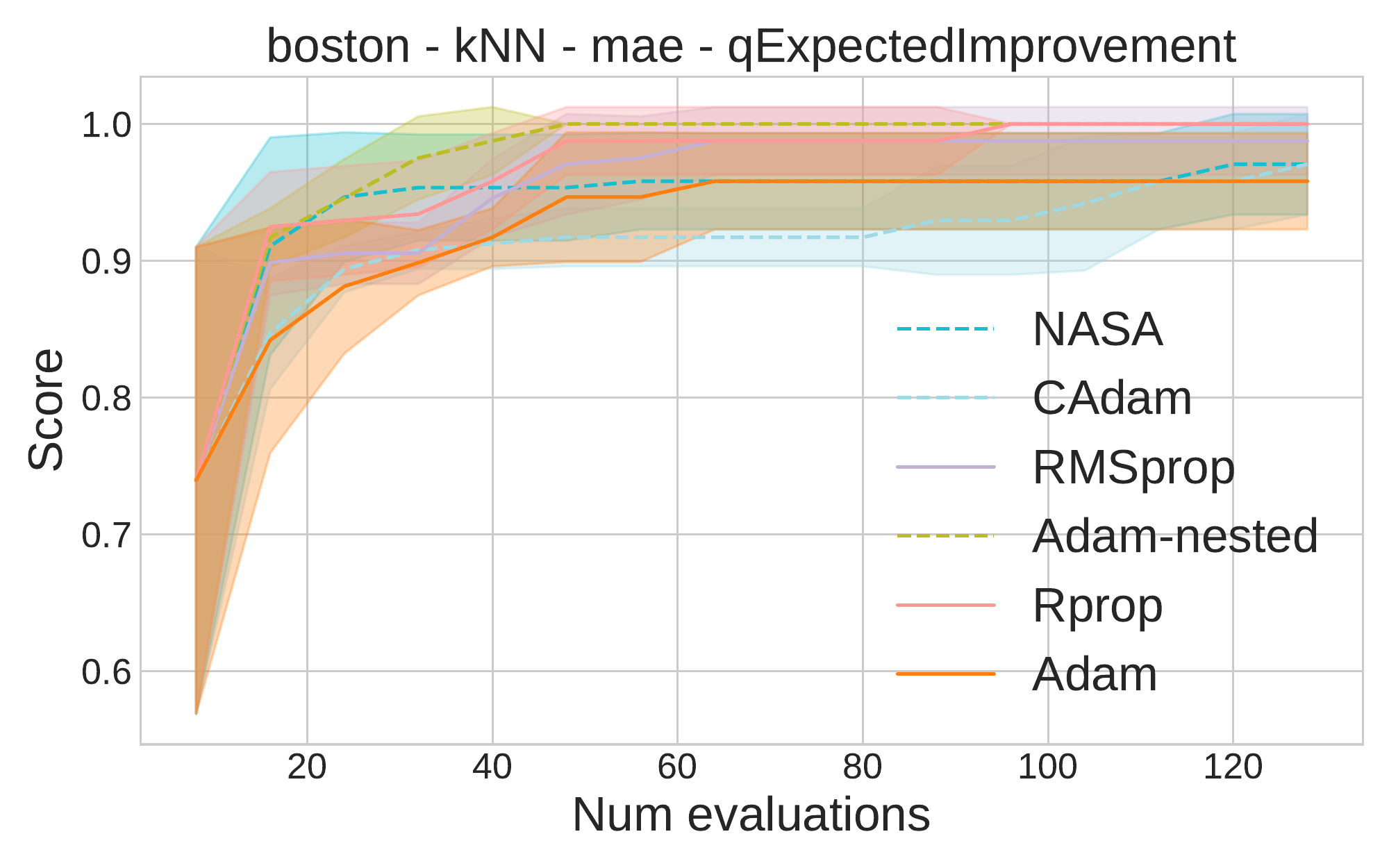}} &  \hspace{-0.5cm}
    \subfloat{\includegraphics[width=0.19\columnwidth, trim={0 0.5cm 0 0.4cm}, clip]{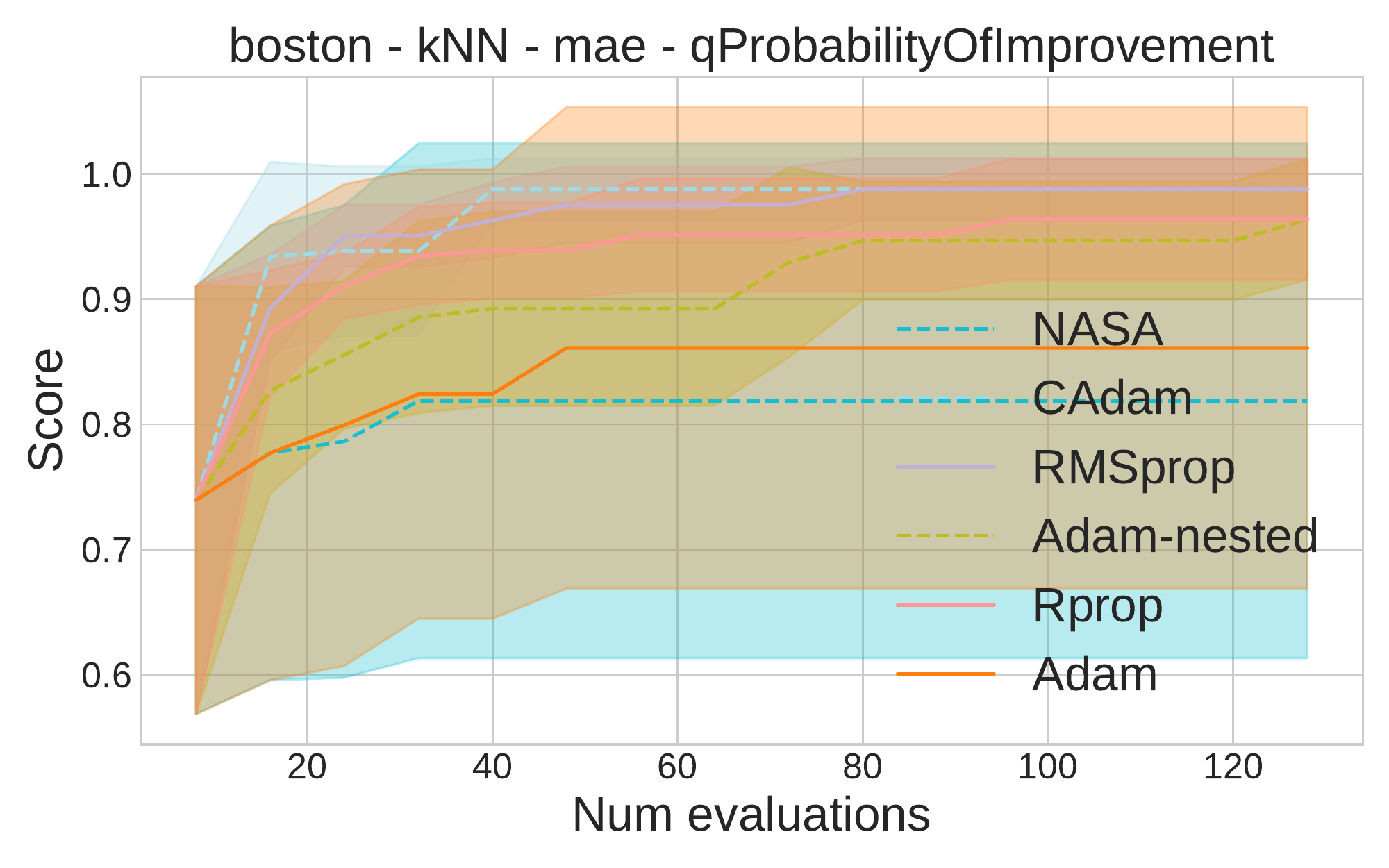}} & \hspace{-0.5cm}
    \subfloat{\includegraphics[width=0.19\columnwidth, trim={0 0.5cm 0 0.4cm}, clip]{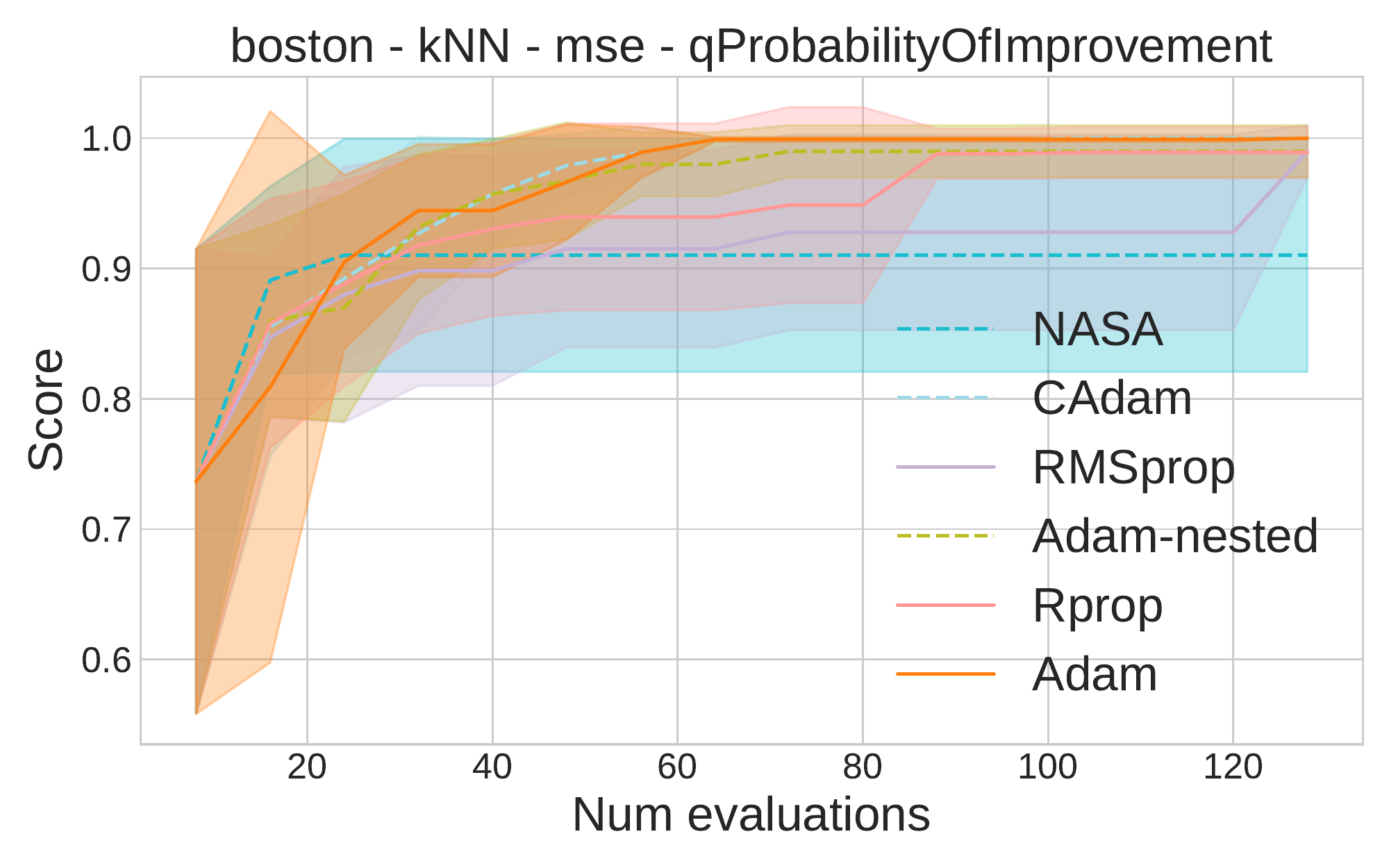}} \\  
    \subfloat{\includegraphics[width=0.19\columnwidth, trim={0 0.5cm 0 0.4cm}, clip]{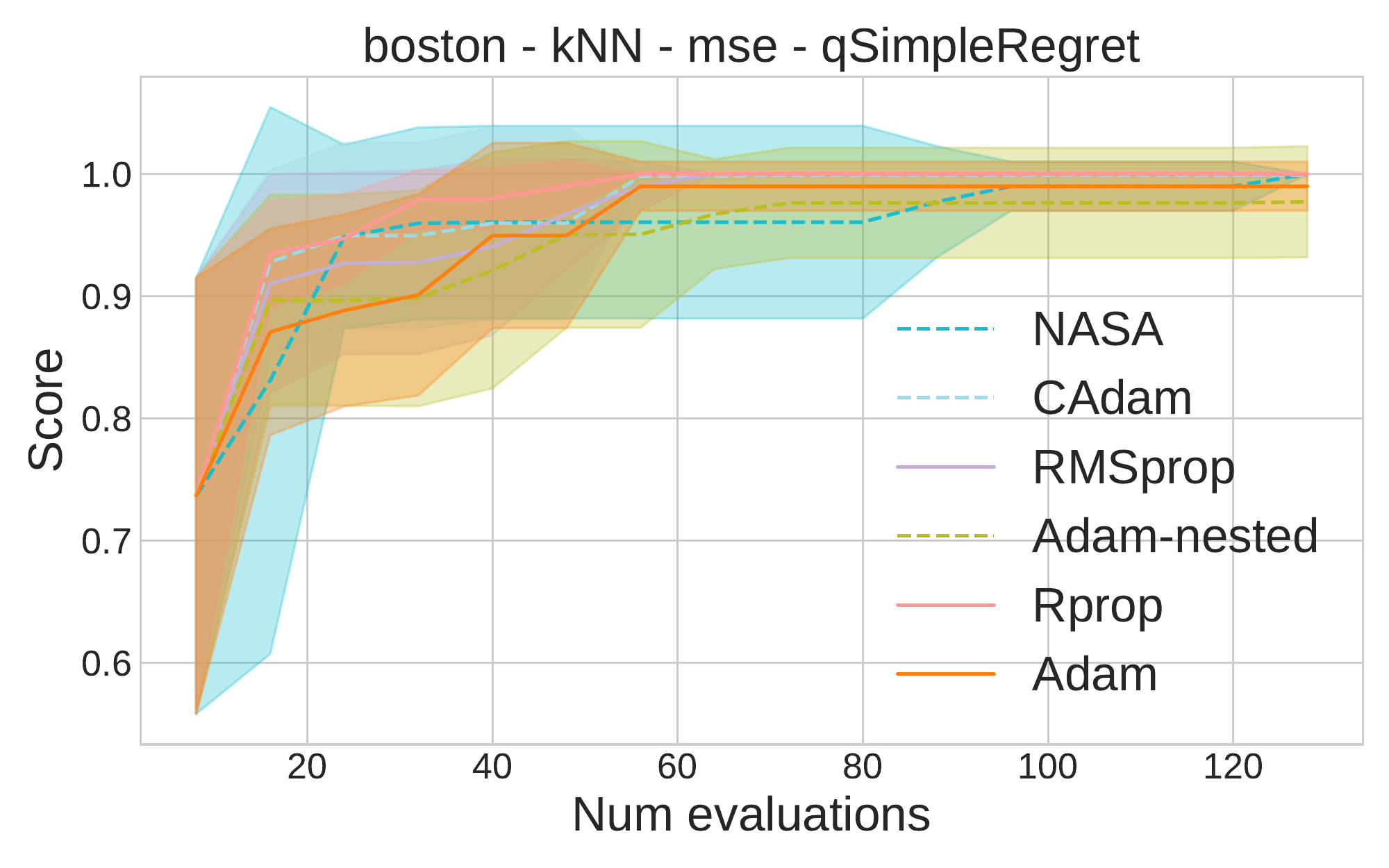}} &   \hspace{-0.5cm} 
    \subfloat{\includegraphics[width=0.19\columnwidth, trim={0 0.5cm 0 0.4cm}, clip]{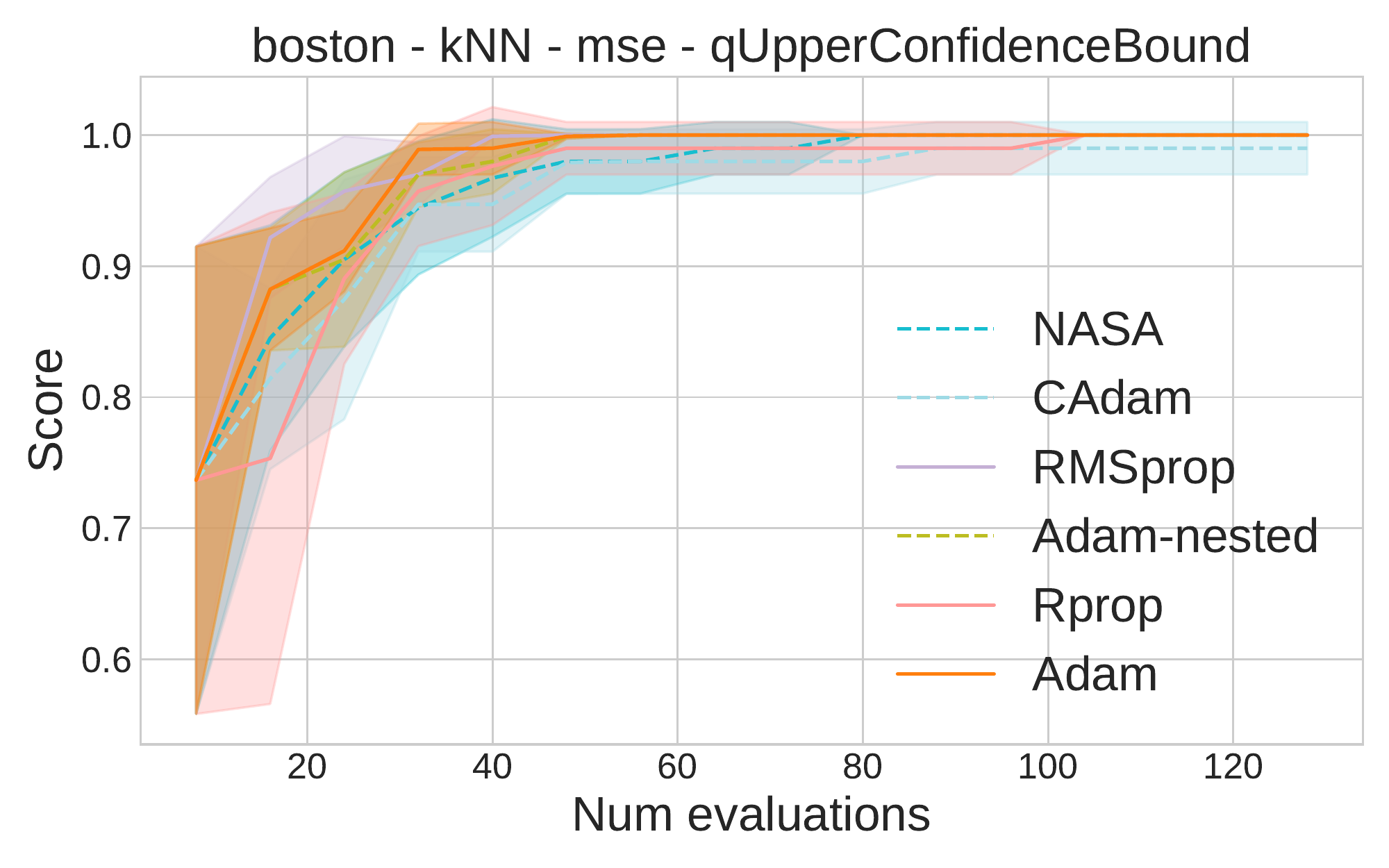}} & \hspace{-0.5cm}
    \subfloat{\includegraphics[width=0.19\columnwidth, trim={0 0.5cm 0 0.4cm}, clip]{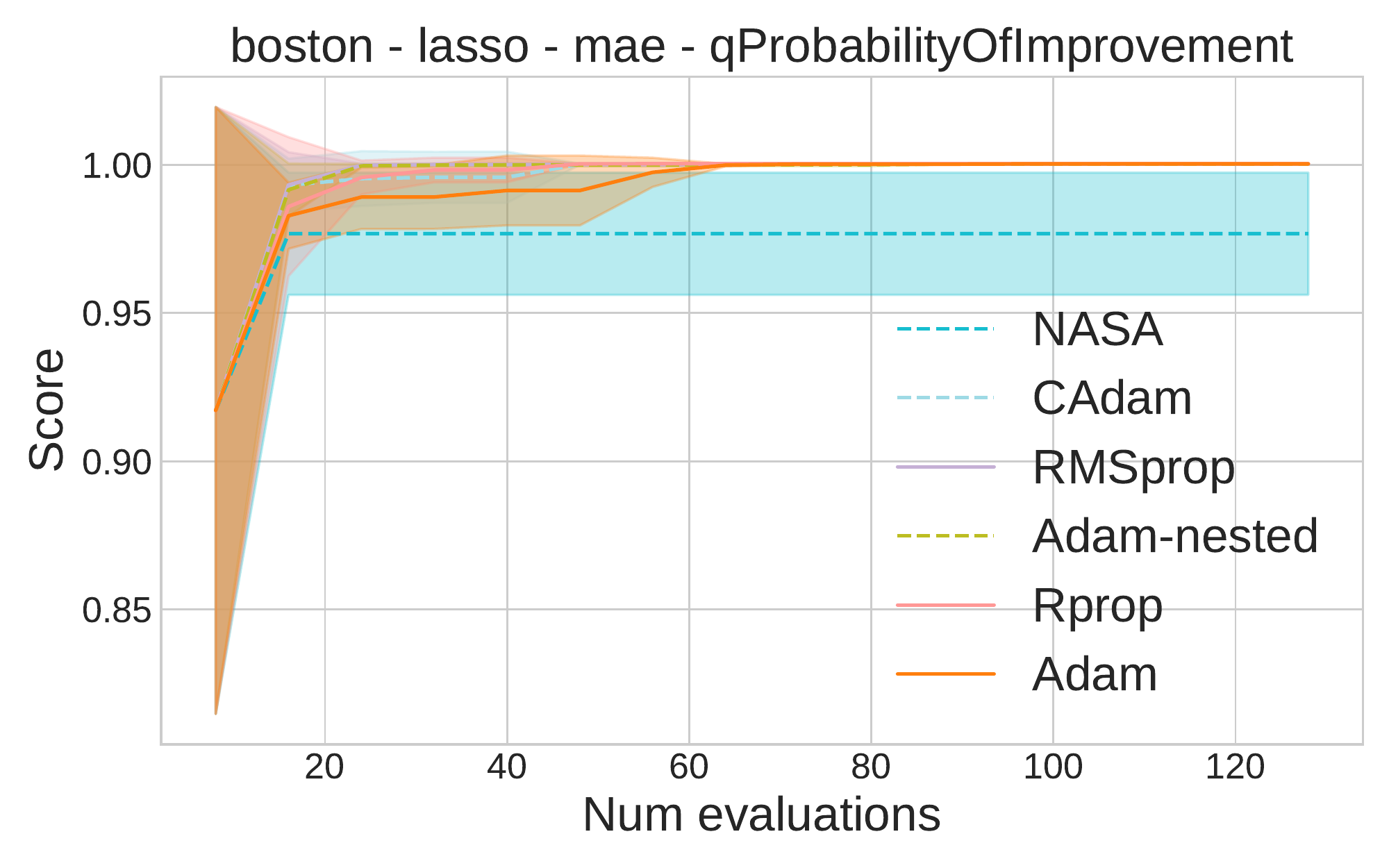}} &  \hspace{-0.5cm}
    \subfloat{\includegraphics[width=0.19\columnwidth, trim={0 0.5cm 0 0.4cm}, clip]{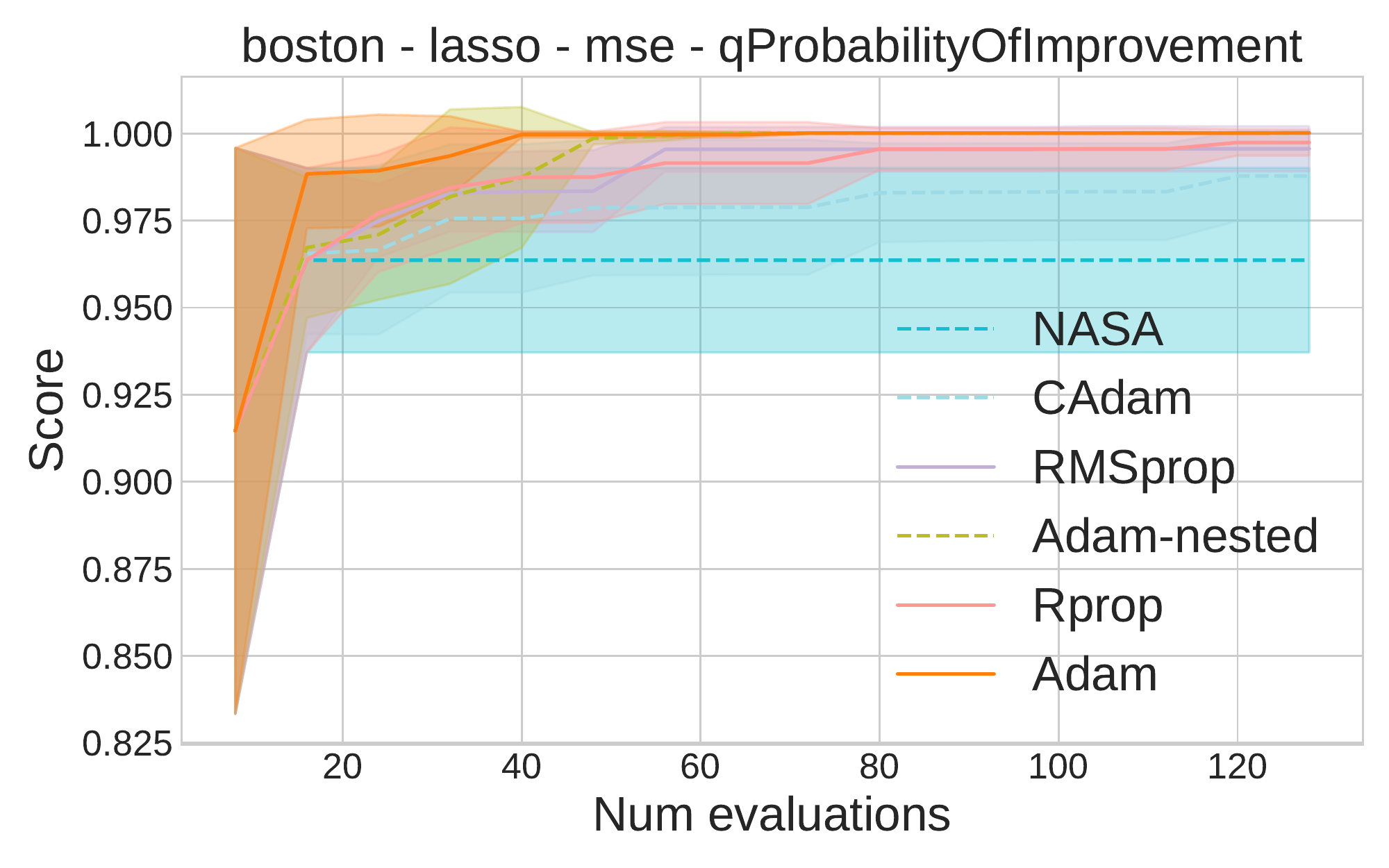}} & \hspace{-0.5cm}
    \subfloat{\includegraphics[width=0.19\columnwidth, trim={0 0.5cm 0 0.4cm}, clip]{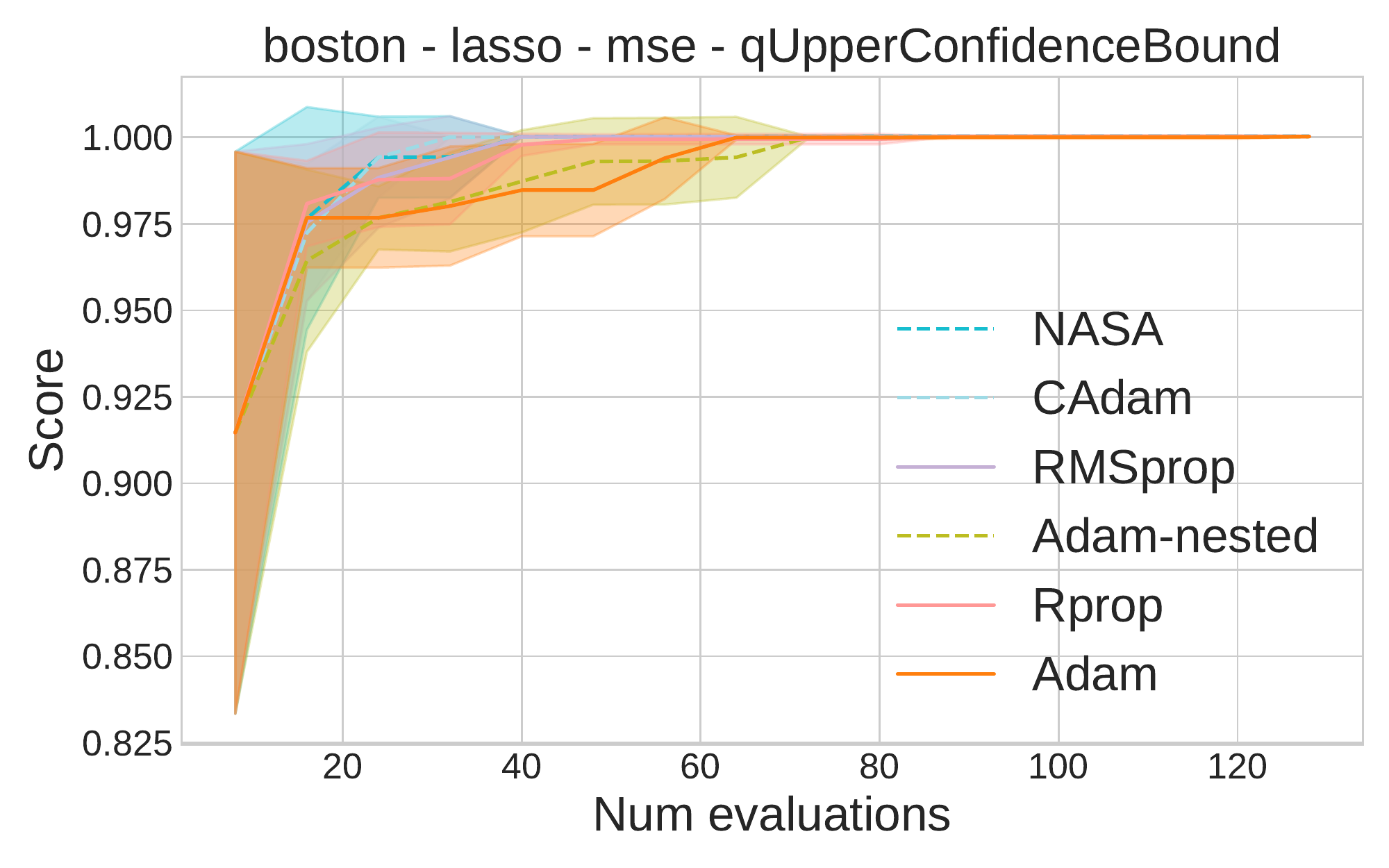}} \\  
    \subfloat{\includegraphics[width=0.19\columnwidth, trim={0 0.5cm 0 0.4cm}, clip]{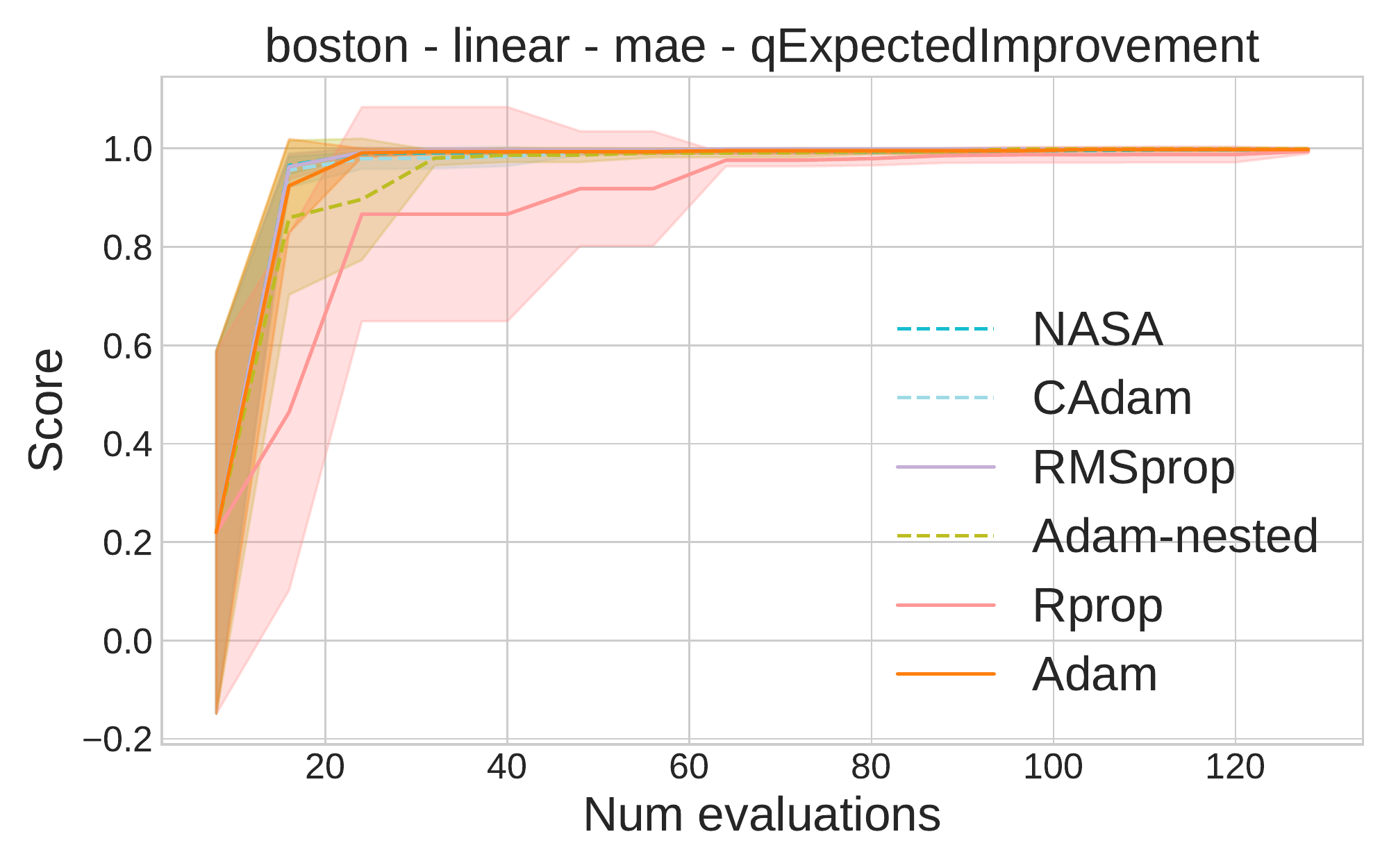}} &   \hspace{-0.5cm} 
    \subfloat{\includegraphics[width=0.19\columnwidth, trim={0 0.5cm 0 0.4cm}, clip]{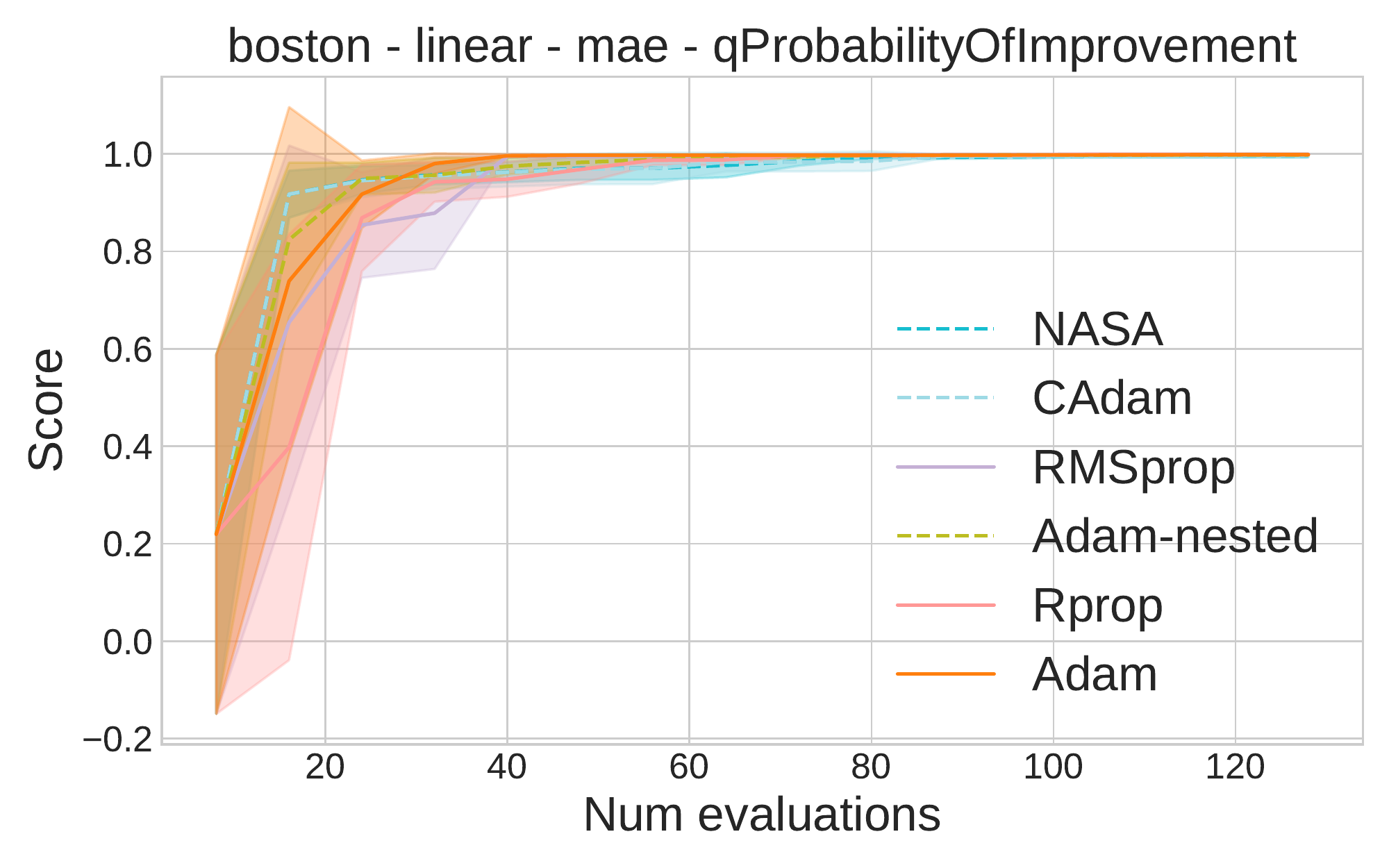}} & \hspace{-0.5cm}
    \subfloat{\includegraphics[width=0.19\columnwidth, trim={0 0.5cm 0 0.4cm}, clip]{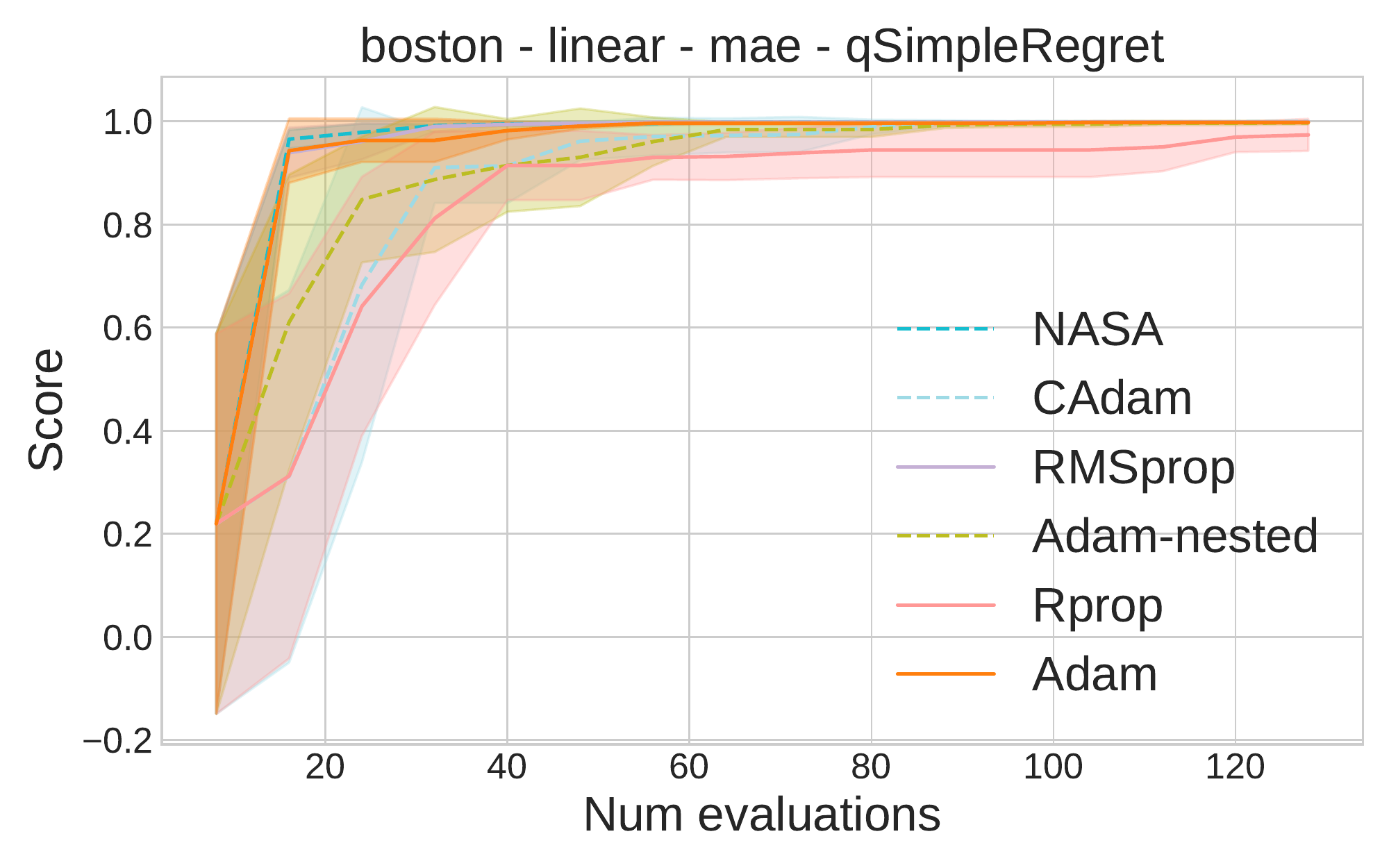}} &  \hspace{-0.5cm}
    \subfloat{\includegraphics[width=0.19\columnwidth, trim={0 0.5cm 0 0.4cm}, clip]{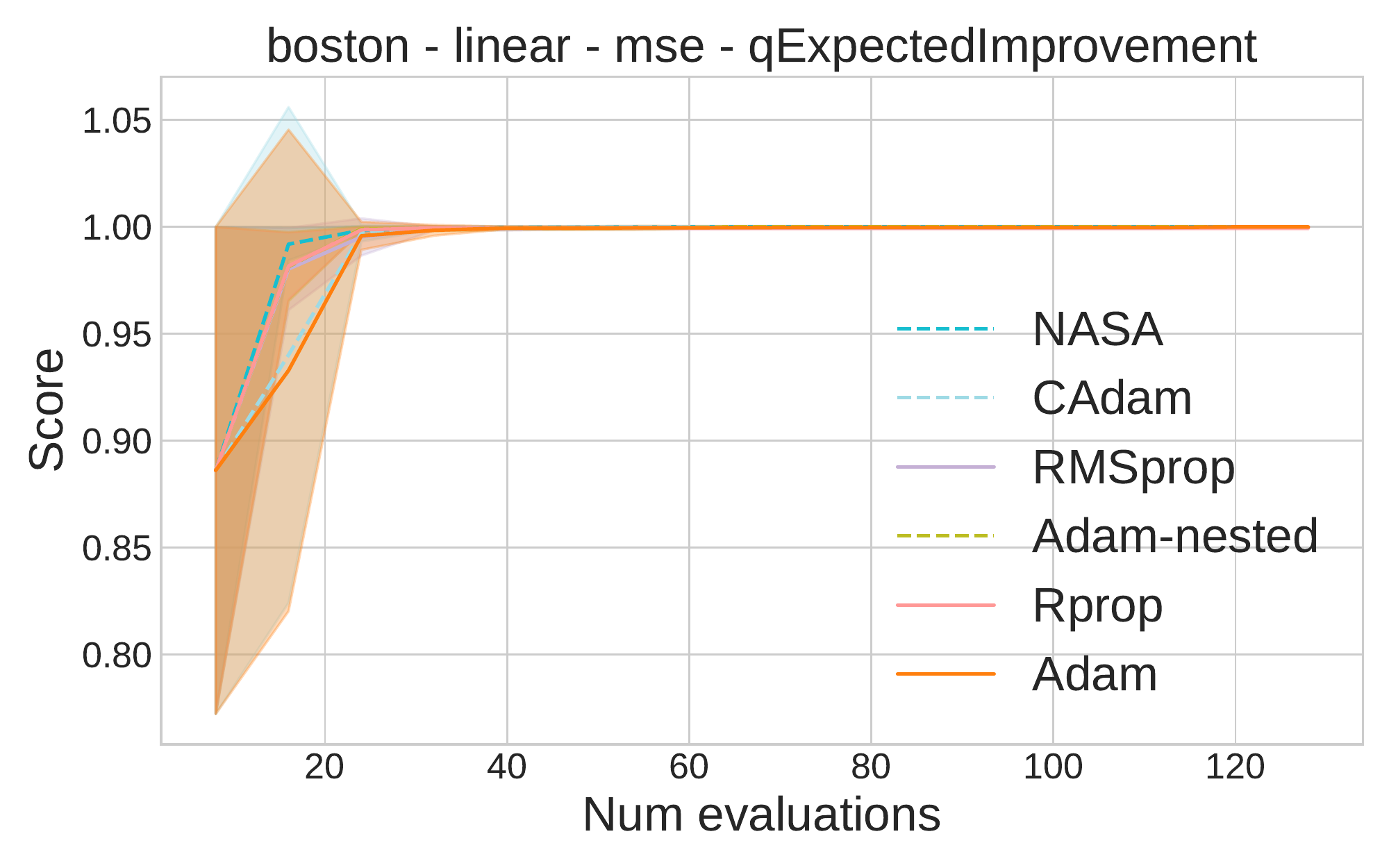}} & \hspace{-0.5cm}
    \subfloat{\includegraphics[width=0.19\columnwidth, trim={0 0.5cm 0 0.4cm}, clip]{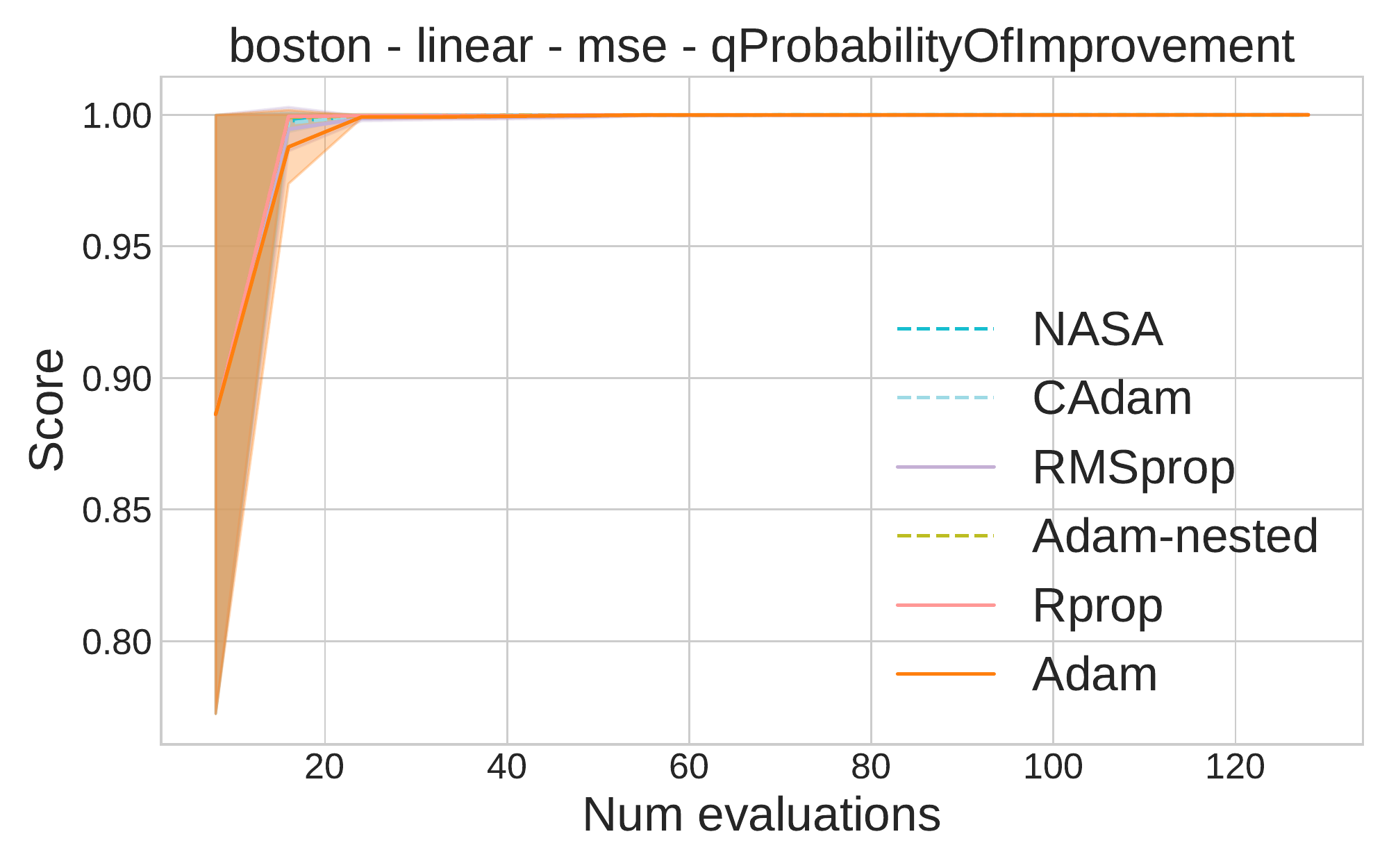}} \\  
    \subfloat{\includegraphics[width=0.19\columnwidth, trim={0 0.5cm 0 0.4cm}, clip]{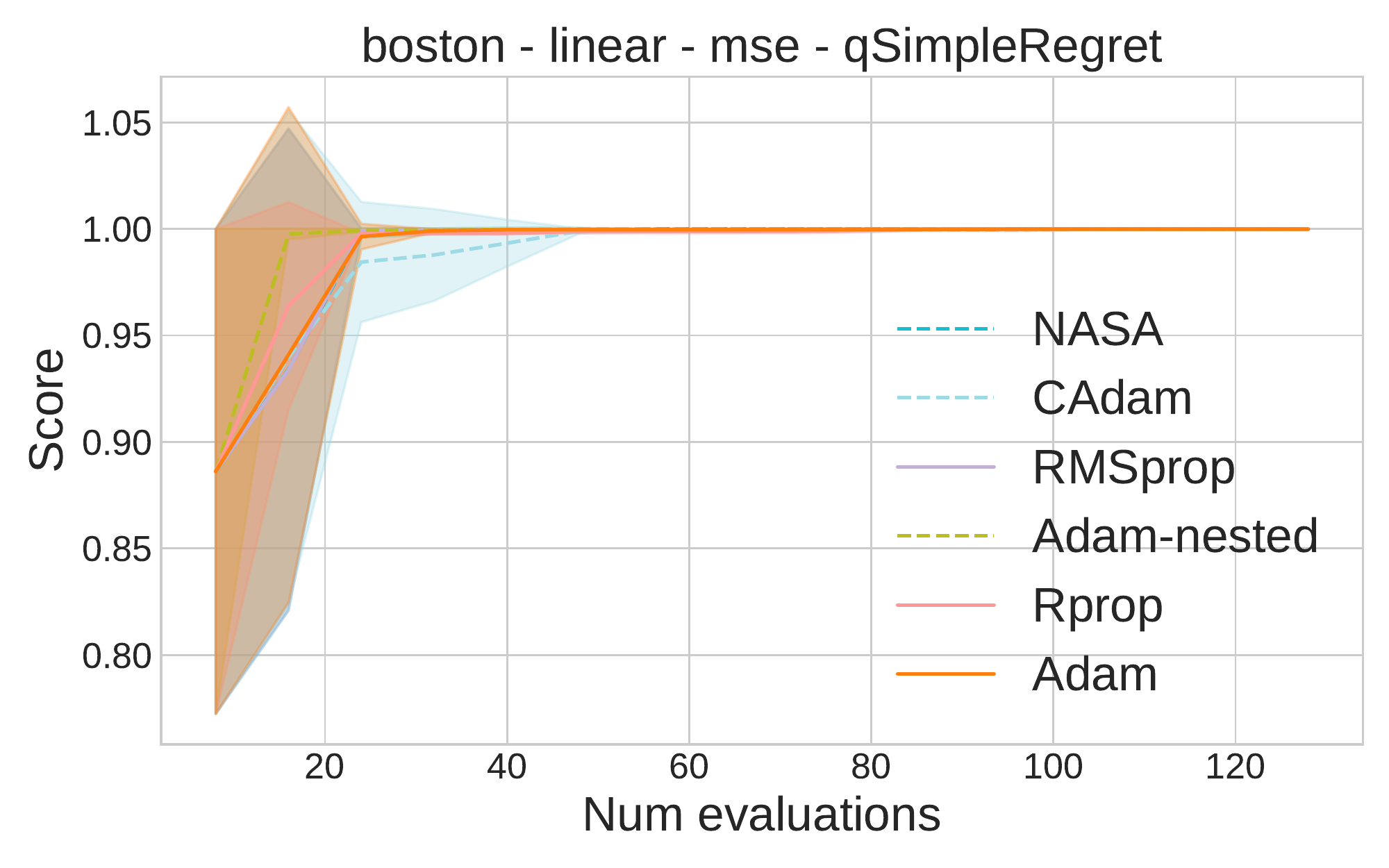}} &   \hspace{-0.5cm} 
    \subfloat{\includegraphics[width=0.19\columnwidth, trim={0 0.5cm 0 0.4cm}, clip]{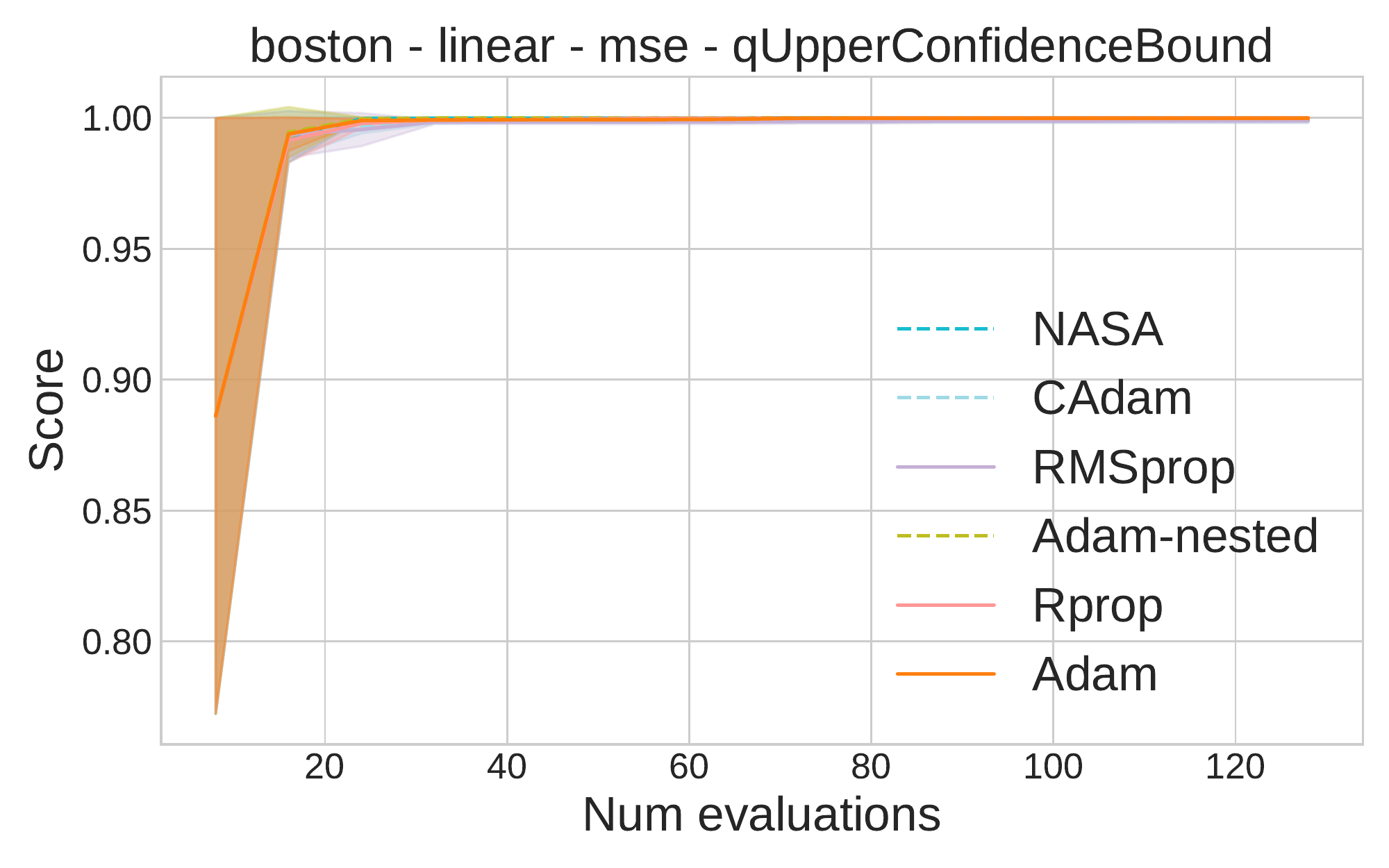}} & \hspace{-0.5cm}
    \subfloat{\includegraphics[width=0.19\columnwidth, trim={0 0.5cm 0 0.4cm}, clip]{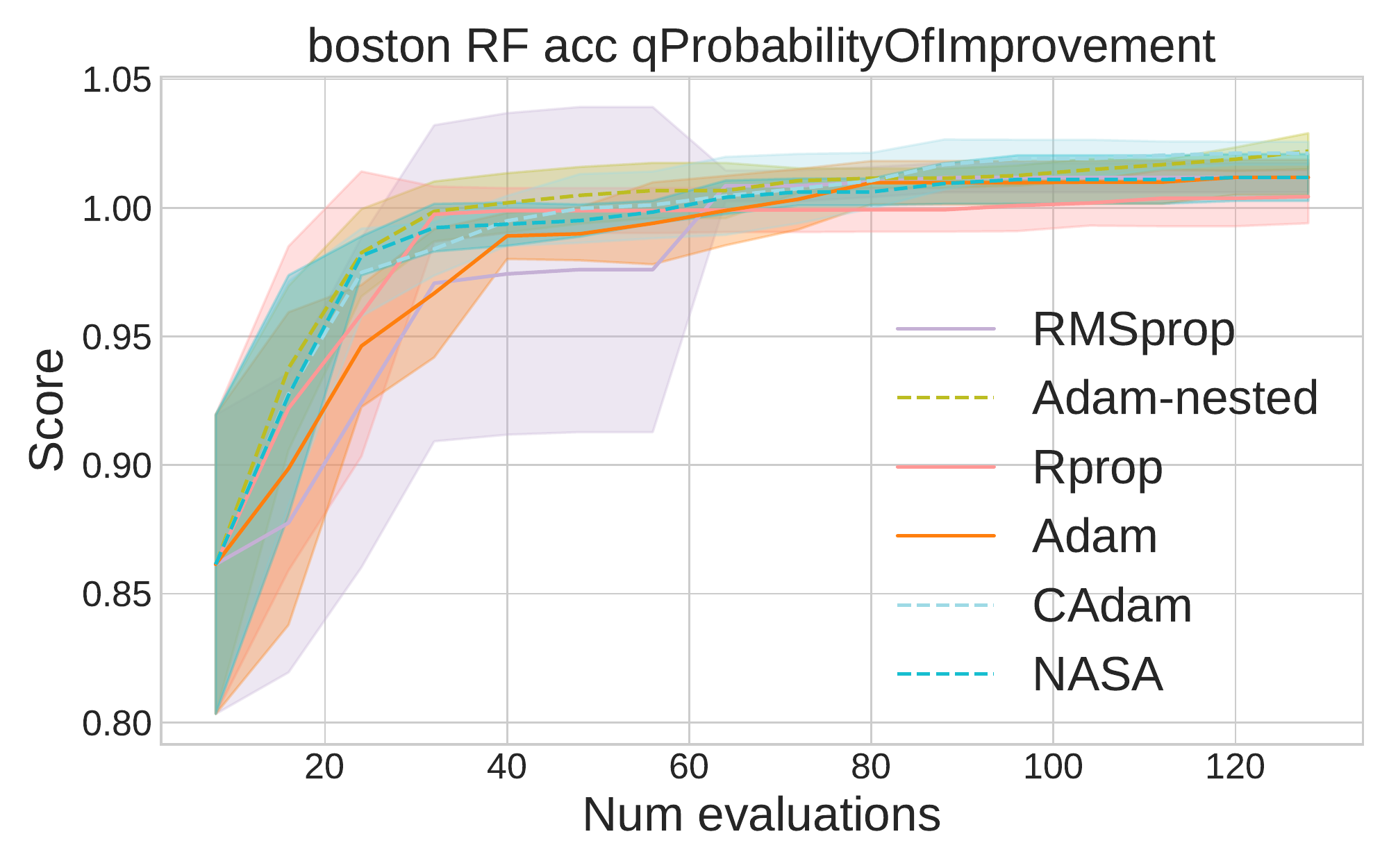}} &  \hspace{-0.5cm}
    \subfloat{\includegraphics[width=0.19\columnwidth, trim={0 0.5cm 0 0.4cm}, clip]{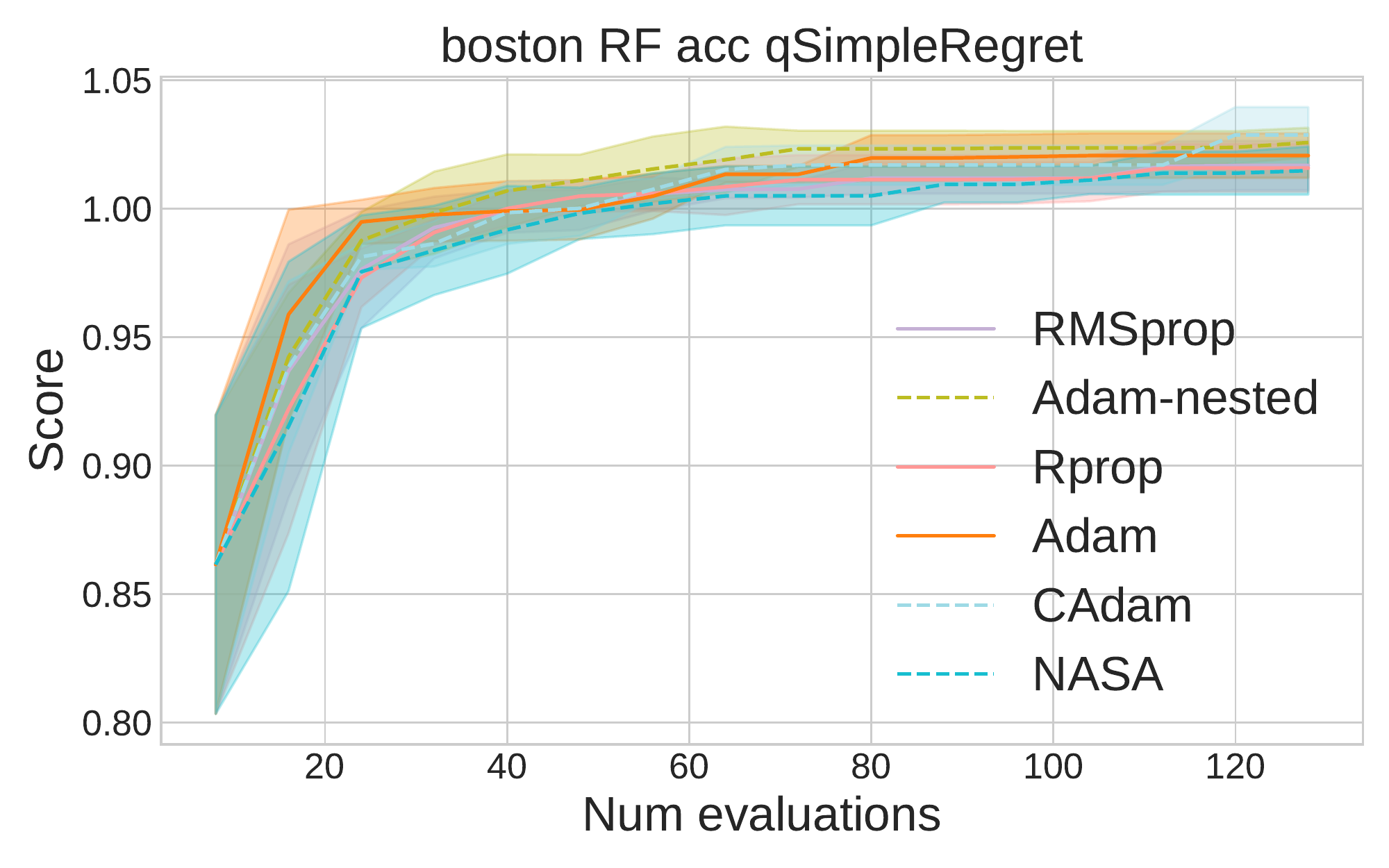}} & \hspace{-0.5cm}
    \subfloat{\includegraphics[width=0.19\columnwidth, trim={0 0.5cm 0 0.4cm}, clip]{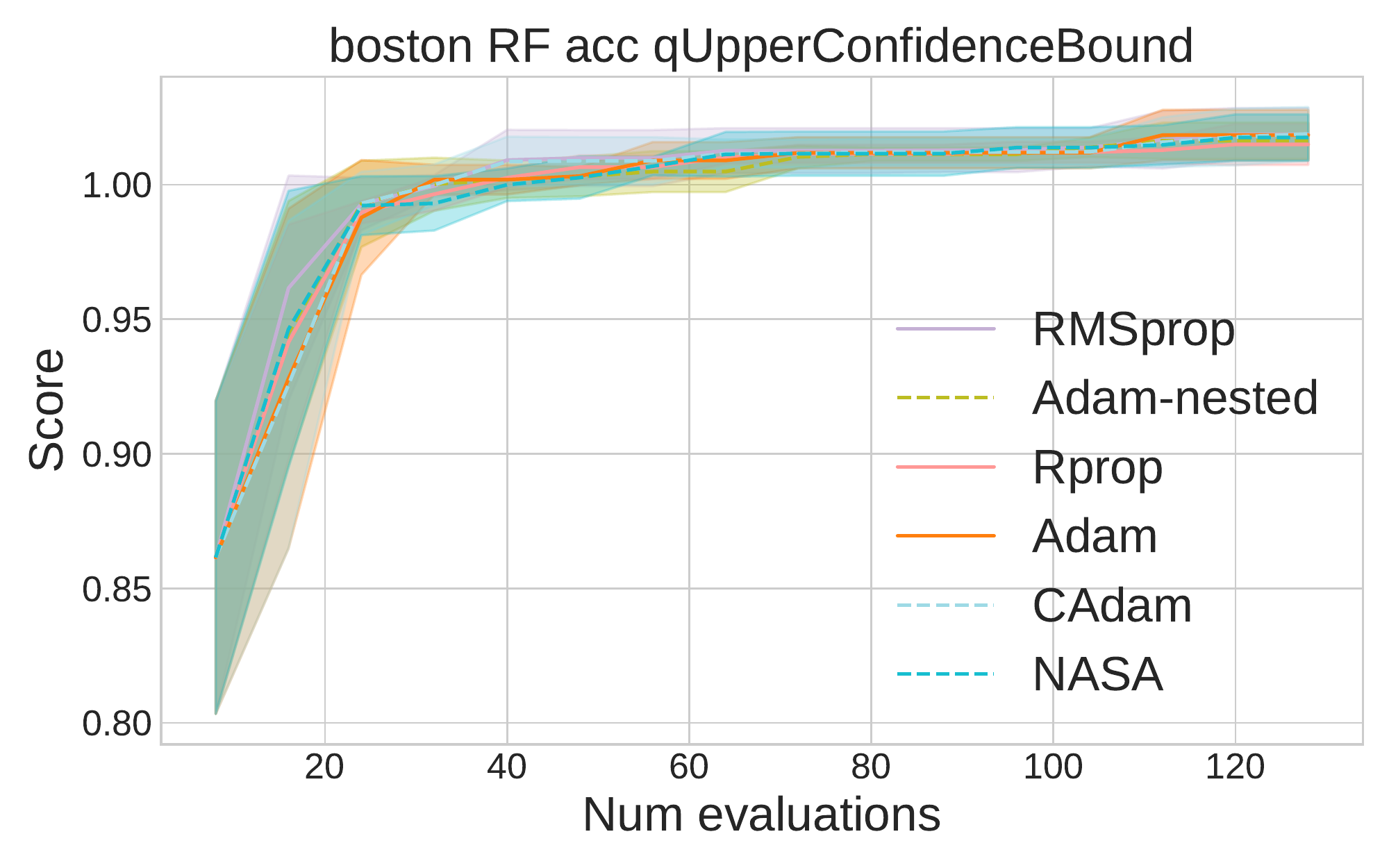}} \\  
    \subfloat{\includegraphics[width=0.19\columnwidth, trim={0 0.5cm 0 0.4cm}, clip]{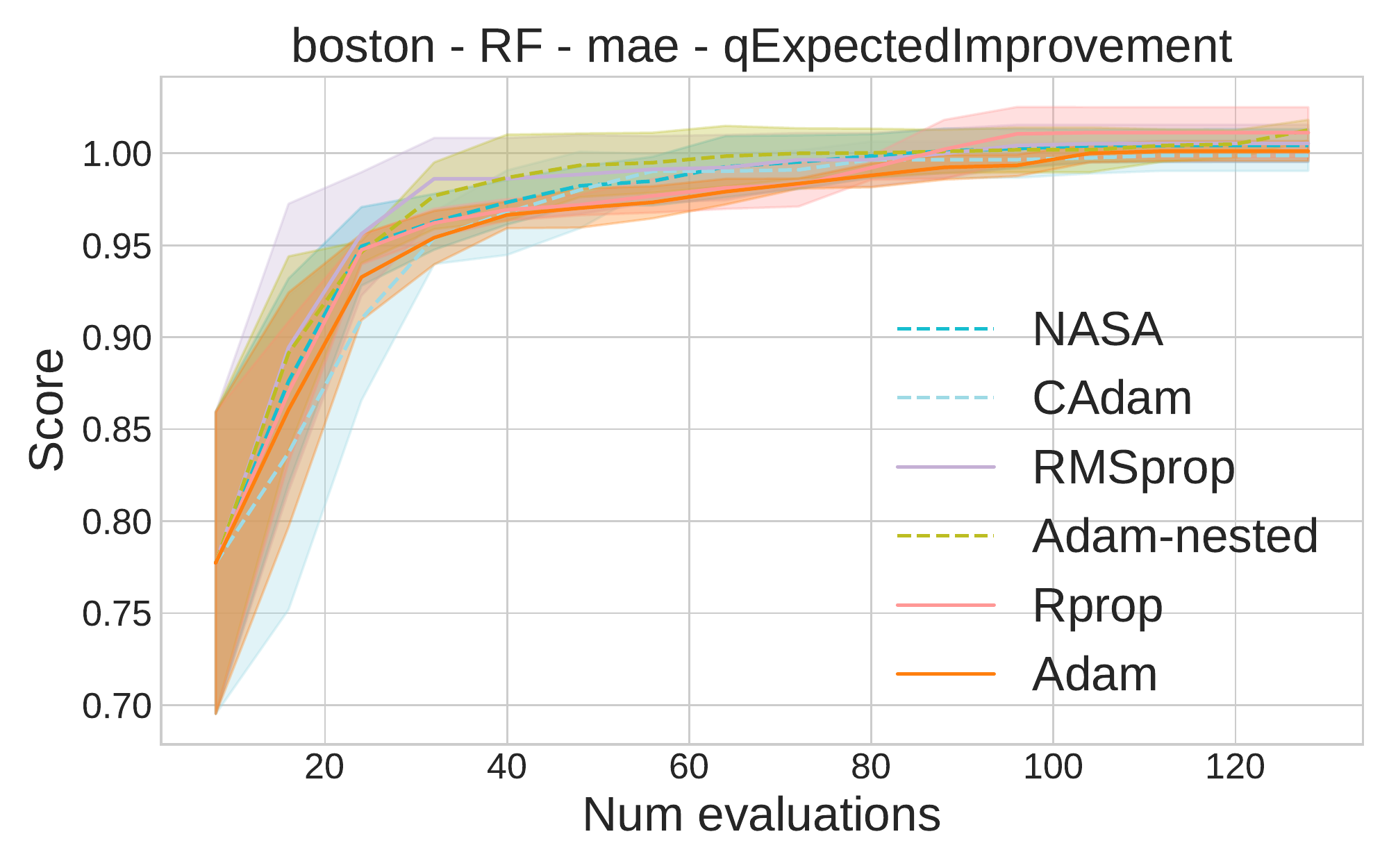}} &   \hspace{-0.5cm} 
    \subfloat{\includegraphics[width=0.19\columnwidth, trim={0 0.5cm 0 0.4cm}, clip]{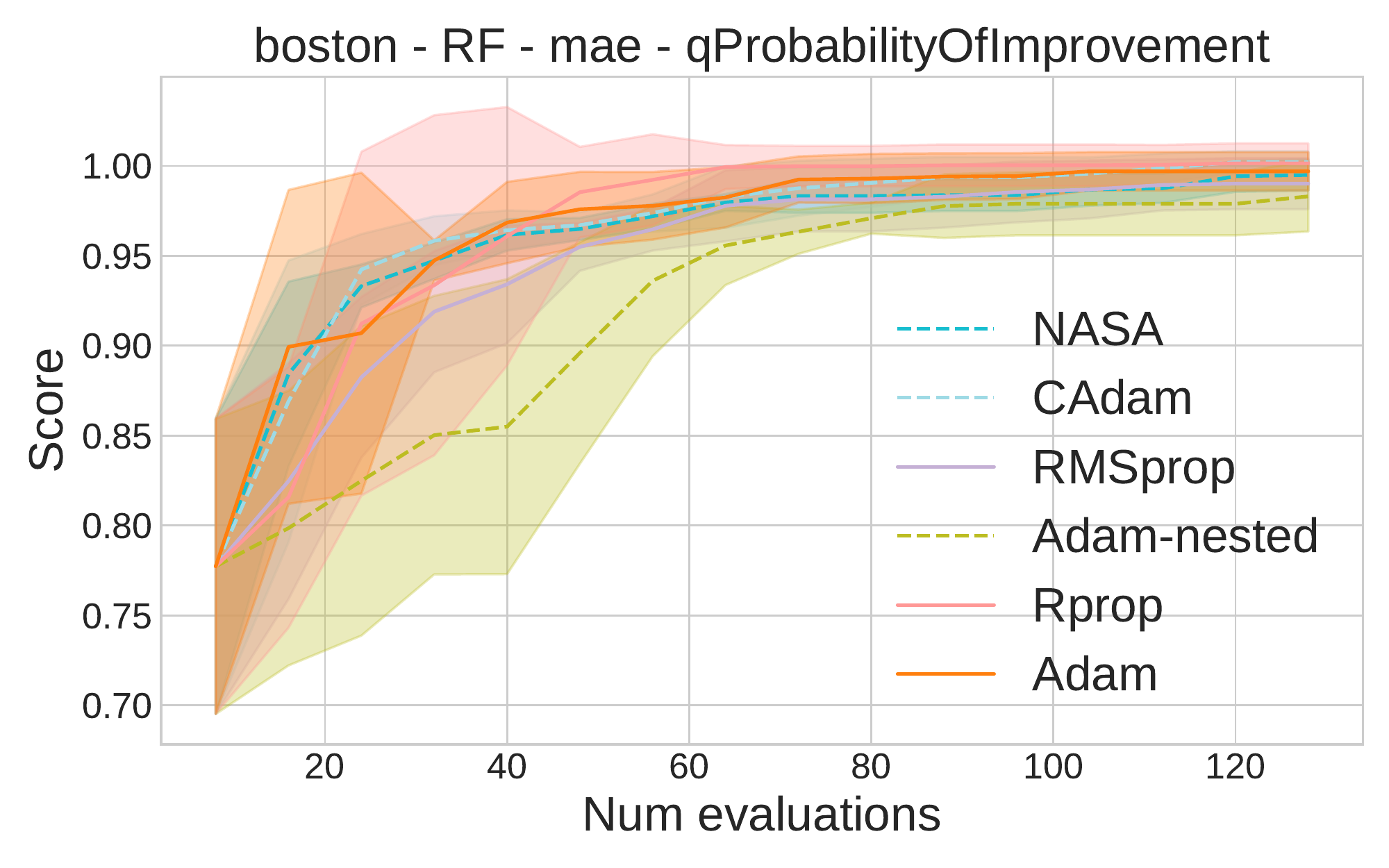}} & \hspace{-0.5cm}
    \subfloat{\includegraphics[width=0.19\columnwidth, trim={0 0.5cm 0 0.4cm}, clip]{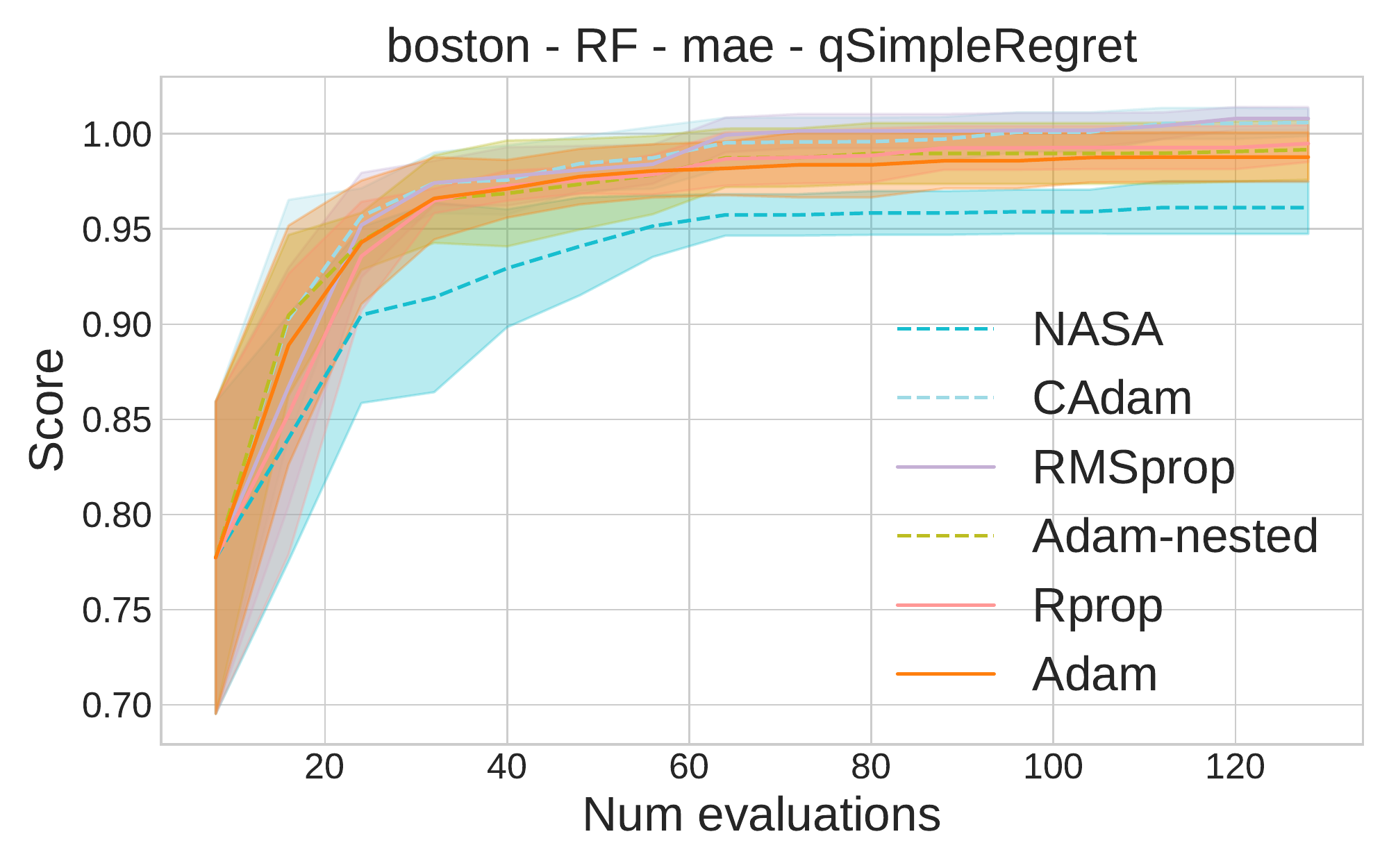}} &  \hspace{-0.5cm}
    \subfloat{\includegraphics[width=0.19\columnwidth, trim={0 0.5cm 0 0.4cm}, clip]{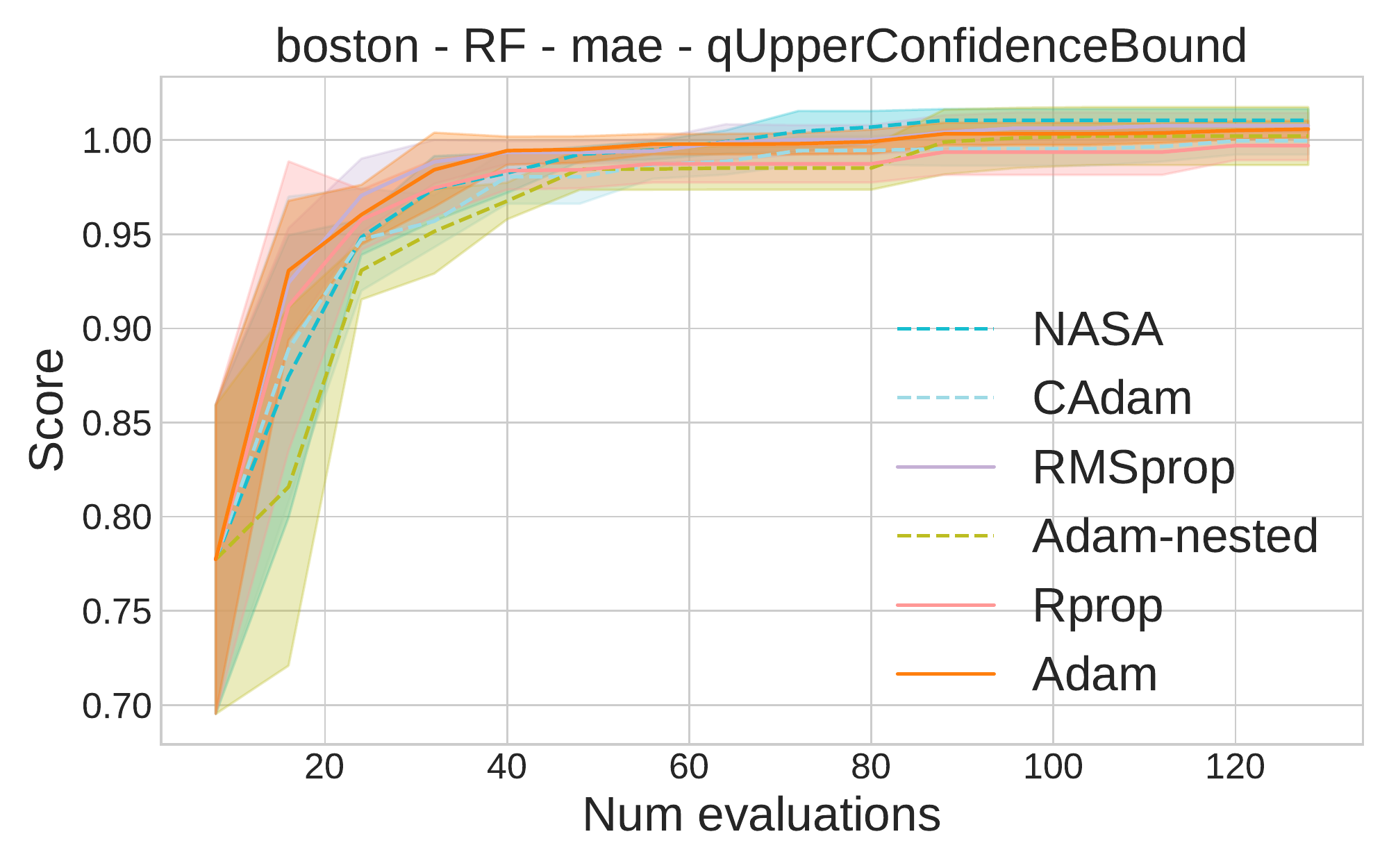}} & \hspace{-0.5cm}
    \subfloat{\includegraphics[width=0.19\columnwidth, trim={0 0.5cm 0 0.4cm}, clip]{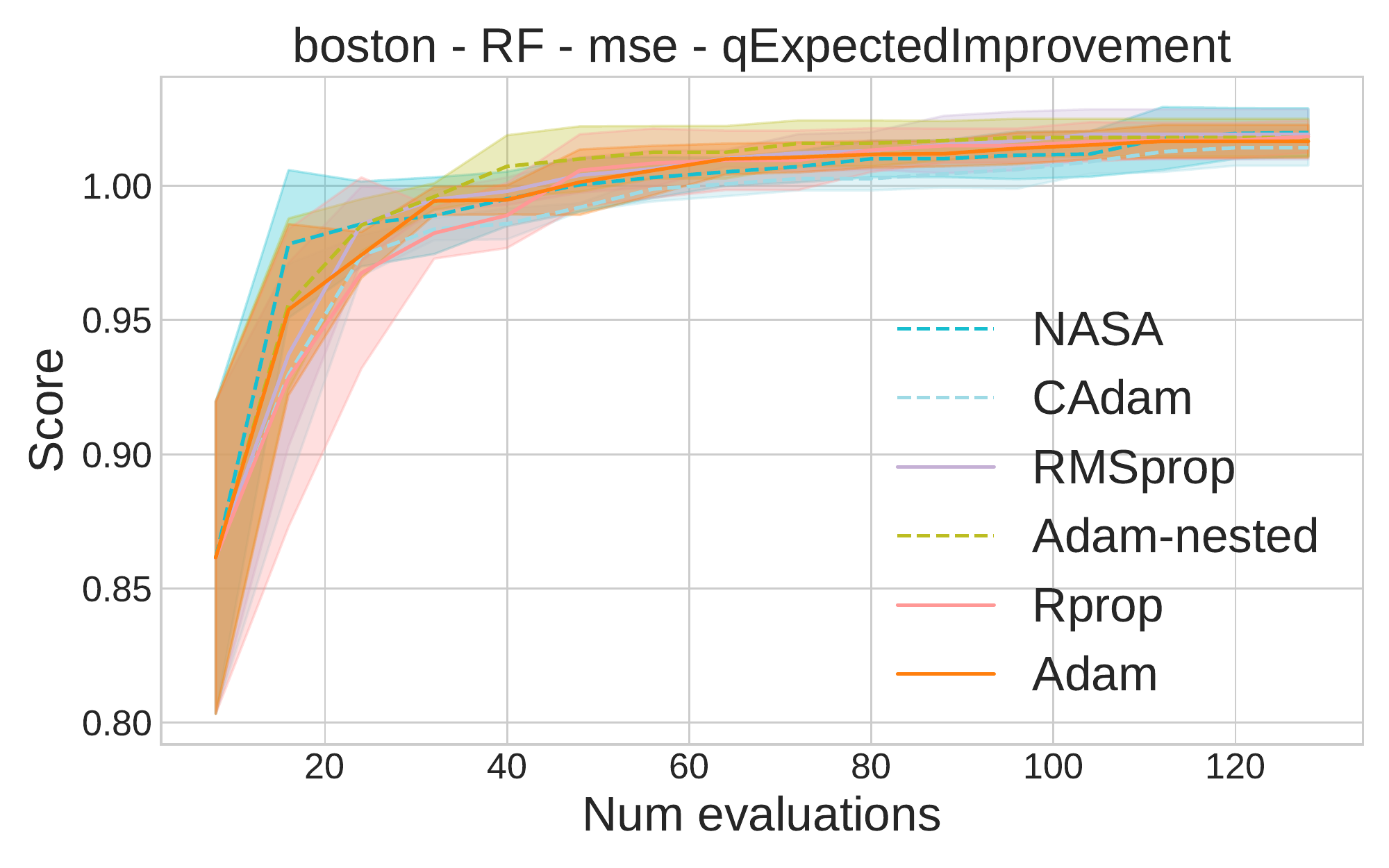}} \\  
    \subfloat{\includegraphics[width=0.19\columnwidth, trim={0 0.5cm 0 0.4cm}, clip]{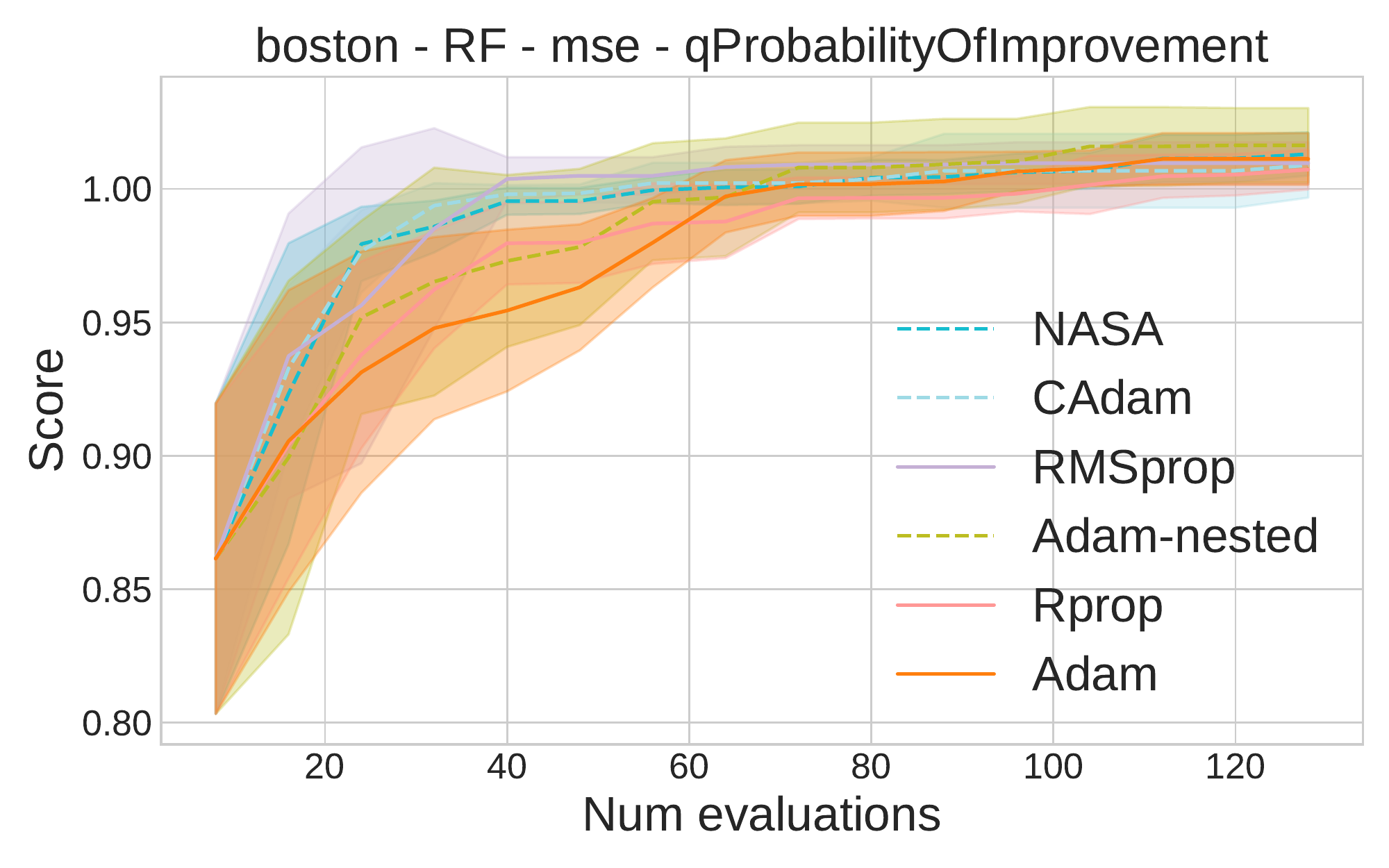}} &   \hspace{-0.5cm} 
    \subfloat{\includegraphics[width=0.19\columnwidth, trim={0 0.5cm 0 0.4cm}, clip]{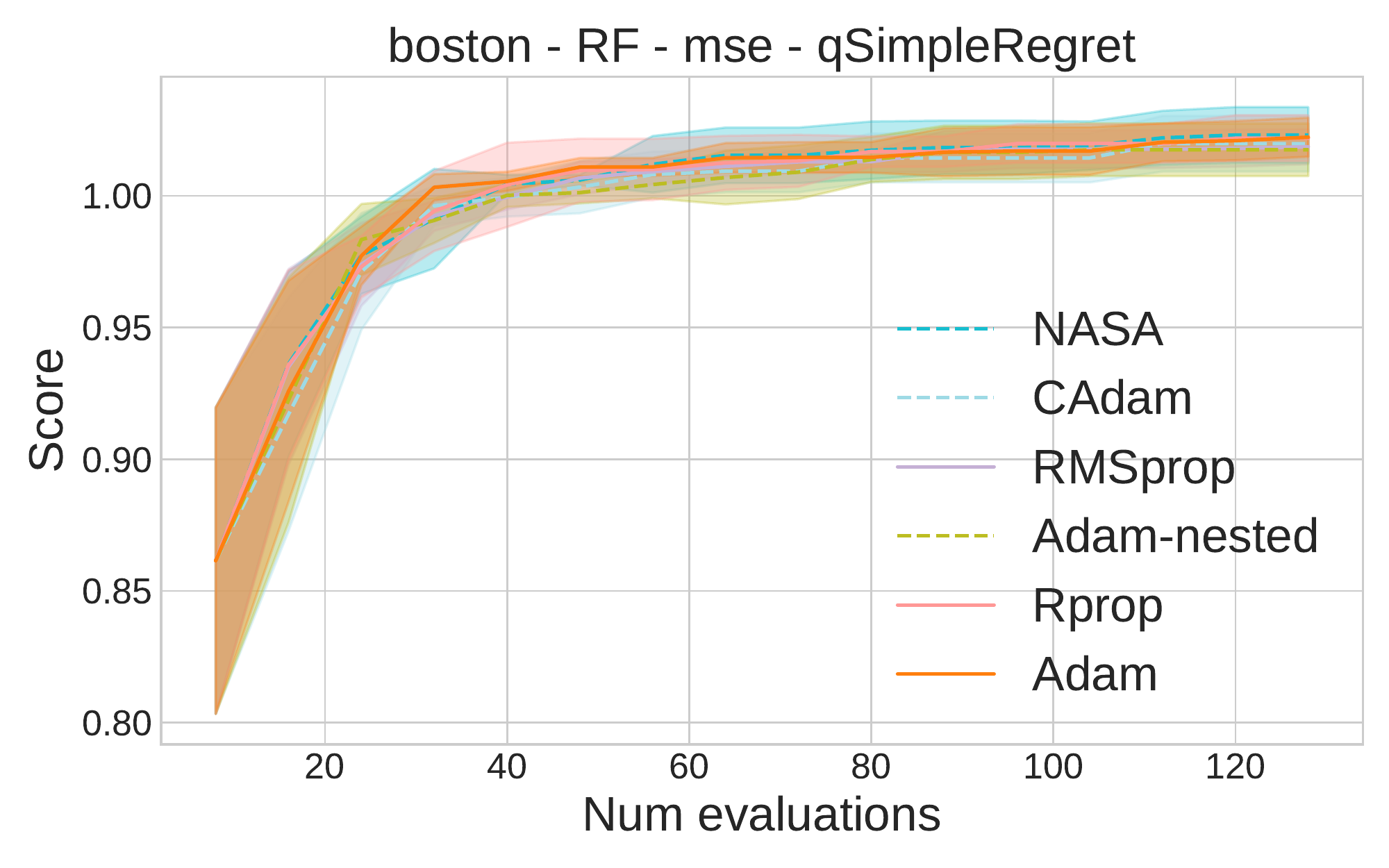}} & \hspace{-0.5cm}
    \subfloat{\includegraphics[width=0.19\columnwidth, trim={0 0.5cm 0 0.4cm}, clip]{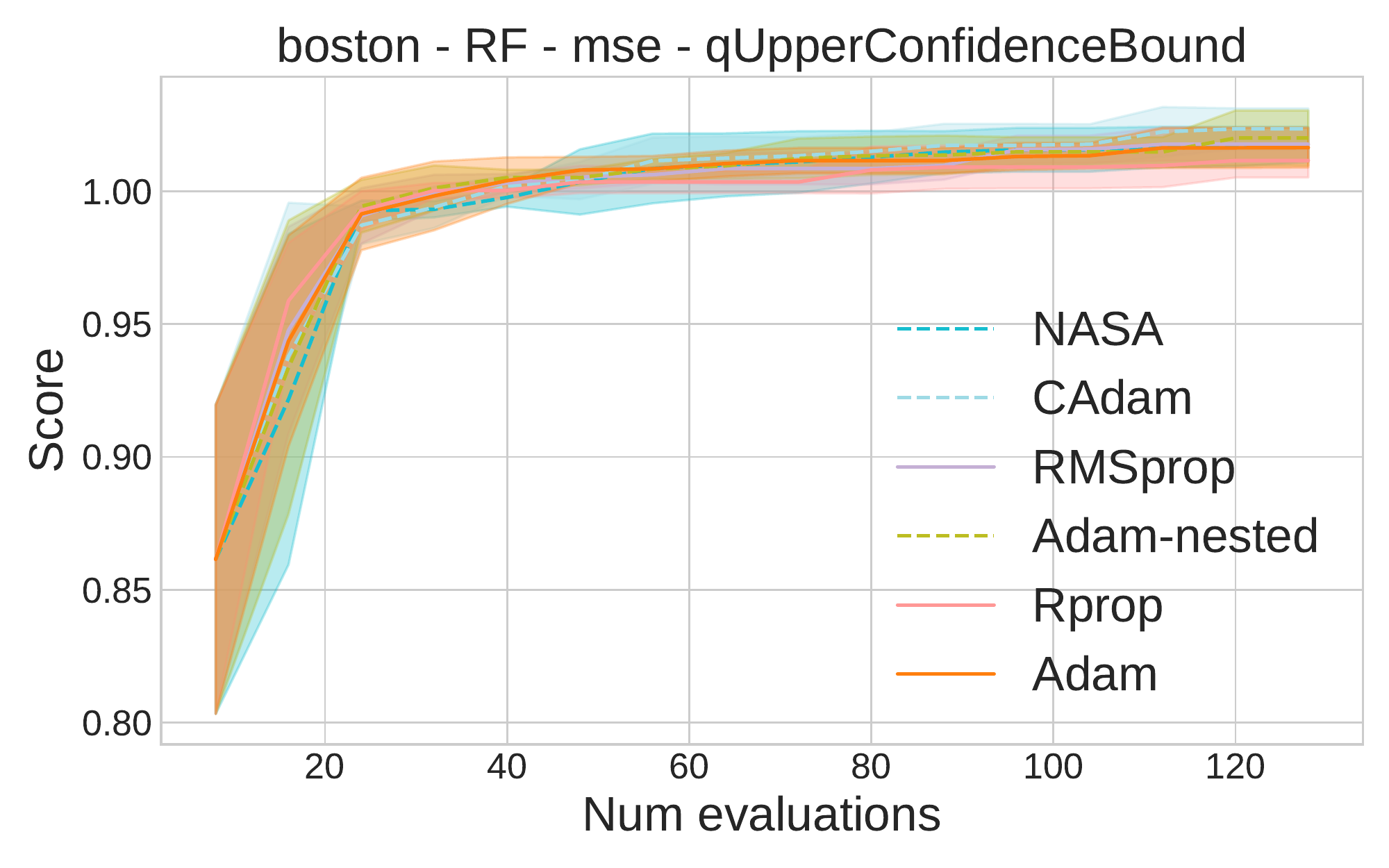}} &  \hspace{-0.5cm}
    \subfloat{\includegraphics[width=0.19\columnwidth, trim={0 0.5cm 0 0.4cm}, clip]{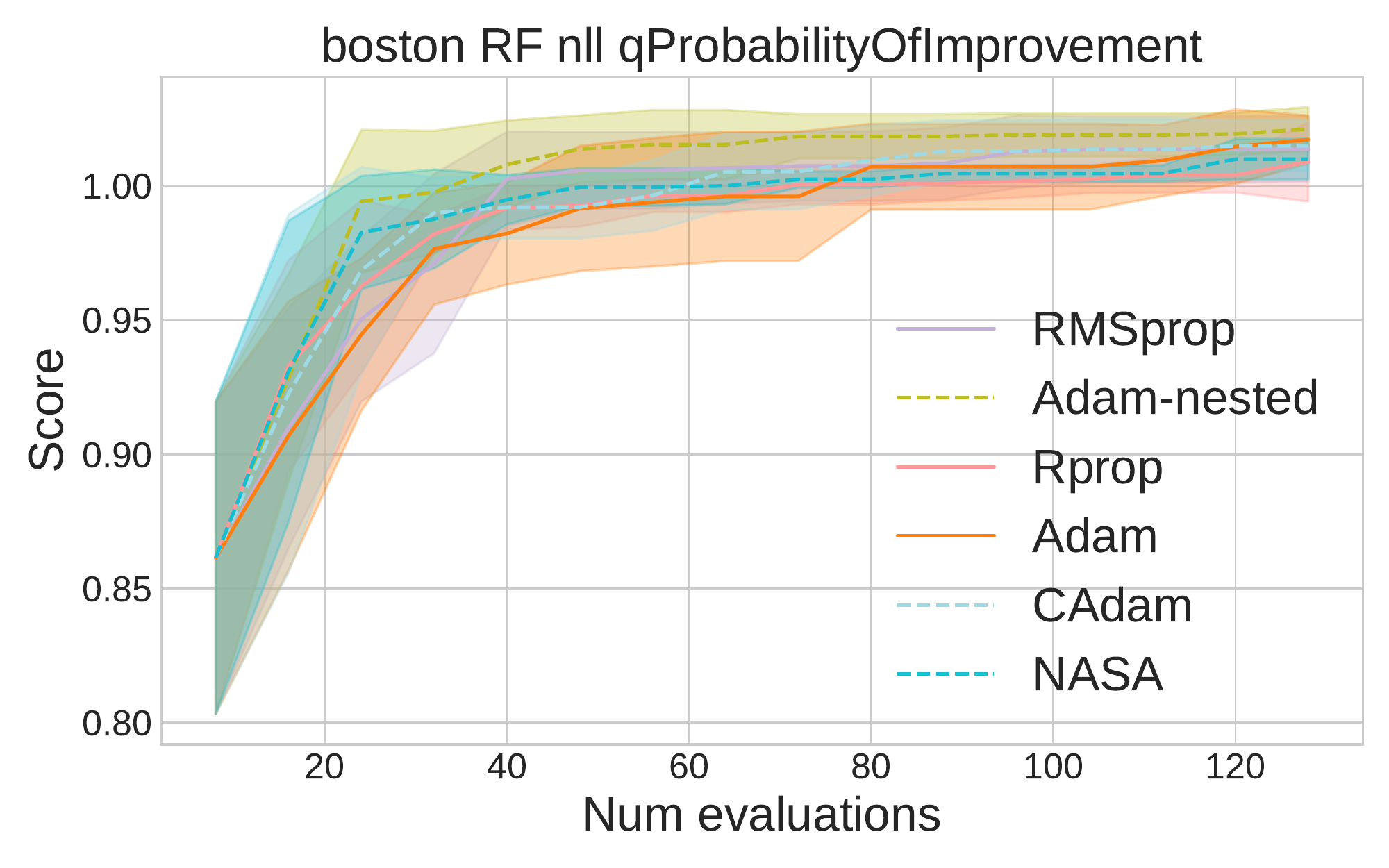}} & \hspace{-0.5cm}
    \subfloat{\includegraphics[width=0.19\columnwidth, trim={0 0.5cm 0 0.4cm}, clip]{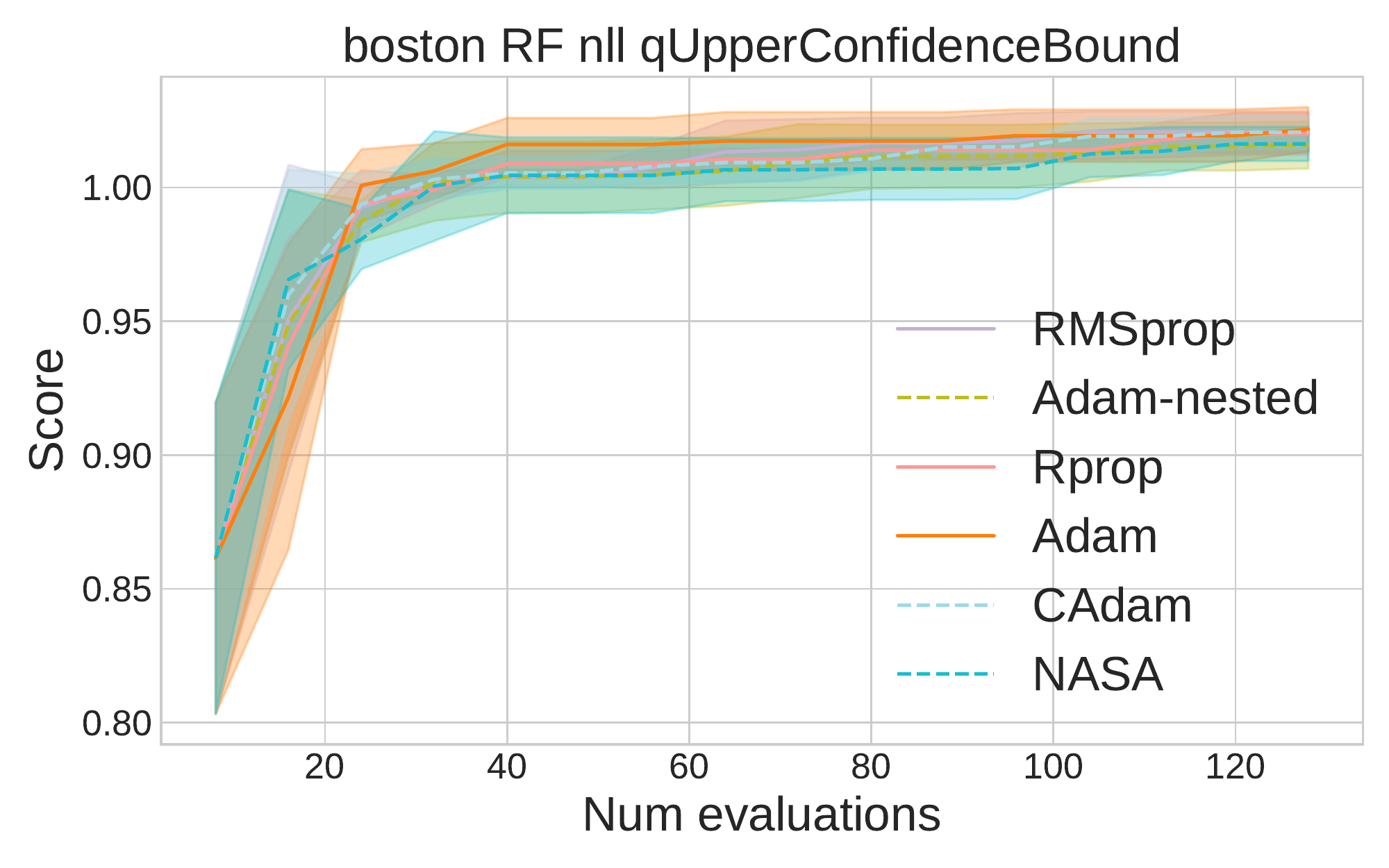}} \\  
    \subfloat{\includegraphics[width=0.19\columnwidth, trim={0 0.5cm 0 0.4cm}, clip]{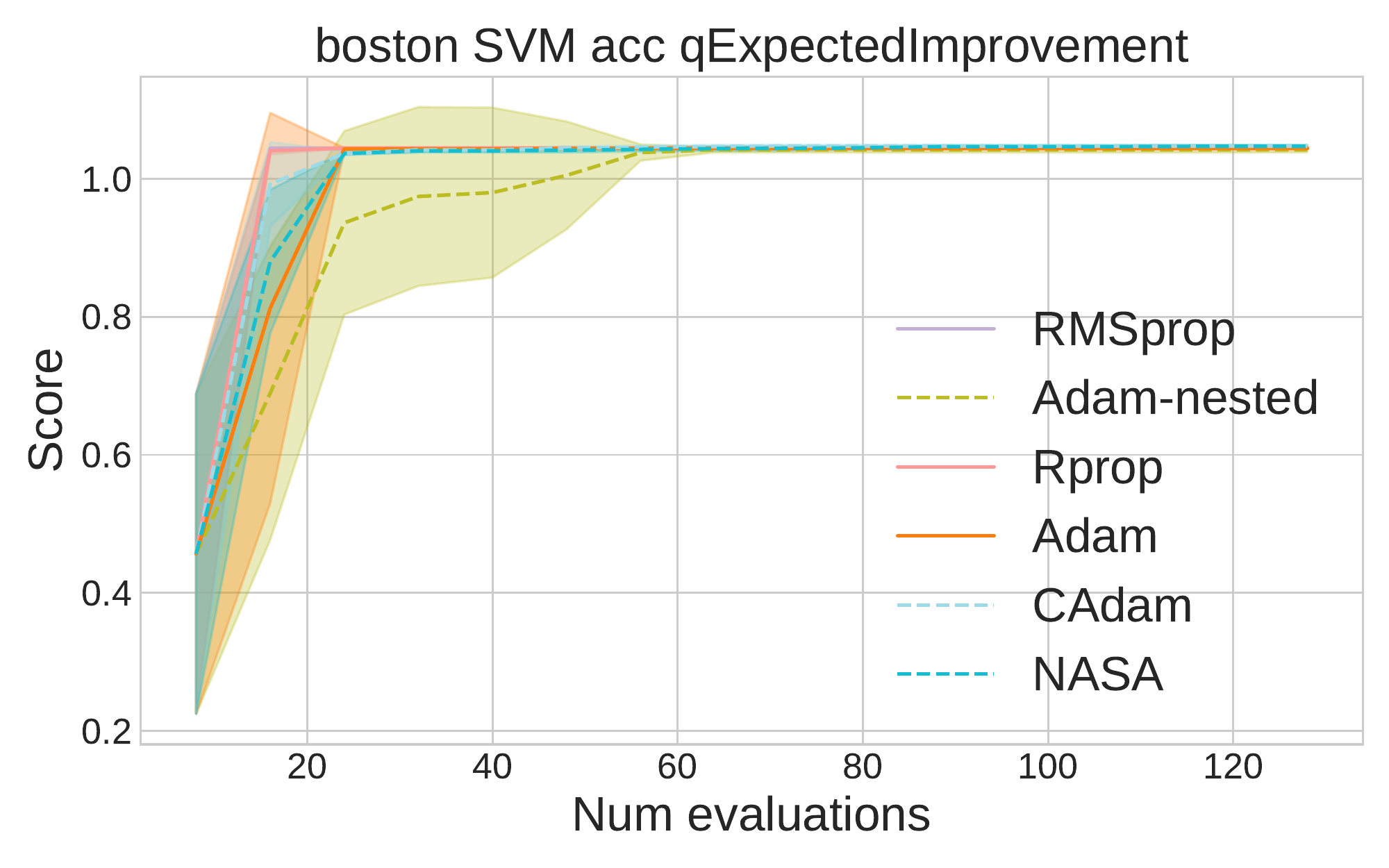}} &   \hspace{-0.5cm} 
    \subfloat{\includegraphics[width=0.19\columnwidth, trim={0 0.5cm 0 0.4cm}, clip]{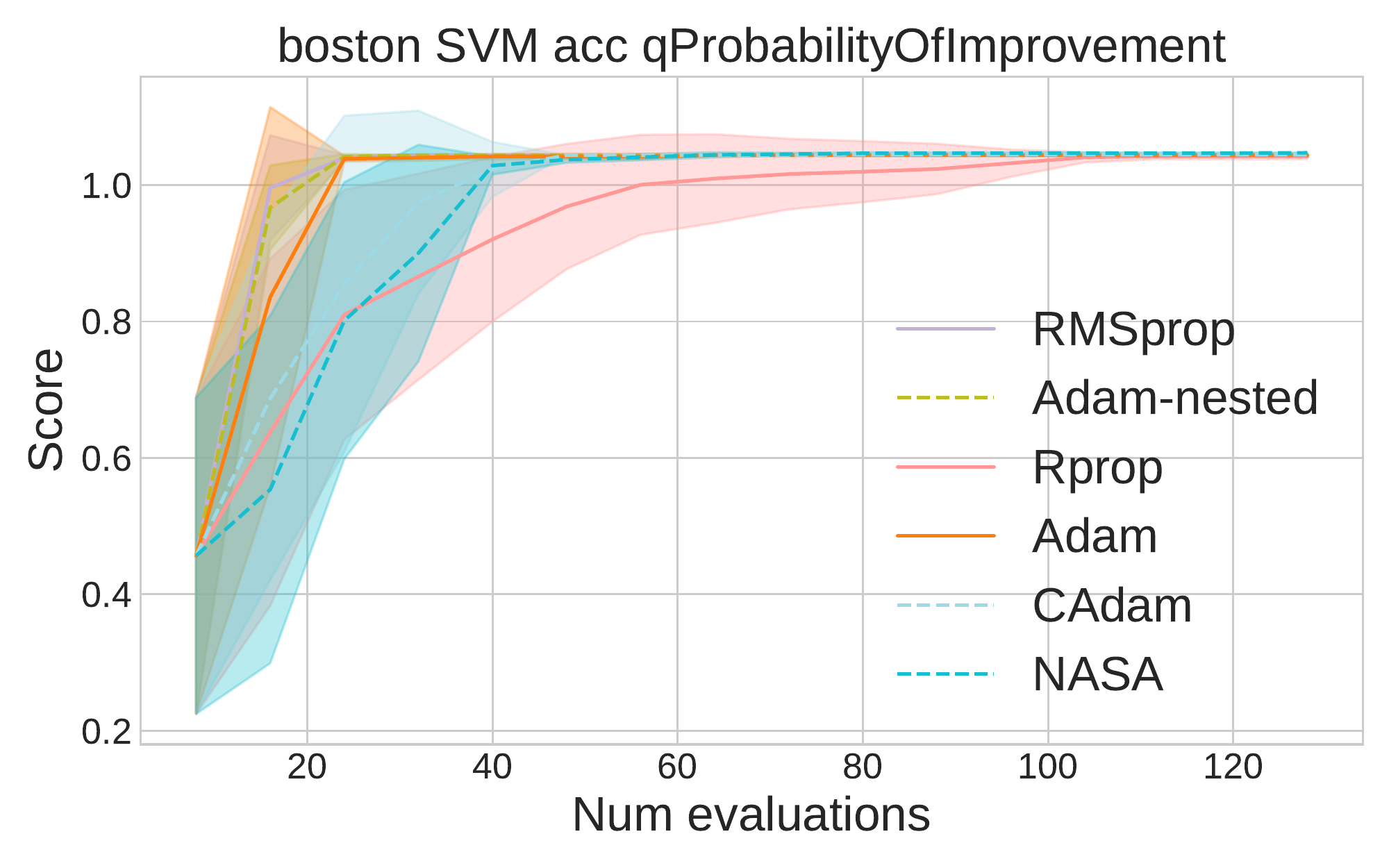}} & \hspace{-0.5cm}
    \subfloat{\includegraphics[width=0.19\columnwidth, trim={0 0.5cm 0 0.4cm}, clip]{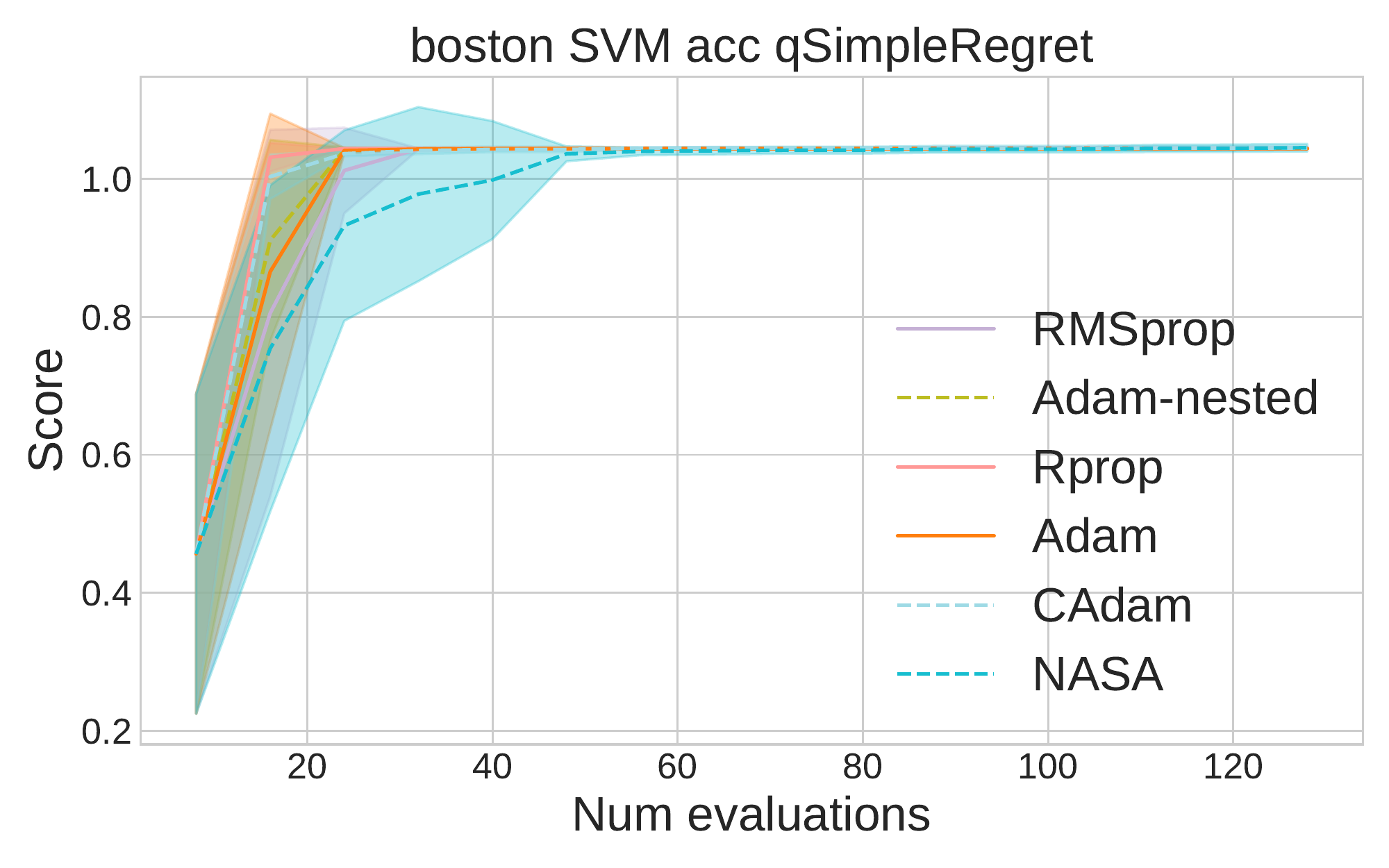}} &  \hspace{-0.5cm}
    \subfloat{\includegraphics[width=0.19\columnwidth, trim={0 0.5cm 0 0.4cm}, clip]{images/boston_RF_nll_qProbabilityOfImprovement_optcount=6}} & \hspace{-0.5cm}
    \subfloat{\includegraphics[width=0.19\columnwidth, trim={0 0.5cm 0 0.4cm}, clip]{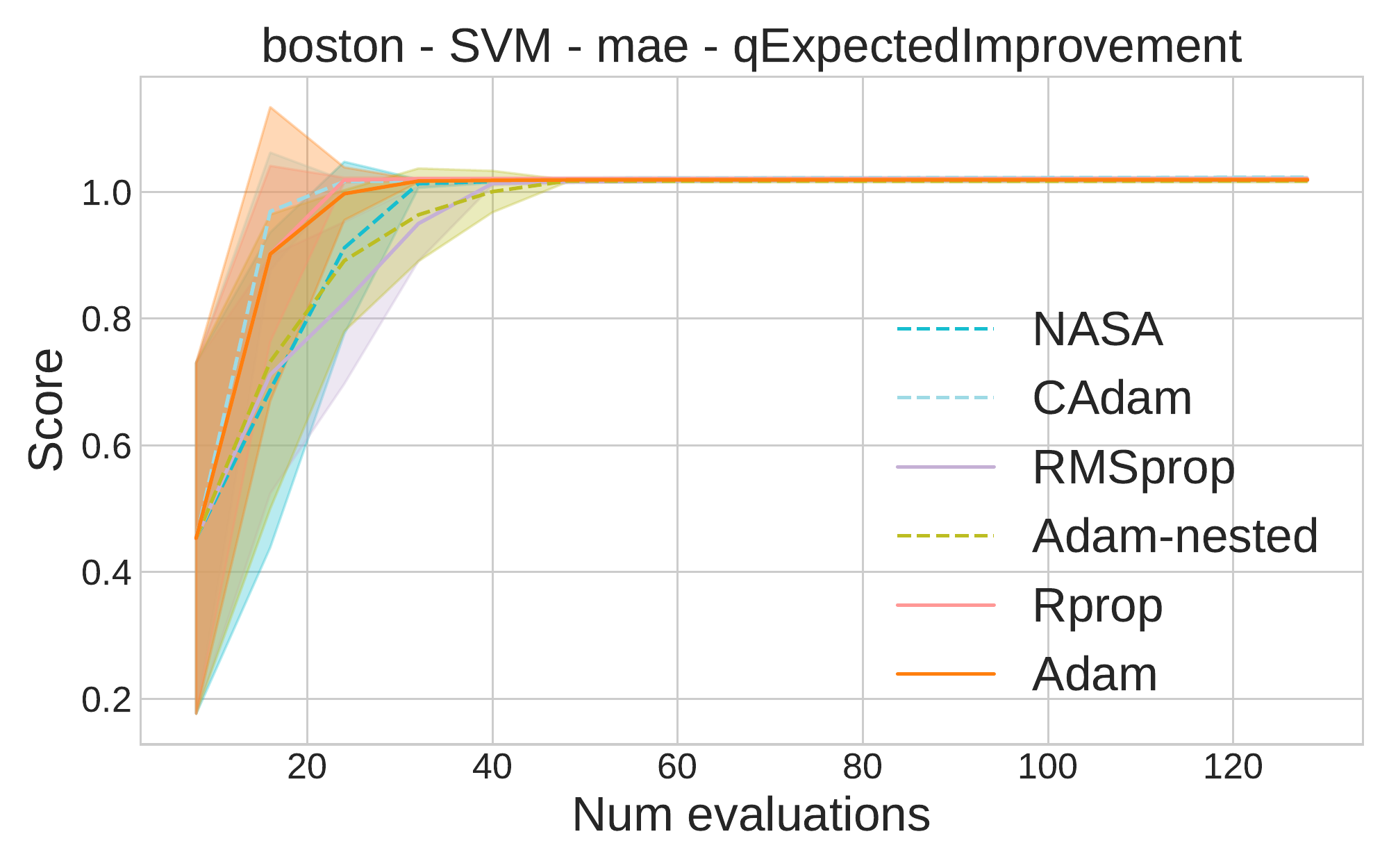}} 
    \end{tabular}
    \caption{}
    \end{figure}
    }
    
    {\renewcommand{\arraystretch}{0}
    \begin{figure}
    \begin{tabular}{ccccc}
    \subfloat{\includegraphics[width=0.19\columnwidth, trim={0 0.5cm 0 0.4cm}, clip]{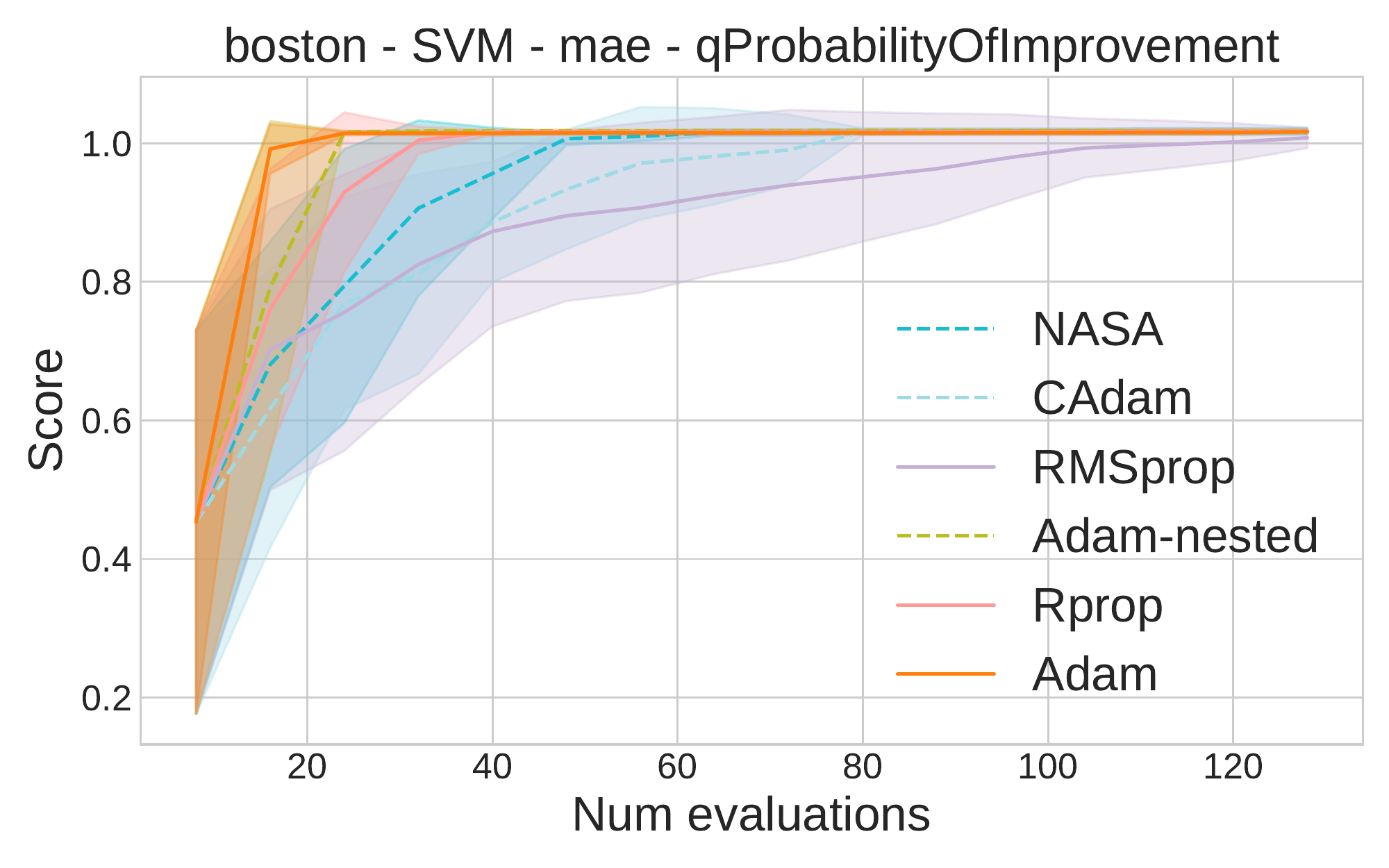}} &  \hspace{-0.5cm}
    \subfloat{\includegraphics[width=0.19\columnwidth, trim={0 0.5cm 0 0.4cm}, clip]{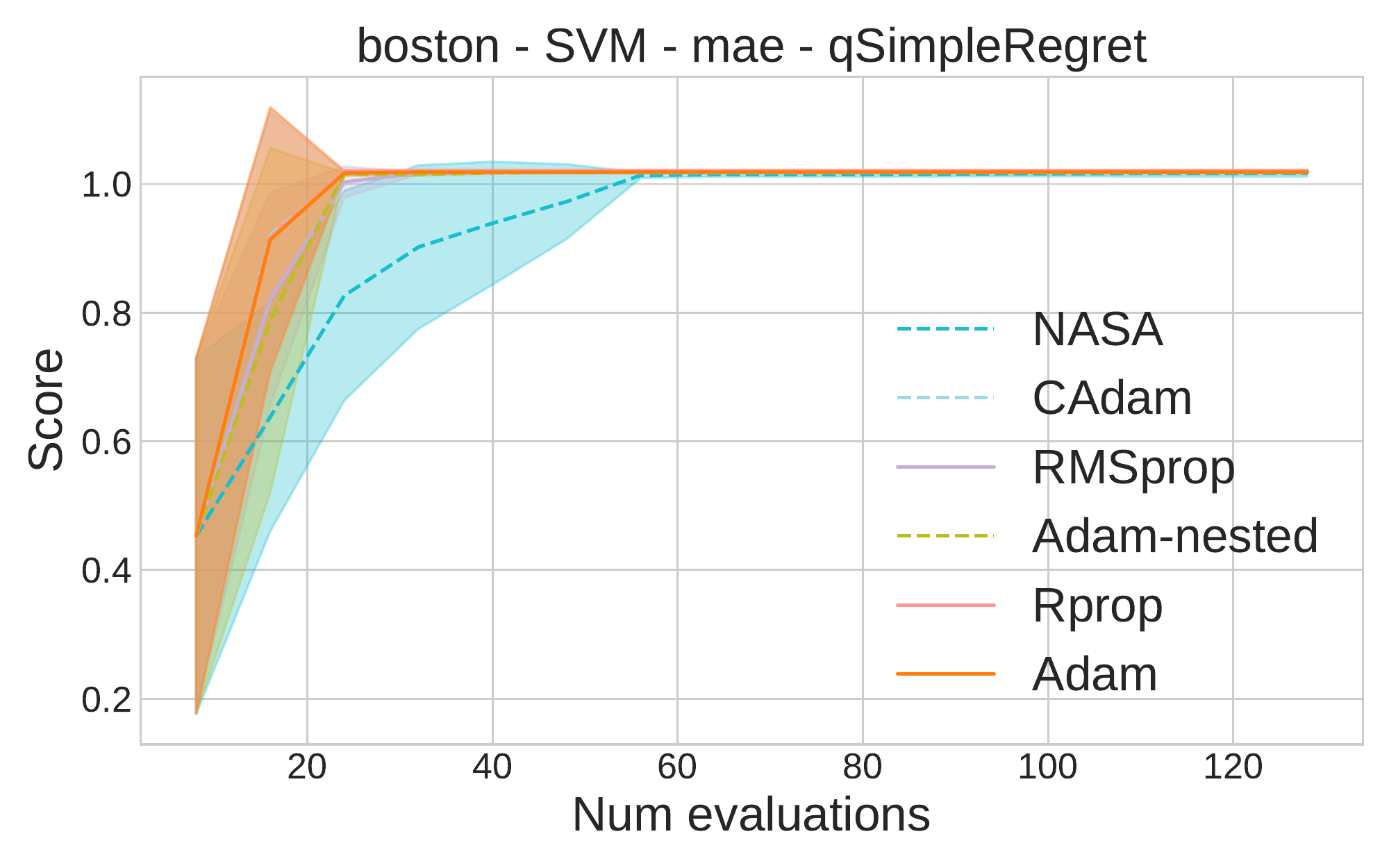}} & \hspace{-0.5cm}
    \subfloat{\includegraphics[width=0.19\columnwidth, trim={0 0.5cm 0 0.4cm}, clip]{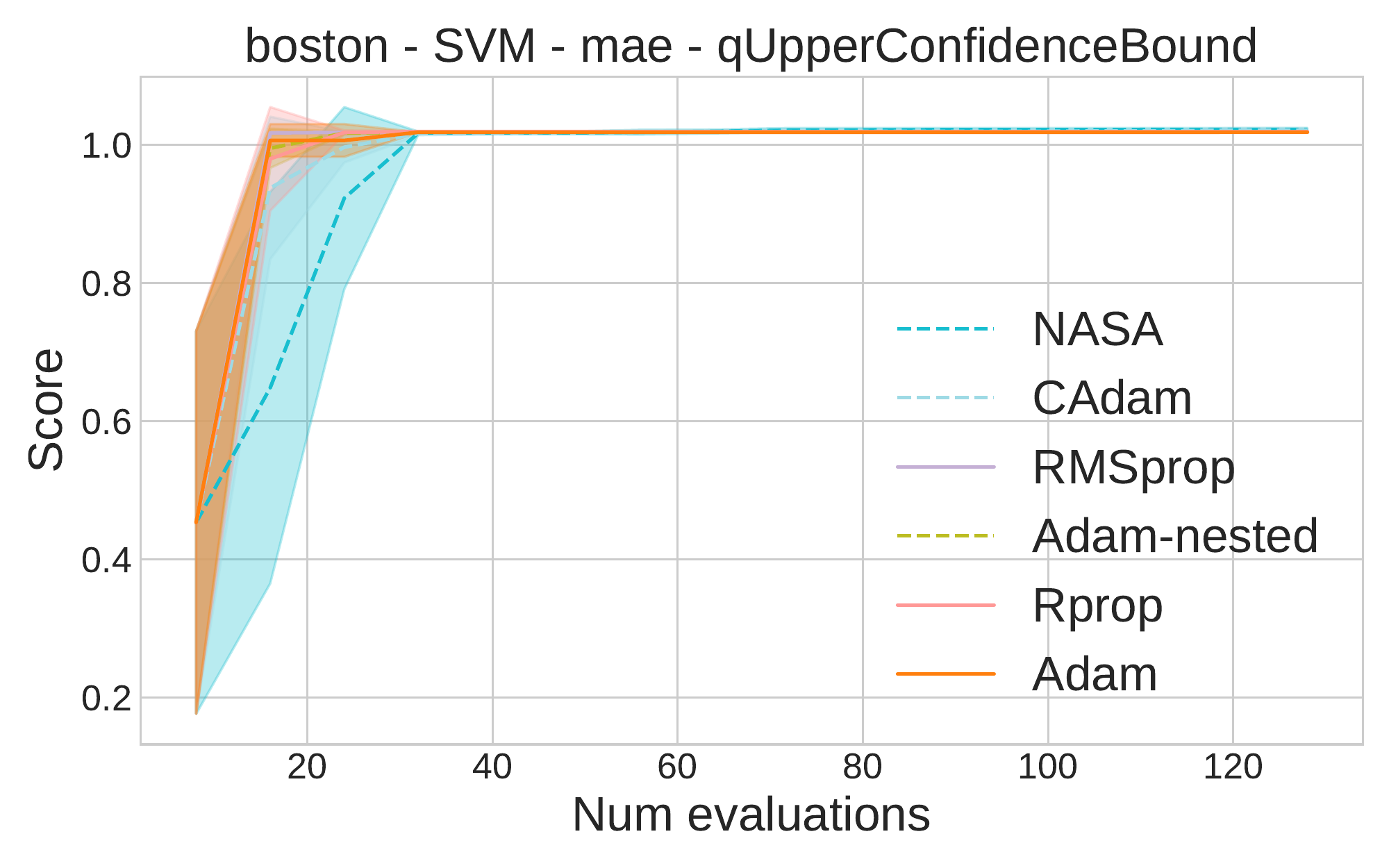}} &  \hspace{-0.5cm}
    \subfloat{\includegraphics[width=0.19\columnwidth, trim={0 0.5cm 0 0.4cm}, clip]{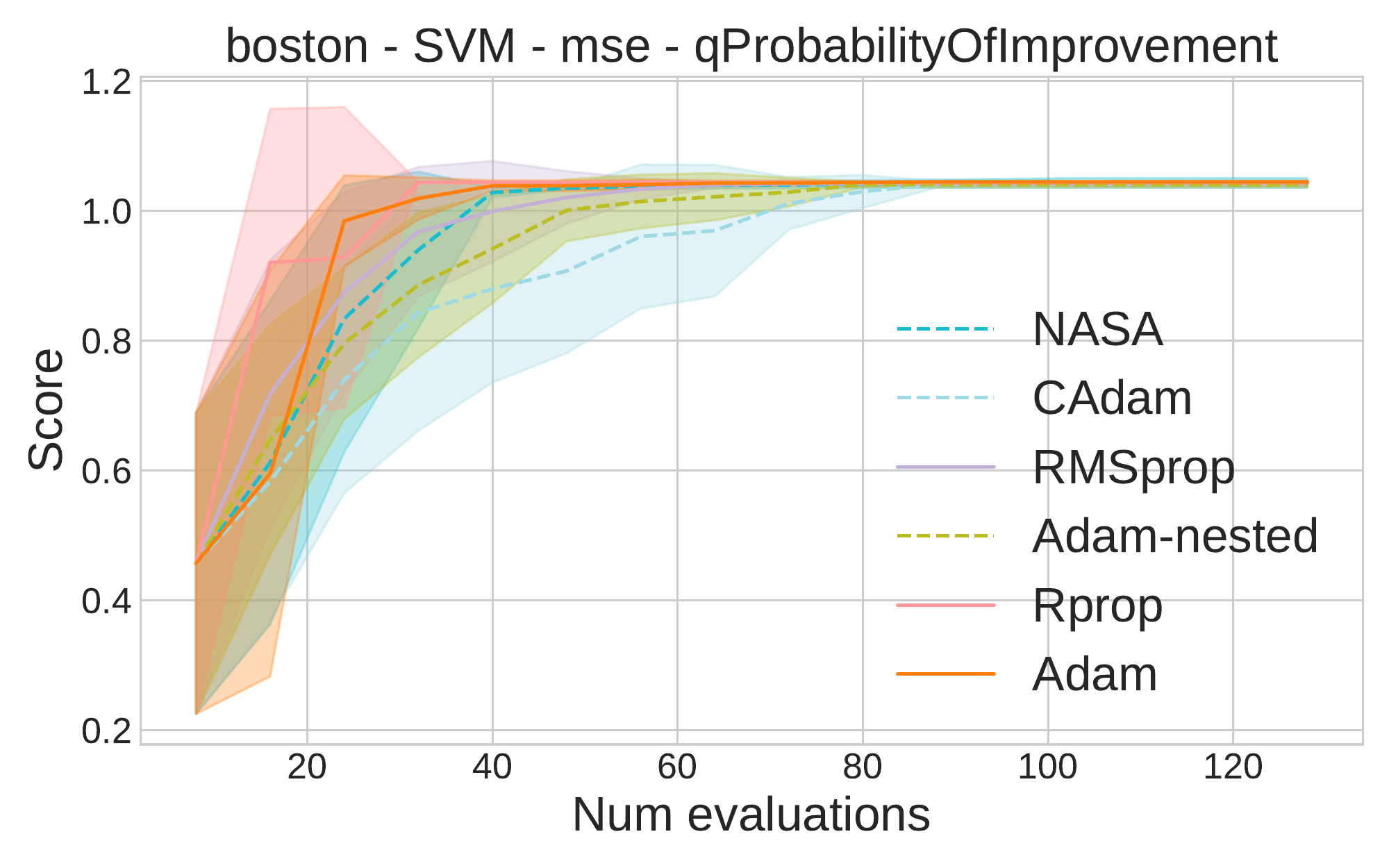}} & \hspace{-0.5cm}
    \subfloat{\includegraphics[width=0.19\columnwidth, trim={0 0.5cm 0 0.4cm}, clip]{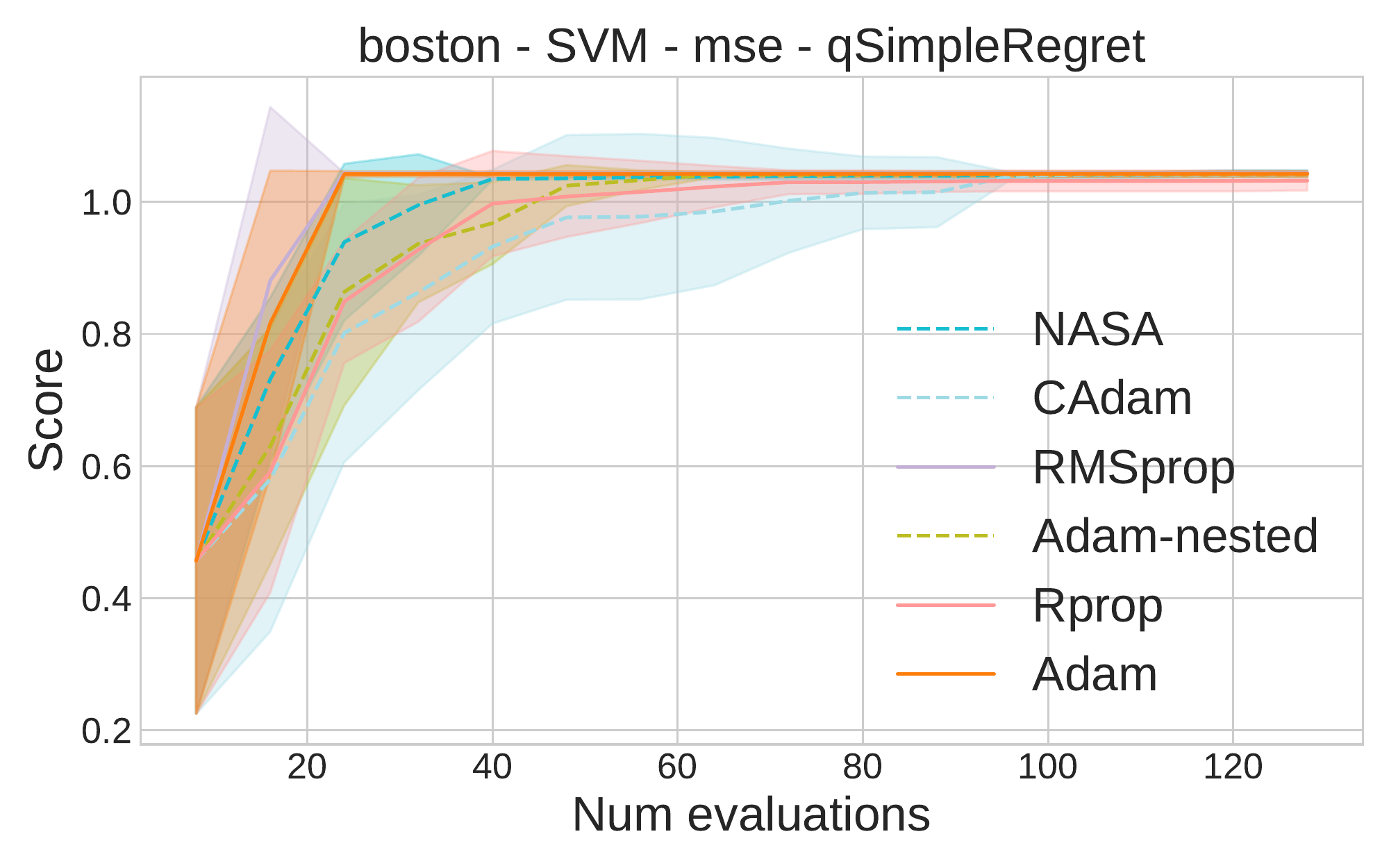}} \\  
    \subfloat{\includegraphics[width=0.19\columnwidth, trim={0 0.5cm 0 0.4cm}, clip]{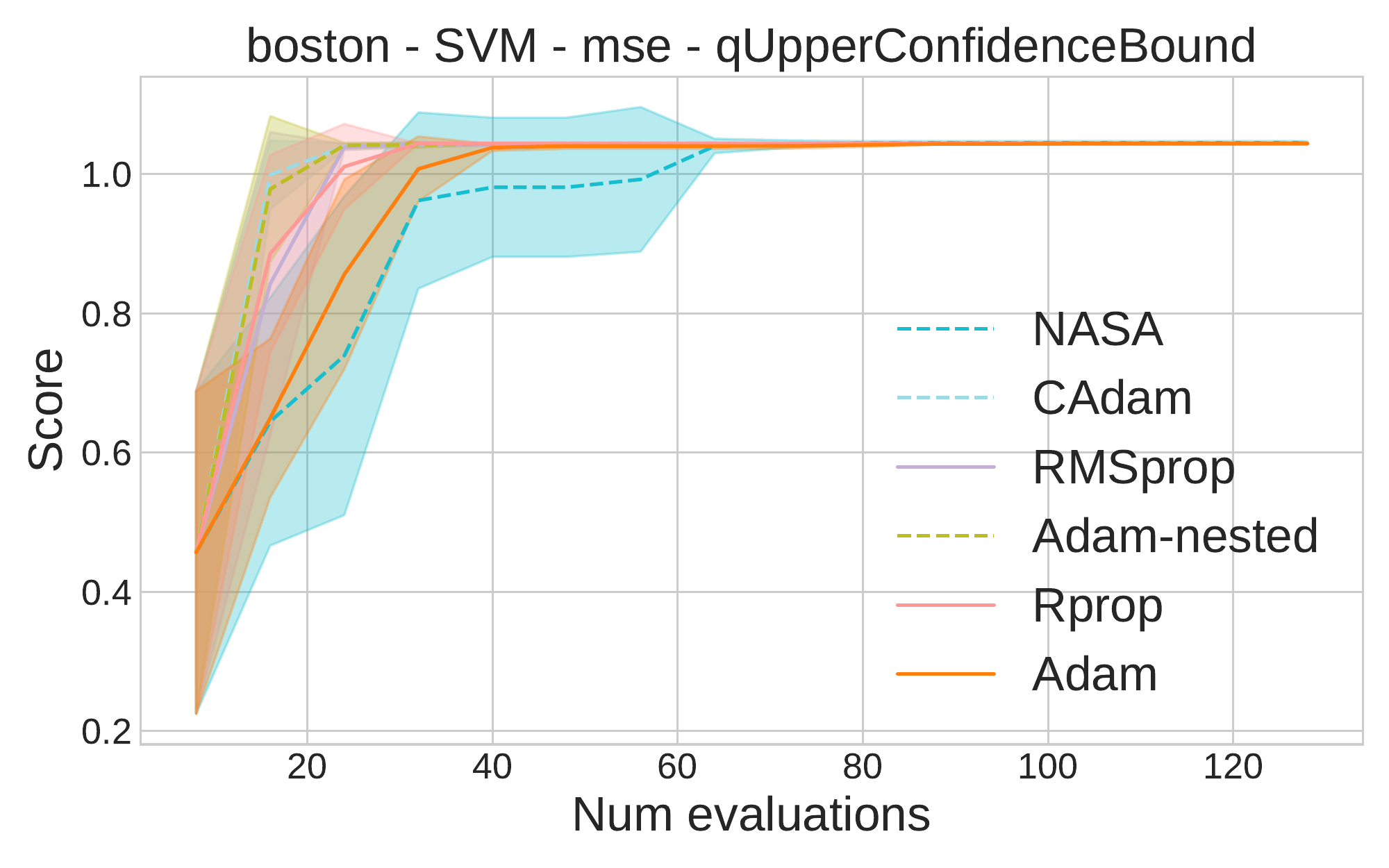}} &   \hspace{-0.5cm} 
    \subfloat{\includegraphics[width=0.19\columnwidth, trim={0 0.5cm 0 0.4cm}, clip]{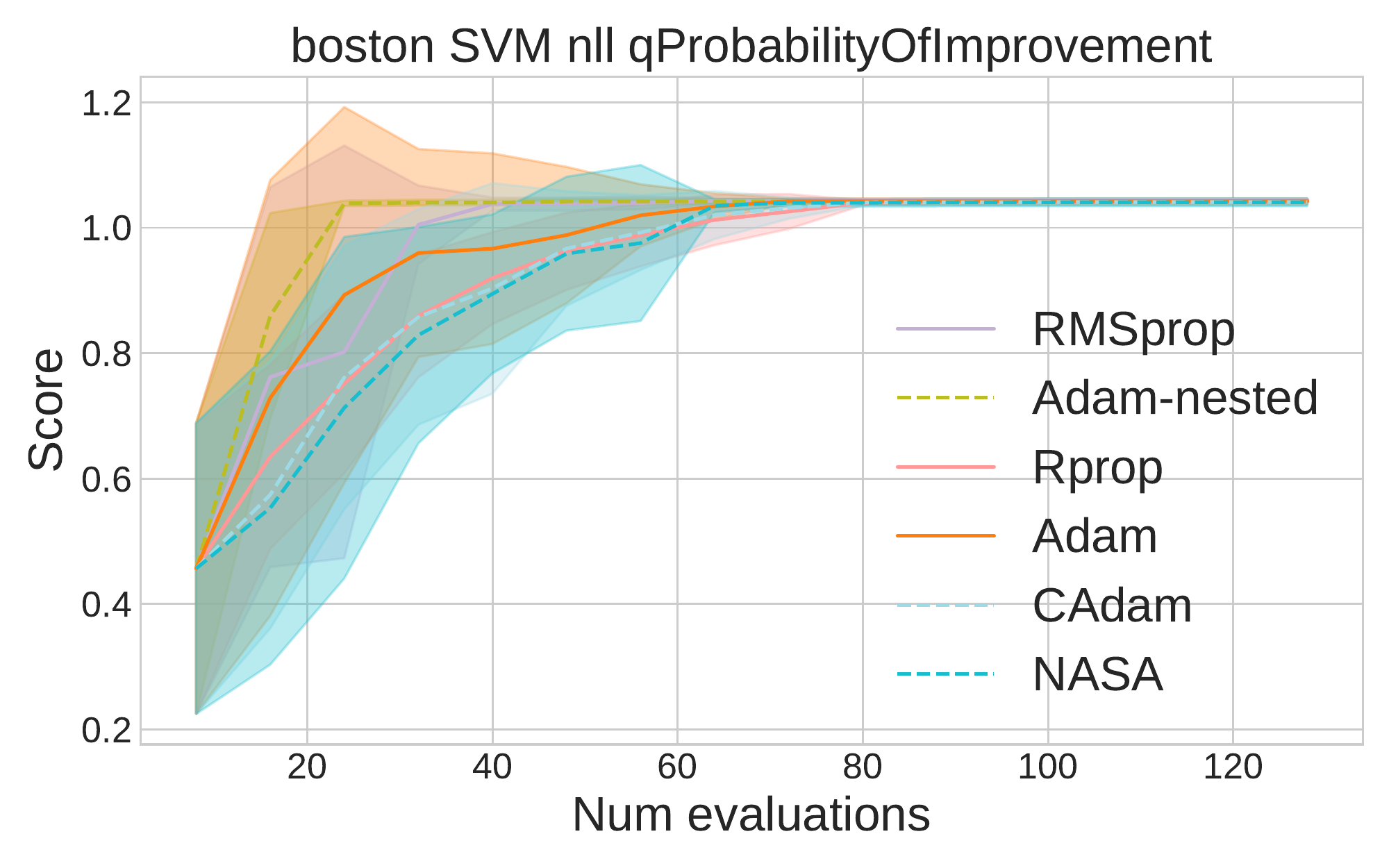}} & \hspace{-0.5cm}
    \subfloat{\includegraphics[width=0.19\columnwidth, trim={0 0.5cm 0 0.4cm}, clip]{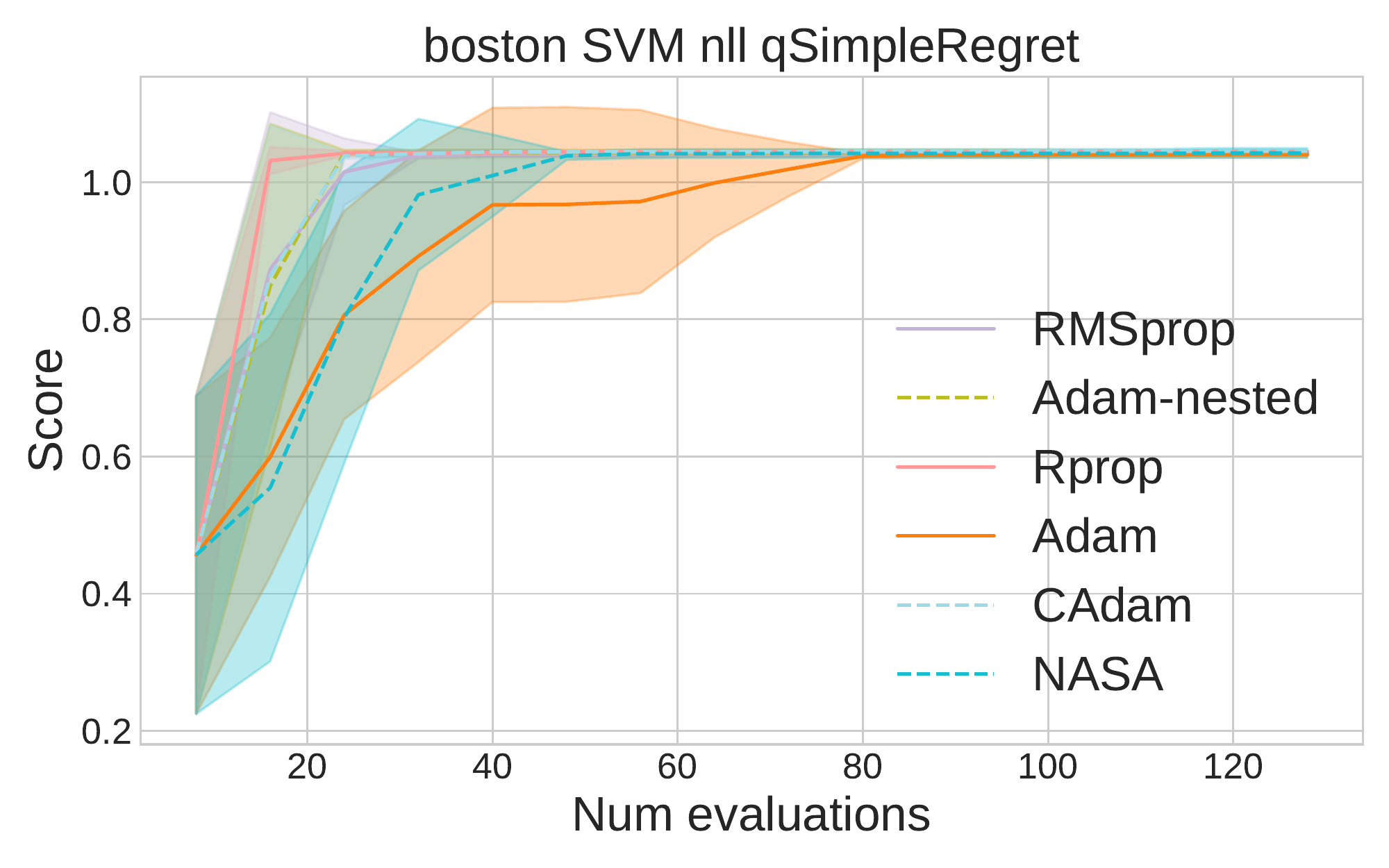}} &  \hspace{-0.5cm}
    \subfloat{\includegraphics[width=0.19\columnwidth, trim={0 0.5cm 0 0.4cm}, clip]{images/boston_SVM_nll_qSimpleRegret_optcount=6}} & \hspace{-0.5cm}
    \subfloat{\includegraphics[width=0.19\columnwidth, trim={0 0.5cm 0 0.4cm}, clip]{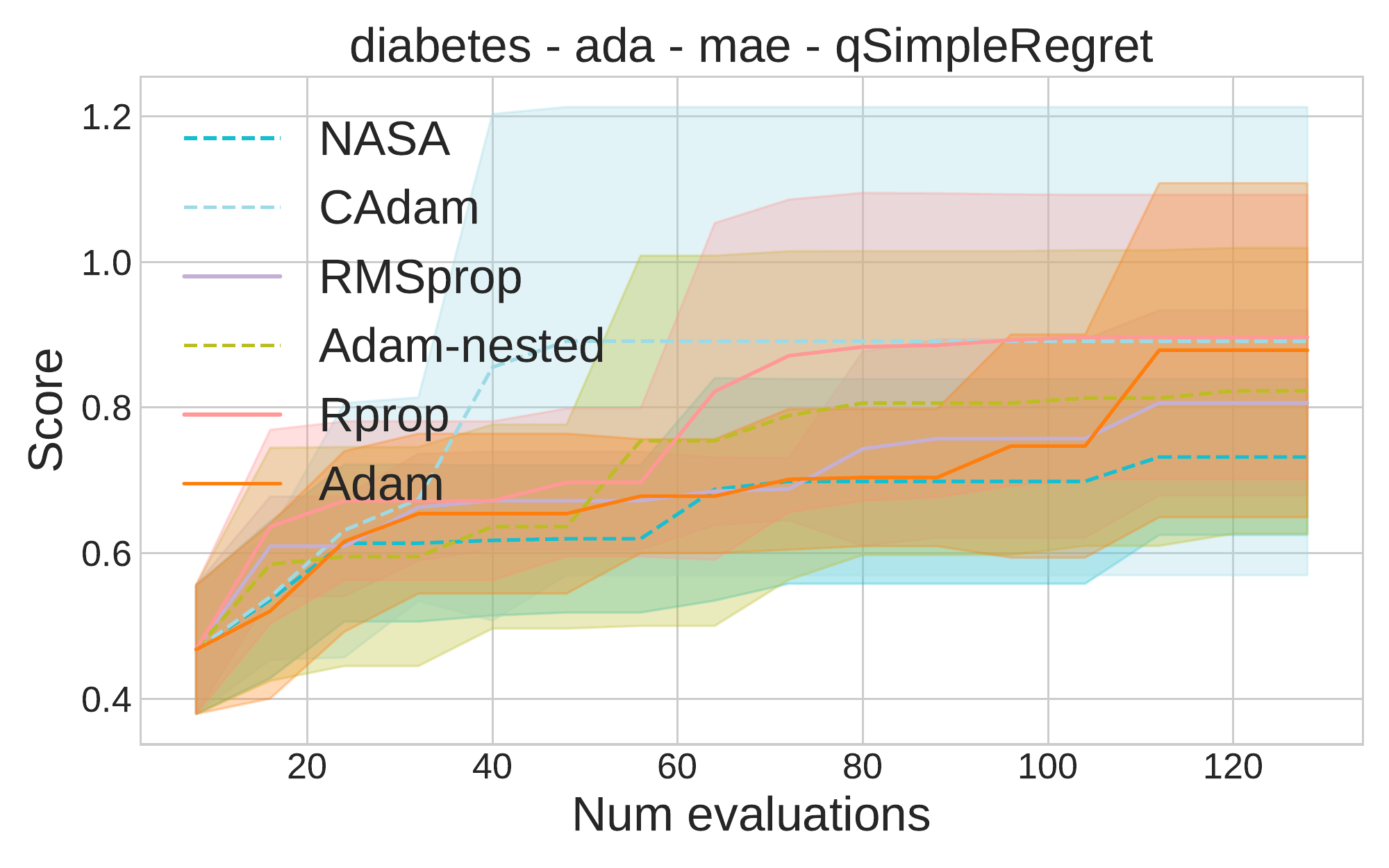}} \\  
    \subfloat{\includegraphics[width=0.19\columnwidth, trim={0 0.5cm 0 0.4cm}, clip]{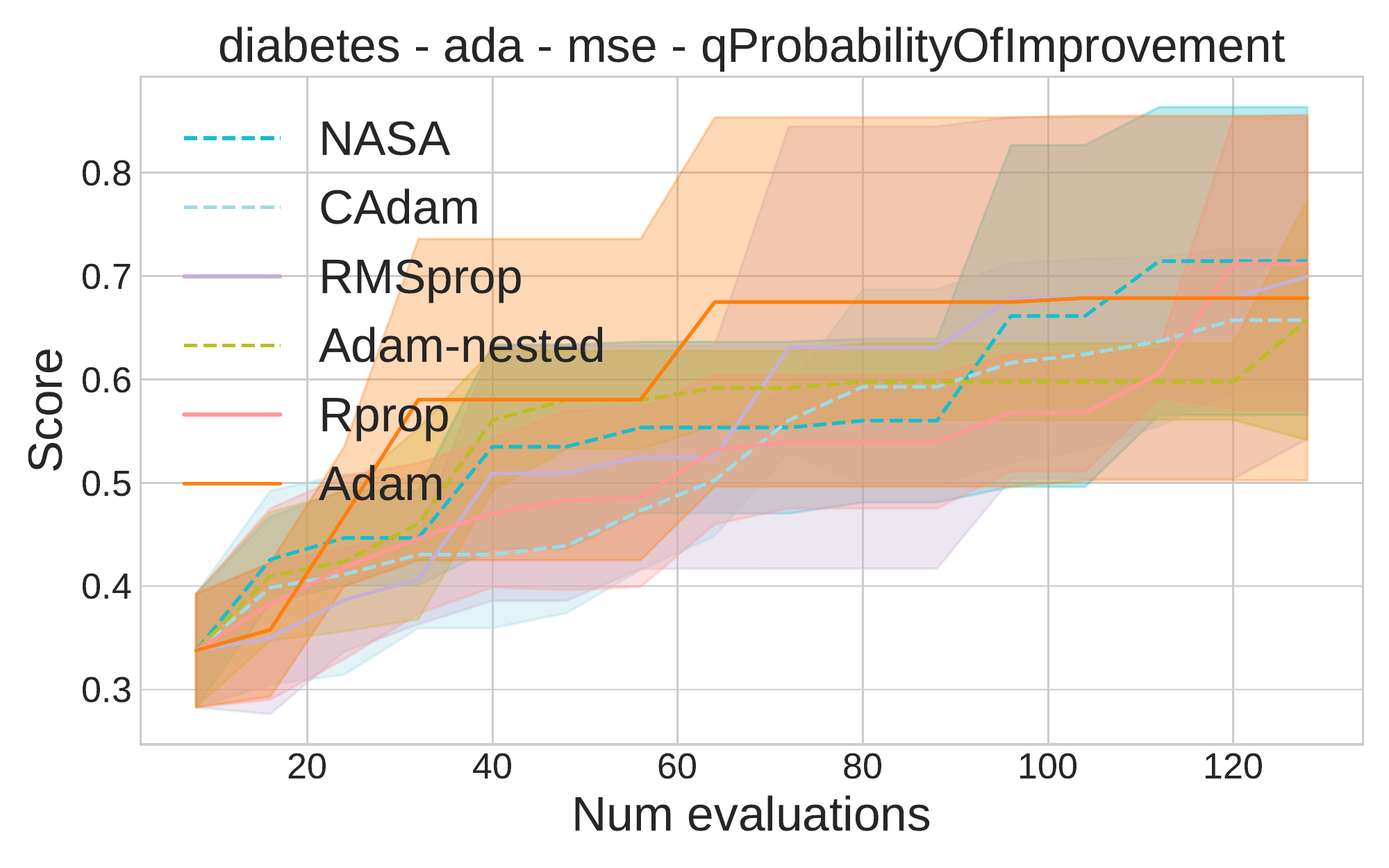}} &   \hspace{-0.5cm} 
    \subfloat{\includegraphics[width=0.19\columnwidth, trim={0 0.5cm 0 0.4cm}, clip]{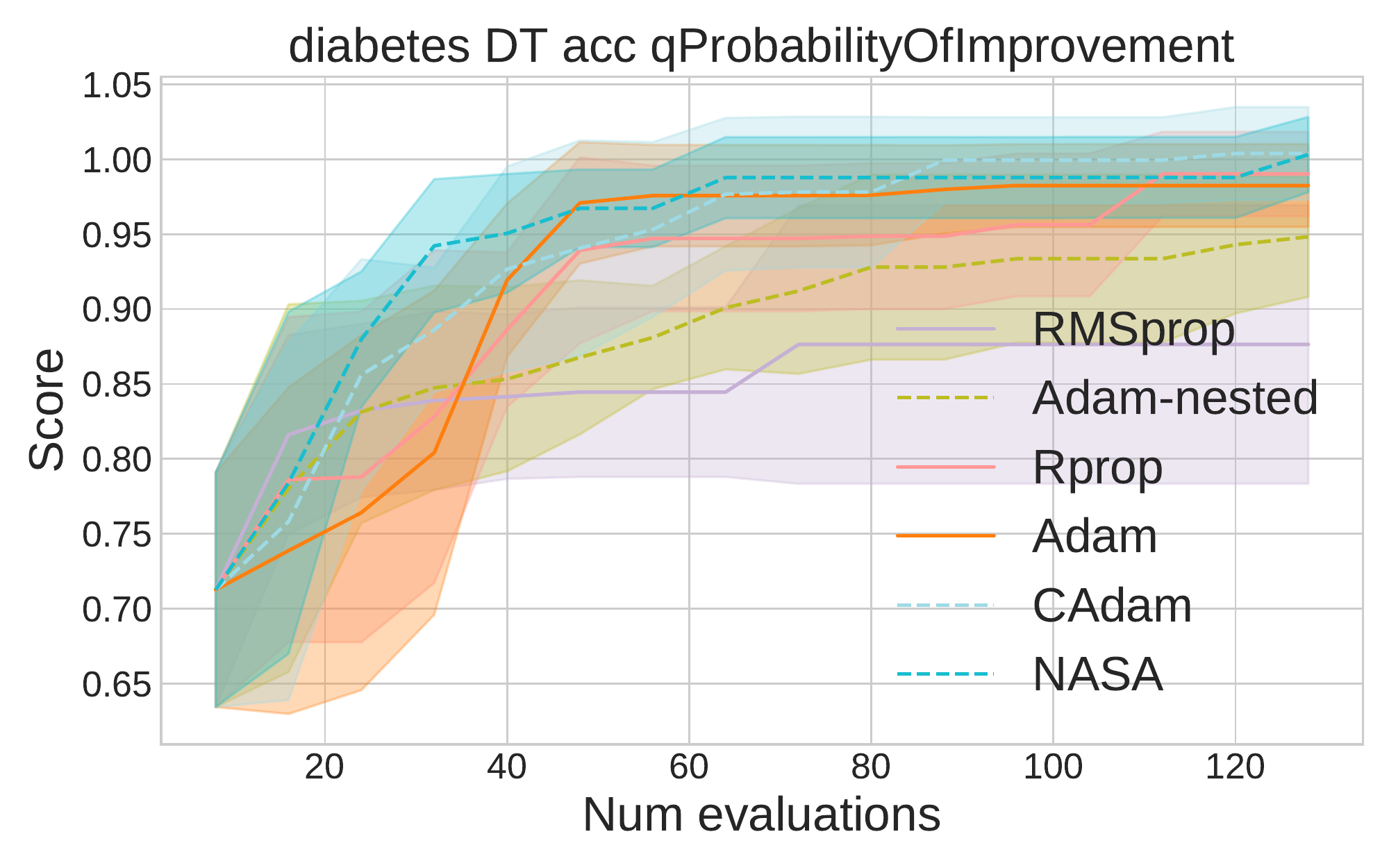}} & \hspace{-0.5cm}
    \subfloat{\includegraphics[width=0.19\columnwidth, trim={0 0.5cm 0 0.4cm}, clip]{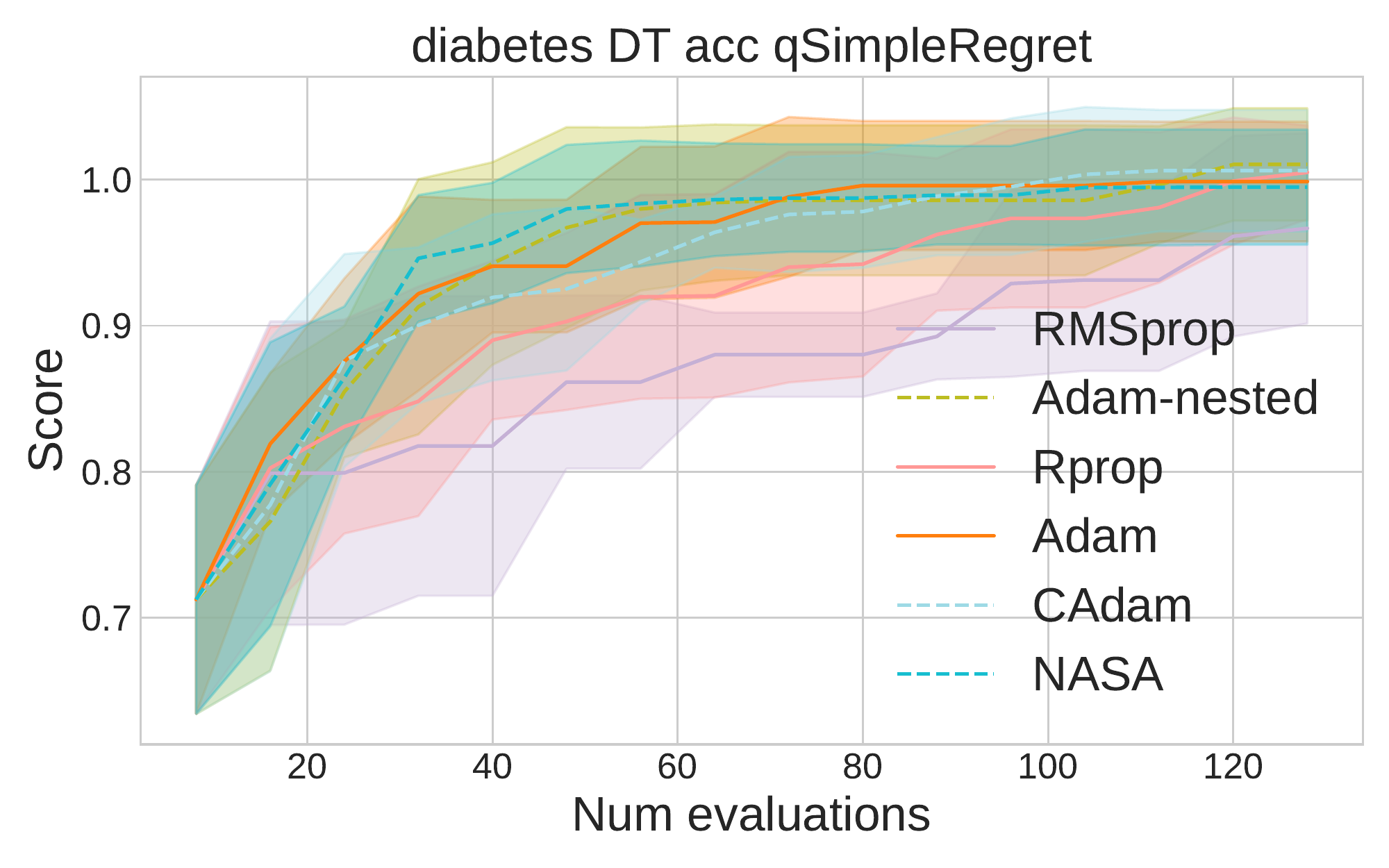}} &  \hspace{-0.5cm}
    \subfloat{\includegraphics[width=0.19\columnwidth, trim={0 0.5cm 0 0.4cm}, clip]{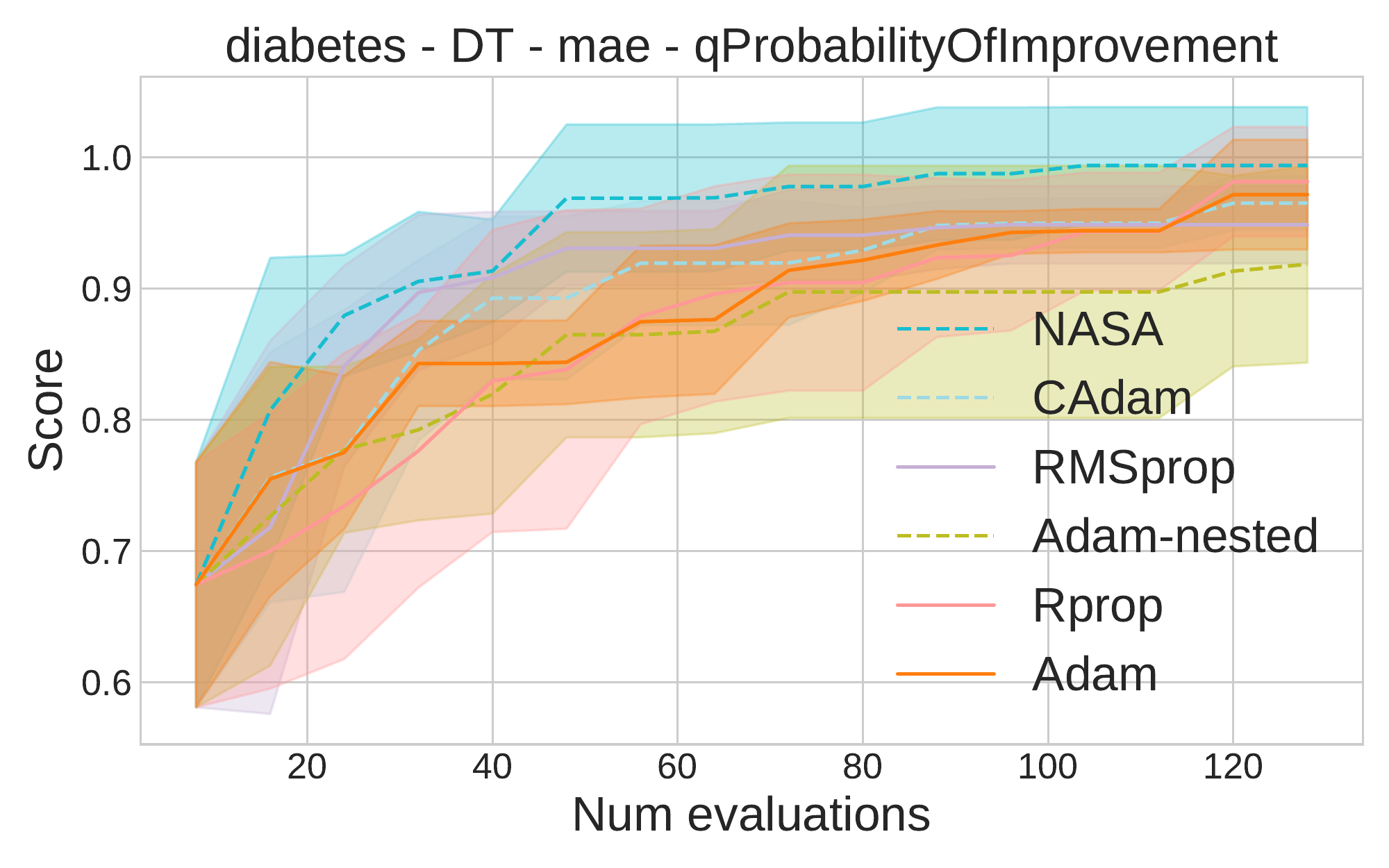}} & \hspace{-0.5cm}
    \subfloat{\includegraphics[width=0.19\columnwidth, trim={0 0.5cm 0 0.4cm}, clip]{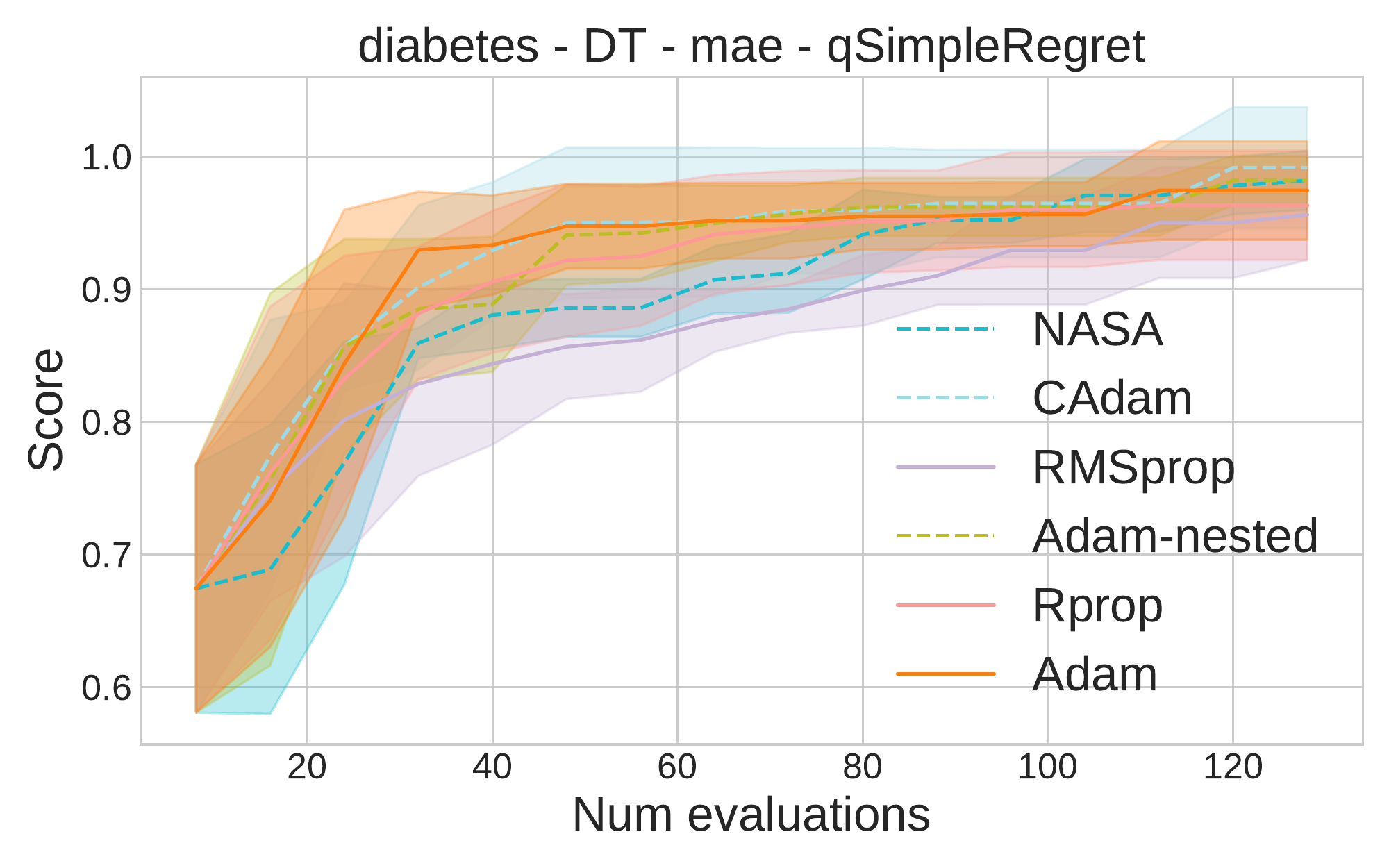}} \\  
    \subfloat{\includegraphics[width=0.19\columnwidth, trim={0 0.5cm 0 0.4cm}, clip]{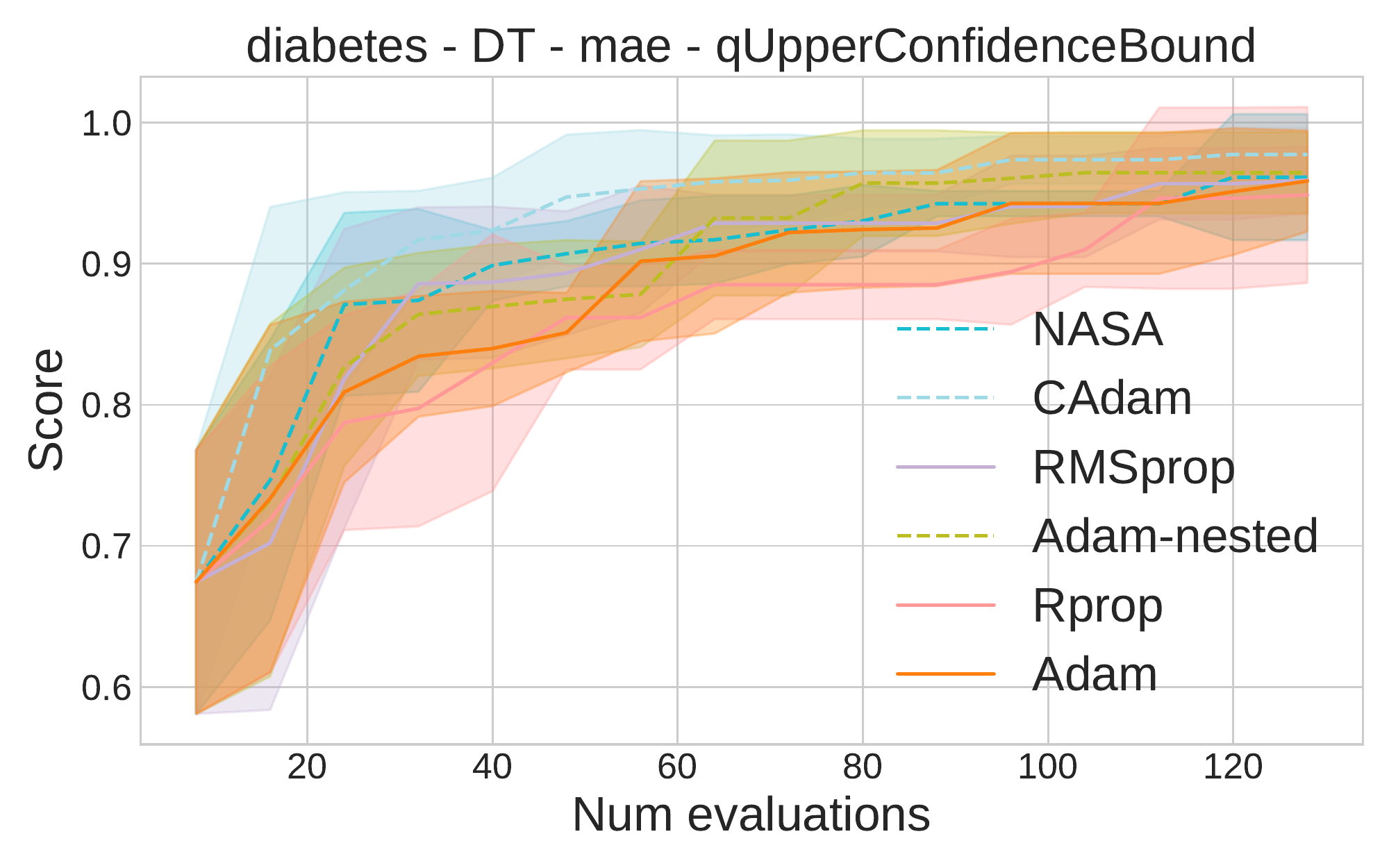}} &   \hspace{-0.5cm} 
    \subfloat{\includegraphics[width=0.19\columnwidth, trim={0 0.5cm 0 0.4cm}, clip]{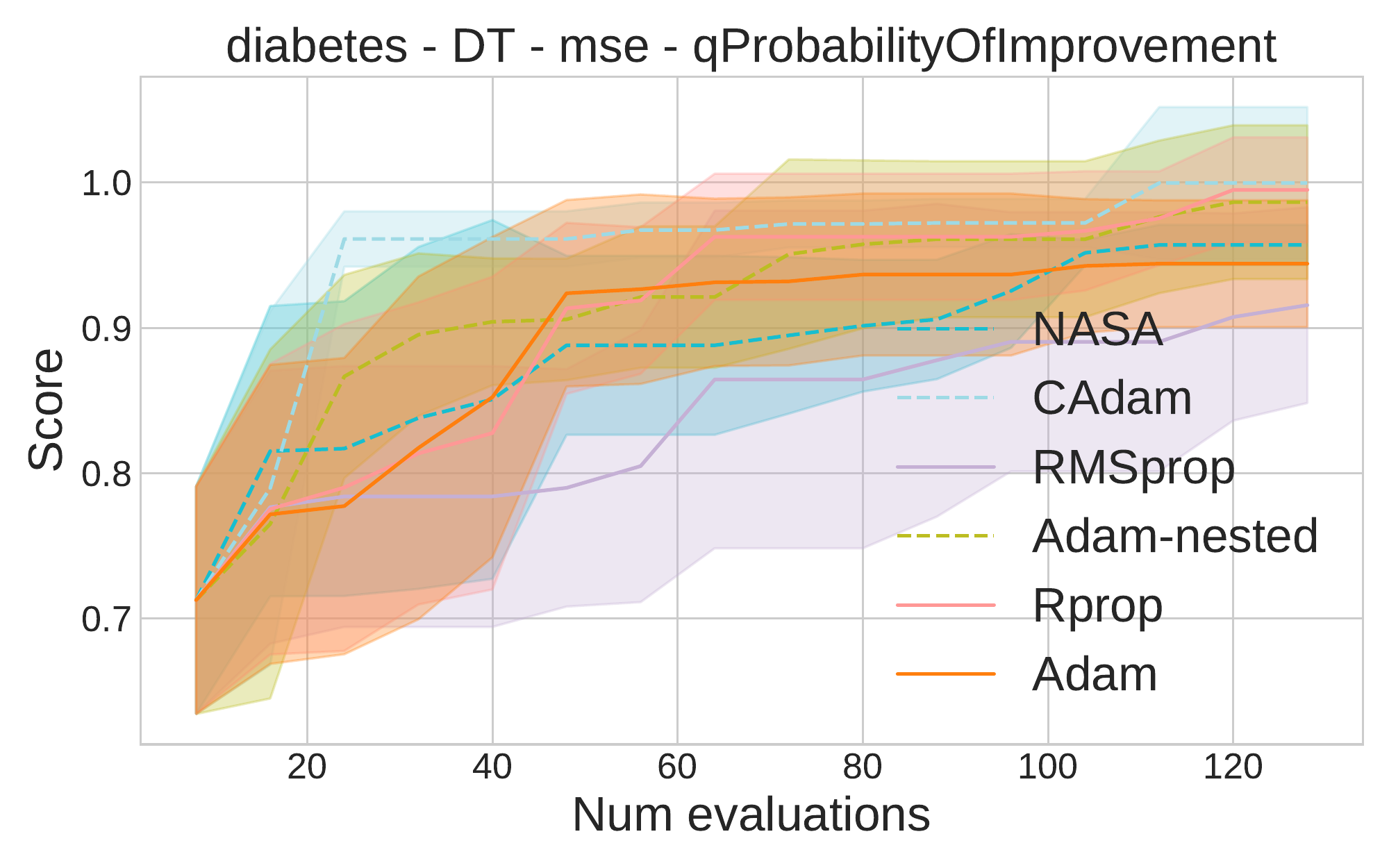}} & \hspace{-0.5cm}
    \subfloat{\includegraphics[width=0.19\columnwidth, trim={0 0.5cm 0 0.4cm}, clip]{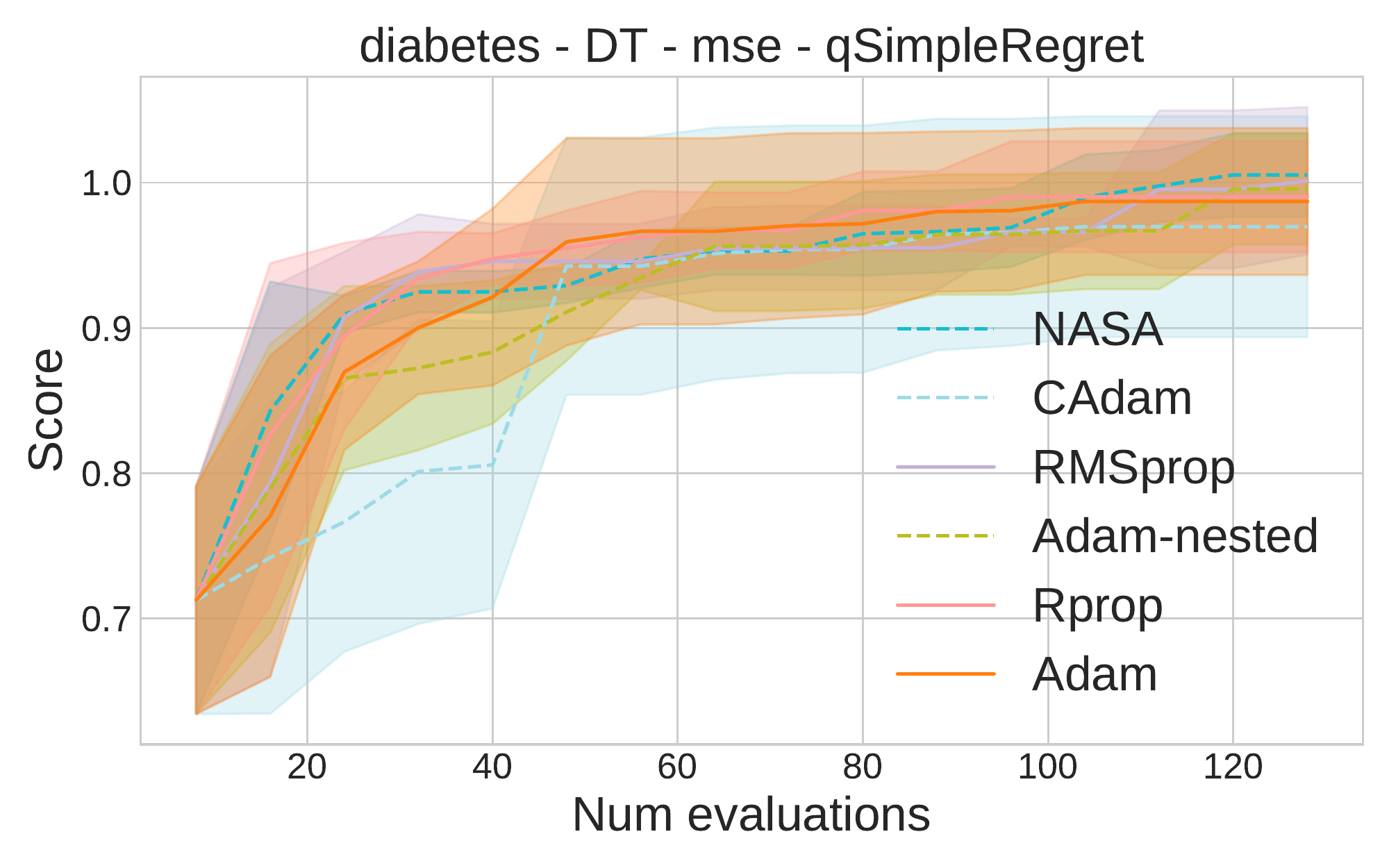}} &  \hspace{-0.5cm}
    \subfloat{\includegraphics[width=0.19\columnwidth, trim={0 0.5cm 0 0.4cm}, clip]{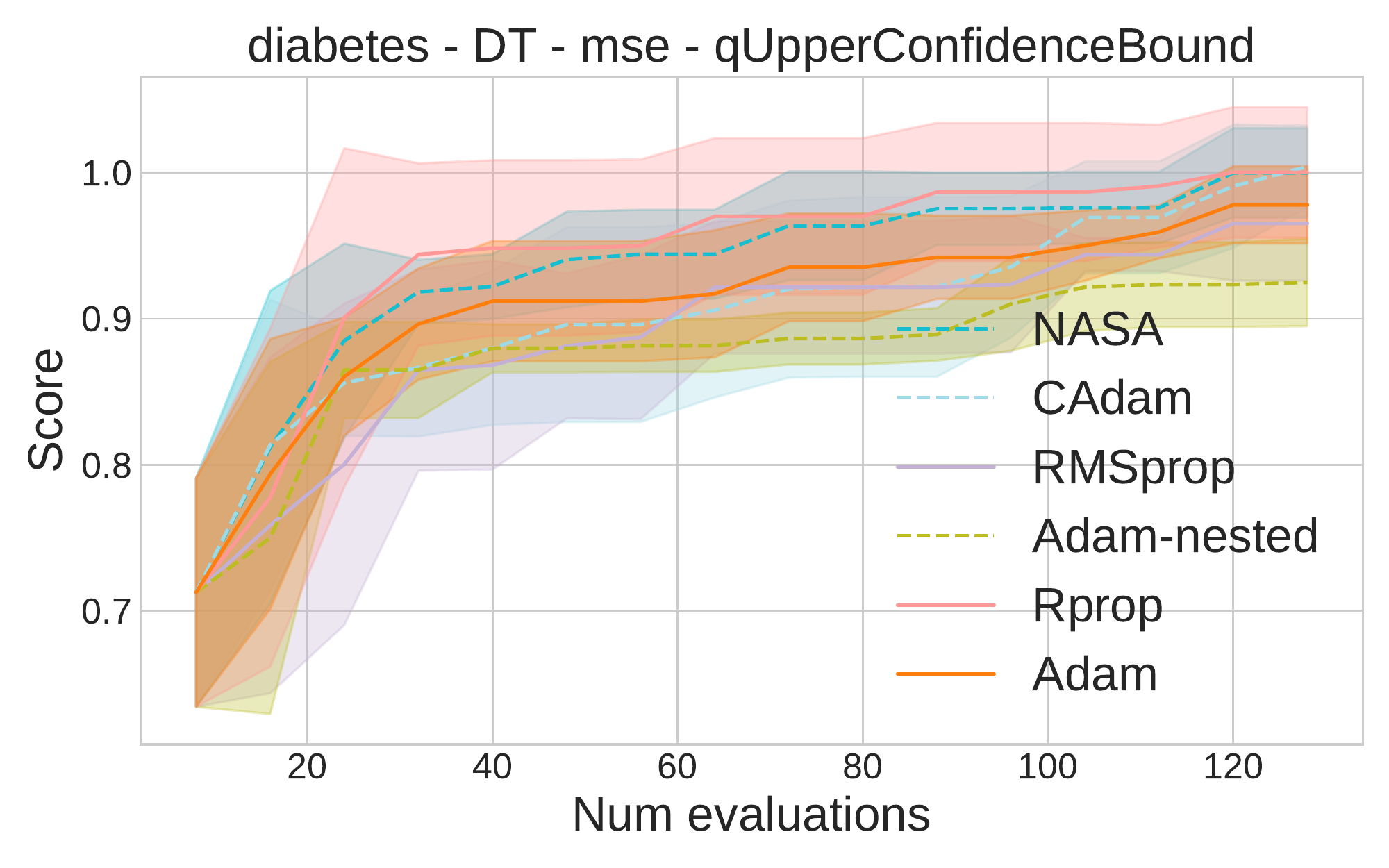}} & \hspace{-0.5cm}
    \subfloat{\includegraphics[width=0.19\columnwidth, trim={0 0.5cm 0 0.4cm}, clip]{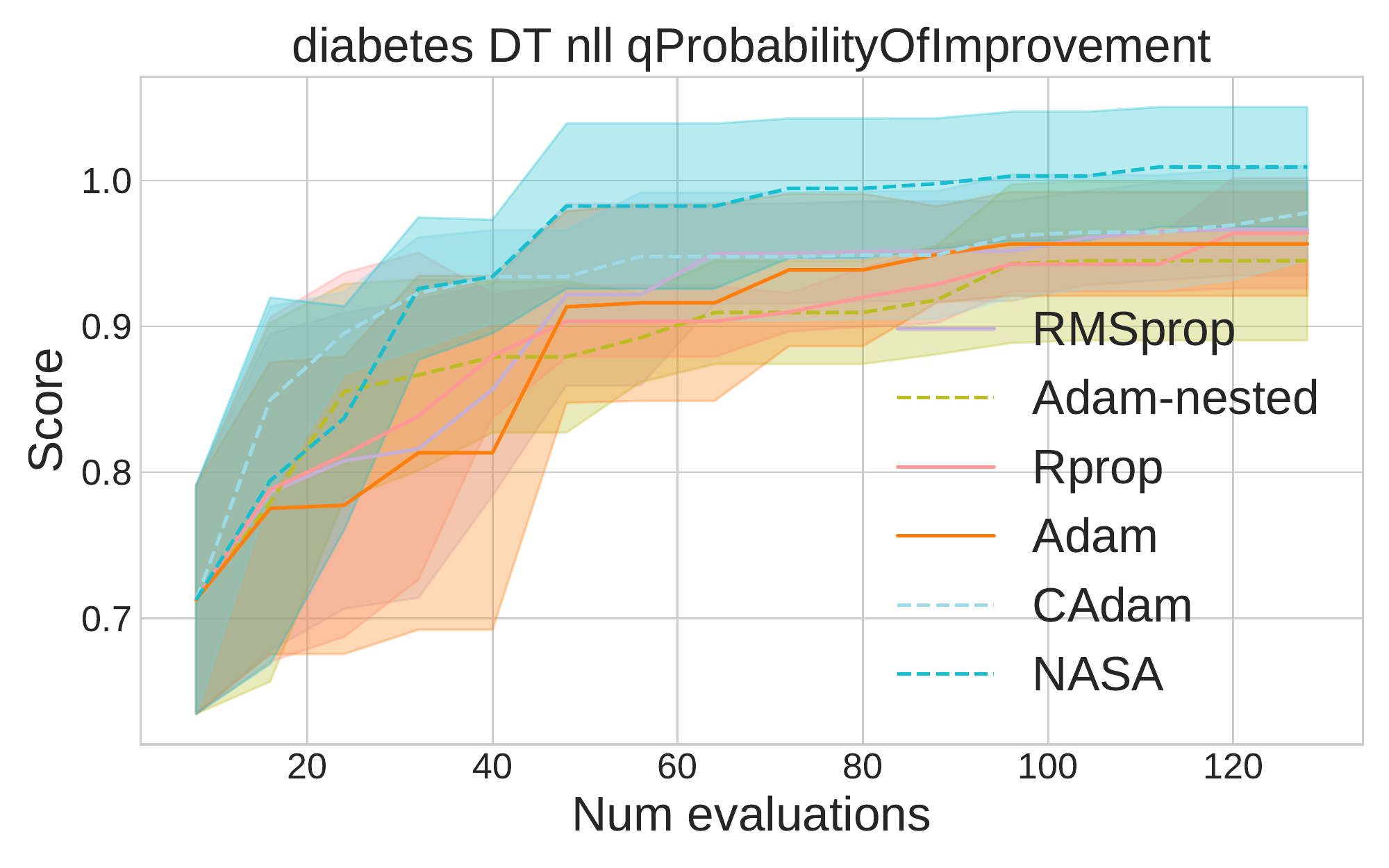}} \\  
    \subfloat{\includegraphics[width=0.19\columnwidth, trim={0 0.5cm 0 0.4cm}, clip]{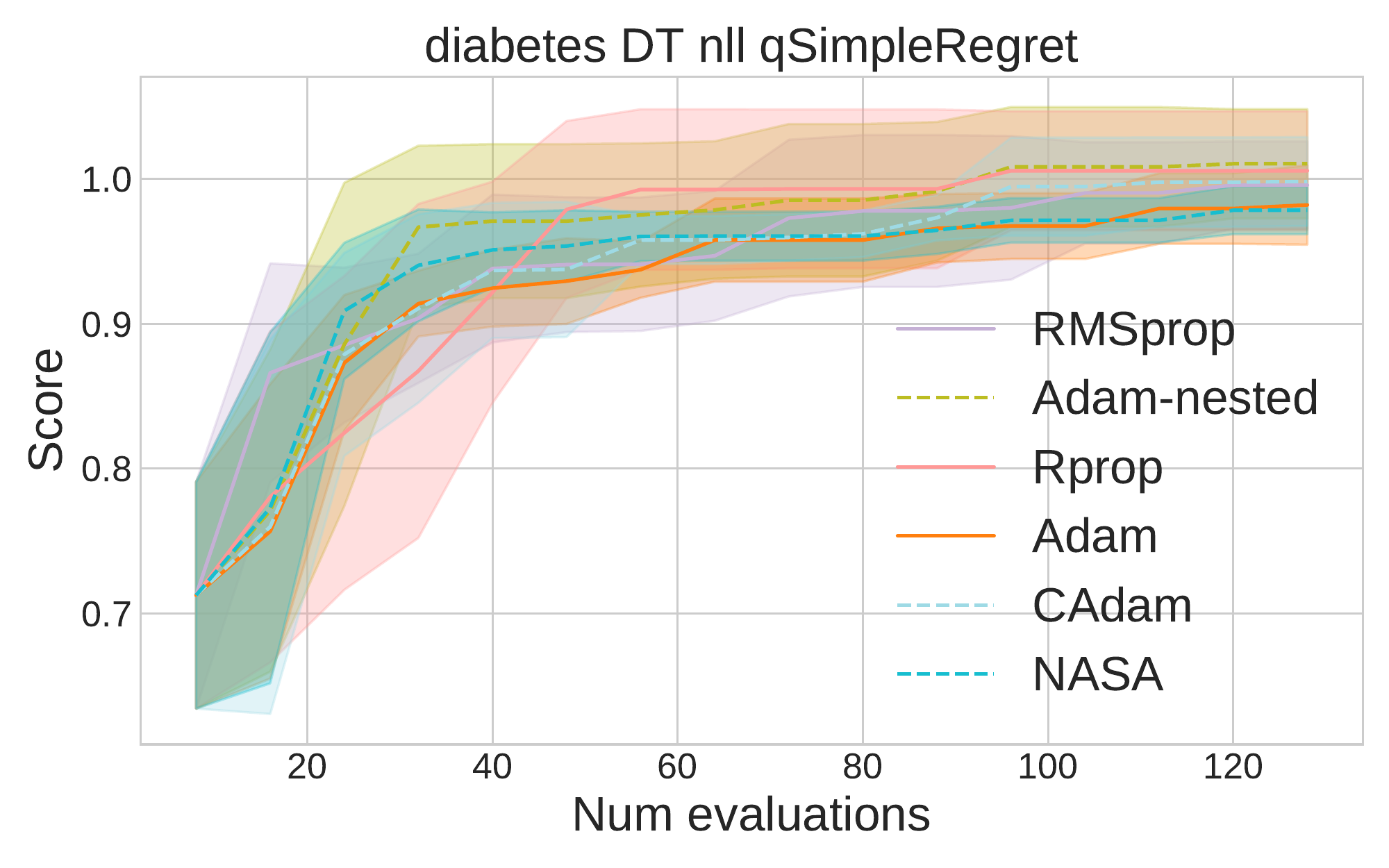}} &   \hspace{-0.5cm} 
    \subfloat{\includegraphics[width=0.19\columnwidth, trim={0 0.5cm 0 0.4cm}, clip]{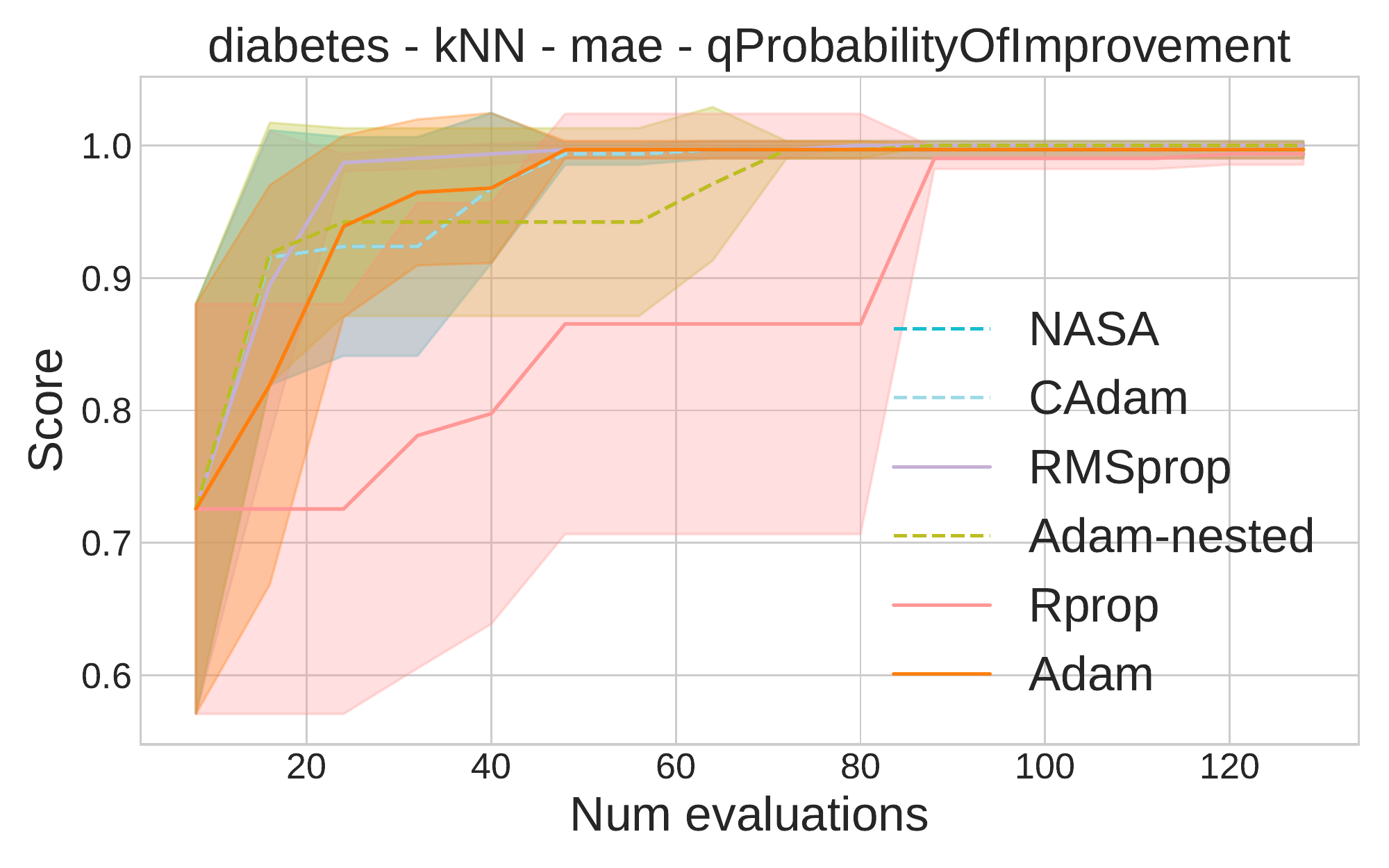}} & \hspace{-0.5cm}
    \subfloat{\includegraphics[width=0.19\columnwidth, trim={0 0.5cm 0 0.4cm}, clip]{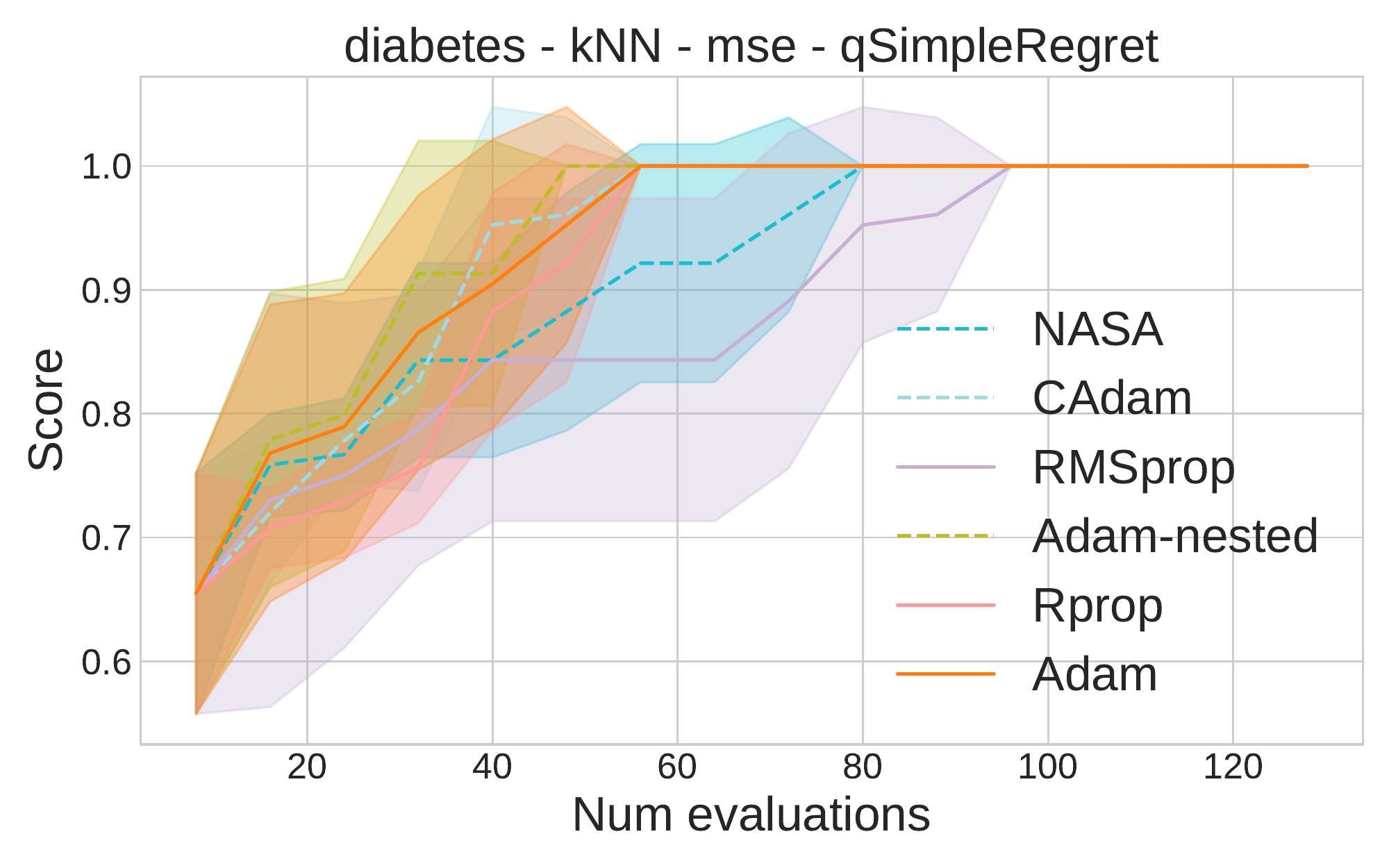}} &  \hspace{-0.5cm}
    \subfloat{\includegraphics[width=0.19\columnwidth, trim={0 0.5cm 0 0.4cm}, clip]{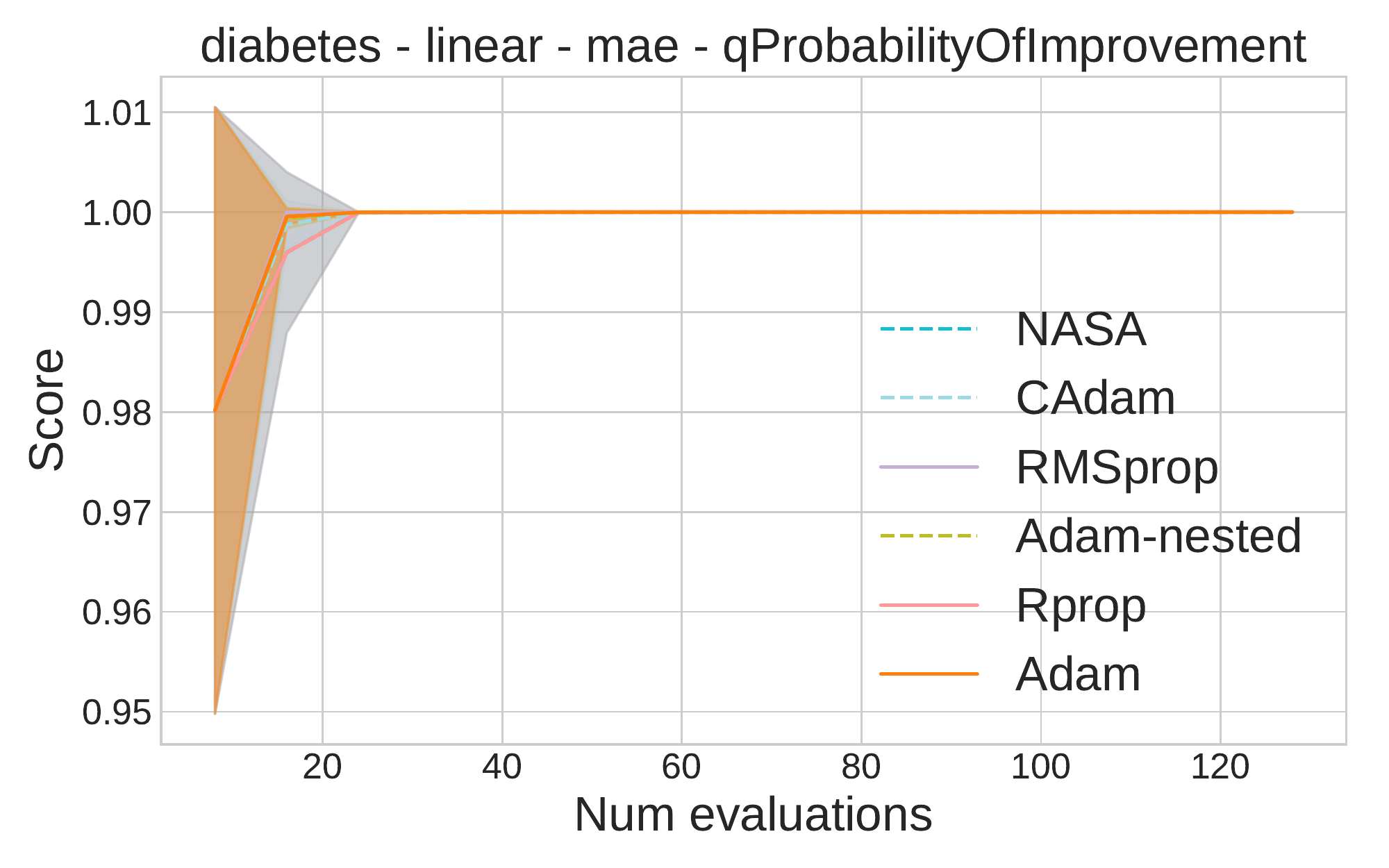}} & \hspace{-0.5cm}
    \subfloat{\includegraphics[width=0.19\columnwidth, trim={0 0.5cm 0 0.4cm}, clip]{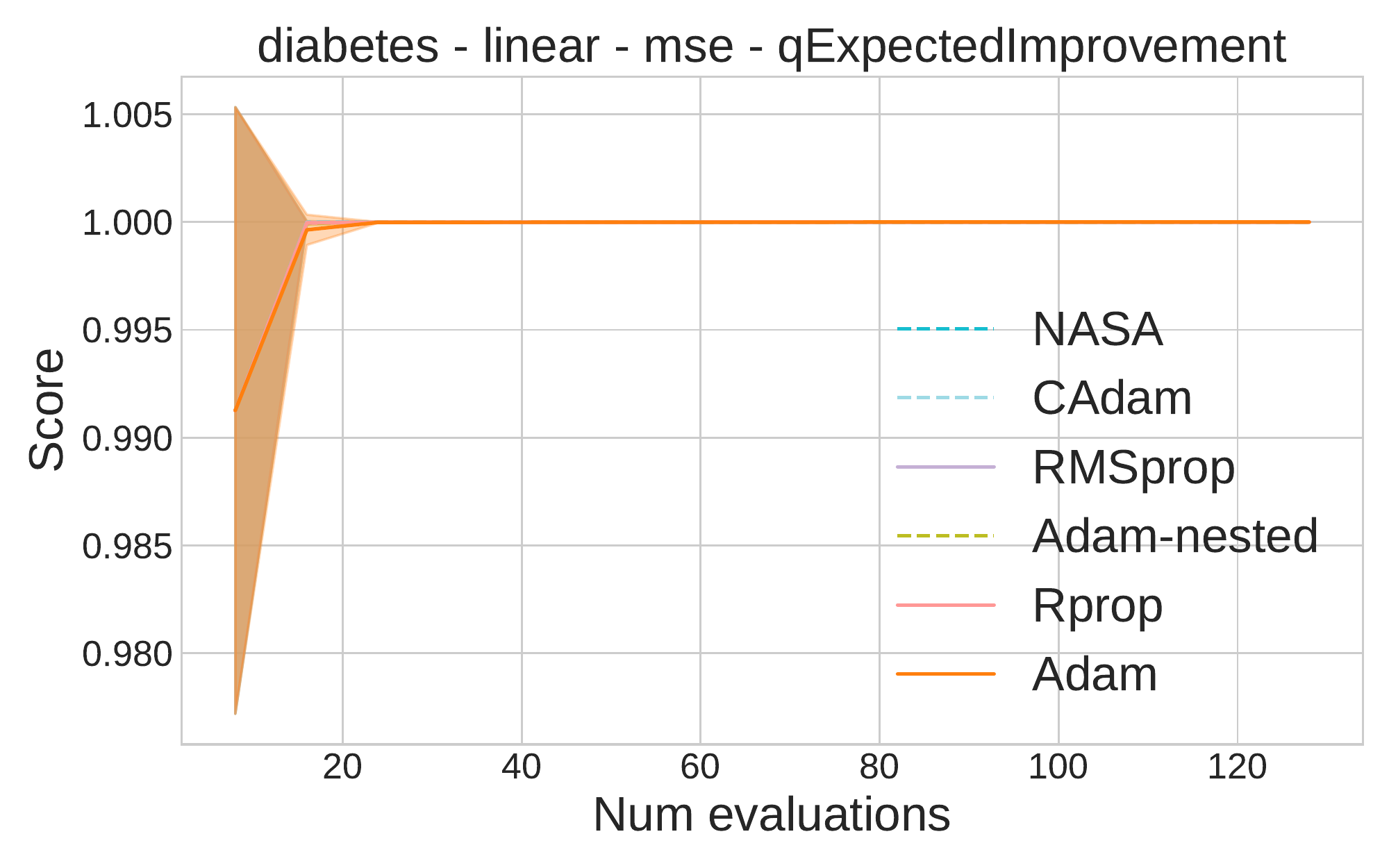}} \\  
    \subfloat{\includegraphics[width=0.19\columnwidth, trim={0 0.5cm 0 0.4cm}, clip]{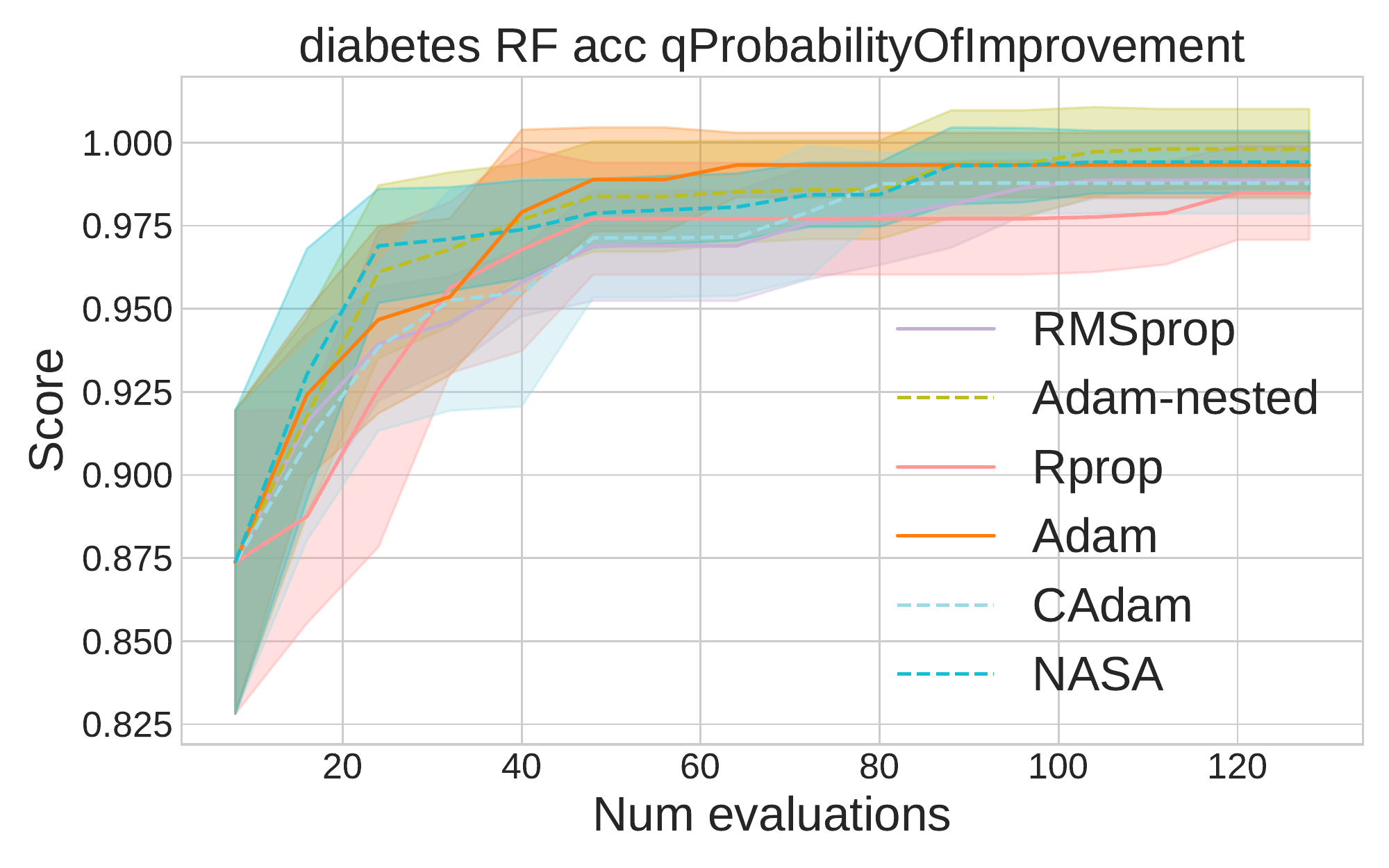}} &   \hspace{-0.5cm} 
    \subfloat{\includegraphics[width=0.19\columnwidth, trim={0 0.5cm 0 0.4cm}, clip]{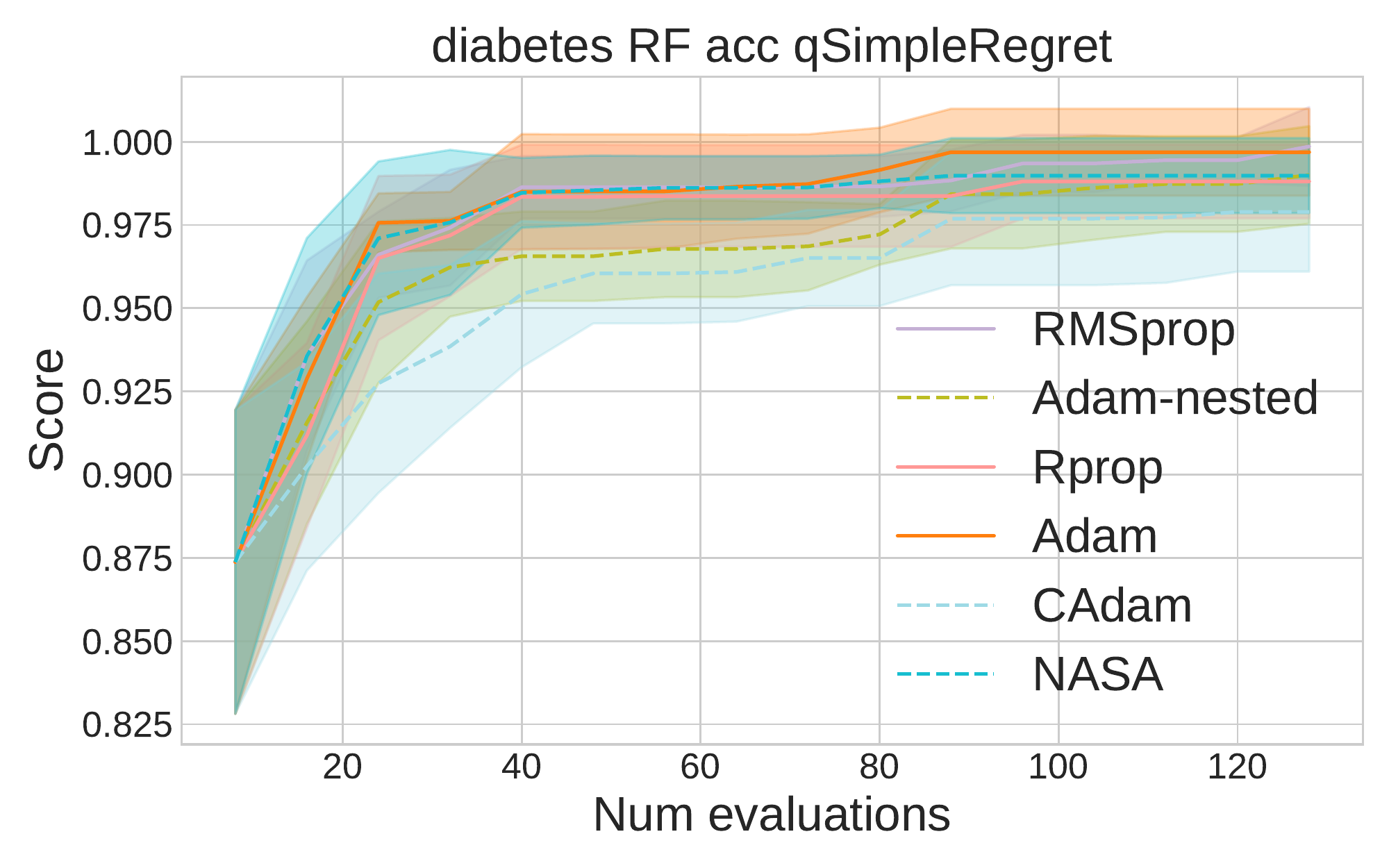}} & \hspace{-0.5cm}
    \subfloat{\includegraphics[width=0.19\columnwidth, trim={0 0.5cm 0 0.4cm}, clip]{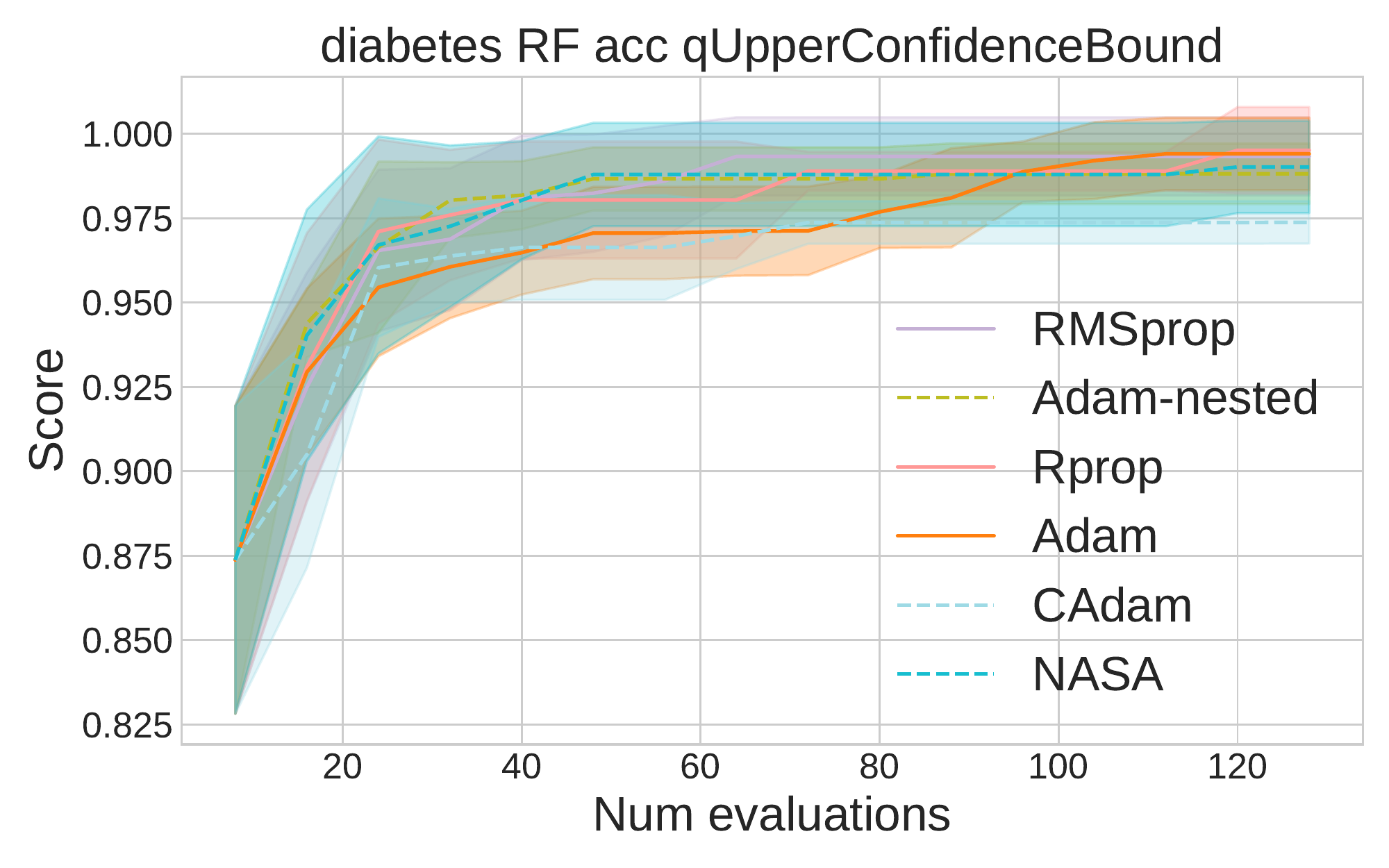}} &  \hspace{-0.5cm}
    \subfloat{\includegraphics[width=0.19\columnwidth, trim={0 0.5cm 0 0.4cm}, clip]{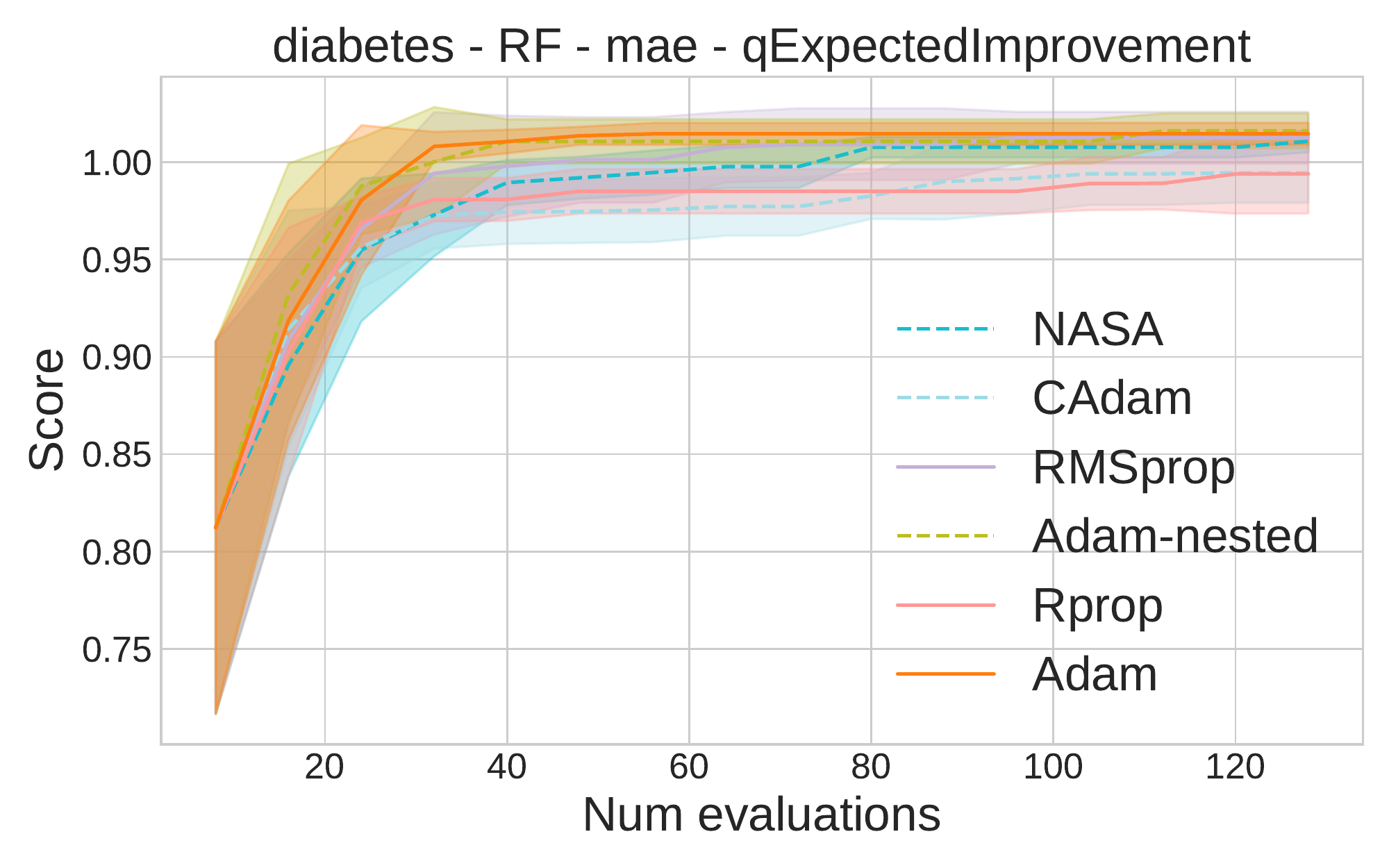}} & \hspace{-0.5cm}
    \subfloat{\includegraphics[width=0.19\columnwidth, trim={0 0.5cm 0 0.4cm}, clip]{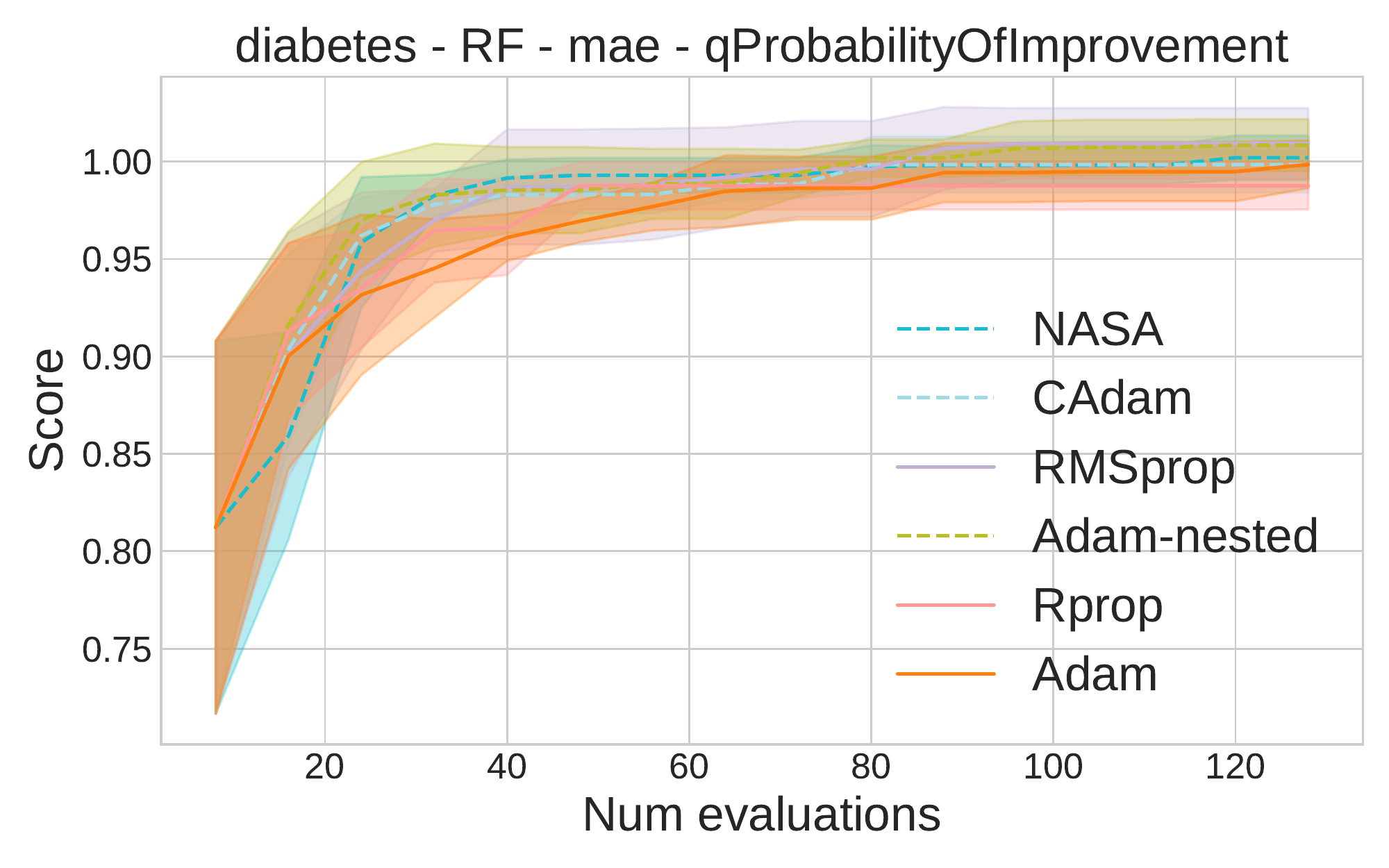}} \\  
    \subfloat{\includegraphics[width=0.19\columnwidth, trim={0 0.5cm 0 0.4cm}, clip]{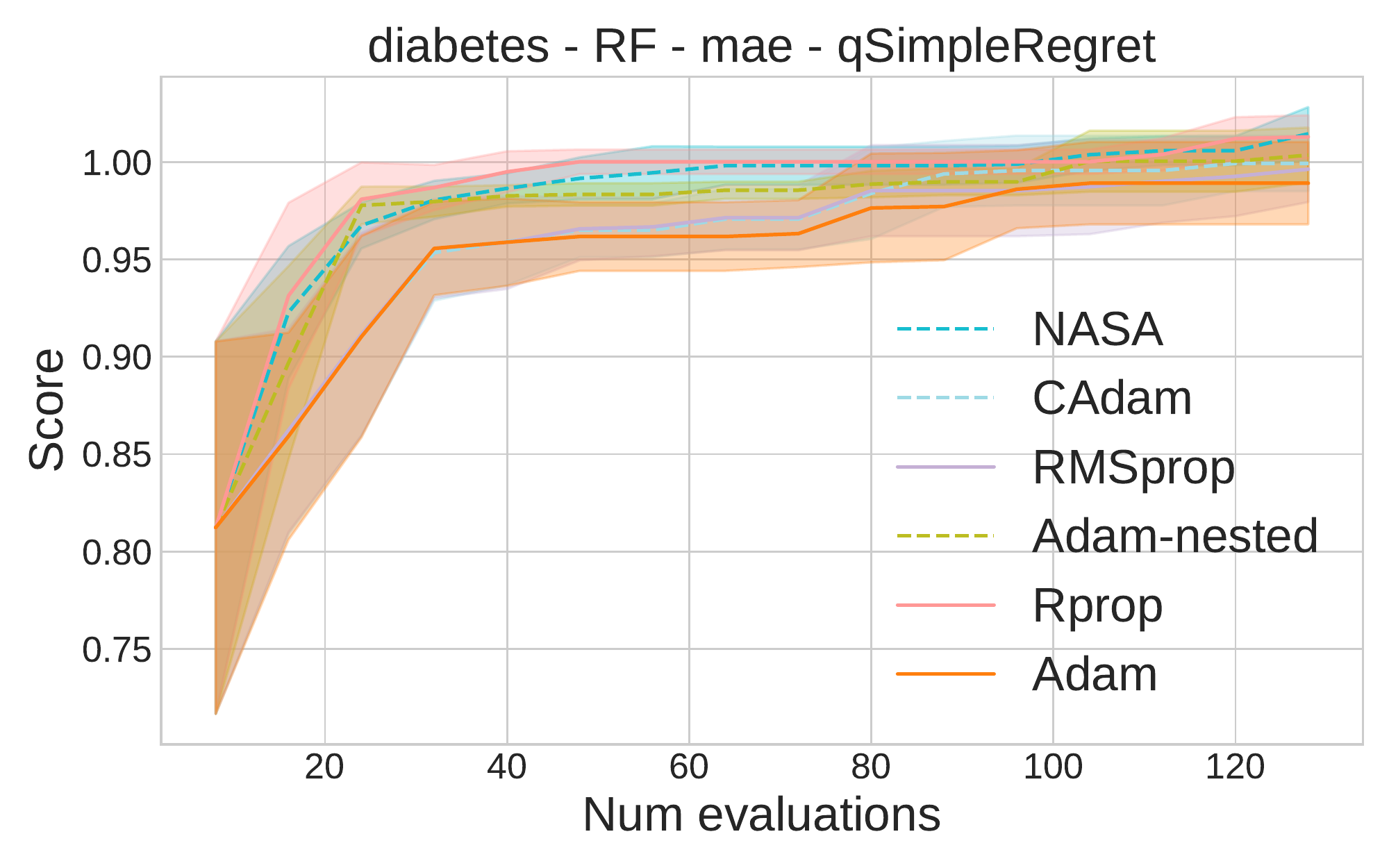}} &   \hspace{-0.5cm} 
    \subfloat{\includegraphics[width=0.19\columnwidth, trim={0 0.5cm 0 0.4cm}, clip]{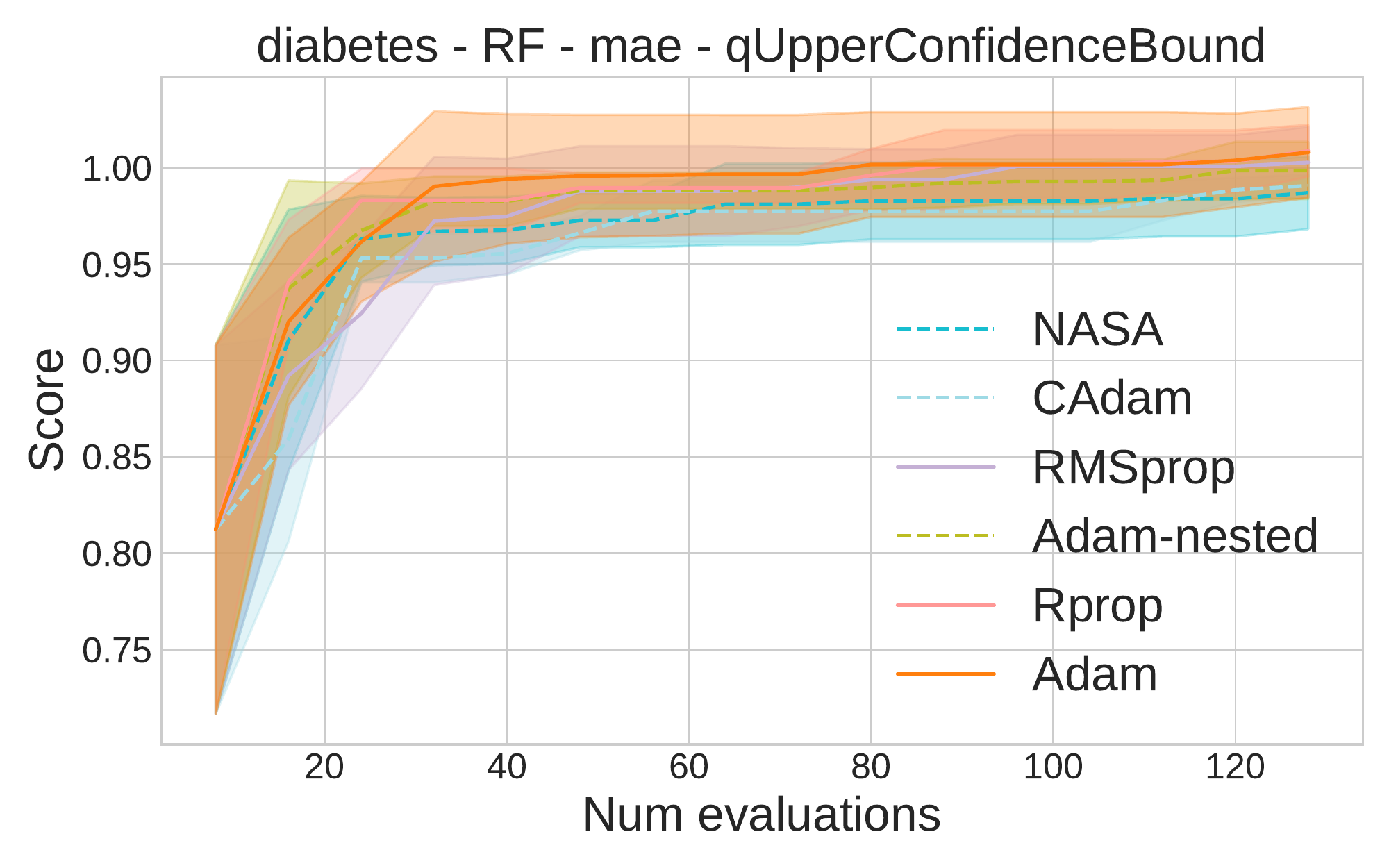}} & \hspace{-0.5cm}
    \subfloat{\includegraphics[width=0.19\columnwidth, trim={0 0.5cm 0 0.4cm}, clip]{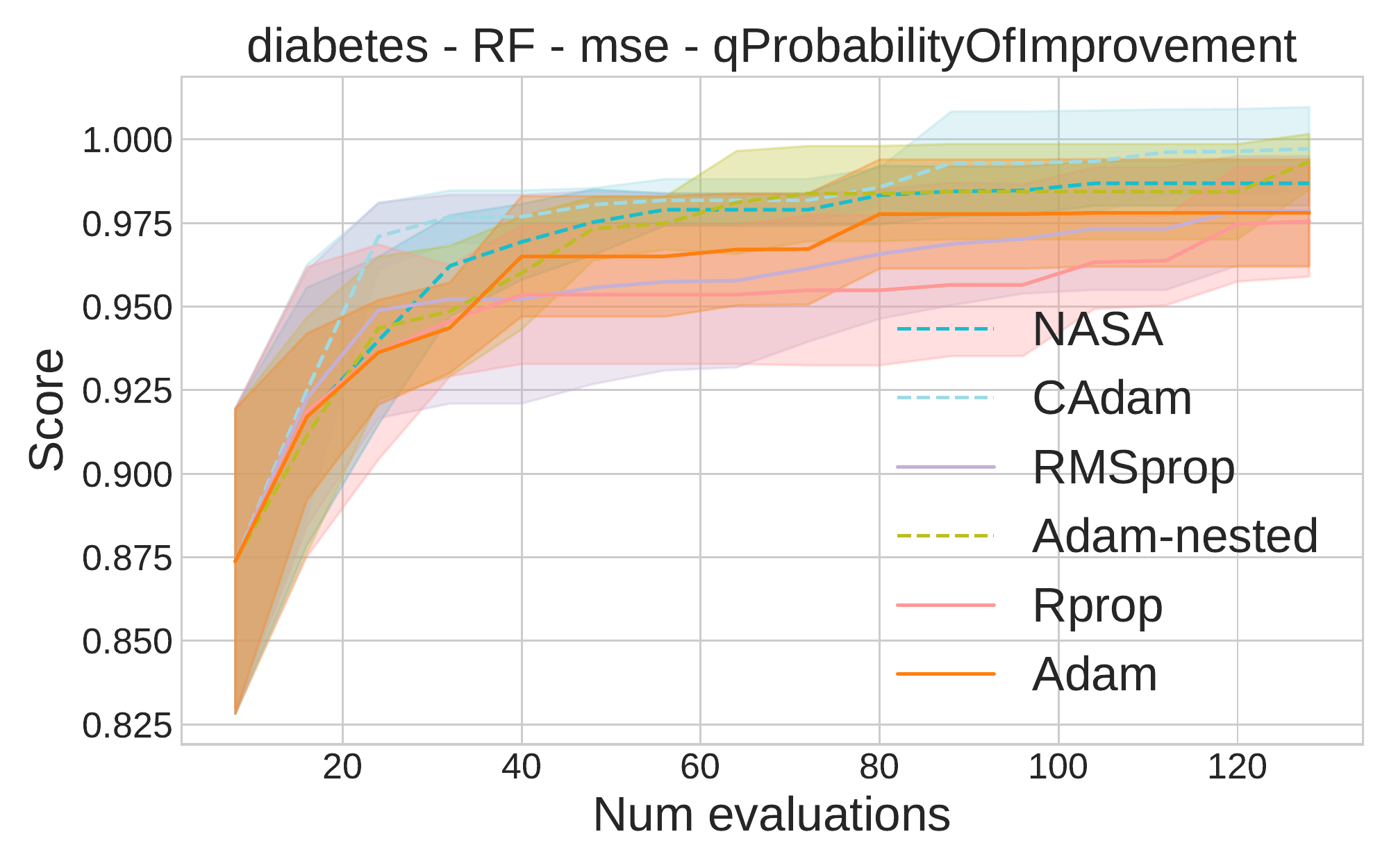}} &  \hspace{-0.5cm}
    \subfloat{\includegraphics[width=0.19\columnwidth, trim={0 0.5cm 0 0.4cm}, clip]{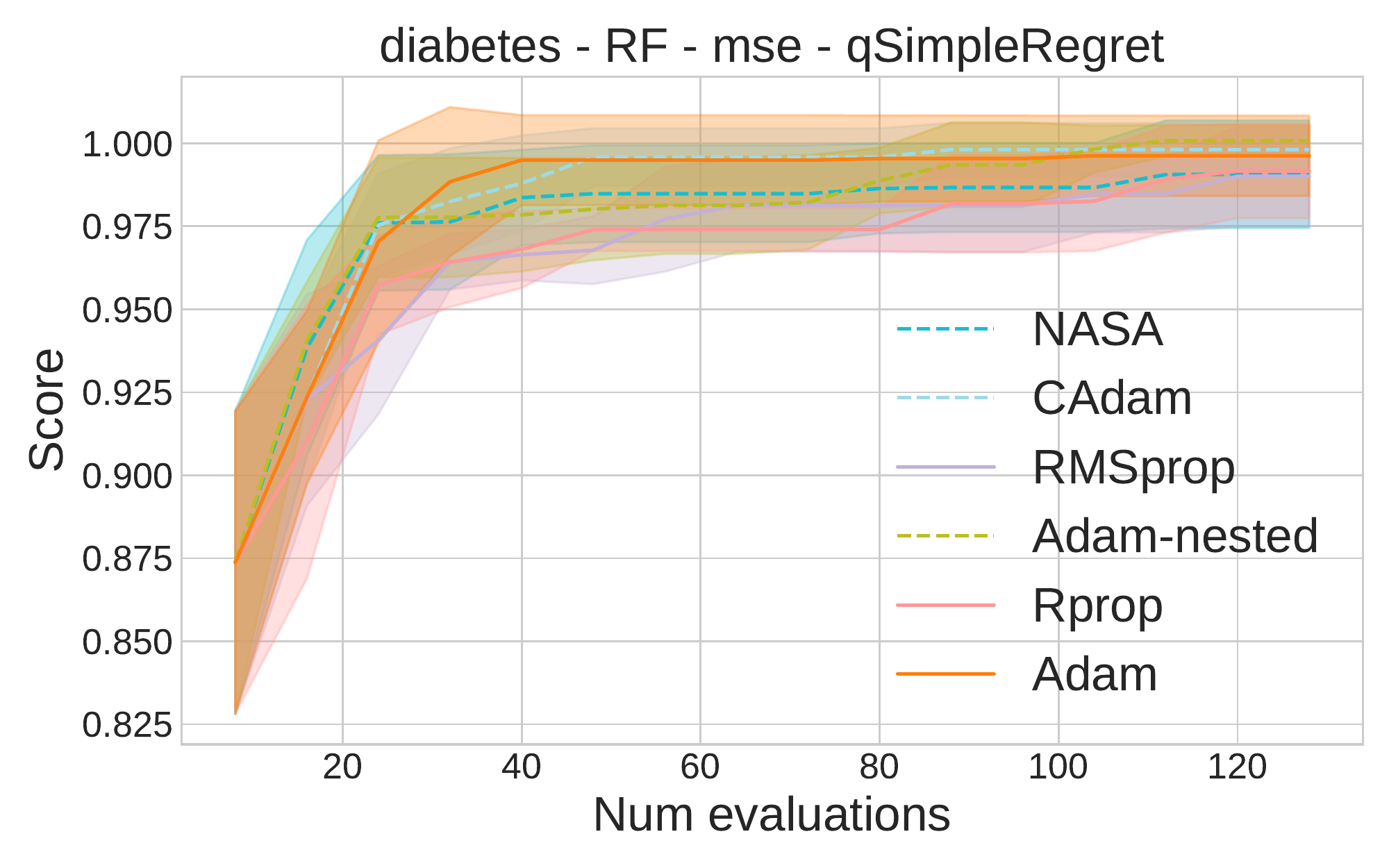}} & \hspace{-0.5cm}
    \subfloat{\includegraphics[width=0.19\columnwidth, trim={0 0.5cm 0 0.4cm}, clip]{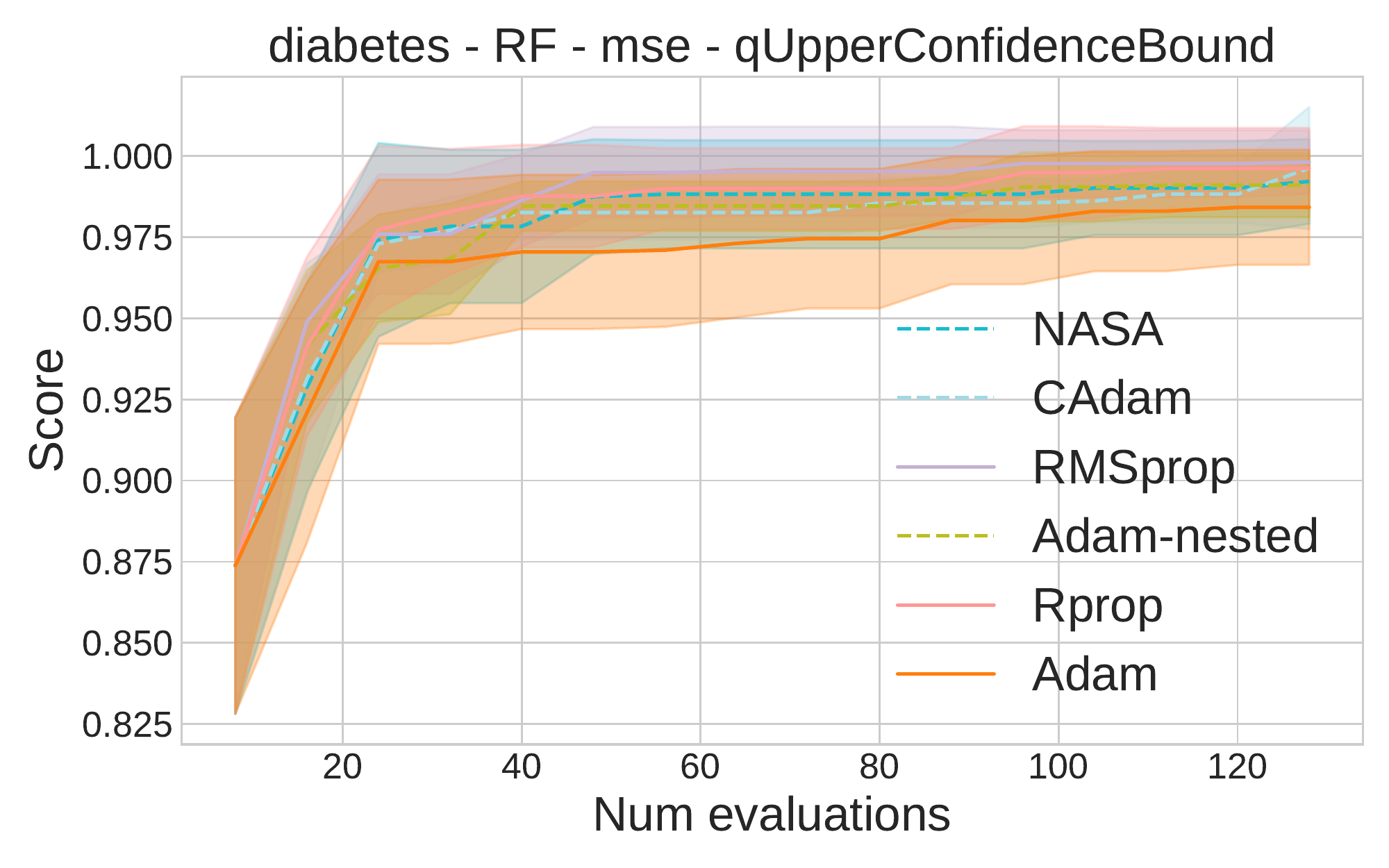}} \\  
    \subfloat{\includegraphics[width=0.19\columnwidth, trim={0 0.5cm 0 0.4cm}, clip]{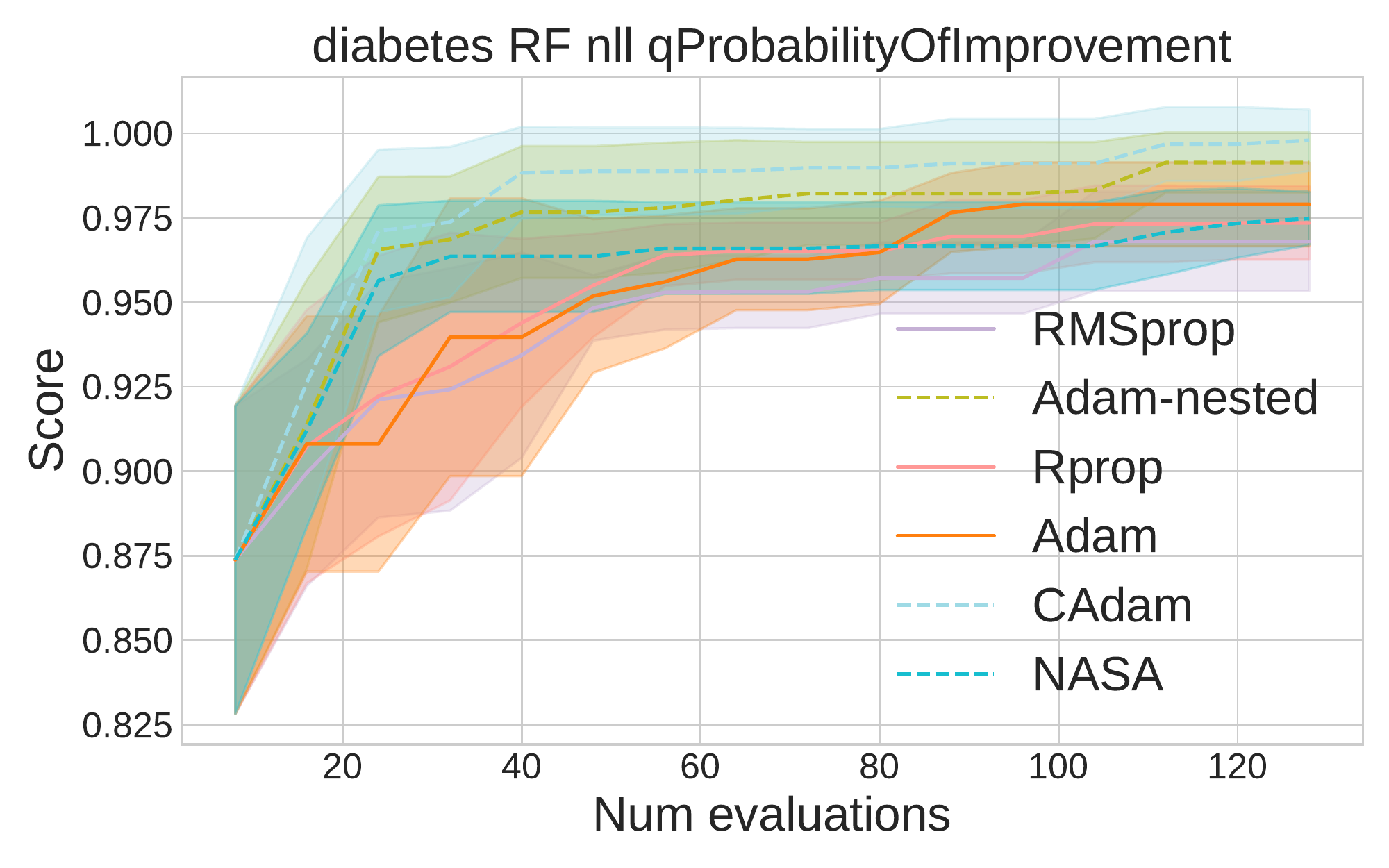}} &   \hspace{-0.5cm} 
    \subfloat{\includegraphics[width=0.19\columnwidth, trim={0 0.5cm 0 0.4cm}, clip]{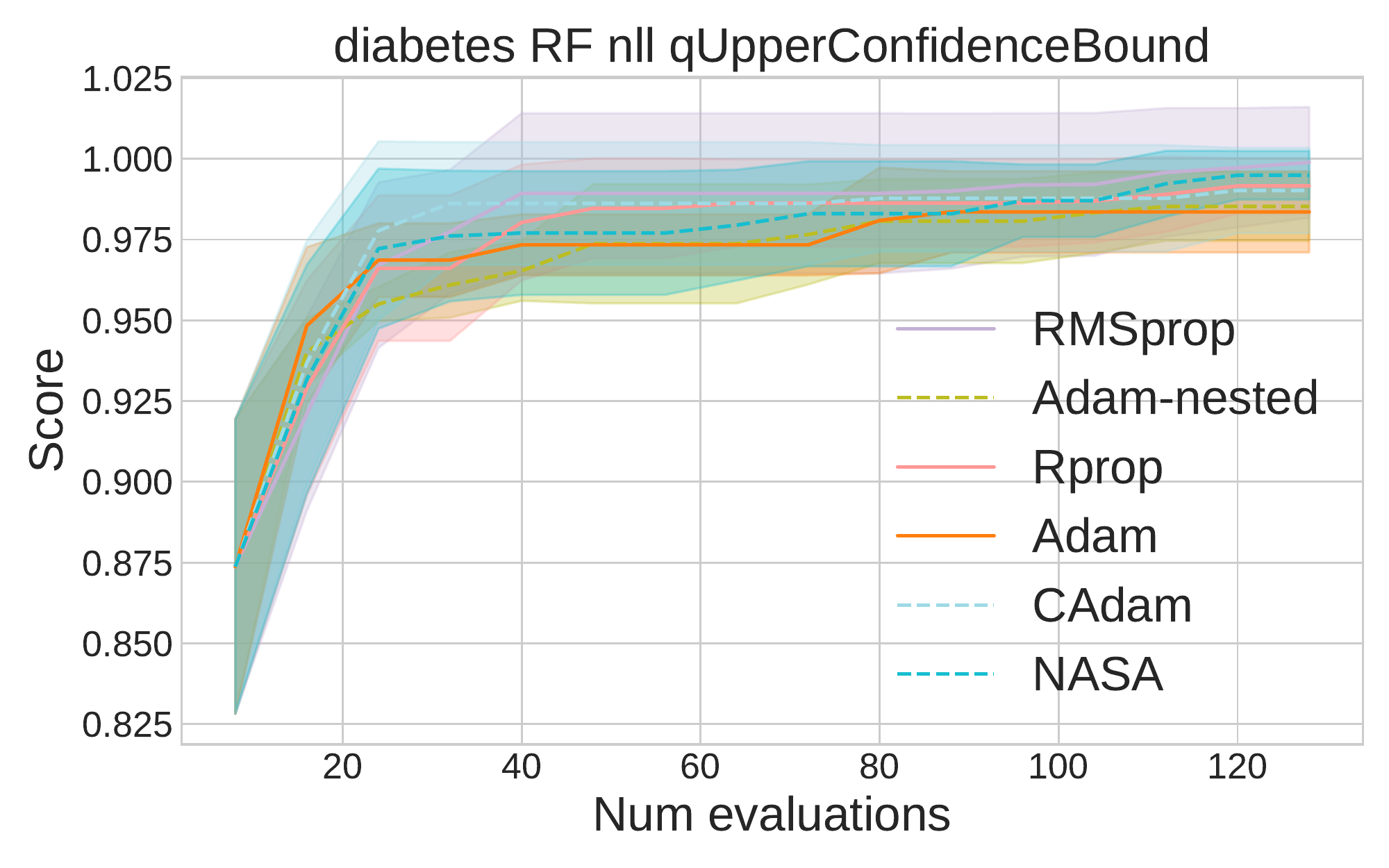}} & \hspace{-0.5cm}
    \subfloat{\includegraphics[width=0.19\columnwidth, trim={0 0.5cm 0 0.4cm}, clip]{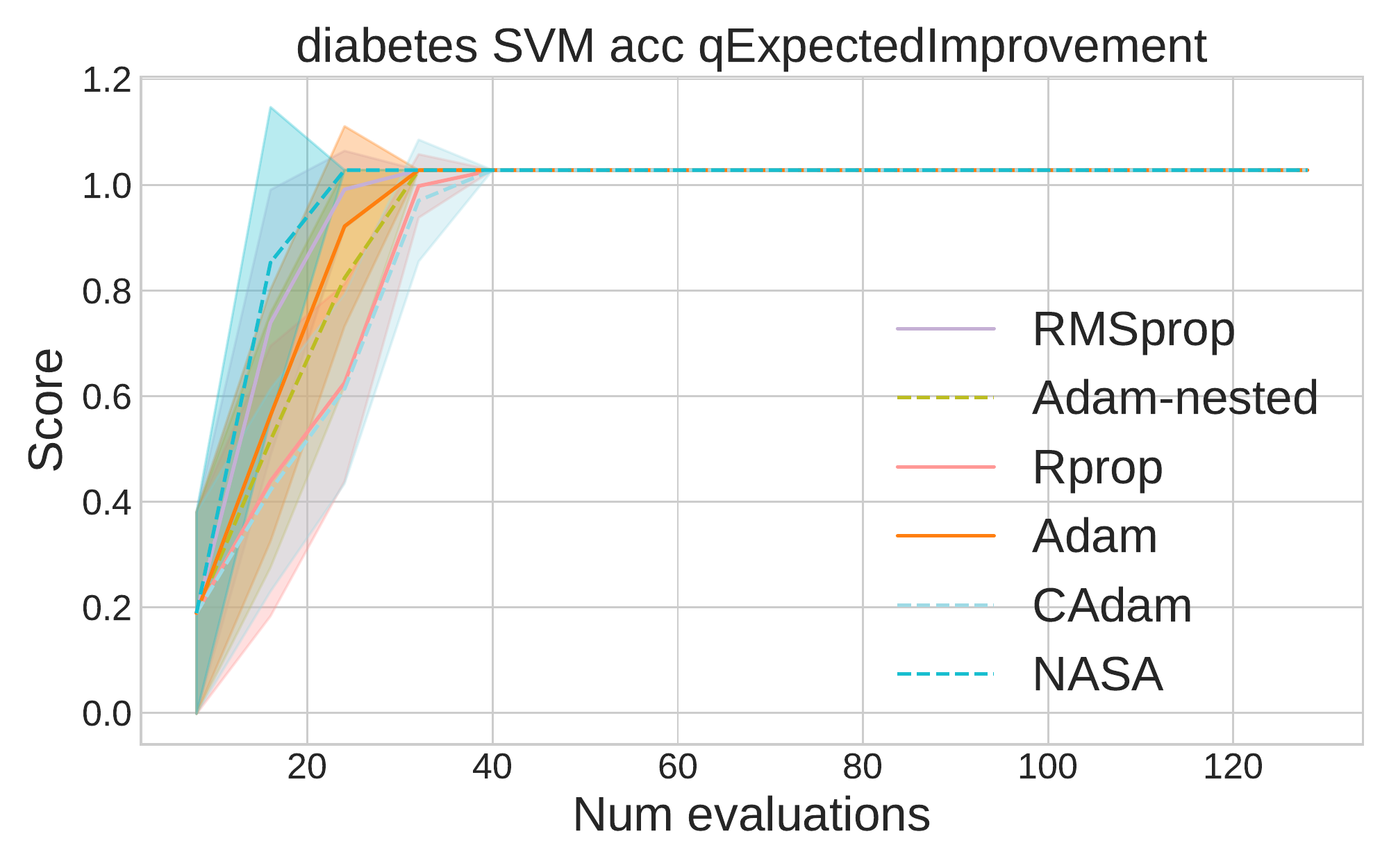}} &  \hspace{-0.5cm}
    \subfloat{\includegraphics[width=0.19\columnwidth, trim={0 0.5cm 0 0.4cm}, clip]{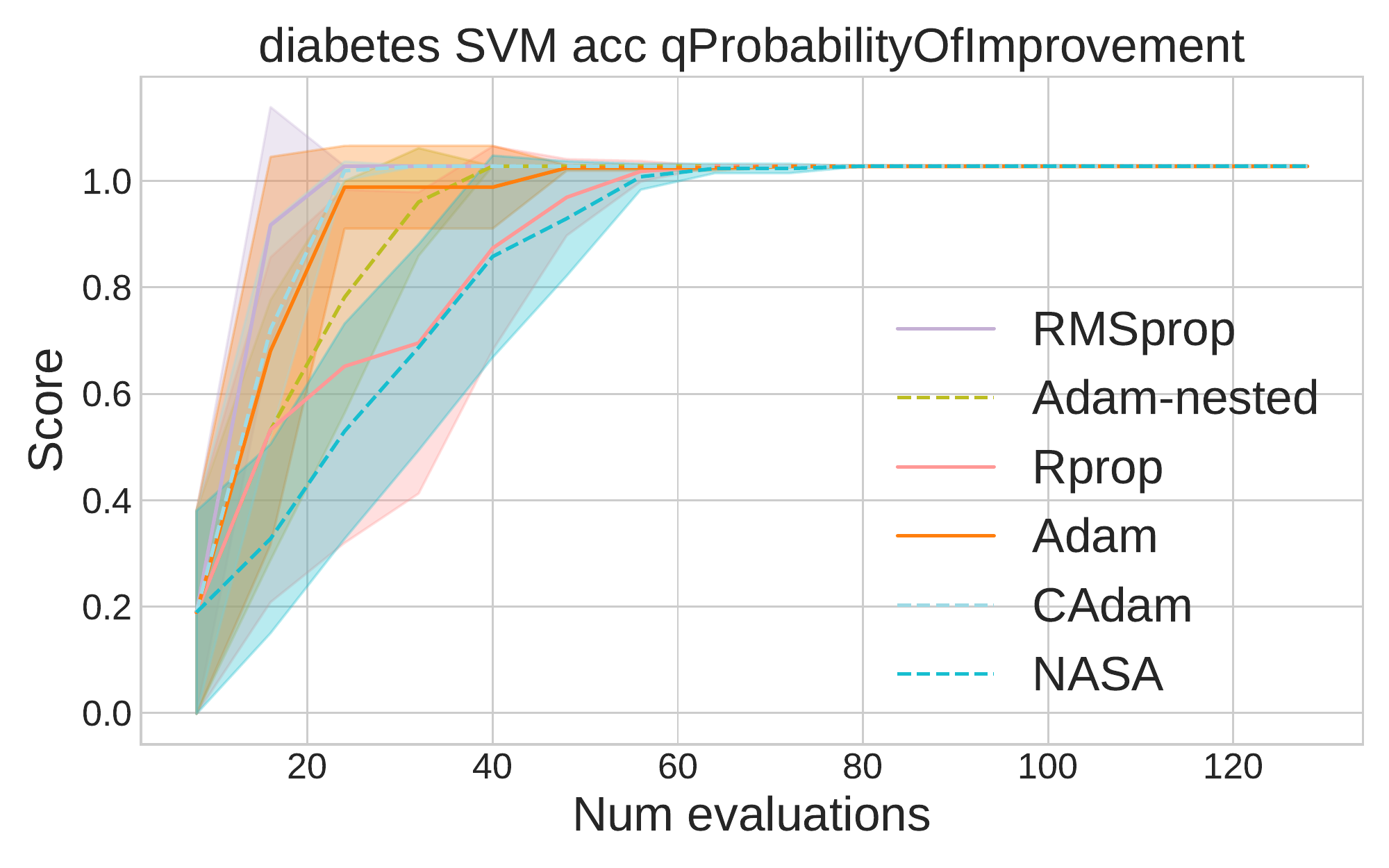}} & \hspace{-0.5cm}
    \subfloat{\includegraphics[width=0.19\columnwidth, trim={0 0.5cm 0 0.4cm}, clip]{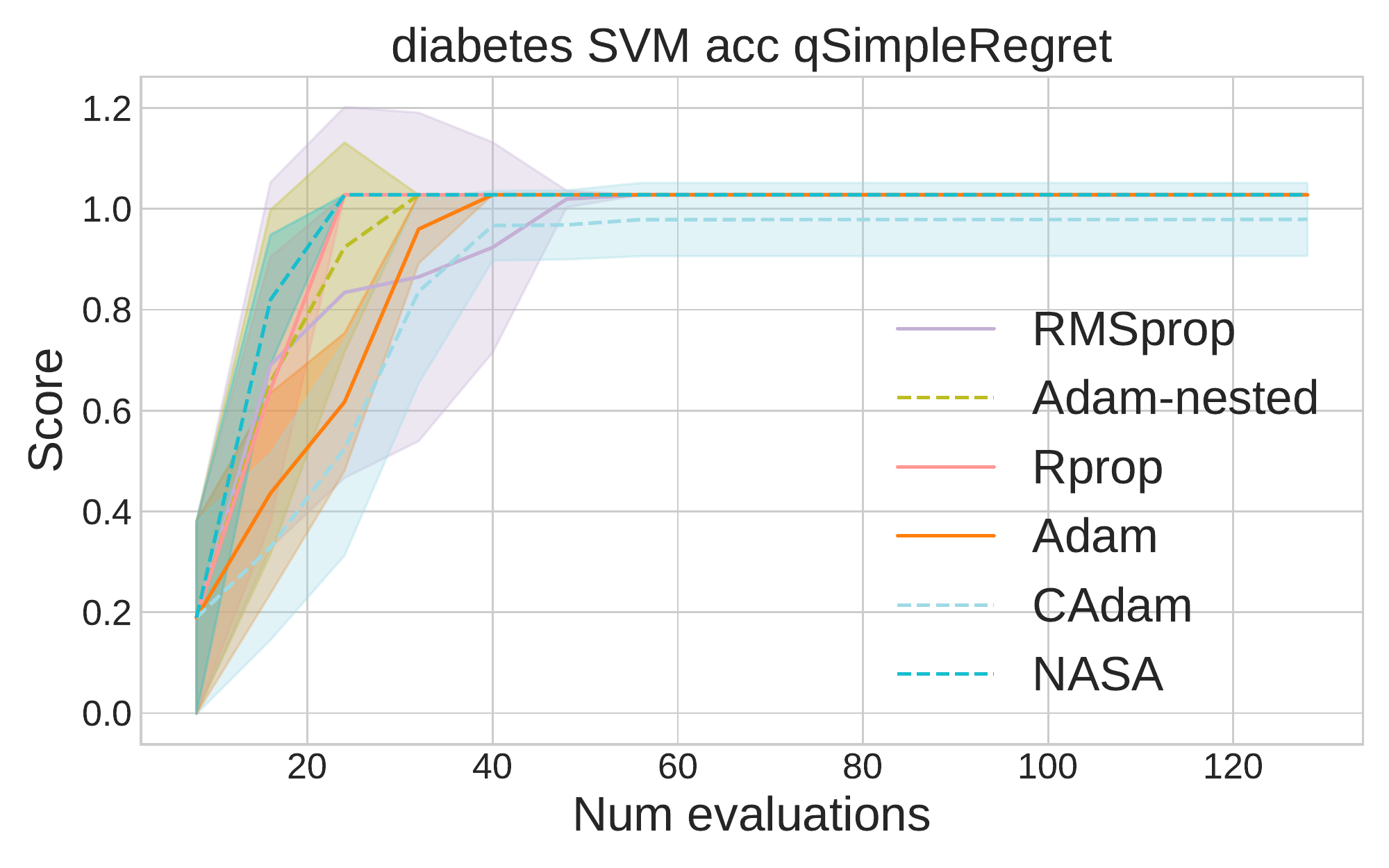}} \\  
    \subfloat{\includegraphics[width=0.19\columnwidth, trim={0 0.5cm 0 0.4cm}, clip]{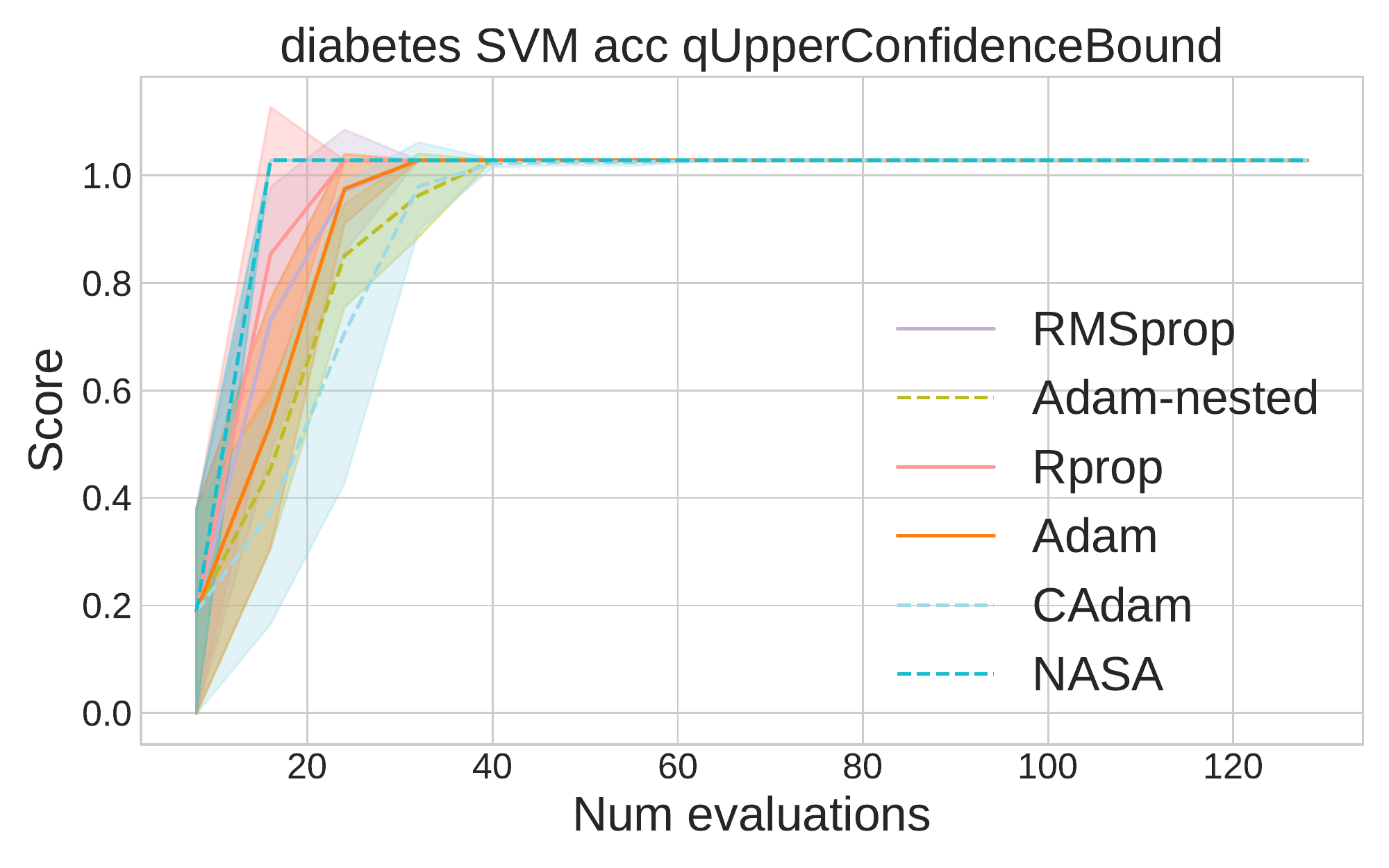}} &   \hspace{-0.5cm} 
    \subfloat{\includegraphics[width=0.19\columnwidth, trim={0 0.5cm 0 0.4cm}, clip]{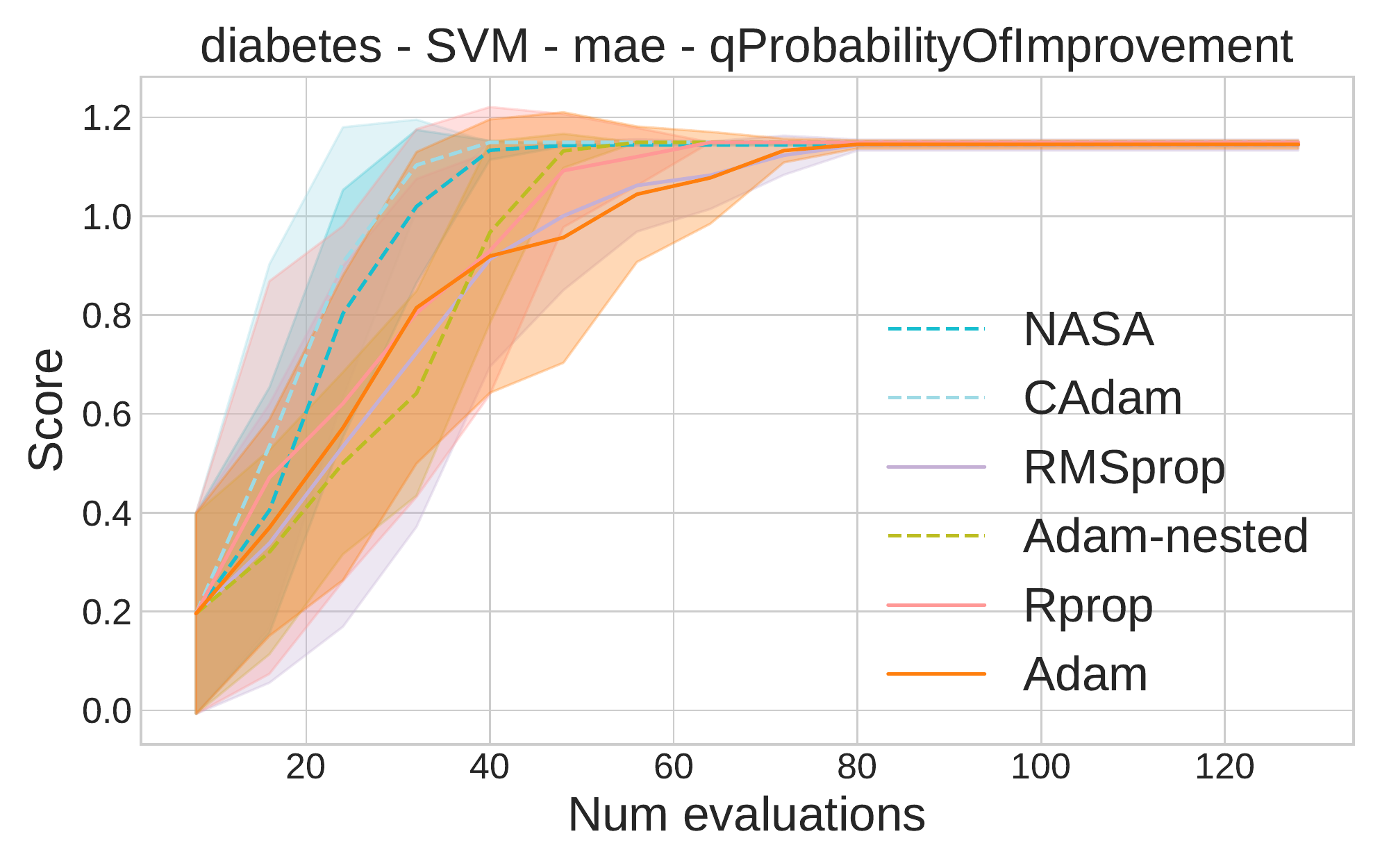}} & \hspace{-0.5cm}
    \subfloat{\includegraphics[width=0.19\columnwidth, trim={0 0.5cm 0 0.4cm}, clip]{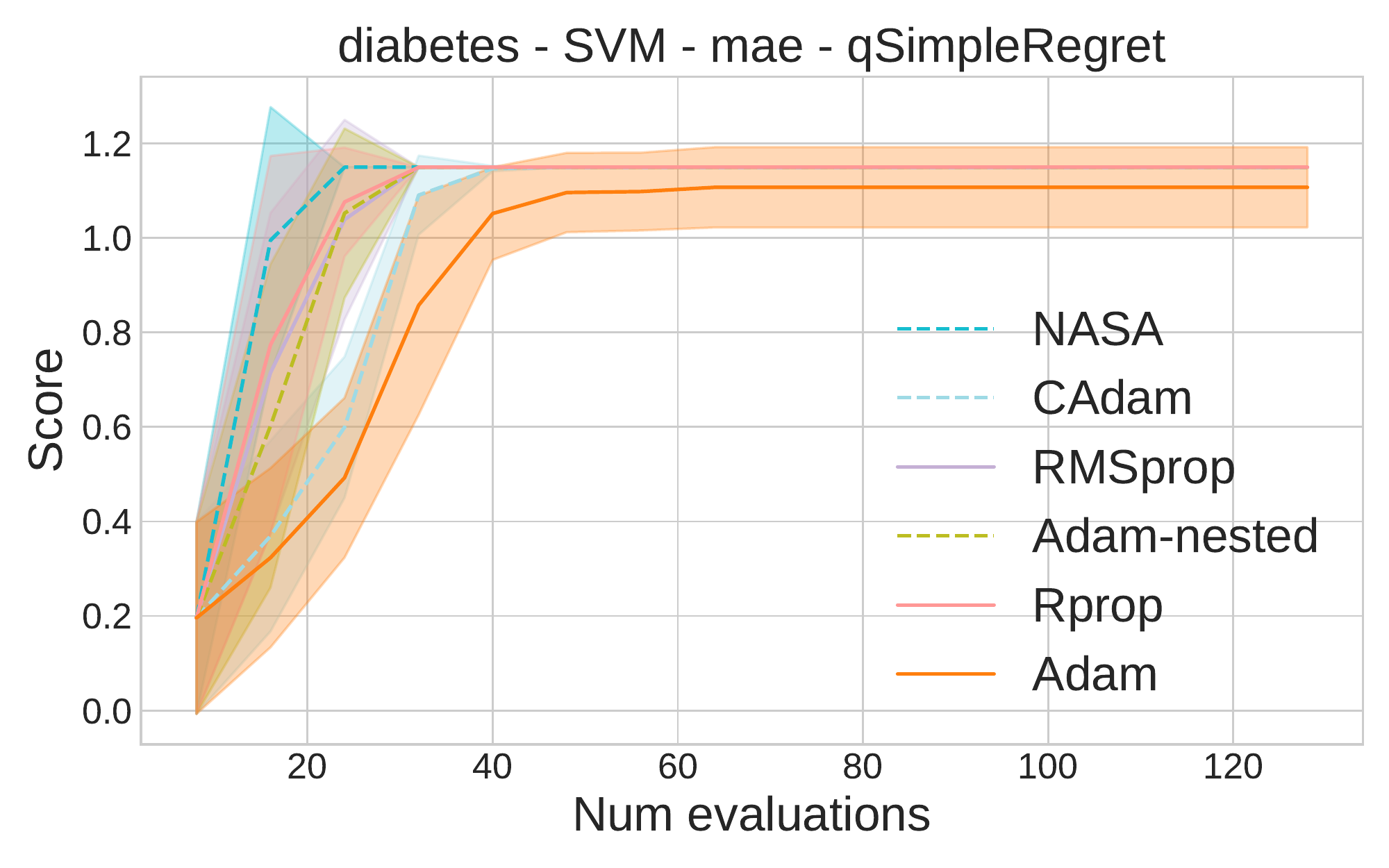}} &  \hspace{-0.5cm}
    \subfloat{\includegraphics[width=0.19\columnwidth, trim={0 0.5cm 0 0.4cm}, clip]{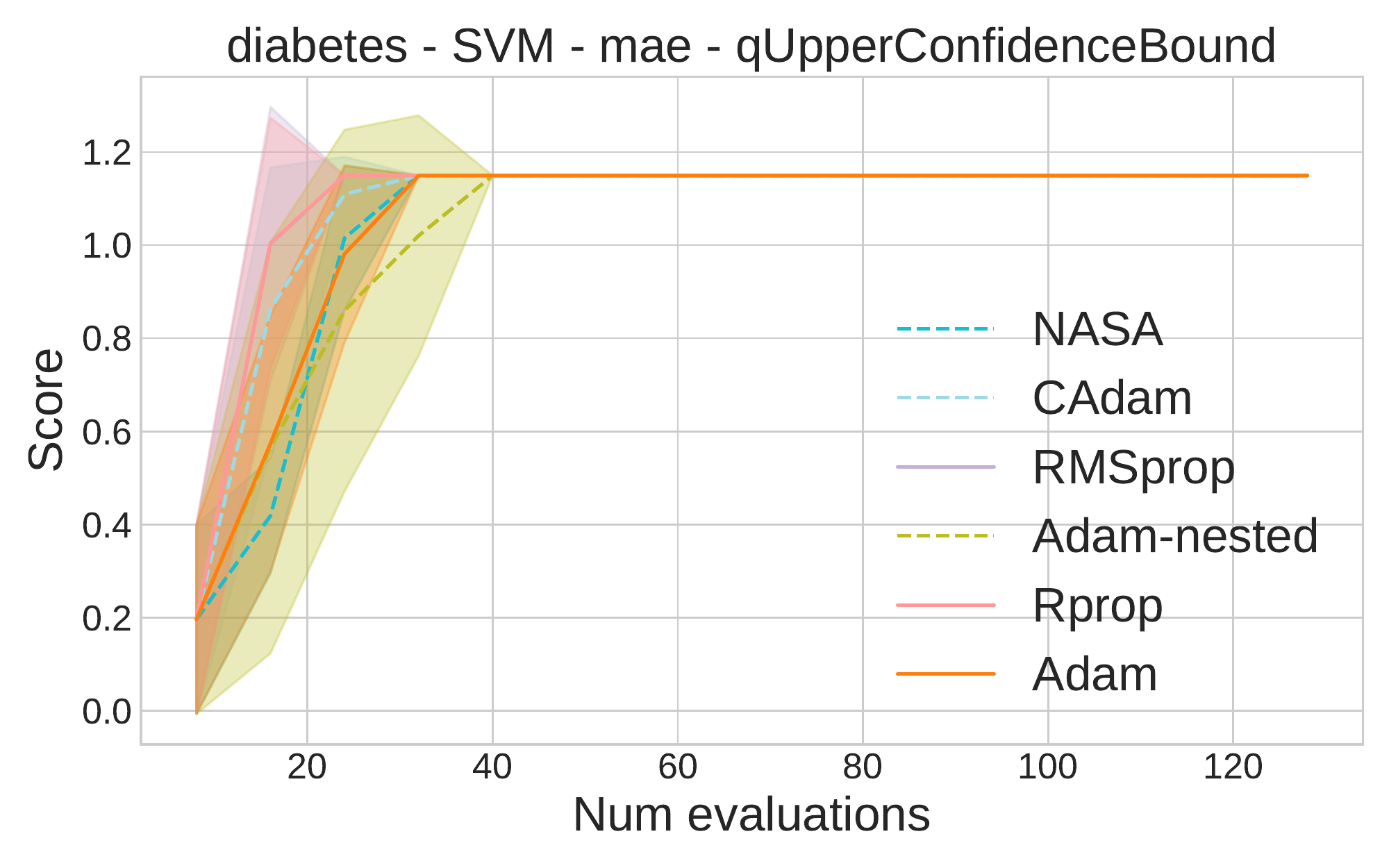}} & \hspace{-0.5cm}
    \subfloat{\includegraphics[width=0.19\columnwidth, trim={0 0.5cm 0 0.4cm}, clip]{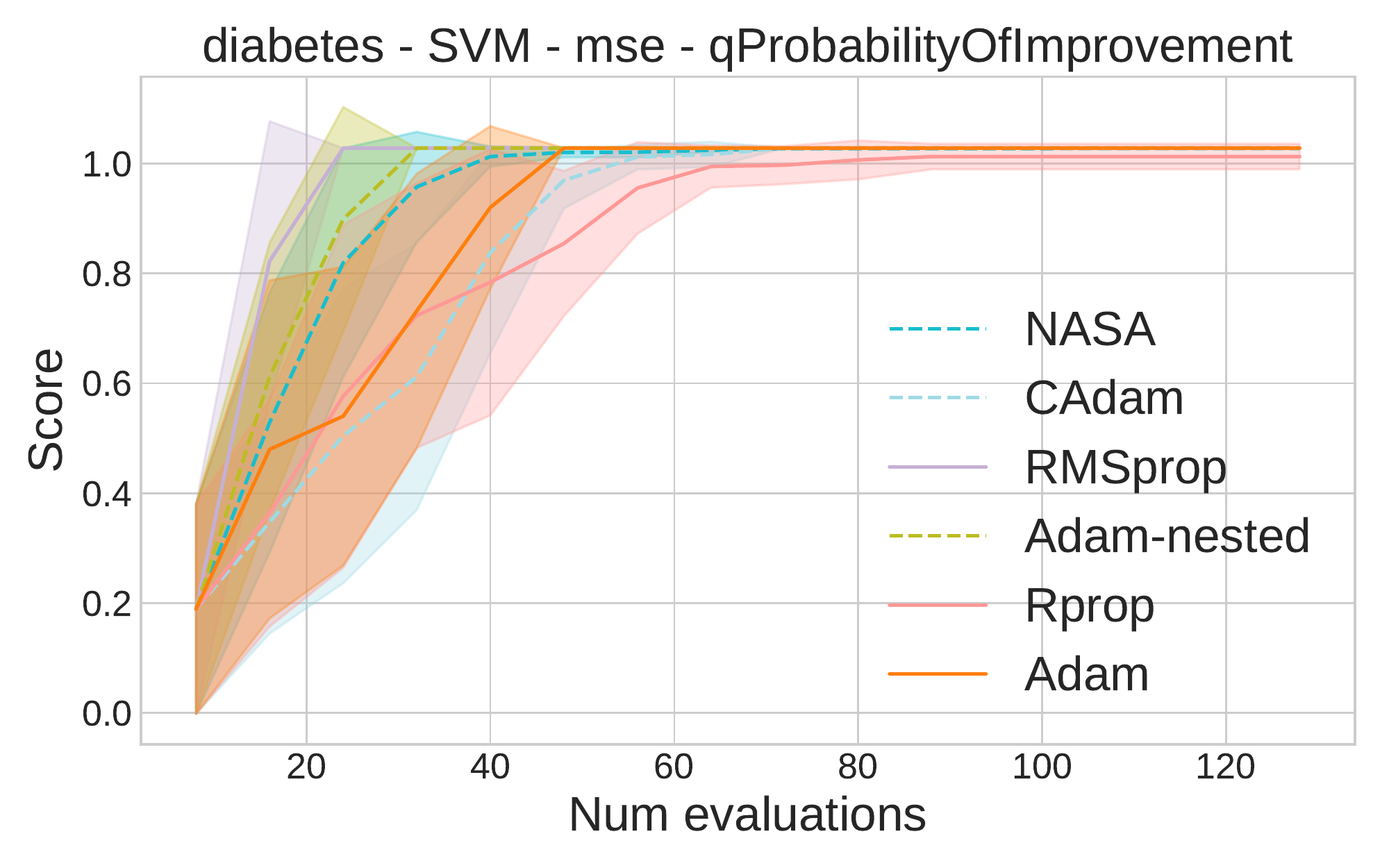}} \\  
    \subfloat{\includegraphics[width=0.19\columnwidth, trim={0 0.5cm 0 0.4cm}, clip]{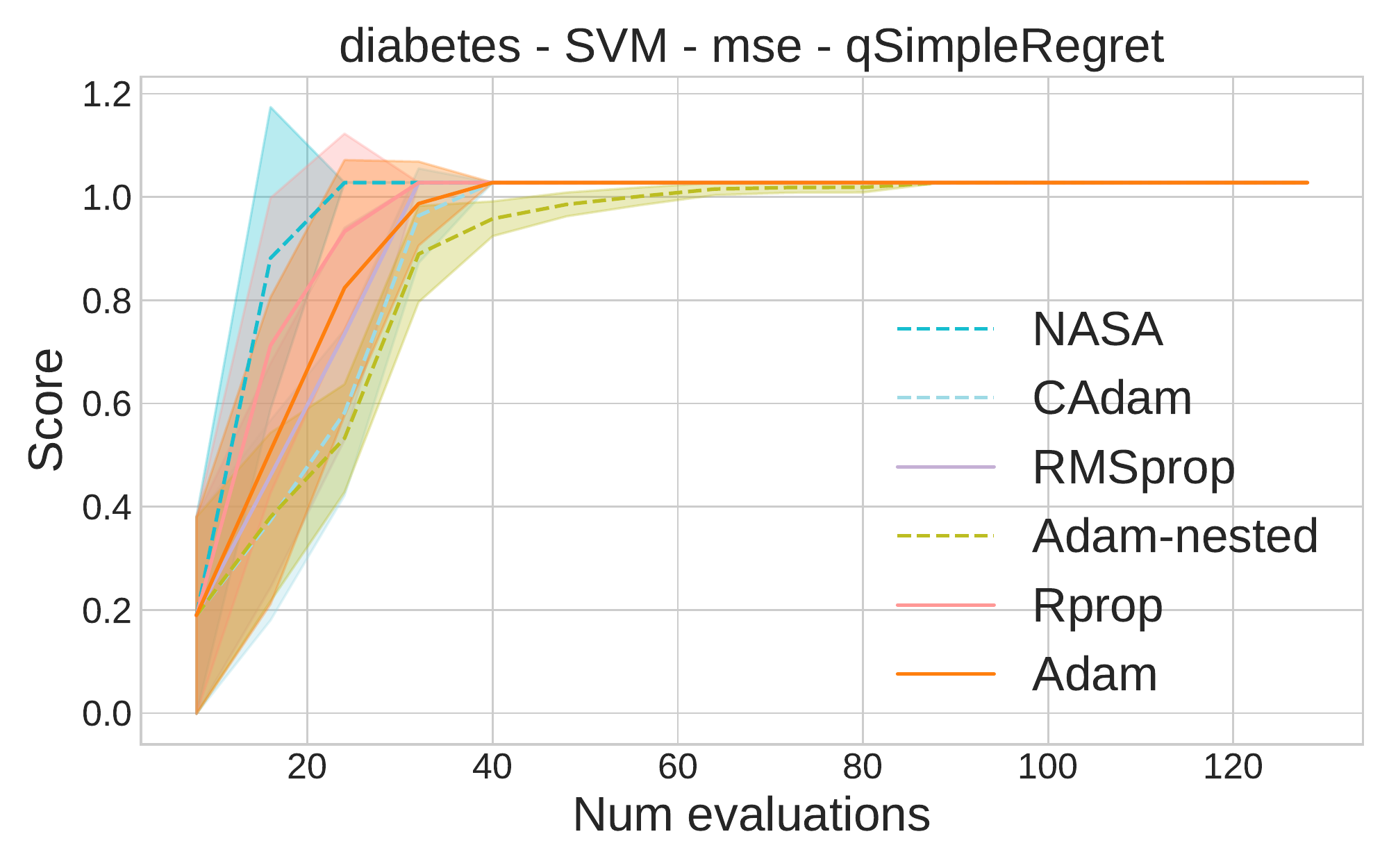}} &   \hspace{-0.5cm} 
    \subfloat{\includegraphics[width=0.19\columnwidth, trim={0 0.5cm 0 0.4cm}, clip]{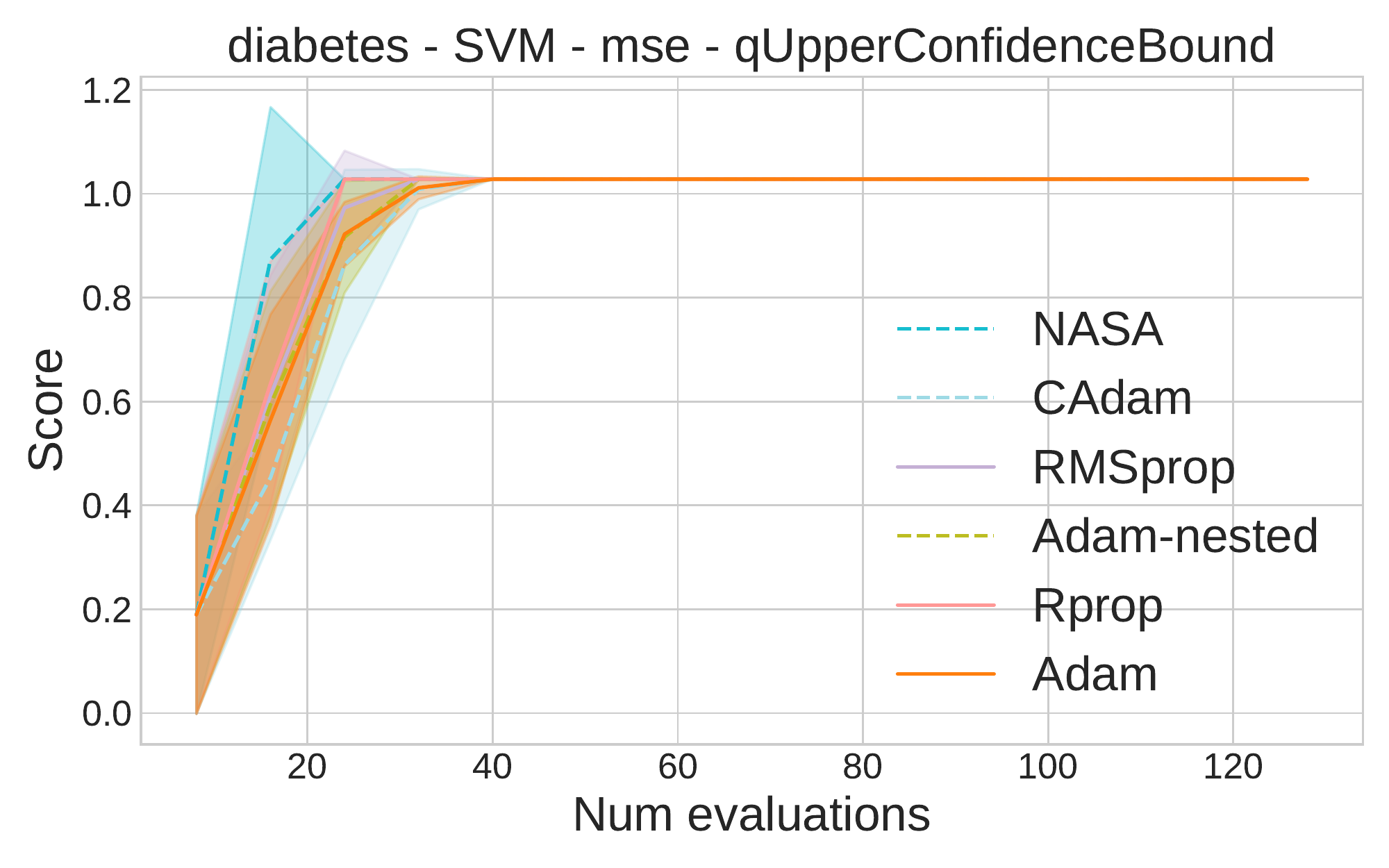}} & \hspace{-0.5cm}
    \subfloat{\includegraphics[width=0.19\columnwidth, trim={0 0.5cm 0 0.4cm}, clip]{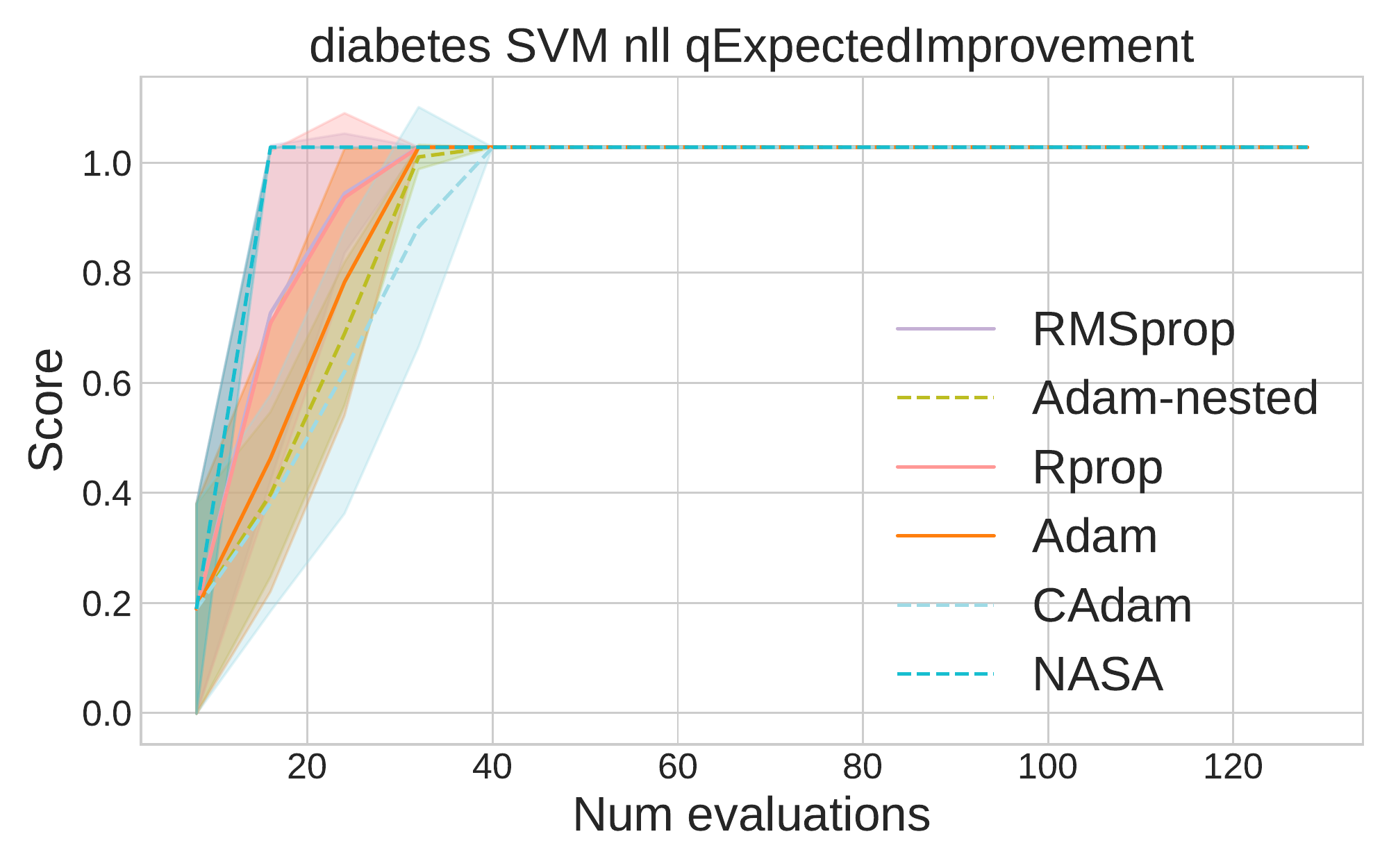}} &  \hspace{-0.5cm}
    \subfloat{\includegraphics[width=0.19\columnwidth, trim={0 0.5cm 0 0.4cm}, clip]{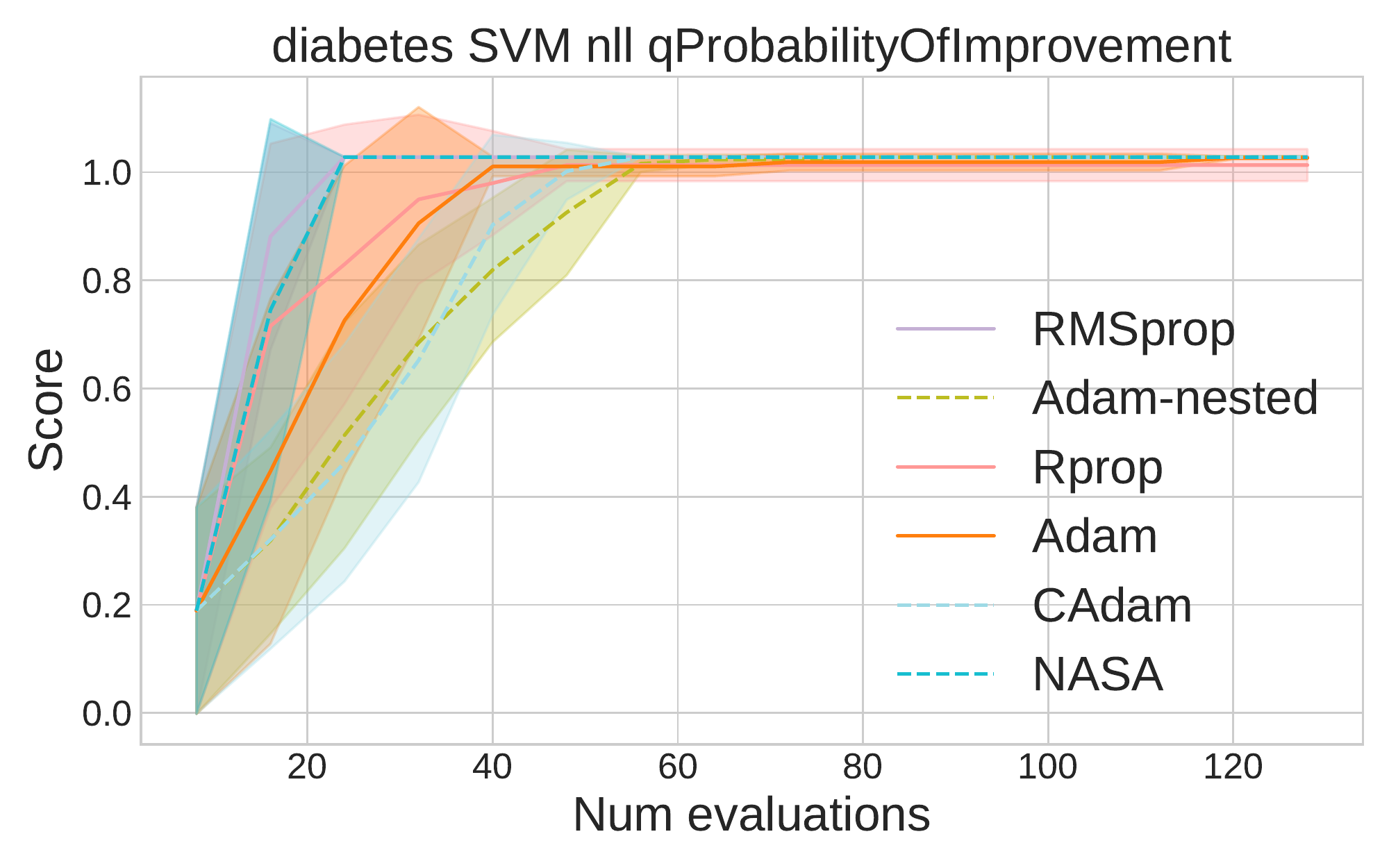}} & \hspace{-0.5cm}
    \subfloat{\includegraphics[width=0.19\columnwidth, trim={0 0.5cm 0 0.4cm}, clip]{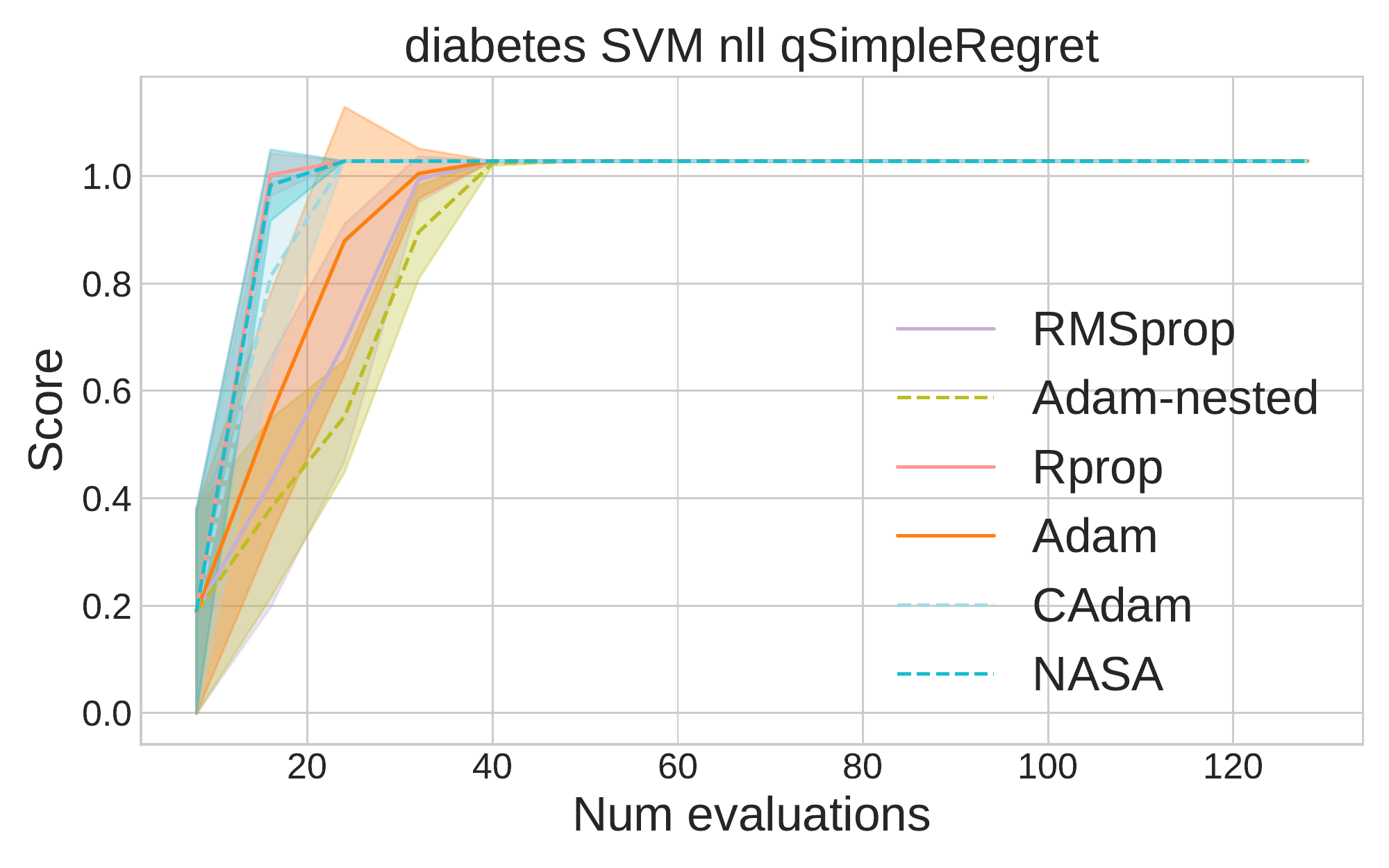}} 
    \end{tabular}
    \caption{}
    \end{figure}
    }

    {\renewcommand{\arraystretch}{0}
    \begin{figure}
    \begin{tabular}{ccccc}
    \subfloat{\includegraphics[width=0.19\columnwidth, trim={0 0.5cm 0 0.4cm}, clip]{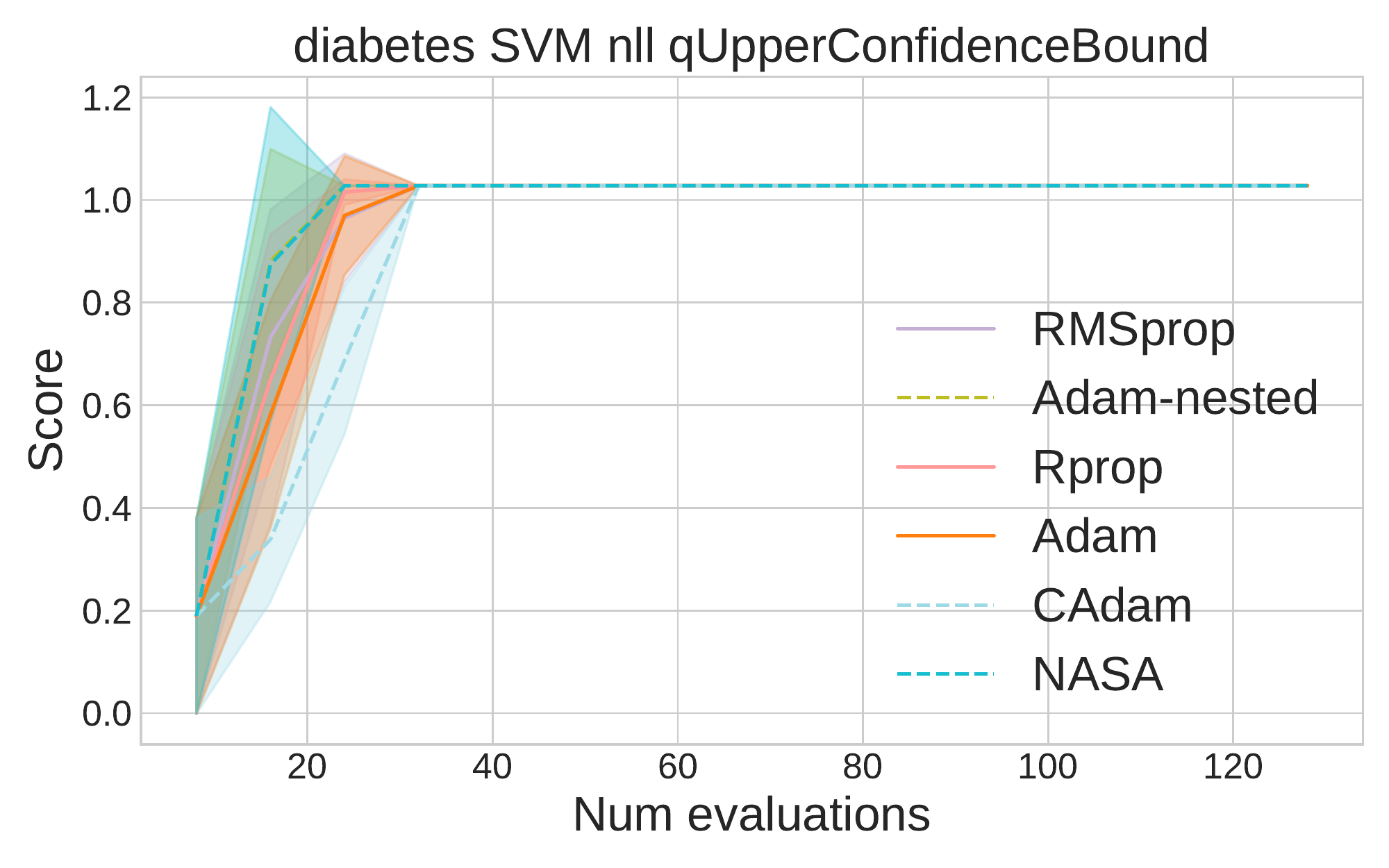}} &  \hspace{-0.5cm}
    \subfloat{\includegraphics[width=0.19\columnwidth, trim={0 0.5cm 0 0.4cm}, clip]{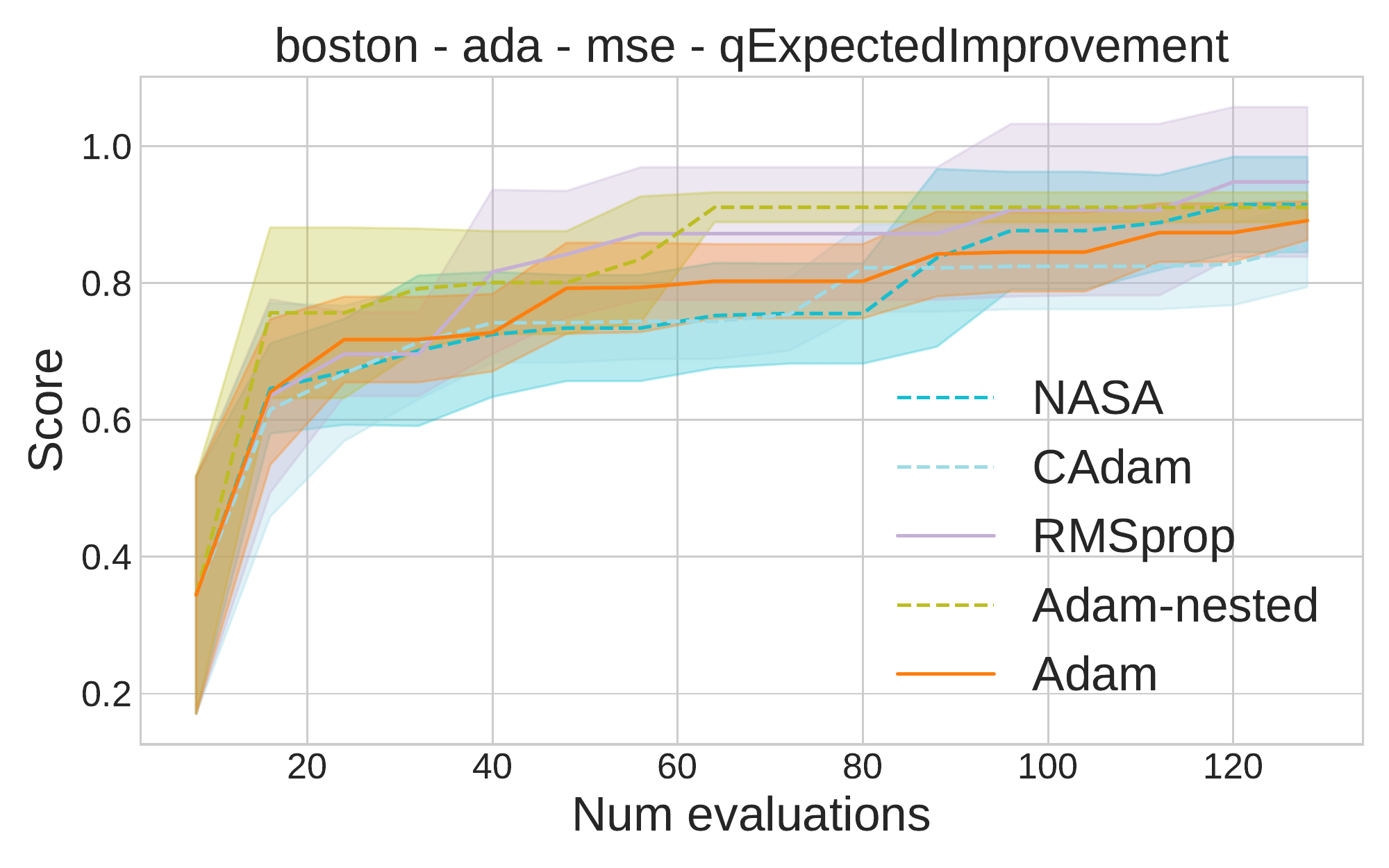}} & \hspace{-0.5cm}
    \subfloat{\includegraphics[width=0.19\columnwidth, trim={0 0.5cm 0 0.4cm}, clip]{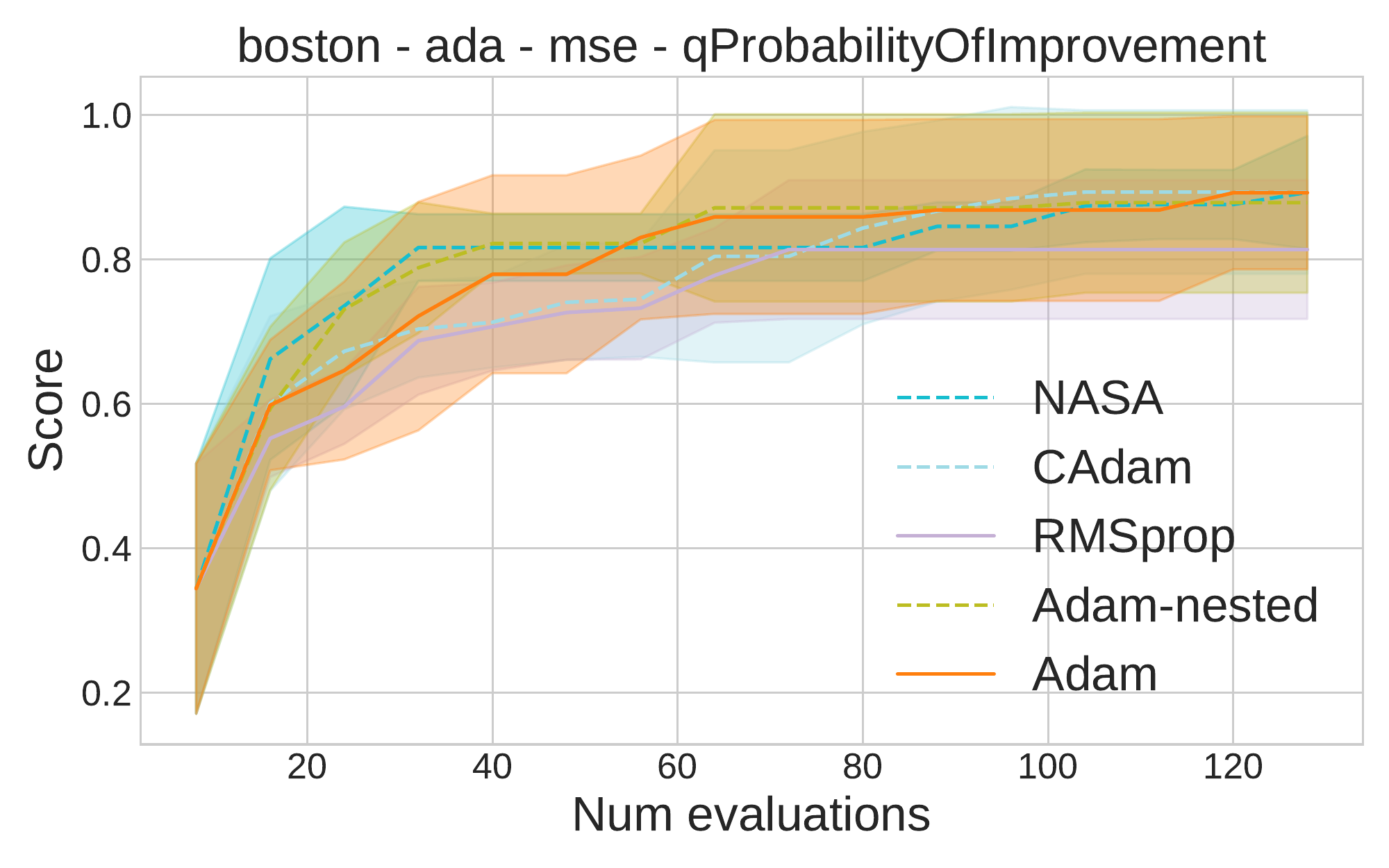}} &  \hspace{-0.5cm}
    \subfloat{\includegraphics[width=0.19\columnwidth, trim={0 0.5cm 0 0.4cm}, clip]{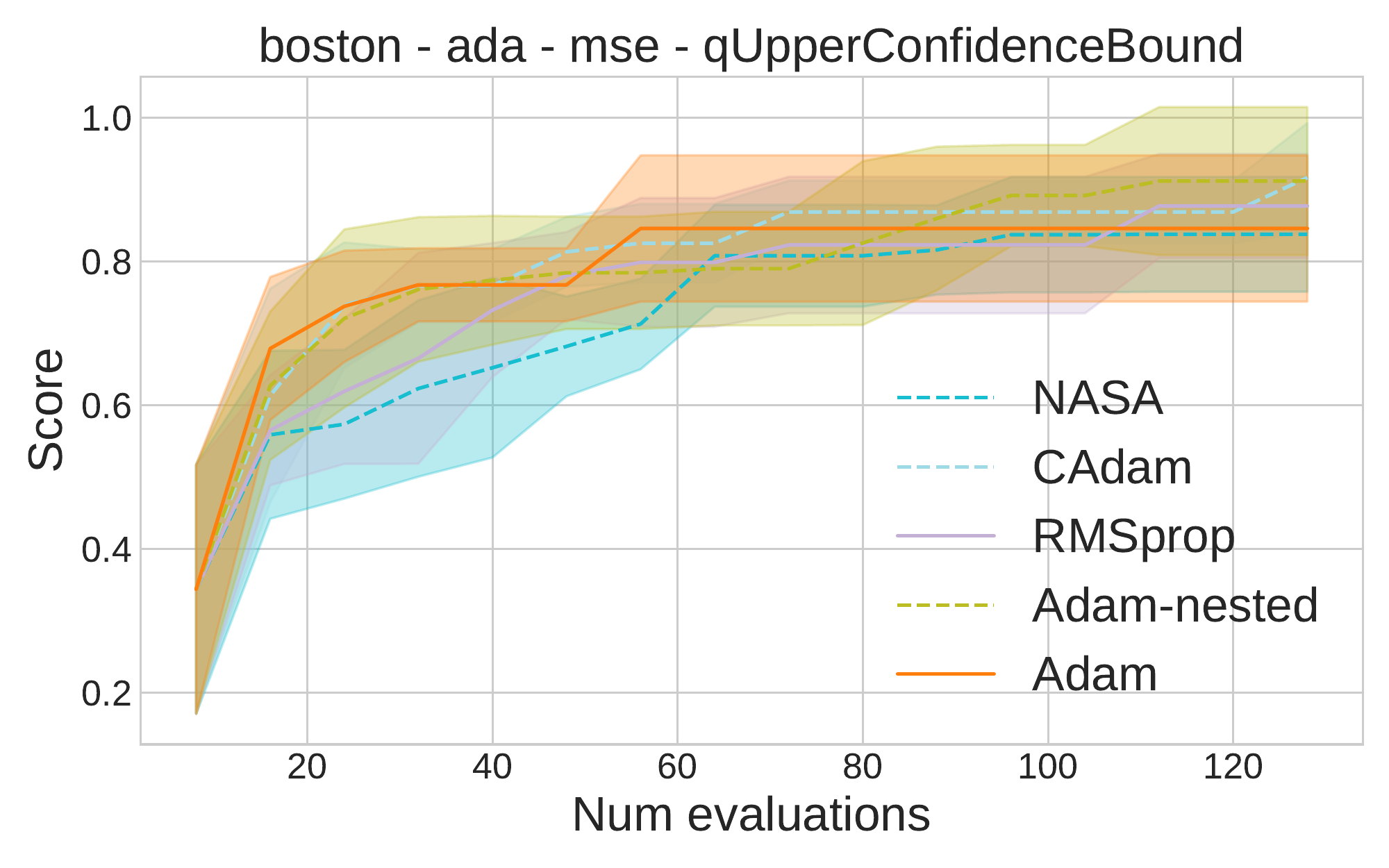}} & \hspace{-0.5cm}
    \subfloat{\includegraphics[width=0.19\columnwidth, trim={0 0.5cm 0 0.4cm}, clip]{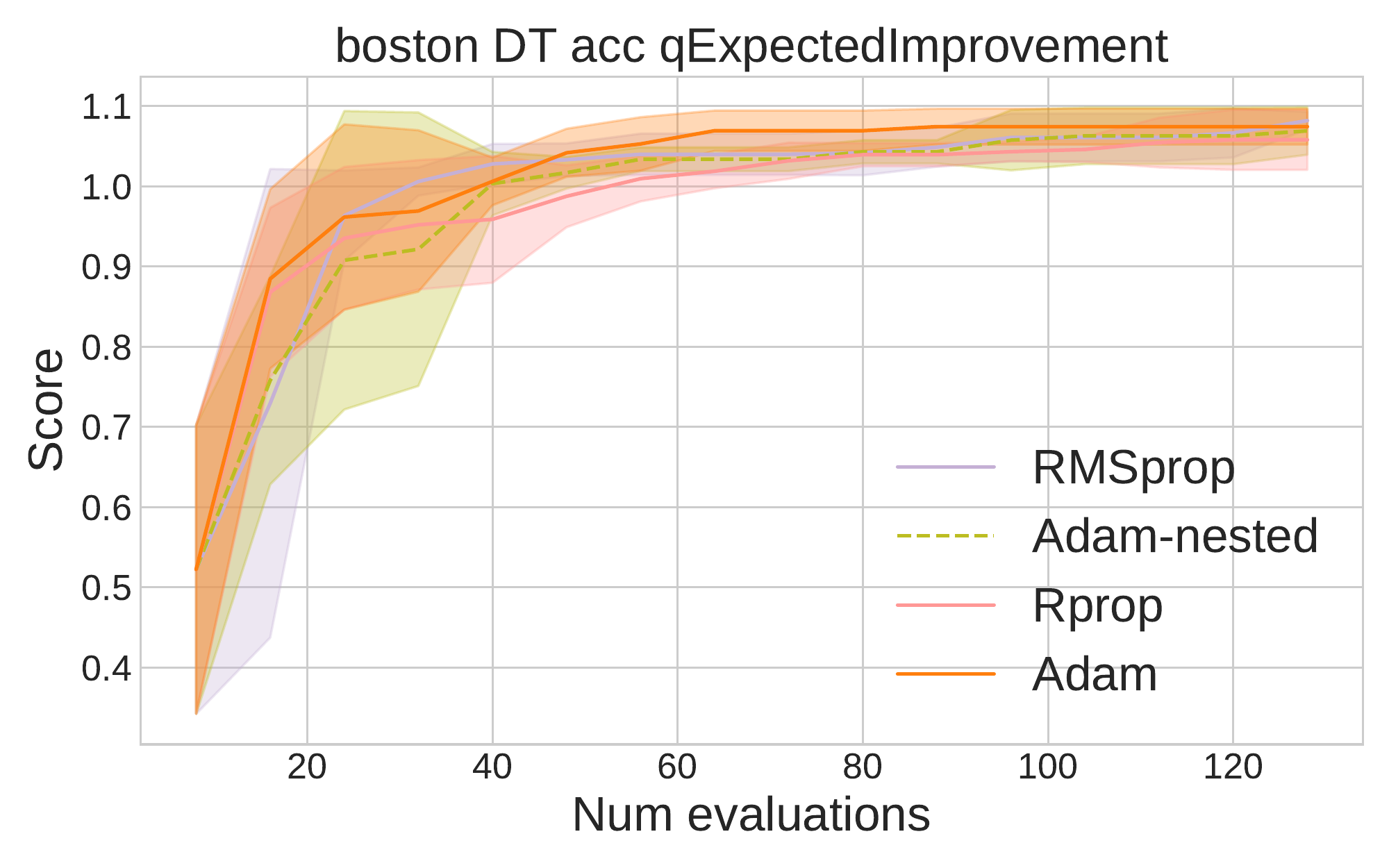}} \\  
    \subfloat{\includegraphics[width=0.19\columnwidth, trim={0 0.5cm 0 0.4cm}, clip]{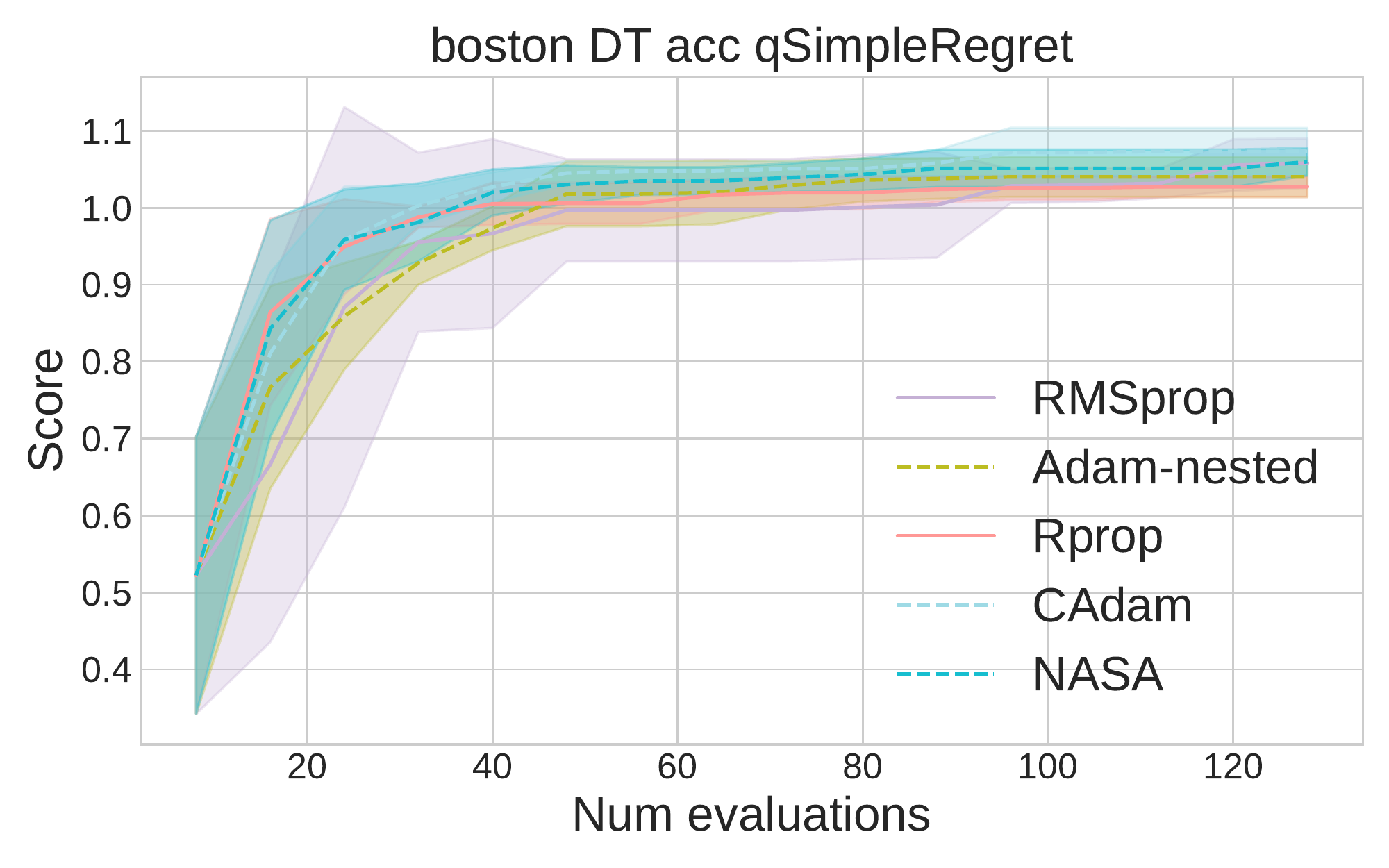}} &   \hspace{-0.5cm} 
    \subfloat{\includegraphics[width=0.19\columnwidth, trim={0 0.5cm 0 0.4cm}, clip]{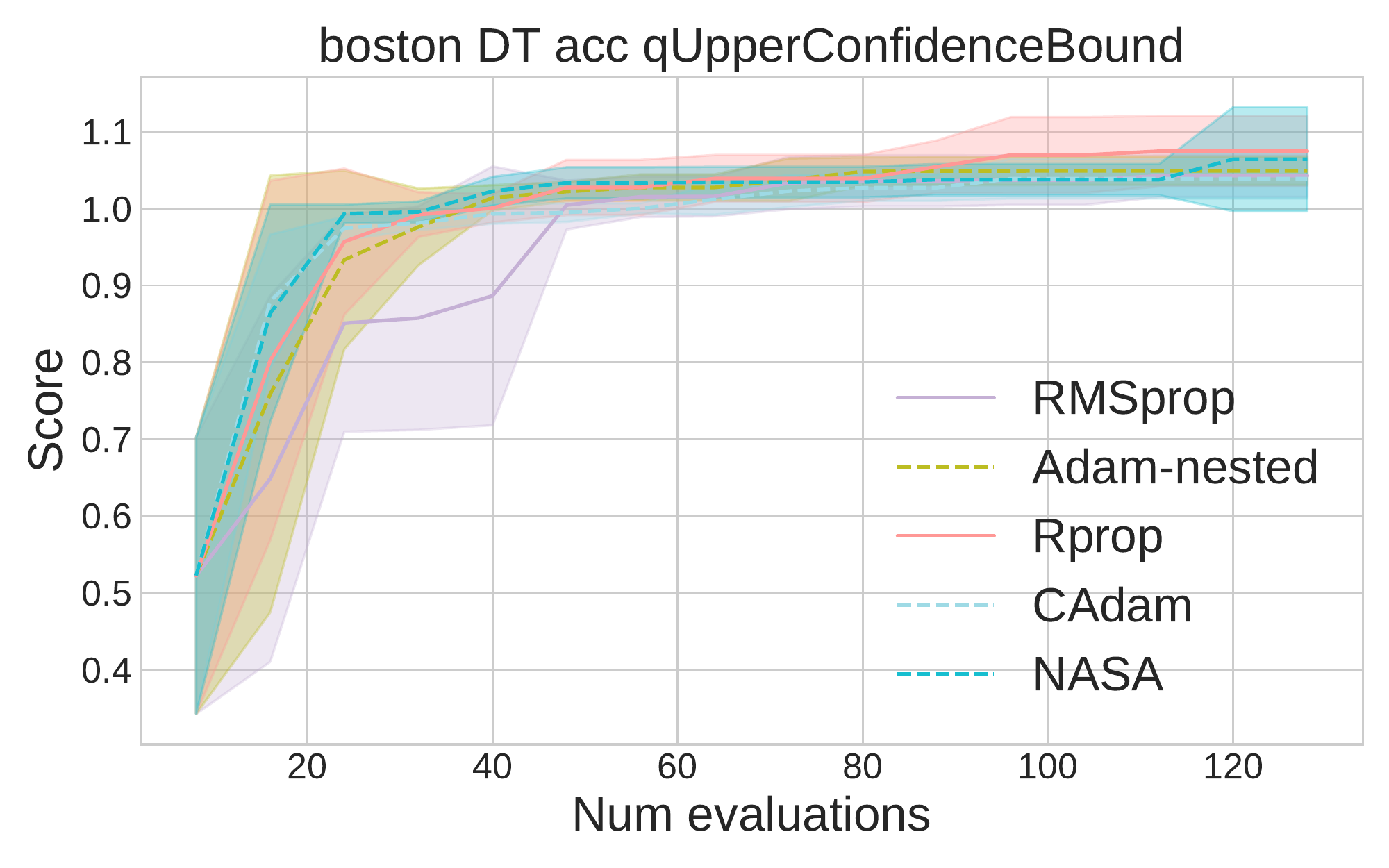}} & \hspace{-0.5cm}
    \subfloat{\includegraphics[width=0.19\columnwidth, trim={0 0.5cm 0 0.4cm}, clip]{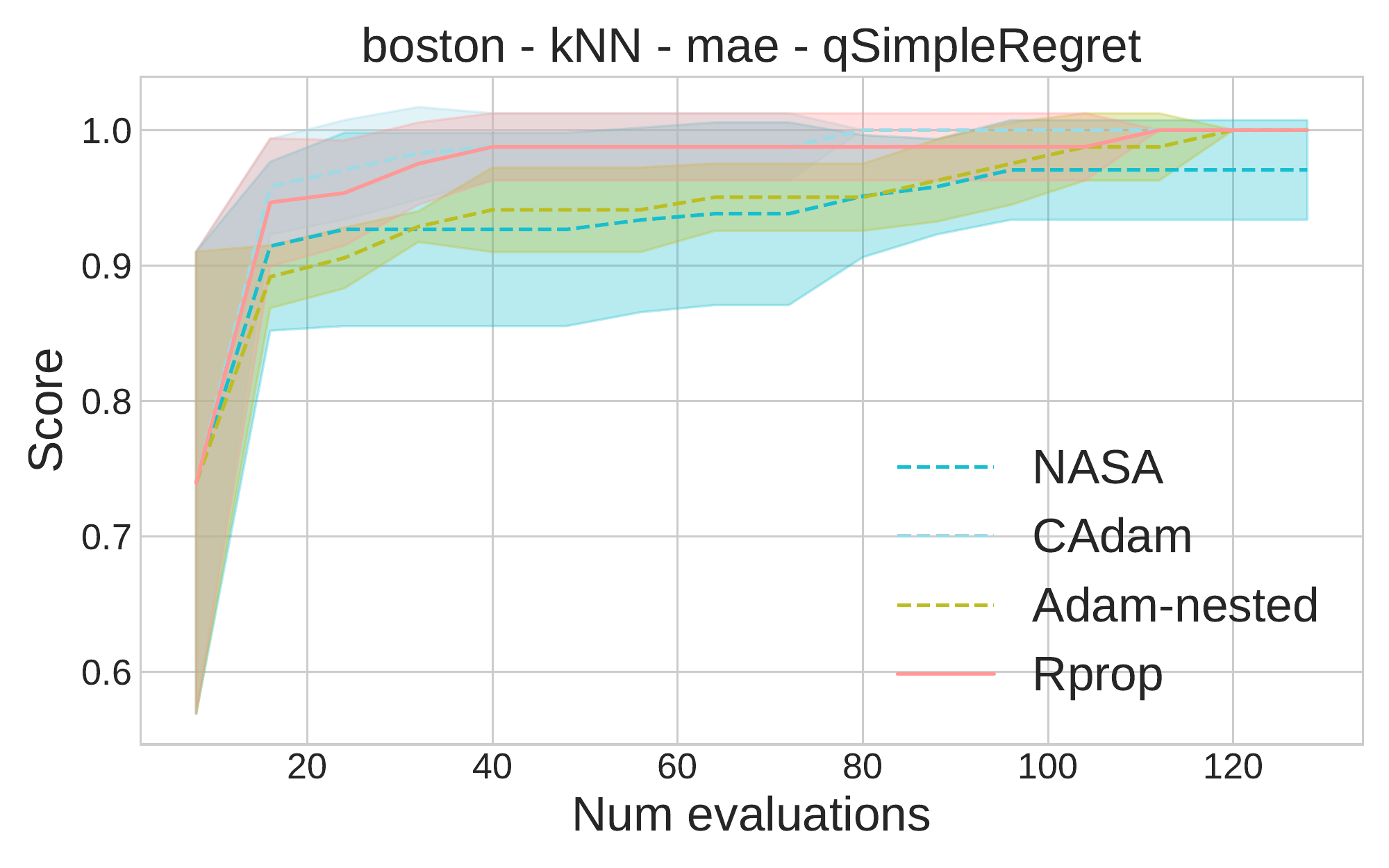}} &  \hspace{-0.5cm}
    \subfloat{\includegraphics[width=0.19\columnwidth, trim={0 0.5cm 0 0.4cm}, clip]{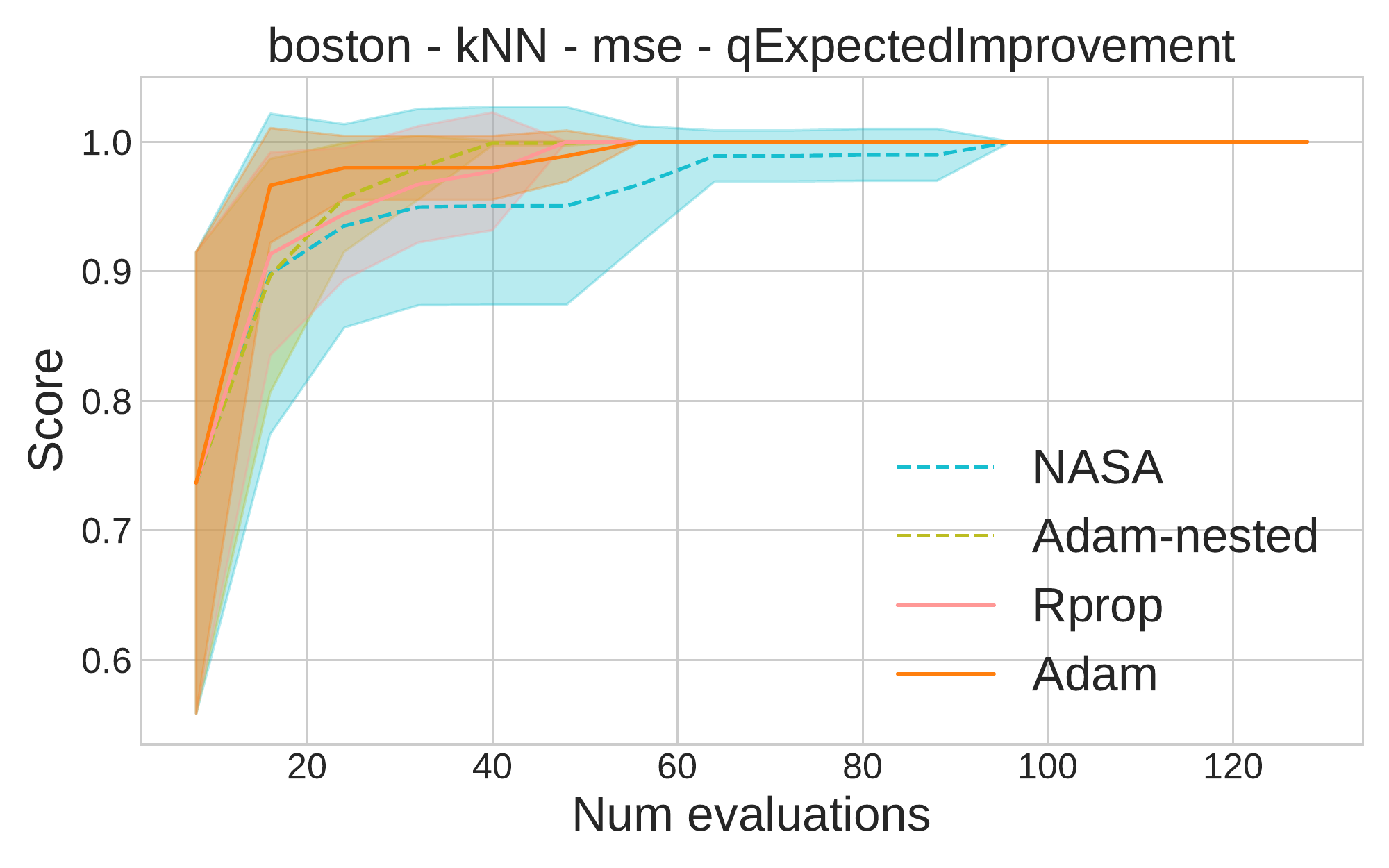}} & \hspace{-0.5cm}
    \subfloat{\includegraphics[width=0.19\columnwidth, trim={0 0.5cm 0 0.4cm}, clip]{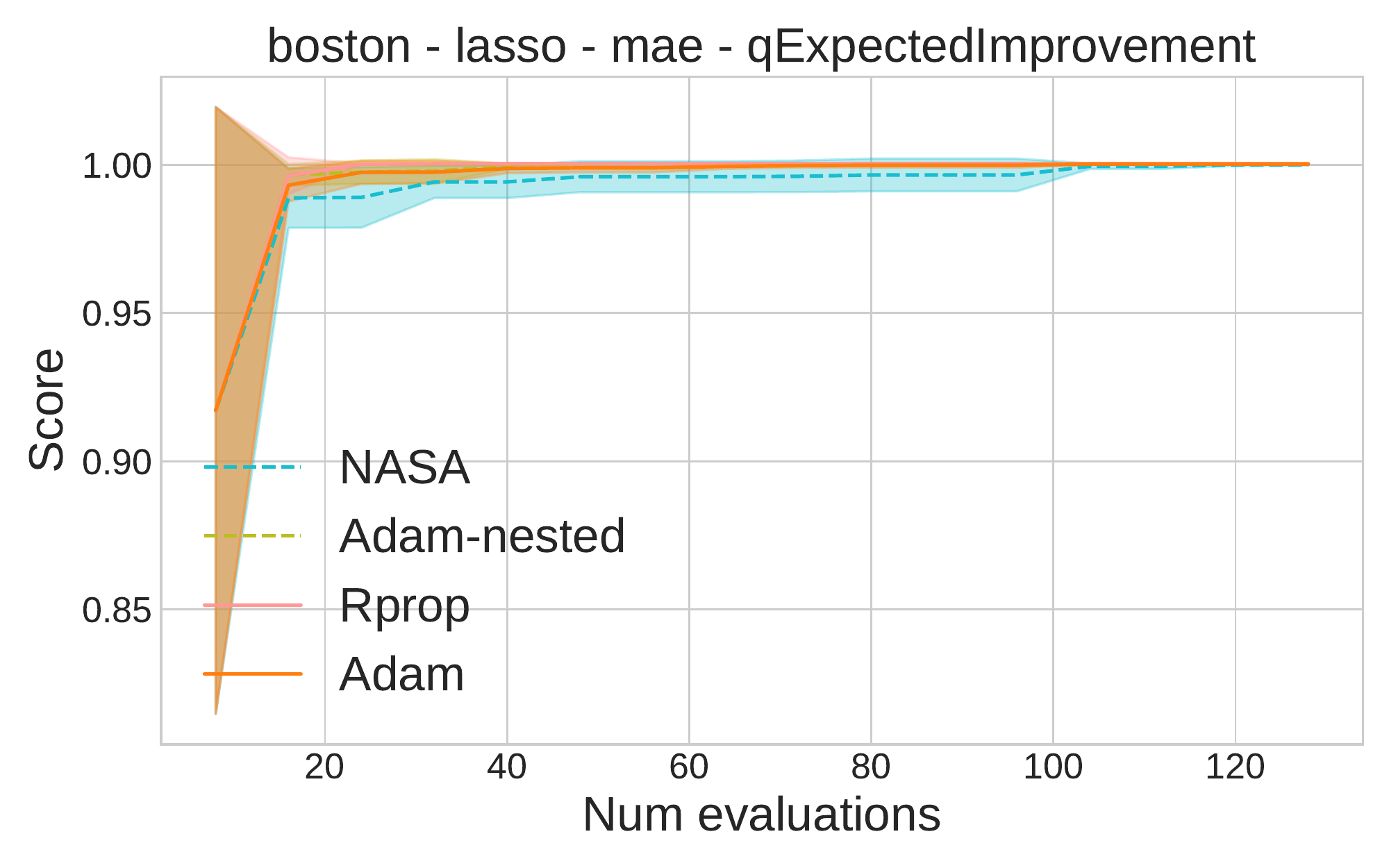}} \\  
    \subfloat{\includegraphics[width=0.19\columnwidth, trim={0 0.5cm 0 0.4cm}, clip]{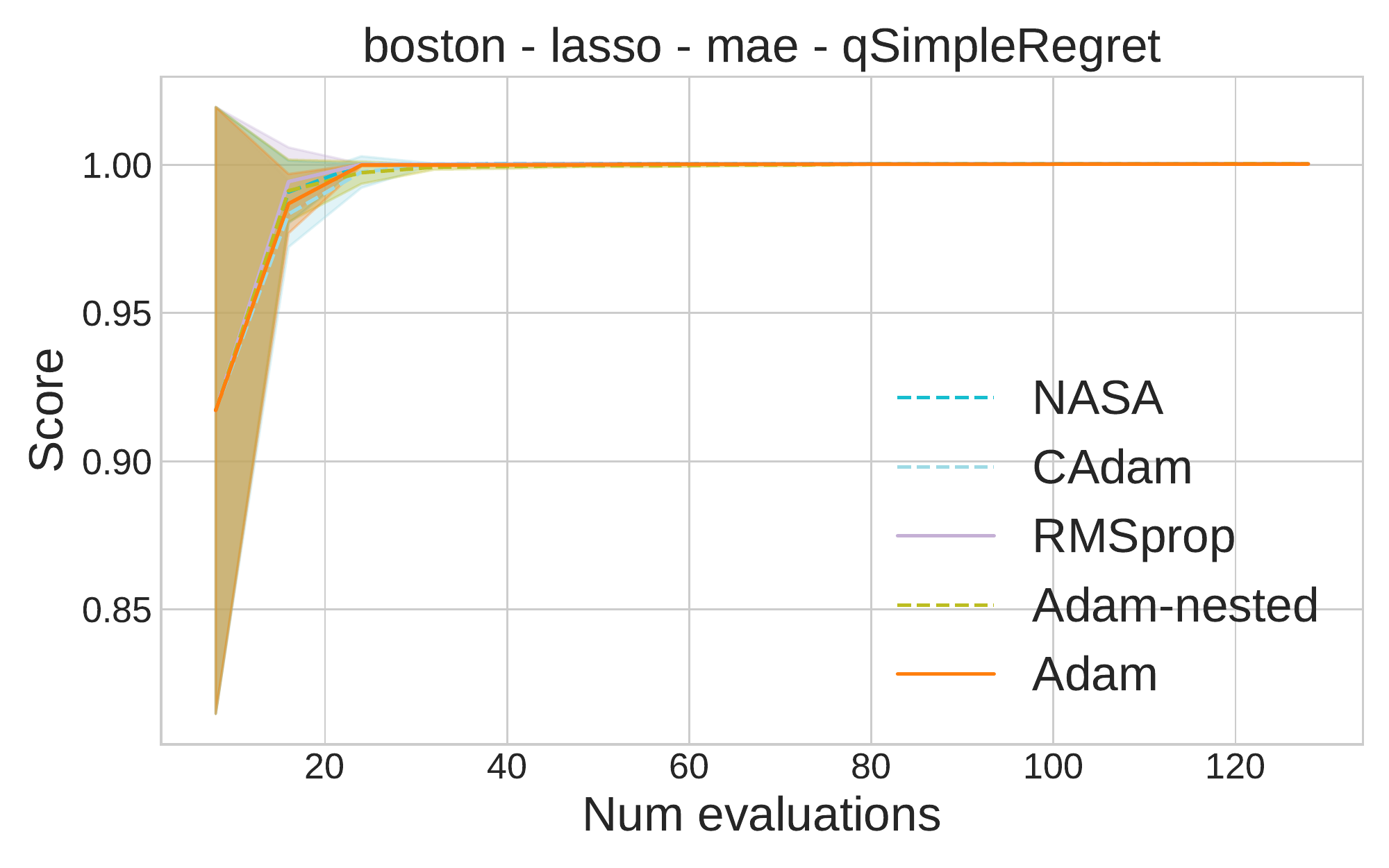}} &   \hspace{-0.5cm} 
    \subfloat{\includegraphics[width=0.19\columnwidth, trim={0 0.5cm 0 0.4cm}, clip]{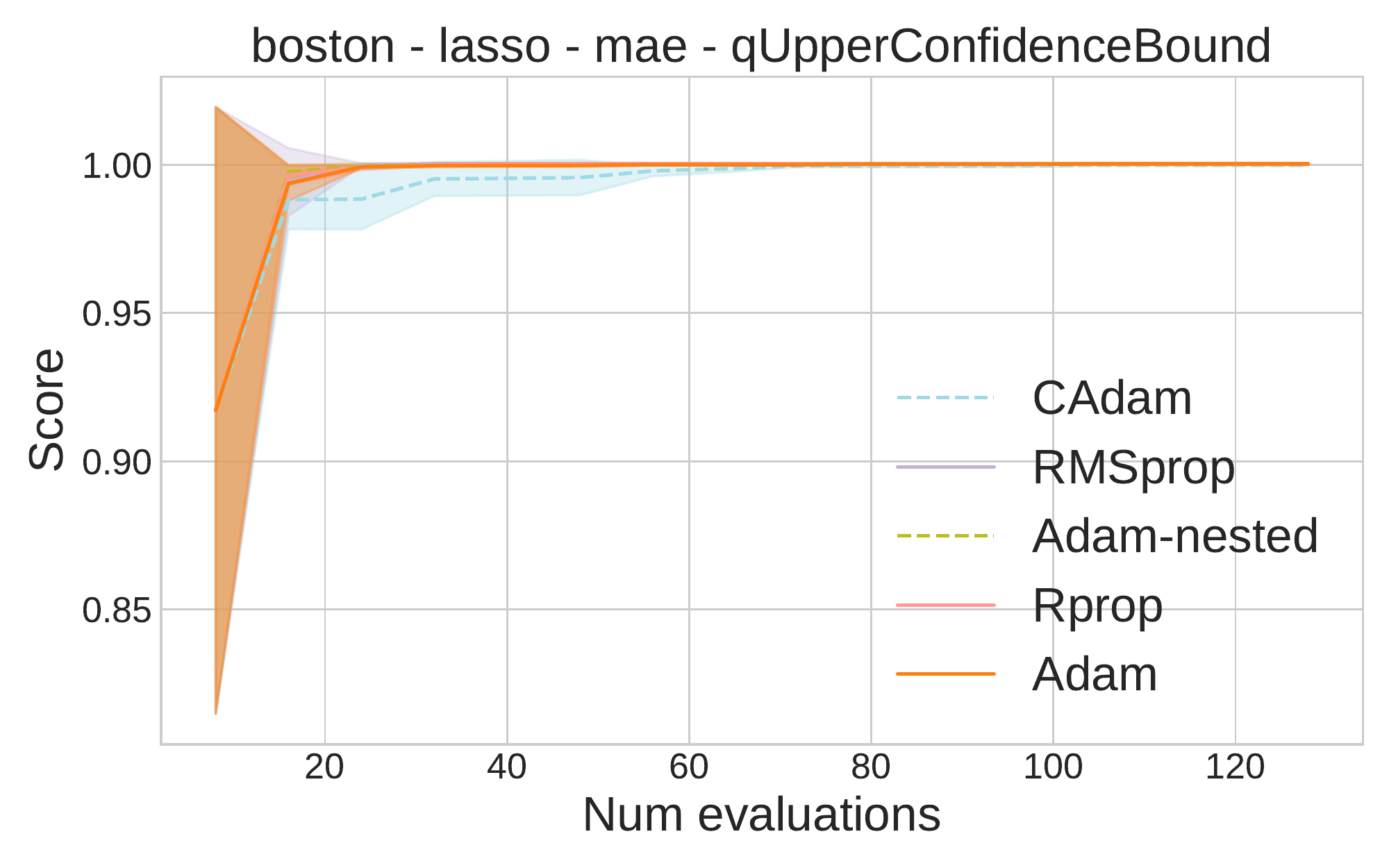}} & \hspace{-0.5cm}
    \subfloat{\includegraphics[width=0.19\columnwidth, trim={0 0.5cm 0 0.4cm}, clip]{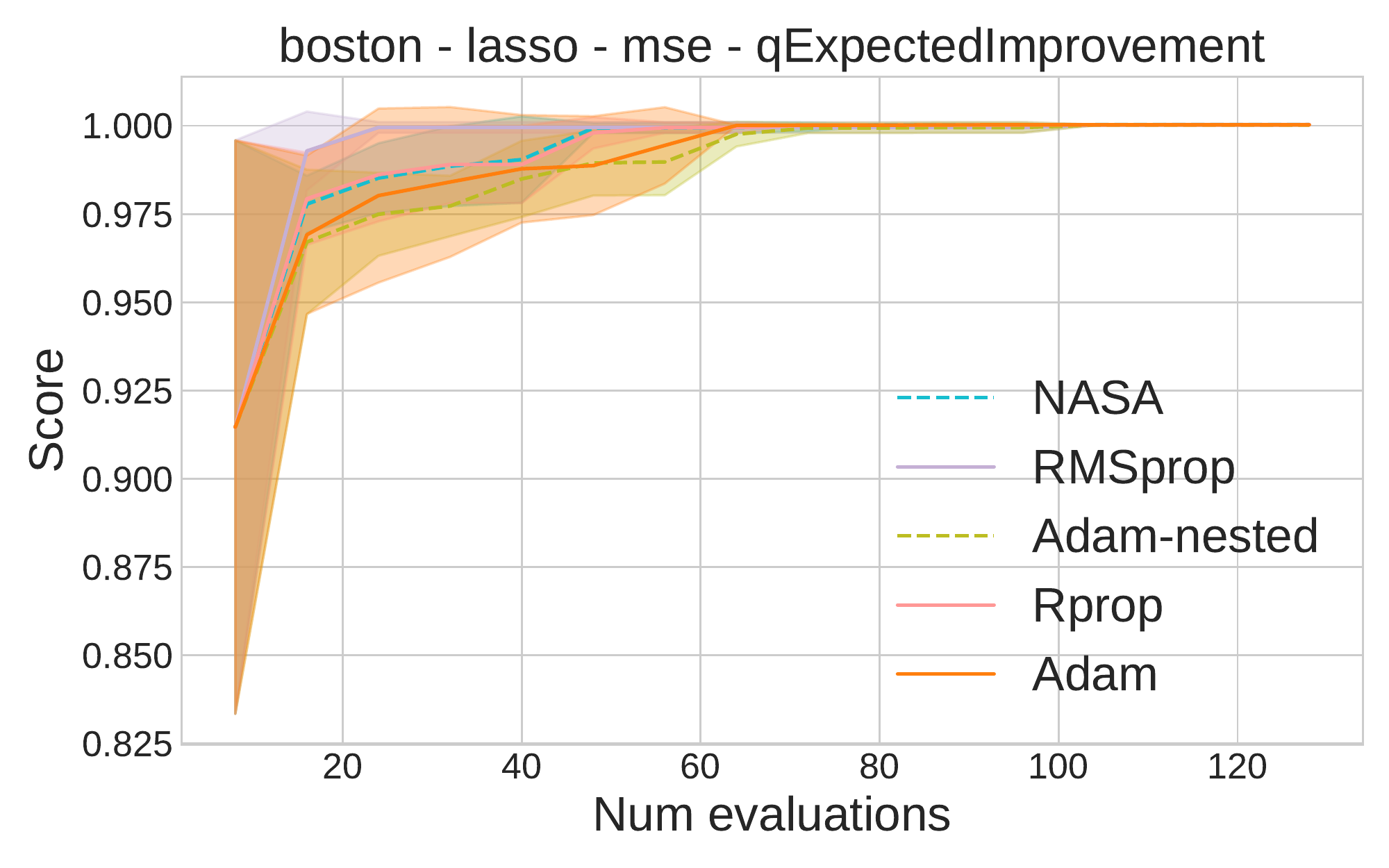}} &  \hspace{-0.5cm}
    \subfloat{\includegraphics[width=0.19\columnwidth, trim={0 0.5cm 0 0.4cm}, clip]{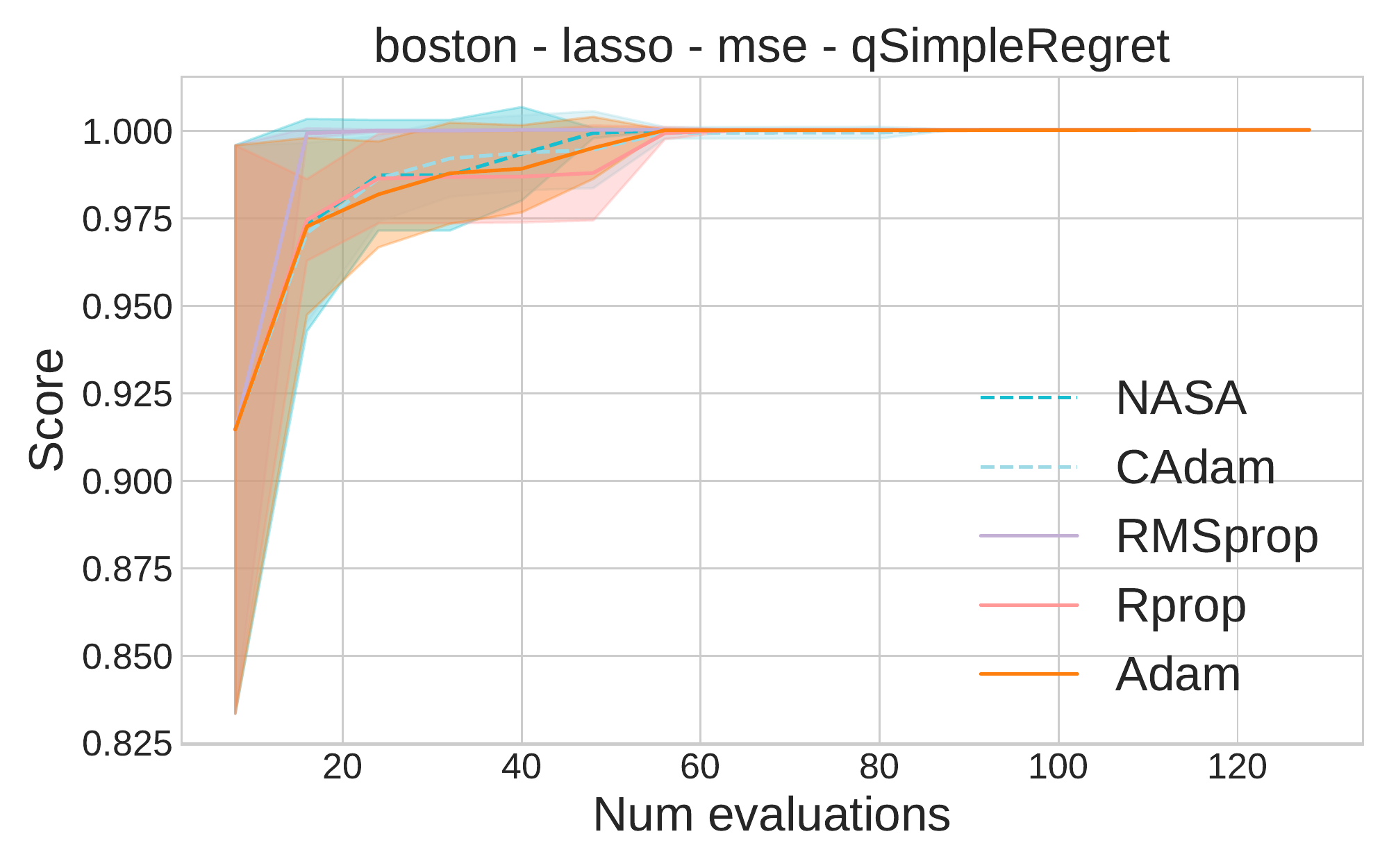}} & \hspace{-0.5cm}
    \subfloat{\includegraphics[width=0.19\columnwidth, trim={0 0.5cm 0 0.4cm}, clip]{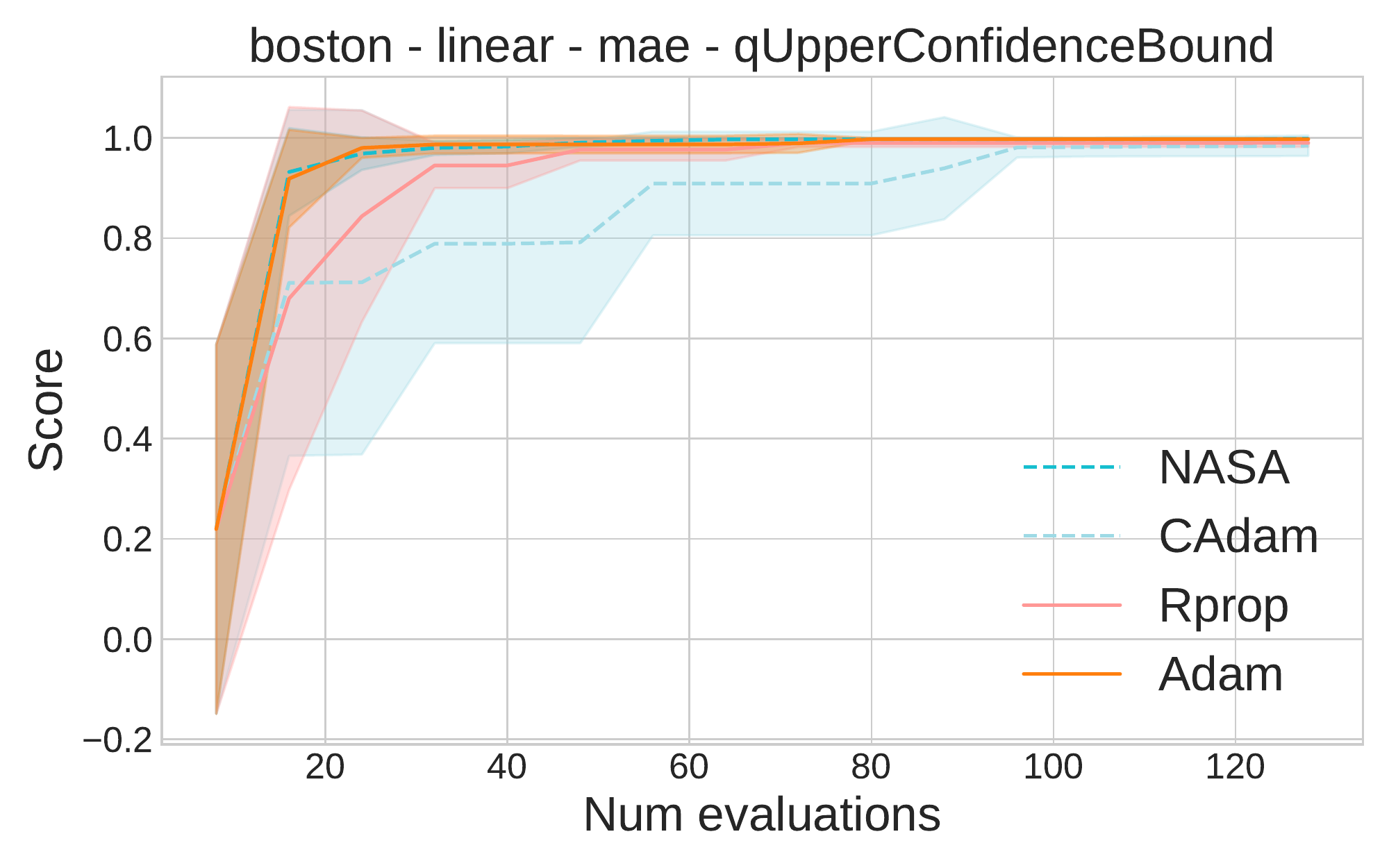}} \\  
    \subfloat{\includegraphics[width=0.19\columnwidth, trim={0 0.5cm 0 0.4cm}, clip]{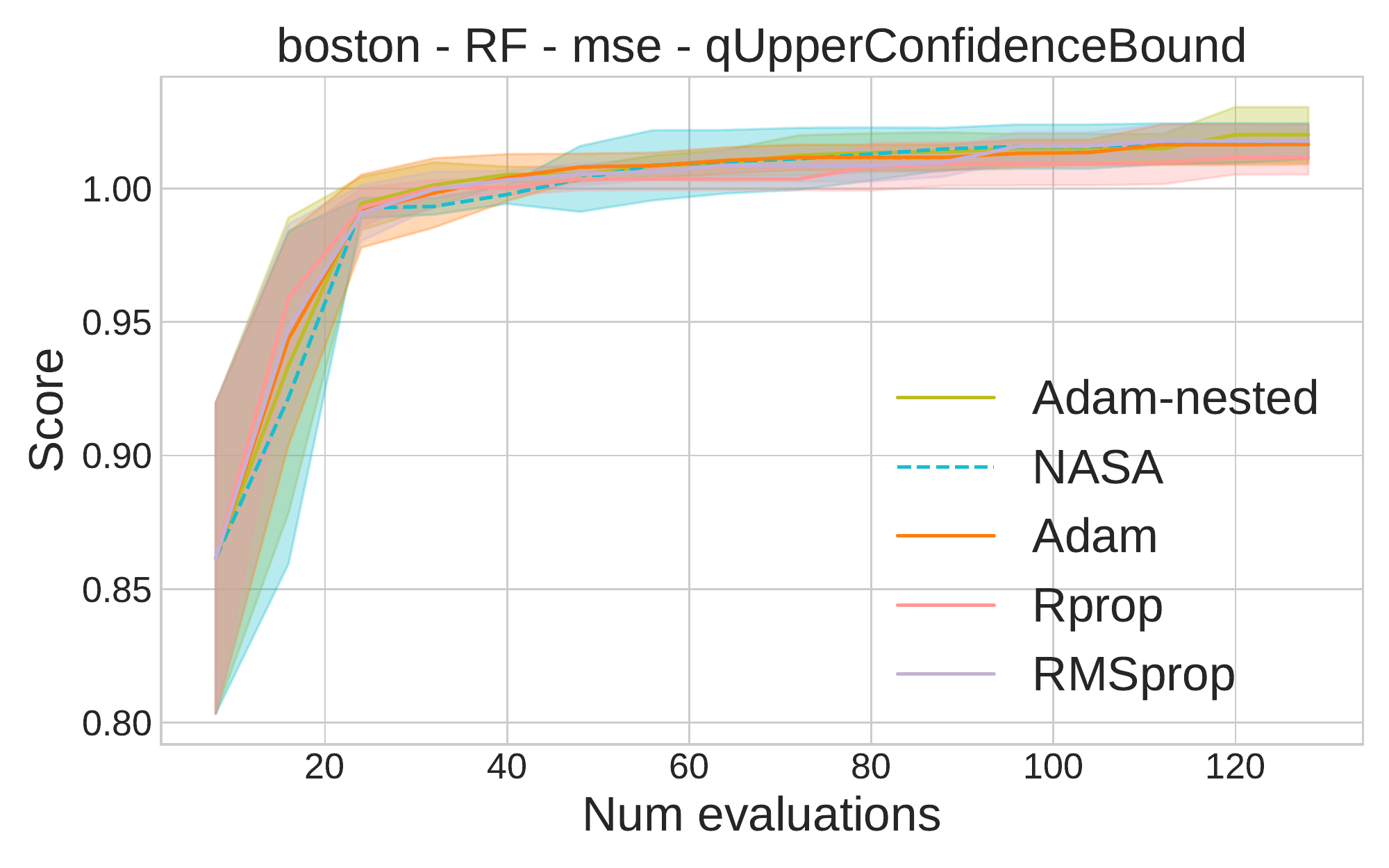}} &   \hspace{-0.5cm} 
    \subfloat{\includegraphics[width=0.19\columnwidth, trim={0 0.5cm 0 0.4cm}, clip]{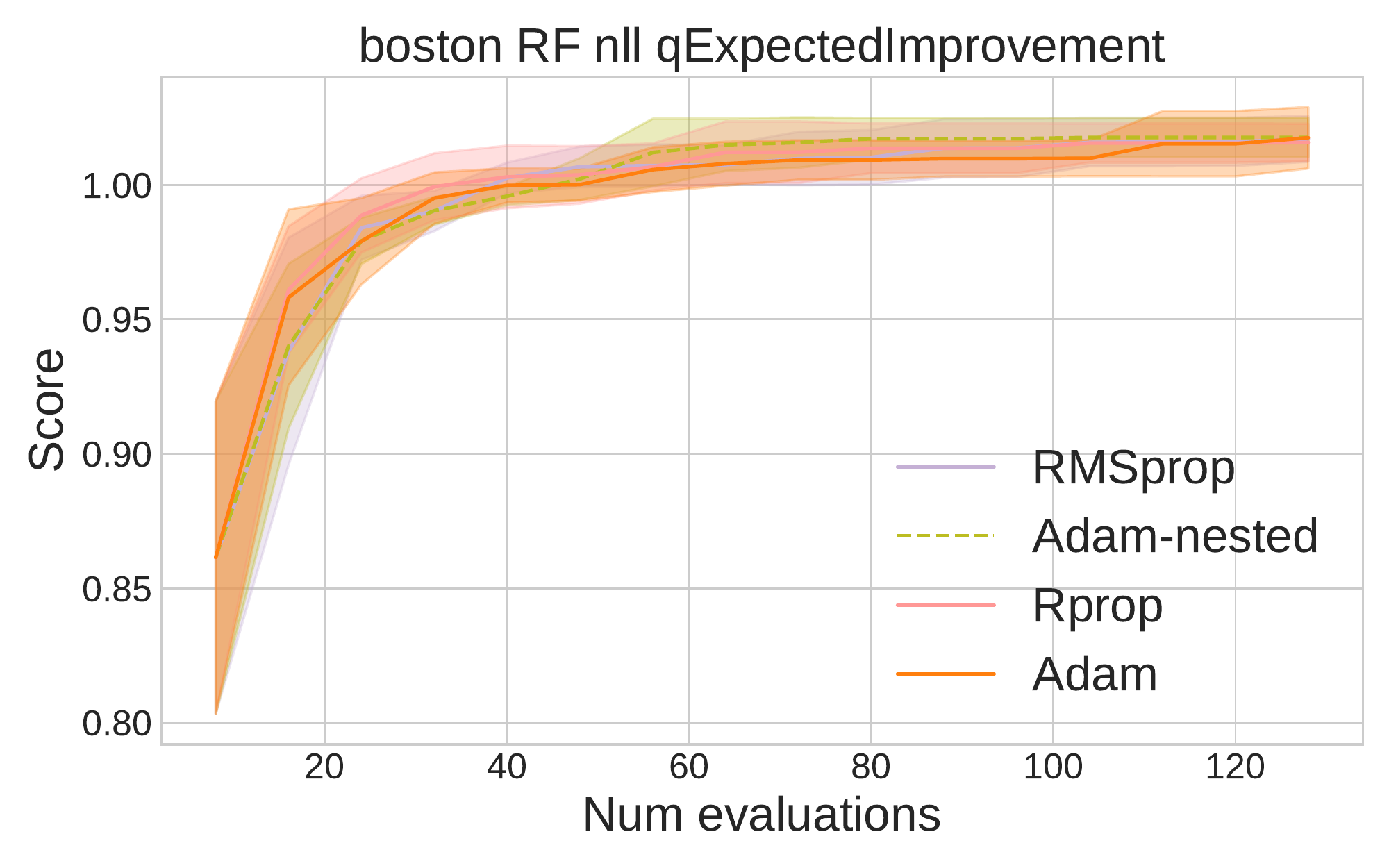}} & \hspace{-0.5cm}
    \subfloat{\includegraphics[width=0.19\columnwidth, trim={0 0.5cm 0 0.4cm}, clip]{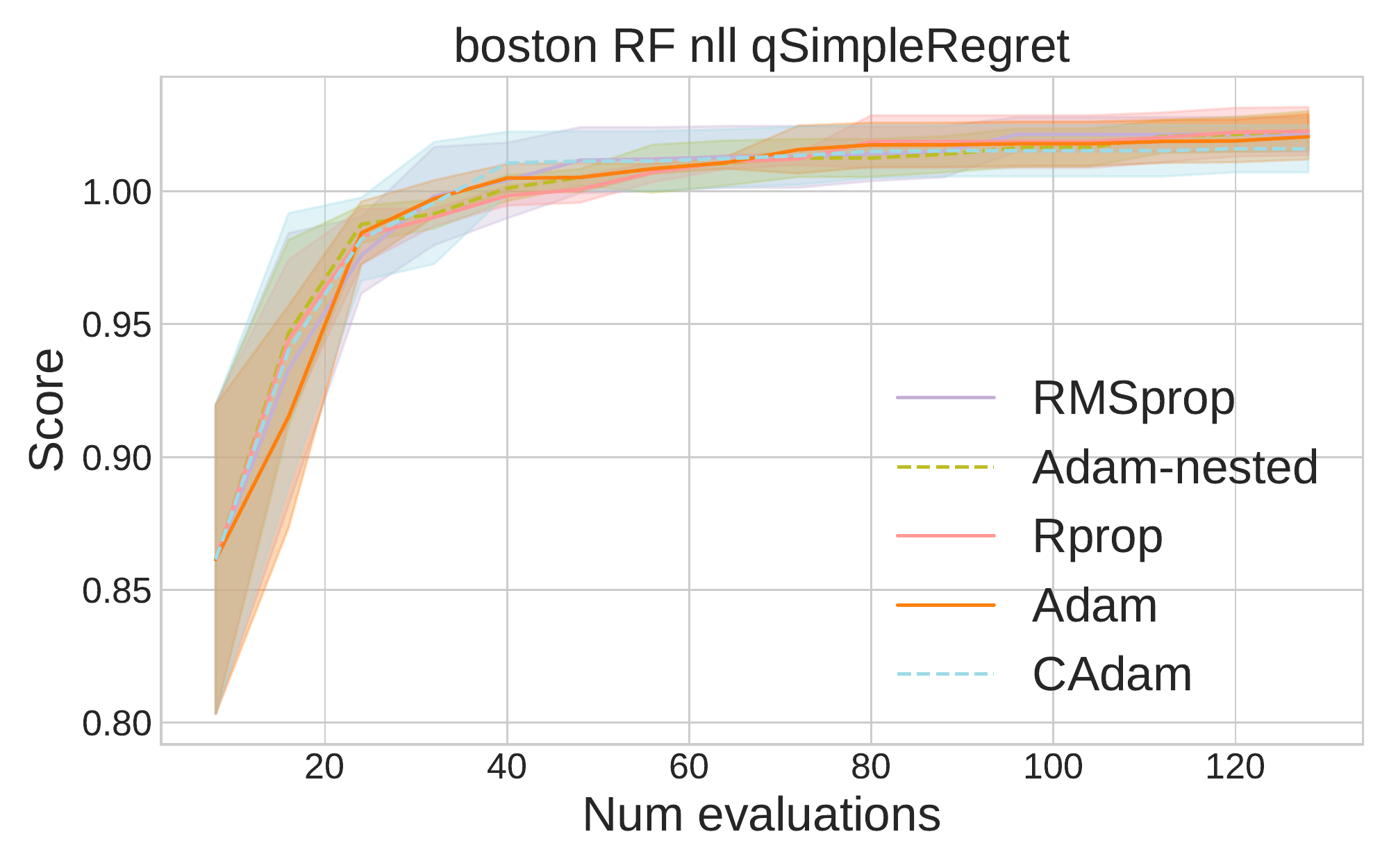}} &  \hspace{-0.5cm}
    \subfloat{\includegraphics[width=0.19\columnwidth, trim={0 0.5cm 0 0.4cm}, clip]{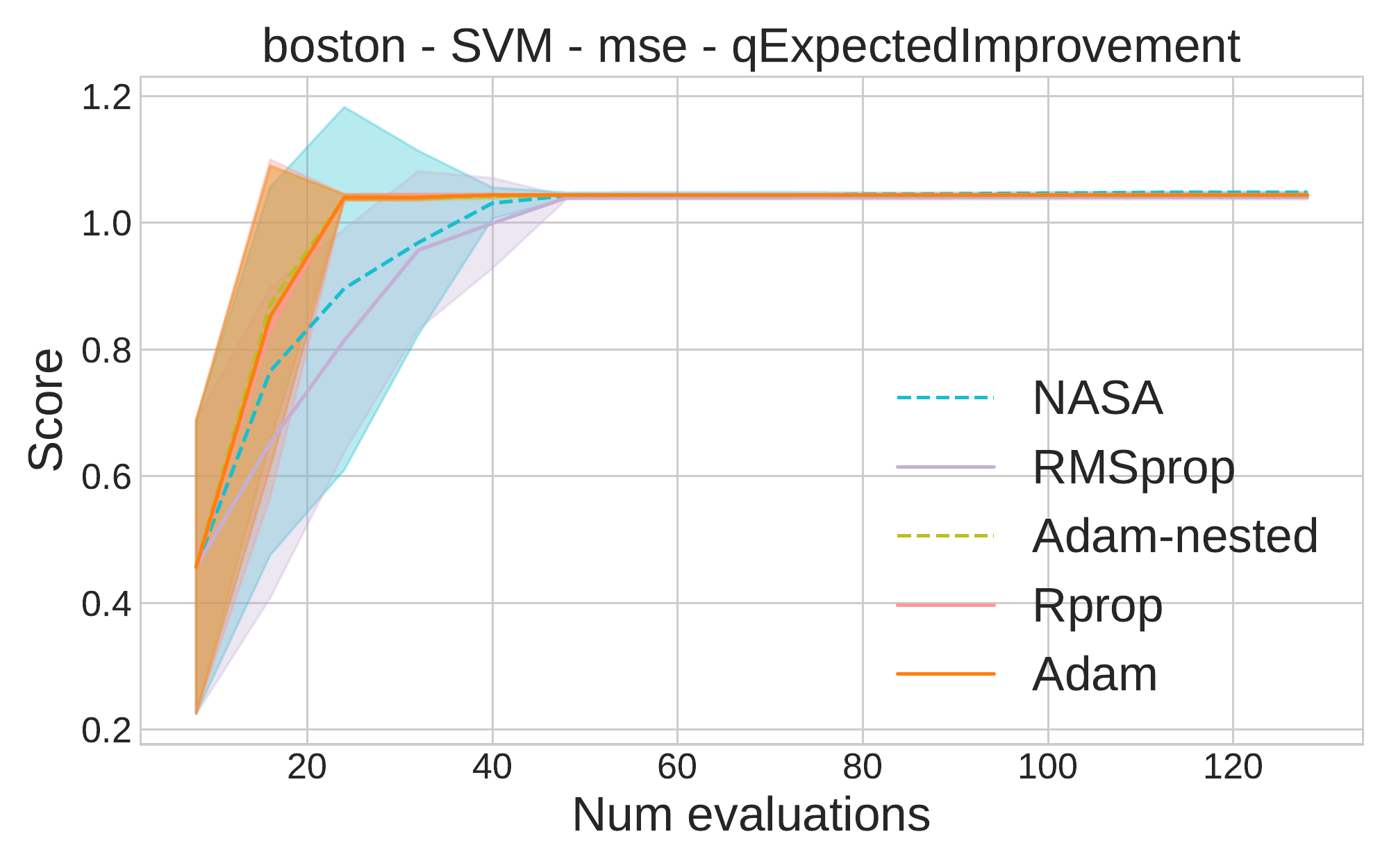}} & \hspace{-0.5cm}
    \subfloat{\includegraphics[width=0.19\columnwidth, trim={0 0.5cm 0 0.4cm}, clip]{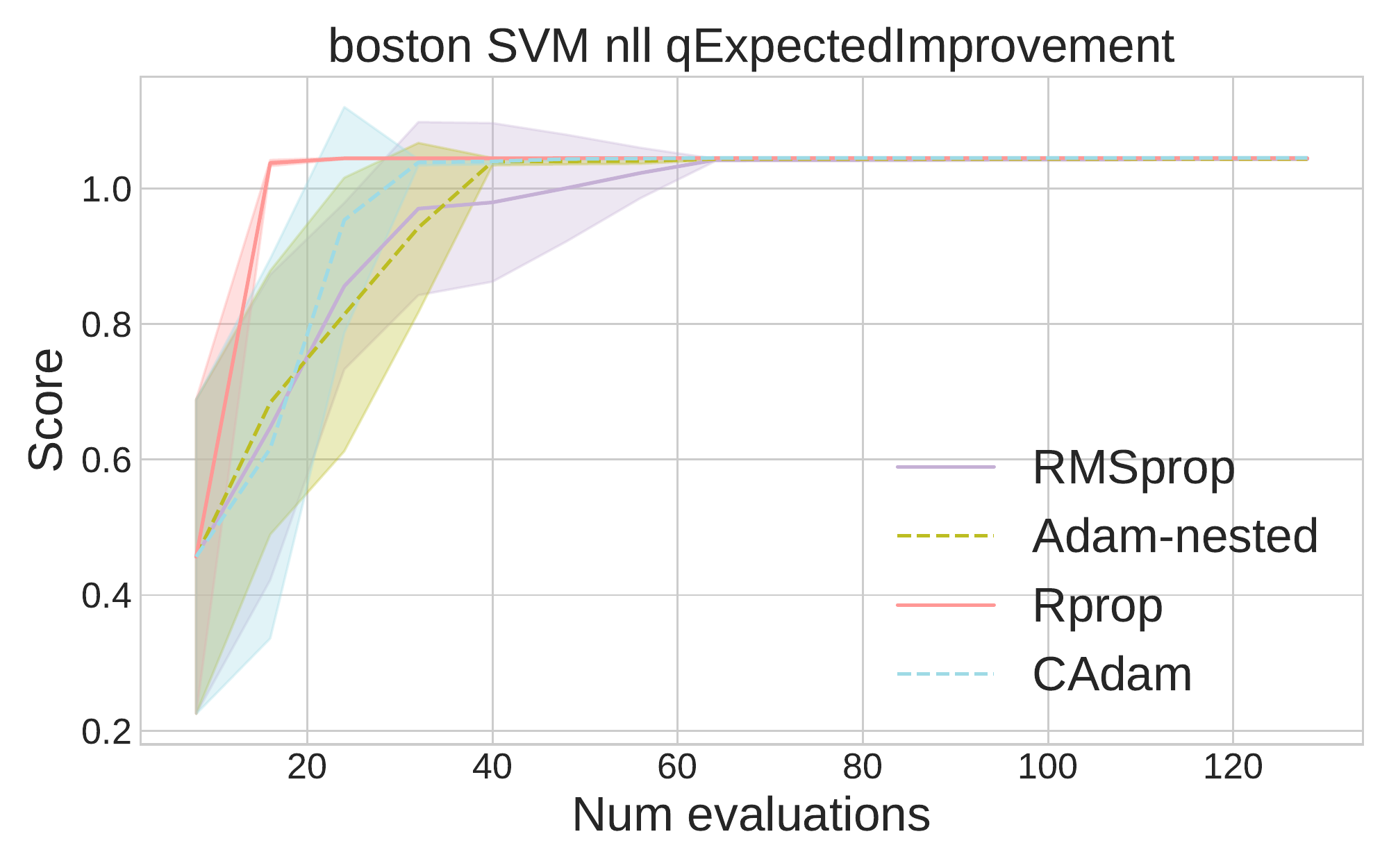}} \\  
    \subfloat{\includegraphics[width=0.19\columnwidth, trim={0 0.5cm 0 0.4cm}, clip]{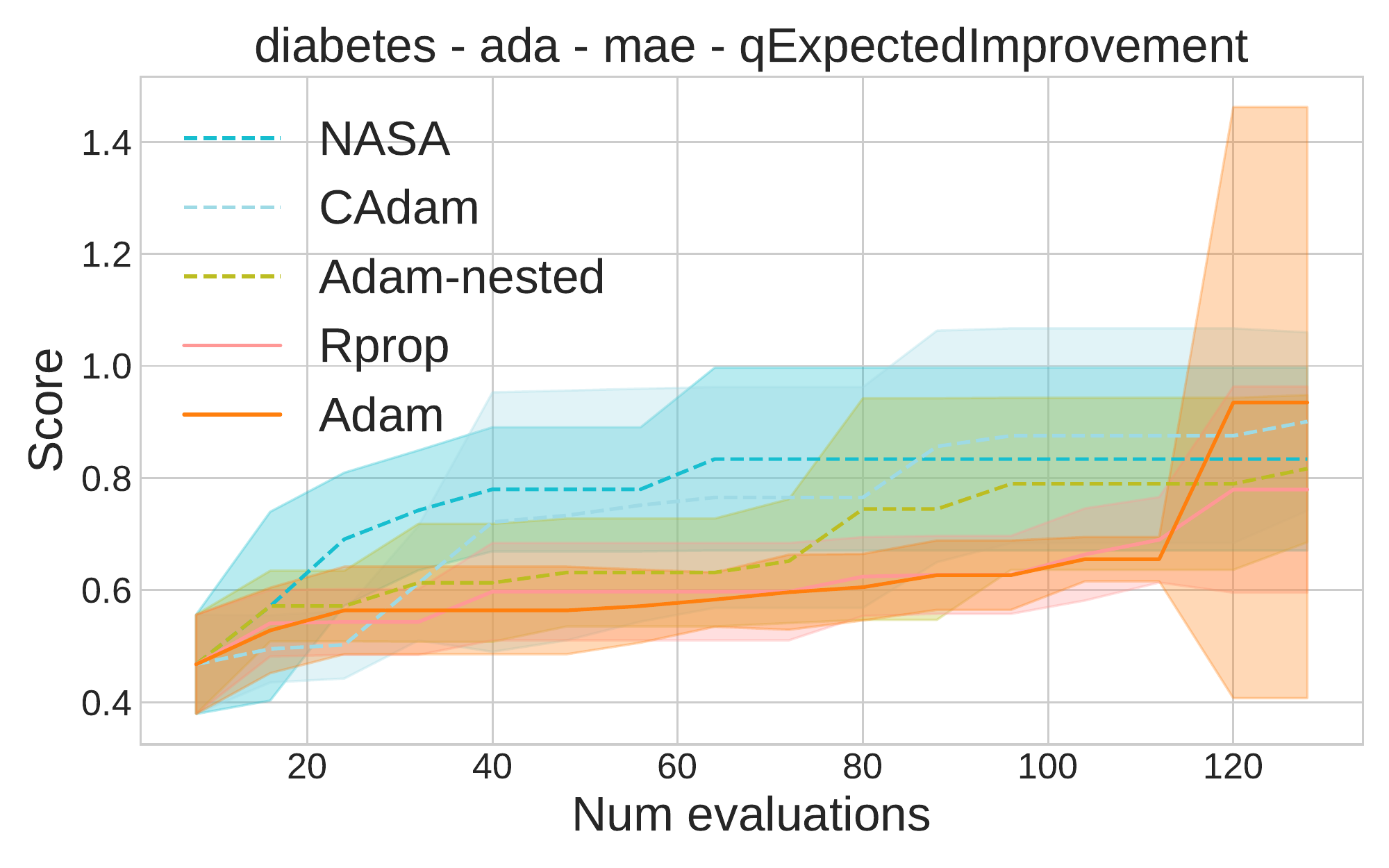}} &  \hspace{-0.5cm} 
    \subfloat{\includegraphics[width=0.19\columnwidth, trim={0 0.5cm 0 0.4cm}, clip]{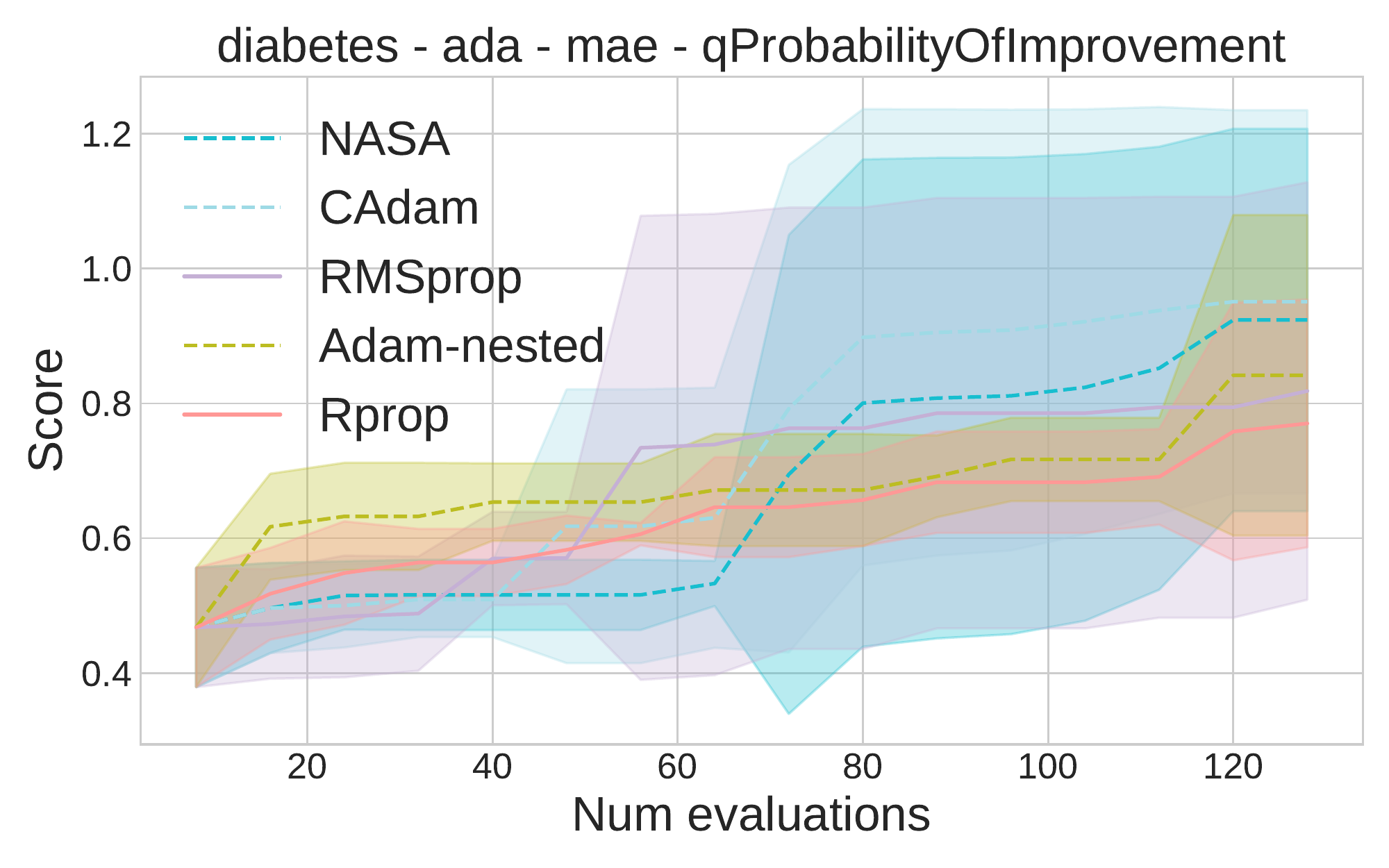}} & \hspace{-0.5cm}
    \subfloat{\includegraphics[width=0.19\columnwidth, trim={0 0.5cm 0 0.4cm}, clip]{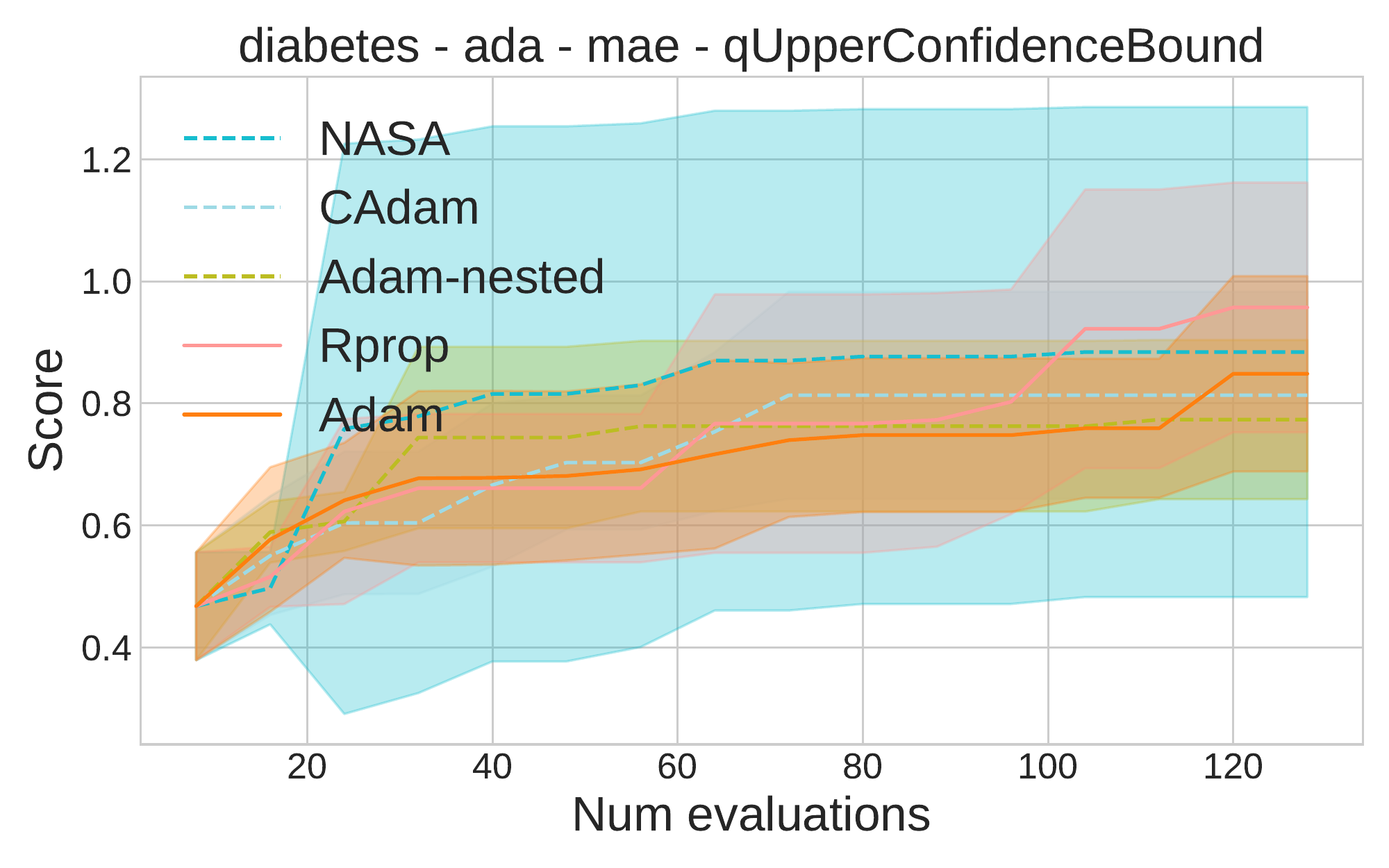}} &  \hspace{-0.5cm}
    \subfloat{\includegraphics[width=0.19\columnwidth, trim={0 0.5cm 0 0.4cm}, clip]{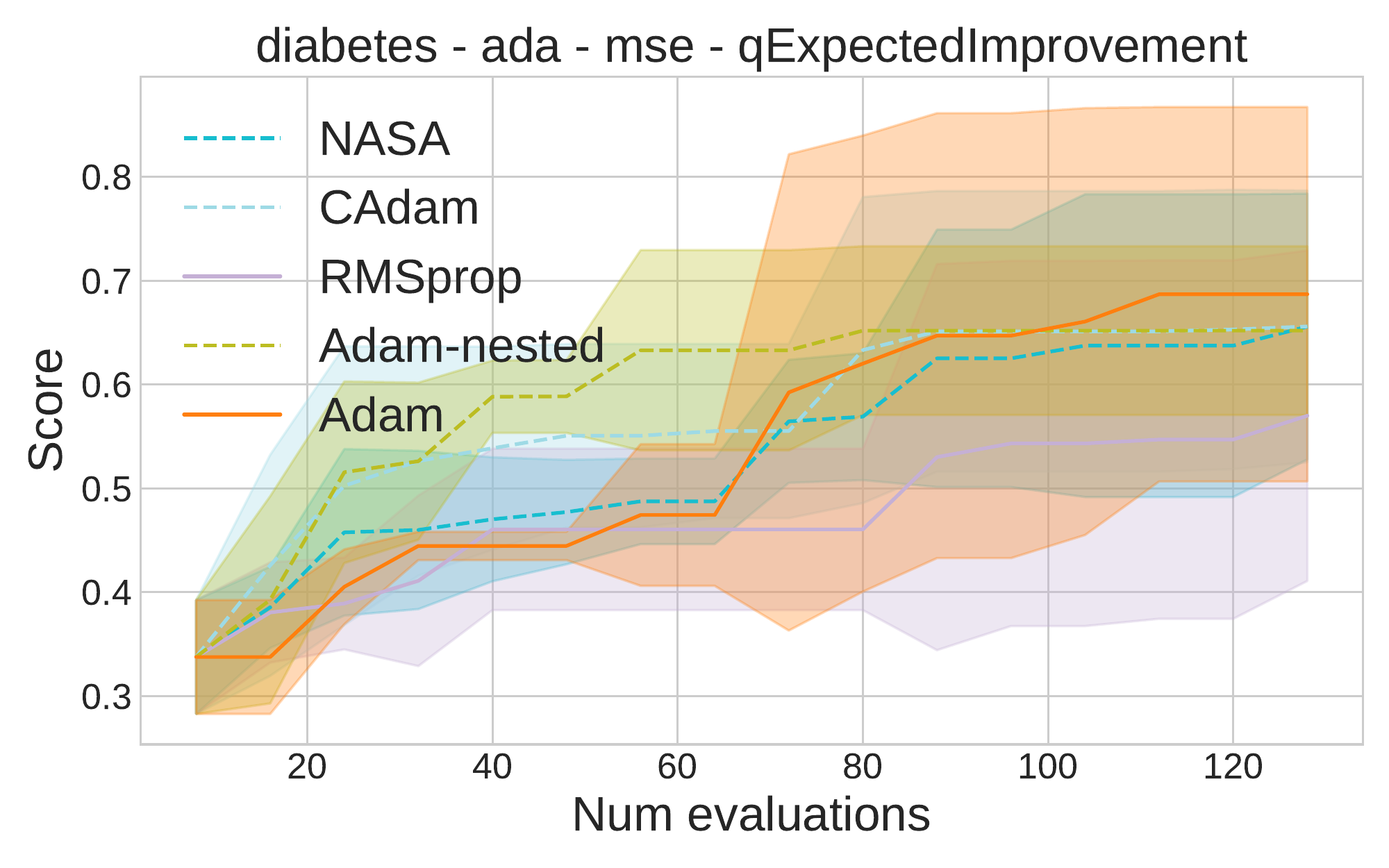}} & \hspace{-0.5cm}
    \subfloat{\includegraphics[width=0.19\columnwidth, trim={0 0.5cm 0 0.4cm}, clip]{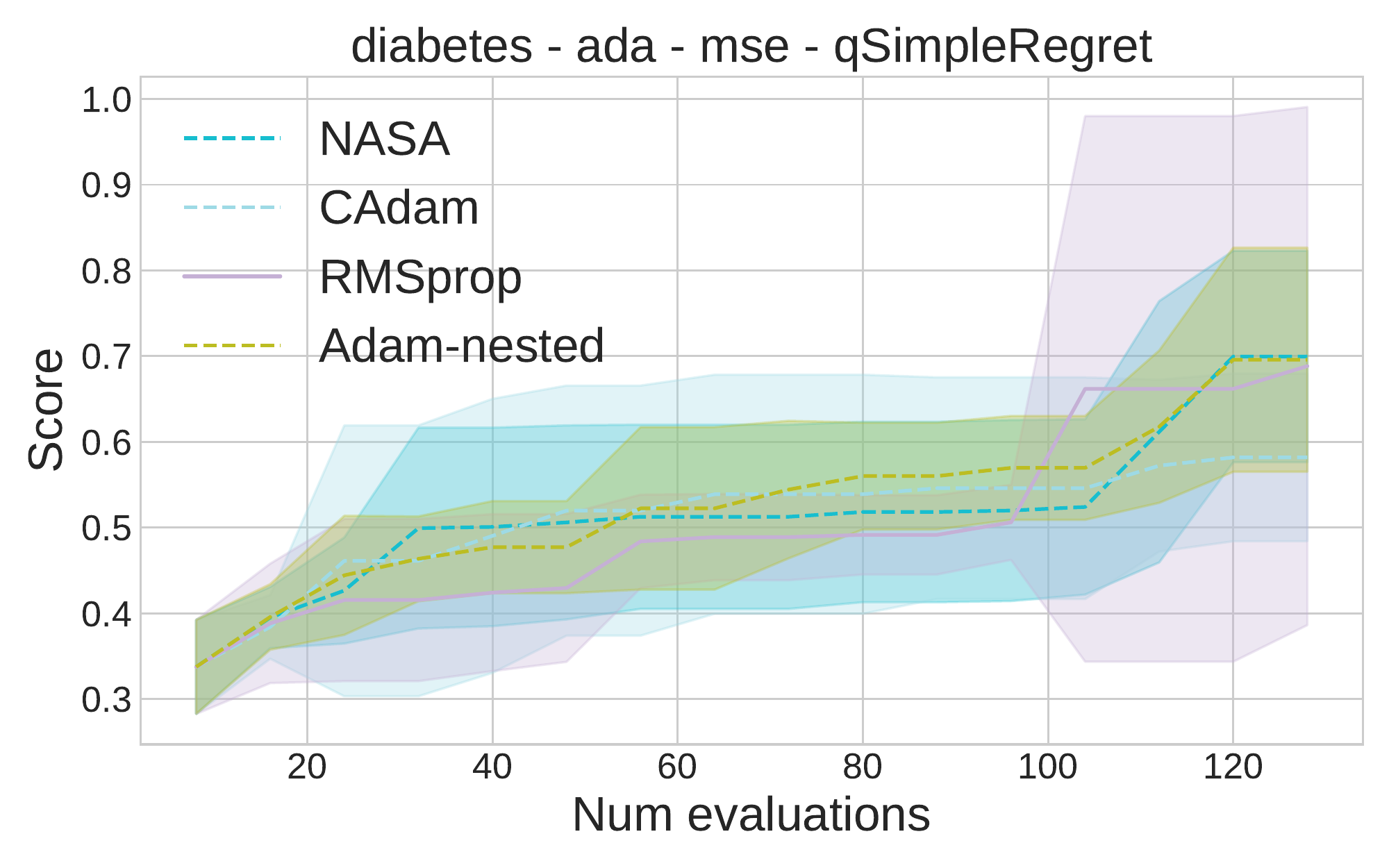}} \\  
    \subfloat{\includegraphics[width=0.19\columnwidth, trim={0 0.5cm 0 0.4cm}, clip]{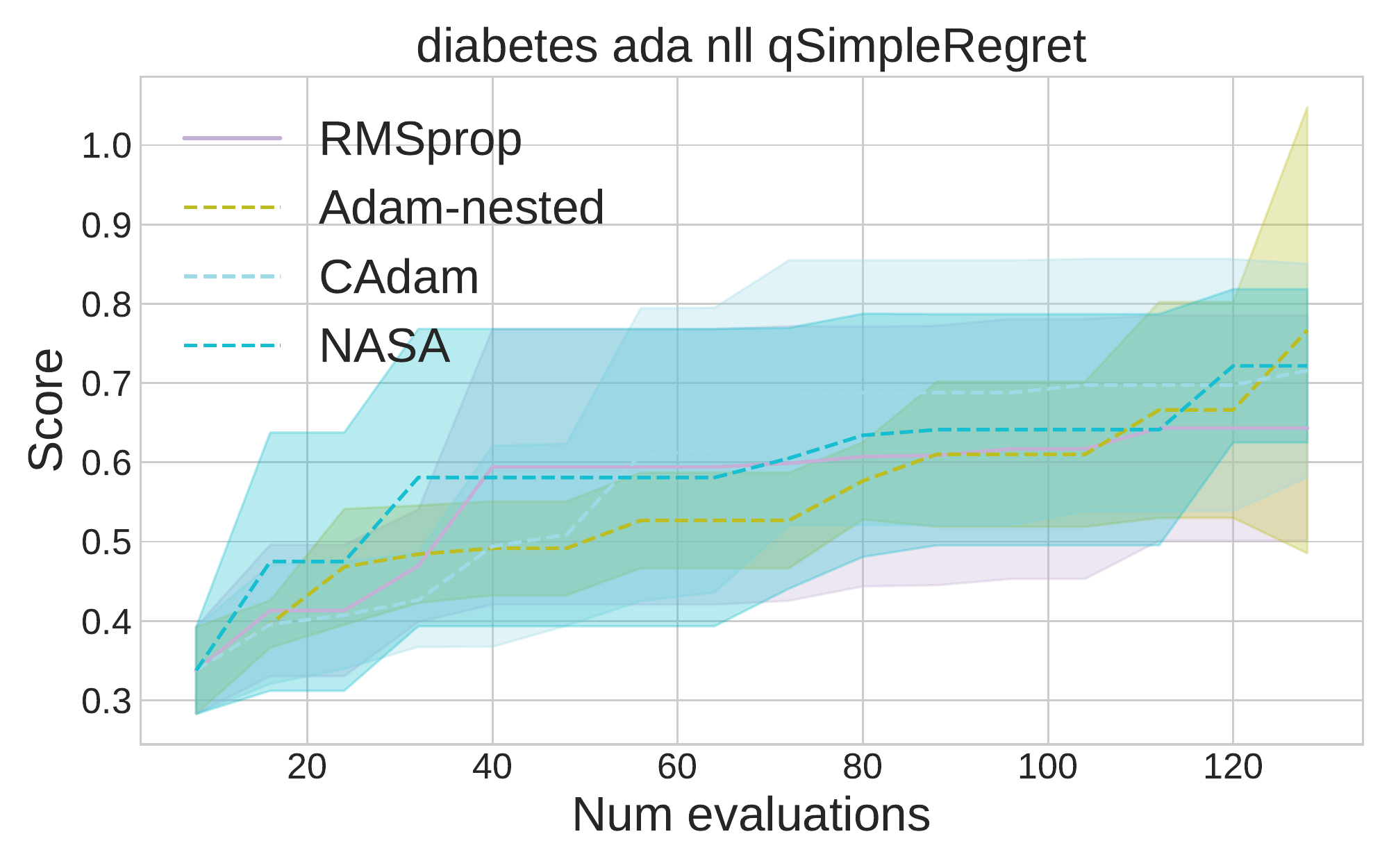}} &   \hspace{-0.5cm} 
    \subfloat{\includegraphics[width=0.19\columnwidth, trim={0 0.5cm 0 0.4cm}, clip]{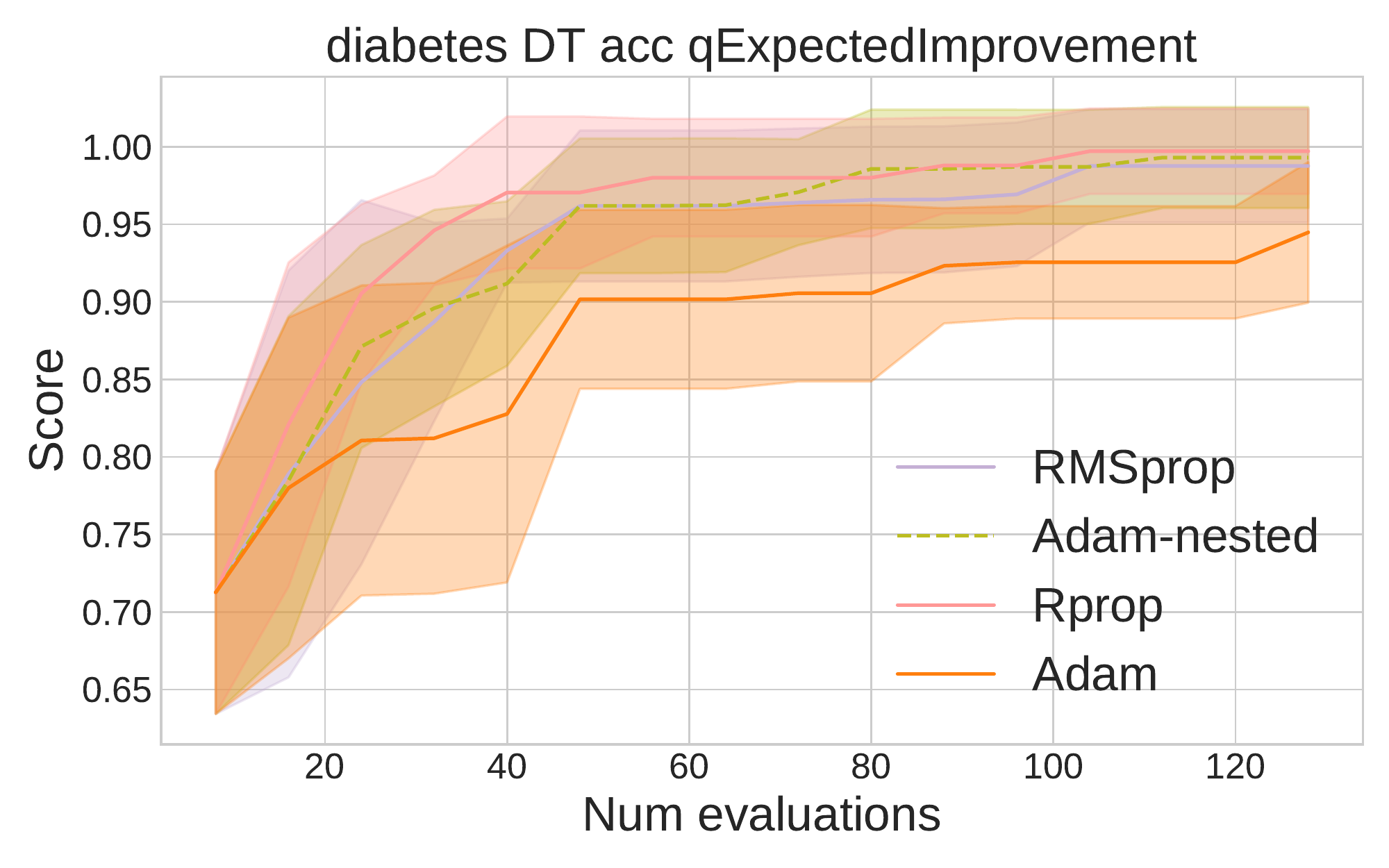}} & \hspace{-0.5cm}
    \subfloat{\includegraphics[width=0.19\columnwidth, trim={0 0.5cm 0 0.4cm}, clip]{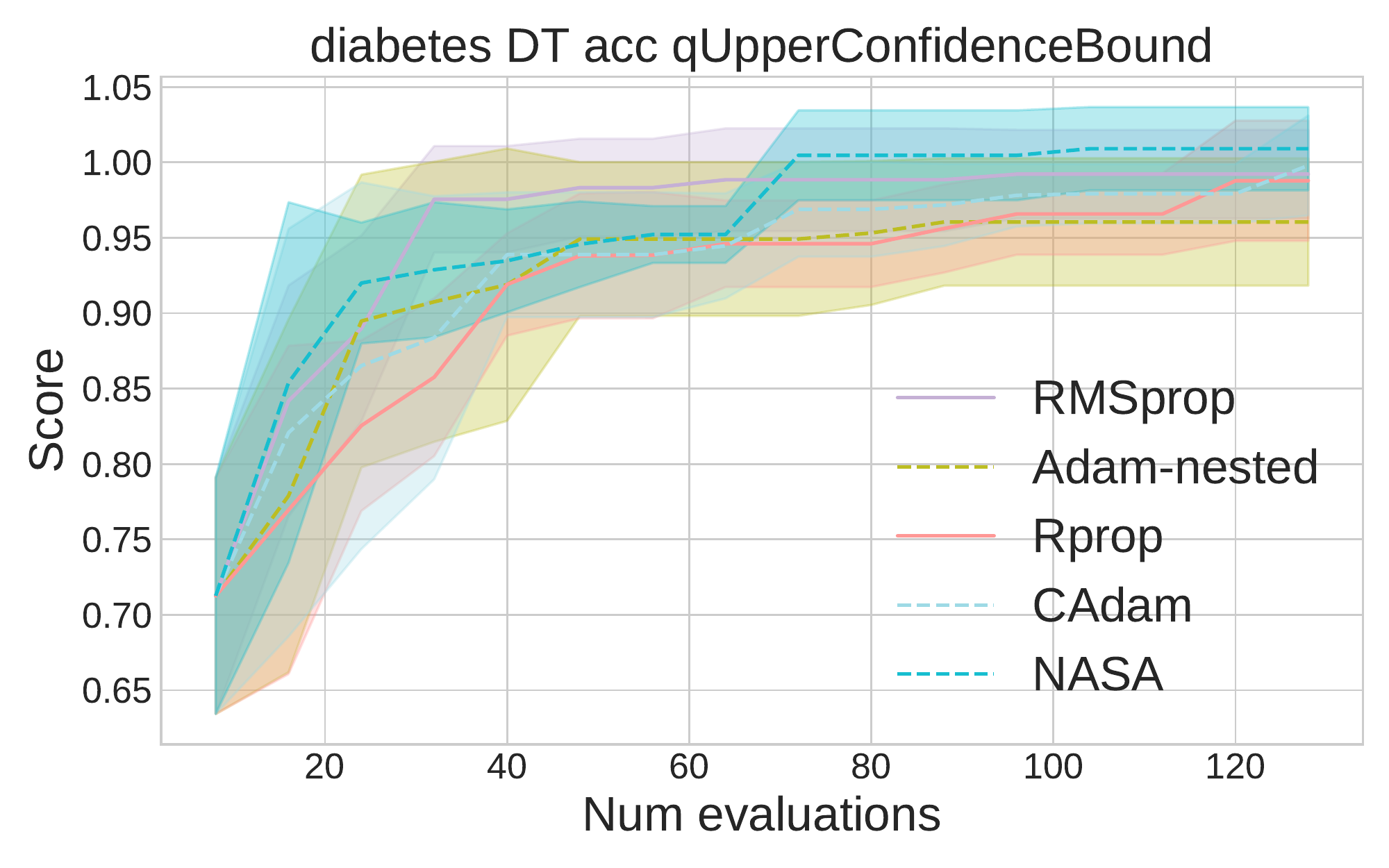}} &  \hspace{-0.5cm}
    \subfloat{\includegraphics[width=0.19\columnwidth, trim={0 0.5cm 0 0.4cm}, clip]{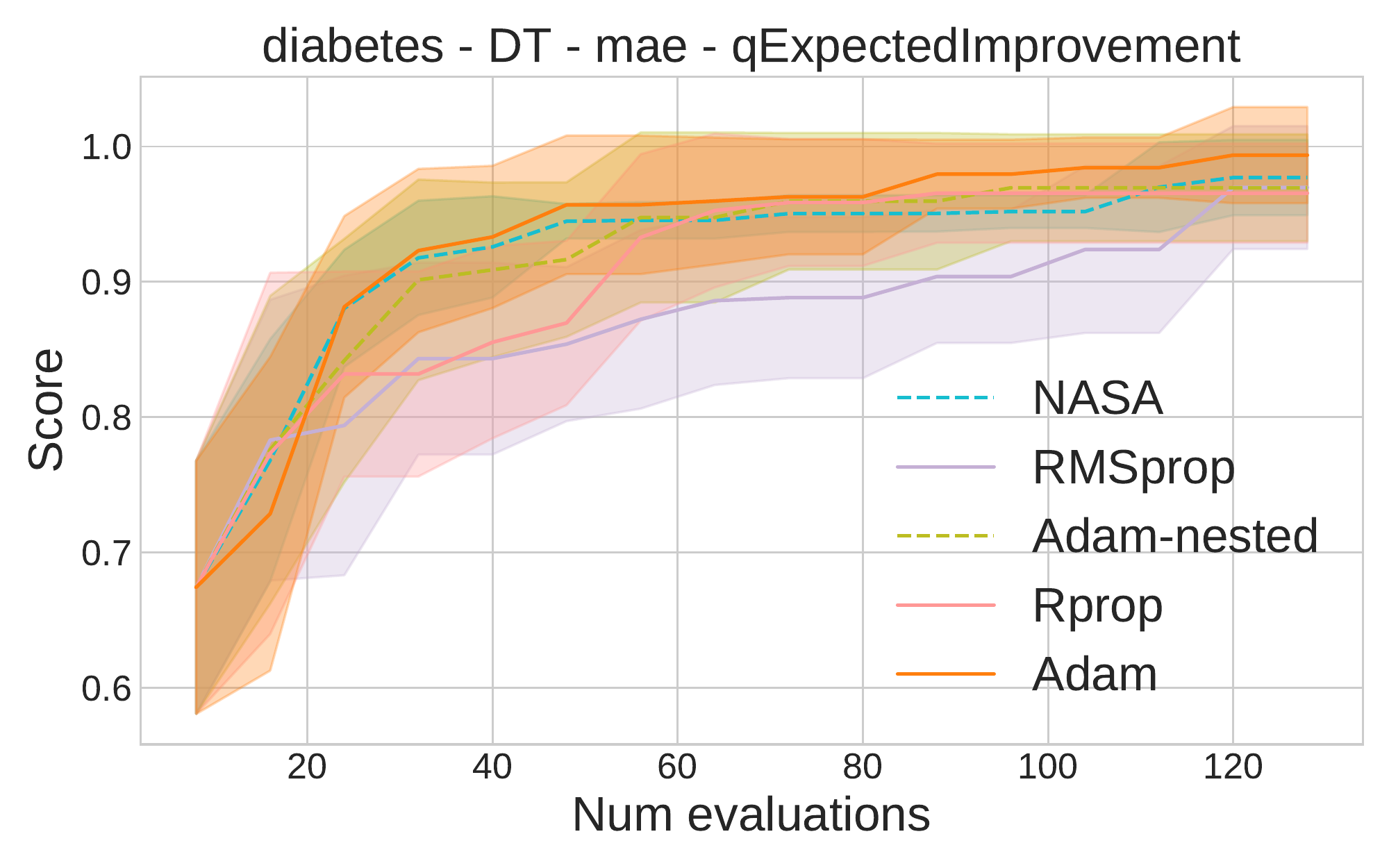}} & \hspace{-0.5cm}
    \subfloat{\includegraphics[width=0.19\columnwidth, trim={0 0.5cm 0 0.4cm}, clip]{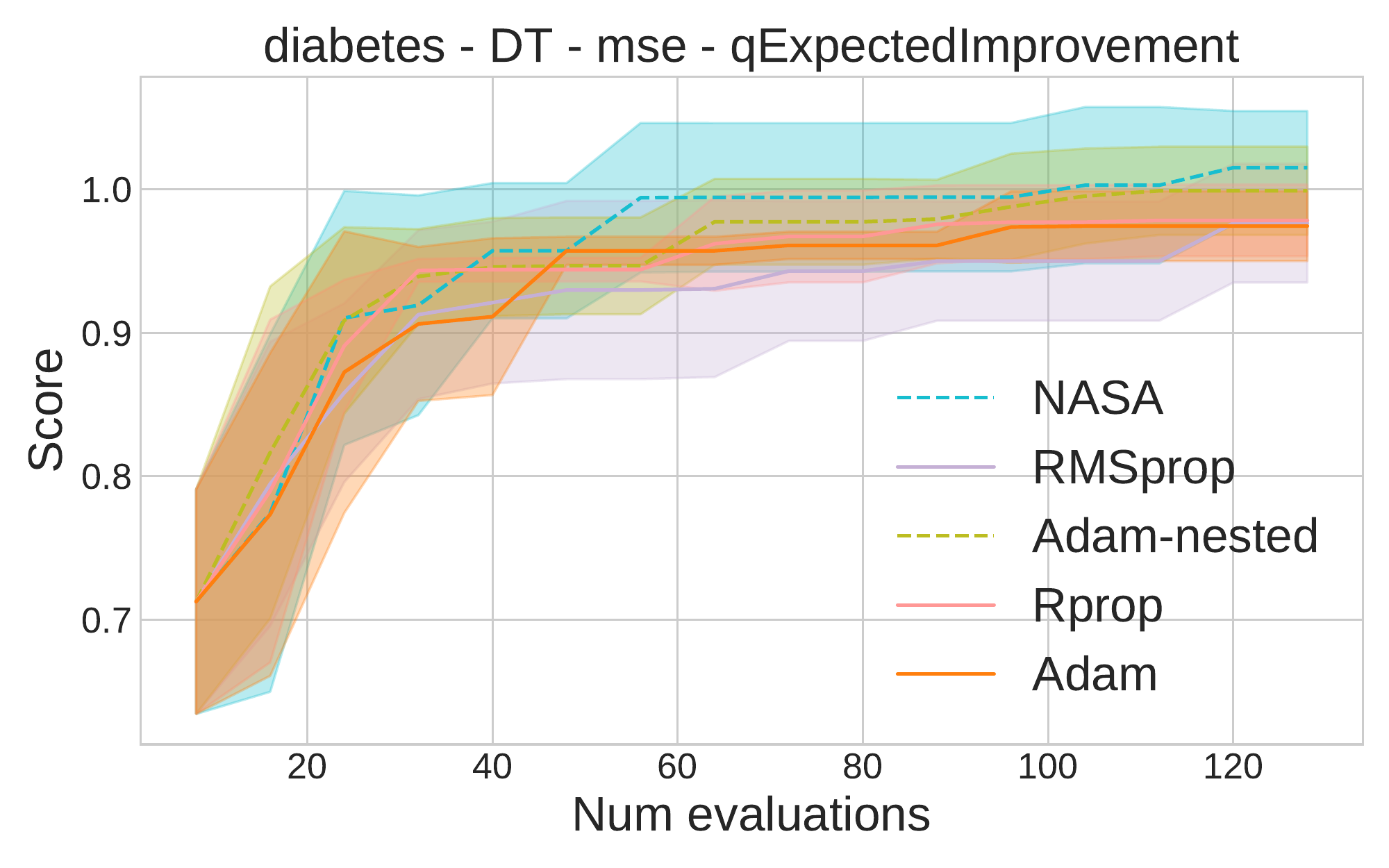}} \\  
    \subfloat{\includegraphics[width=0.19\columnwidth, trim={0 0.5cm 0 0.4cm}, clip]{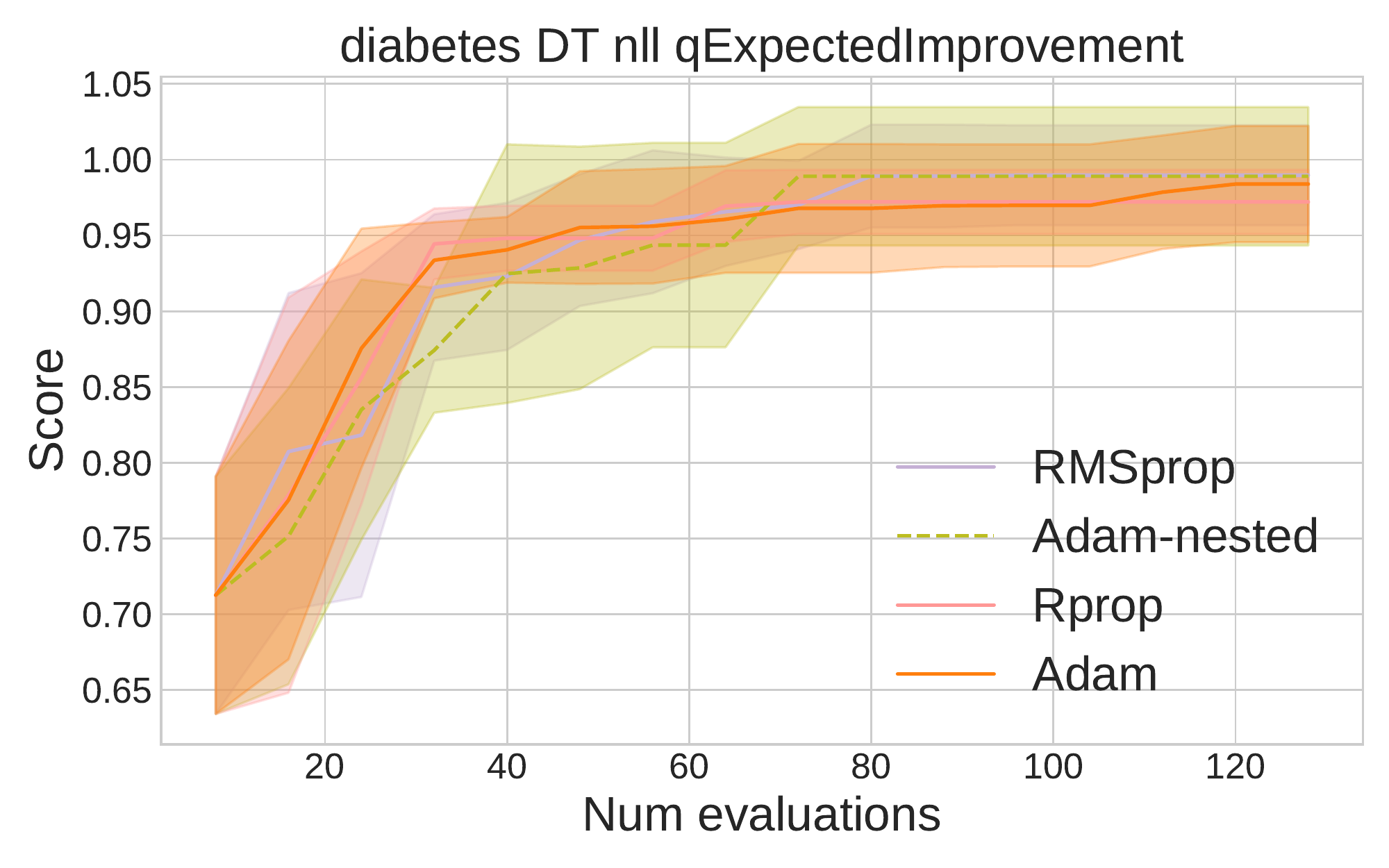}} &   \hspace{-0.5cm} 
    \subfloat{\includegraphics[width=0.19\columnwidth, trim={0 0.5cm 0 0.4cm}, clip]{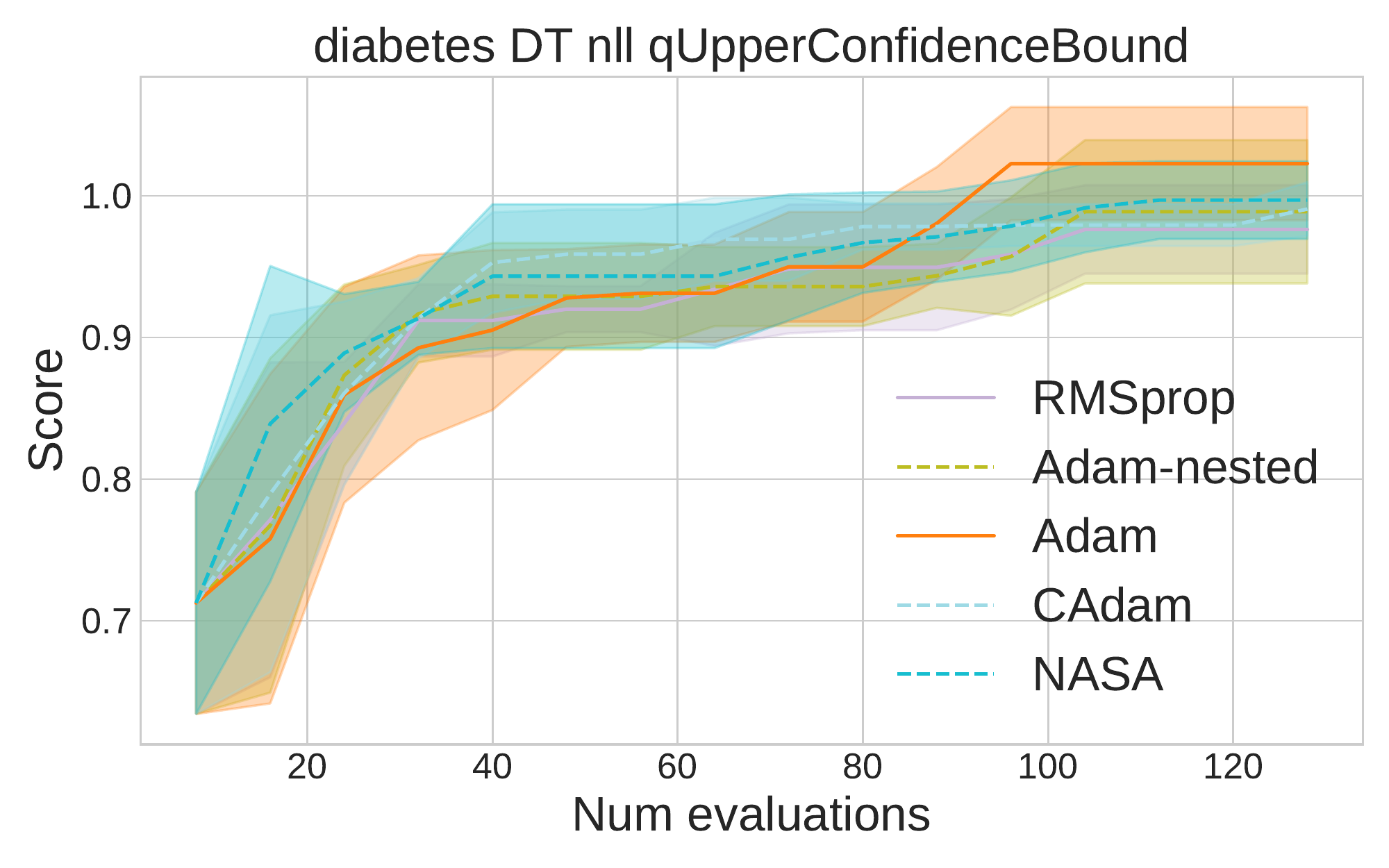}} & \hspace{-0.5cm}
    \subfloat{\includegraphics[width=0.19\columnwidth, trim={0 0.5cm 0 0.4cm}, clip]{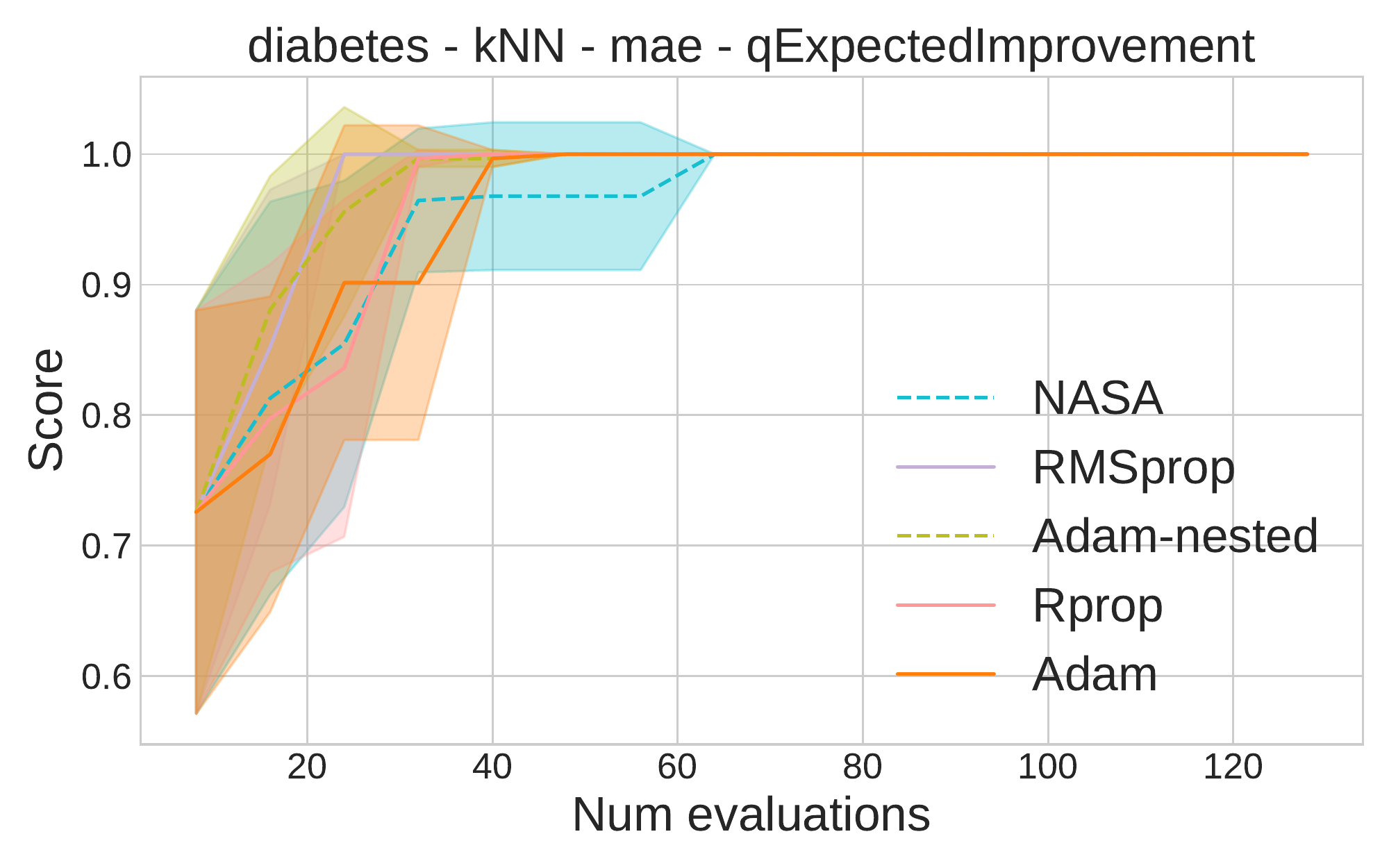}} &  \hspace{-0.5cm}
    \subfloat{\includegraphics[width=0.19\columnwidth, trim={0 0.5cm 0 0.4cm}, clip]{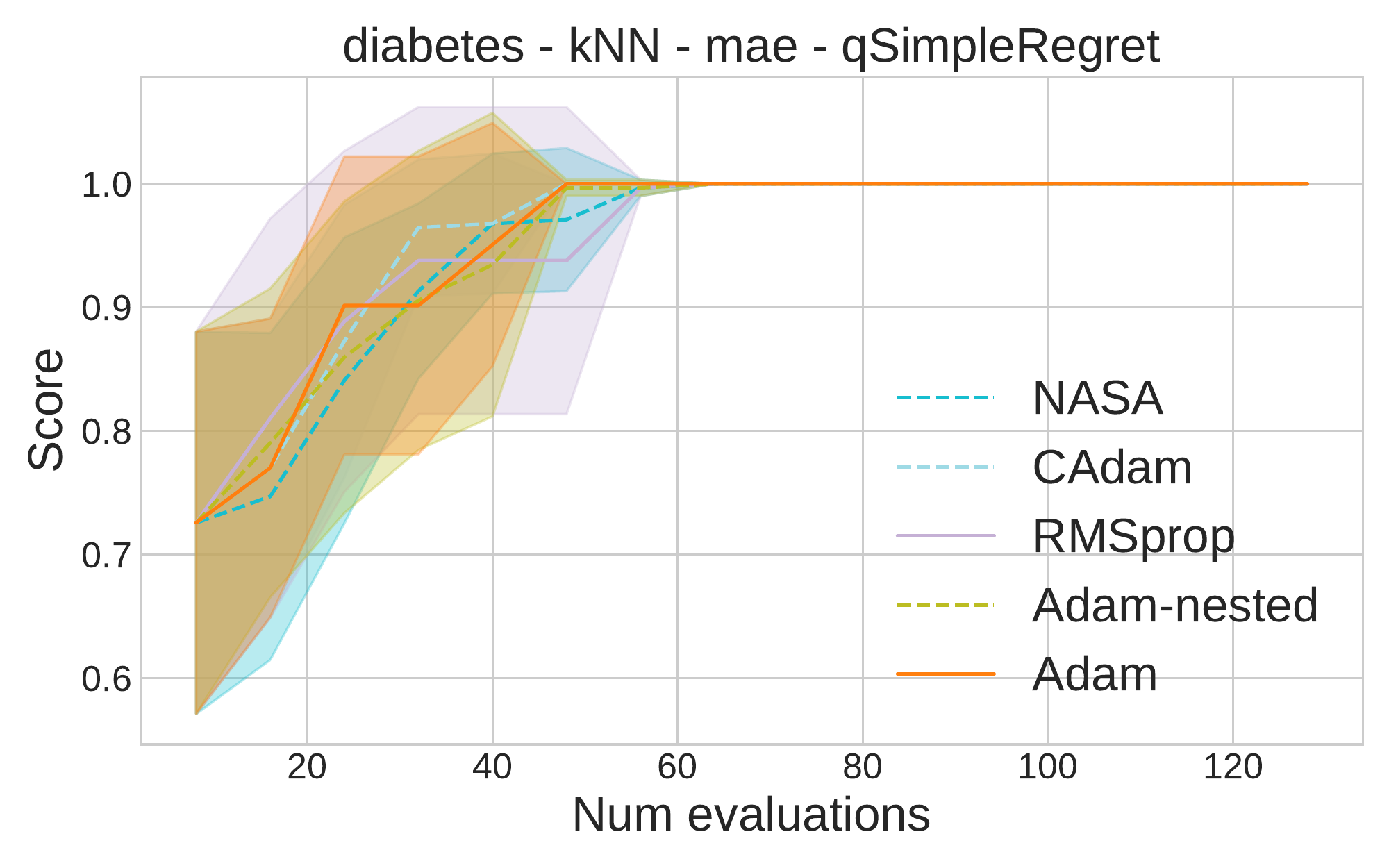}} & \hspace{-0.5cm}
    \subfloat{\includegraphics[width=0.19\columnwidth, trim={0 0.5cm 0 0.4cm}, clip]{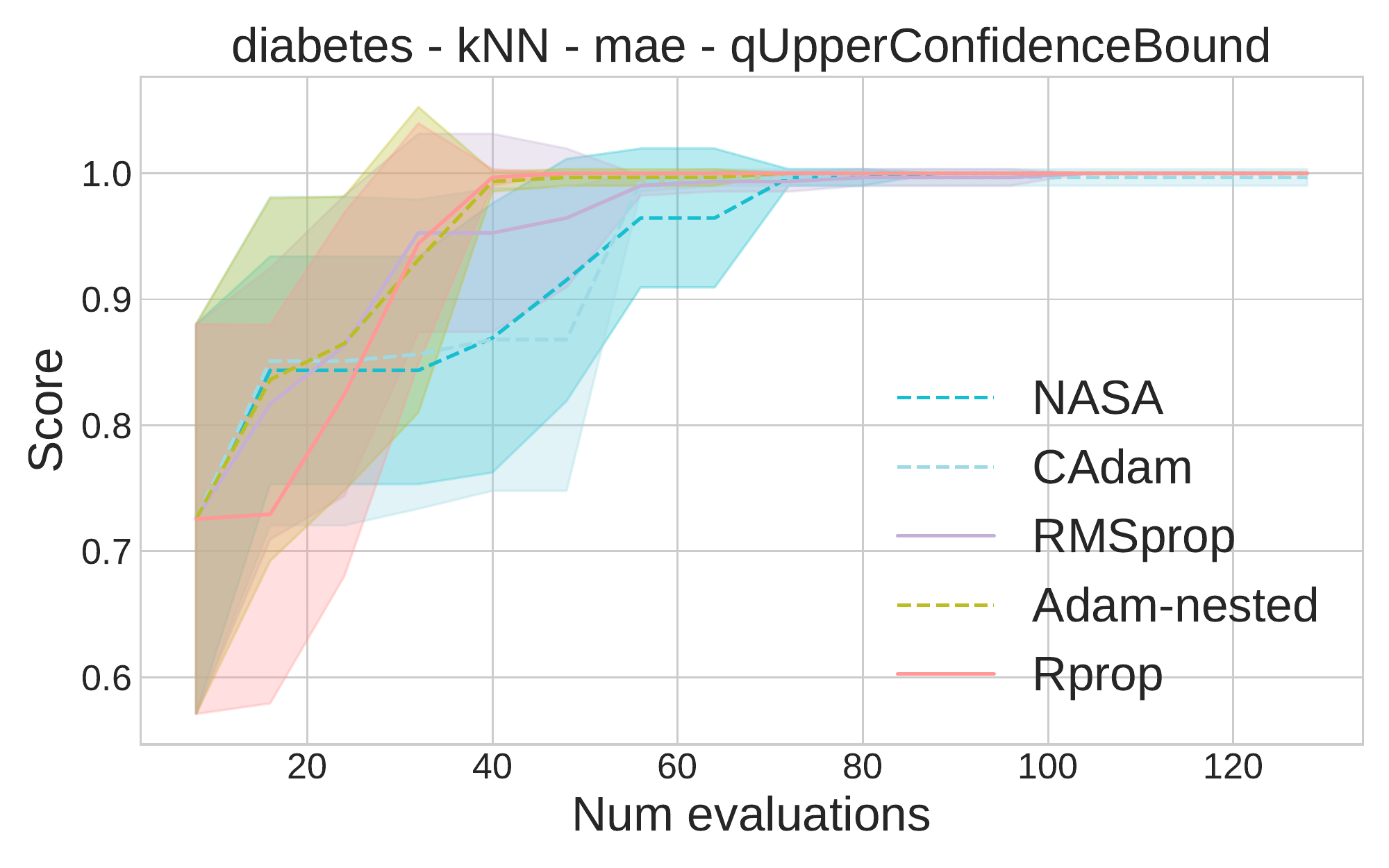}} \\  
    \subfloat{\includegraphics[width=0.19\columnwidth, trim={0 0.5cm 0 0.4cm}, clip]{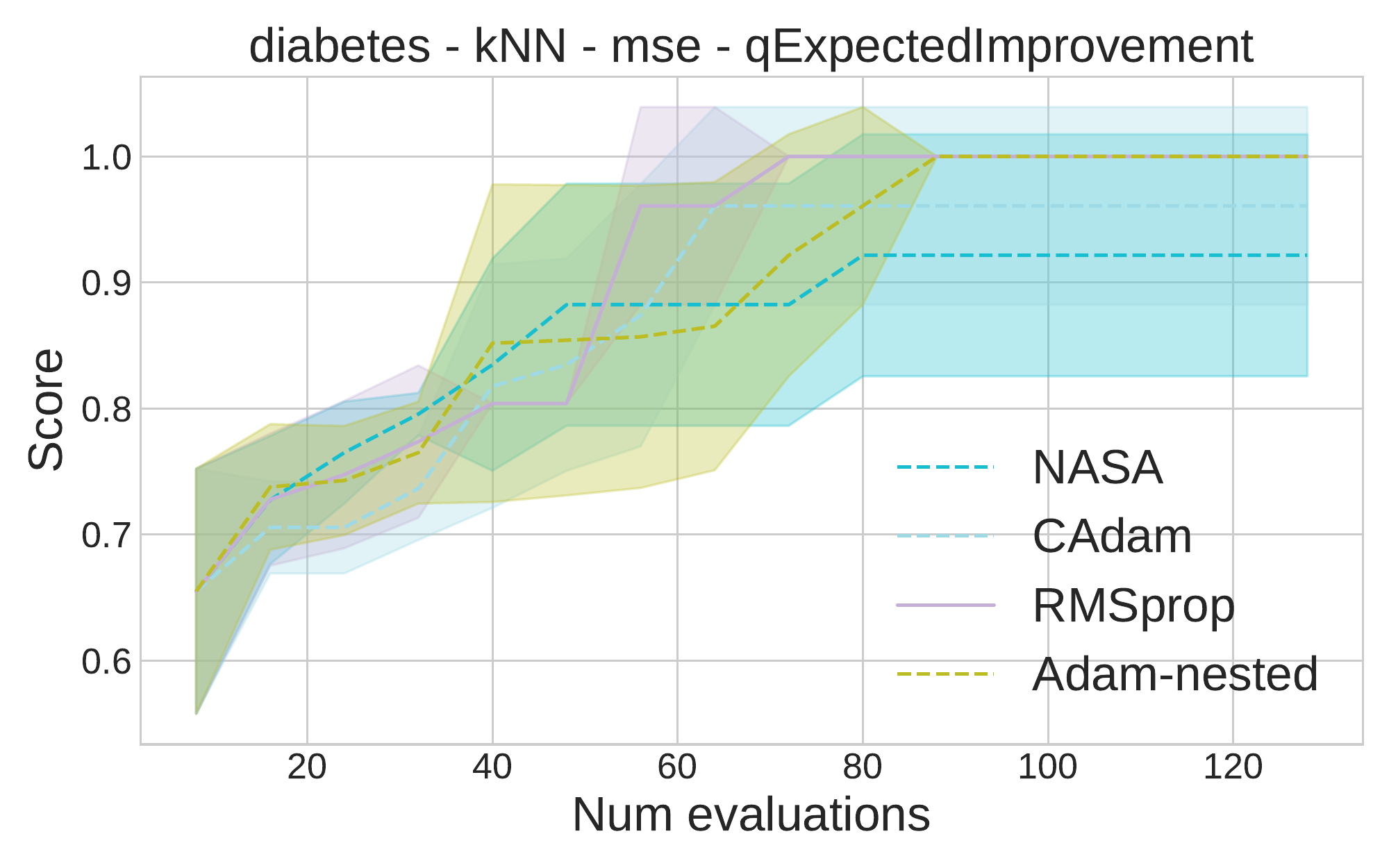}} &   \hspace{-0.5cm} 
    \subfloat{\includegraphics[width=0.19\columnwidth, trim={0 0.5cm 0 0.4cm}, clip]{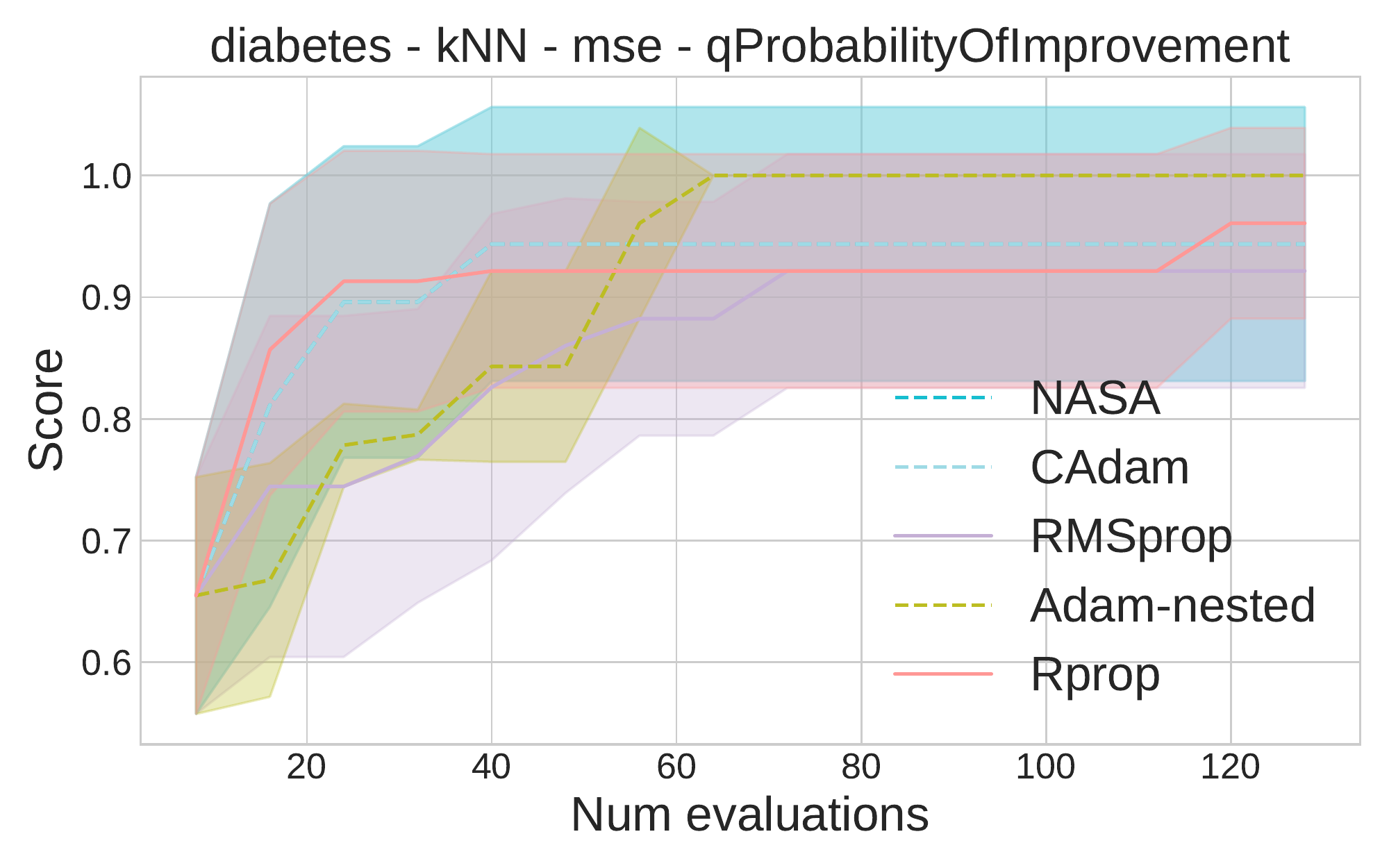}} & \hspace{-0.5cm}
    \subfloat{\includegraphics[width=0.19\columnwidth, trim={0 0.5cm 0 0.4cm}, clip]{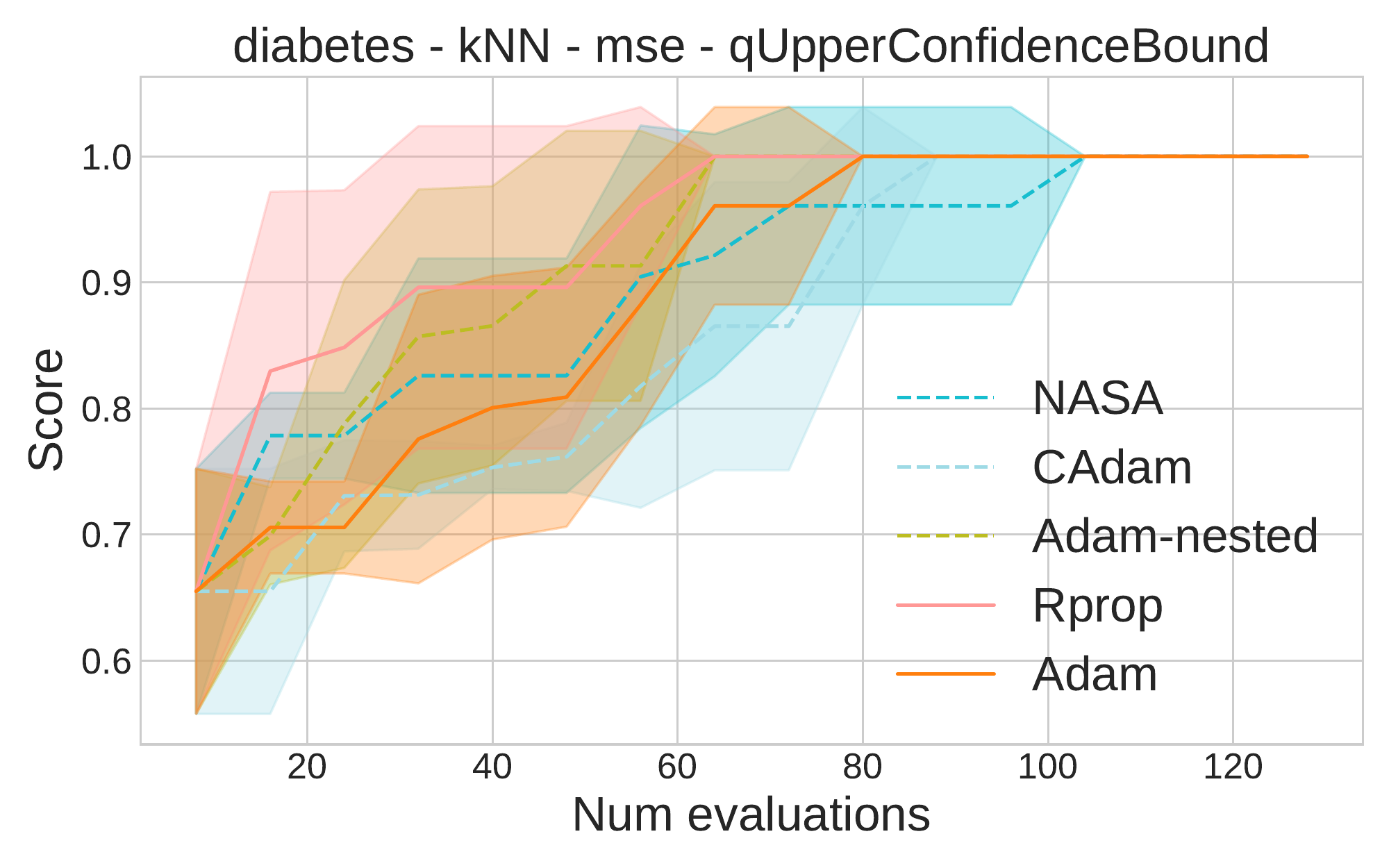}} &  \hspace{-0.5cm}
    \subfloat{\includegraphics[width=0.19\columnwidth, trim={0 0.5cm 0 0.4cm}, clip]{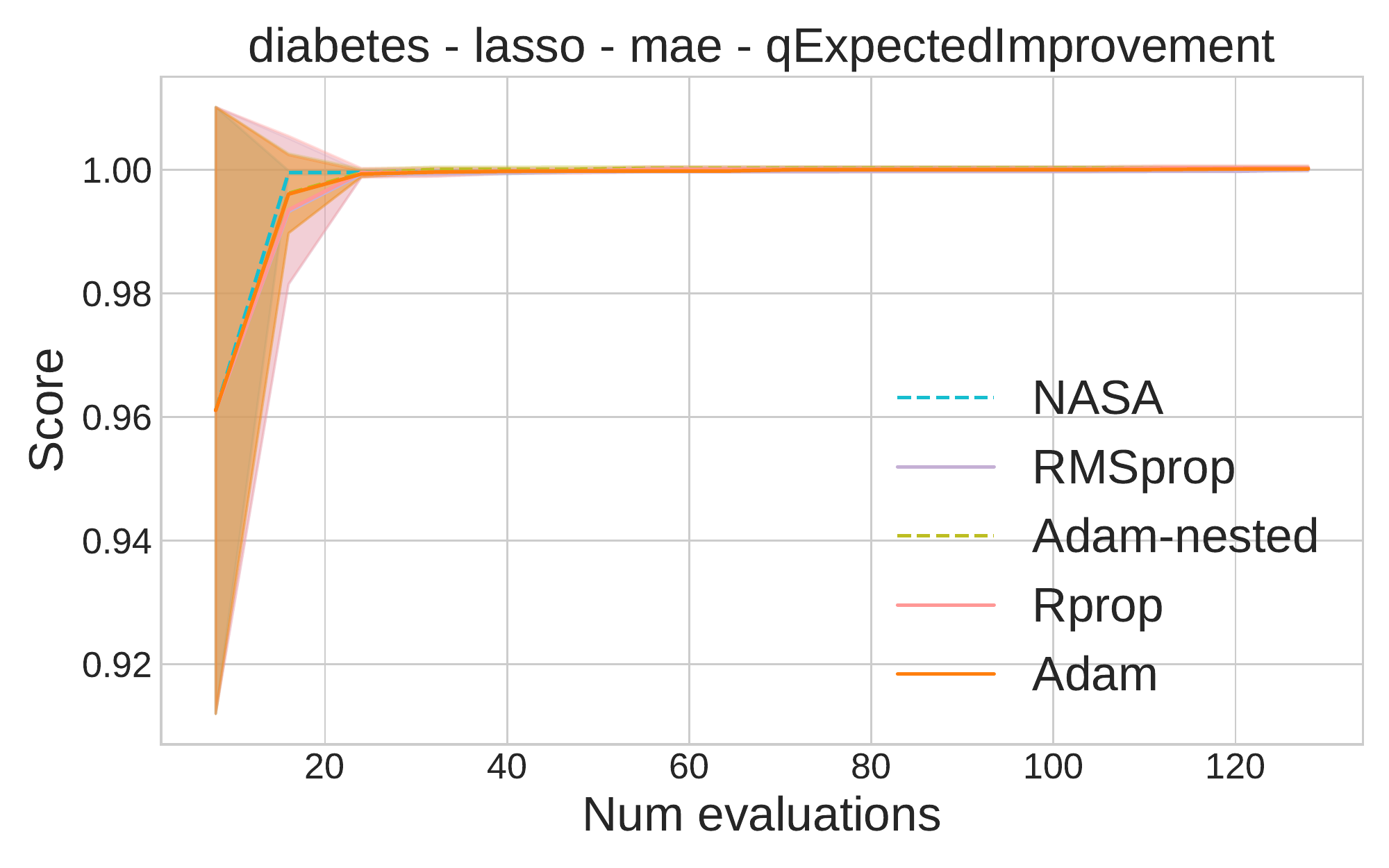}} & \hspace{-0.5cm}
    \subfloat{\includegraphics[width=0.19\columnwidth, trim={0 0.5cm 0 0.4cm}, clip]{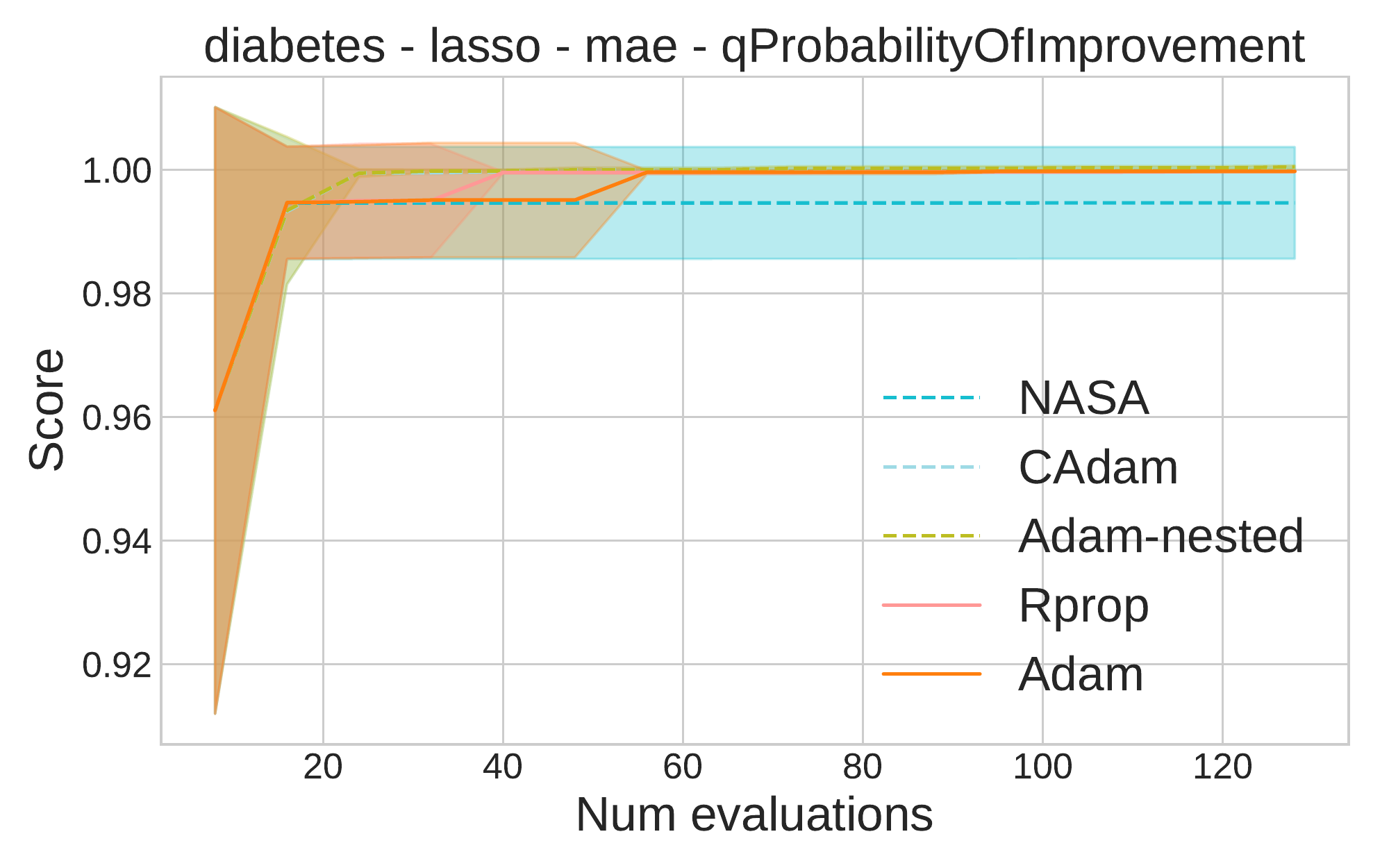}} \\  
    \subfloat{\includegraphics[width=0.19\columnwidth, trim={0 0.5cm 0 0.4cm}, clip]{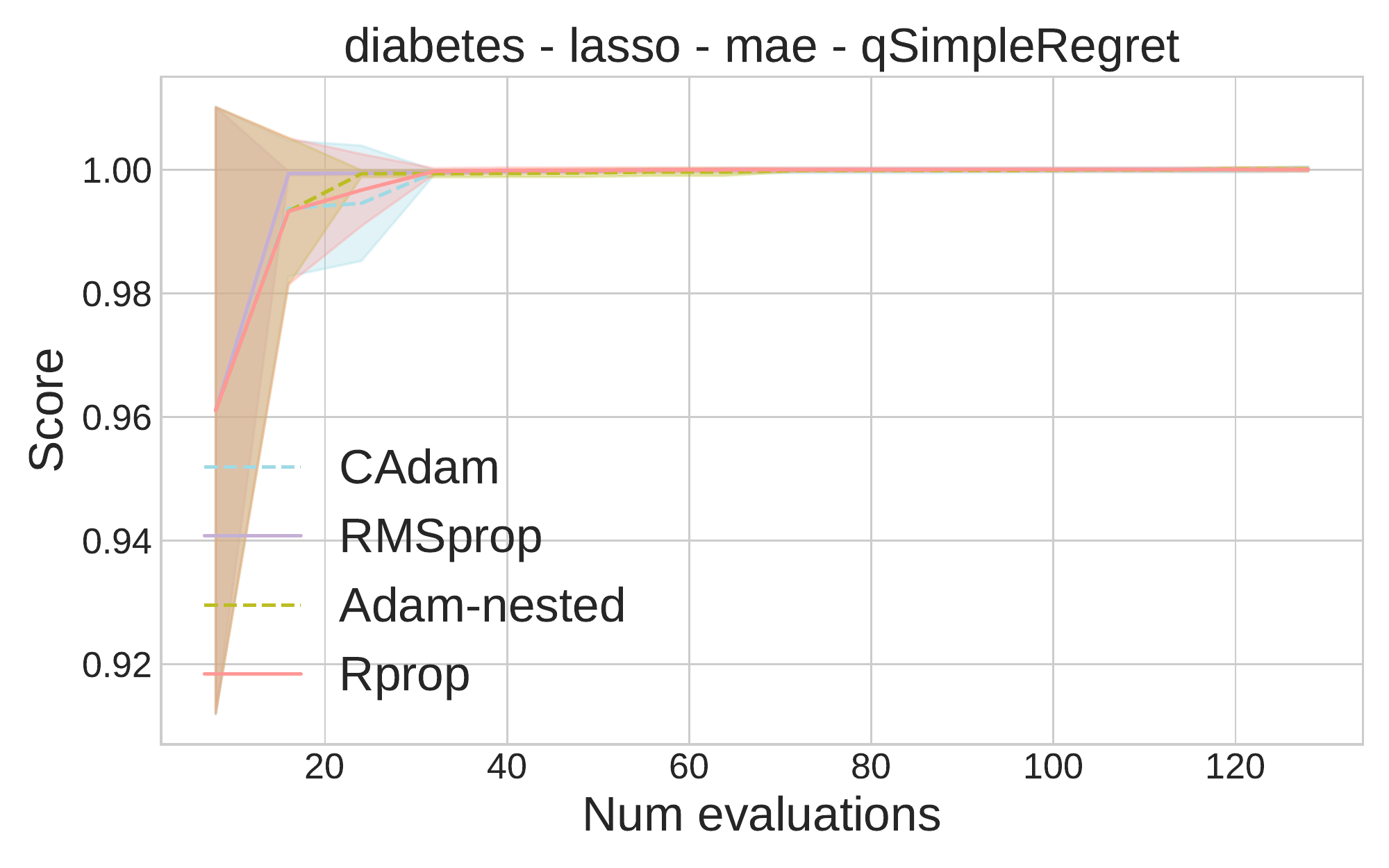}} &   \hspace{-0.5cm} 
    \subfloat{\includegraphics[width=0.19\columnwidth, trim={0 0.5cm 0 0.4cm}, clip]{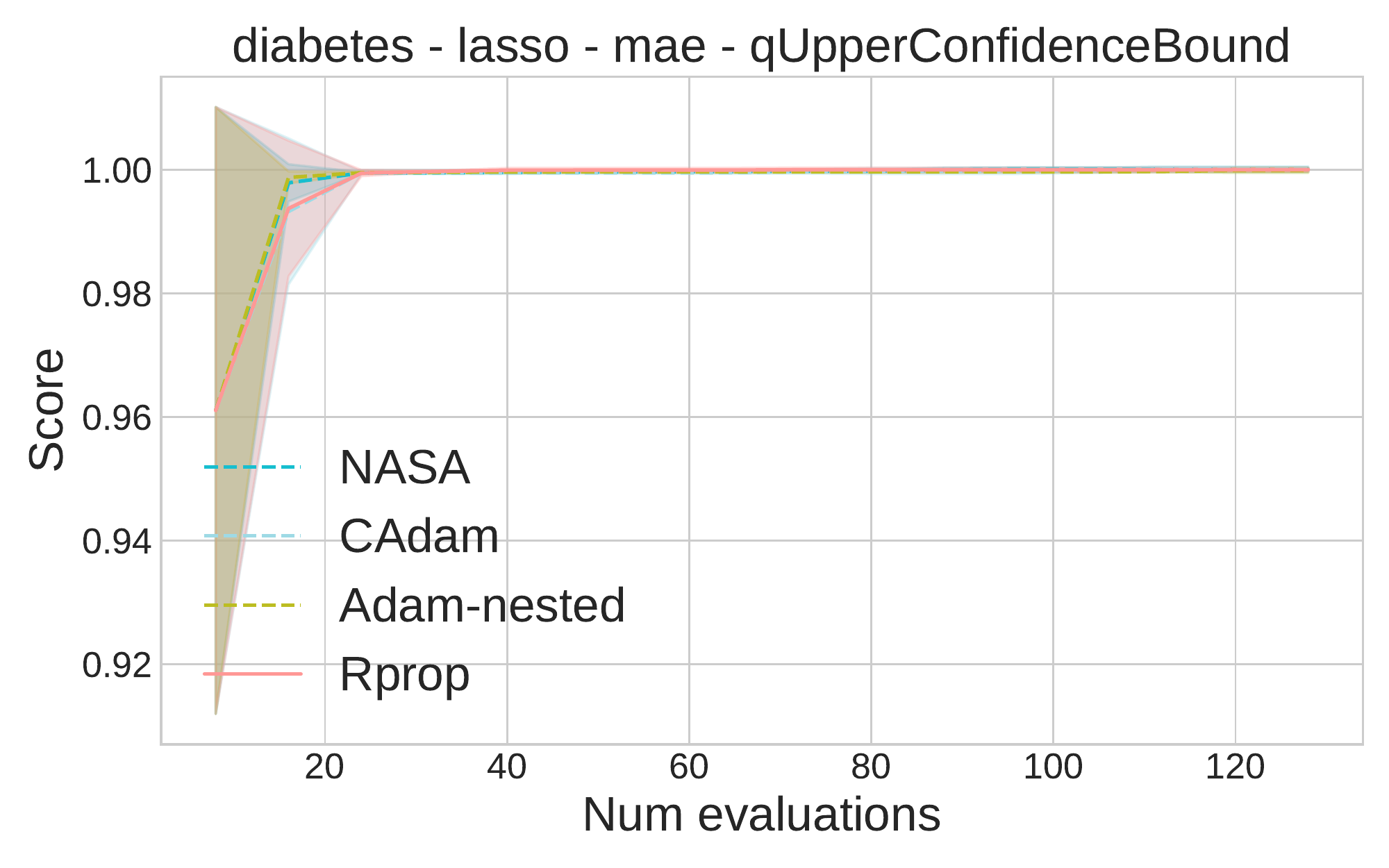}} & \hspace{-0.5cm}
    \subfloat{\includegraphics[width=0.19\columnwidth, trim={0 0.5cm 0 0.4cm}, clip]{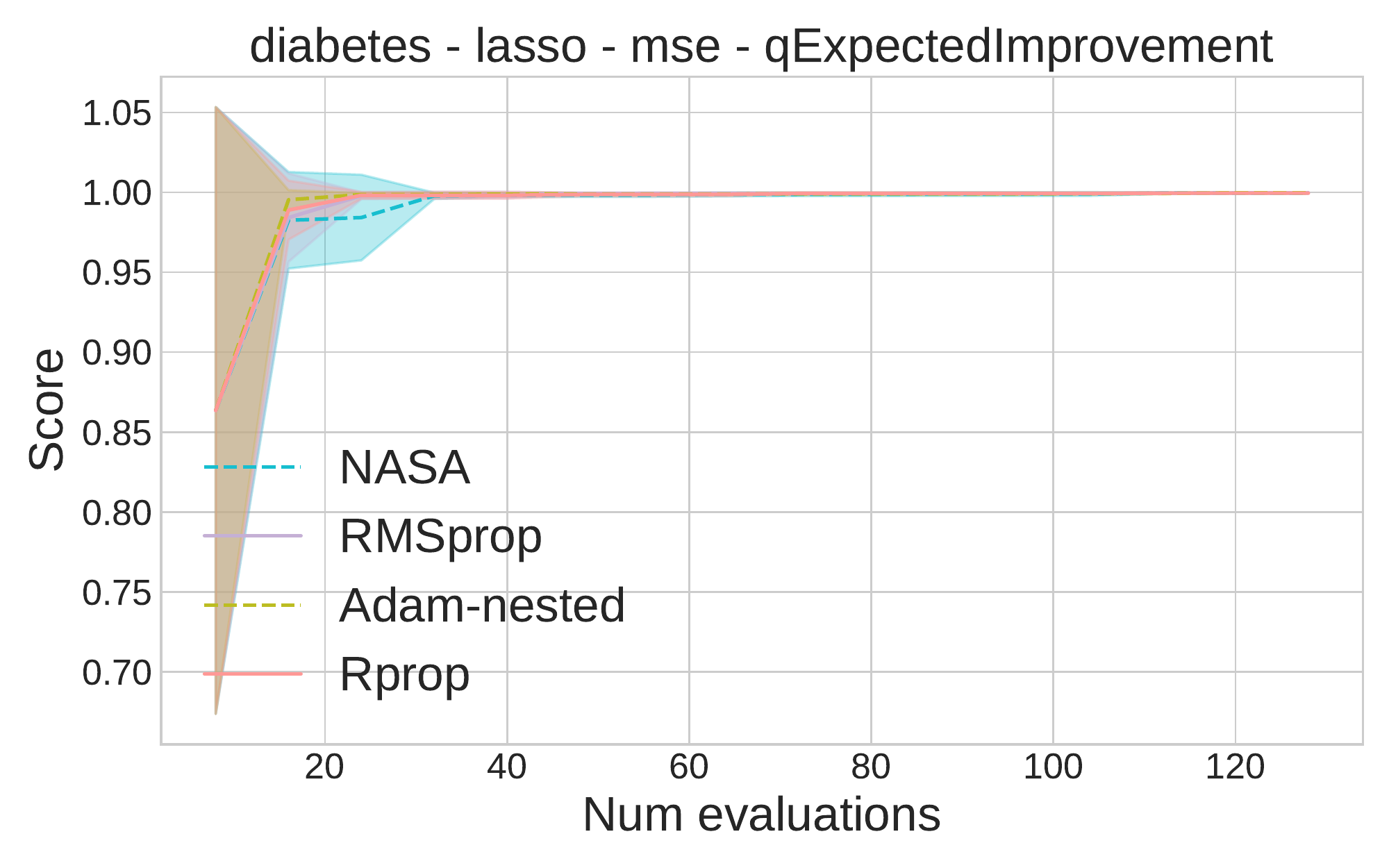}} &  \hspace{-0.5cm}
    \subfloat{\includegraphics[width=0.19\columnwidth, trim={0 0.5cm 0 0.4cm}, clip]{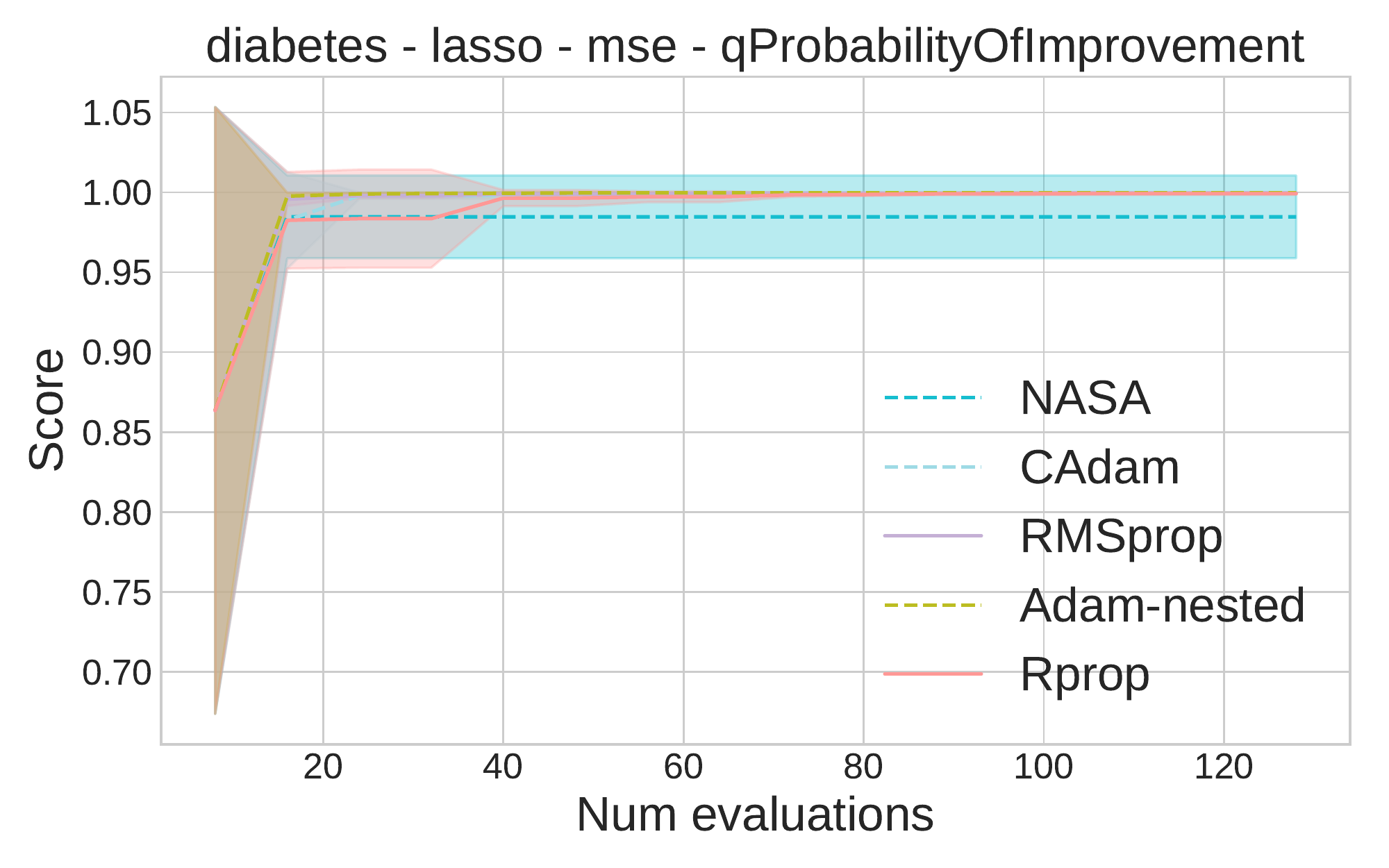}} & \hspace{-0.5cm}
    \subfloat{\includegraphics[width=0.19\columnwidth, trim={0 0.5cm 0 0.4cm}, clip]{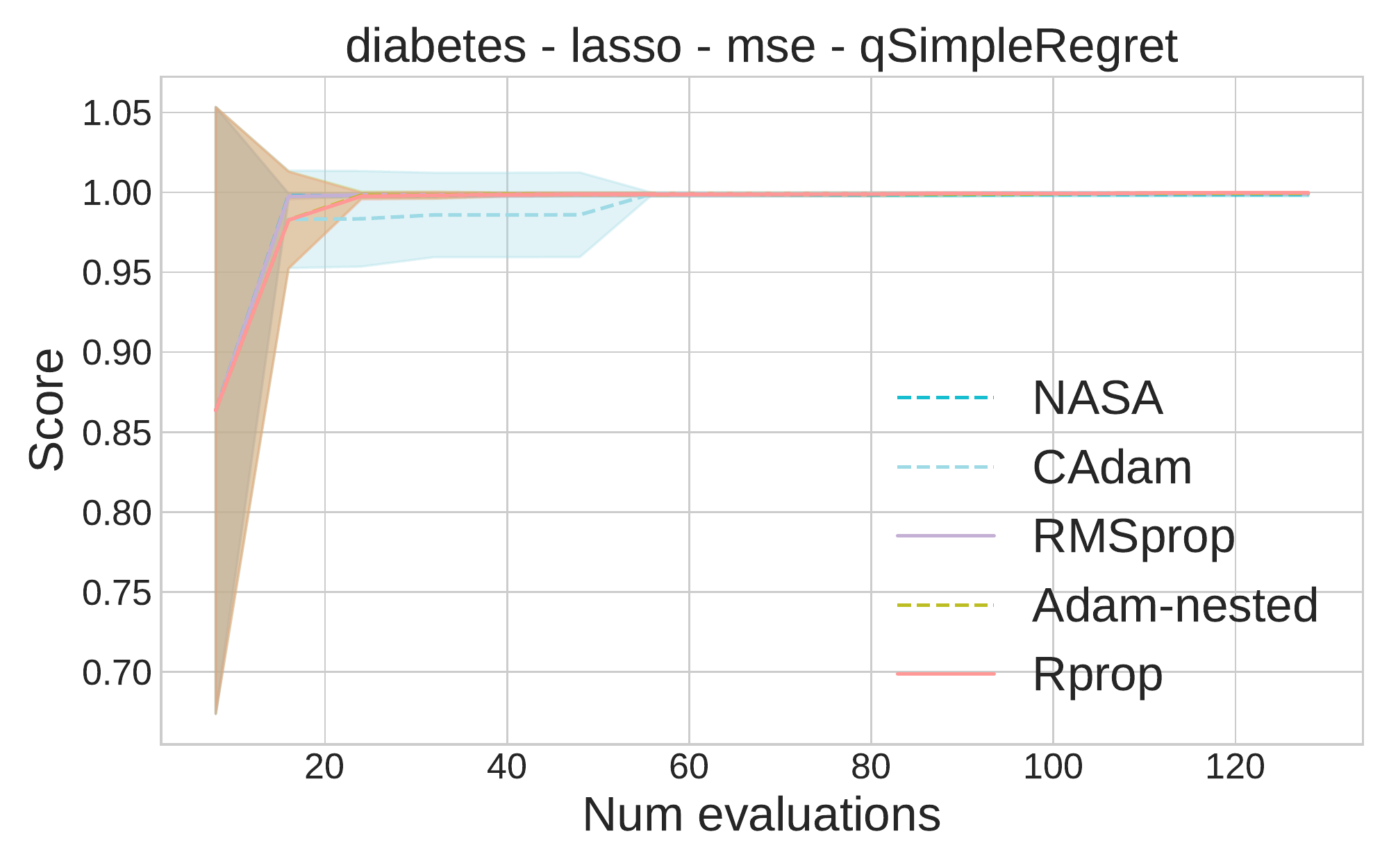}} \\  
    \subfloat{\includegraphics[width=0.19\columnwidth, trim={0 0.5cm 0 0.4cm}, clip]{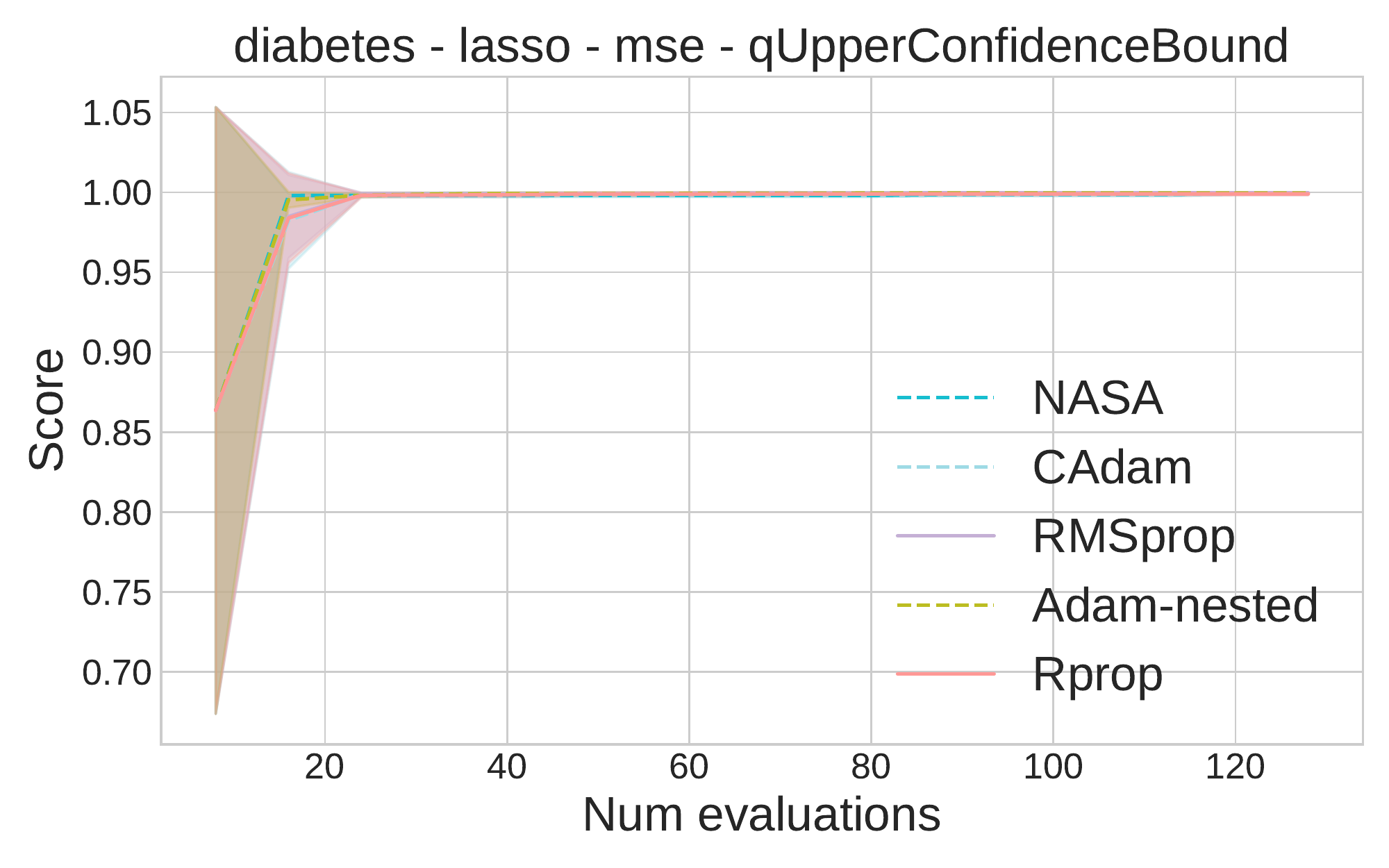}} &   \hspace{-0.5cm} 
    \subfloat{\includegraphics[width=0.19\columnwidth, trim={0 0.5cm 0 0.4cm}, clip]{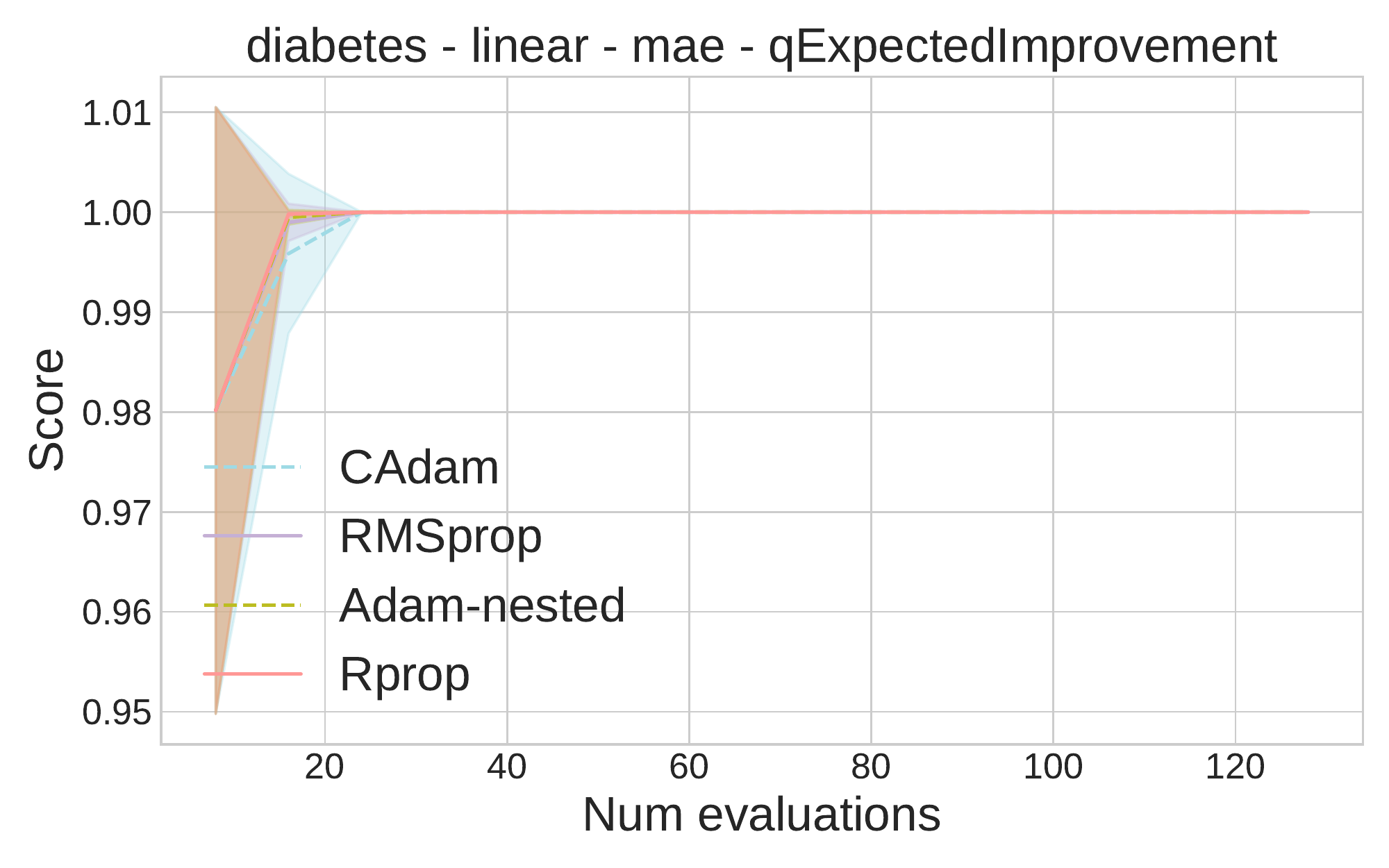}} & \hspace{-0.5cm}
    \subfloat{\includegraphics[width=0.19\columnwidth, trim={0 0.5cm 0 0.4cm}, clip]{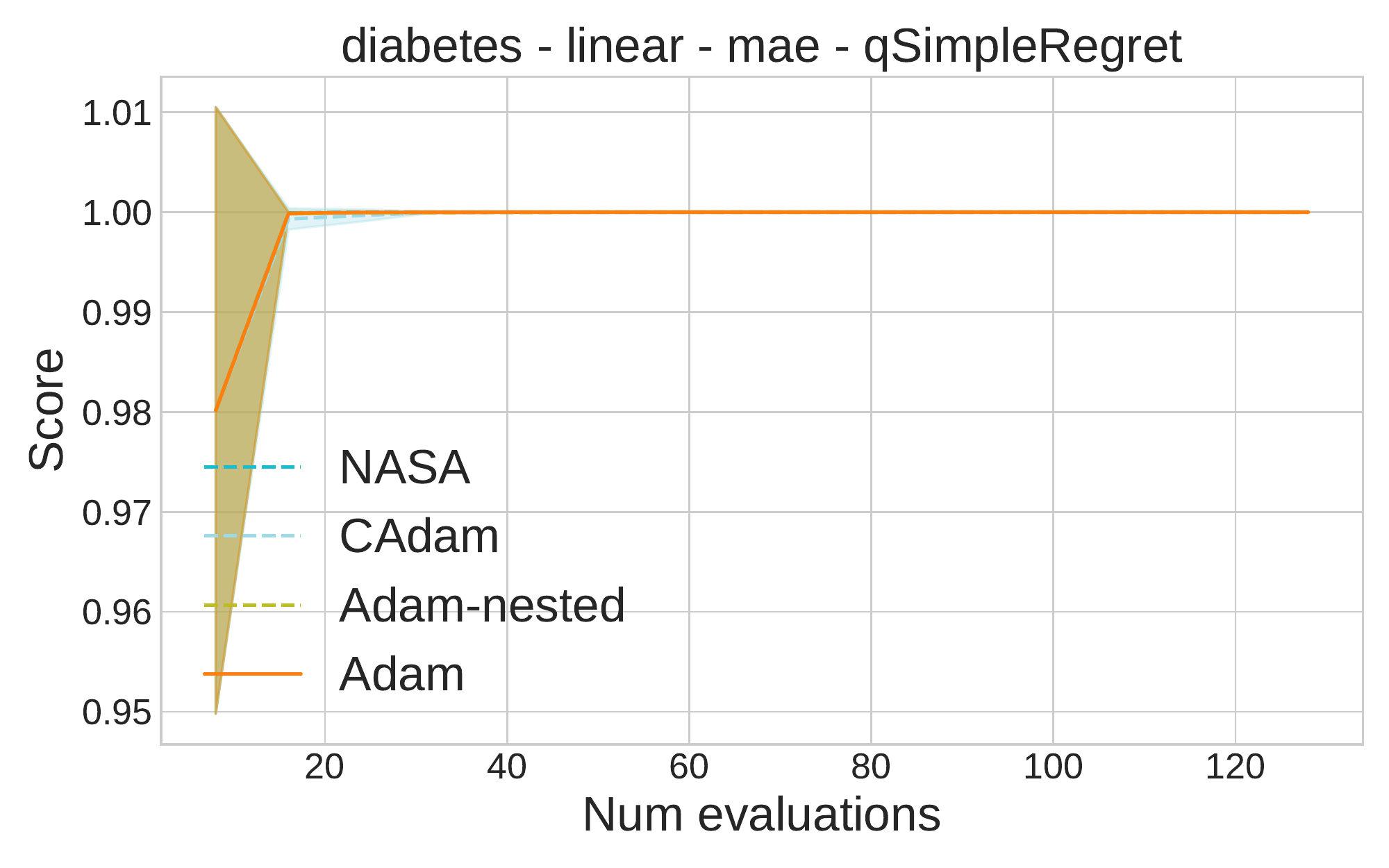}} &  \hspace{-0.5cm}
    \subfloat{\includegraphics[width=0.19\columnwidth, trim={0 0.5cm 0 0.4cm}, clip]{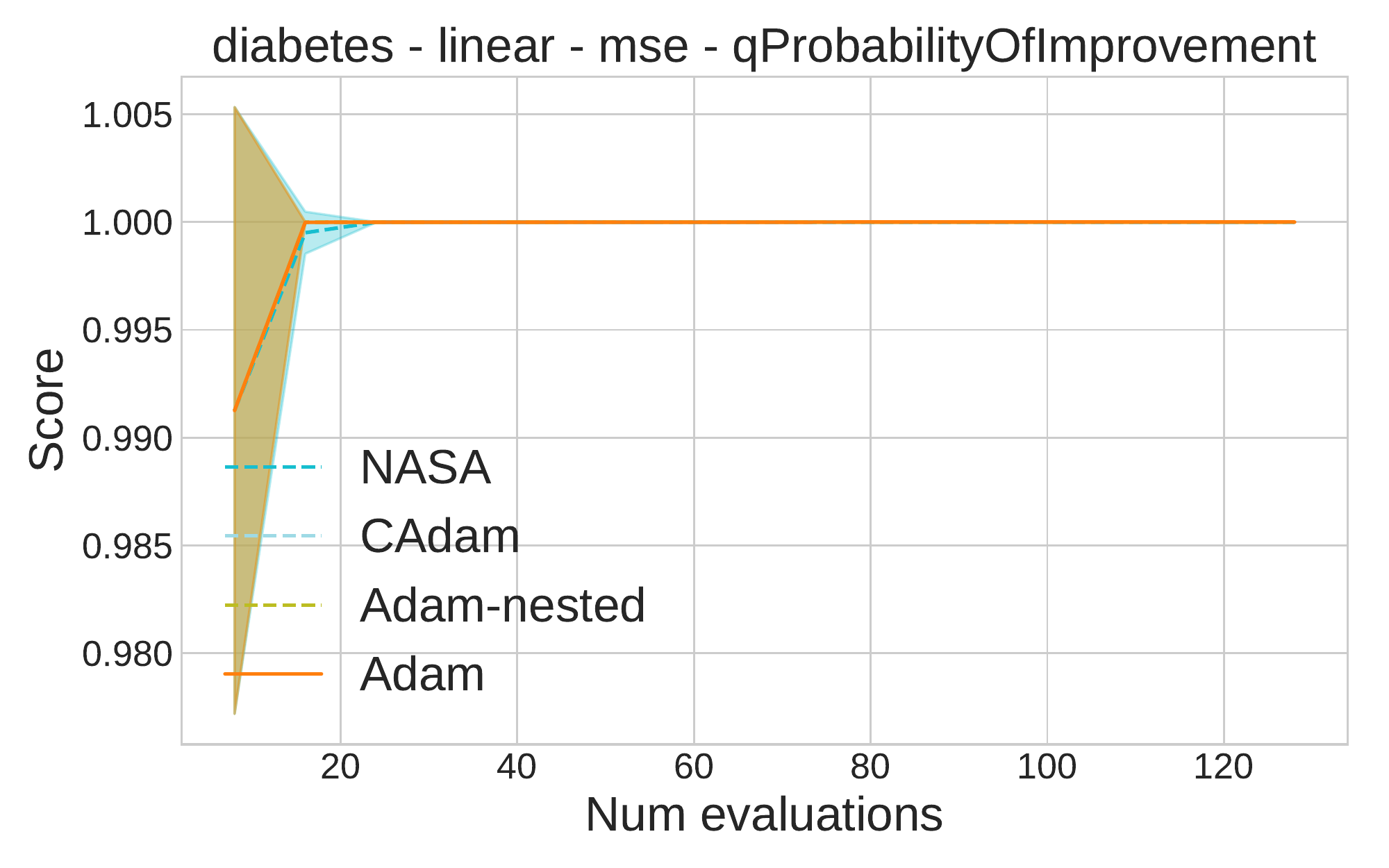}} & \hspace{-0.5cm}
    \subfloat{\includegraphics[width=0.19\columnwidth, trim={0 0.5cm 0 0.4cm}, clip]{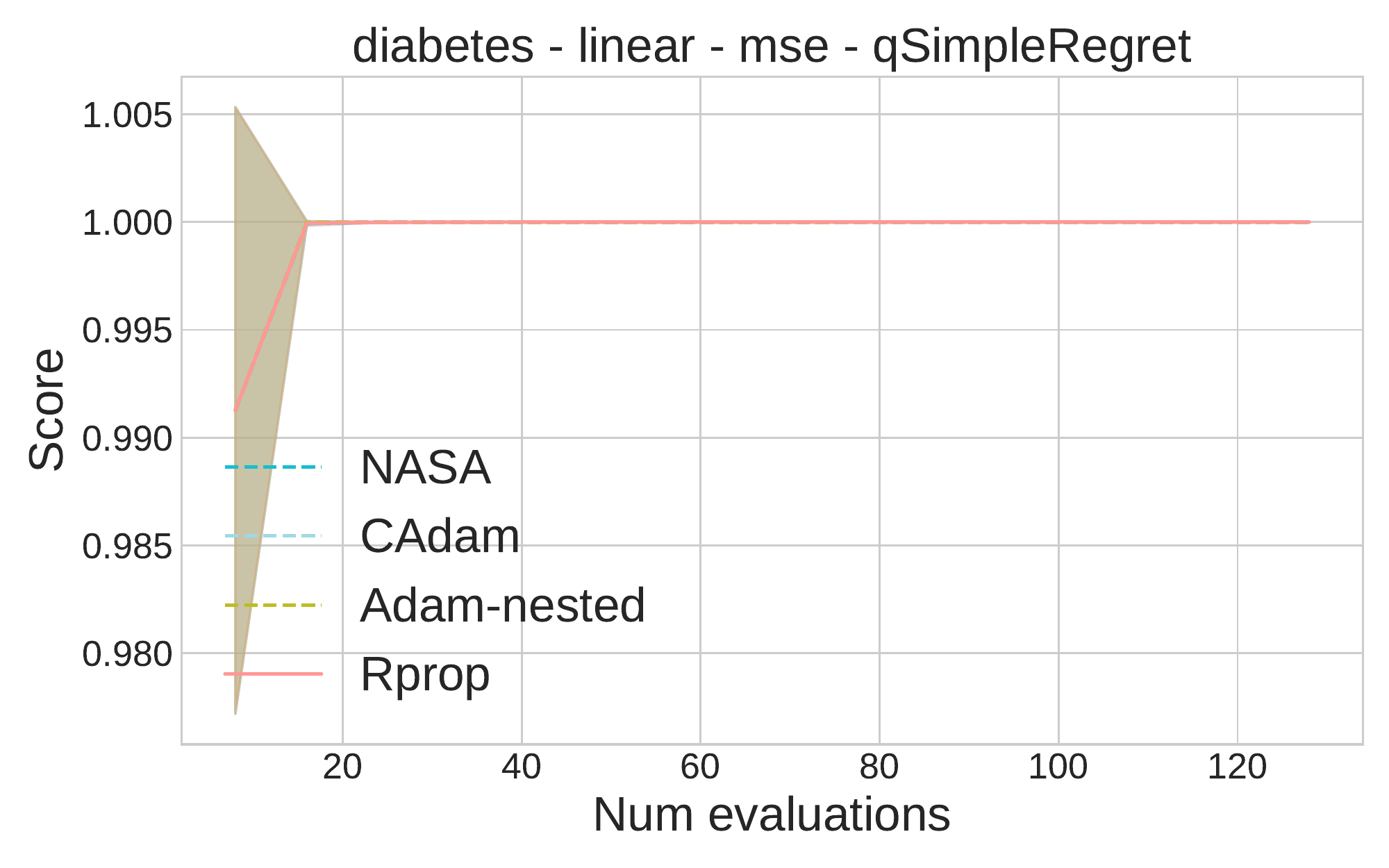}} 
    \end{tabular}
    \caption{}
    \end{figure}
    }
    
    {\renewcommand{\arraystretch}{0}
    \begin{figure}
    \begin{tabular}{ccccc}
    \subfloat{\includegraphics[width=0.19\columnwidth, trim={0 0.5cm 0 0.4cm}, clip]{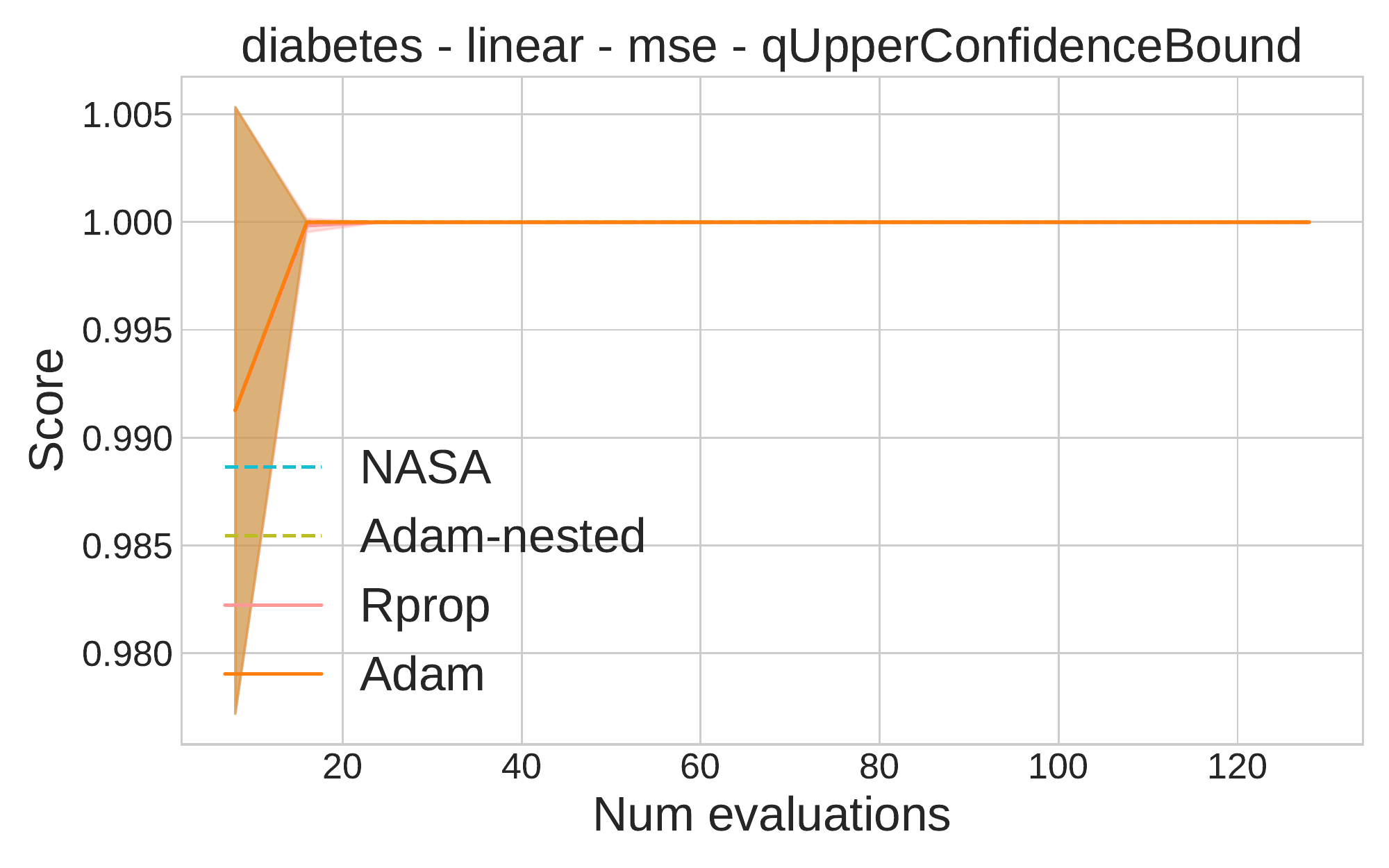}} &  \hspace{-0.5cm}
    \subfloat{\includegraphics[width=0.19\columnwidth, trim={0 0.5cm 0 0.4cm}, clip]{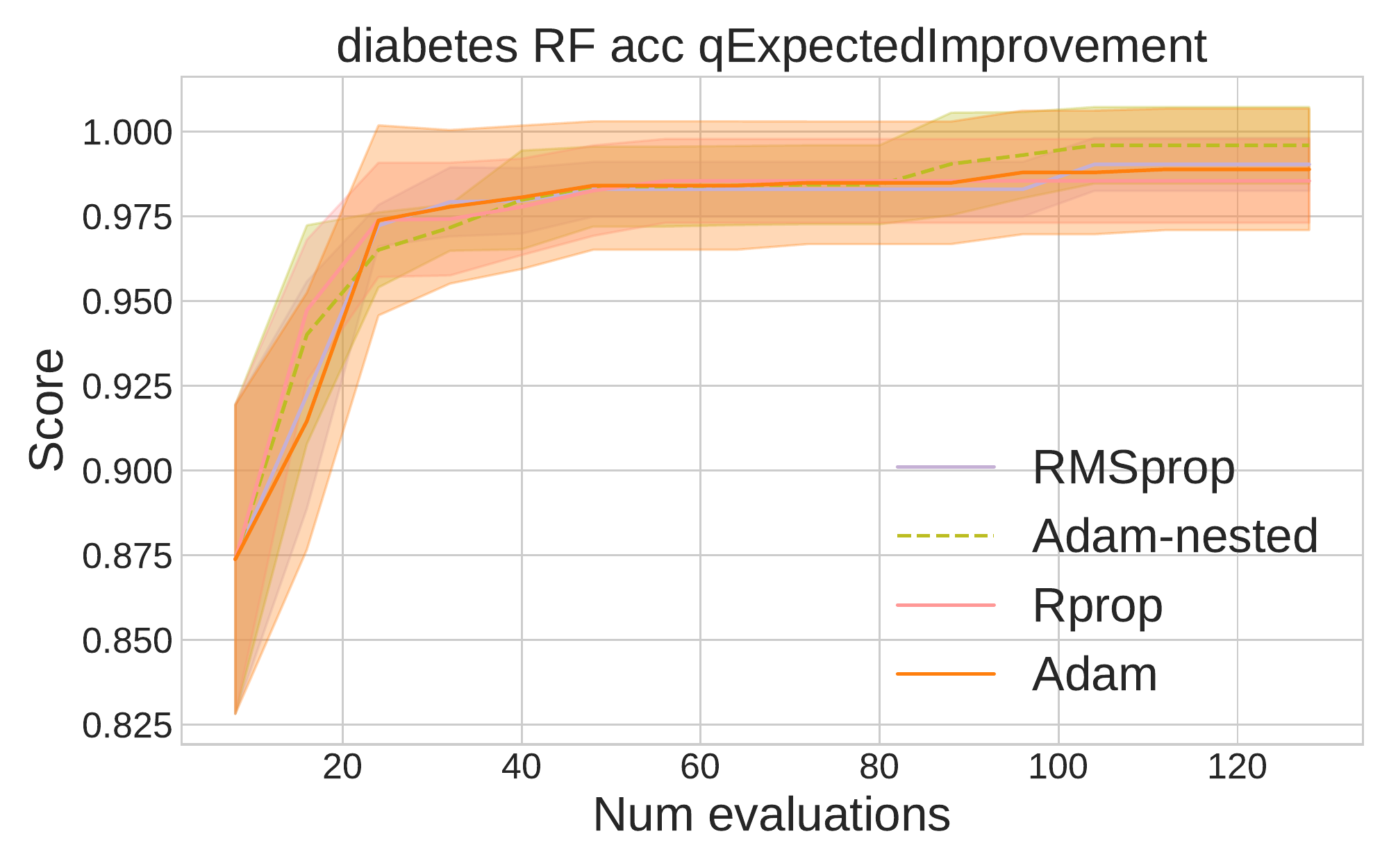}} & \hspace{-0.5cm}
    \subfloat{\includegraphics[width=0.19\columnwidth, trim={0 0.5cm 0 0.4cm}, clip]{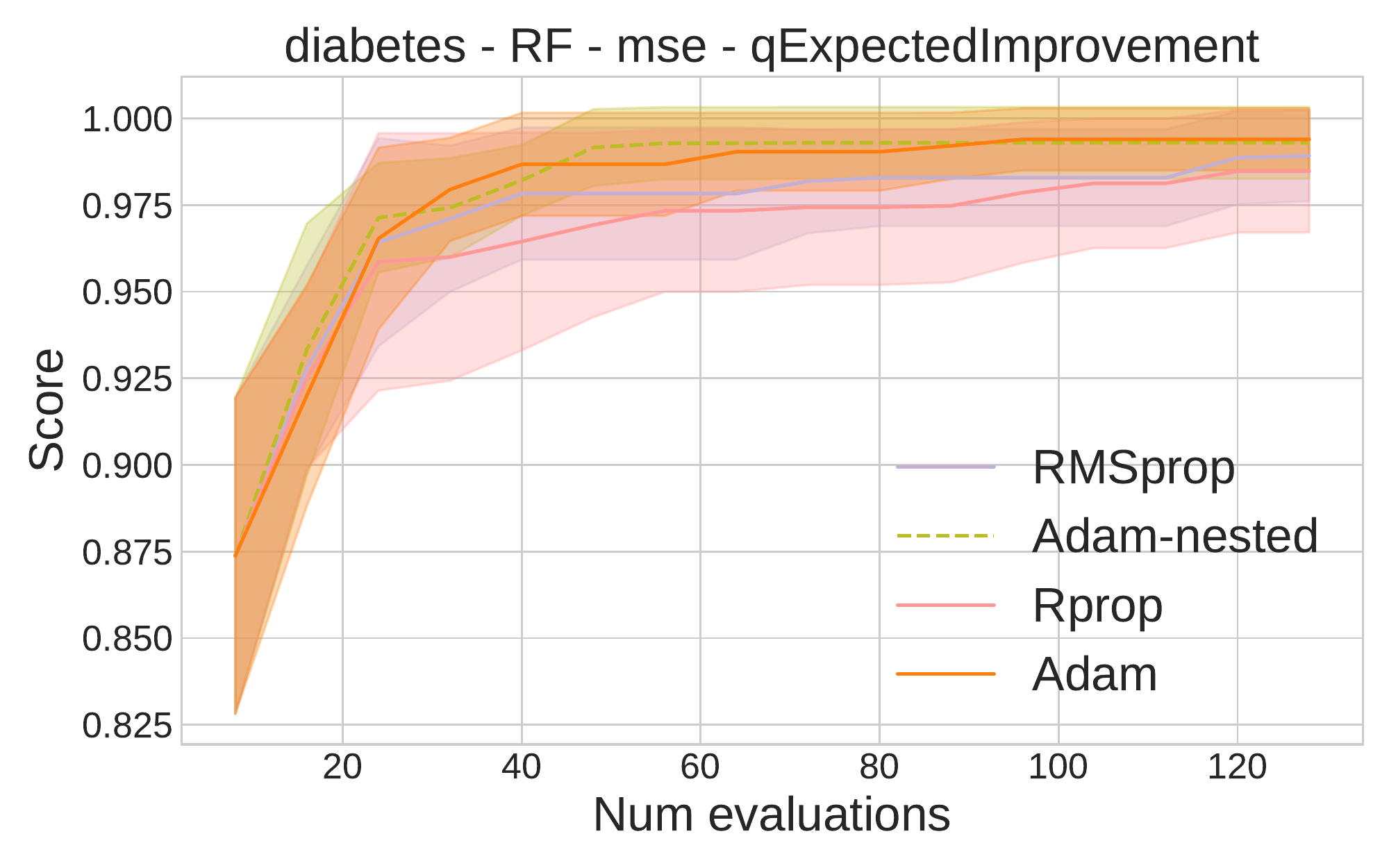}} &
    \subfloat{\includegraphics[width=0.19\columnwidth, trim={0 0.5cm 0 0.4cm}, clip]{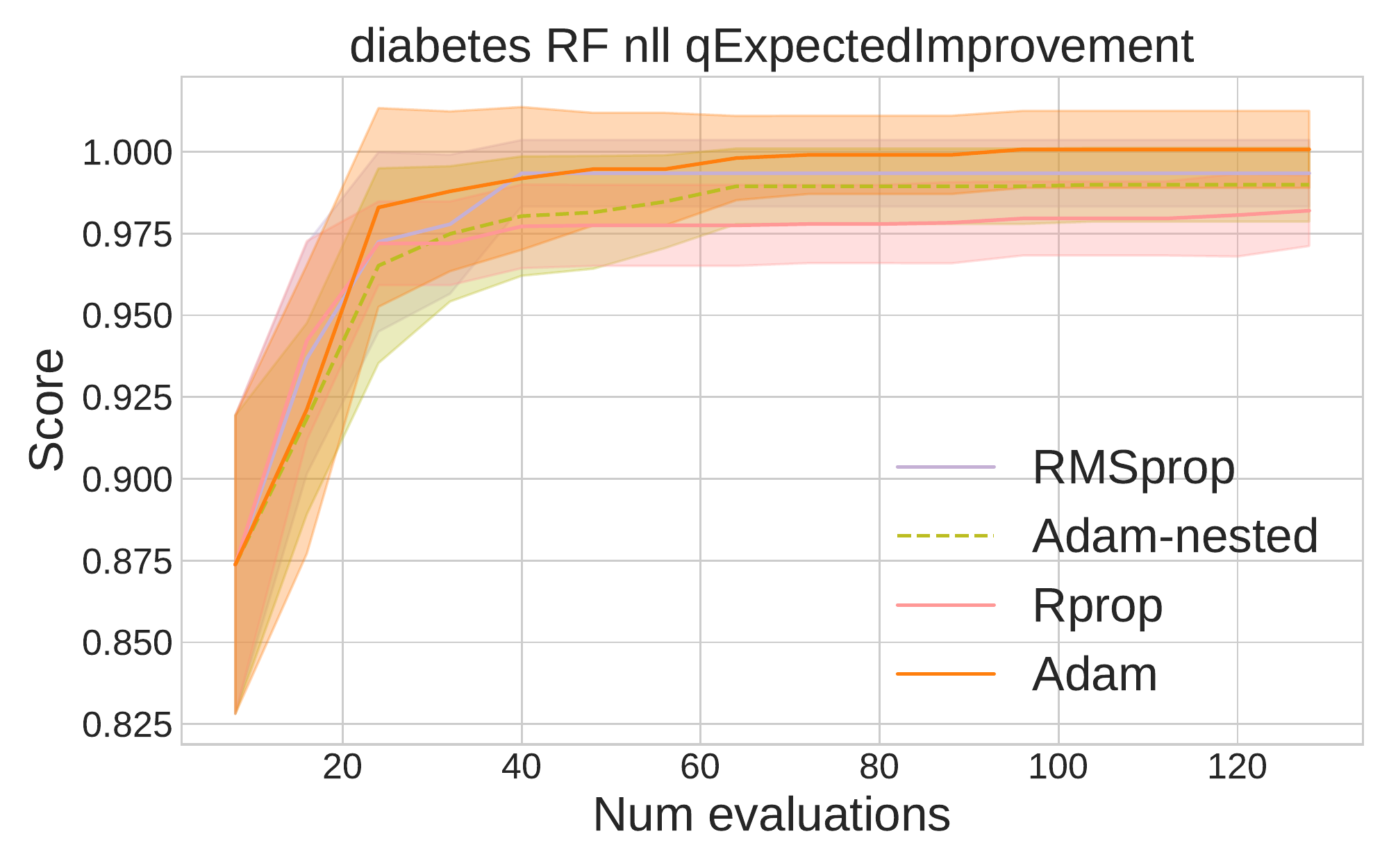}} &  \hspace{-0.5cm}
    \subfloat{\includegraphics[width=0.19\columnwidth, trim={0 0.5cm 0 0.4cm}, clip]{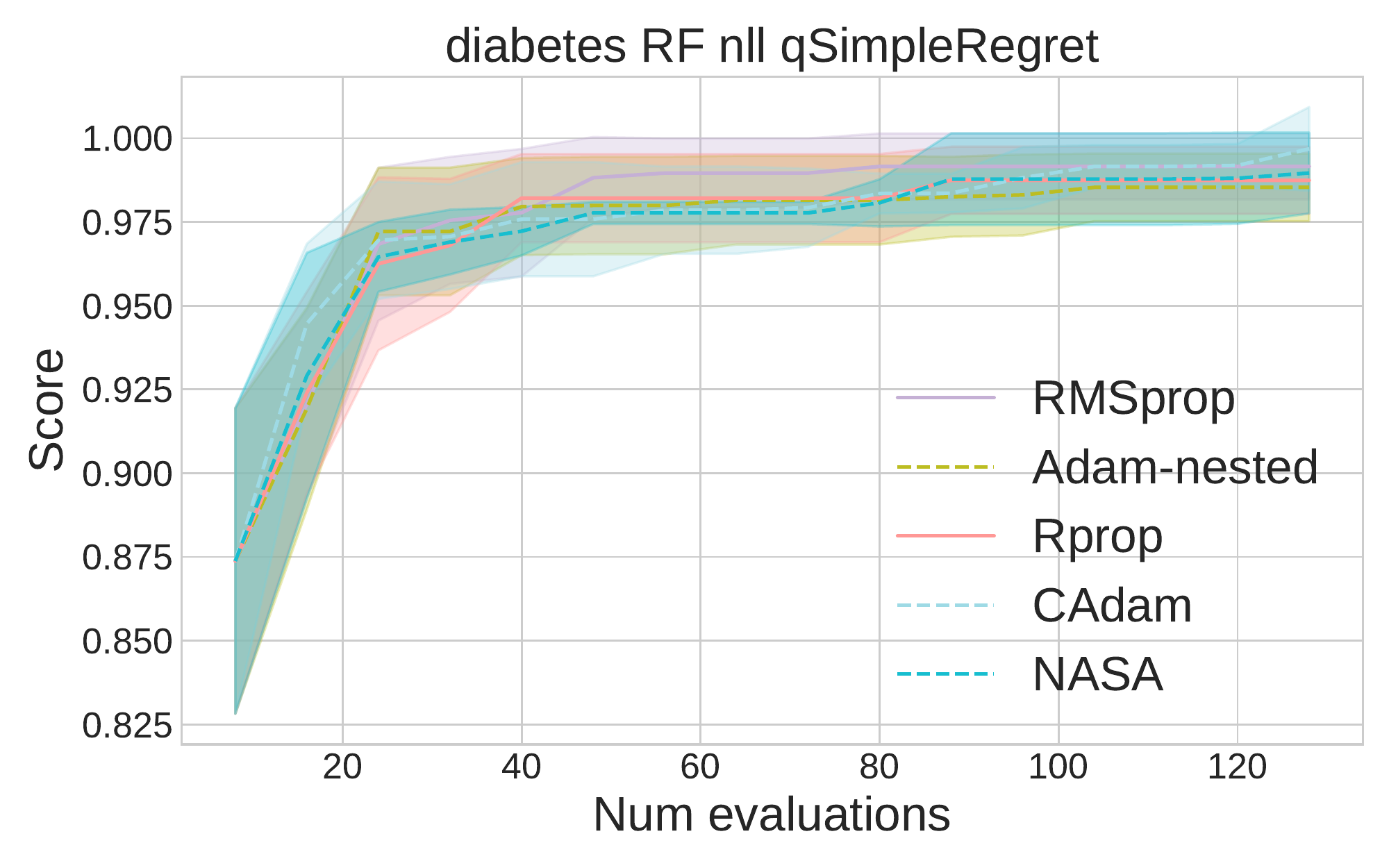}} \\  
    \subfloat{\includegraphics[width=0.19\columnwidth, trim={0 0.5cm 0 0.4cm}, clip]{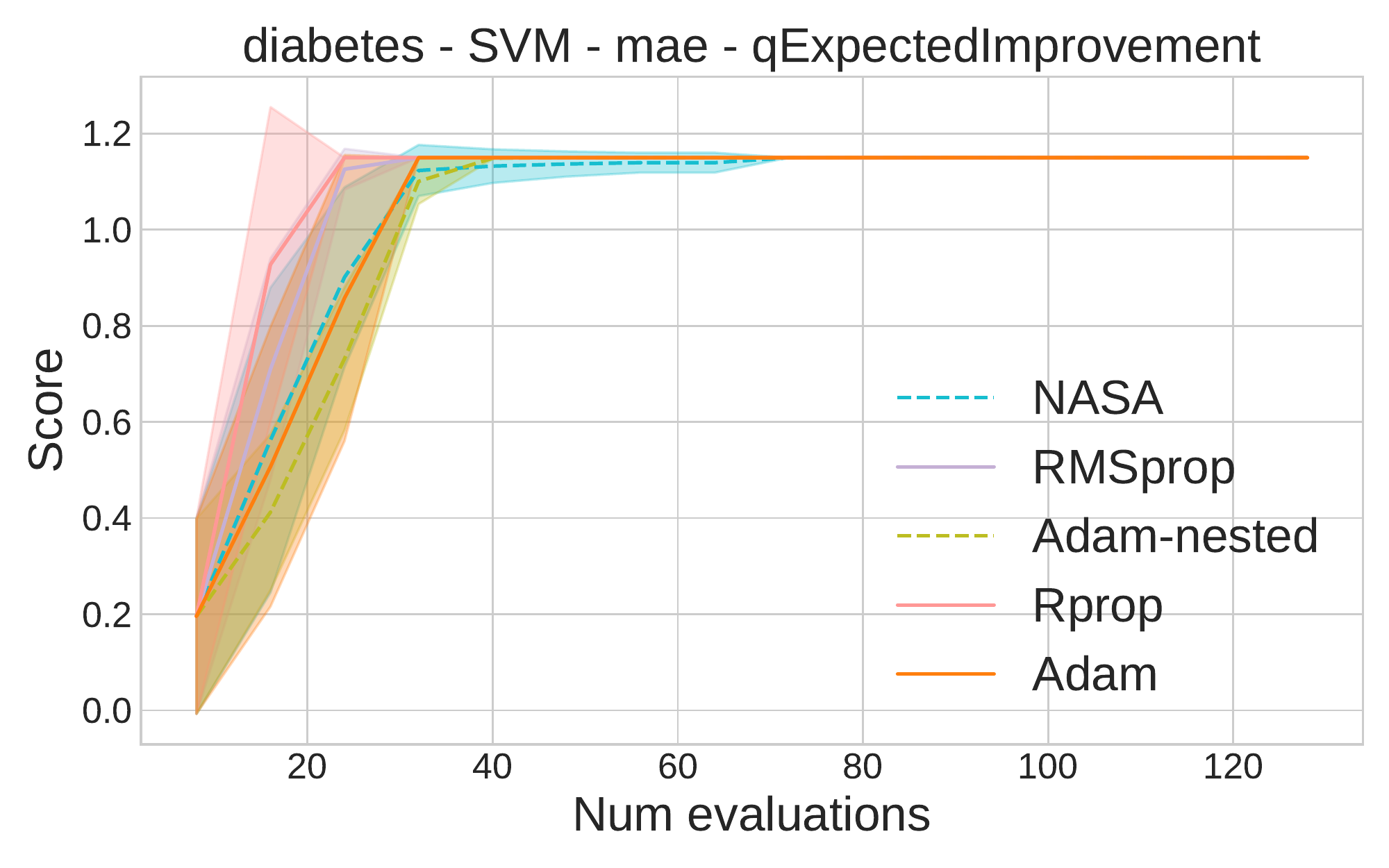}} &   \hspace{-0.5cm} 
    \subfloat{\includegraphics[width=0.19\columnwidth, trim={0 0.5cm 0 0.4cm}, clip]{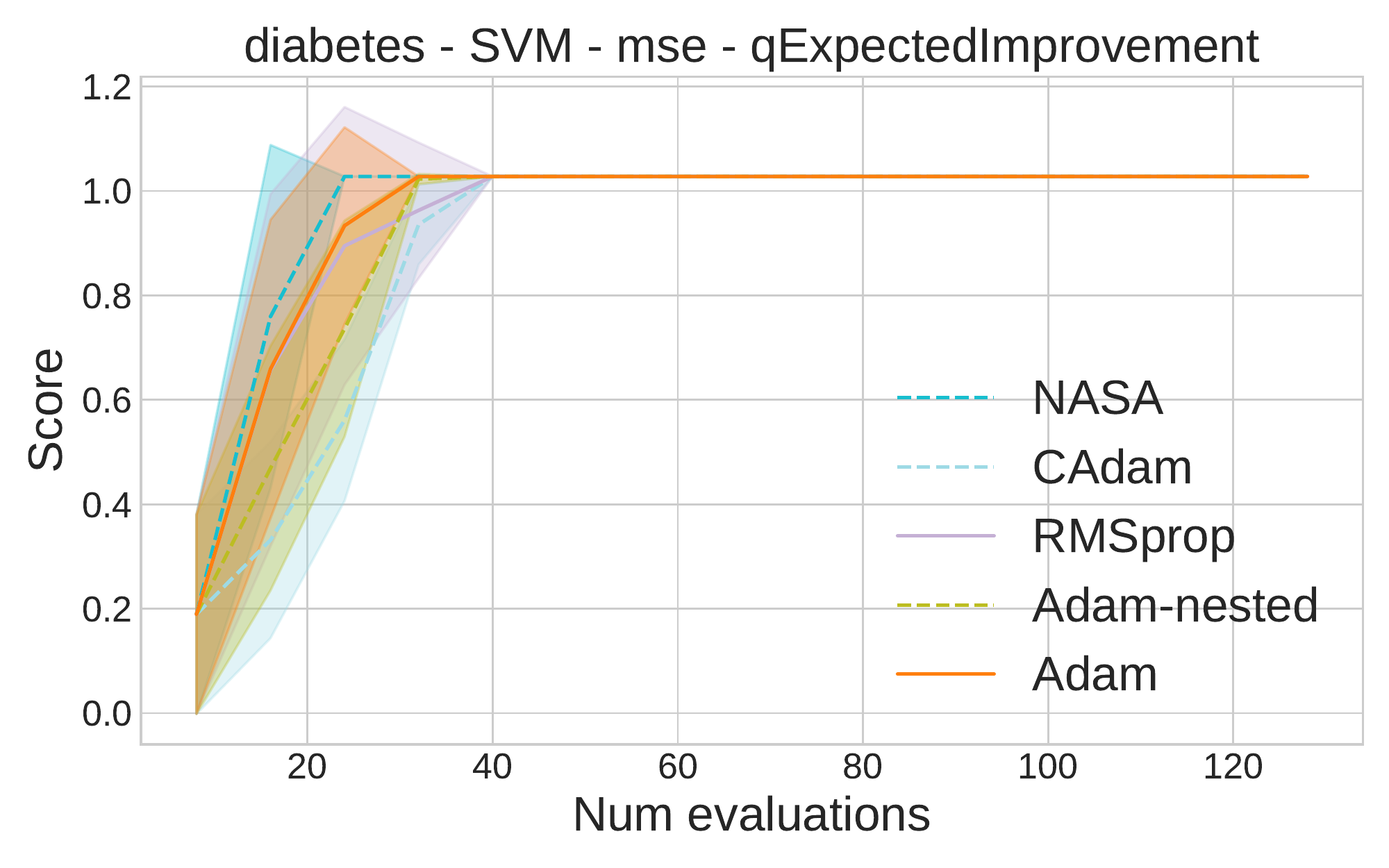}} & \hspace{-0.5cm}
    \end{tabular}
    \caption{}
    \end{figure}
    }

    \section{Hyperparameter Settings}\label{App:HyperParam}
    
    As explained in Section \ref{Sec:Exps}, the performances of first-order optimisers that are reported have been obtained after a hyperparameter tuning phase. We show in Table \ref{tab:non-compositional-tuning} and \ref{tab:compositional-tuning} the hyperparameters that have been tuned for each optimiser, along with their tuning domains. Tuning has been performed using the BO library \textsc{GPyOpt} with a total of $33$ configurations tested for each optimiser.
    
    \begin{table}[h]
        \centering
        \begin{tabularx}{\textwidth}{l c c X c}
        \toprule
        Optimiser & Reference & Parameters & Tuning domain & Scheduled \\ \hline
    \textbf{Adam} & App.~\ref{App:ADAM} & $\textnormal{lr}$  & $\mathcal{LU}(10^{-5}, 0.3)$ & \cmark\\ 
        \ & \ & $\beta_1$  & $\mathcal{U}(0.05, 0.999)$ & \ \\
        \ & \ & $\beta_2$ & $\mathcal{LU}(0.9, 1 - 10^{-6})$ & \ \\
        \ & \ & $w_d$ &  $\mathcal{LU}(10^{-8}, 0.1)$ & \ \\
      \textbf{AdamW} & App.~\ref{App:AdamW} & $\textnormal{lr}$  & $\mathcal{LU}(10^{-5}, 0.3)$ & \cmark\\ 
        \ & \ & $\beta_1$  & $\mathcal{U}(0.05, 0.999)$ & \ \\
        \ & \ & $\beta_2$  & $\mathcal{LU}(0.9, 1 - 10^{-6})$ & \ \\
        \ & \ & $w_d$  & $\mathcal{LU}(10^{-8}, 0.1)$ & \ \\
        
      \textbf{Adadelta} & App.~\ref{App:AdaDelta} & $\textnormal{lr}$  & $\mathcal{LU}(10^{-5}, 0.3)$ & \cmark\\ 
        \ & \ & $\rho$  & $\mathcal{LU}(0, 0.999)$ & \ \\
        \ & \ & $w_d$  & $\mathcal{LU}(10^{-8}, 0.1)$ & \ \\
        
        \textbf{Adagrad} & App.~\ref{App:AdaGrad} & $\textnormal{lr}$  & 
        $\mathcal{LU}(10^{-5}, 0.3)$ & \\ 
        \ & \ & $\textnormal{lr}_d$  & $\mathcal{LU}(10^{-7}, 10)$ & \ \\
        \ & \ & $\delta$  & $\mathcal{LU}(10^{-8}, .3)$ & \ \\
        \ & \ & $w_d$  & $\mathcal{LU}(10^{-8}, 0.1)$ & \ \\
    
        \textbf{SGA}& App.~\ref{App:SGA} & $\textnormal{lr}$  & $\mathcal{LU}(10^{-5}, 0.3)$ & \cmark\\ 
        \ & \ & $\rho$  & $\mathcal{U}(0, 1)$ & \ \\
        \ & \ & $\Delta$  & $\mathcal{U}(0,1)$ & \ \\
        \ & \ & $\textnormal{nesterov}$  & $\mathcal{U}\left\{0,1\right\}$ & \ \\
        \ & \ & $w_d$  & $\mathcal{LU}(10^{-8}, 0.1)$ & \ \\
        
        \textbf{Rprop} & App.~\ref{App:Rprop} & $\textnormal{lr}$  & $\mathcal{LU}(10^{-5}, 0.3)$ & \cmark\\ 
        \ & \ & $\eta_1$  & $\mathcal{U}(0, 1)$ & \ \\
        \ & \ & $\eta_2$  & $\mathcal{U}(1,3)$ & \ \\
        
        \textbf{RMSprop} & App.~\ref{App:RMSProp} & $\textnormal{lr}$  & $\mathcal{LU}(10^{-5}, 0.3)$ & \cmark\\ 
        \ & \ & $\rho$  & $\mathcal{U}(0, 1)$ & \ \\
        \ & \ & $\alpha$  & $\mathcal{LU}(10^{-6}, .3)$ & \ \\
        \ & \ & $\textnormal{centering}$  & $\mathcal{U}\left\{0,1\right\}$ & \ \\
        \ & \ & $w_d$  & $\mathcal{LU}(10^{-8}, 0.1)$ & \ \\
        
        \textbf{Adamos} & App.~\ref{App:AdamOS} & $\textnormal{lr}$  & $\mathcal{LU}(10^{-5}, 1.)$ & \\ 
        \ & \ & $\mu$  & $\mathcal{LU}(0.1, 0.999)$ & \ \\
        \ & \ & $C_\gamma$  & $\mathcal{U}(0.5, 1)$ & \ \\
        \ & \ & $\alpha_d$  & $\mathcal{U}(0.02,0.5)$ & \ \\
        \ & \ & $\mu_d$  & $\mathcal{U}(0.8, 1.2)$ & \ \\
        \ & \ & ${\gamma_2}_d$  & $\mathcal{U}(0.2, 0.8)$ & \ \\
    \bottomrule
    \end{tabularx}
        \caption{Selected first-order non-compositional optimisers used in our experiments together with their tuning domains. Searches in a log-uniform domain are denoted by $\mathcal{LU}$, while searches in a uniform continuous (resp. discrete) domain is denoted by $\mathcal{U}()$ (resp. $\mathcal{U}\{\}$). The learning rate scheduling, marked with the $\checkmark$ sign in the last column, is an exponential decay schedule with a multiplicative factor $\gamma$ tuned in $\mathcal{LU}(10^{-7}, 0.3)$ along with the other optimiser hyperparameters. }
        \label{tab:non-compositional-tuning}
    \end{table}

    \begin{table}
        \centering
        \begin{tabularx}{\textwidth}{X c c X c}
        \toprule
        Optimiser & Reference & Parameters & Tuning domain & Scheduled \\ \hline
        \textbf{CAdam} & \citep{tutunov2020cadam} & $\textnormal{lr}$  & $\mathcal{LU}(10^{-5}, 1.)$ & \\ 
        \ & \ & $\beta$  & $\mathcal{LU}(0.001, 0.999)$ & \ \\
        \ & \ & $\mu$  & $\mathcal{LU}(0.1, 0.999)$ & \ \\
        \ & \ & $C_\gamma$  & $\mathcal{U}(0.5, 1)$ & \ \\
        \ & \ & $\alpha_d$  & $\mathcal{U}(0.02,0.5)$ & \ \\
        \ & \ & $\mu_d$  & $\mathcal{U}(0.8, 1.2)$ & \ \\
        \ & \ & ${\gamma_2}_d$  & $\mathcal{U}(0.2, 0.8)$ & \ \\
        
        \textbf{NASA} & \citep{ghadimi2020single} & $\textnormal{a}$  & $\mathcal{U}(0.1, 10)$ & \\ 
        \ & \ & $b$  & $\mathcal{U}(0.1, 10)$ & \ \\
        \ & \ & $\beta$  & $\mathcal{U}(0.1, 10)$ & \ \\
        \ & \ & $\gamma$  & $\mathcal{U}(0.5, 1.)$ & \ \\
        
        \textbf{SCGD} & \citep{2017_Wang} & $\textnormal{\textnormal{lr}}$  & $\mathcal{LU}(-4, 1)$ & \\ 
        \ & \ & $\textnormal{lr}_d$  & $\mathcal{U}(0.4, .95)$ & \ \\
        \ & \ & $\beta$  & $\mathcal{LU}(0.1, 0.999)$ & \ \\
        \ & \ & $\beta_d$  & $\mathcal{U}(0.2, 0.8)$ & \ \\
        
        \textbf{ASCGD} & \citep{2017_Wang} & $\textnormal{\textnormal{lr}}$  & $\mathcal{LU}(-4, 1)$ & \\ 
        \ & \ & $\textnormal{lr}_d$  & $\mathcal{U}(0.4, .95)$ & \ \\
        \ & \ & $\beta$  & $\mathcal{LU}(0.1, 0.999)$ & \ \\
        \ & \ & $\beta_d$  & $\mathcal{U}(0.25, 0.85)$ & \ \\
        \bottomrule
      \end{tabularx}
        \caption{Selected first-order non-compositional optimisers used in our experiments together with their tuning domains.}
        \label{tab:compositional-tuning}
    \end{table}
    \end{document}